\documentclass[10pt,journal,compsoc]{IEEEtran}

\ifCLASSOPTIONcompsoc
  \usepackage[nocompress]{cite}
\else
  \usepackage{cite}
\fi

\ifCLASSINFOpdf
   \usepackage[pdftex]{graphicx}
\else

\fi

\usepackage{url}
\usepackage{array}
\usepackage{color}
\usepackage{amsmath}
\usepackage{amssymb}
\usepackage{multirow}
\usepackage{threeparttable}

\newcommand{\tabincell}[2]{\begin{tabular}{@{}#1@{}}#2\end{tabular}}

\newcommand{\mhy}[1]{\textcolor[rgb]{0.0,0.0,0.0}{#1}}
\newcommand{\add}[1]{\textcolor[rgb]{0.0,0.0,0.0}{#1}}

\begin{document}

\title{Large-Field Contextual Feature Learning for \\Glass Detection}

\author{Haiyang Mei, \ Xin Yang, \ Letian Yu, \ Qiang Zhang, \ Xiaopeng Wei, \ and \ Rynson W.H. Lau
\IEEEcompsocitemizethanks{\IEEEcompsocthanksitem Haiyang Mei, Xin Yang, Letian Yu, Qiang Zhang, and Xiaopeng Wei are with the Faculty of Electronic Information and Electrical Engineering, Dalian University of Technology, Dalian, 116024, China. E-mail: \protect\url{mhy666@mail.dlut.edu.cn}; \protect\url{xinyang@dlut.edu.cn}; \protect\url{letianyu@mail.dlut.edu.cn}; \protect\url{zhangq@dlut.edu.cn}; \protect\url{xpwei@dlut.edu.cn}.
\IEEEcompsocthanksitem Rynson W.H. Lau is with the Department of Computer Science, City University of Hong Kong. E-mail: \protect\url{Rynson.Lau@cityu.edu.hk}.
\IEEEcompsocthanksitem \mhy{Project homepage: \protect\url{https://mhaiyang.github.io/}.}
}
\thanks{Corresponding authors: Xin Yang, Xiaopeng Wei, and Rynson W.H. Lau.}
}

\markboth{IEEE Transactions on Pattern Analysis and Machine Intelligence}%
{Mei \MakeLowercase{\textit{et al.}}: Large-Field Contextual Feature Learning for Glass Detection}

\IEEEtitleabstractindextext{%
\begin{abstract}
Glass is very common in our daily life. Existing computer vision systems neglect it and thus may have severe consequences, \emph{e.g.}, a robot may crash into a glass wall. However, sensing the presence of glass is not straightforward. The key challenge is that arbitrary objects/scenes can appear behind the glass.
In this paper, we propose an important problem of detecting glass surfaces from a single RGB image. To address this problem, we construct the first large-scale glass detection dataset (GDD) and propose a novel glass detection network, called GDNet-B, which explores abundant contextual cues in a large field-of-view via a novel large-field contextual feature integration (LCFI) module and integrates both high-level and low-level boundary features with a boundary feature enhancement (BFE) module. Extensive experiments demonstrate that our GDNet-B achieves satisfying glass detection results on the images within and beyond the GDD testing set.
\mhy{We further validate the effectiveness and generalization capability of our proposed GDNet-B by applying it to other vision tasks, including mirror segmentation and salient object detection.}
\mhy{Finally, we show the potential applications of glass detection and discuss possible future research directions.}
\end{abstract}

}

\maketitle

\IEEEdisplaynontitleabstractindextext

\IEEEpeerreviewmaketitle

\IEEEraisesectionheading{\section{Introduction}\label{sec:introduction}}
\IEEEPARstart{G}{lass} is a non-crystalline, often transparent, amorphous solid that has widespread practical and decorative usages, \emph{e.g.}, window panes, glass doors, and glass walls. Such glass regions can have a critical impact to the existing vision systems (\emph{e.g.}, depth prediction and instance segmentation) as demonstrated in Figure \ref{fig:teaser}. They may further affect intelligent decisions in many applications, such as robotic navigation and drone tracking (\emph{i.e.}, the robot/drone might crash into the glass). Hence, it is essential for vision systems to be able to detect and segment glass from input images.

\def\wteazer{0.24\linewidth}
\def\hteazerone{.65in}
\def\hteazertwo{.52in}
\def\hteazerthree{.7in}
\begin{figure}
	\setlength{\tabcolsep}{1.6pt}
	\centering
	\begin{tabular}{cccc}
		& \includegraphics[width=\wteazer, height=2.5mm]{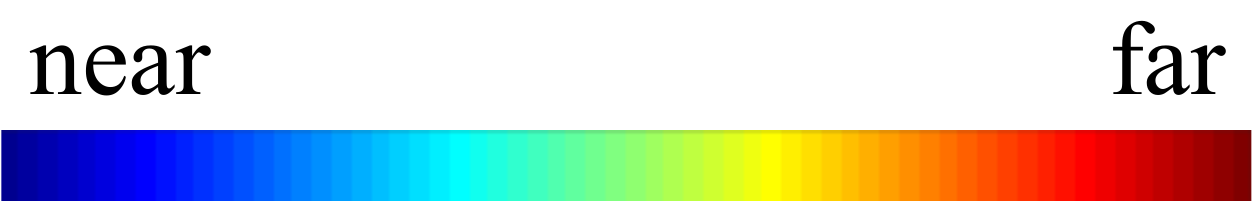} & & \includegraphics[width=\wteazer, height=2.5mm]{figure/teaser/bar.pdf} \\
		
		\includegraphics[width=\wteazer, height=\hteazerone]{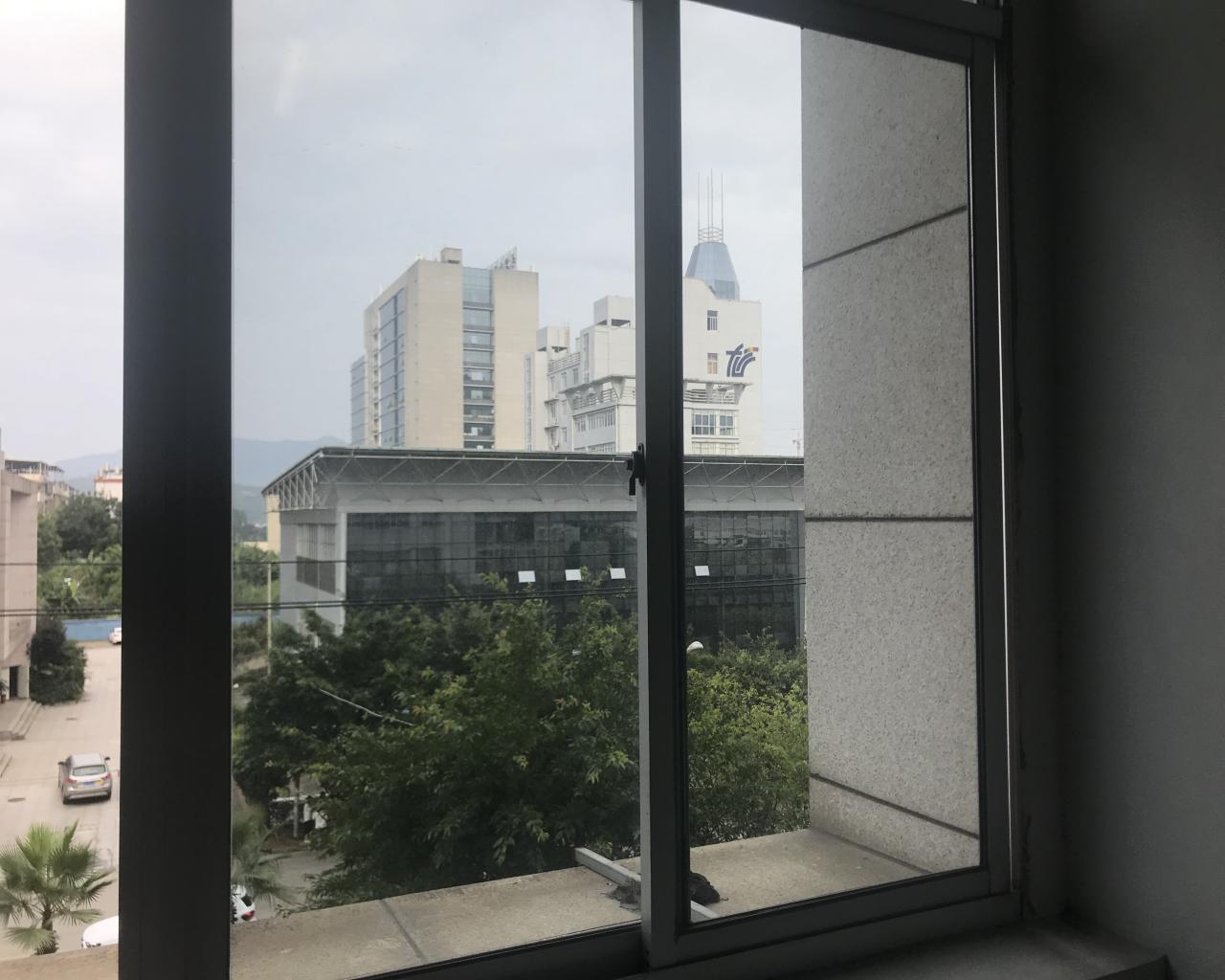}&
		\includegraphics[width=\wteazer, height=\hteazerone]{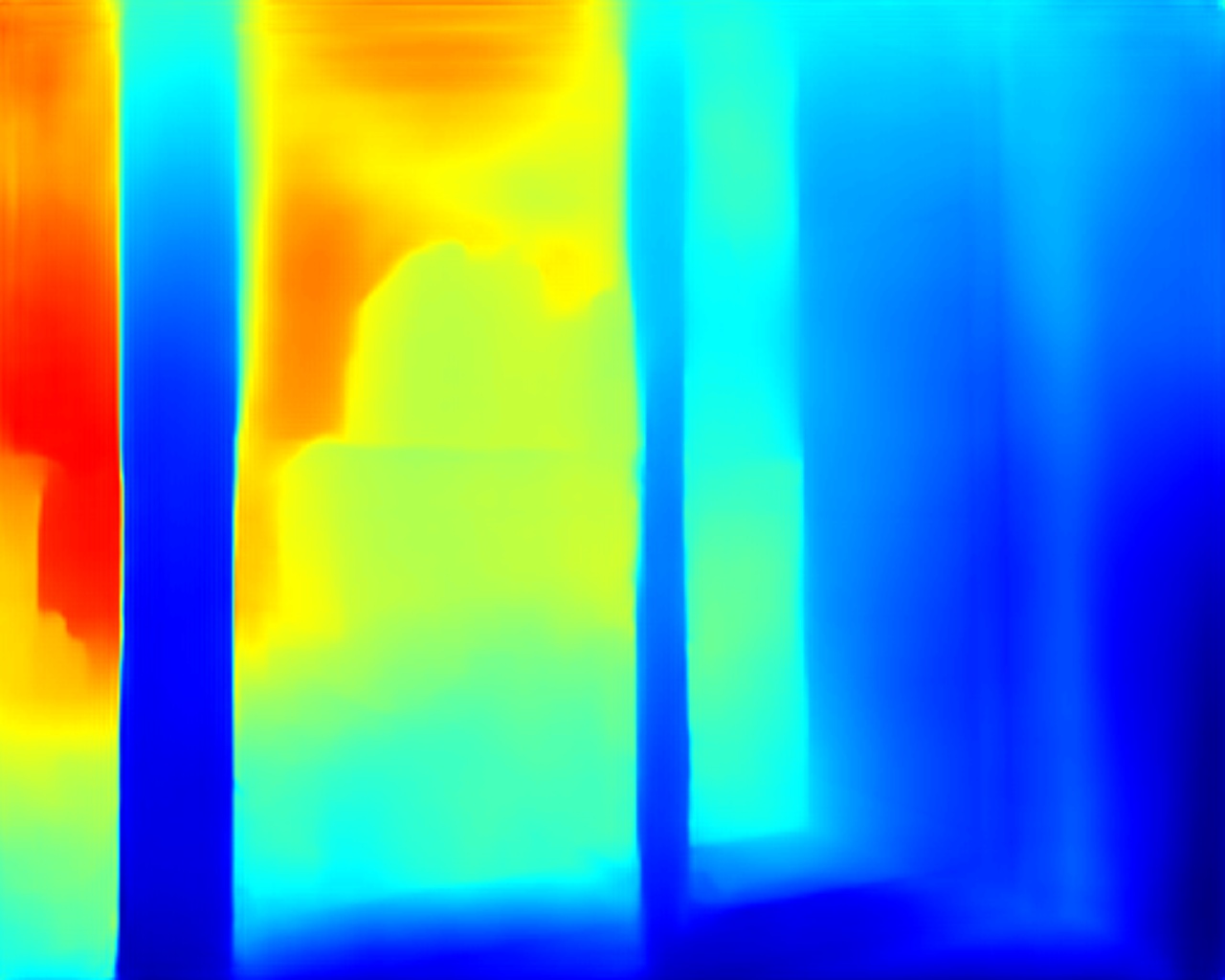}&
		\includegraphics[width=\wteazer, height=\hteazerone]{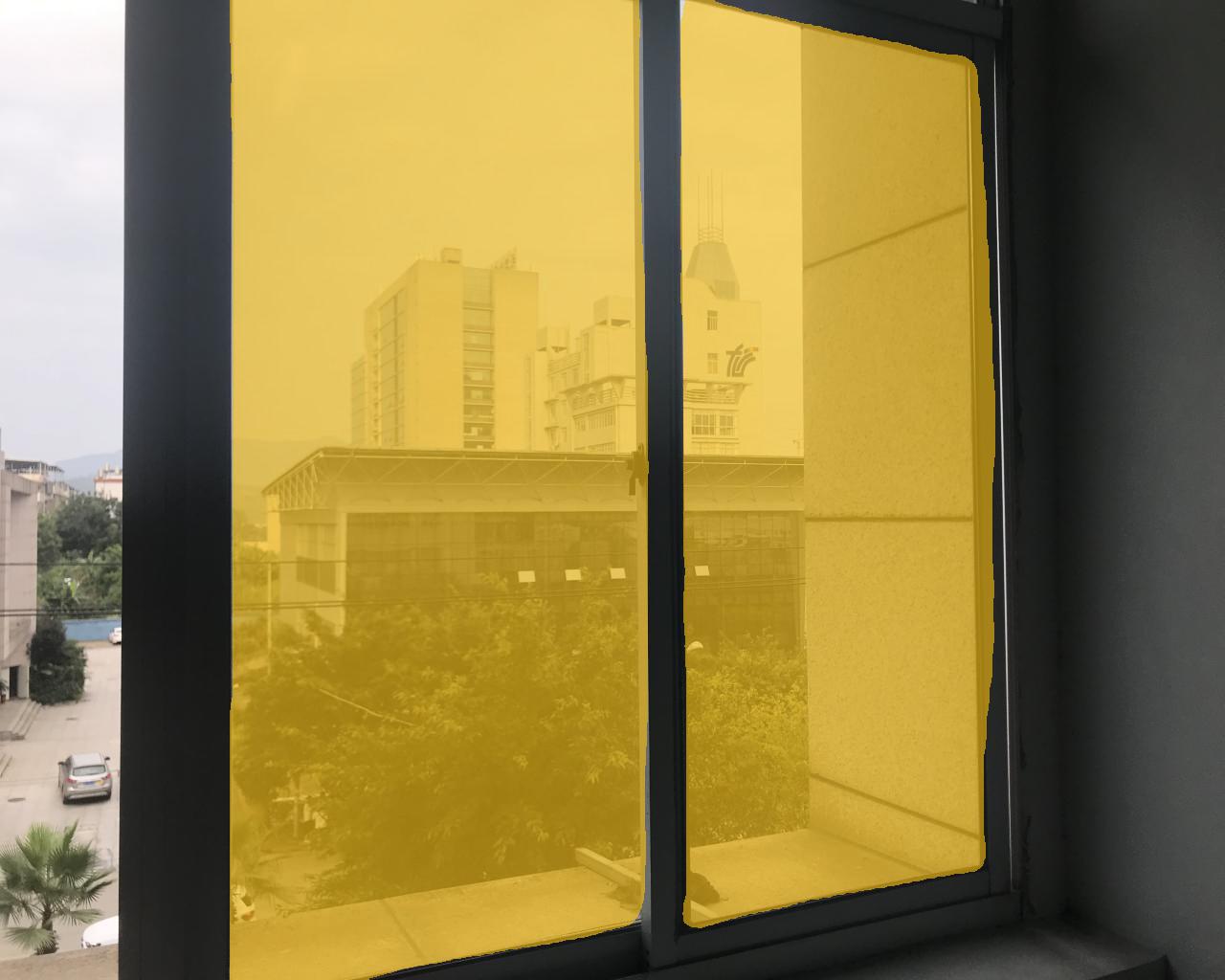}&
		\includegraphics[width=\wteazer, height=\hteazerone]{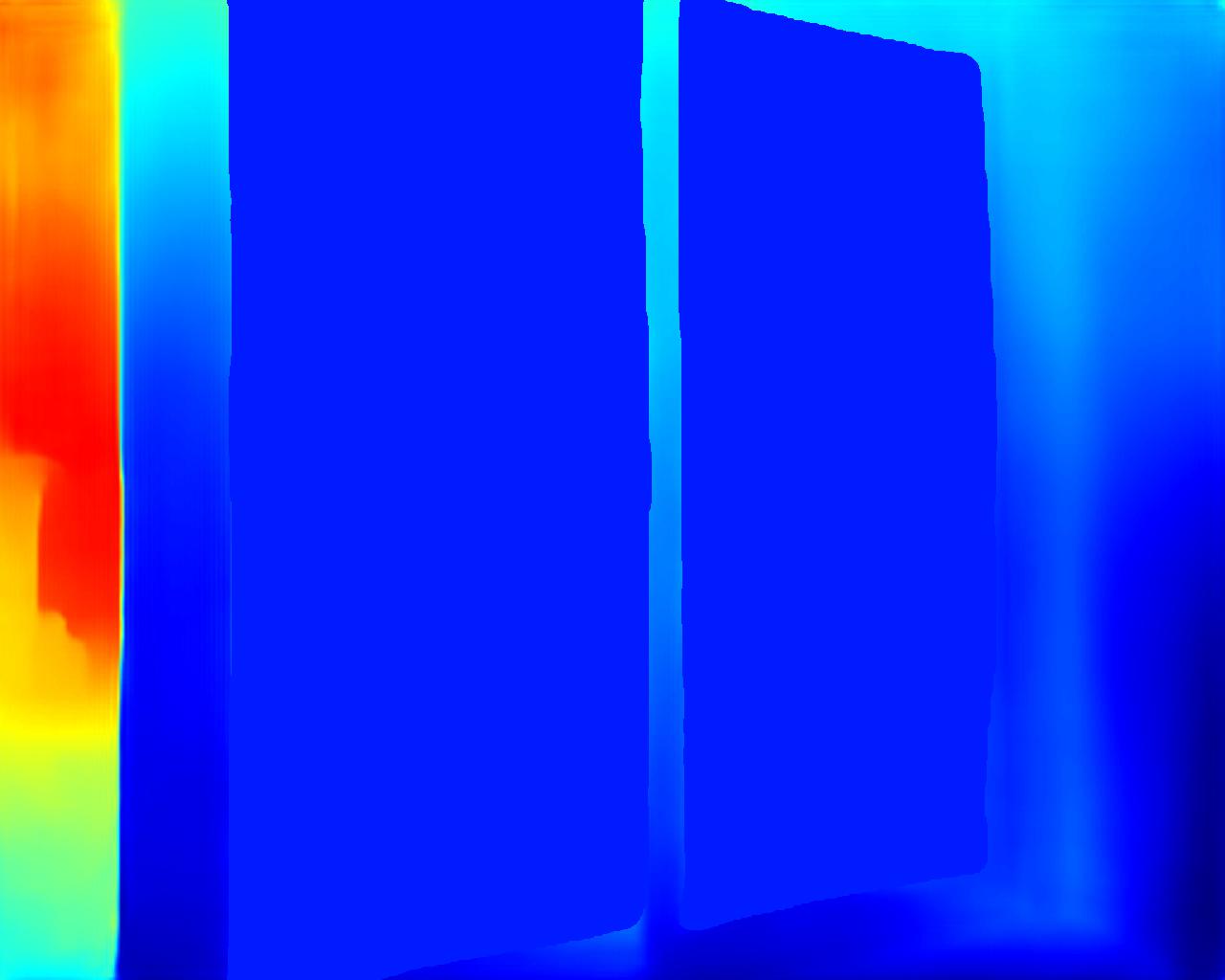}\\
		
		\includegraphics[width=\wteazer, height=\hteazertwo]{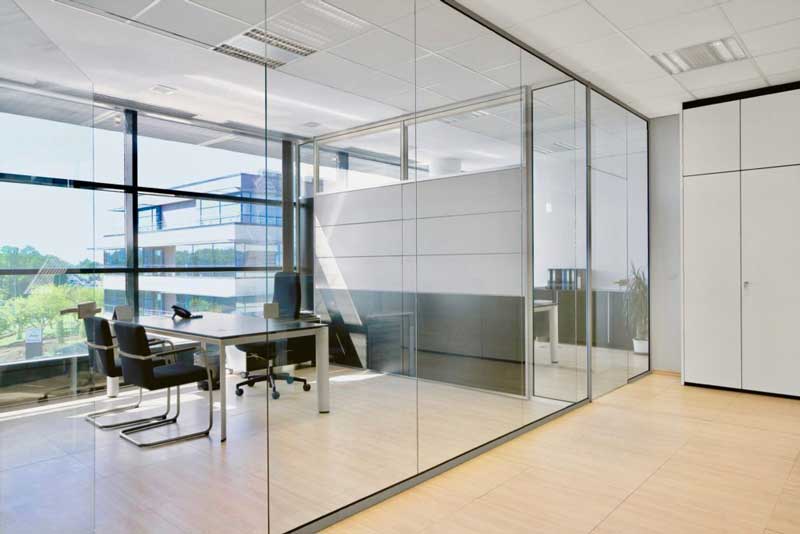}&
		\includegraphics[width=\wteazer, height=\hteazertwo]{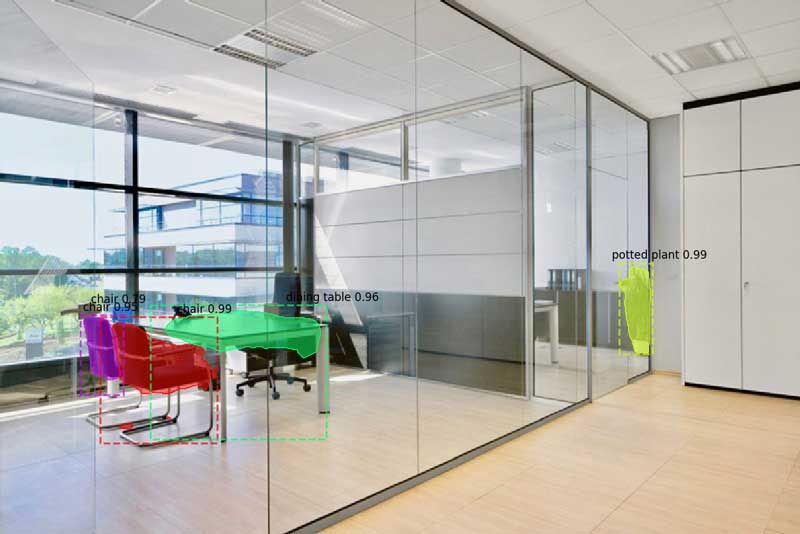}&
		\includegraphics[width=\wteazer, height=\hteazertwo]{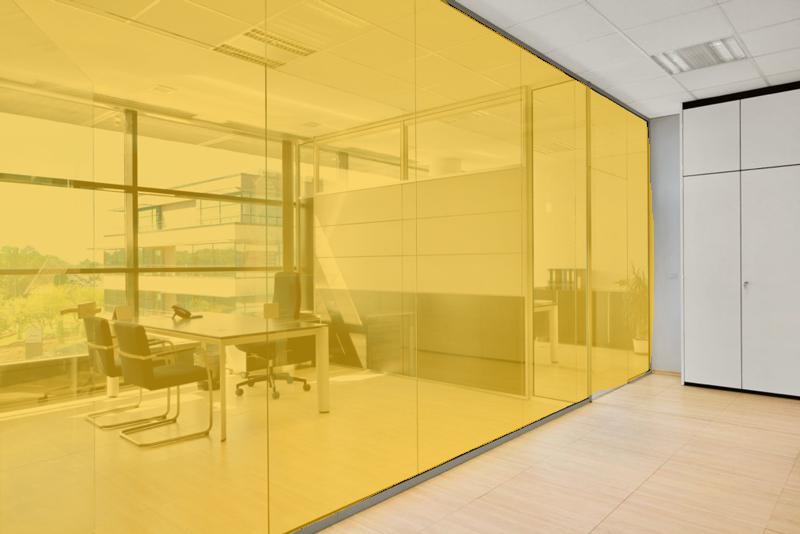}&
		\includegraphics[width=\wteazer, height=\hteazertwo]{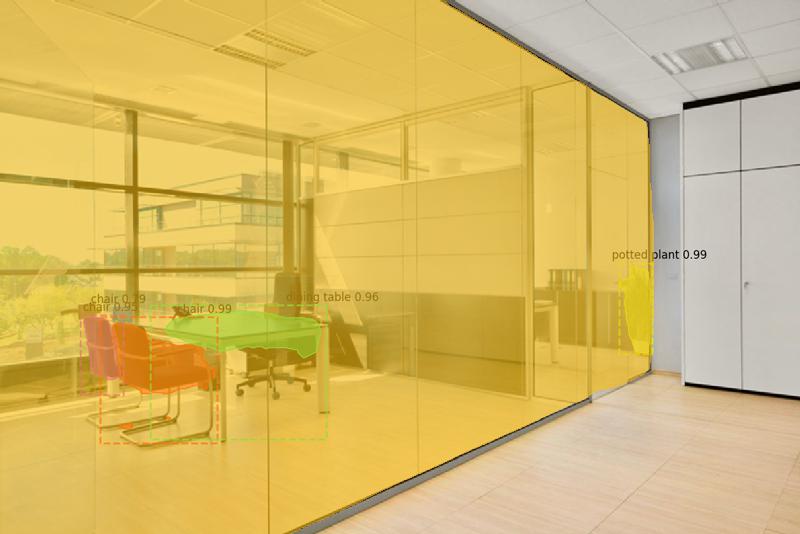}\\
		
		\includegraphics[width=\wteazer, height=\hteazerthree]{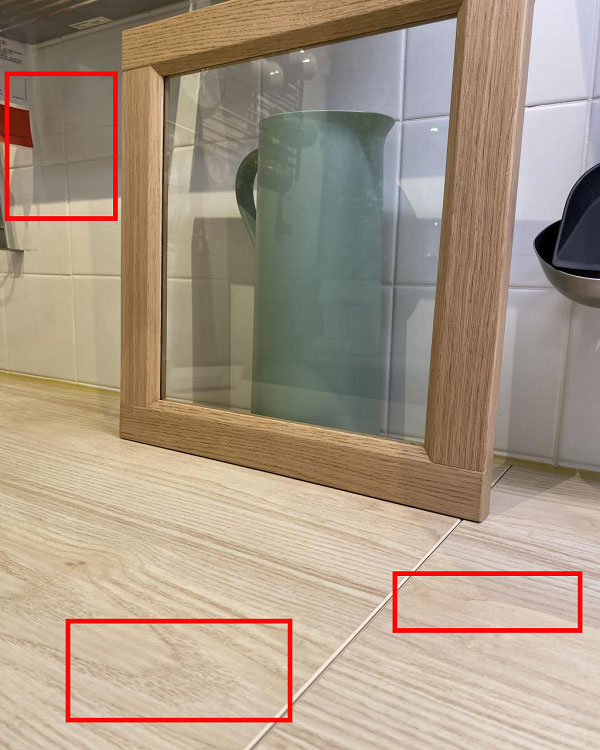}&
		\includegraphics[width=\wteazer, height=\hteazerthree]{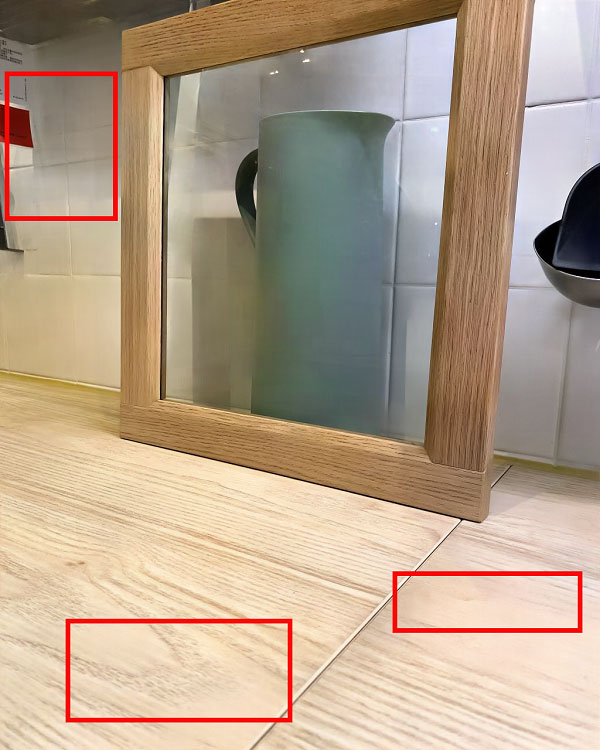}&
		\includegraphics[width=\wteazer, height=\hteazerthree]{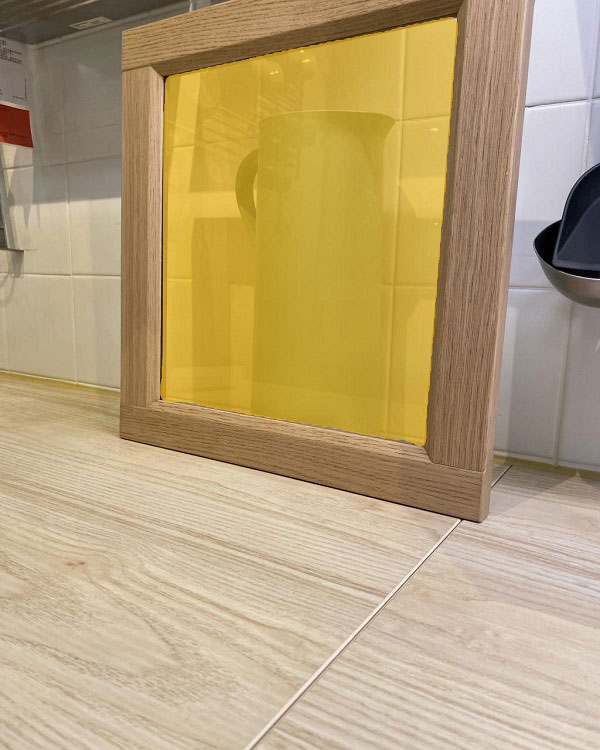}&
		\includegraphics[width=\wteazer, height=\hteazerthree]{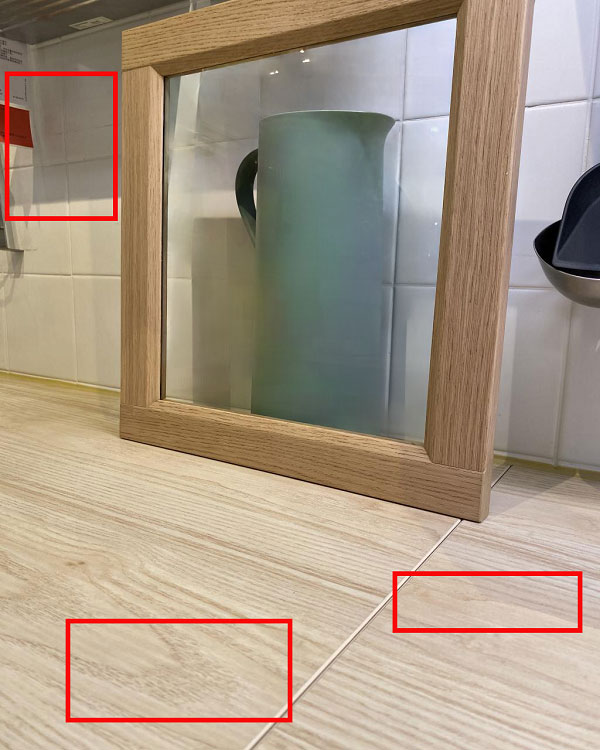}\\
		
		\scriptsize{(a) glass images} & \scriptsize{(b) w/o correction} & \scriptsize{(c) our results} & \scriptsize{(d) w/ correction} \\
		
	\end{tabular}
	\caption{Problems with glass in existing vision tasks. In depth prediction, existing method~\cite{Li_2018_CVPR_teaser} wrongly predicts the depth of the scene behind the glass, instead of the depth of the glass (1\emph{st} row of (b)). For instance segmentation, Mask RCNN~\cite{He_2017_ICCV} only segments the instances behind the glass, not aware that they are actually behind the glass (2\emph{nd} row of (b)). Besides, if we directly apply an existing singe-image reflection removal (SIRR) method~\cite{Yang_2018_ECCV_teaser} to an image that is only partially covered by glass, the non-glass region can be corrupted (3\emph{rd} row of (b)). Our method can detect the glass (c) and then help correct these failure cases (d).}
	\label{fig:teaser}
    \vspace{-4mm}
\end{figure}

Some small glass-made objects, such as cup and wine glass, can be detected well by \mhy{existing methods, \emph{e.g.}, \cite{He_2017_ICCV,Cai_2018_CVPR_cascadercnn,huang2019mask_msrcnn,Kalra_2020_CVPR}}, as these glass objects exhibit certain fixed patterns. However, automatically detecting glass regions from images like the ones shown in Figure \ref{fig:teaser}(a) is an extremely challenging task. Due to the fact that a glass region does not have a fixed pattern, \emph{i.e.}, arbitrary objects/scene can appear behind the glass. This makes the glass fundamentally different from other common objects that have been well-addressed by the state-of-the-art segmentation methods~\cite{He_2017_ICCV}. Meanwhile, directly applying existing salient object detection methods~\cite{liu2018picanet,qin2019basnet} to detect glass is inappropriate, as not all glass regions are salient.
\mhy{Besides, the recent mirror segmentation method \cite{yang2019mirrornet} may segment mirrors by detecting the content contrast between inside and outside the mirror. However, as the content presented by the transparent glass is mainly inherited from the background rather than totally reflecting the scene in front of it like a mirror, the appearance of the glass is typically similar with the image background, making the glass detection problem more difficult.}

To address the glass detection problem, a straightforward solution is to apply a reflection/boundary detector for glass detection. Unfortunately, this may fail if the glass has only weak/partial reflections or ambiguous boundary due in some complex scene, \emph{e.g.}, the second image in Figure \ref{fig:teaser}(a). In general, humans can identify the existence of glass well. We observe that humans typically would combine different contextual information to infer whether and where glass exists. These contexts not only include low-level cues (\emph{e.g.}, the color difference between inside and outside of the glass, blur/bright spot/ghost caused by reflection), but also high-level contexts (\emph{e.g.}, relations between different objects). This inspires us to leverage abundant contextual features for glass detection.

In this paper, we address the glass detection problem from two aspects. First, we construct a large-scale glass detection dataset (GDD), which consists of 3,916 high-quality images with glass and corresponding glass masks, covering various daily-life scenes. \mhy{Second, we propose a novel glass detection network (GDNet-B), in which multiple well-designed large-field contextual feature integration (LCFI) modules and two boundary feature enhancement (BFE) modules are embedded to harvest abundant low-level as well as high-level contexts from a large receptive field and to integrate both high-level and low-level boundary information, respectively,} for accurately detecting glass of different sizes in various scenes.

To sum up, our contributions are as follows:
\begin{itemize}
	\item We contribute the first large-scale glass detection dataset (GDD) with glass images in diverse scenes and corresponding manually labeled glass masks.
	\item \mhy{We propose a novel glass detection network (GDNet-B) with a well-designed large-field contextual feature integration (LCFI) module and a boundary feature enhancement (BFE) module for accurate glass detection, by exploring abundant contextual cues from a large receptive field and integrating boundary information.}
	\item We achieve superior glass detection results on the GDD testing set, by comparing with state-of-the-art models fine-tuned for glass detection.
\end{itemize}

\mhy{A preliminary version of this work was presented at CVPR 2020 \cite{Mei_2020_CVPR}. This submission extends the work in the following four aspects:}
\begin{itemize}
	\item \mhy{We add a summary of existing representative works on discriminative context learning (Section \ref{sec:relatedwork}).}
	\item \mhy{We introduce the boundary cue into the glass detection process via embedding a well-designed boundary feature enhancement (BFE) module into the original GDNet \cite{Mei_2020_CVPR}, making our boundary-aware model (termed as GDNet-B) achieve a new state-of-the-art glass detection performance.}
	\item \mhy{We demonstrate the effectiveness and generalization capability of our proposed GDNet-B by applying it to mirror segmentation (Section \ref{sec:mirror}) and salient object detection (Section \ref{sec:saliency}) tasks.}
	\item \mhy{We analyze the limitations of our method, show the potential applications of glass detection, and discuss possible future research directions (Section \ref{sec:discussion}).}
\end{itemize}



\section{Related Work}\label{sec:relatedwork}
In this section, we first briefly review state-of-the-art methods from relevant fields, including semantic/scene/instance segmentation, salient object detection, specific region detection/segmentation, and single image reflection removal. Then, we discuss some representative works on discriminative context learning.

\textbf{Semantic/scene/instance segmentation.}
Semantic segmentation aims to segment and parse a given image into different regions associated with semantic categories of discrete objects. Scene segmentation further considers stuff when assigning a label for each pixel.
Recently, great progress has been achieved benefited by the advances of deep neural networks.  Based on fully convolutional networks (FCNs) \cite{Long_2015_CVPR}, state-of-the-art models typically leverage multi-scale context aggregation or
exploit more discriminative contexts to achieve high segmentation performance.
For example, Chen \emph{et al.} \cite{chen2017deeplab} introduce an atrous spatial pyramid pooling (ASPP) to capture multi-scale context information. Zhao \emph{et al.} \cite{zhao2017pyramid} employ a pyramid pooling module to aggregate local and global context.
Ding \emph{et al.} \cite{Ding_2018_CVPR} explore contextual contrasted features to boost the segmentation performance of small objects. Zhang \emph{et al.} \cite{Zhang_2018_CVPR_enc} introduce a channel attention mechanism to capture global context. Fu \emph{et al.} \cite{Fu_2019_CVPR} leverage channel- and spatial-wise non-local attention modules to capture contextual features with long-range dependencies. Huang \emph{et al.} \cite{Huang_2019_ICCV} further propose a criss-cross attention module to efficiently capture information from long-range dependencies.

Instance segmentation aims to differentiate individual instances of the each category. A typical method is Mask-RCNN \cite{He_2017_ICCV}, which adds a branch of the object detection network Faster-RCNN \cite{ren2017faster} and achieves good results. PANet \cite{liu2018path} further adds bottom-up paths to aggregates multi-level features for detection and segmentation.

However, applying the above segmentation approaches for glass detection (\emph{i.e.}, treating glass as one of the object categories) may not be appropriate as arbitrary objects/scenes can appear behind glass, making  glass fundamentally different from other objects. In this paper, we focus on the glass detection problem and formulate it as a binary classification problem (\emph{i.e.}, glass or non-glass).

\textbf{Salient object detection (SOD).}
A lot of image-based SOD methods have been proposed in the past decades. Early methods are mainly based on low-level hand-crafted visual features as well as heuristic saliency priors (\emph{e.g.}, color \cite{achanta2009frequency} and contrast \cite{klein2011center,cheng2014global}). These features, however, have limited capability to help distinguish between salient and non-salient objects.
%
Recently, many state-of-the-art deep models are proposed. They utilize the integration of different levels of features to enhance network performances.
Specifically, Liu \emph{et al.} \cite{Liu_2016_CVPR} progressively integrate local context information to predict saliency maps. Zhang \emph{et al.} \cite{Zhang_2017_ICCV} propose a generic framework to integrate multi-level features at different resolutions. Zhang \emph{et al.} \cite{Zhang_2018_CVPR_progressive} introduce an attention guided network to selectively integrate multi-level information in a progressive manner. Zhang \emph{et al.} \cite{Zhang_2018_CVPR_bd} design a bi-directional message passing module with a gated function to integrate multi-level features. Wang \emph{et al.} \cite{Wang_2019_CVPR} integrate high-level and low-level features by performing both top-down and bottom-up saliency inferences in an iterative and cooperative manner.

In general, the content presented by a glass region is from a real scene, which may or may not contain salient objects. Existing SOD methods may not be able to detect the whole glass region well.

\textbf{Specific region detection/segmentation.}
We briefly review three binary classification tasks: shadow detection, water hazard detection, and mirror segmentation.

Shadow detection aims to detect shadows from a single input image. Hu \emph{et al.} \cite{Hu_2018_CVPR} address the shadow detection problem by analyzing image context in a direction-aware manner. Zhu \emph{et al.} \cite{Zhu_2018_ECCV} combine local and global contexts for shadow detection. Zheng \emph{et al.} \cite{zheng2019distraction} consider shadow distractions in the shadow detection task. In general, there is an intensity difference between shadow region and non-shadow region, while glass typically does not have such an obvious intensity difference between inside and outside of the glass region, making the glass detection problem more difficult to address.

Water hazard detection is to detect water in puddles and flooded areas, on and off the road, to reduce the risk to autonomous cars. The reflection on the water surface typically is an inverted and disturbed transform of the sky or nearby objects above the water surface. Han \emph{et al.} \cite{Han_2018_ECCV} present a reflection attention unit to match this pattern in the vertical direction. However, reflections on the glass can be generated from arbitrary directions and thus applying this method may not be suitable.

Mirror segmentation is a newly proposed research topic that aims to segment mirror regions from a single RGB image. Yang \emph{et al.} \cite{yang2019mirrornet} observe that there exists both high-level and low-level discontinuities between inside and outside of the mirror and leverage contextual contrasted features to segment mirrors.
\mhy{As the content presented in the transparent glass region is mainly inherited from the background rather than totally reflecting the scene in front of it like a mirror, the appearance of the glass is typically similar with the image background, leading to the little contrasts in terms of both semantics and low-level features between the glass region and its surroundings.}
Therefore, utilizing contextual contrasted features to detect glass may not obtain the desired results.

\textbf{Single image reflection removal (SIRR).} Reflection is a frequently-encountered source of image corruption when shooting through a glass surface. Such corruptions can be addressed via a single-image reflection removal (SIRR) task. Traditional SIRR methods employ different priors (\emph{e.g.}, sparsity \cite{levin2007user}, smoothness \cite{wan2016depth, Li_2014_CVPR}, and ghost \cite{Shih_2015_CVPR}) to exploit the special properties of the transmitted and reflection layers. In recent deep-learning-based methods, edge information \cite{Fan_2017_ICCV, Wan_2018_CVPR}, perceptual loss \cite{Zhang_2018_CVPR_sirr} and adversarial loss \cite{Wei_2019_CVPR} are used to improve the recovered transmitted layer.

SIRR can be seen as an image enhancement problem. It aims to recognize where the reflections are and then remove them to enhance the visibility of the background scene. The ultimate goal of our glass detection problem is not to recognize only the reflections but to detect the whole glass region, which may contain only partial or weak reflections.

\mhy{\textbf{Discriminative context learning.} Contextual information plays an important role in achieving high performance for many computer vision tasks. Many works are devoted to exploiting more discriminative contexts to enhance the ability of feature representation. For example, Chen \emph{et al.} \cite{chen2017deeplab} introduce an atrous spatial pyramid pooling (ASPP) module to capture multi-scale contextual information, which is then adopted by many methods such as \cite{Qiao_2020_CVPR_Matting,liu2021prior,Liu_2021_ICCV_Matting,qiao2021cpral}. Zhao \emph{et al.} \cite{Zhao_2017_CVPR_pspnet} employ a pyramid pooling module to aggregate local and global contexts. \add{Liu \emph{et al.} \cite{liu2018receptive} design a receptive field block (RFB) to simulate the configuration in terms of the size and eccentricity of the receptive fields in the human visual system via multi-scale contexts perception for enhancing the feature discriminability and robustness.} \add{In addition, multi-scale adaptive/dynamic contexts are proposed in \cite{He_2019_CVPR_adaptive,He_2019_ICCV_dynamic,wu2016cartoon}, multi-level contexts are extracted in \cite{yang2019drfn,Zhang_2020_ICME,nie2012compact,tian2020weakly,tian2021learning}, switchable contexts are used in \cite{lin2018scn}, large-field contextual features are captured in \cite{peng2017large}, direction-aware contexts are explored in \cite{Hu_2018_CVPR_dsc,wang2019spatial}, and contextual contrasted features are leveraged in \cite{Ding_2018_CVPR_contrast,yang2019mirrornet,Mei_2021_CVPR_PDNet,Lin_2020_CVPR_PMD,Lin_2021_CVPR_GSD}.}
Besides, a non-local operation \cite{Wang_2018_CVPR_non-local} is proposed to capture long-range contextual dependencies, and its variants are applied to various vision tasks, {\it e.g.}, \add{image segmentation \cite{Fu_2019_CVPR_danet,Huang_2019_ICCV_ccnet,Mei_2021_CVPR_PFNet,xie2021segmenting}} and image restoration \cite{li2018non,liu2018non,Dai_2019_CVPR_second-order}.}

\mhy{Our method differs from the above works in that we explore rich contextual information in a very large field-of-view. \add{Note that \cite{peng2017large} only extracts large-field contexts via spatially separable convolutions at a fixed scale while our proposed LCFI module can explore multi-scale large-field contexts.} We validate the effectiveness and generalization capability of the proposed large-field contextual feature learning with experiments.}

\def\wsample{0.08\linewidth}
\def\hsample{.65in}
\begin{figure*}[t]
	\setlength{\tabcolsep}{1.5pt}
	\centering
	\begin{tabular}{cccccccccccc}
		\includegraphics[width=\wsample, height=\hsample]{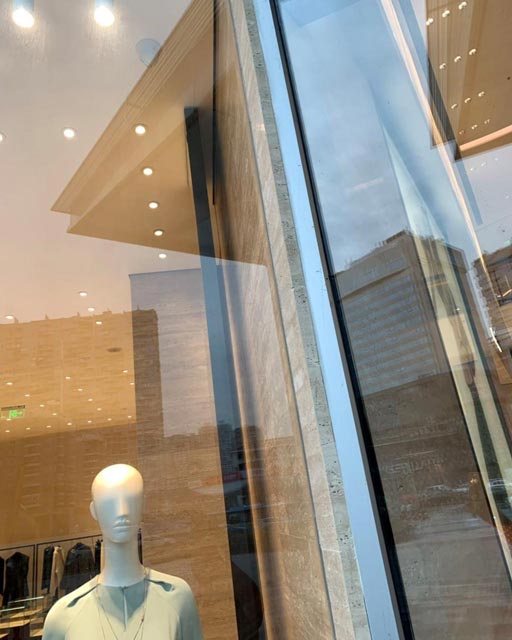}&
		\includegraphics[width=\wsample, height=\hsample]{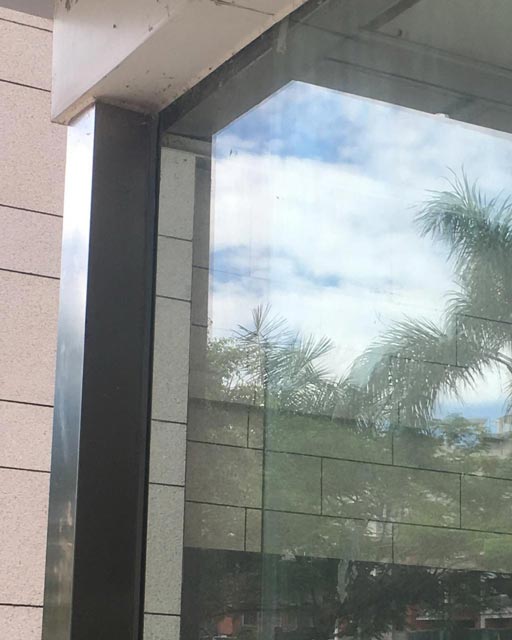}&
		\includegraphics[width=\wsample, height=\hsample]{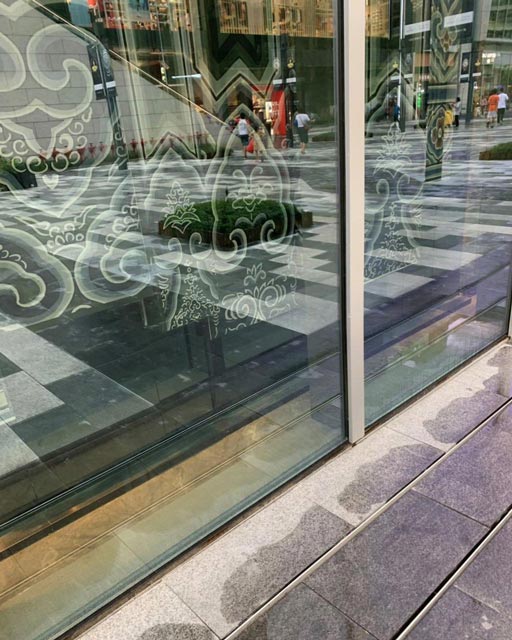}&
		\includegraphics[width=\wsample, height=\hsample]{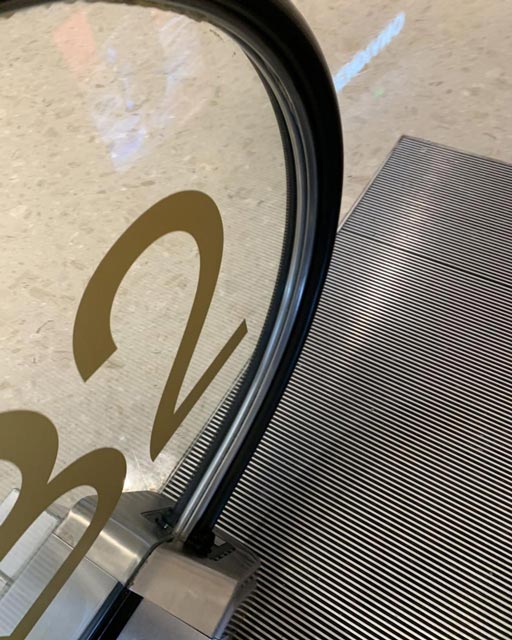}&
		\includegraphics[width=\wsample, height=\hsample]{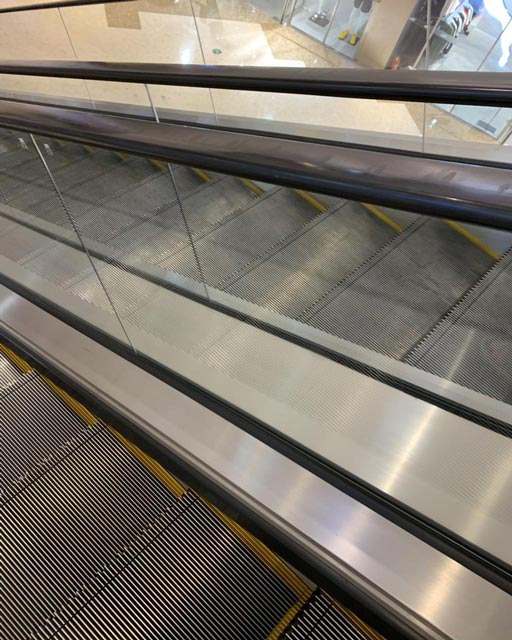}&
		\includegraphics[width=\wsample, height=\hsample]{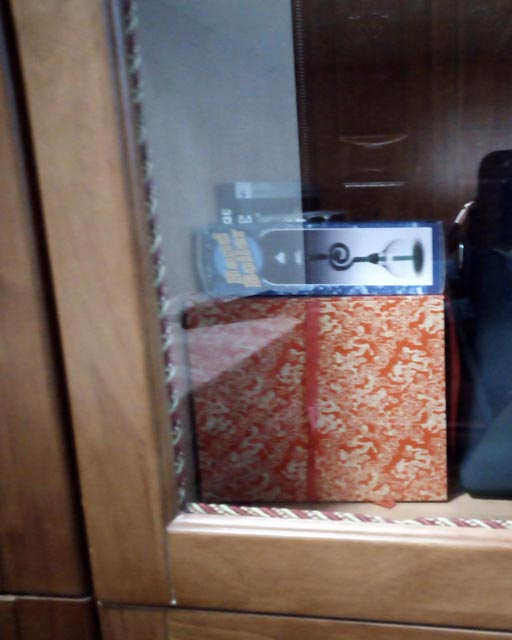}&
		\includegraphics[width=\wsample, height=\hsample]{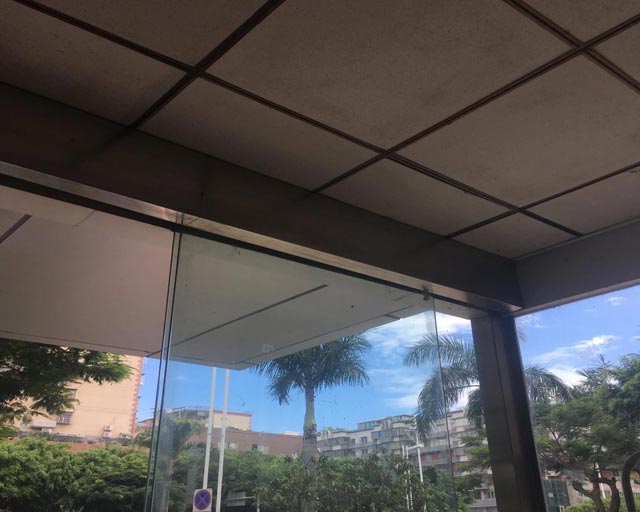}&
		\includegraphics[width=\wsample, height=\hsample]{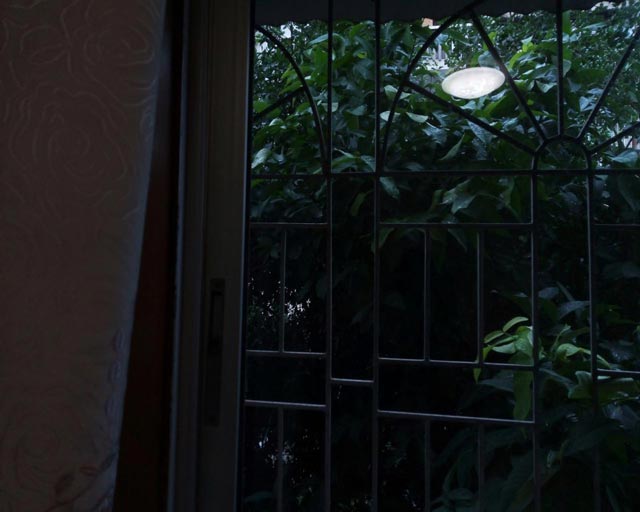}&
		\includegraphics[width=\wsample, height=\hsample]{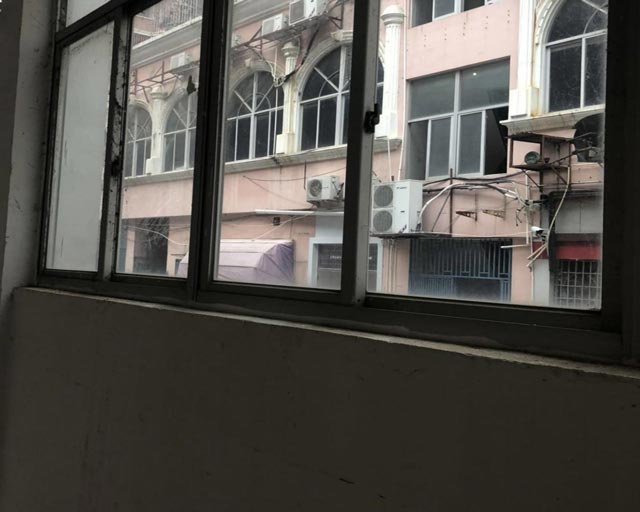}&
		\includegraphics[width=\wsample, height=\hsample]{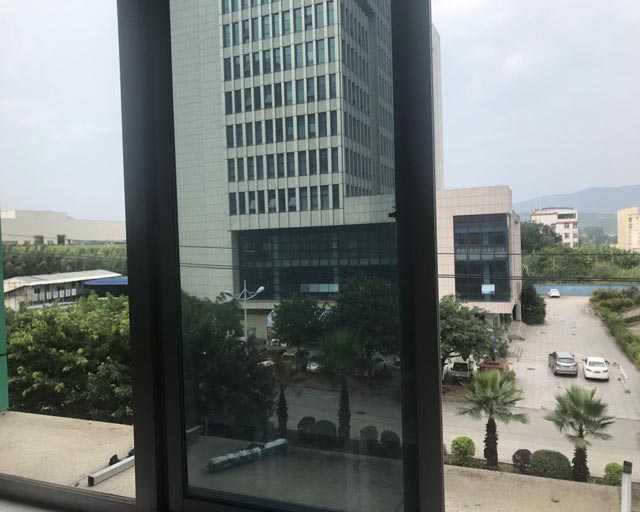}&
		\includegraphics[width=\wsample, height=\hsample]{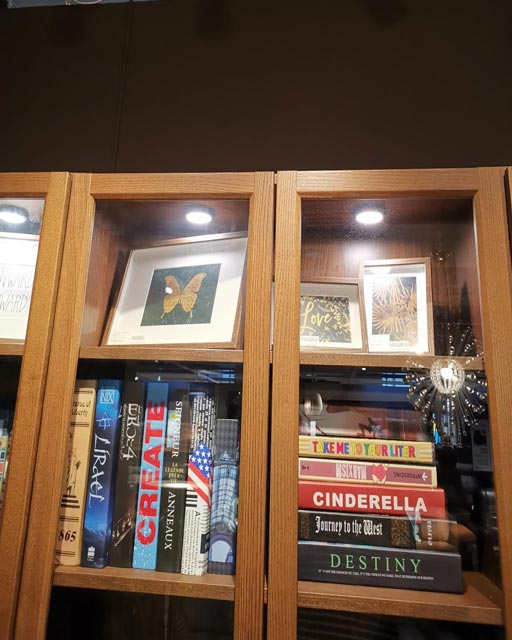}&
		\includegraphics[width=\wsample, height=\hsample]{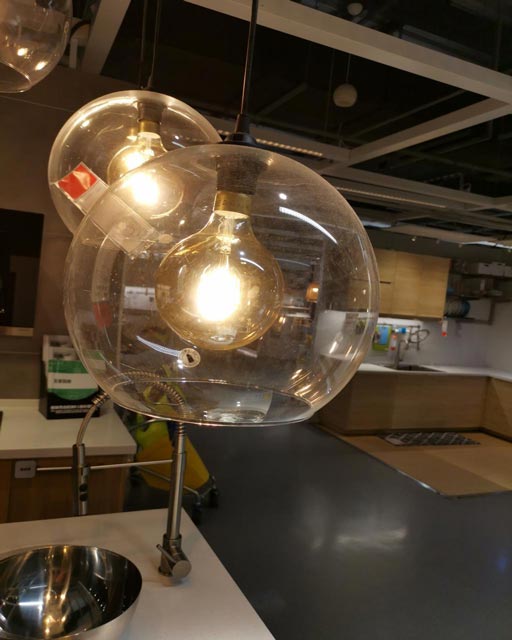} \\
		
		\includegraphics[width=\wsample, height=\hsample]{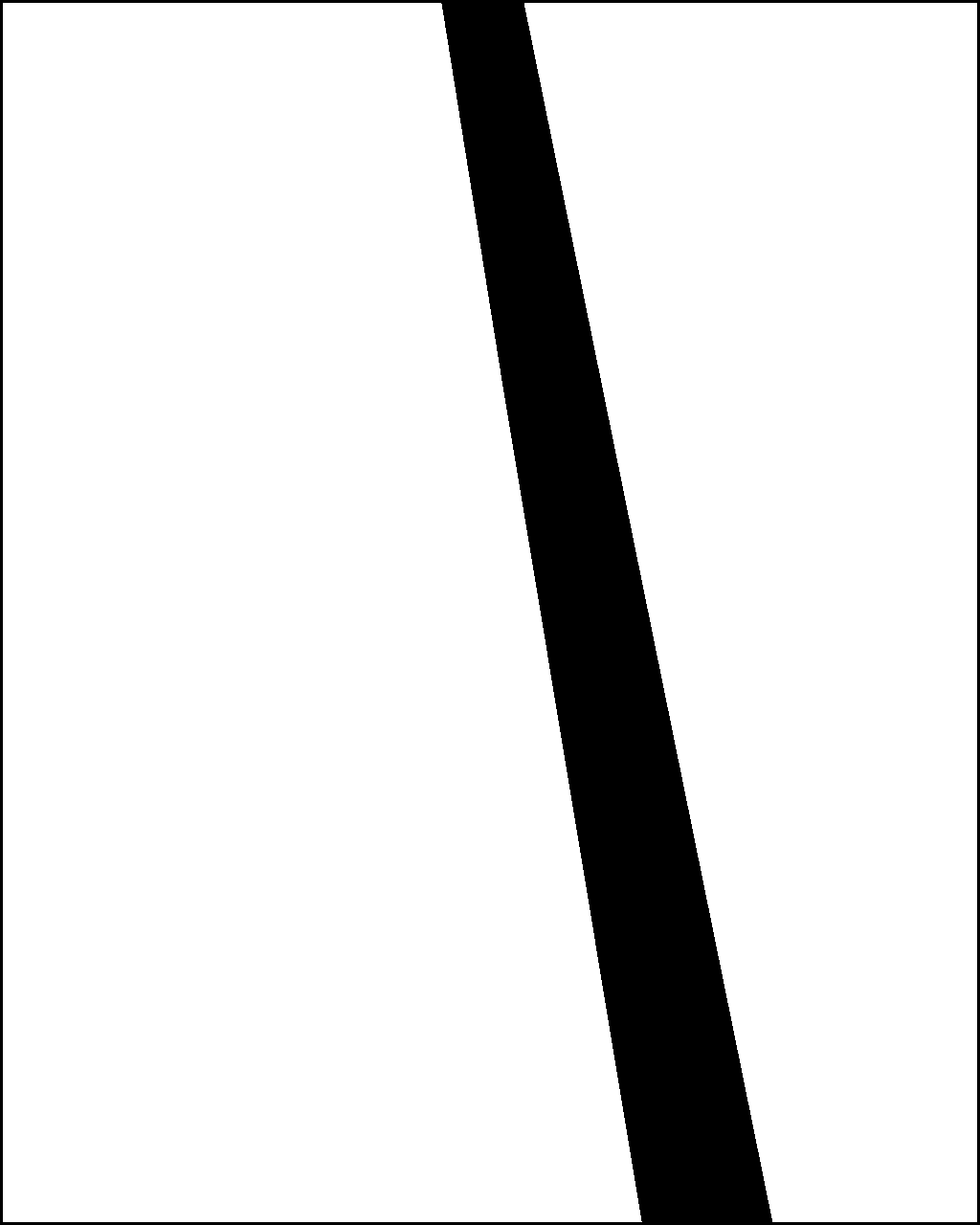}&
		\includegraphics[width=\wsample, height=\hsample]{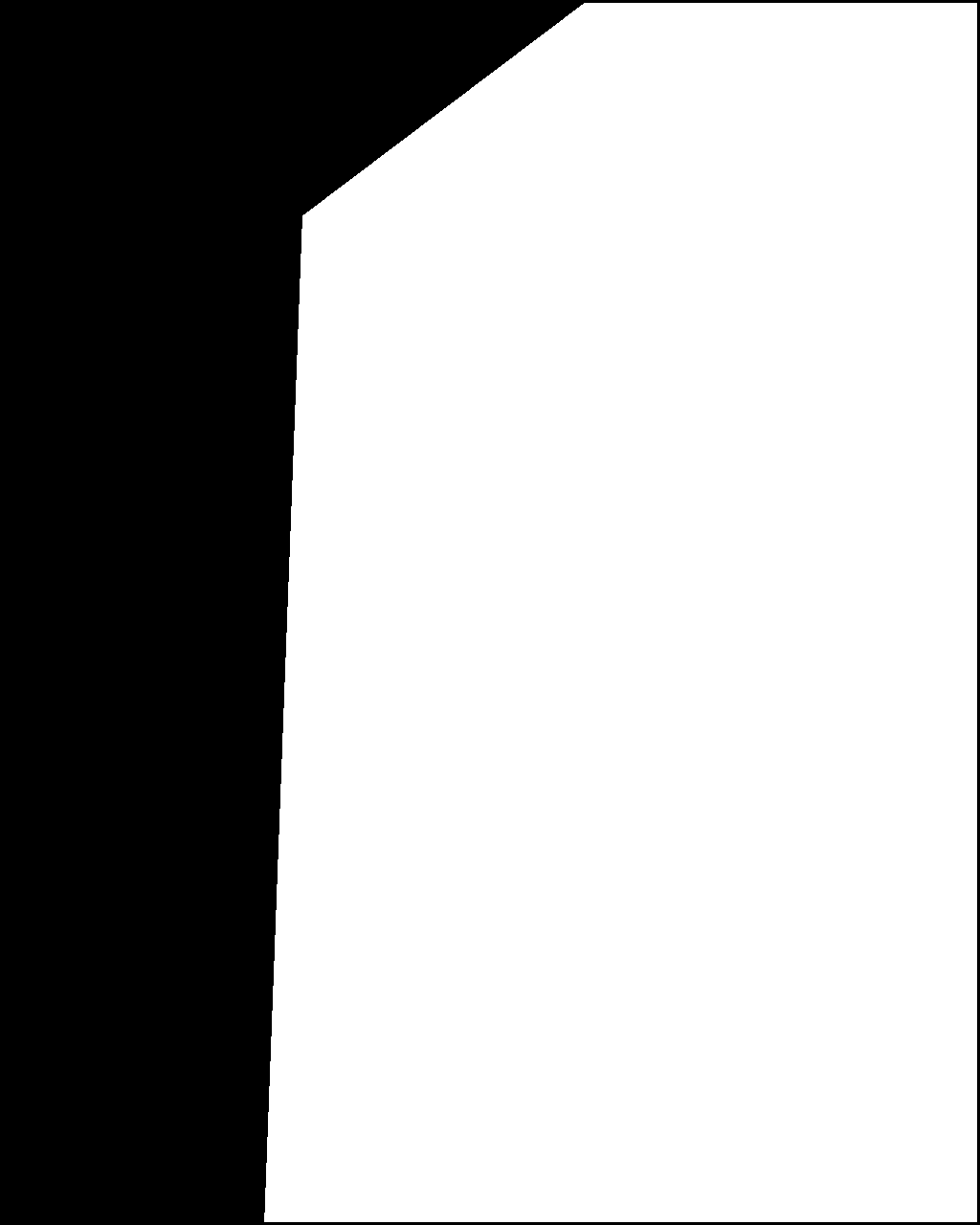}&
		\includegraphics[width=\wsample, height=\hsample]{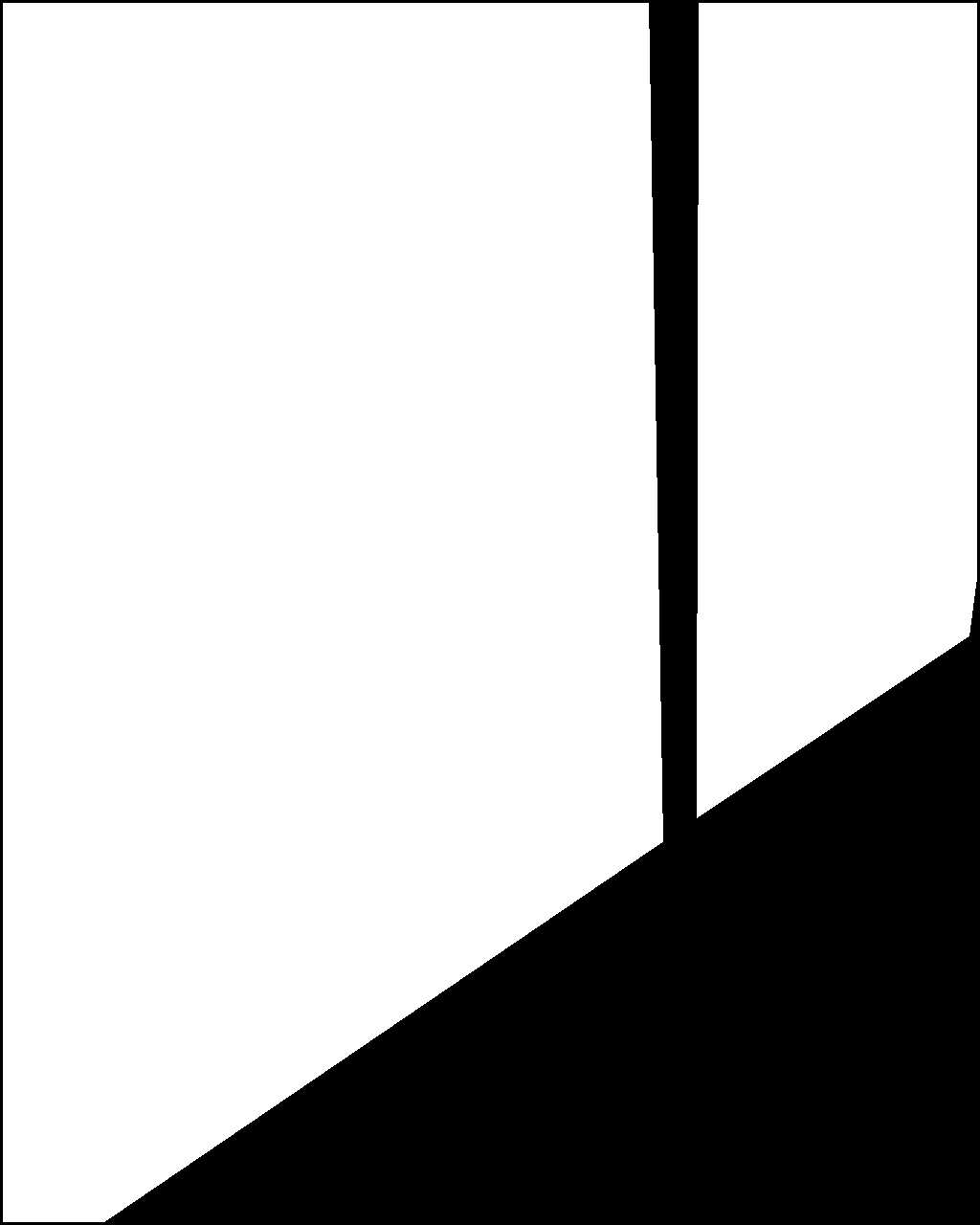}&
		\includegraphics[width=\wsample, height=\hsample]{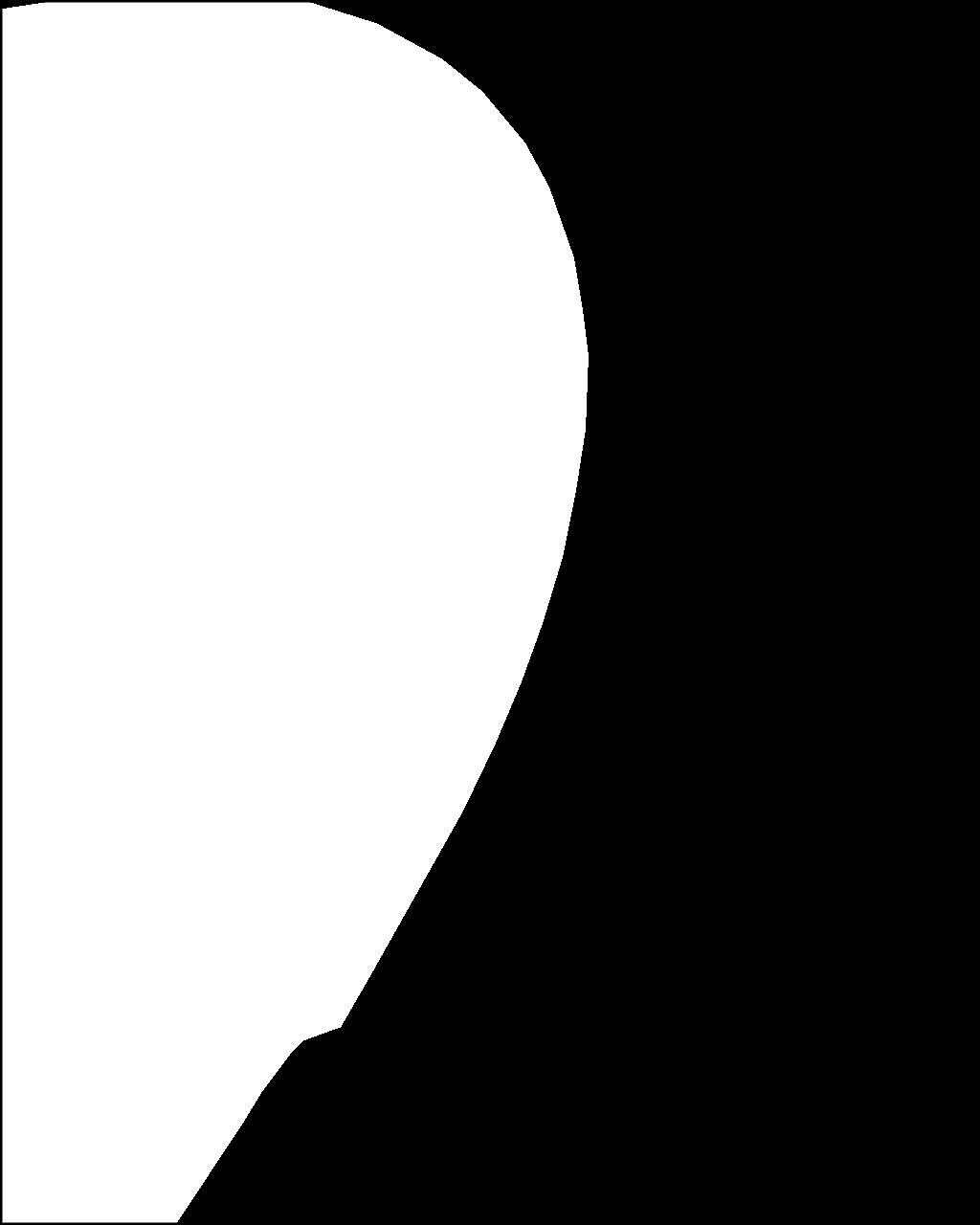}&
		\includegraphics[width=\wsample, height=\hsample]{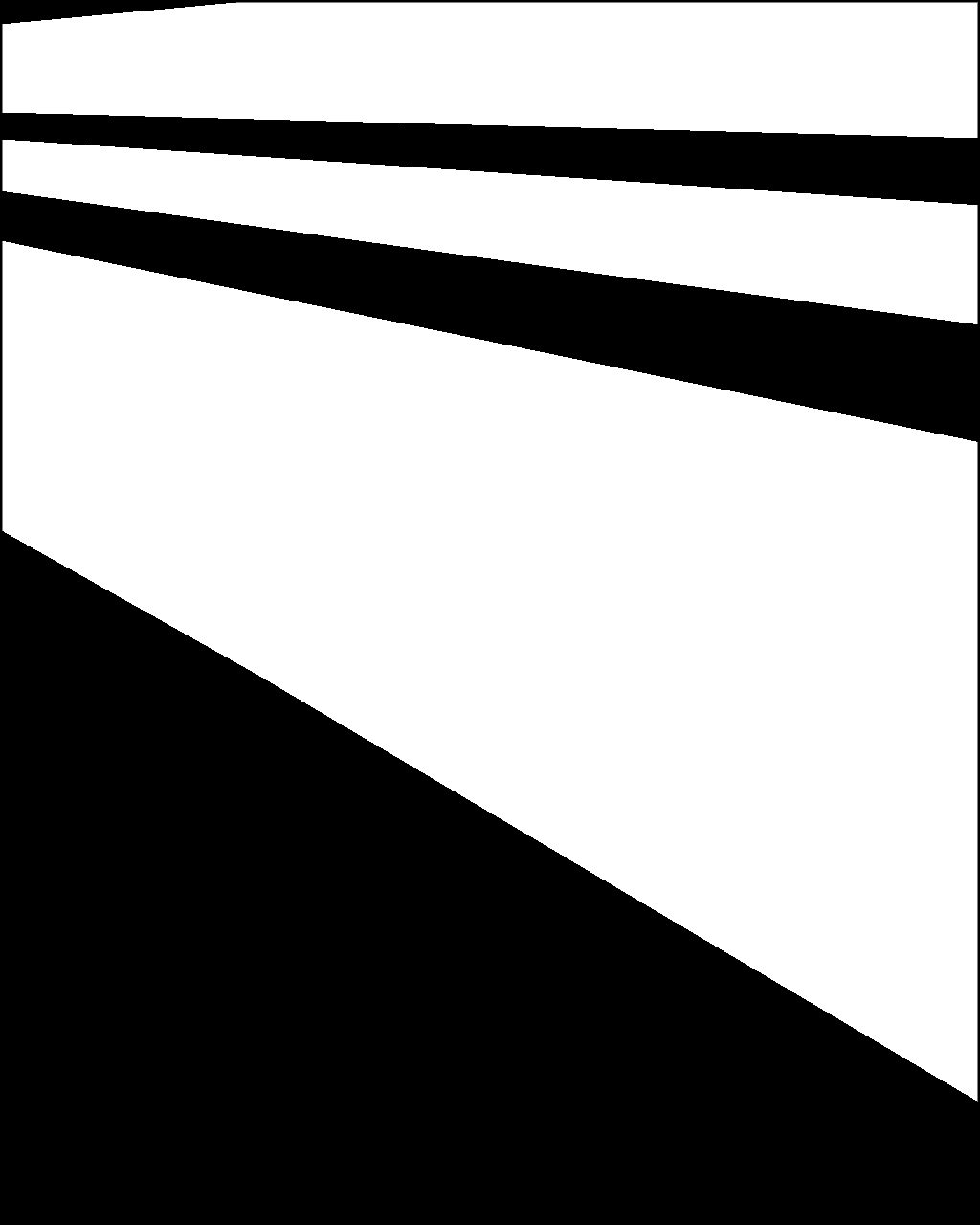}&
		\includegraphics[width=\wsample, height=\hsample]{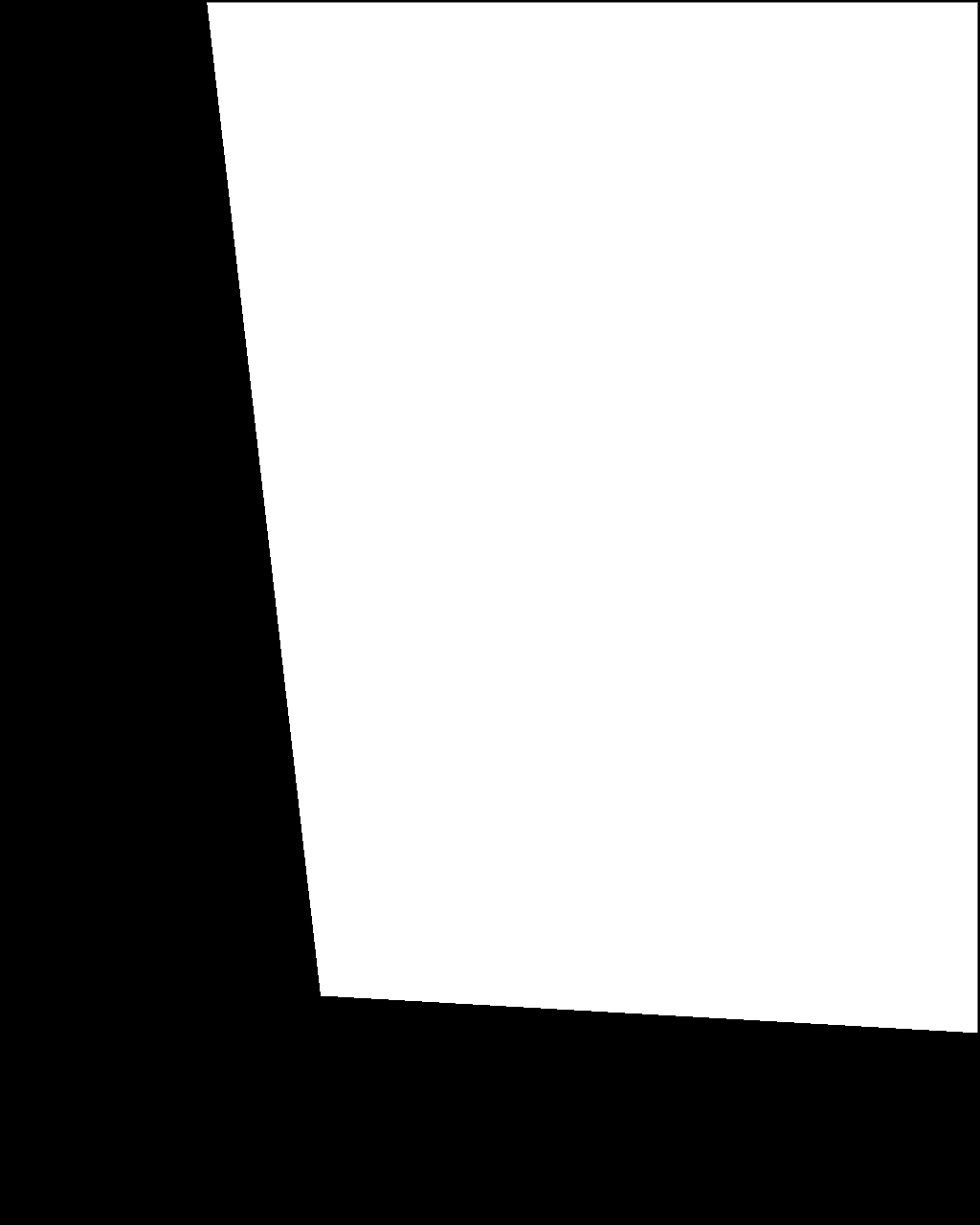}&
		\includegraphics[width=\wsample, height=\hsample]{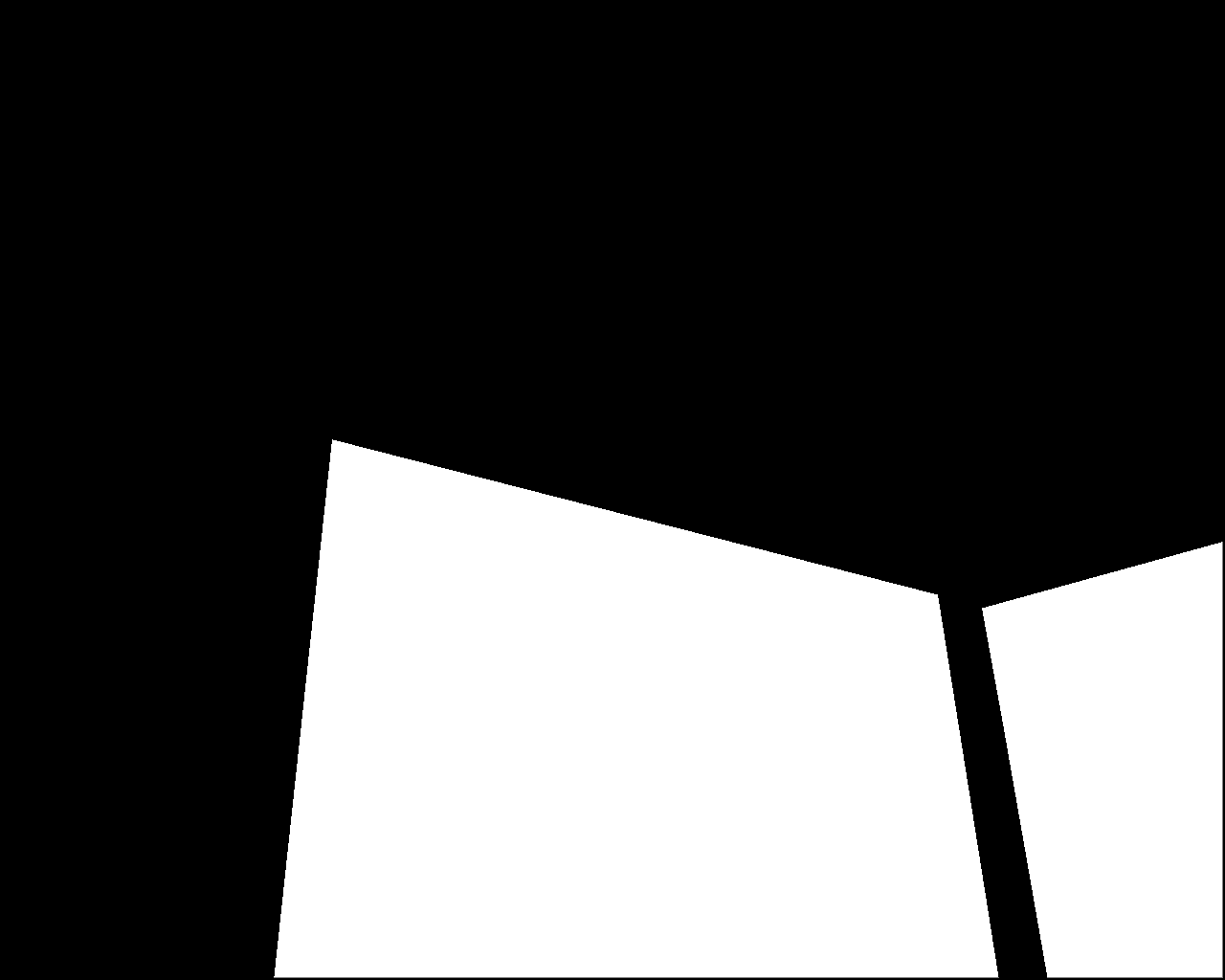}&
		\includegraphics[width=\wsample, height=\hsample]{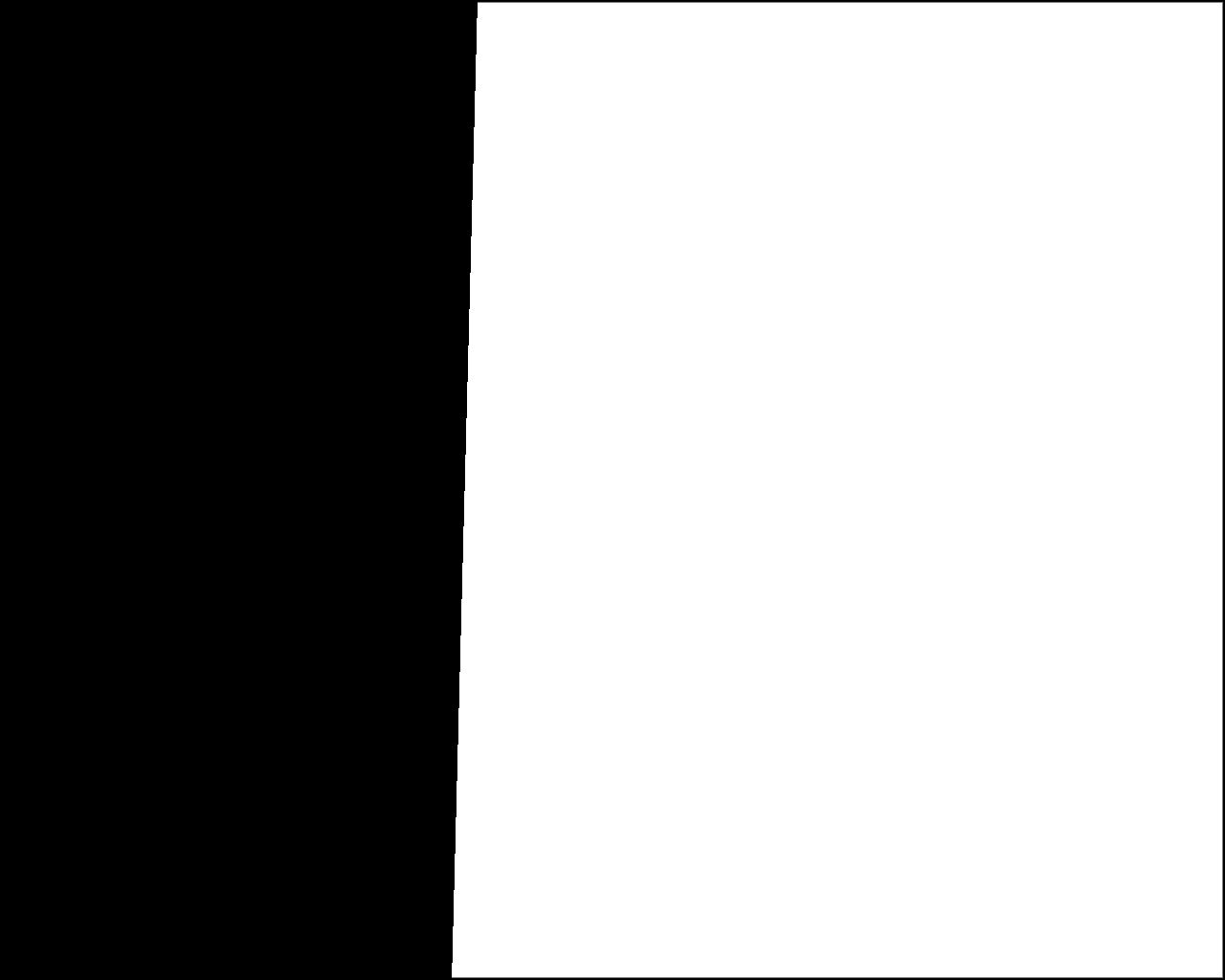}&
		\includegraphics[width=\wsample, height=\hsample]{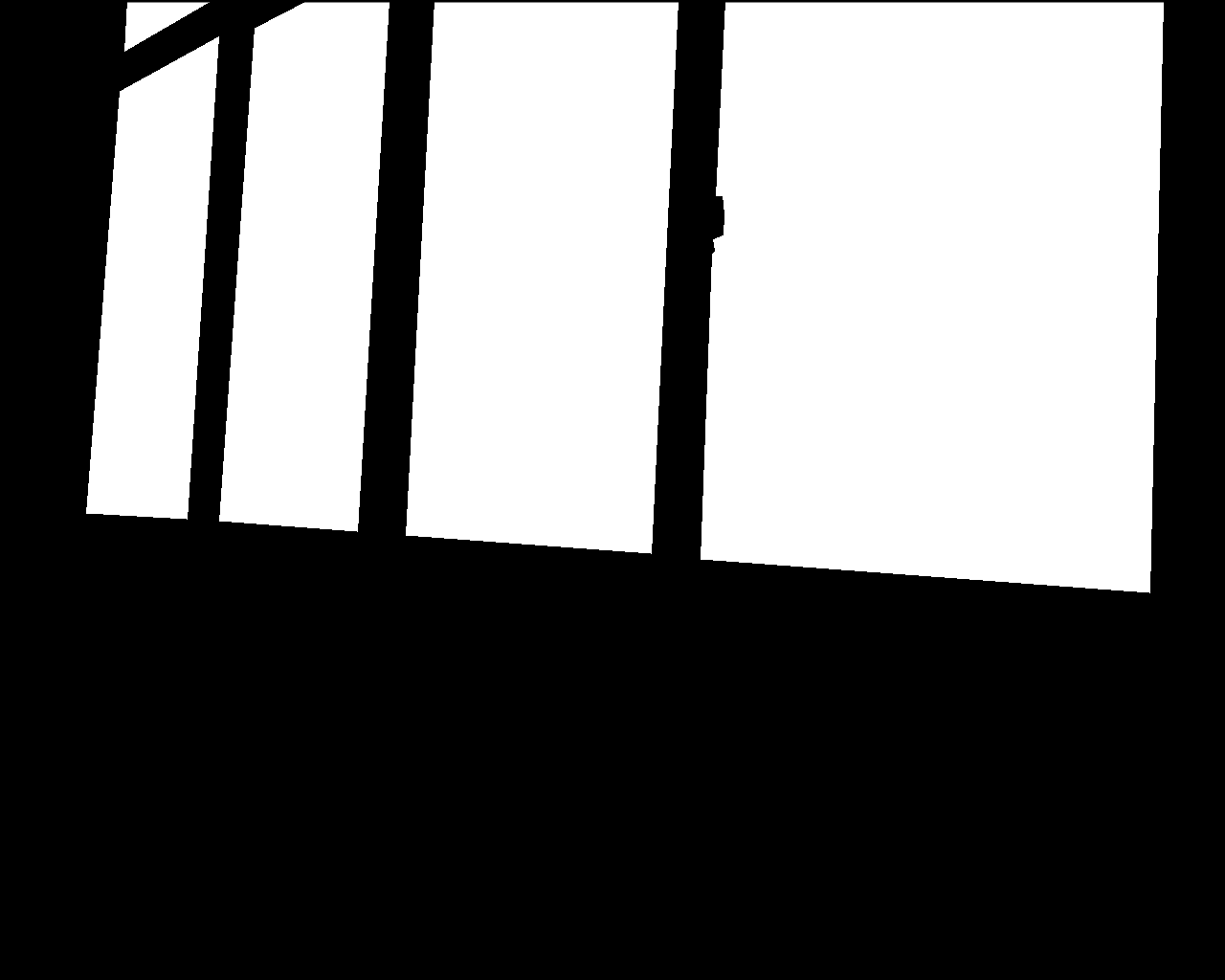}&
		\includegraphics[width=\wsample, height=\hsample]{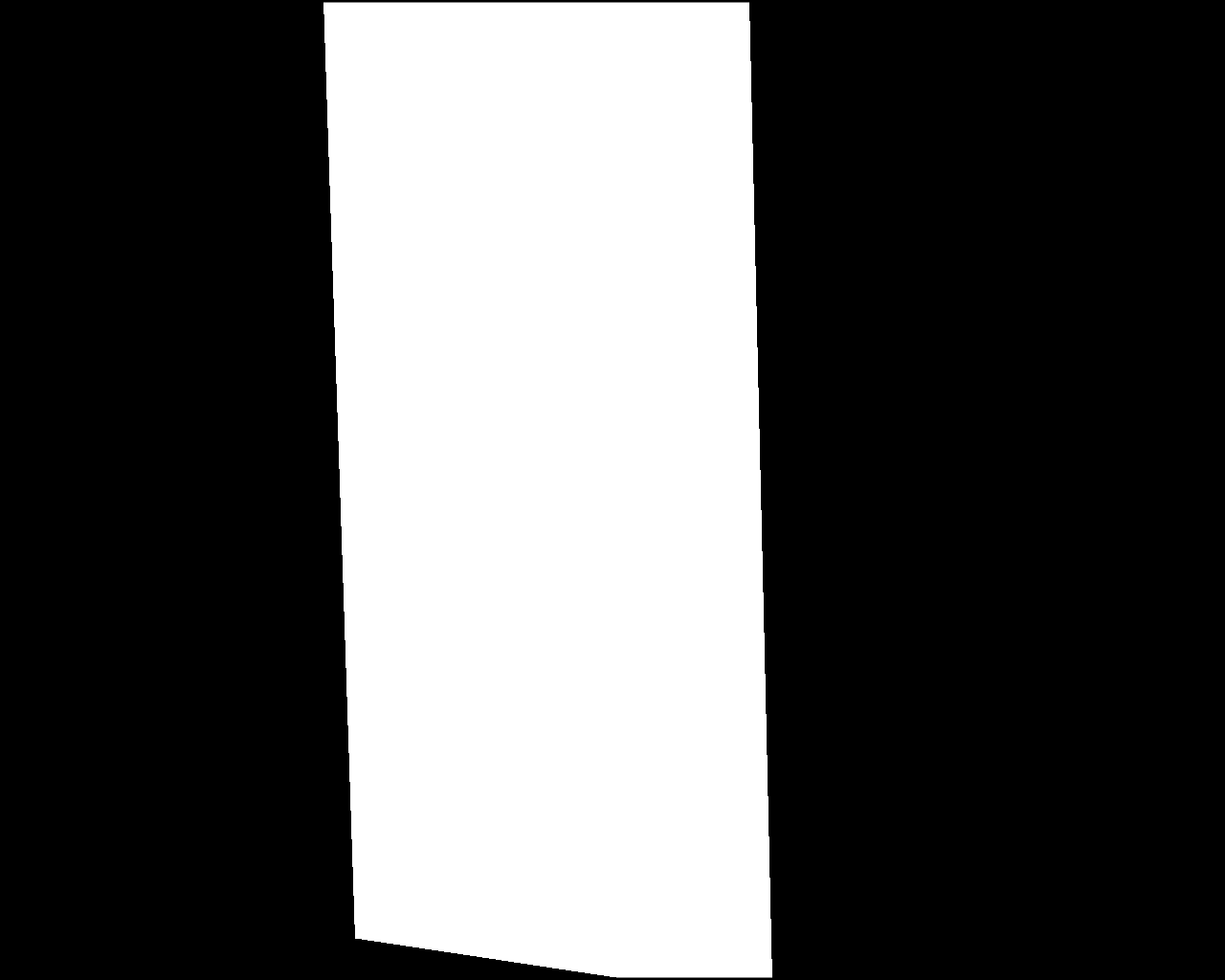}&
		\includegraphics[width=\wsample, height=\hsample]{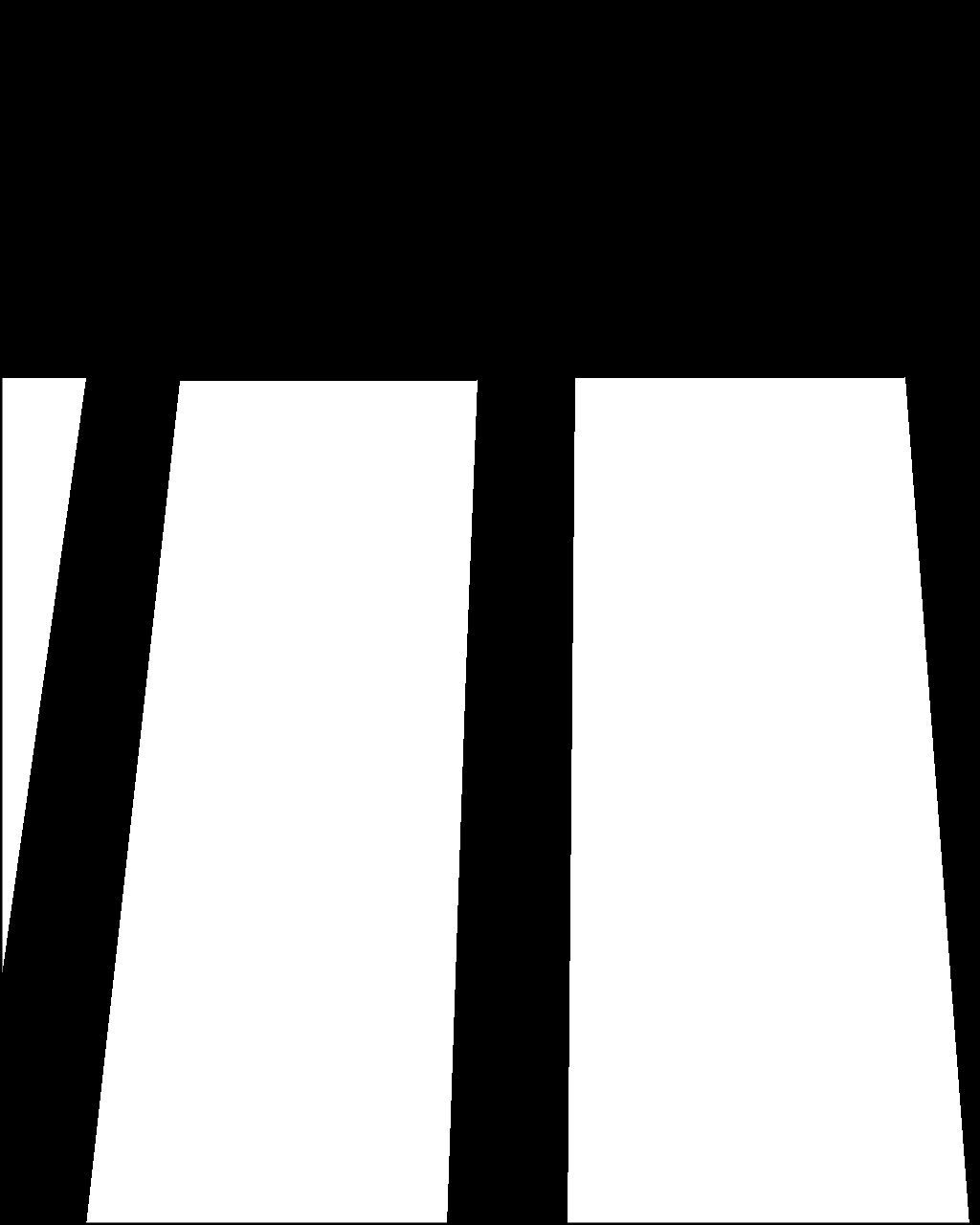}&
		\includegraphics[width=\wsample, height=\hsample]{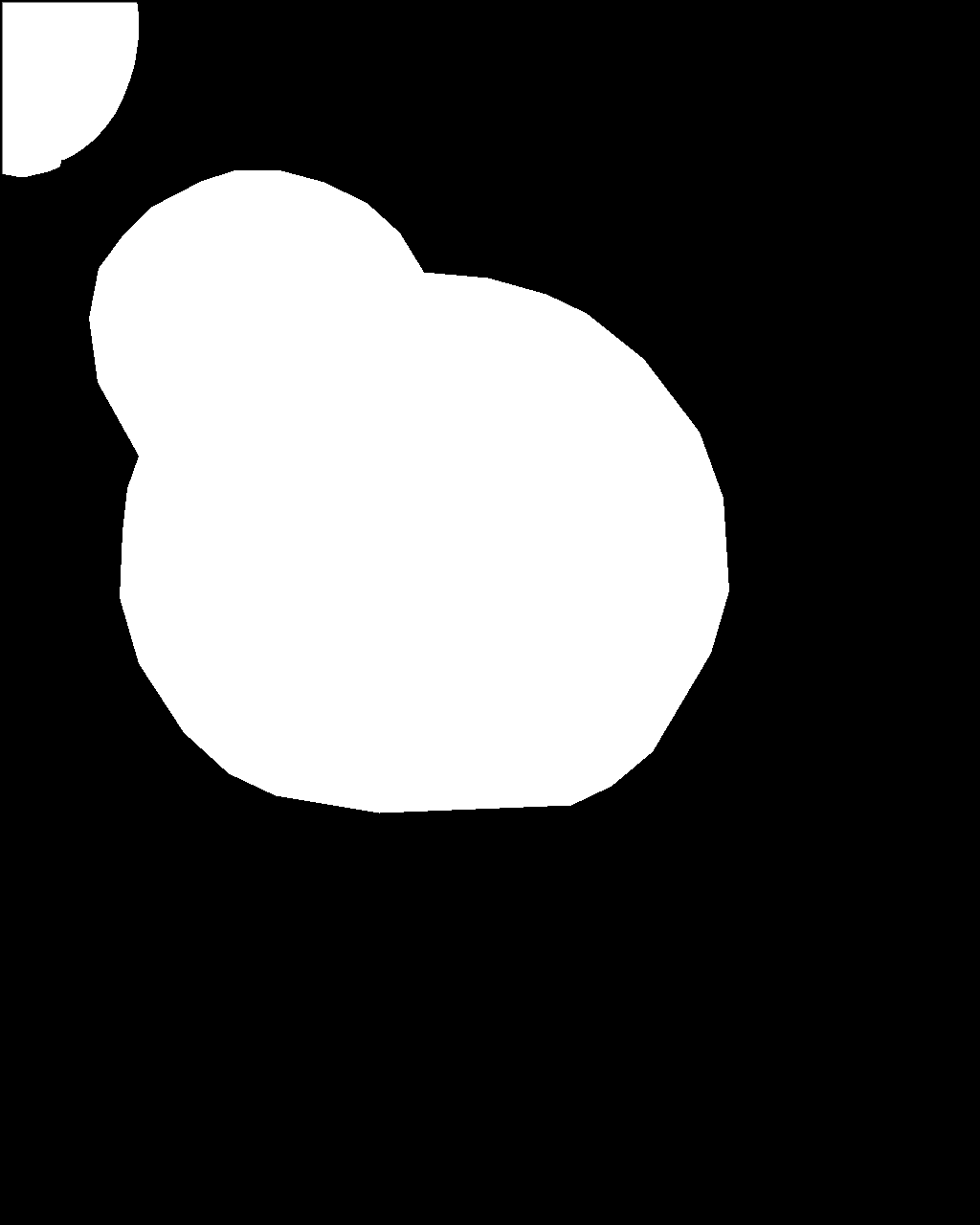} \\
		
		
	\end{tabular}
	\caption{Example glass image/mask pairs in our glass detection dataset (GDD). It shows that GDD covers diverse types of glass in daily-life scenes.}
	\label{fig:dataset}
\end{figure*}

\def\wdataset{0.25\linewidth}
\def\hdataset{.9in}
\begin{figure*}[t]
	\setlength{\tabcolsep}{1pt}
	\centering
	\begin{tabular}{cccc}
		\includegraphics[width=\wdataset, height=\hdataset]{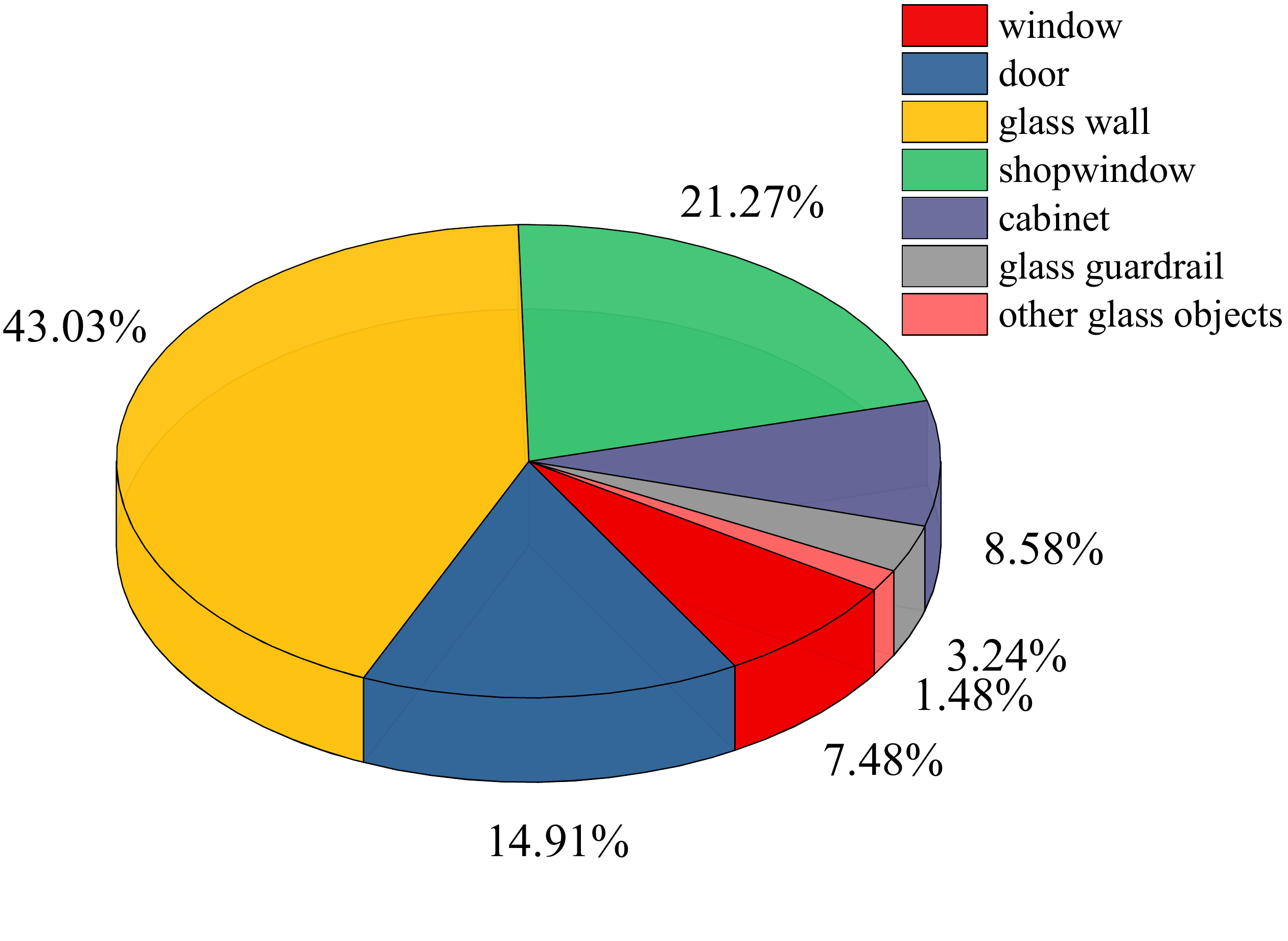} &
		\includegraphics[width=\wdataset, height=\hdataset]{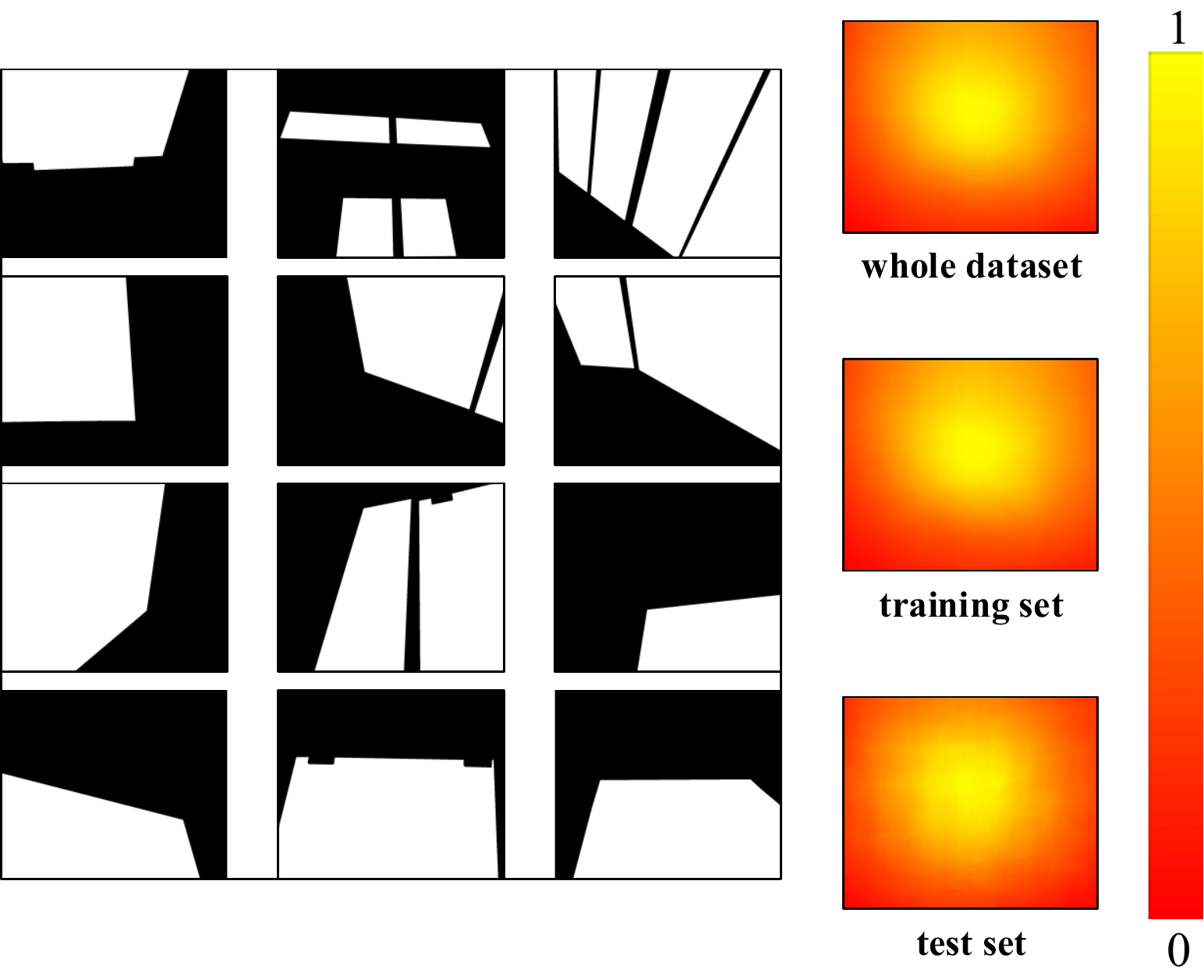} &
		\includegraphics[width=\wdataset, height=\hdataset]{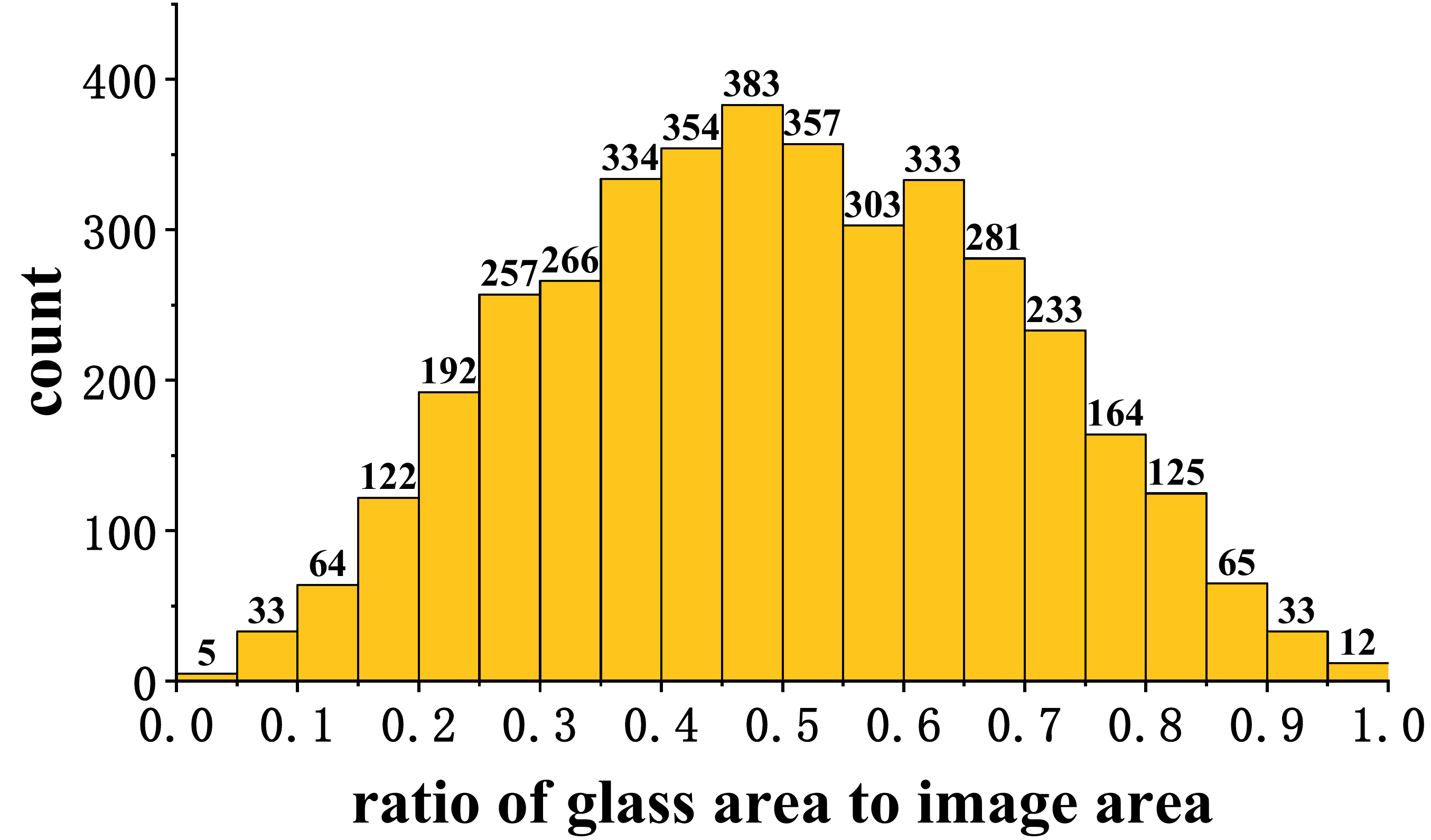} &
		\includegraphics[width=\wdataset, height=\hdataset]{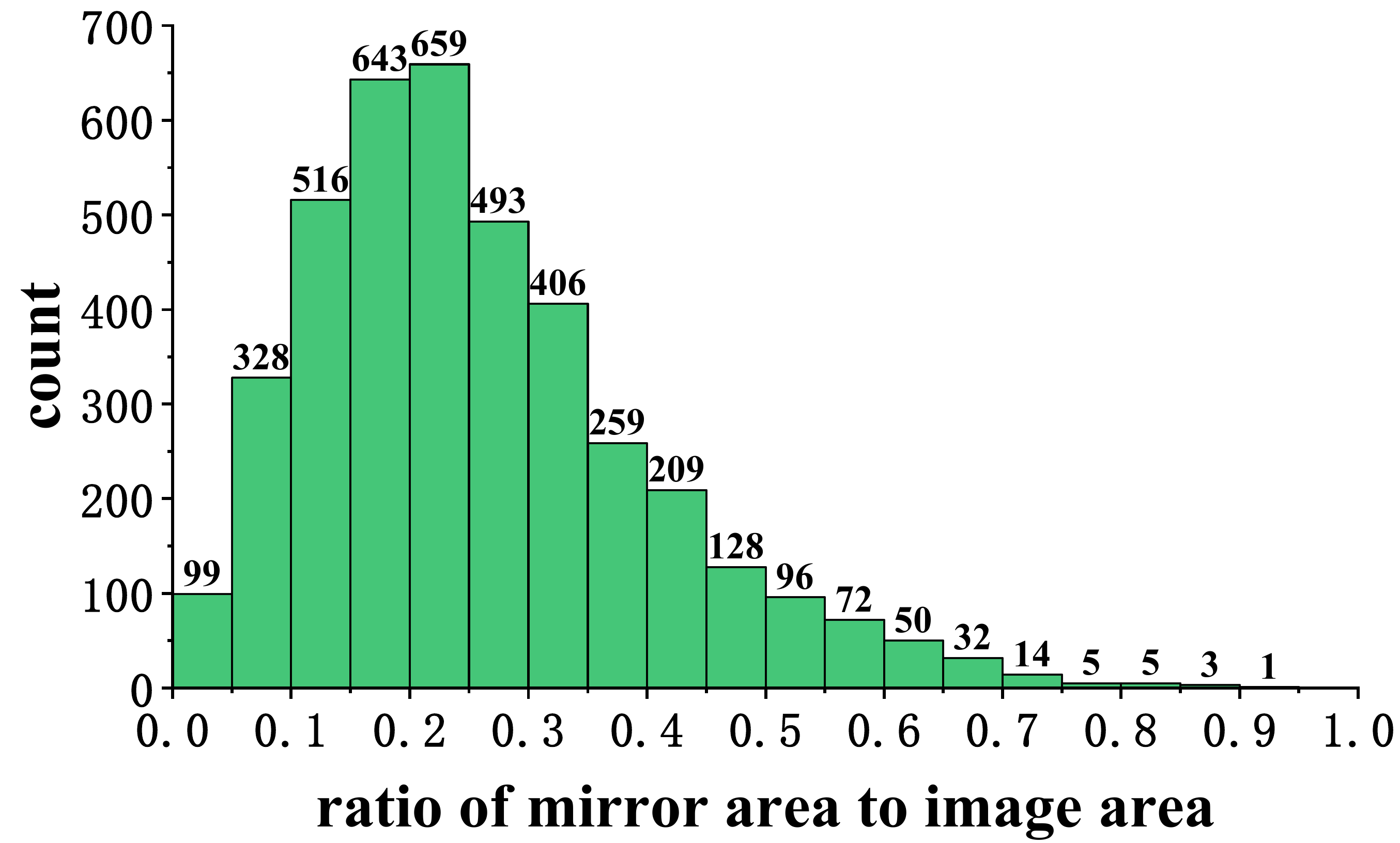} \\
		{\small (a) glass type distribution} & {\small (b) glass location distribution} & {\small (c) glass area distribution} & {\small (d) mirror area distribution} \\
	\end{tabular}
	\caption{Statistics of our dataset. We show that GDD has glass with reasonable property distributions in terms of type, location and area.}
	\label{fig:statistic}
\end{figure*}

\section{A New Dataset for Glass Detection}
\mhy{The emergence of new tasks and datasets (\emph{e.g.}, \cite{Cordts_2016_CVPR_cityscape,everingham2015the,Zhou_2017_CVPR,yang2019mirrornet,tan2021night}) has led to rapid progress in various areas of computer vision. For example, ImageNet \cite{deng2009imagenet} revolutionizes the use of deep models for visual recognition. With this in mind, our goals for studying and developing a new dataset for glass detection are: (\romannumeral1) to propose a new meaningful yet challenging task, (\romannumeral2) to promote research on this new topic, and (\romannumeral3) to spark novel ideas.}
In this paper, we contribute a large-scale glass detection dataset (GDD), which contains 3,916 pairs of glass and glass mask images. To the best of our knowledge, GDD is the first large-scale benchmark specifically for glass detection.

\textbf{Dataset construction.}
\add{Our glass images are captured with some latest cameras and smart phones. The pixel-level glass masks (with pixels inside the glass region set to 1 and 0 otherwise) are labeled by professional annotators with a professional labeling tool. The glass mask is checked by three viewers to ensure labeling accuracy.}
Our constructed glass detection dataset, GDD, covers diverse daily-life scenes (\emph{e.g.}, bathroom, office, street, and mall), in which 2,827 images are taken from indoor scenes and 1,089 images are taken from outdoor scenes. Figure \ref{fig:dataset} shows some example glass and glass mask images in GDD. For dataset split, 2,980 images are randomly selected for training and the remaining 936 images are used for testing.

\textbf{Dataset analysis.}
To validate the diversity of GDD and how challenging it is, we show its statistics as follows:
\begin{itemize}
	\item \textbf{Glass type.} As shown in Figure \ref{fig:statistic}(a), there are various types of common glass in GDD (\emph{e.g.}, shopwindow, glass wall, glass door, glass guardrail, and glass on window and cabinet). Other relatively small glass objects such as glass bulbs and glass clocks are also included. The reason that such glass objects occupy only a small ratio in GDD is that in this work, we aim to detect relatively large transparent glass that could contribute critical effect to scene understanding. The small glass objects are mainly to add diversity.
	\item \textbf{Glass location.} Our GDD has glass located at different positions of the image, as illustrated in Figure \ref{fig:statistic}(b). We further compute probability maps that indicate how likely each pixel belongs to a glass region, to show the location distributions of glass in GDD. The overall spatial distribution tends to be centered, as glass is typically large and covers the center. Besides, the glass spatial distributions for the training/test splits are consistent to those of the whole dataset.
	\item \textbf{Glass area.} We define the size of the glass region as a proportion of pixels in the image. In Figure \ref{fig:statistic}(c), we can see that the glass in our GDD varies in a wide range in terms of size, and much of it falls in the range of [0.2, 0.8]. Glass falling in the range of (0, 0.2] represents small glass objects or glass corners. Such small glass regions can be easily cluttered with diverse background objects/scenes. Glass falling in the range of (0.8, 1.0) is typically located close to the camera. In this situation, the content of the image is dominated by the complicated scene behind the glass. Extreme cases, \emph{i.e.}, glass area equals to 0 or 1, are not included in GDD. Compared with the distribution of the mirrors in the mirror segmentation dataset (MSD) \cite{yang2019mirrornet} as shown in Figure \ref{fig:statistic}(d), glass in our GDD typically has a larger area, which means that more objects/scenes would be presented inside the glass, making GDD more challenging.
\end{itemize}



\section{Methodology}
\mhy{Considering that there may exist only a few or very weak reflections on the glass and the whole glass region has no fixed patterns, \emph{i.e.}, arbitrary objects/scenes can appear behind the glass in real-world scenes, we do not leverage the off-the-shelf physical model to capture reflections nor focus on exploring a specific physical cue to detect the whole glass region.}
Instead, we observe that humans can identify the existence of glass well, by considering contextual information in terms of low-level cues (\emph{e.g.}, color difference between inside and outside the glass, blur/bright spot/ghost caused by reflection) as well as high-level contexts (\emph{e.g.}, relations between different objects) and boundary information. This inspires us to leverage both abundant contextual features and boundary features for glass detection.

To this end, first, we propose a novel Large-field Contextual Feature Integration (LCFI) block to extract abundant contextual features from a large field for context inference and glass localization. Second, based on the LCFI block, a novel LCFI module is designed to effectively integrate multi-scale large-field contextual features for detecting glass of different sizes. Third, we embed multiple LCFI modules to the glass detection network (GDNet) to obtain large-field contextual features of different levels.
\mhy{Finally, we apply two boundary feature enhancement (BFE) modules to integrate both high-level and low-level boundary cues for the accurate glass detection under various scenes.}

\begin{figure*}[tbp]
	\begin{center}
		\includegraphics[width = 1\linewidth]{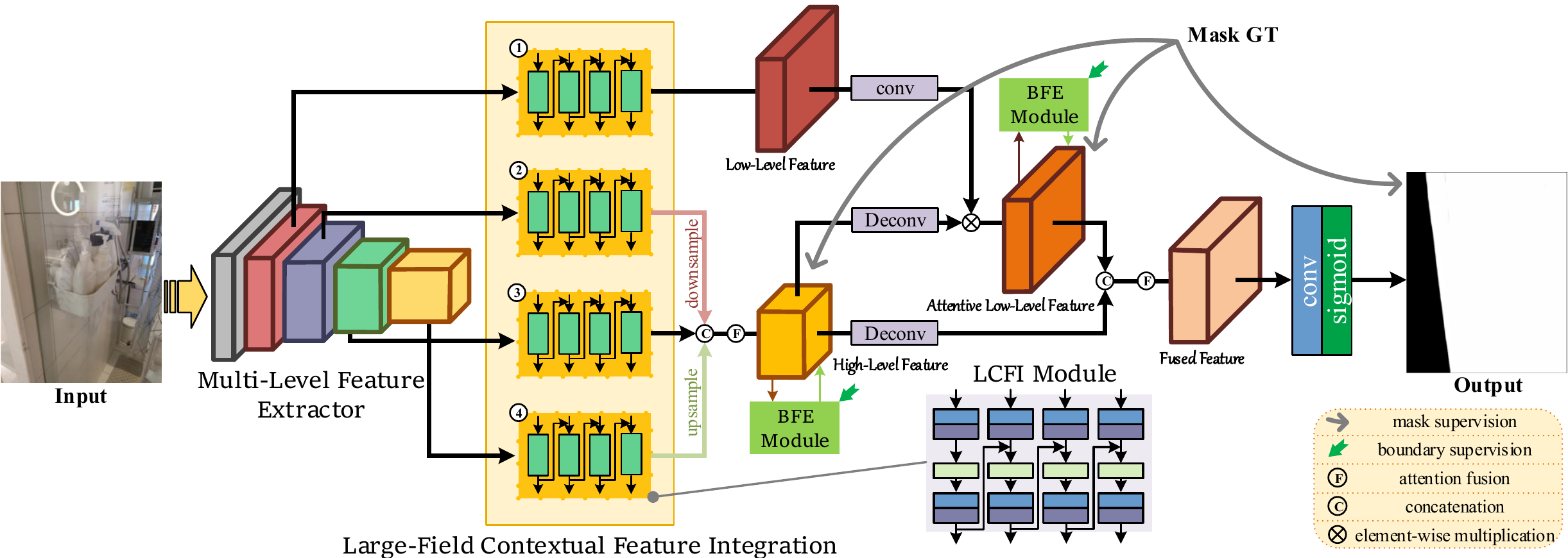}
	\end{center}
	\caption{The pipeline of the proposed GDNet-B. First, we use the pre-trained ResNeXt-101 \cite{Xie_2017_CVPR} as a multi-level feature extractor (MFE) to obtain features of different levels. Second, we embed four LCFI modules to the last four layers of MFE, to learn large-field contextual features at different levels. Third, the outputs of the last three LCFI modules are concatenated and fused via an attention module \cite{Woo_2018_ECCV} to generate high-level large-field contextual features. An attention map is then learned from these high-level large-field contextual features and used to guide the low-level large-field contextual features, \emph{i.e.}, the output of the first LCFI module, to focus more on glass regions. Fourth, we apply two BFE modules on the high-level/attentive low-level large-field contextual features to further perceive and integrate boundary cues. Finally, we combine high-level and attentive low-level large-field contextual features by concatenation and attention \cite{Woo_2018_ECCV} operations to produce the final glass detection map.}
	\label{fig:GDNet}
	\vspace{7pt}
\end{figure*}

\begin{figure*}[t]
	\begin{center}
		\includegraphics[width = 1\linewidth]{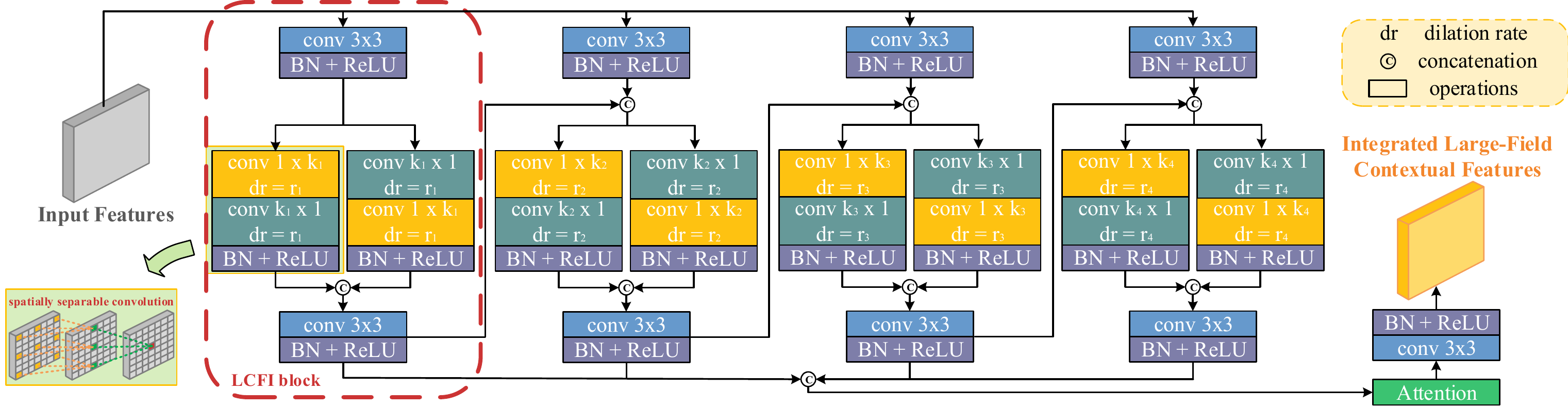}
	\end{center}
	\caption{The structure of the proposed LCFI module. The input features are passed through four parallel LCFI blocks, and the outputs of all LCFI blocks are fused to generate multi-scale large-field contextual features. In each LCFI block (red dashed box), input features are fed to two parallel spatially separable convolutions with opposite convolution orders to obtain large-field contextual features with different characteristics. The output of the current LCFI block is then fed to the next LCFI block to be further processed in a larger field.}
	\label{fig:LCFI}
	\vspace{5pt}
\end{figure*}

\subsection{Network Overview}
\begin{figure}[tbp]
	\begin{center}
		\includegraphics[width = 1\linewidth]{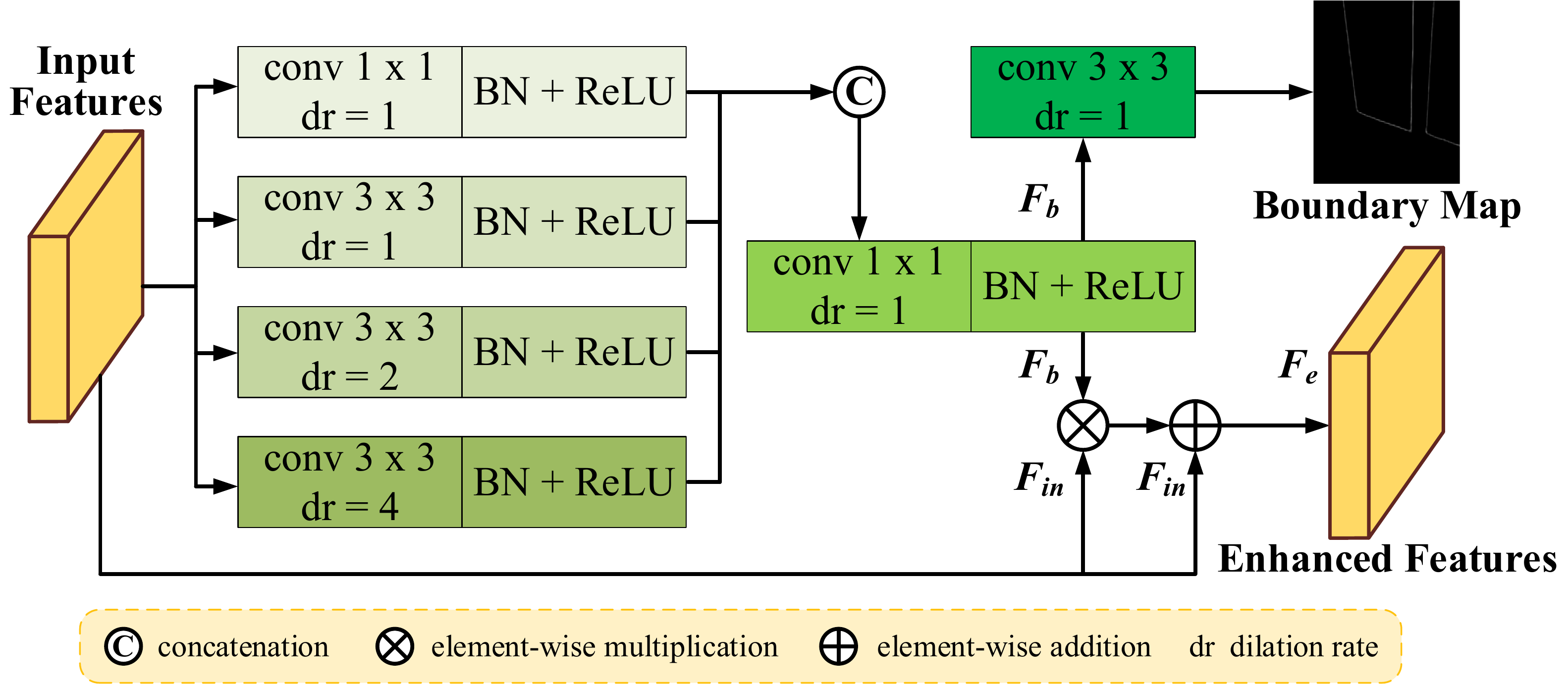}
	\end{center}
	\caption{\mhy{The structure of the proposed BFE module. The input features are passed through four parallel branches, and the outputs of all branches are fused to generate glass boundary features, which are then used to predict the glass boundary map and complement the input features.}}
	\label{fig:BFE}
\end{figure}

Figure \ref{fig:GDNet} presents the proposed glass detection network (GDNet-B). It employs the LCFI module (Figure \ref{fig:LCFI}) to learn large-field contextual features and the BFE module (Figure \ref{fig:BFE}) to integrate boundary information. Given a single input image, we first feed it into the multi-level feature extractor (MFE) to harvest features of different levels, which are further fed into four proposed LCFI modules to learn large-field contextual features. The outputs of the last three LCFI modules are fused to generate high-level large-field contextual features, which are then used to guide the first LCFI module to focus more on the glass regions as it extracts low-level large-field contextual features. \mhy{We then apply the well-designed BFE module on both the high-level features and attentive low-level features to perceive and integrate boundary cues.} Finally, we fuse high-level and attentive low-level large-field contextual features to produce the final glass detection result.

\subsection{Large-Field Contextual Feature Integration}
Figure \ref{fig:LCFI} illustrates the structure of our LCFI module. Given the input features, the LCFI module aims to efficiently and effectively extract and integrate multi-scale large-field contextual features, for the purpose of detecting glass of different sizes.

\textbf{LCFI block.} The LCFI is designed to efficiently extract abundant contextual information from a large field for context inference and glass localization. The common practice to obtain larger context information is to use convolutions with large kernels or dilated convolutions. However, large kernels would result in heavy computation and a large dilation rate would lead to sparse sampling. Non-local operations \cite{Wang_2018_CVPR} could provide long-range dependencies but also suffer from huge computation. Here, we propose to use spatially separable convolutions. Spatially separable convolutions are convolutions that can be separated across their spatial axes and enable densely connections within a large region in the feature map while at the same time using fewer parameters than the traditional convolutions. Thus, we propose to use them to achieve the goal of efficiently extracting abundant contexts from a large field:
\begin{equation}\label{equ:lcf}
	F_c = \aleph(conv_h(conv_v(F))),
\end{equation}
where $F$ denotes the input features. $conv_v$ and $conv_h$ refer to vertical convolution with a kernel size of $k\times 1$ and horizontal convolution with a kernel size of $1\times k$, respectively. $\aleph$ represents batch normalization and ReLU operations. $F_{c}$ denotes the extracted large-field contextual features.


As the content inside a glass region is typically complicated, contextual features with different characteristics are needed to eliminate the ambiguity. Thus, we use another spatially separable convolution with reverse convolution order, \emph{i.e.}, $\aleph(conv_v(conv_h(F)))$, to extract complementary large-field contextual features.
\mhy{As the contextual cues in the vertical and horizontal directions are usually not exactly the same, the intermediate features extracted by the first vertical/horizontal convolutions in the two paths would have different characteristics. Thus, the final features of the two paths are different and may be complementary as they are further learned by convolutions with orthogonal directions.}
Besides, we adopt dilated spatially separable filters to ensure that more contexts can be explored in a larger field. Finally, the large-field contextual features extracted from the two parallel paths are fused by a $3 \times 3$ convolution followed by BN and ReLU. The tasks of the LCFI block can be formulated as:
\begin{equation}\label{equ:lcfi}
	\begin{split}
		F_{lcfi}  = & \ \aleph(conv_2(concat(\aleph(conv_v(conv_h(F_{l}))), \\
		& \ \ \ \ \ \ \ \aleph(conv_h(conv_v(F_{l})))))), \\
		F_{l} = & \ \aleph(conv_1(F_{in})),
	\end{split}
\end{equation}
where $F_{in}$ denotes the input features of the LCFI block and $F_{lcfi}$ denotes the integrated large-field contextual features. $conv_1$ and $conv_2$ are local convolutions with a $3 \times 3$ kernel.

\textbf{LCFI module.} Glass captured in an image can vary in size (Figure \ref{fig:statistic}(a)). Given the kernel size $k$ and dilation rate $r$, the LCFI block extracts contextual features from a large field of a fixed size. On the one hand, if this field is not large enough to cover the whole glass region, incomplete detection may occur. On the other hand, if this field is too large for a small glass region, too much noise would be introduced and cause a false positive detection. To address this problem, contexts of different scales should be considered. Hence, based on the LCFI block, we propose a LCFI module to harvest contextual features from large fields of different scales. Specifically, we feed the input features into four parallel LCFI blocks and fuse their outputs using an attention module \cite{Woo_2018_ECCV}. To further explore more contextual features, we add information flow between adjacent LCFI blocks, \emph{i.e.}, we feed the output of a current LCFI block to the next LCFI block. By doing so, local features $F^i_{l}$ and large-field contextual features from the previous block $F^{i-1}_{lcfi}$ are combined and further processed by the current LCFI block. For the spatially separable convolutions used in the four LCFI blocks, the kernel size $k$ is set to 3, 5, 7, 9, and the dilation rate $dr$ is set to 1, 2, 3, 4, respectively.

Although we draw inspiration (\emph{i.e.}, adding information flow between different paths/blocks) from the integrated successive dilation (ISD) module in \cite{Su_2019_ICCV} in our module design, the proposed LCFI module is inherently different from ISD in both motivation and implementation. The ISD module aims to extract invariant features for the salient object embedded in various contexts while our LCFI module is designed to locate glass of different sizes by exploring contextual information from a large field of different scales. Besides, ISD uses $3 \times 3$ convolutions with a large dilation rate (\emph{e.g.}, $r$=16) to capture large-field contexts. We argue that the contexts extracted in this way are insufficient for complete glass detection (see ``Base + LCFI w/ local'' in Figure \ref{fig:visual_lcfi}). Instead, in each LCFI block, we leverage spatially separable convolutions to explore abundant contextual cues from a large field (see ``Base + LCFI'' in Figure \ref{fig:visual_lcfi}).

\subsection{\mhy{Boundary Feature Enhancement}}
\mhy{Since the boundary is a strong cue for humans to distinguish a glass region, we further introduce the boundary cue to help boost the glass detection accuracy. Integrating boundary information can help glass detection in both segmentation and localization. Based on this idea, we propose a boundary feature enhancement (BFE) module to learn and integrate glass boundary features.}

\mhy{Figure \ref{fig:BFE} illustrates the structure of the proposed BFE module. Given the input features $F_{in}$, we first feed them into a multi-scale boundary feature extractor, which contains four parallel branches with different receptive fields (1, 3, 5, 9) and each branch consists of a convolution layer, a batch normalization (BN) layer and a ReLU nonlinearity operation. We then concatenate the outputs of all four branches and fuse them via a $1\times 1$ convolution followed by BN and ReLU, to obtain the glass boundary features $F_{b}$. Finally, we extract the boundary-enhanced features $F_{e}$ by aggregating the boundary features $F_{b}$ and the input features $F_{in}$ as:
	\begin{equation}\label{equ:bfe}
		F_{e} = F_b \otimes F_{in} \oplus F_{in},
	\end{equation}
where $\otimes$ denotes element-wise multiplication and $\oplus$ denotes element-wise addition. We employ a $3\times 3$ convolution on the glass boundary features $F_{b}$ to predict the glass boundary map, which is supervised by the glass boundary ground truth generated by computing the gradients on the glass mask ground truth, to encourage $F_b$ to obtain discriminative boundary features. Benefiting from the rich location information and boundary information in the glass boundary features $F_{b}$, the boundary-enhanced features $F_{e}$ can help the network locate glass regions, especially their boundaries more accurately.
As shown in Figure \ref{fig:GDNet}, we apply the BFE module on both the high-level features and attentive low-level features to harvest both high-level and low-level boundary-enhanced features, leading to a boundary-aware model.}

\add{Note that our method as well as \cite{xie2020segmenting,Su_2019_ICCV_banet} explore the boundary cues for the specific task. Our BFE module differs from the method in \cite{xie2020segmenting} in both the generation and use of boundary features. \cite{xie2020segmenting} uses an additional boundary stream (\textit{i.e.}, an encoder-decoder branch) to obtain the boundary features and then fuse these boundary features with different levels of main stream features. Our BFE module can generate the boundary features from the given input features that needs to be enhanced. As different levels of features typically require different levels of boundary features (referred to as  ``HL'' and ``LL'' maps in Figure \ref{fig:visual_bfe}, for an example), our BFE would be more effective and flexible. \cite{Su_2019_ICCV_banet} only conducts the boundary feature enhancement on the lowest level of features while we apply our BFE module on both the high-level and low-level features to harvest both high-level and low-level boundary-enhanced features, leading to an omni-boundary-aware model.}

\subsection{Loss Function}
For optimizing GDNet-B, we adopt three types of losses, \emph{i.e.}, binary cross-entropy (BCE) loss $\ell_{bce}$ \cite{boer2005a}, IoU loss $\ell_{iou}$ \cite{qin2019basnet} and edge loss $\ell_{edge}$ \cite{Zhao_2019_CVPR}, during the training process.
Specifically, for high-level large-field contextual features, we combine BCE loss and IoU loss, \emph{i.e.}, $\mathcal{L}_{h} = \ell_{bce} + \ell_{iou}$, to force them to explore high-level cues for complete glass detection. For attentive low-level large-field contextual features, we want them to provide low-level cues for predicting glass maps with clear boundaries. Thus, we combine BCE loss and edge loss, \emph{i.e.}, $\mathcal{L}_{l} = \ell_{bce} + \ell_{edge}$. The edge loss would implicitly help the network find glass boundaries. For the final output, the complete detection with clear glass boundary is desired. Hence, we combine BCE loss, IoU loss and edge loss, \emph{i.e.}, $\mathcal{L}_f  = \ell_{bce} + \ell_{iou} + \ell_{edge}$. We further use BCE loss to supervise the glass boundary map predicted by the BFE module. Finally, the overall loss function is:
\begin{equation}\label{gdnet_loss}
	\mathcal{L}_{overall} = w_h\mathcal{L}_h + w_l\mathcal{L}_l + w_{f}\mathcal{L}_{f} + w_h^b\mathcal{L}_h^b + w_l^b\mathcal{L}_l^b,
\end{equation}
where $w_h$, $w_l$, and $w_f$ represent the balancing parameters for $\mathcal{L}_h$, $\mathcal{L}_l$, and $\mathcal{L}_f$, respectively. $\mathcal{L}_h^b$ and $\mathcal{L}_l^b$ denote the losses between the glass boundary ground truth and the glass boundary prediction of the BFE module applied on the high-level features and attentive low-level features, respectively. And $w_h^b$ and $w_l^b$ represent the balancing parameters for $\mathcal{L}_h^b$ and $\mathcal{L}_l^b$.



\section{Experiments on Glass Detection}
\subsection{Experimental Settings}
\textbf{Implementation details.}
We have implemented GDNet-B on the PyTorch framework \cite{paszke2019pytorch}. \mhy{A PC with an Intel eight-core i7-7700K 4.2 GHz CPU, 32GB RAM and an NVIDIA GTX 1080Ti GPU (with 11GB memory) is used for both training and testing.}

For training, input images are resized to a resolution of $416\times416$ and are augmented by horizontally random flipping. The parameters of the multi-level feature extractor are initialized by the pre-trained ResNeXt-101 network \cite{Xie_2017_CVPR} and the other parameters are initialized randomly. Stochastic gradient descent (SGD) with a momentum of $0.9$ and weight decay of $5\times10^{-4}$ is used to optimize the network. We adjust the learning rate by the poly strategy \cite{liu2015parsenet} with a basic learning rate of $0.001$ and a power of $0.9$. The batch size is set to $6$ and the balancing parameters $w_h$, $w_l$, $w_f$, $w_h^b$, and $w_l^b$ are empirically set to $1$, $1$, $1$, $10$, and $50$, respectively. \mhy{Training GDNet-B takes about $22$ hours for $190$ epochs.}

\mhy{For testing, the input image is first resized to the resolution of $416\times416$ for network inference, and the output mask is then resized back to the original size of the input image. Both resizing processes use bilinear interpolation.}
\mhy{There is no post-processing (such as the fully connected conditional random field (CRF) \cite{krahenbuhl2011efficient}) needed to further enhance the final output. The average inference time of each image in GDD testing set is 0.034 seconds.}

\textbf{Evaluation metrics.}
For a comprehensive evaluation, we adopt five widely used metrics for quantitatively evaluating the glass detection performance: intersection over union ($IoU$), pixel accuracy ($PA$), maximum F-measure ($F_\beta^m$), mean absolute error ($M$), and balance error rate ($BER$).

The intersection of union ($IoU$) and pixel accuracy ($PA$) are widely used in the semantic segmentation field, which are defined as:
\mhy{
	\begin{equation}\label{equ:iou}
		IoU = \frac{\sum\limits_{i=1}^{H}\sum\limits_{j=1}^{W}(G(i,j)*P_b(i,j))}{\sum\limits_{i=1}^{H}\sum\limits_{j=1}^{W}(G(i,j)+P_b(i,j)-G(i,j)*P_b(i,j))},
\end{equation}}
\mhy{
	\begin{equation}\label{equ:pa}
		PA = \frac{1}{H\times W}\sum\limits_{i=1}^{H}\sum\limits_{j=1}^{W}\textbf{1}(G(i,j)=P_b(i,j)),
\end{equation}}
\mhy{where $G$ and $P_b$ are the ground truth mask and the predicted mask binarized with a threshold of 0.5, respectively. $H$ and $W$ are the height and width of the predicted mask. $\textbf{1}(\cdot)$ is the indicator function.}

We also adopt the F-measure and mean absolute error metrics from the salient object detection field. F-measure is a harmonic mean of precision and recall, defined as:
\begin{equation}\label{equ:f}
	F_\beta=\frac{(1+\beta ^2)Precision \times Recall}{\beta ^2Precision + Recall},
\end{equation}
where $\beta^2$ is set to be $0.3$ to emphasize more on precision over recall, as suggested in \cite{cheng2014global}. \mhy{The precision and recall of a detection map are computed by comparing the binarized predicted mask against the ground truth mask. 
	\mhy{Under a specific binarizing threshold, we can obtain a pair of average precision and recall over all binarized saliency maps in a dataset.} By varying the threshold from 0 to 1 with a step of $\frac{1}{255}$, we obtain $256$ precision-recall pairs and 256 $F_\beta$ scores. We report the maximum $F_\beta$ ($F_\beta^m$) in the comparison of glass detection performance.}

\mhy{Given the predicted mask $P$ and the ground truth mask $G$, the mean absolute error ($M$) can be calculated by the element-wise difference between $P$ and $G$,
	\begin{equation}\label{equ:mae}
		M = \frac{1}{H\times W}\sum_{i=1}^{H}\sum_{j=1}^{W}|P(i,j)-G(i,j)|,
	\end{equation}
	where $P(i,j)$ indicates the probability score at location $(i,j)$ in the predicted mask and $H$ and $W$ are the map's height and width, respectively.}

The last metric is the balance error rate ($BER$), which is a standard metric in the shadow detection field, defined as:
\begin{equation}\label{equ:ber}
	BER = (1 - \frac{1}{2}(\frac{TP}{N_{p}} + \frac{TN}{N_{n}})) \times 100,
\end{equation}
where $TP$, $TN$, $N_p$, and $N_n$ represent the numbers of true positive pixels, true negative pixels, glass pixels, and non-glass pixels, respectively.

Note that for $IoU$, $PA$ and $F_\beta^m$, it is the higher the better, while for $M$ and $BER$, it is the lower the better.

\subsection{Comparison with the State-of-the-arts}
\textbf{Compared methods.} As a first attempt to detect glass from a single RGB image, we validate the effectiveness of our GDNet by comparing it with 18 methods selected from other related fields \mhy{according to the following criteria: (\romannumeral1) classical architectures, (\romannumeral2) recently published, and (\romannumeral3) achieving state-of-the-art performance in a specific field.} Specifically, we choose ICNet \cite{Zhao_2018_ECCV_icnet}, PSPNet \cite{zhao2017pyramid}, DenseASPP \cite{Yang_2018_CVPR}, BiSeNet \cite{Yu_2018_ECCV}, PSANet \cite{Zhao_2018_ECCV_psanet}, DANet \cite{Fu_2019_CVPR} and CCNet \cite{Huang_2019_ICCV} from the semantic segmentation field; DSS \cite{HouPami19Dss}, PiCANet \cite{liu2018picanet}, RAS \cite{Chen_2018_ECCV}, R\textsuperscript{3}Net \cite{deng2018r3net}, CPD \cite{Wu_2019_CVPR}, PoolNet \cite{Liu_2019_CVPR}, BASNet \cite{qin2019basnet} and EGNet \cite{zhao2019EGNet} from the salient object detection field; DSC \cite{Hu_2018_CVPR} and BDRAR \cite{Zhu_2018_ECCV} from the shadow detection field; MirrorNet \cite{yang2019mirrornet} from the mirror segmentation field; \add{and TransLab \cite{xie2020segmenting}, Trans2Seg \cite{xie2021segmenting} and GDNet \cite{Mei_2020_CVPR} from the glass segmentation field.} For a fair comparison, we use either their publicly available codes or the implementations with recommended parameter settings. All models are retrained on the GDD training set.

\begin{table}[tbp]
	\centering
	\caption{Quantitative comparison to the state-of-the-art methods on the GDD testing set. All methods are re-trained on the GDD training set. * denotes using CRFs~\cite{krahenbuhl2011efficient} for post-processing. ``Statistics" means thresholding glass location statistics from our training set as a glass mask for detection. The first, second, and third best results are marked in \textcolor[rgb]{1,0,0}{\textbf{red}}, \textcolor[rgb]{0,1,0}{\textbf{green}}, and \textcolor[rgb]{0,0,1}{\textbf{blue}}, respectively. Our method achieves the state-of-the-art performances on all five evaluation metrics.}
	\setlength{\tabcolsep}{2.4pt}
	\centering
	\begin{tabular}{p{2.4cm}|p{1cm}<{\centering}|p{1cm}<{\centering}|p{1cm}<{\centering}|p{1cm}<{\centering}|p{.85cm}<{\centering}}
		\hline
		\hline
		Methods & IoU$\uparrow$ & PA$\uparrow$ & $F_\beta^m$$\uparrow$ & M$\downarrow$ & BER$\downarrow$ \\
		\hline
		\hline
		Statistics & 40.75 & 0.584 & 0.564 & 0.451 & 39.31 \\
		\hline
		\hline
		ICNet \cite{Zhao_2018_ECCV_icnet} & 69.59 & 0.836 & 0.821 & 0.164 & 16.10 \\
		PSPNet \cite{zhao2017pyramid} & 84.06 & 0.916 & 0.906 & 0.084 & 8.79 \\
		DenseASPP \cite{Yang_2018_CVPR} & 83.68 & 0.919 & 0.911 & 0.081 & 8.66 \\
		BiSeNet \cite{Yu_2018_ECCV} & 80.00 & 0.894 & 0.883 & 0.106 & 11.04 \\
		PSANet \cite{Zhao_2018_ECCV_psanet} & 83.52 & 0.918 & 0.909 & 0.082 & 9.09 \\
		DANet \cite{Fu_2019_CVPR} & 84.15 & 0.911 & 0.901 & 0.089 & 8.96 \\
		CCNet \cite{Huang_2019_ICCV} & 84.29 & 0.915 & 0.904 & 0.085 & 8.63 \\
		\hline
		\hline
		DSS \cite{HouPami19Dss} & 80.24 & 0.898 & 0.890 & 0.123 & 9.73 \\
		PiCANet \cite{liu2018picanet} & 83.73 & 0.916 & 0.909 & 0.093 & 8.26 \\
		RAS \cite{Chen_2018_ECCV} & 80.96 & 0.902& 0.895 & 0.106 & 9.48 \\
		R\textsuperscript{3}Net* \cite{deng2018r3net} & 76.71 & 0.869 & 0.869 & 0.132 & 13.85 \\
		CPD \cite{Wu_2019_CVPR} & 82.52 & 0.907 & 0.903 & 0.095 & 8.87 \\
		PoolNet \cite{Liu_2019_CVPR} & 81.92 & 0.907& 0.900 & 0.100 & 8.96 \\
		BASNet \cite{qin2019basnet} & 82.88 & 0.907 & 0.896 & 0.094 & 8.70 \\
		EGNet \cite{zhao2019EGNet} & 85.04 & 0.920 & \textcolor[rgb]{0,0,1}{\textbf{0.916}} & 0.083 & 7.43 \\
		\hline
		\hline
		DSC \cite{Hu_2018_CVPR} & 83.56 & 0.914 & 0.911 & 0.090 & 7.97 \\
		BDRAR* \cite{Zhu_2018_ECCV} & 80.01 & 0.902 & 0.908 & 0.098 & 9.87 \\
		\hline
		\hline
		MirrorNet* \cite{yang2019mirrornet} & \textcolor[rgb]{0,0,1}{\textbf{85.07}} & 0.918 & 0.903 & 0.083 & 7.67 \\
		\hline
		\hline
		\add{TransLab} \cite{xie2020segmenting} & 81.64 & 0.903 & 0.892 & 0.097 & 9.70 \\
		\add{Trans2Seg} \cite{xie2021segmenting} & 84.41 & \textcolor[rgb]{0,0,1}{\textbf{0.922}} & 0.905 & \textcolor[rgb]{0,0,1}{\textbf{0.078}} & \textcolor[rgb]{0,0,1}{\textbf{7.36}} \\
		GDNet \cite{Mei_2020_CVPR} & \textcolor[rgb]{0,1,0}{\textbf{87.63}} & \textcolor[rgb]{0,1,0}{\textbf{0.939}} & \textcolor[rgb]{0,1,0}{\textbf{0.937}} & \textcolor[rgb]{0,1,0}{\textbf{0.063}} & \textcolor[rgb]{0,1,0}{\textbf{5.62}} \\
		\hline
		\hline
		\textbf{GDNet-B (ours)} & \textcolor[rgb]{1,0,0}{\textbf{87.83}} & \textcolor[rgb]{1,0,0}{\textbf{0.941}} & \textcolor[rgb]{1,0,0}{\textbf{0.939}} & \textcolor[rgb]{1,0,0}{\textbf{0.061}} & \textcolor[rgb]{1,0,0}{\textbf{5.52}} \\
		\hline
		\hline
	\end{tabular}
	\label{tab:comparison}
\end{table}

\textbf{Evaluation on the GDD testing set.} Table \ref{tab:comparison} reports the quantitative results of glass detection on the proposed GDD testing set. We can see that our method outperforms all the other state-of-the-art methods on all five metrics.
Figure \ref{fig:visual} shows the qualitative comparison of our method with the state-of-the-art methods. We can see that our method is capable of accurately detecting small glass regions (\textit{e.g.}, the 1\textit{st} row), large glass regions (\textit{e.g.}, the 2\textit{nd} row), and multiple glass regions (\textit{e.g.}, the 3\textit{rd} row). This is mainly because multi-scale contextual features extracted by the LCFI module can help the network locate and segment glass better. While the state-of-the-arts are typically confused by the non-glass regions which share similar boundaries/appearances with the glass regions, our method can successfully eliminate such ambiguities and detect only the real glass regions (\textit{e.g.}, 4-6\textit{th} rows). This is mainly contributed by the proposed large-field contextual feature learning, which provides abundant contextual information for context inference and glass localization. Benefited by the glass boundary feature enhancement, our GDNet-B achieve better glass detection results than GDNet \cite{Mei_2020_CVPR} (\textit{e.g.}, the last four rows).

\begin{table}[tbp]
	\centering
	\caption{Ablation studies on the effectiveness of the LCFI module. ``Base'' denotes GDNet \cite{Mei_2020_CVPR} with all LCFI modules removed. ``one scale'' and ``two scales'' denote that there are one and two LCFI blocks in the LCFI module. ``local'' denotes replacing the spatially separable convolutions in LCFI with local convolutions and keeping the parameters approximately the same. Based on ``local'', ``sparse'' adopts dilated convolutions to achieve a similar receptive field as spatially separable convolutions. ``one path'' denotes that there is only one spatially separable convolution path in each LCFI block. Our LCFI module contains four LCFI blocks and each of them contains two parallel paths.}
	\begin{tabular}{p{4cm}|p{1cm}<{\centering}p{0.9cm}<{\centering}p{0.8cm}<{\centering}}
		\hline
		\hline
		Networks & IoU$\uparrow$ & $F_\beta^m$$\uparrow$ & BER$\downarrow$ \\
		\hline
		\hline
		Base  & 84.89 & 0.923 & 7.40 \\
		\hline
		Base + LCFI w/ one scale   & 86.22 & 0.931 & 6.51 \\
		Base + LCFI w/ two scales   & 86.78 & 0.932 & 6.34 \\
		\hline
		Base + LCFI w/ local    & 86.93 & 0.932 & 6.36 \\
		Base + LCFI w/ sparse   & 87.13 & 0.933 & 5.88 \\
		\hline
		Base + LCFI w/ one path & 87.31 & 0.935 & 5.81\\
		\hline
		Base + LCFI (\textit{i.e.}, GDNet \cite{Mei_2020_CVPR}) & \textbf{87.63} & \textbf{0.937} & \textbf{5.62} \\
		\hline
		\hline
	\end{tabular}
	\label{tab:ablation_lcfi}
\end{table}

\def\winternal{0.19\linewidth}
\def\hinternal{.5in}
\begin{figure}[tbp]
	\setlength{\tabcolsep}{1.2pt}
	\centering
	\begin{tabular}{ccccc}
		\includegraphics[width=\winternal, height=\hinternal]{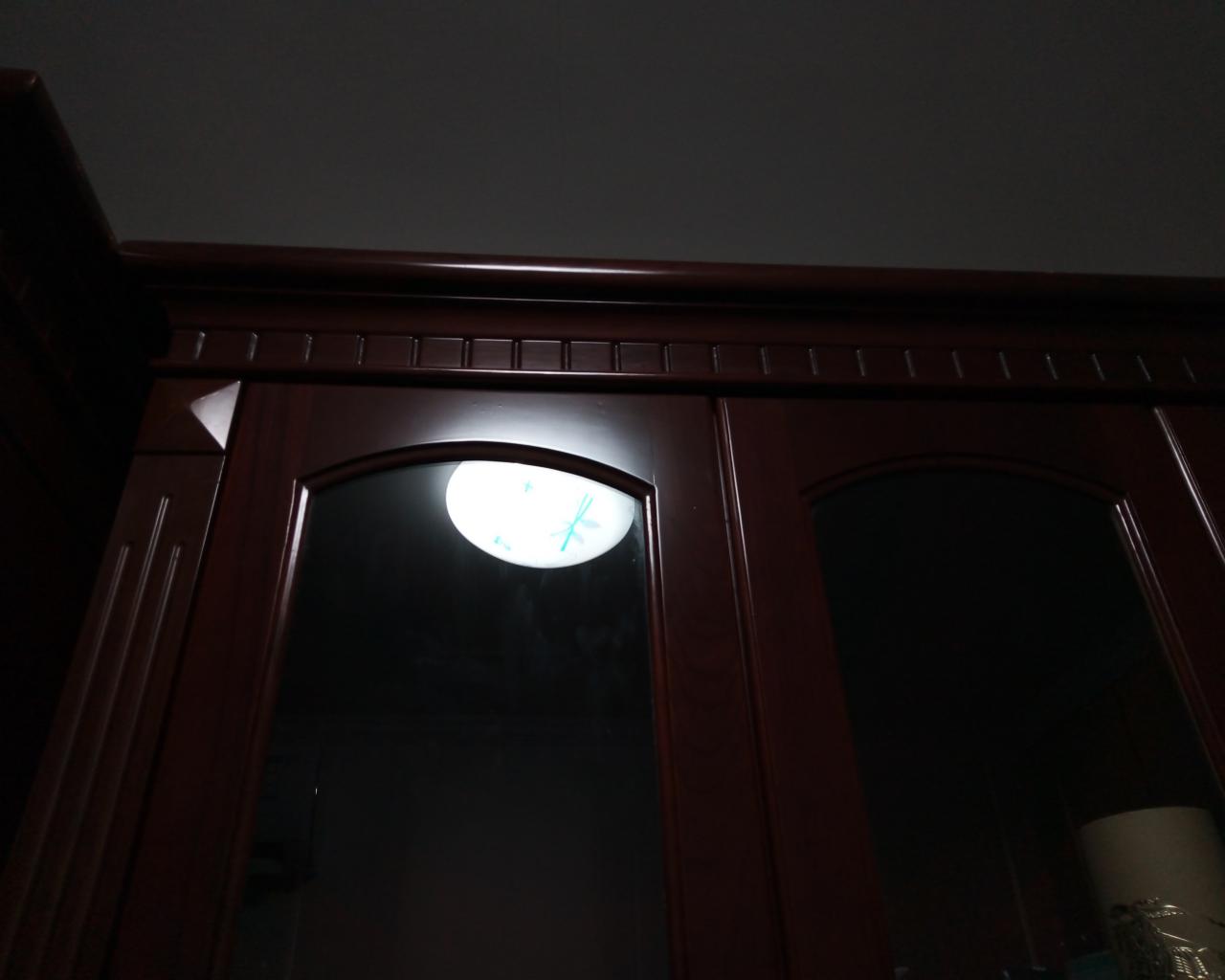}&
		\includegraphics[width=\winternal, height=\hinternal]{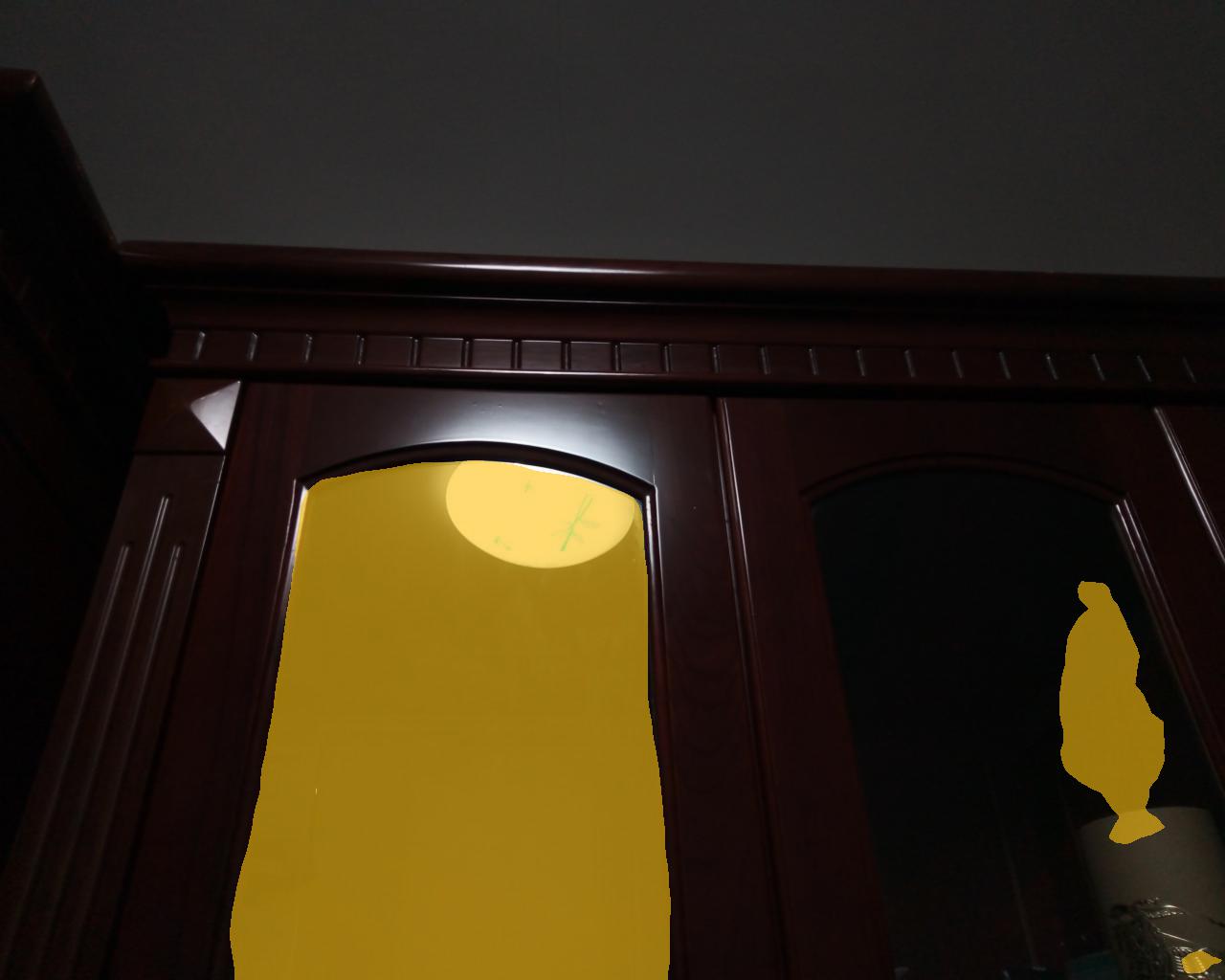}&
		\includegraphics[width=\winternal, height=\hinternal]{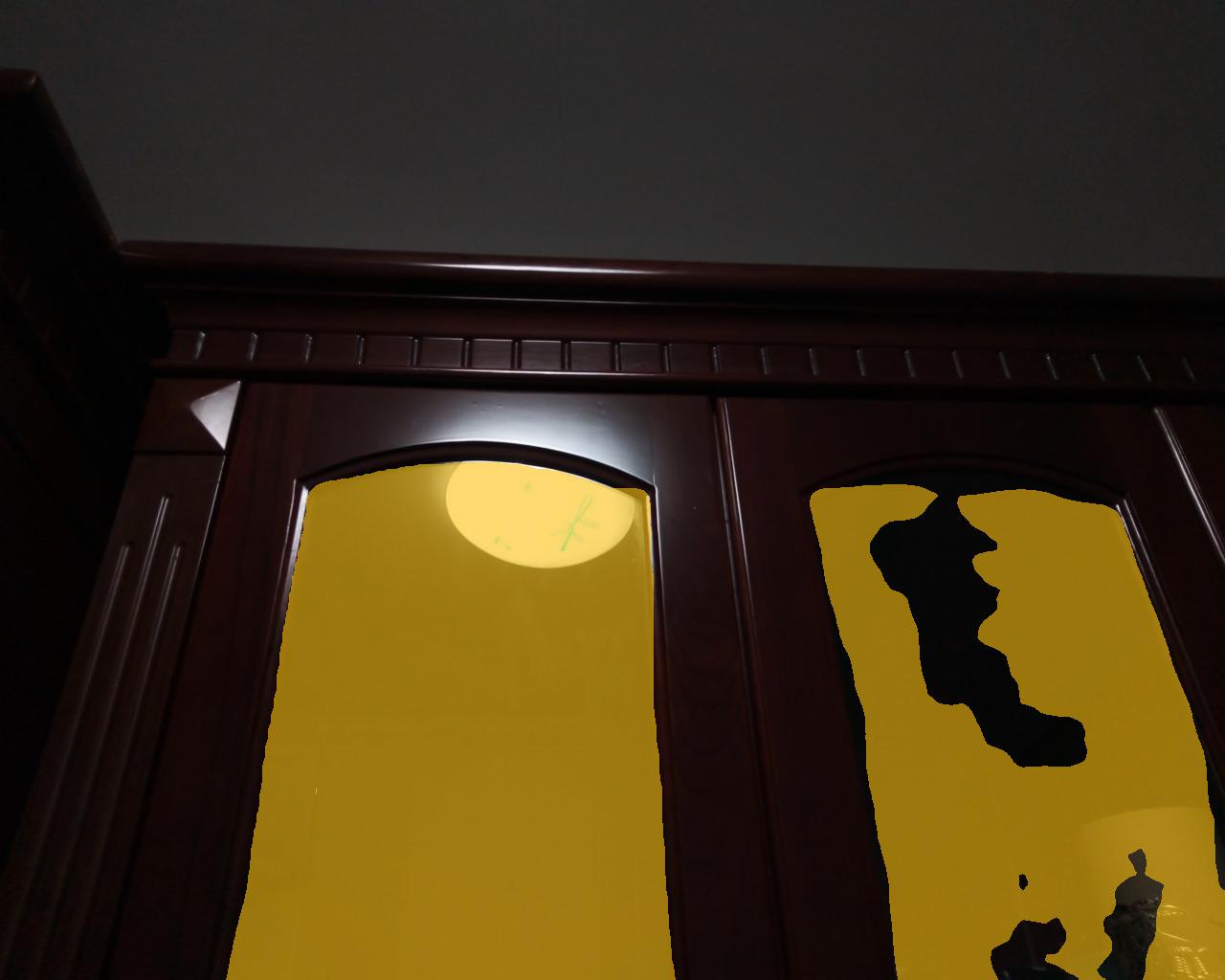}&
		\includegraphics[width=\winternal, height=\hinternal]{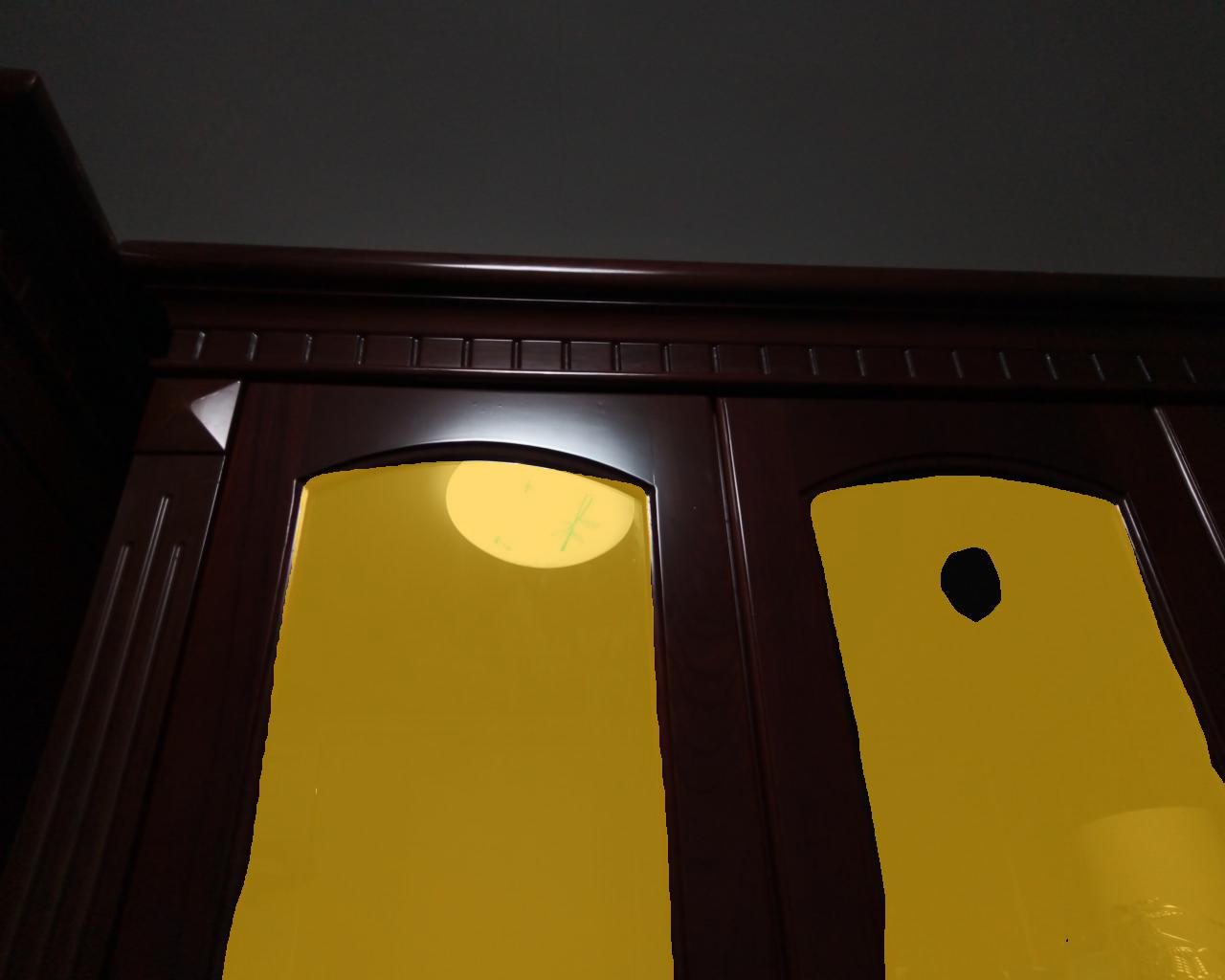}&
		\includegraphics[width=\winternal, height=\hinternal]{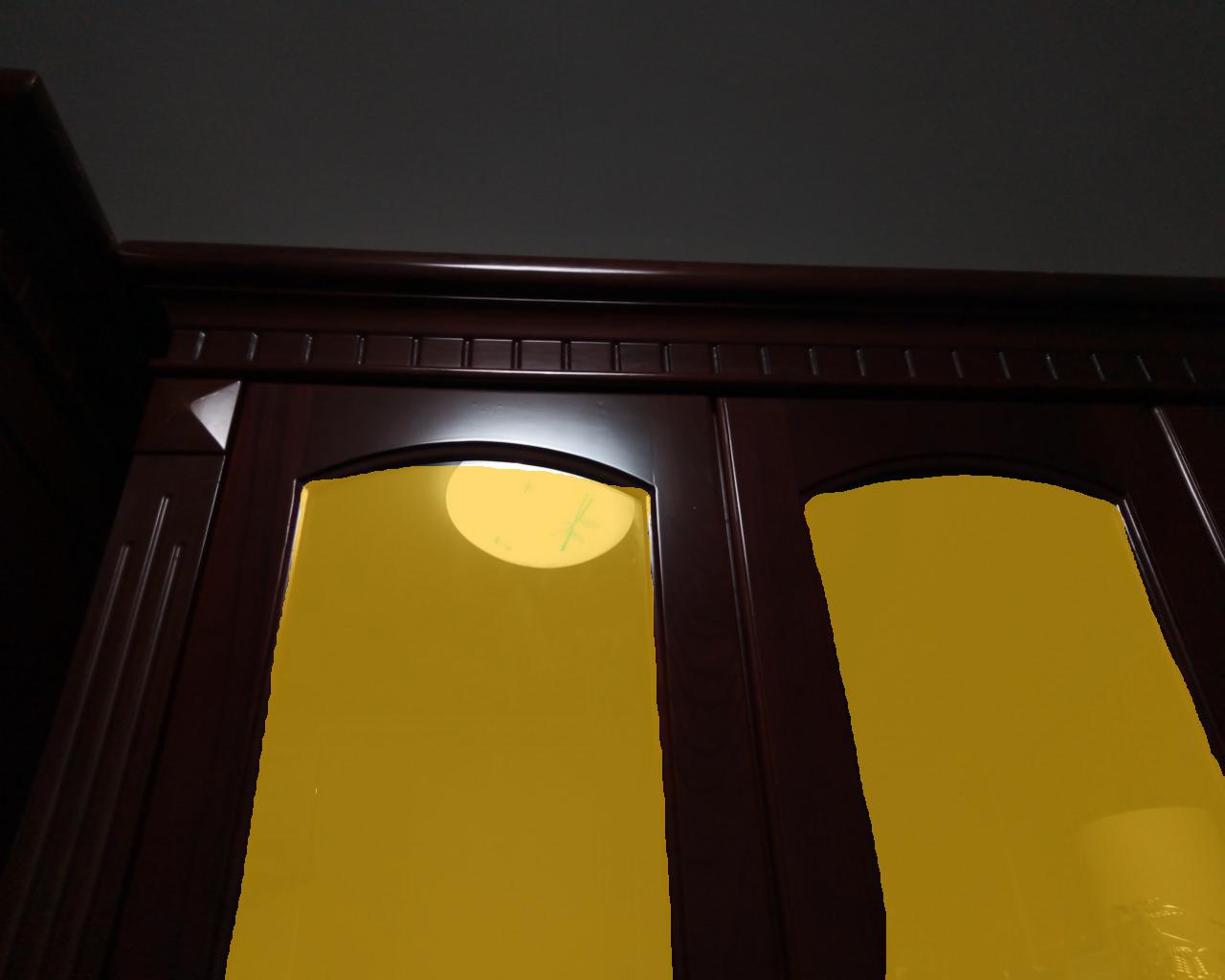} \\
		
		\includegraphics[width=\winternal, height=\hinternal]{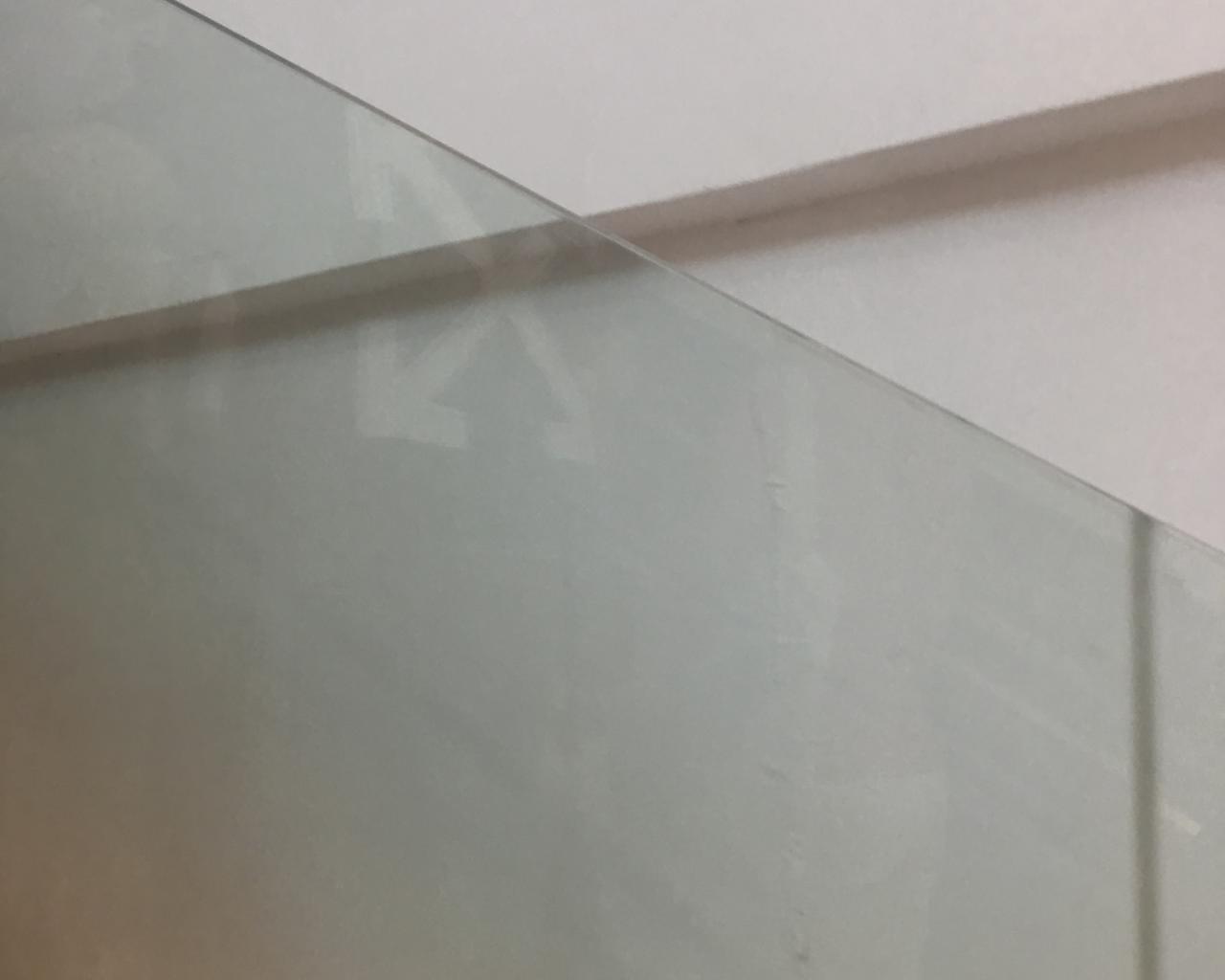}&
		\includegraphics[width=\winternal, height=\hinternal]{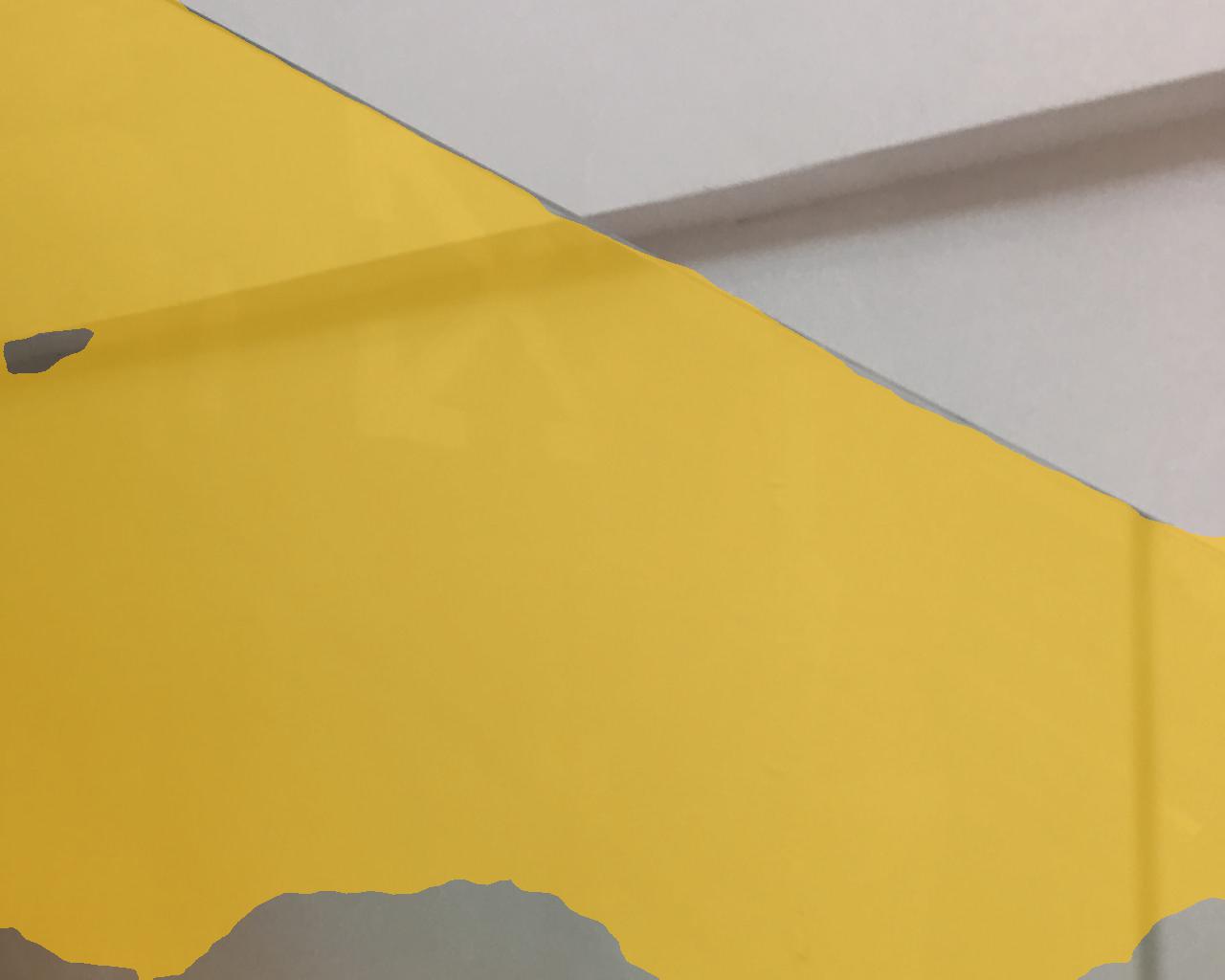}&
		\includegraphics[width=\winternal, height=\hinternal]{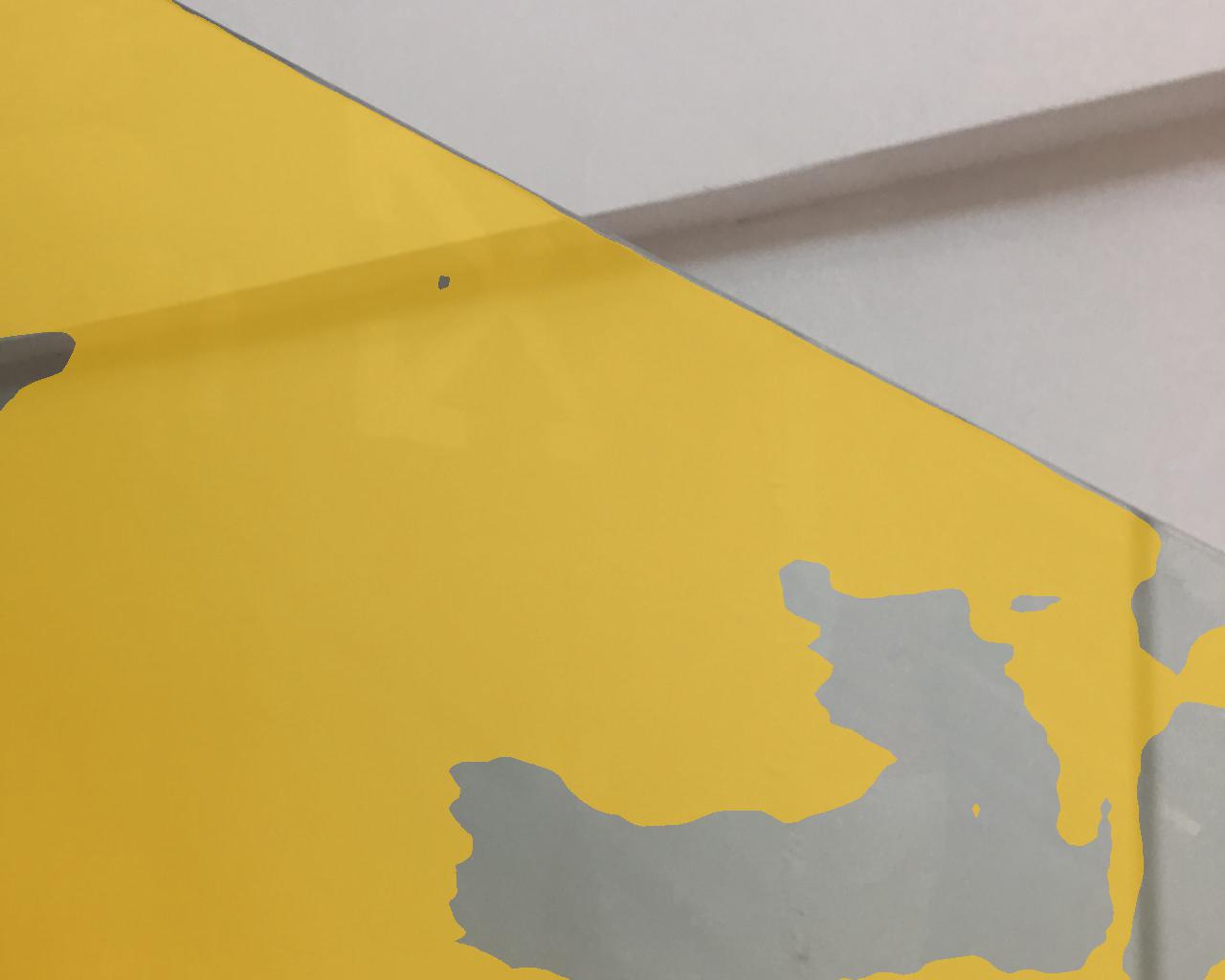}&
		\includegraphics[width=\winternal, height=\hinternal]{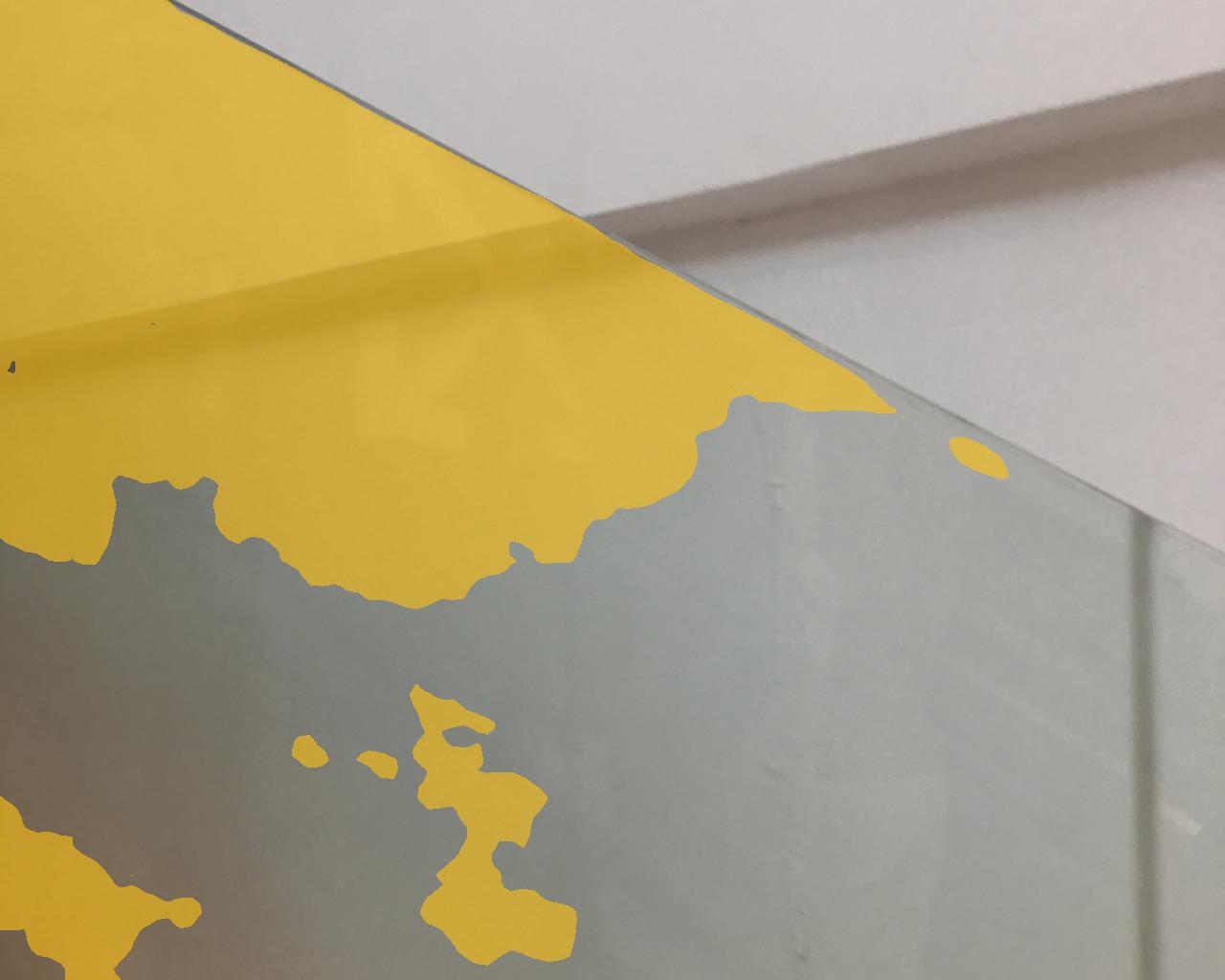}&
		\includegraphics[width=\winternal, height=\hinternal]{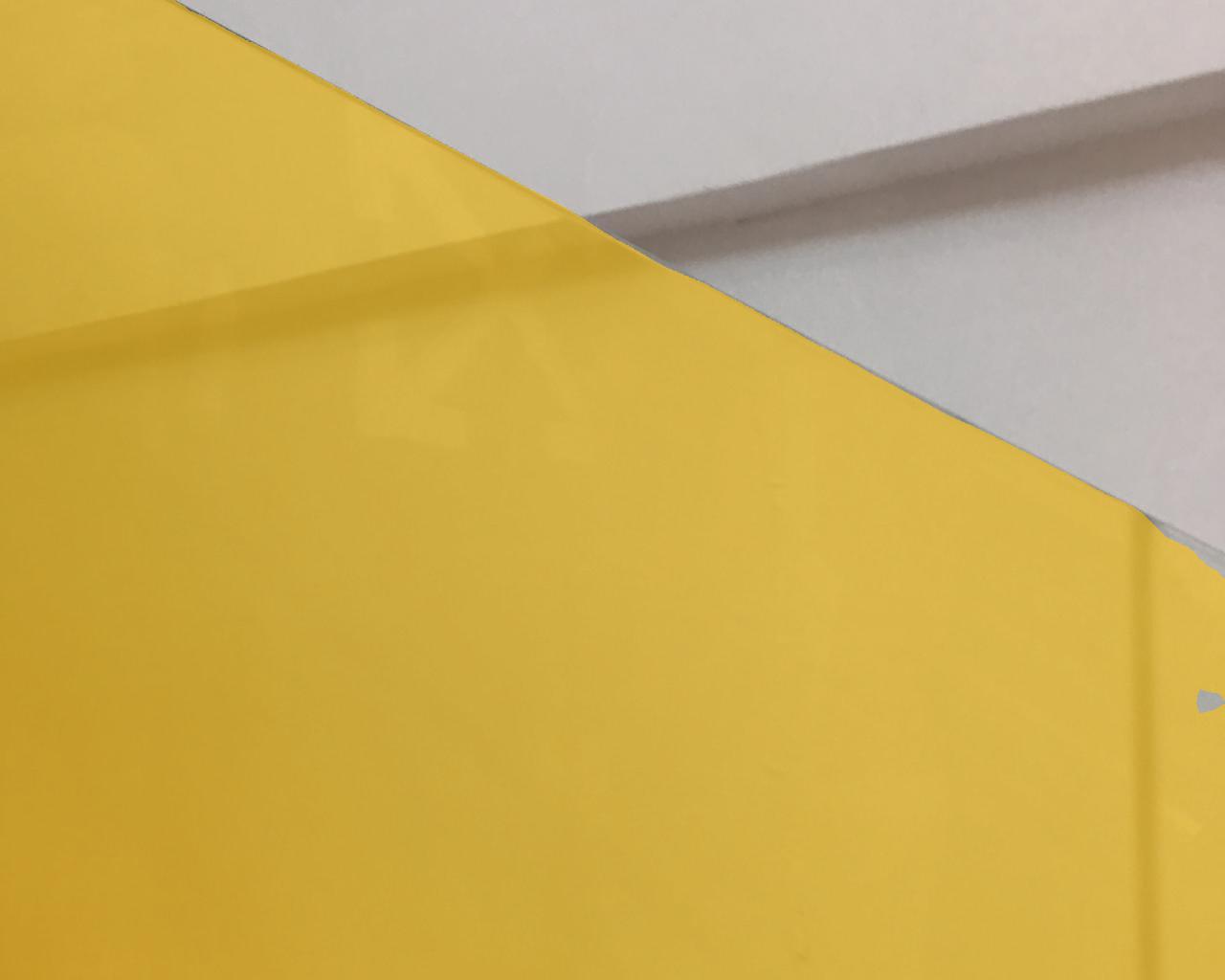} \\
		
		\footnotesize{Image} & \footnotesize{Base} & \tabincell{c}{\footnotesize{Base + LCFI}\\\footnotesize{w/ local}} & \tabincell{c}{\footnotesize{Base + LCFI}\\\footnotesize{w/ sparse}} & \footnotesize{Base + LCFI} \\
		
	\end{tabular}
	\caption{Ablation analysis on the effectiveness of the LCFI module.}
	\label{fig:visual_lcfi}
\end{figure}

\subsection{Component Analysis}
\mhy{We conduct ablation studies \add{on GDD} to validate the effectiveness of two key components tailored for accurate glass detection (\emph{i.e.}, the LCFI module and the BFE module) \add{as well as the impact of different multi-level feature extractors}, and report the comparison results in Table \ref{tab:ablation_lcfi}, Table \ref{tab:ablation_bfe}, \add{Table \ref{tab:ablation_backbone}}, Figure \ref{fig:visual_lcfi}, and Figure \ref{fig:visual_bfe}.}

\def\wvisual{0.078\linewidth}
\def\hvisualshort{.47in}
\def\hvisualtall{.59in}
\begin{figure*}
	\setlength{\tabcolsep}{1.6pt}
	\centering
	\scriptsize
	\begin{tabular}{cccccccccccc}
		\includegraphics[width=\wvisual, height=\hvisualtall]{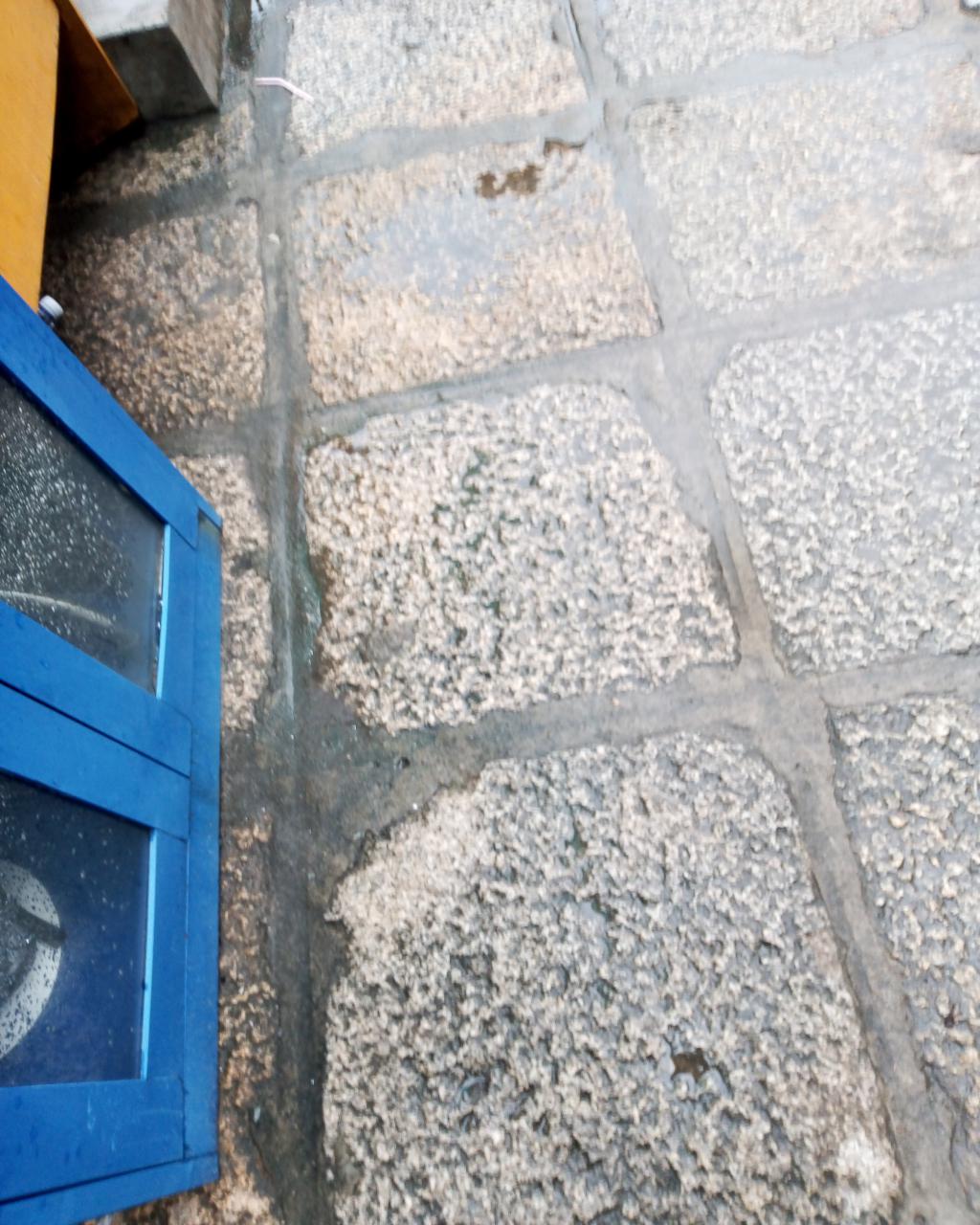}&
		\includegraphics[width=\wvisual, height=\hvisualtall]{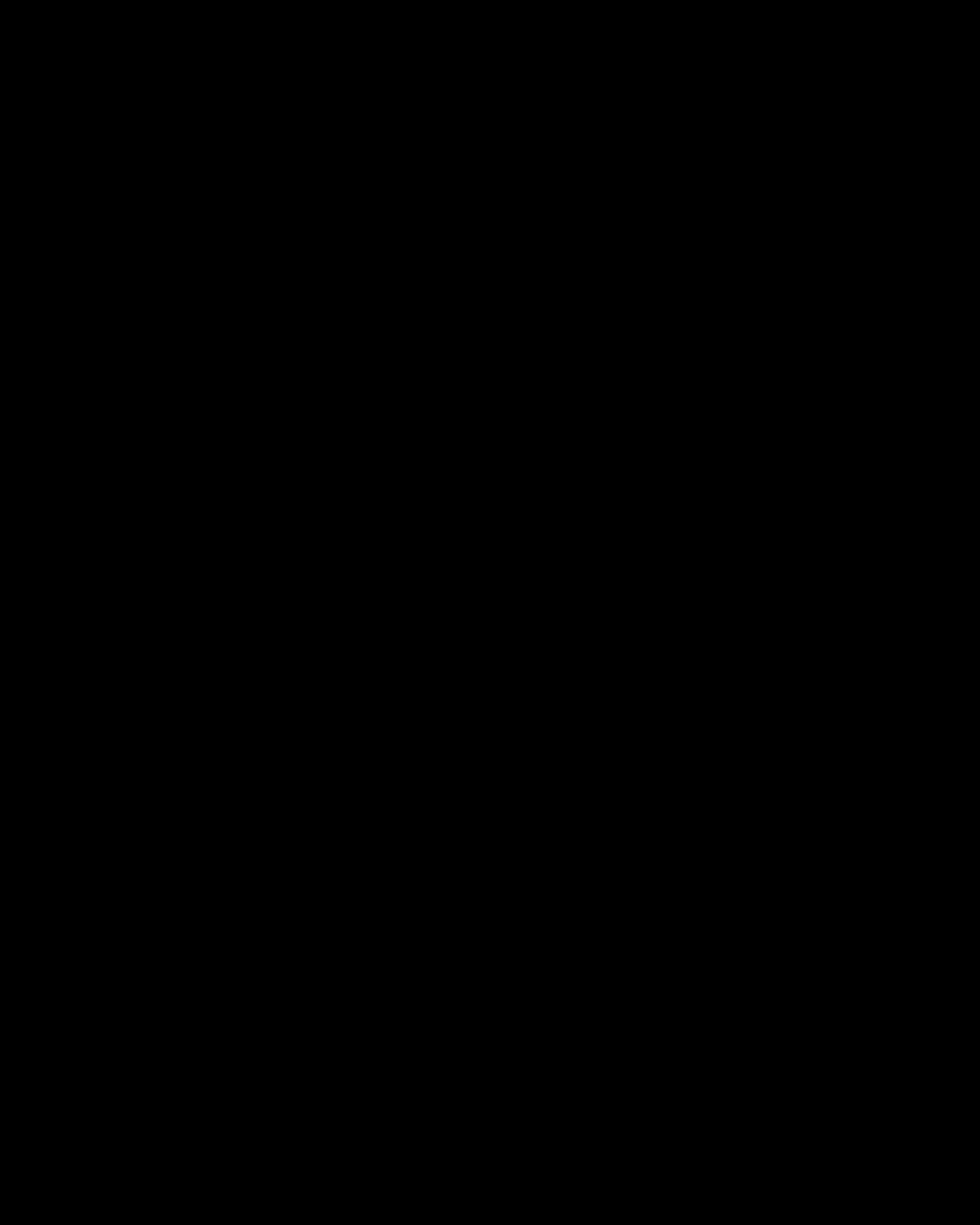}&
		\includegraphics[width=\wvisual, height=\hvisualtall]{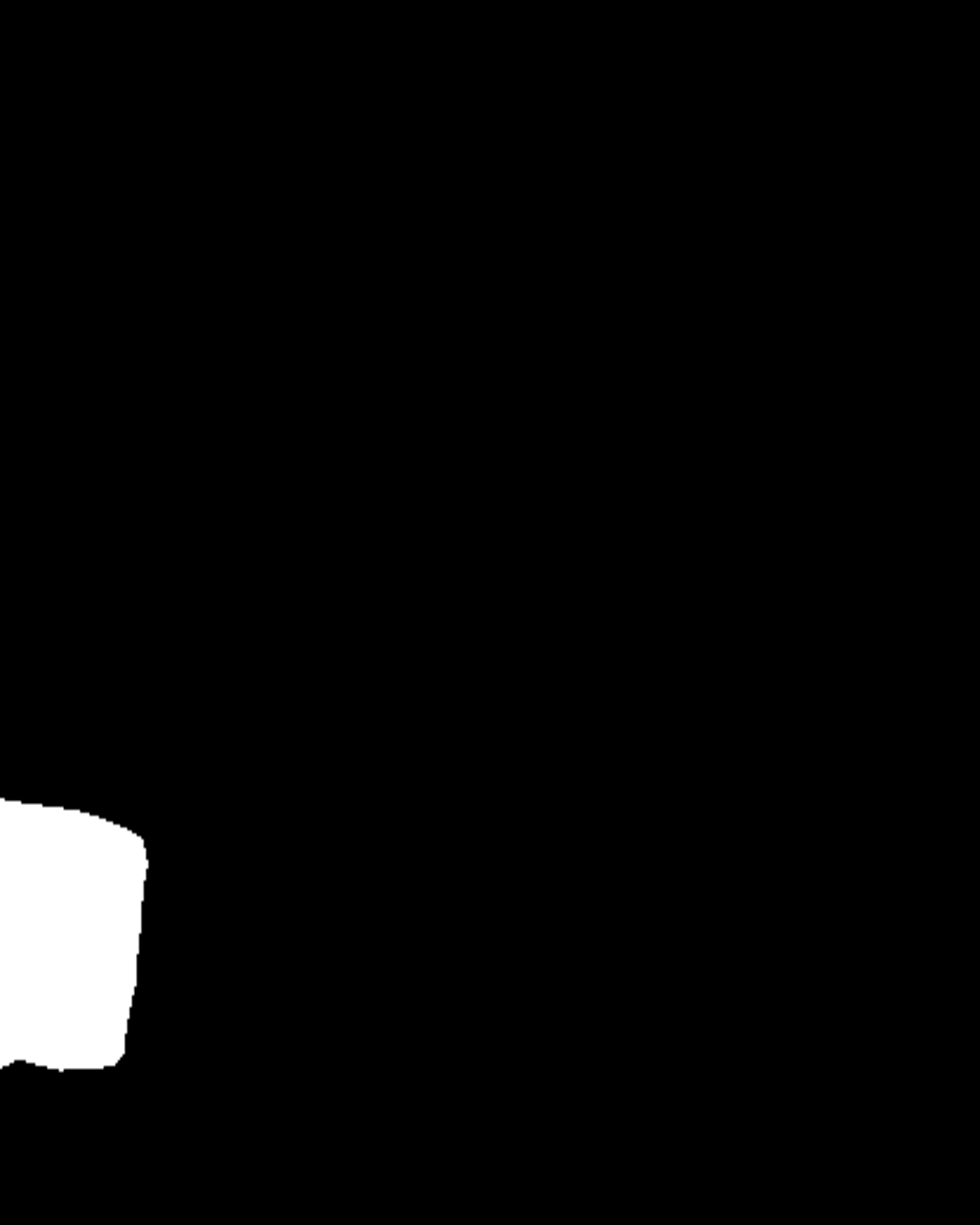}&
		\includegraphics[width=\wvisual, height=\hvisualtall]{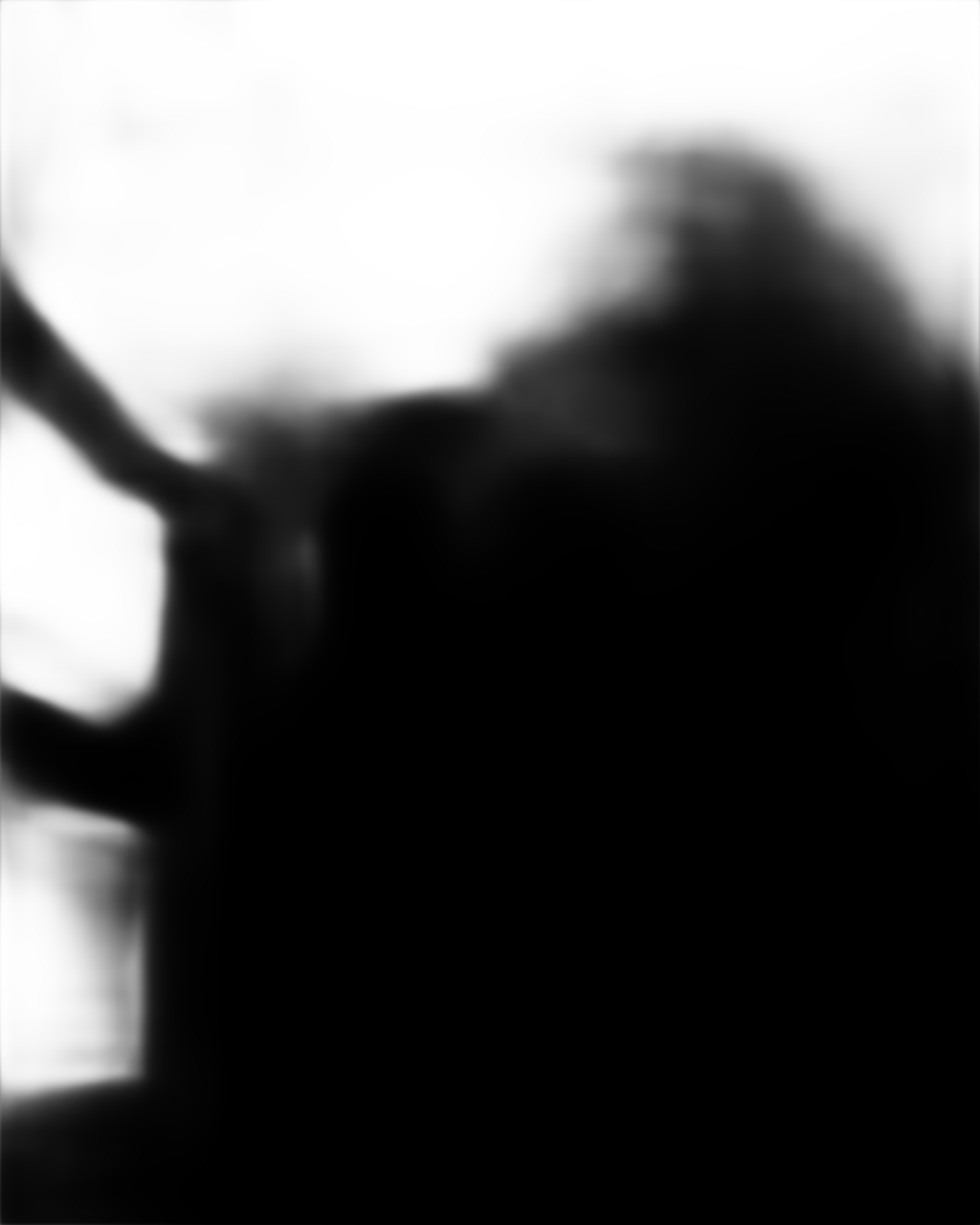}&
		\includegraphics[width=\wvisual, height=\hvisualtall]{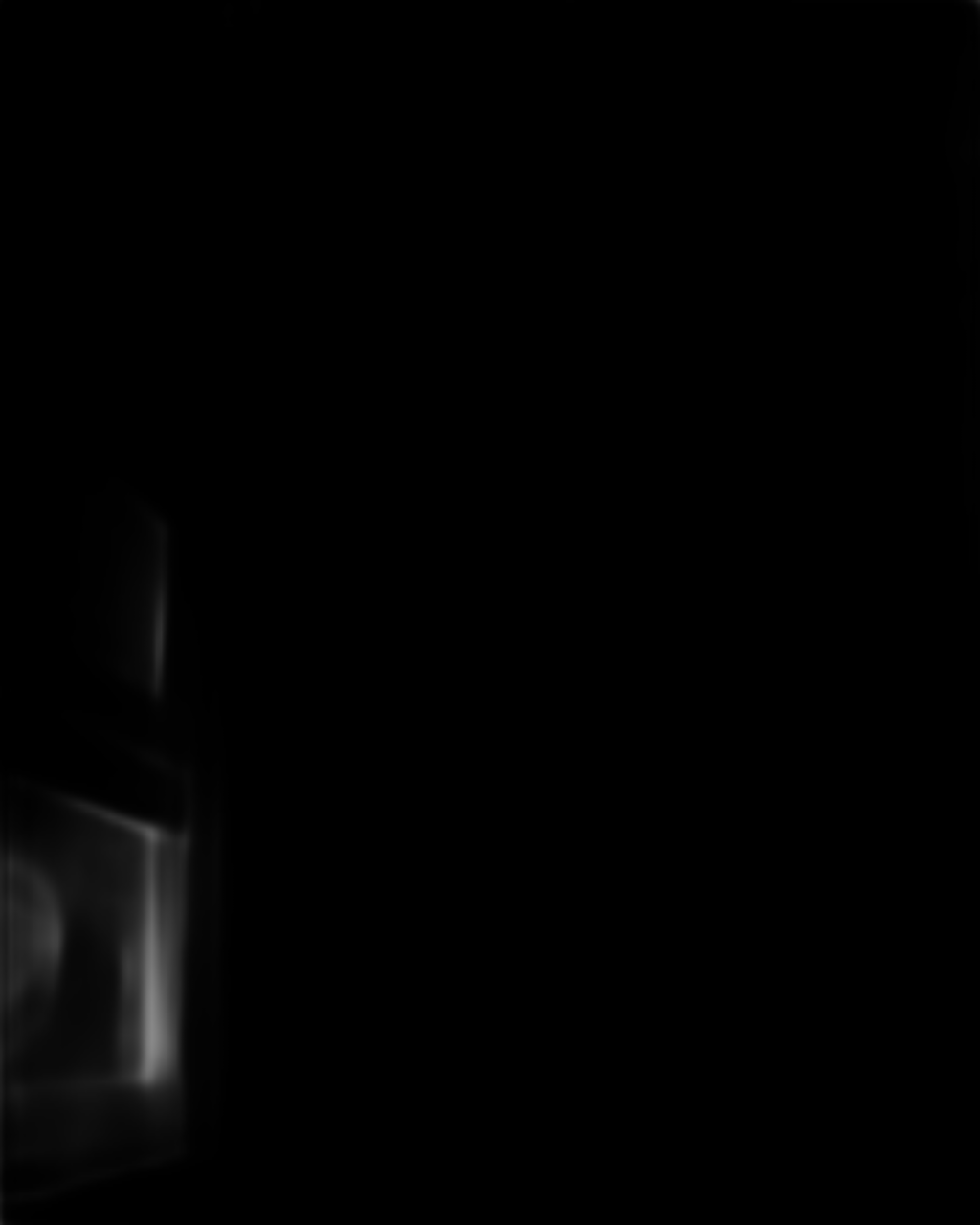}&
		\includegraphics[width=\wvisual, height=\hvisualtall]{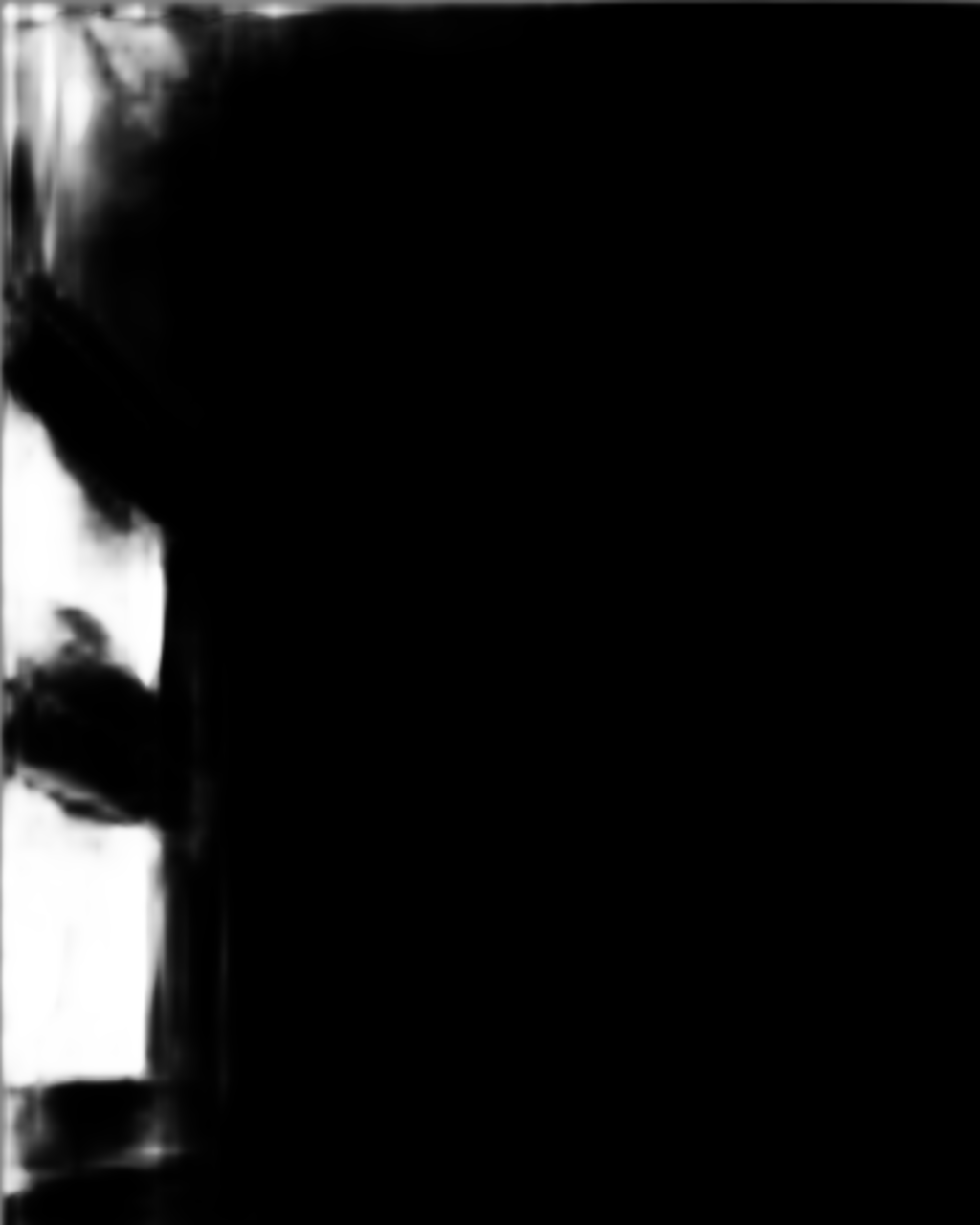}&
		\includegraphics[width=\wvisual, height=\hvisualtall]{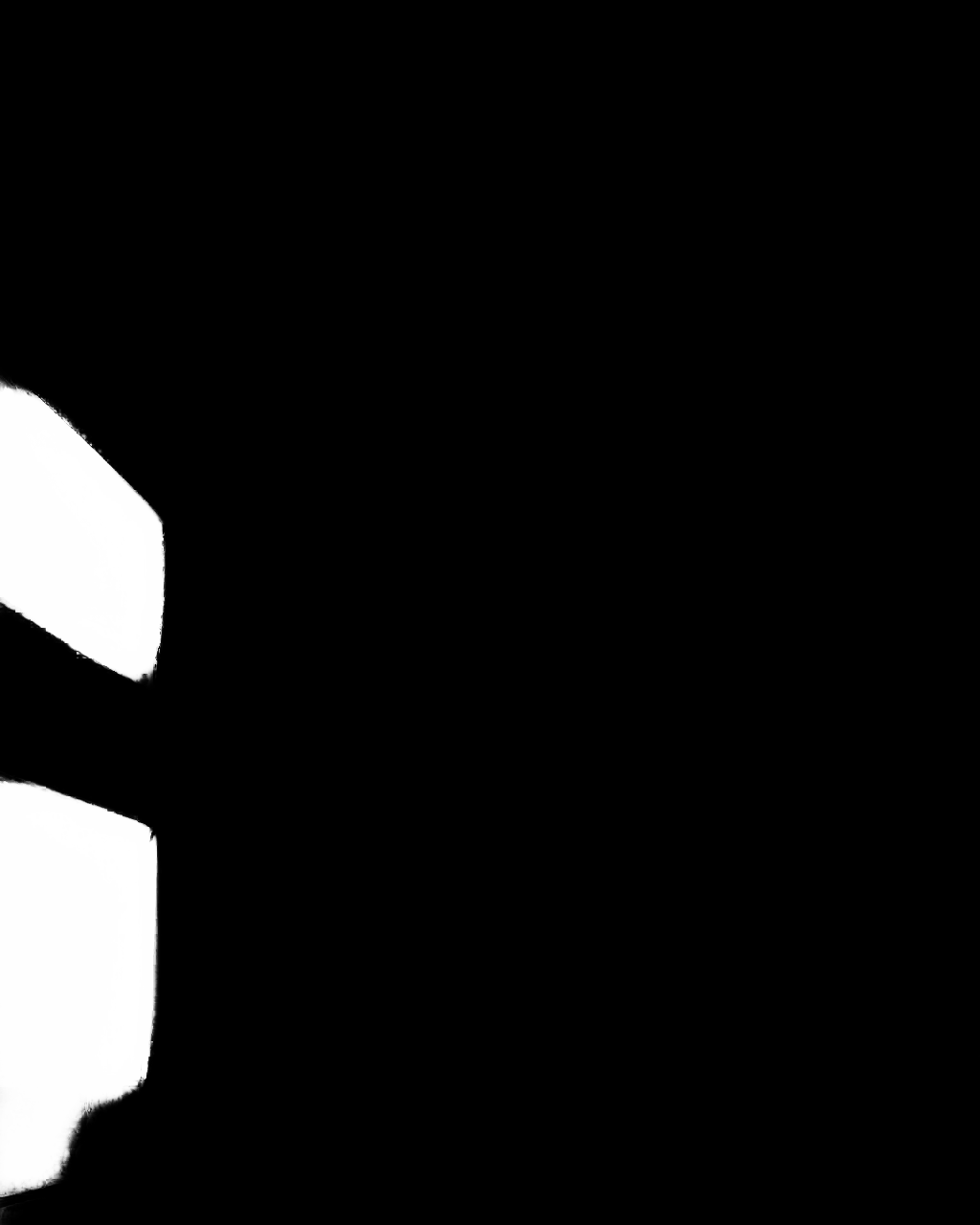}&
		\includegraphics[width=\wvisual, height=\hvisualtall]{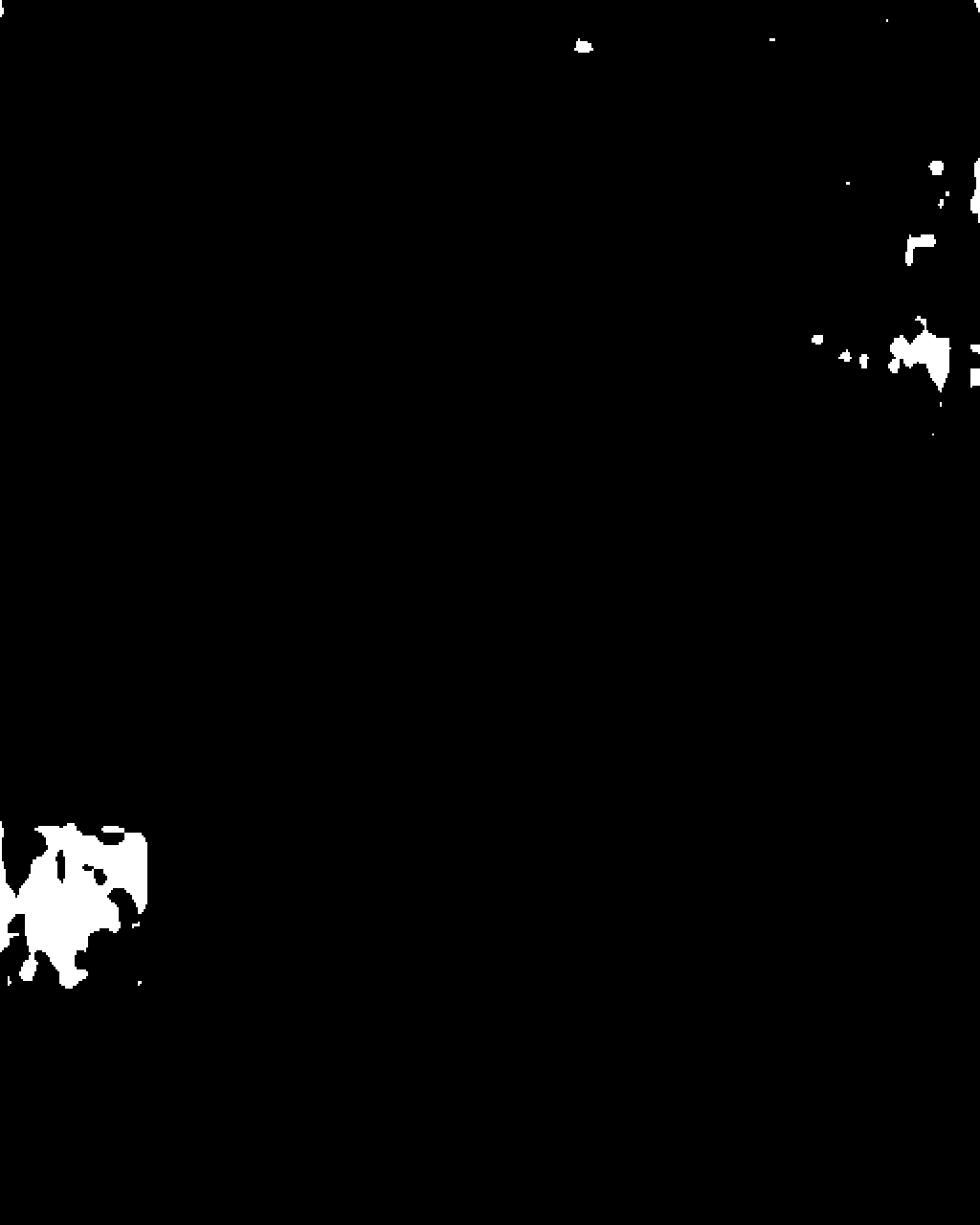}&
		\includegraphics[width=\wvisual, height=\hvisualtall]{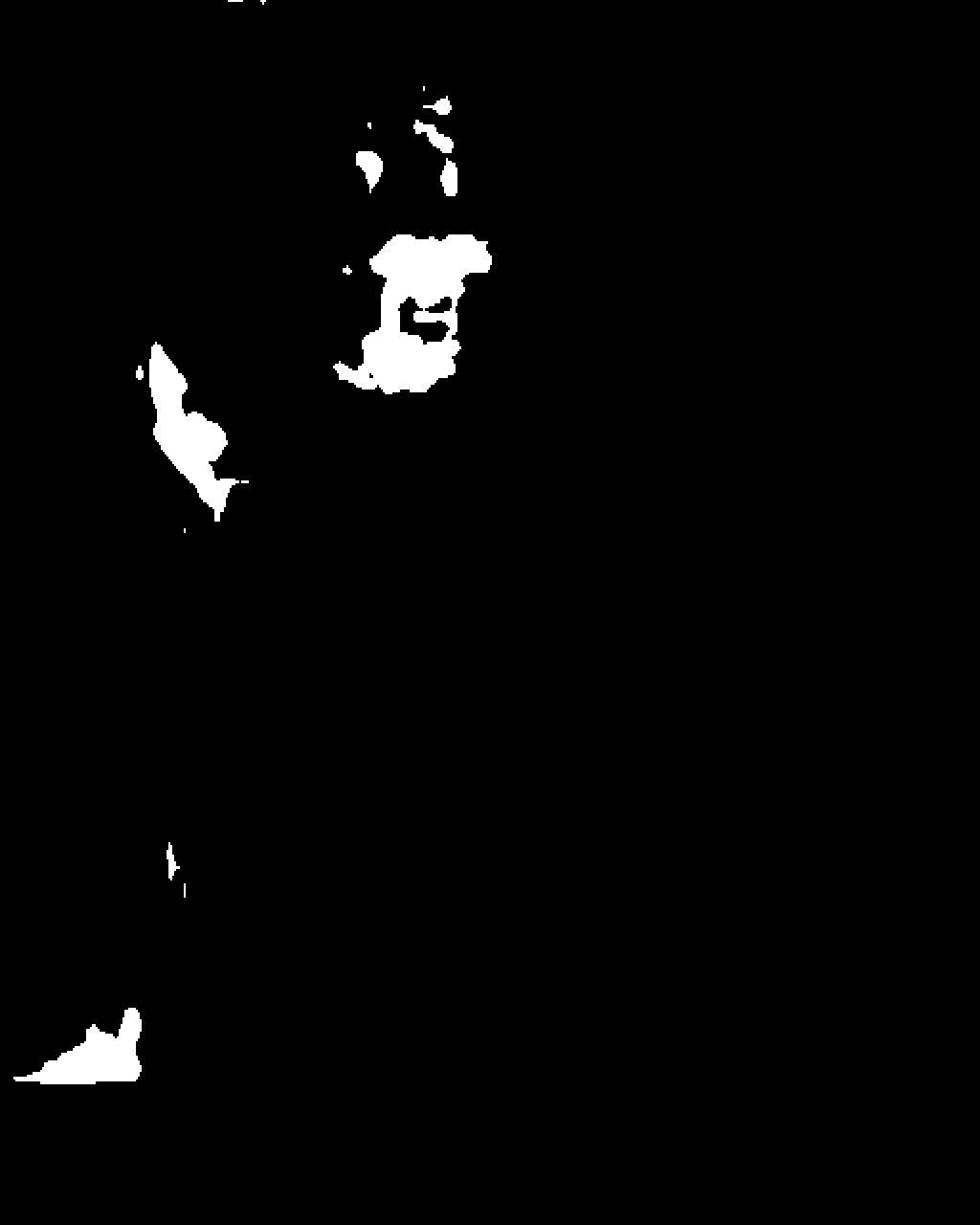}&
		\includegraphics[width=\wvisual, height=\hvisualtall]{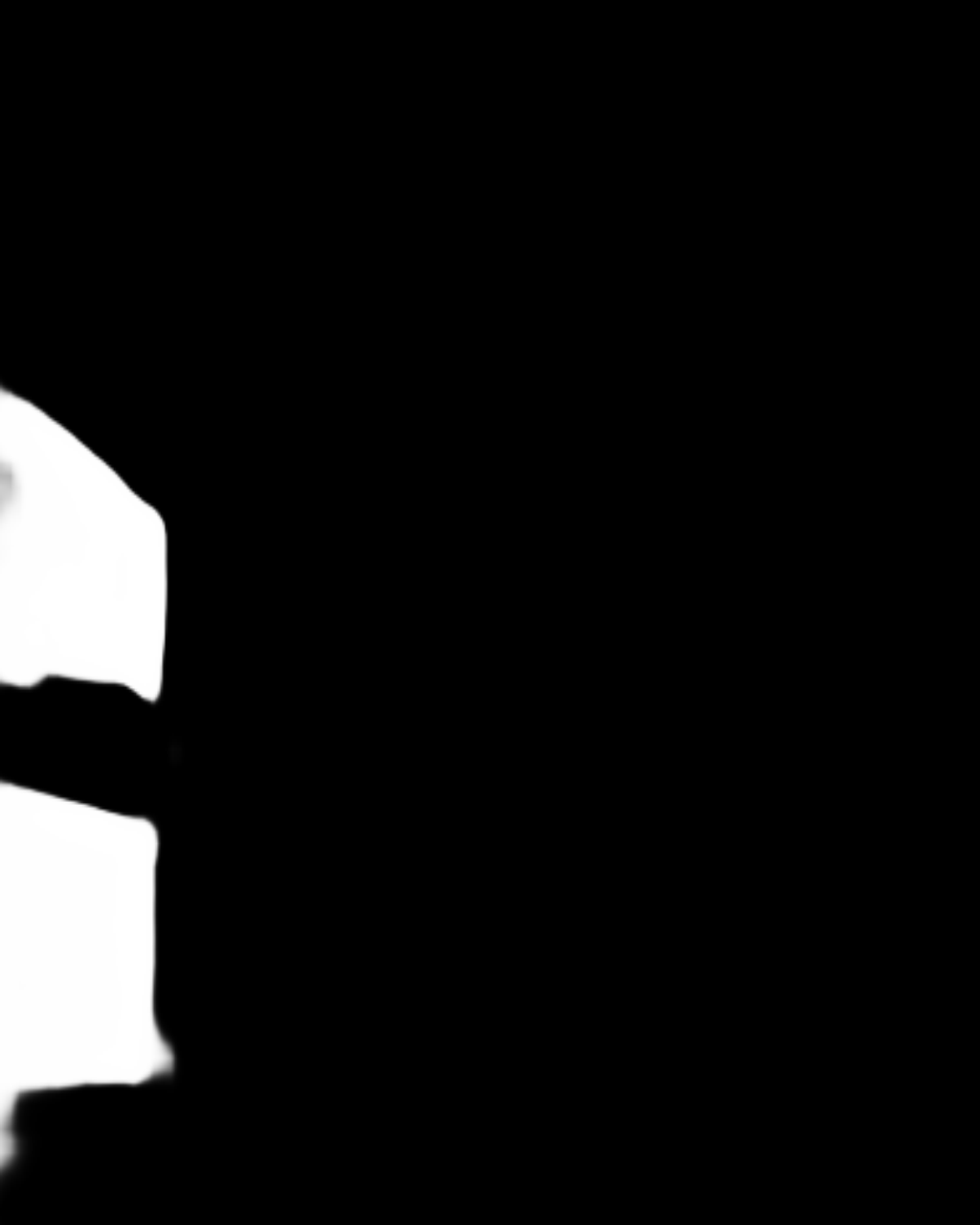}&
		\includegraphics[width=\wvisual, height=\hvisualtall]{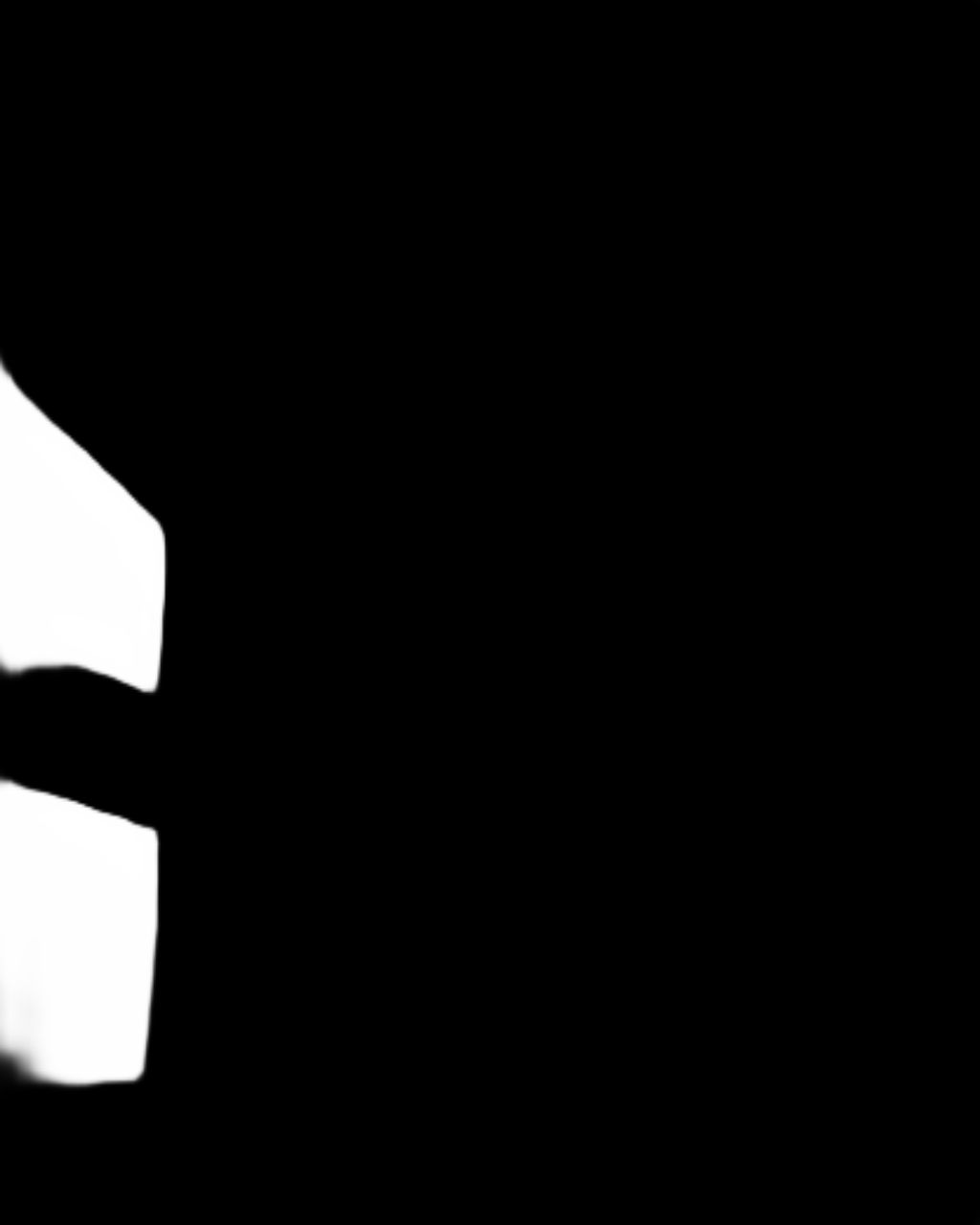}&
		\includegraphics[width=\wvisual, height=\hvisualtall]{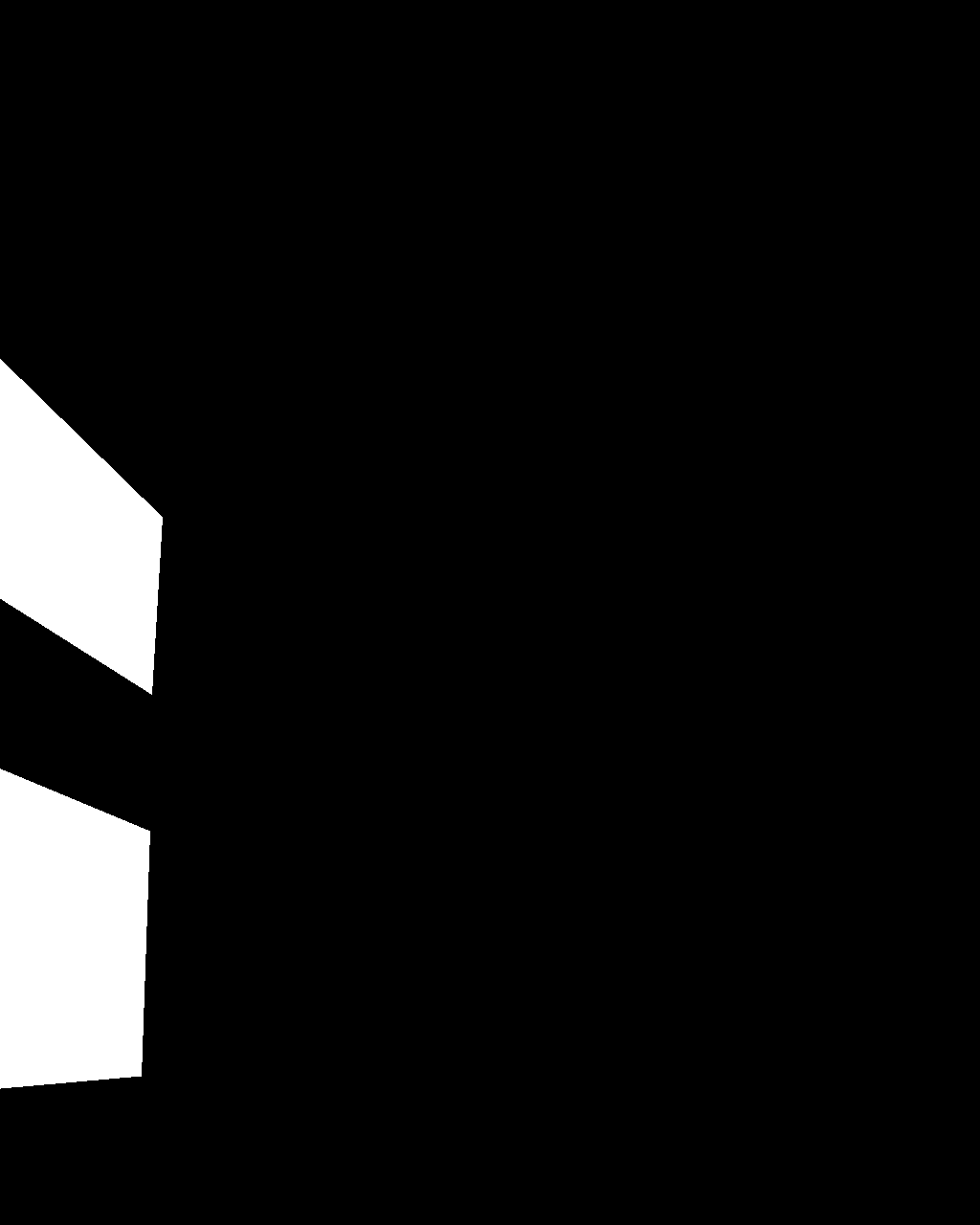} \\	
		
		\includegraphics[width=\wvisual, height=\hvisualtall]{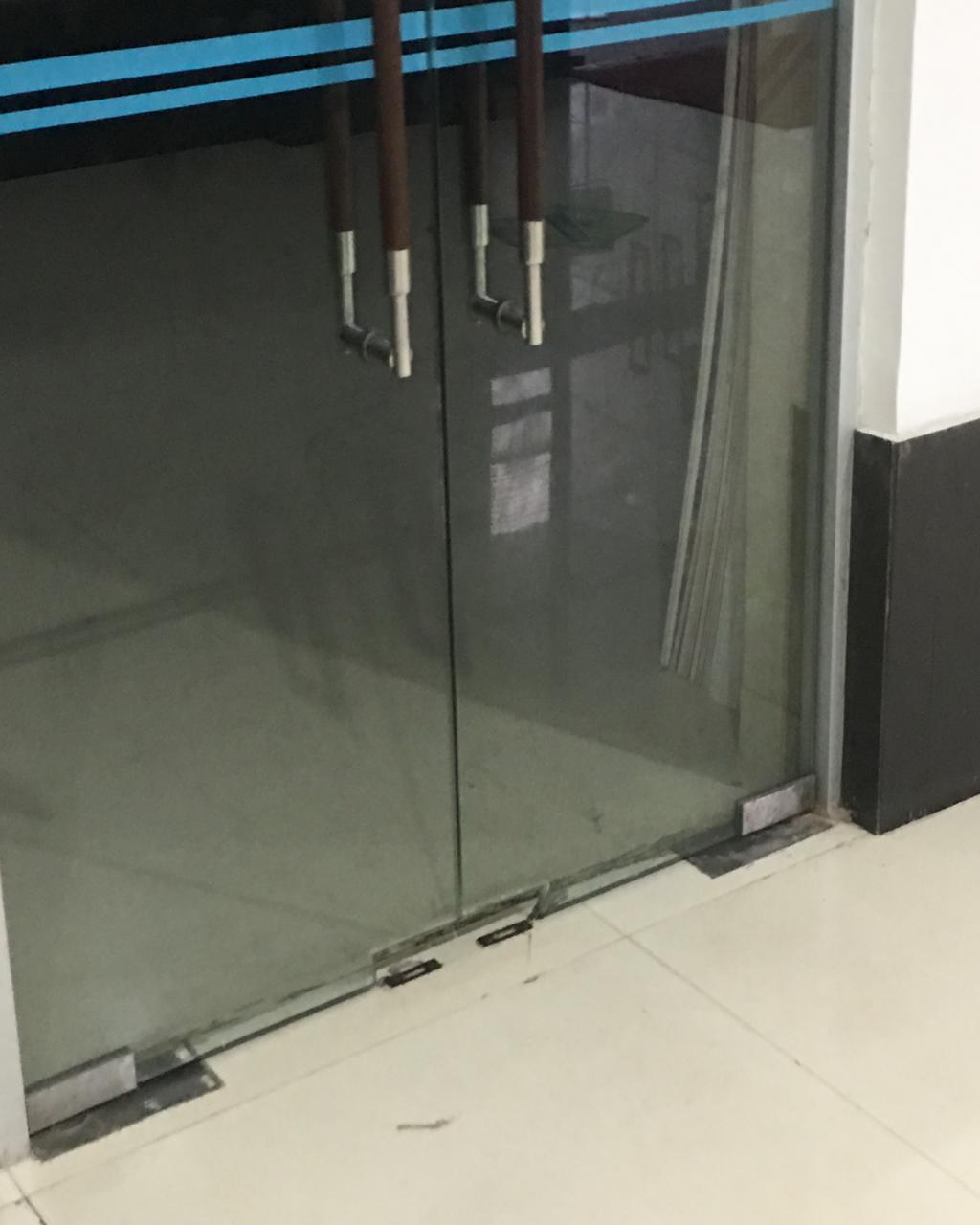}&
		\includegraphics[width=\wvisual, height=\hvisualtall]{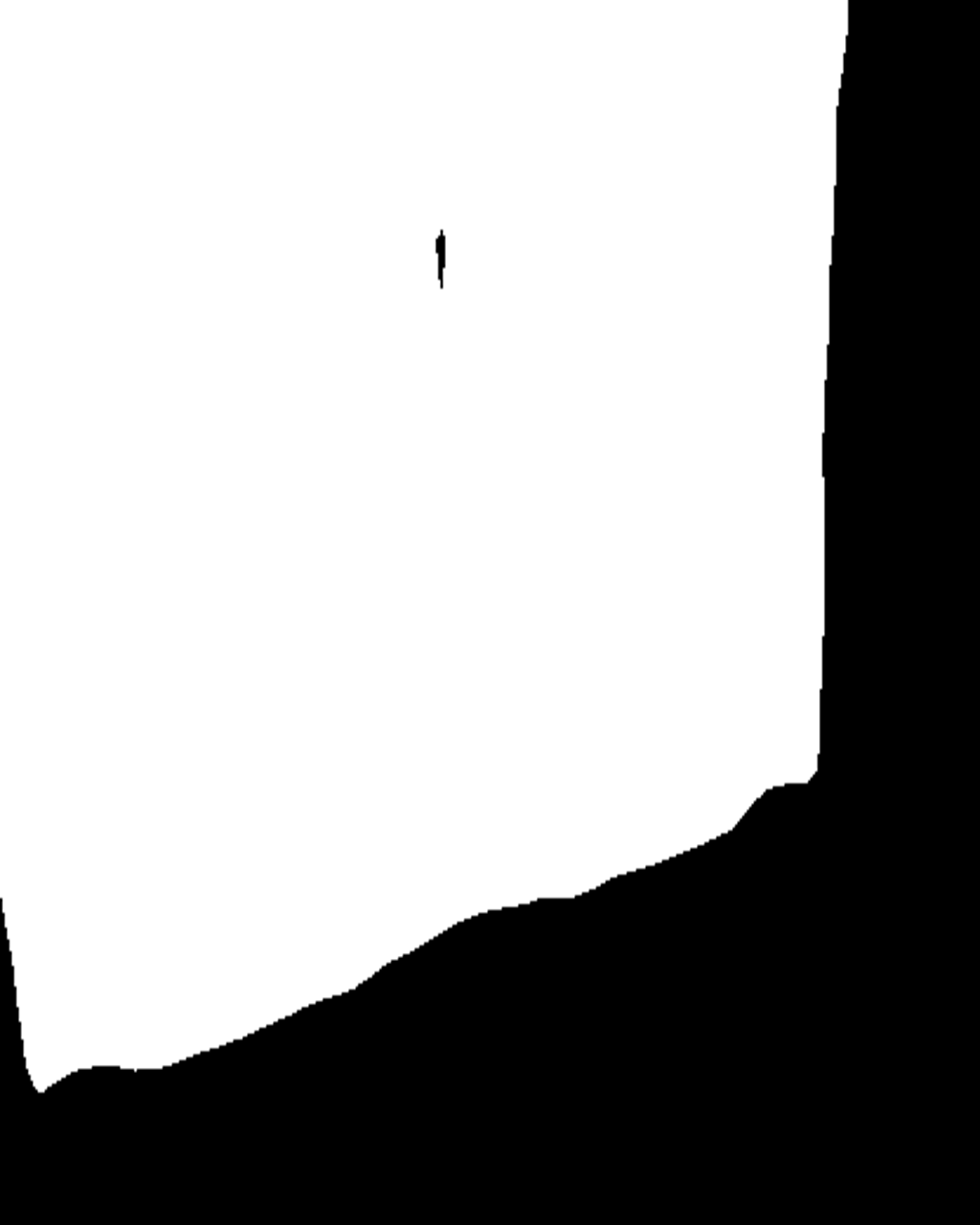}&
		\includegraphics[width=\wvisual, height=\hvisualtall]{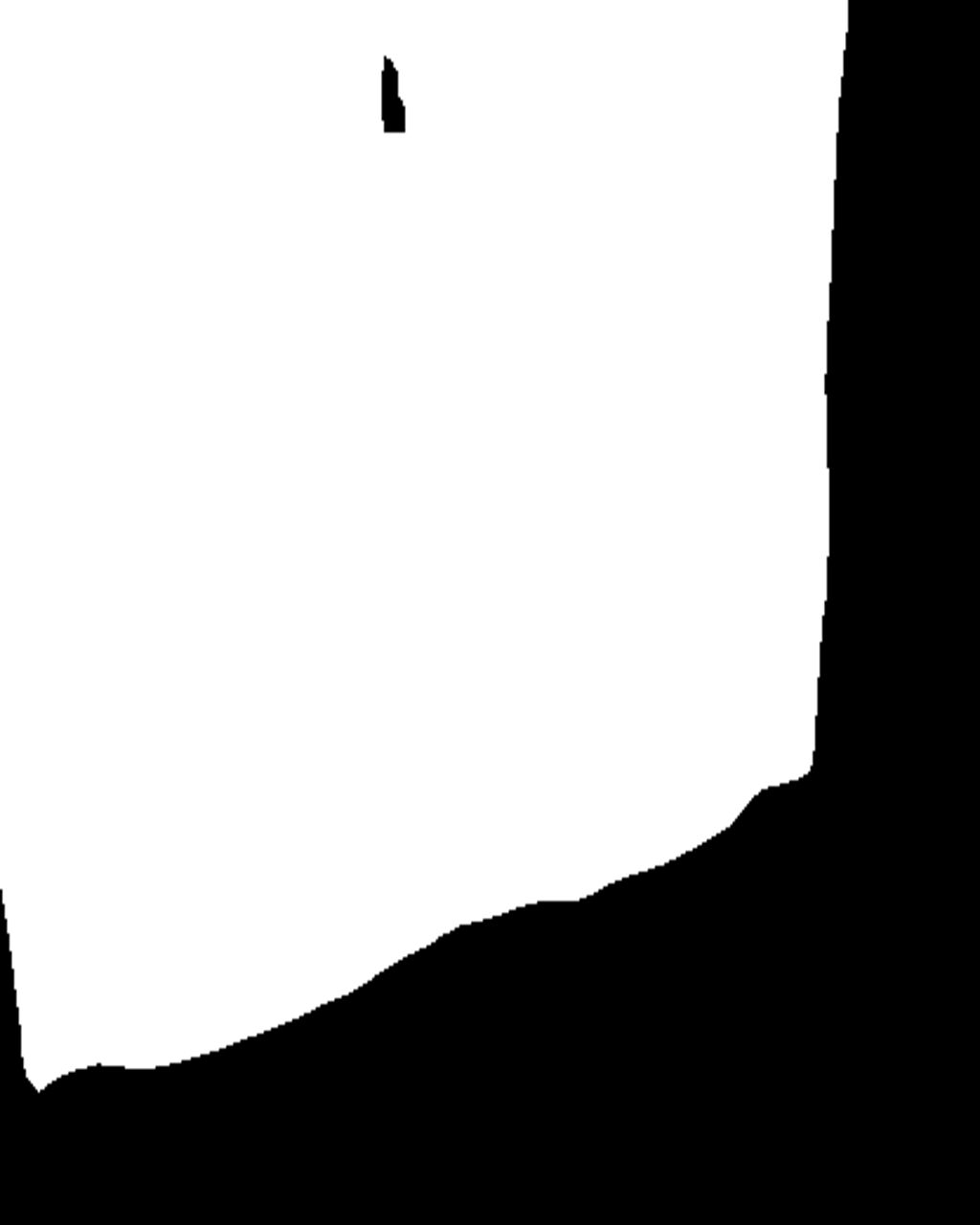}&
		\includegraphics[width=\wvisual, height=\hvisualtall]{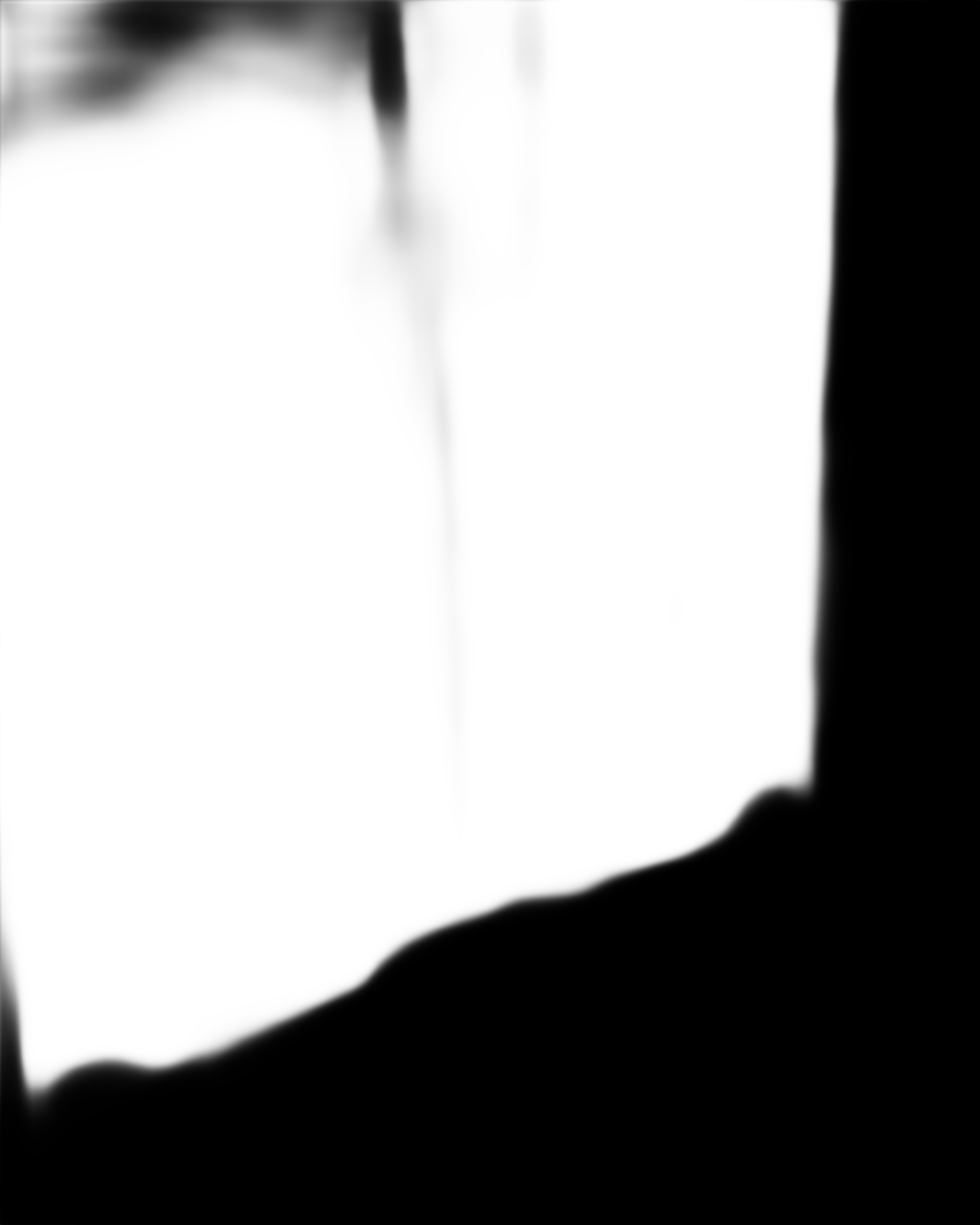}&
		\includegraphics[width=\wvisual, height=\hvisualtall]{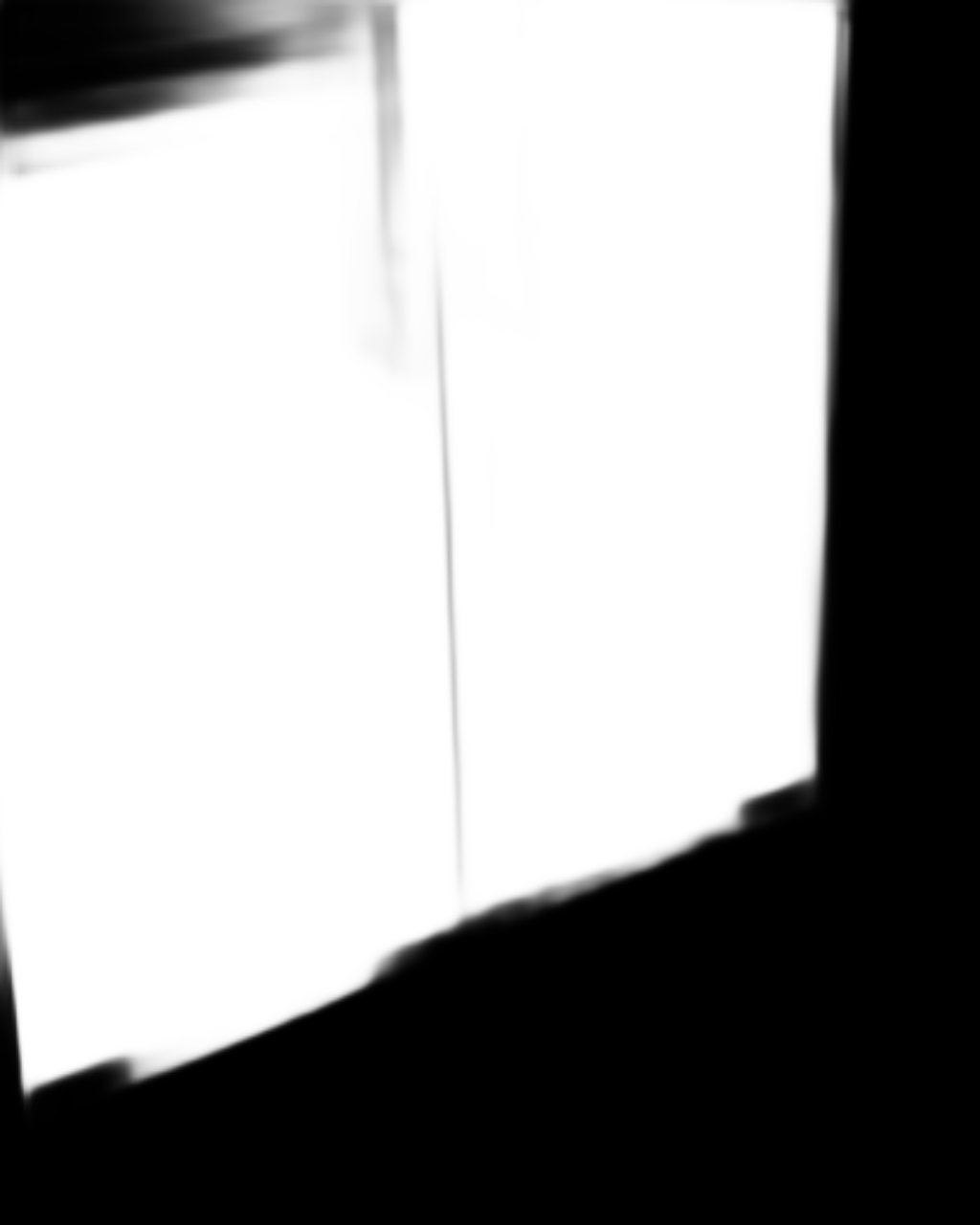}&
		\includegraphics[width=\wvisual, height=\hvisualtall]{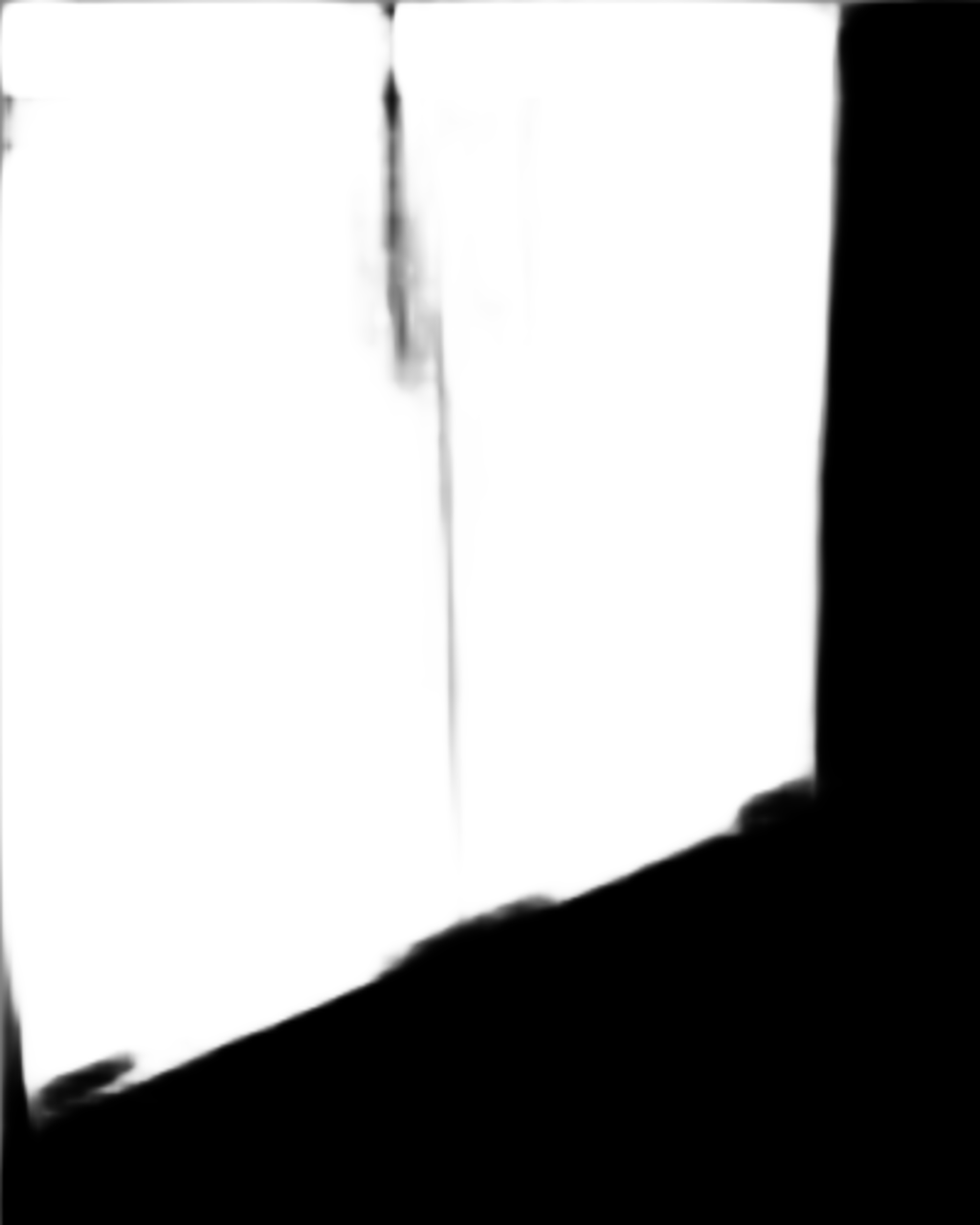}&
		\includegraphics[width=\wvisual, height=\hvisualtall]{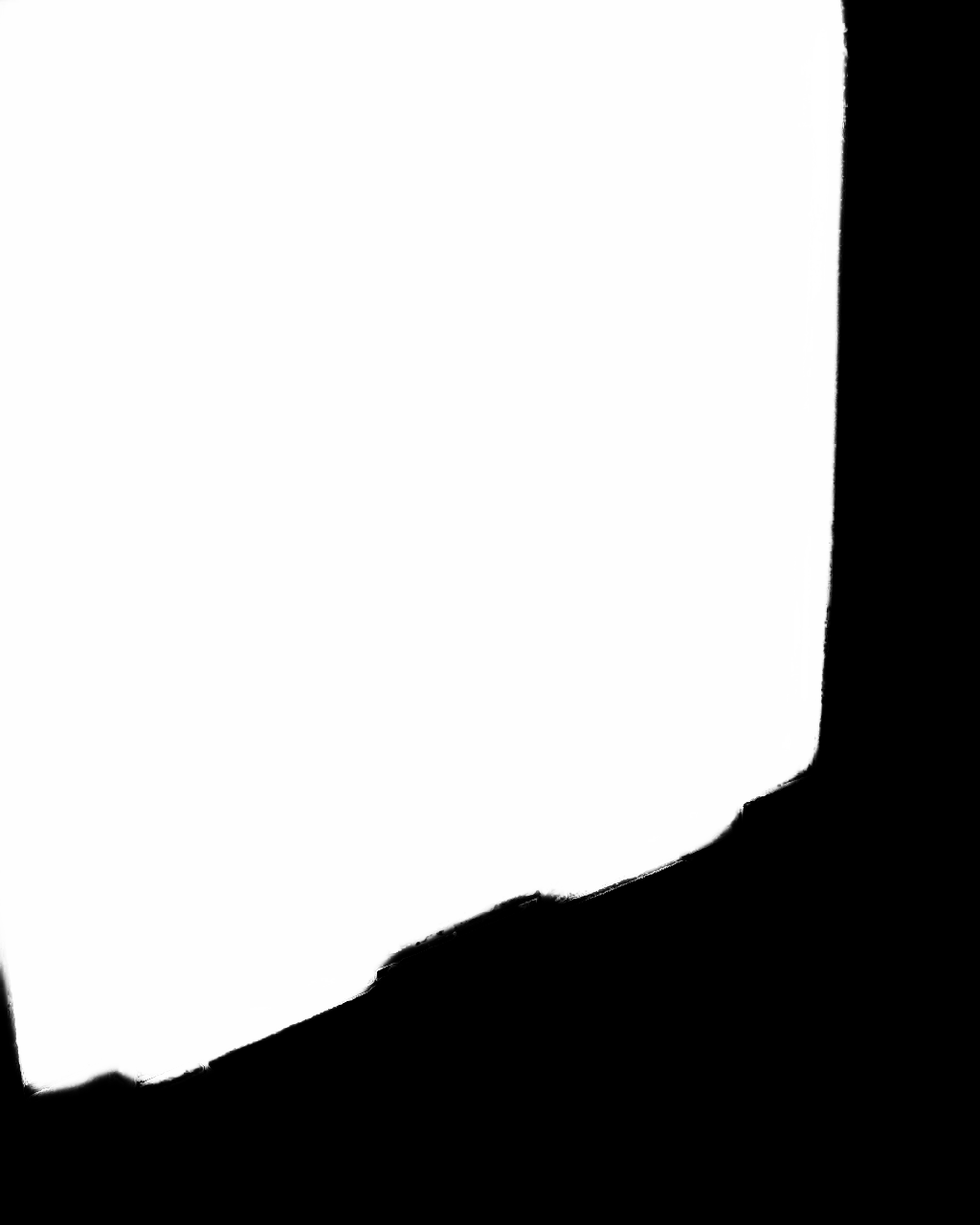}&
		\includegraphics[width=\wvisual, height=\hvisualtall]{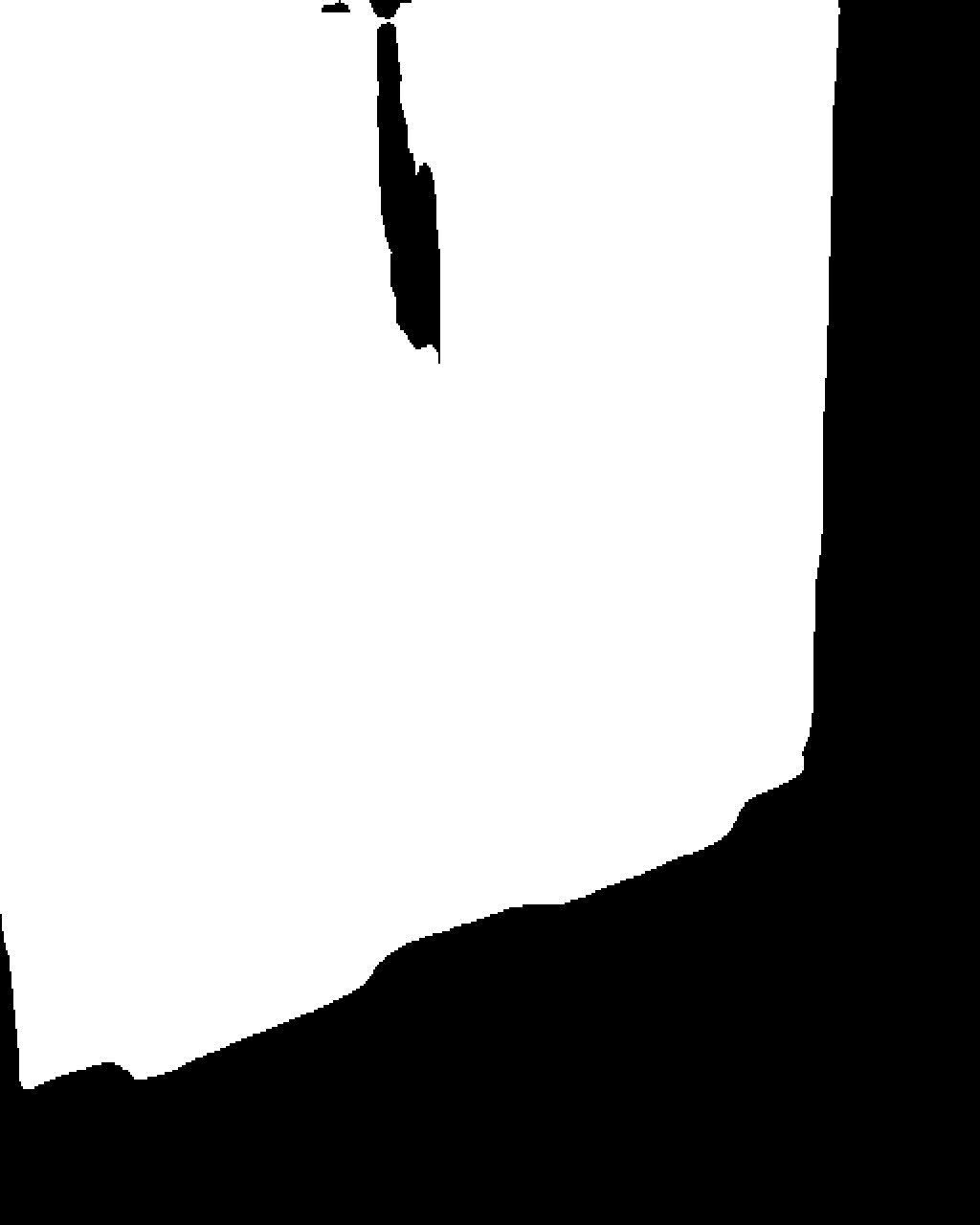}&
		\includegraphics[width=\wvisual, height=\hvisualtall]{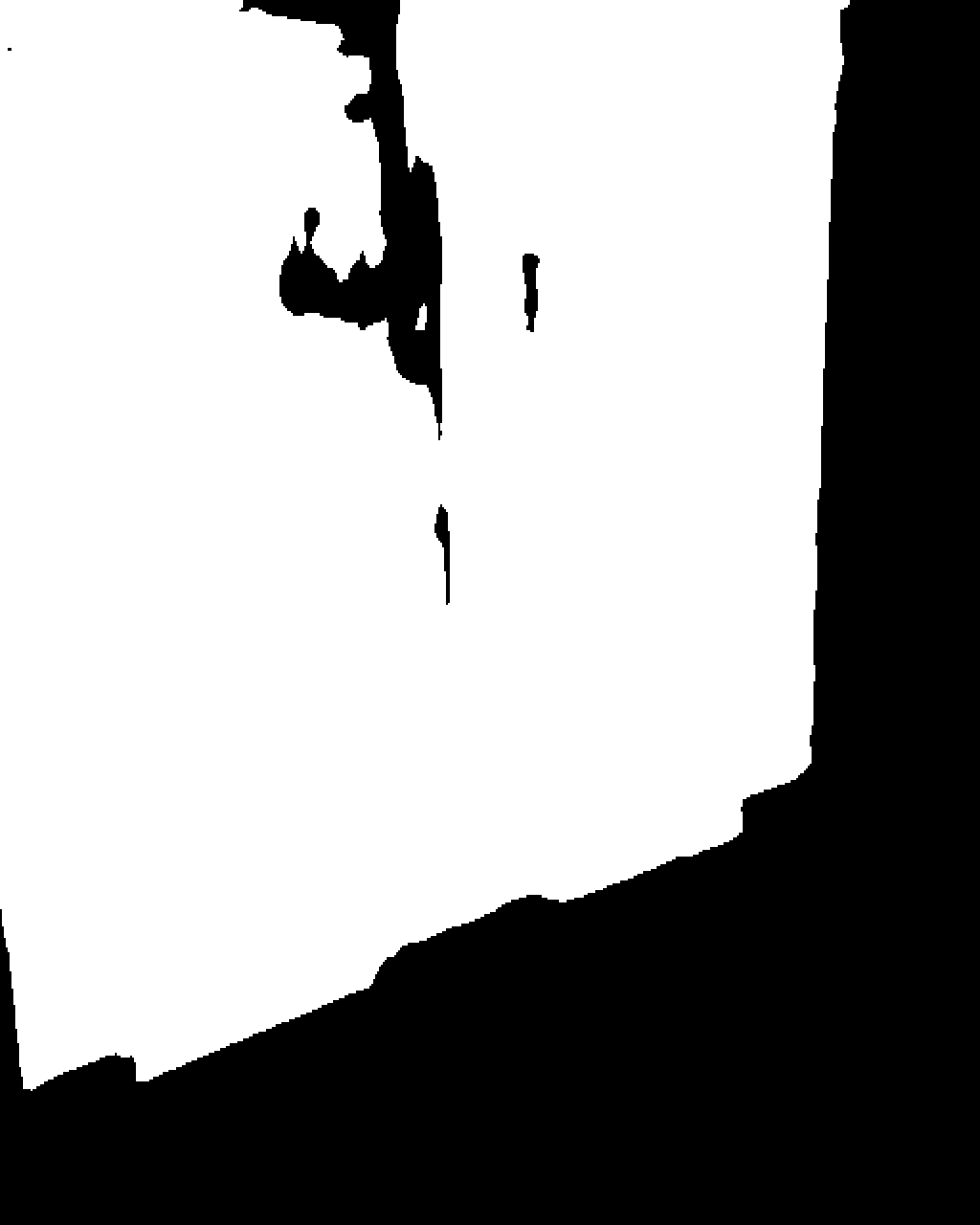}&
		\includegraphics[width=\wvisual, height=\hvisualtall]{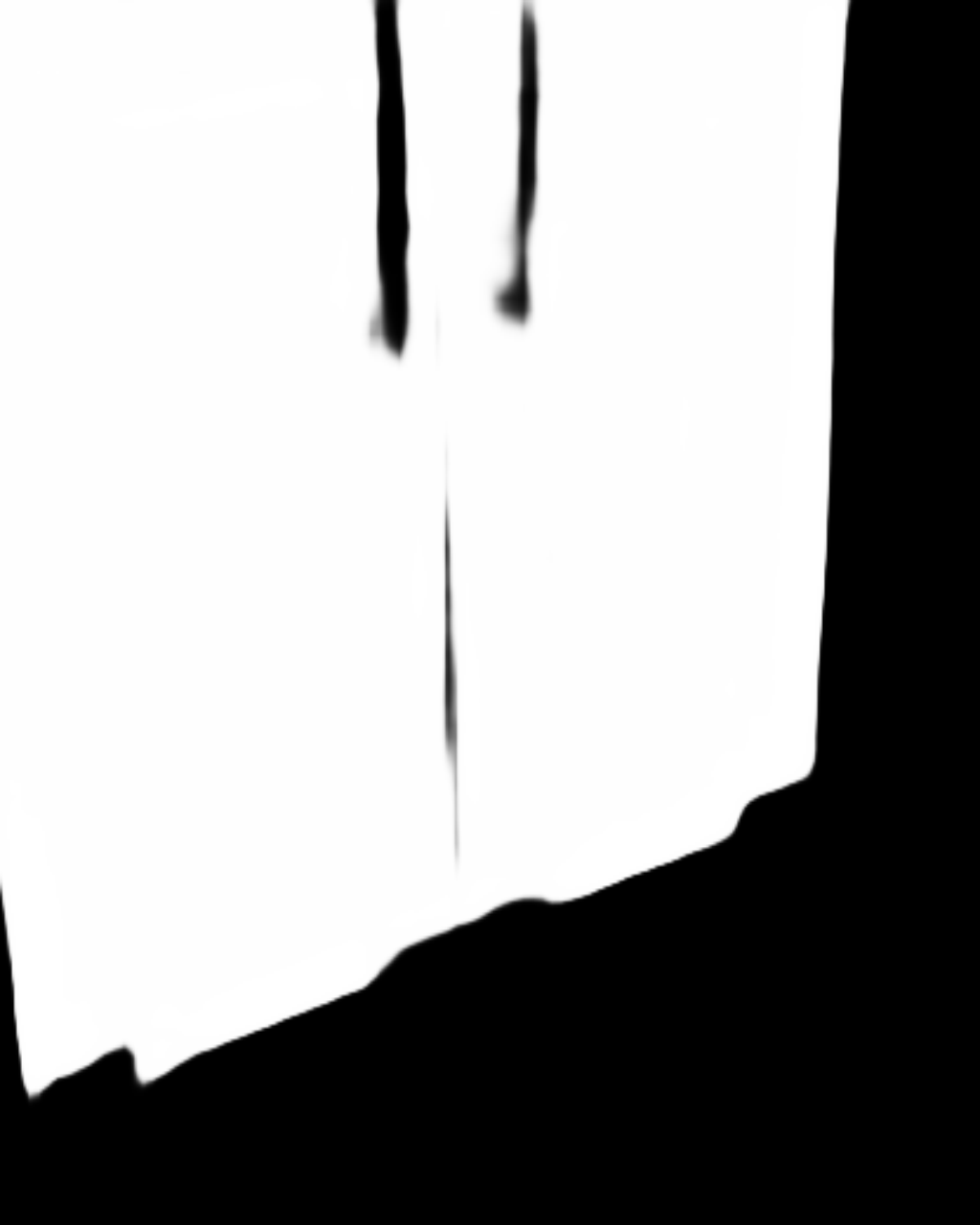}&
		\includegraphics[width=\wvisual, height=\hvisualtall]{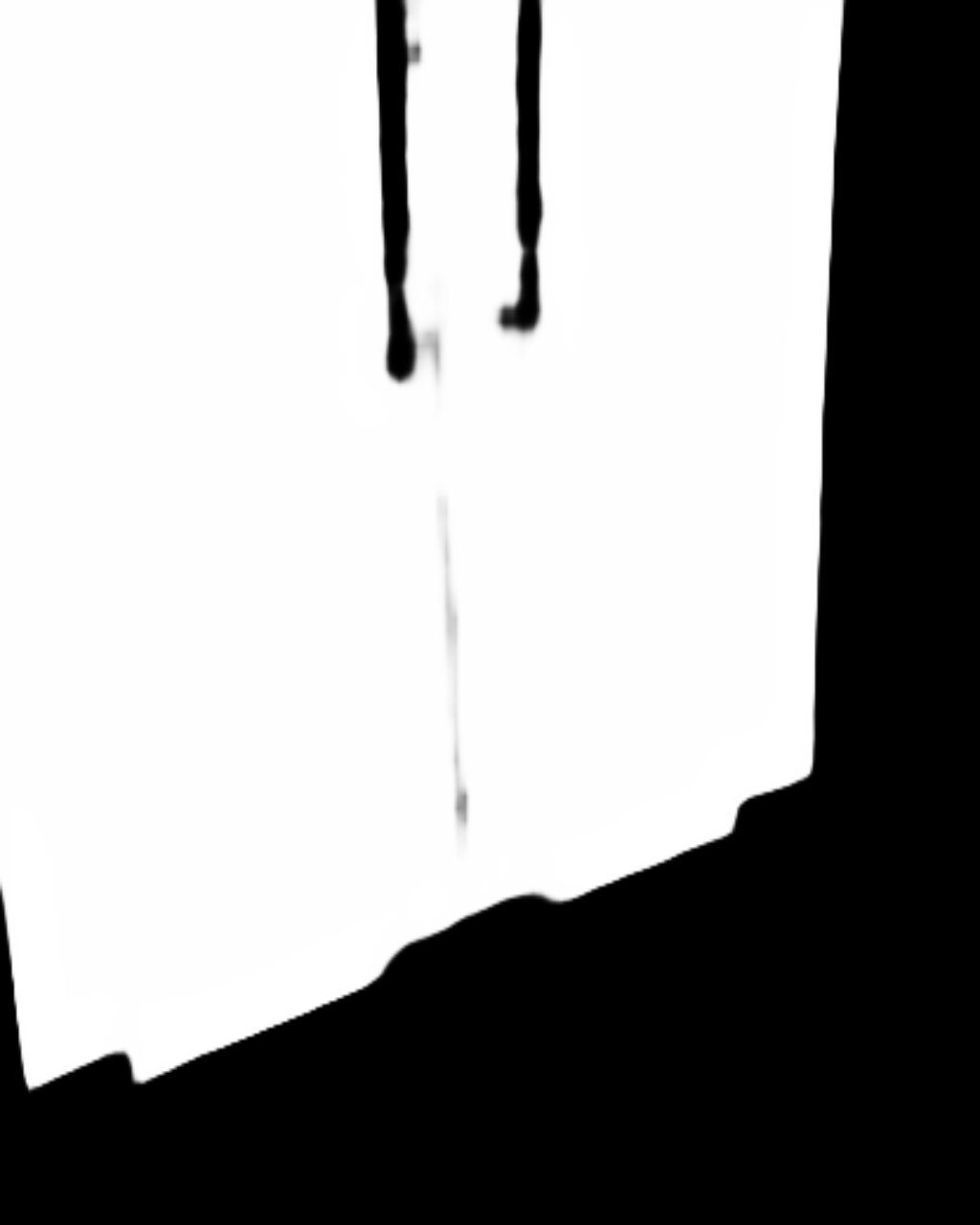}&
		\includegraphics[width=\wvisual, height=\hvisualtall]{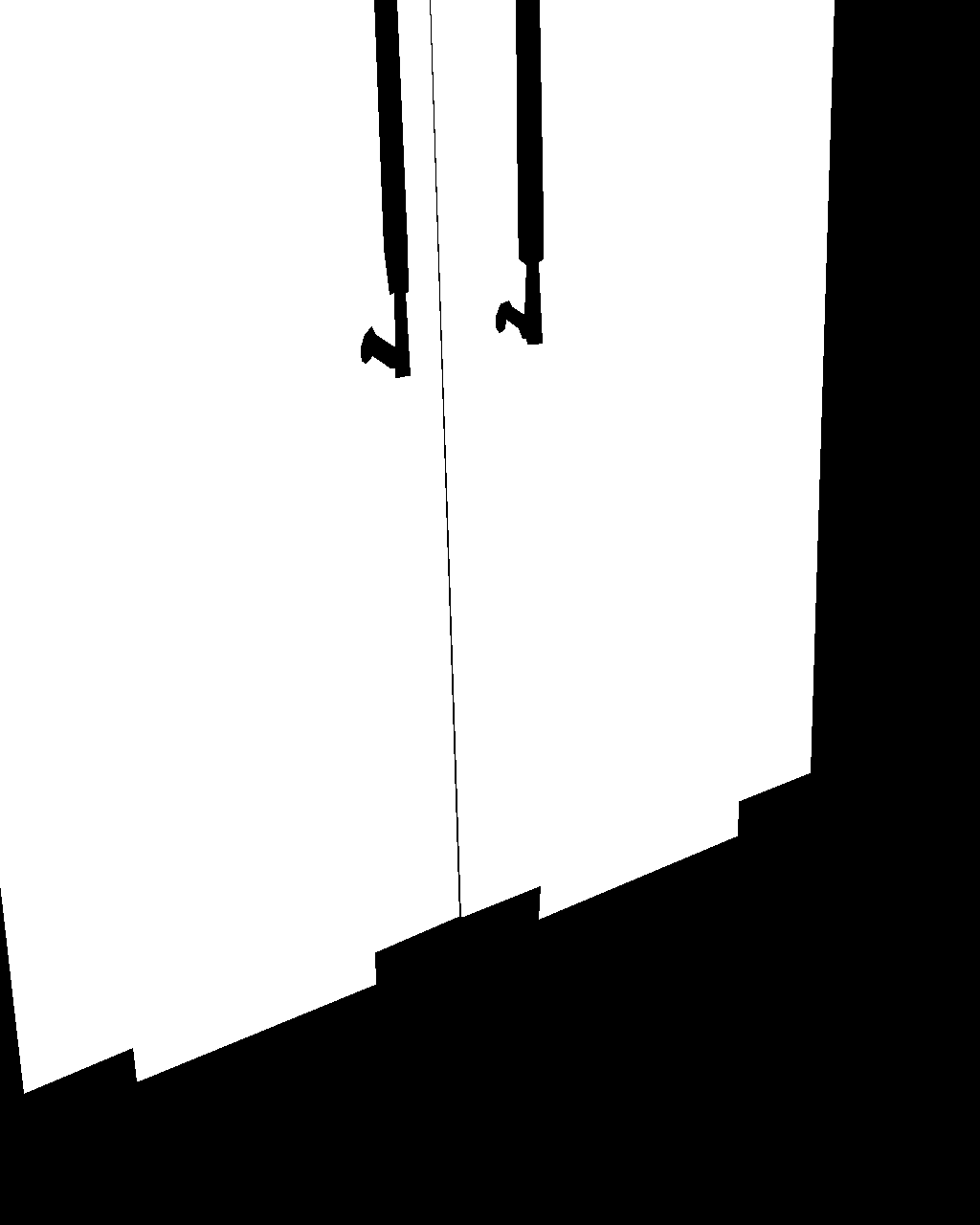} \\
		
		\includegraphics[width=\wvisual, height=\hvisualshort]{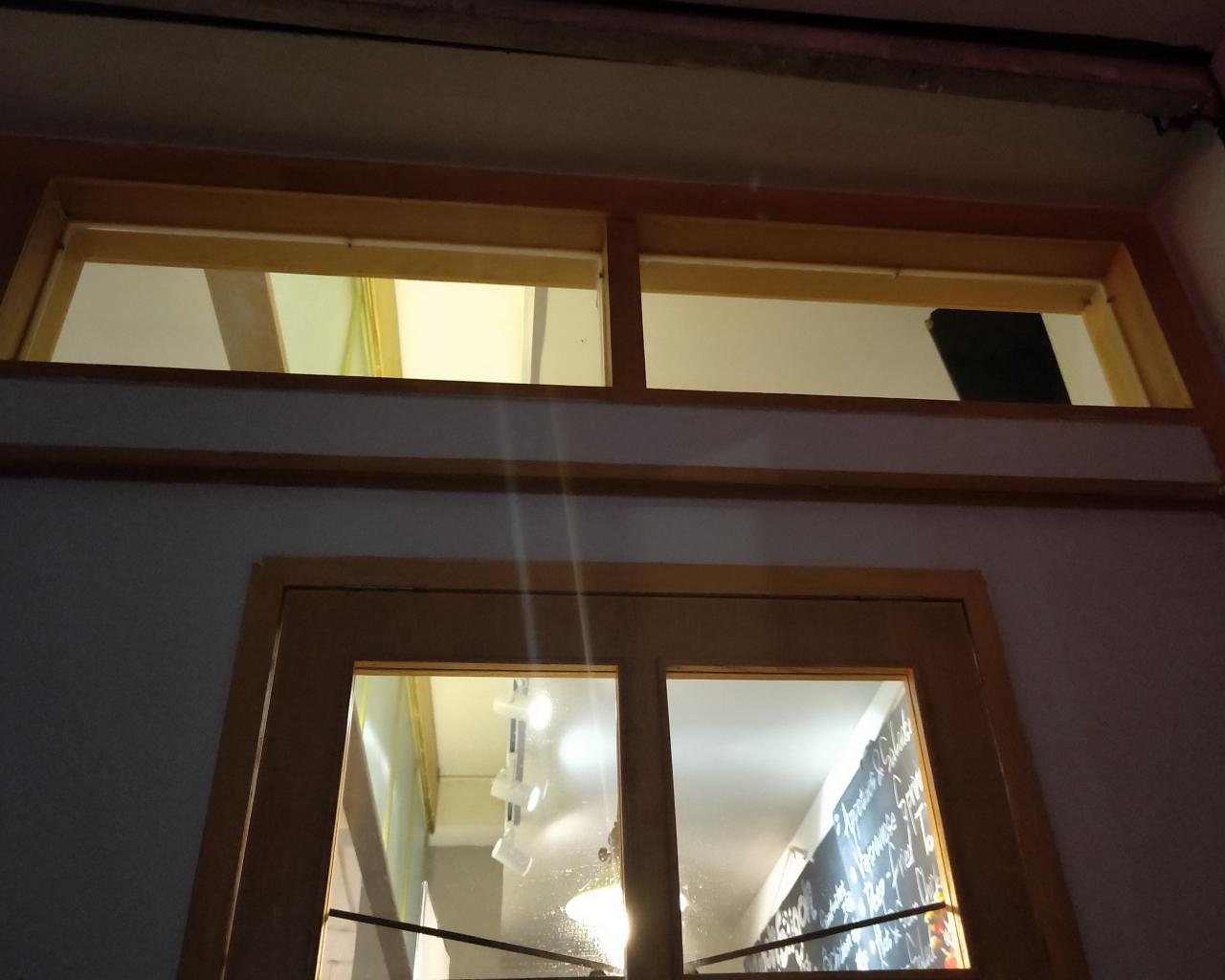}&
		\includegraphics[width=\wvisual, height=\hvisualshort]{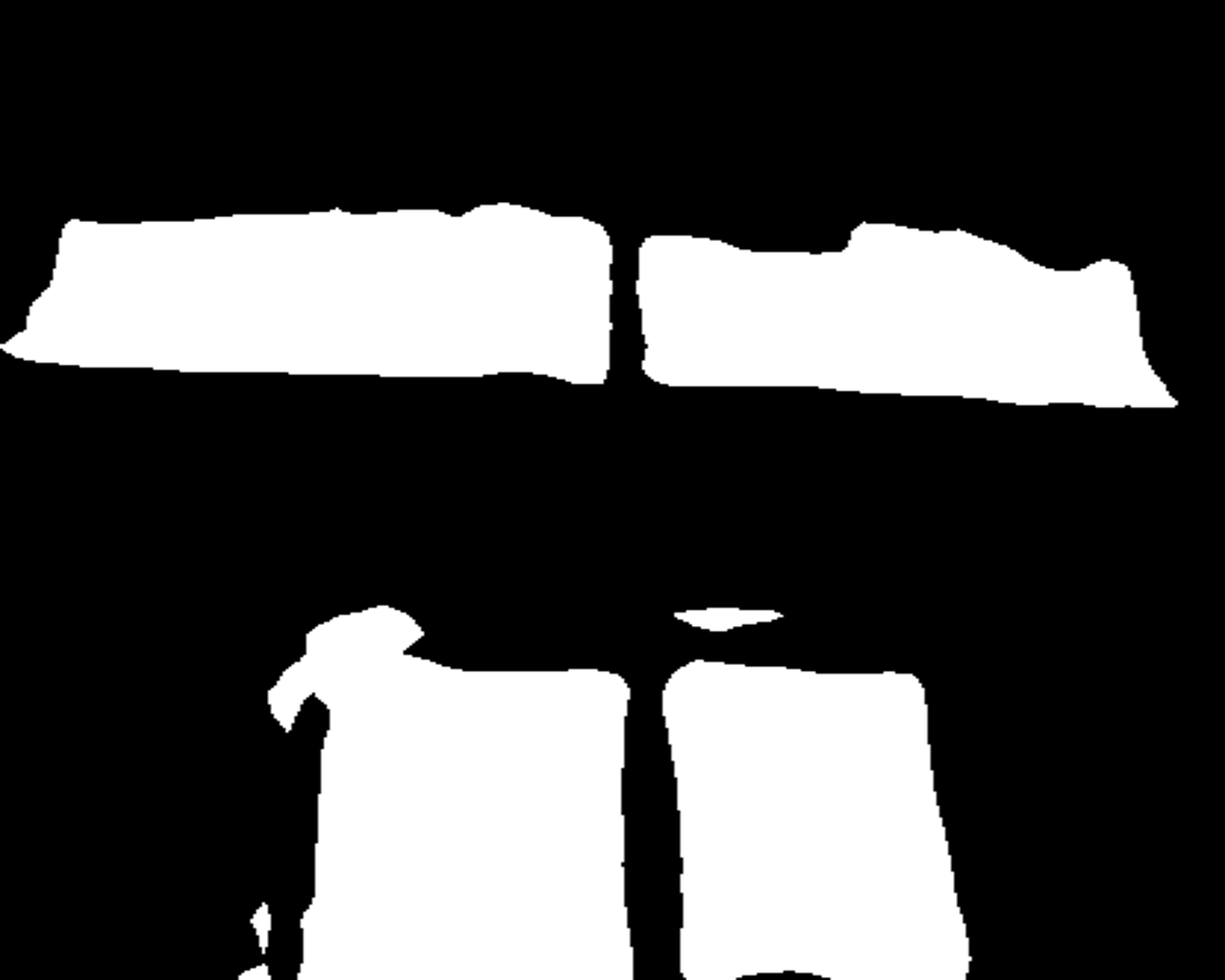}&
		\includegraphics[width=\wvisual, height=\hvisualshort]{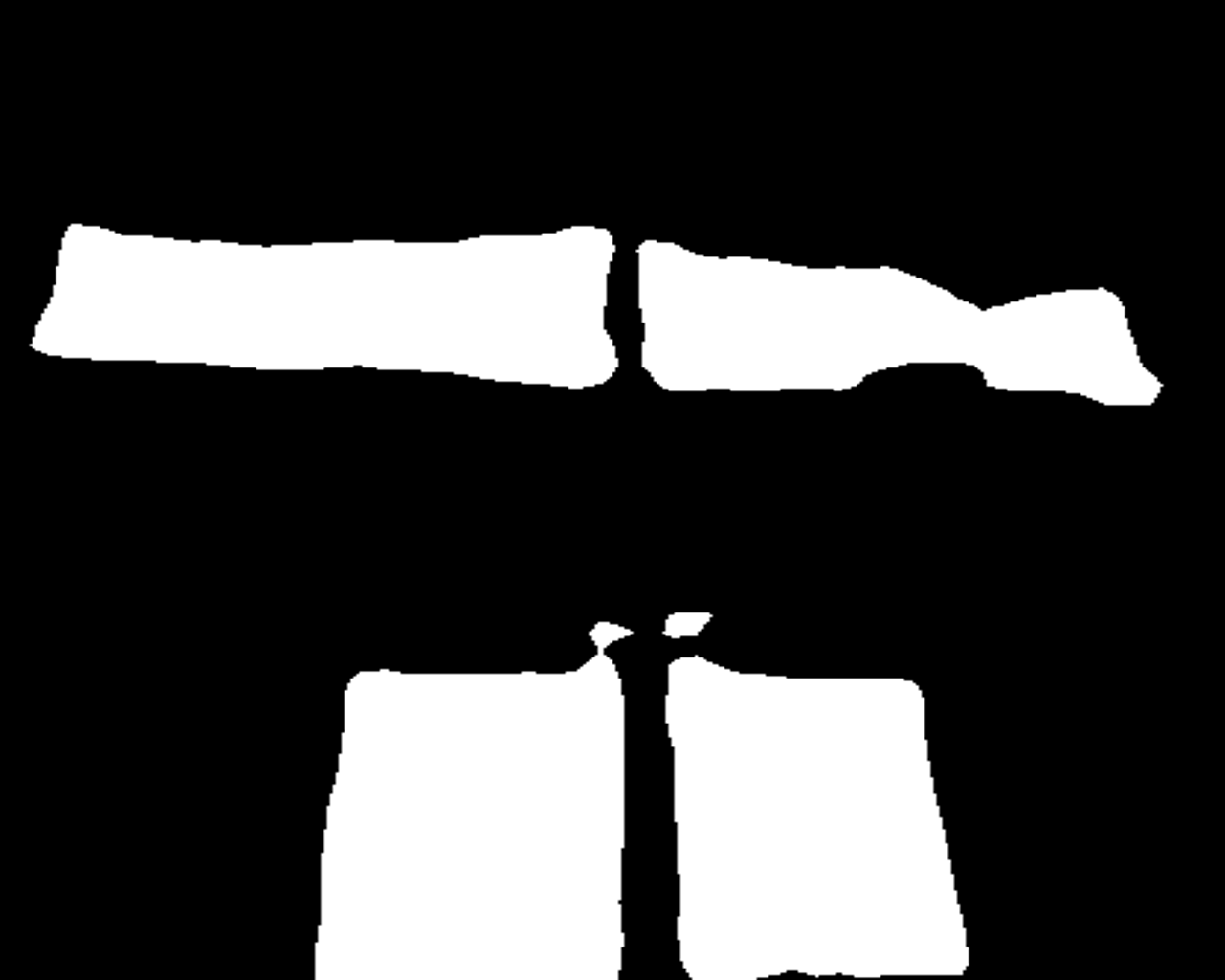}&
		\includegraphics[width=\wvisual, height=\hvisualshort]{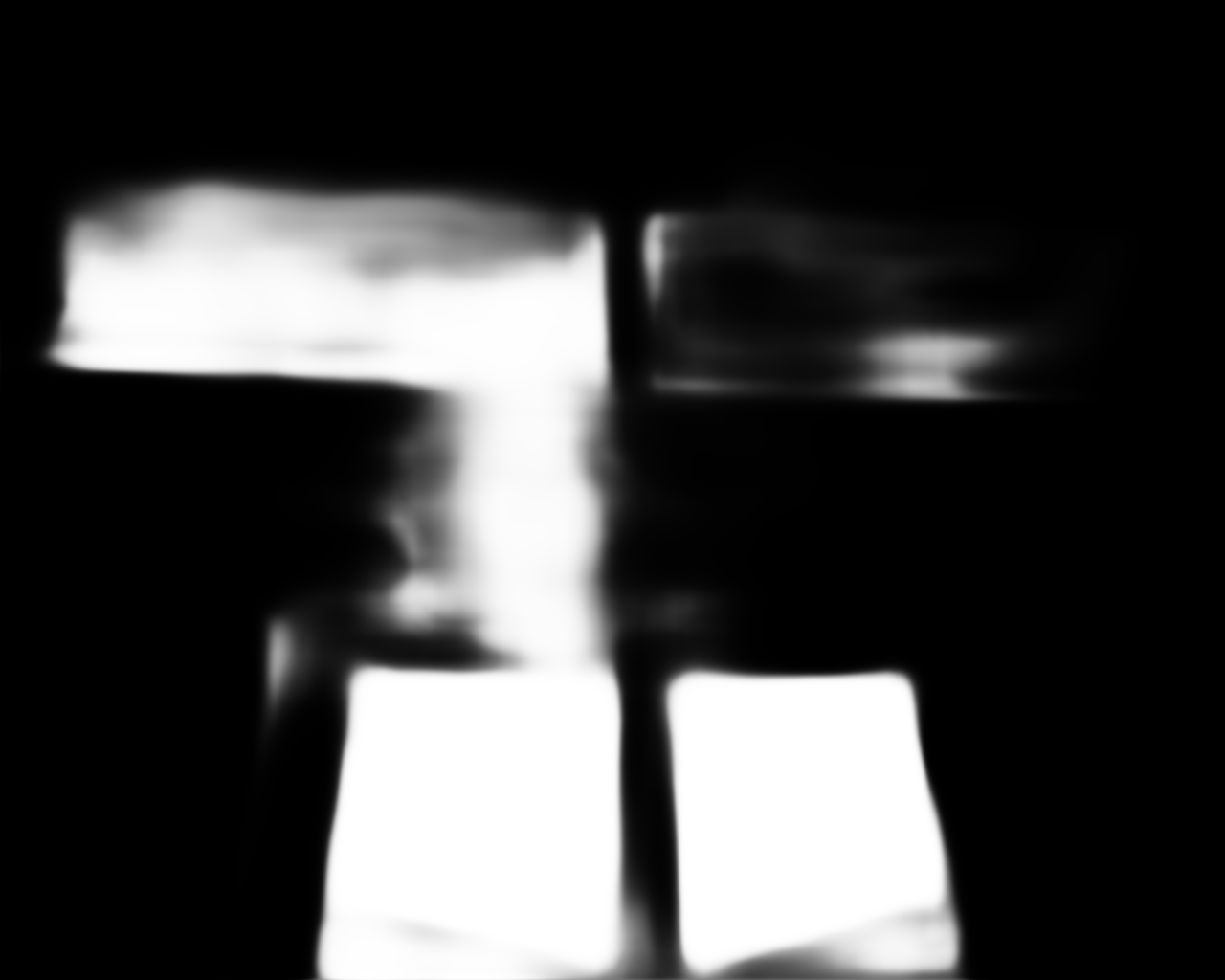}&
		\includegraphics[width=\wvisual, height=\hvisualshort]{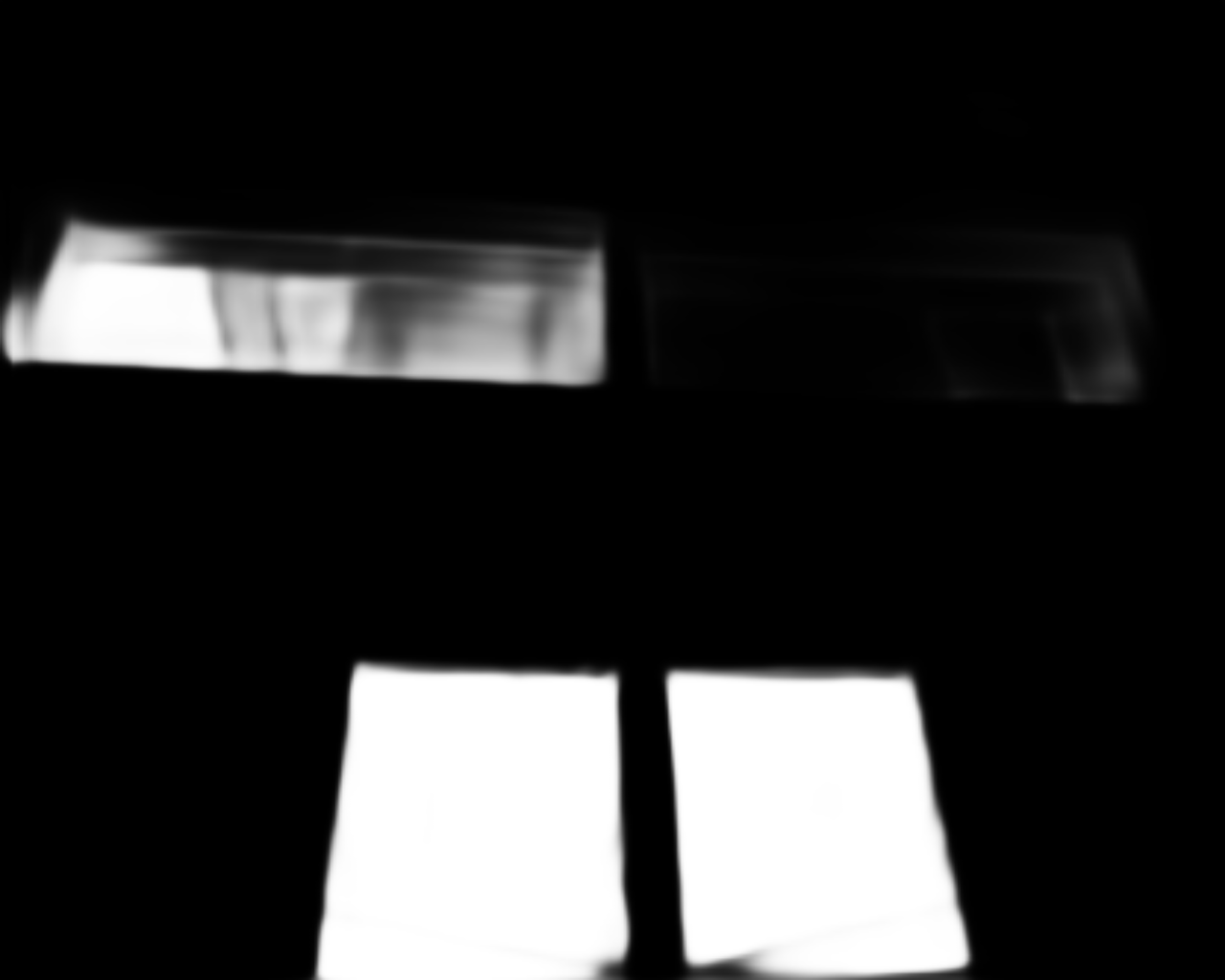}&
		\includegraphics[width=\wvisual, height=\hvisualshort]{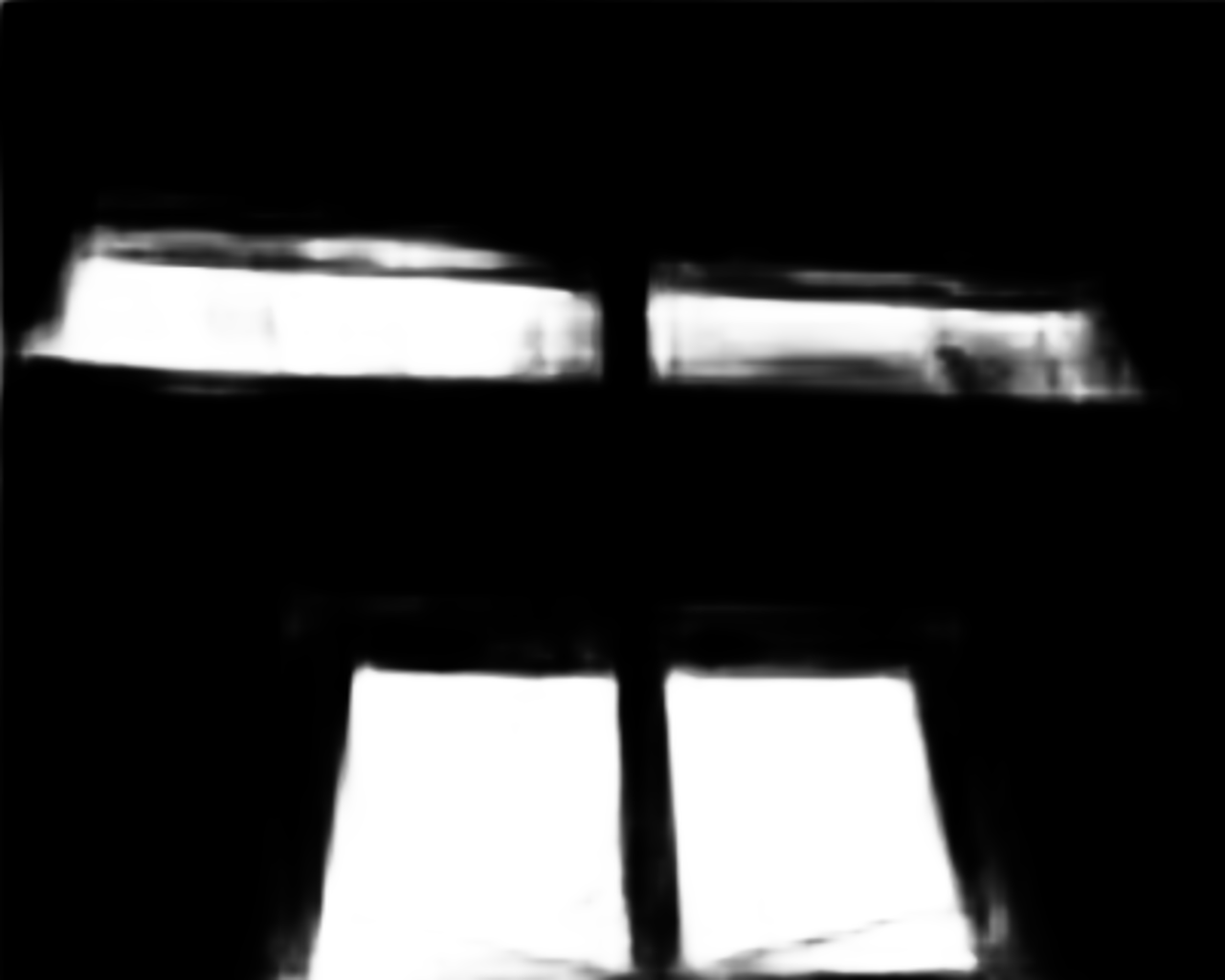}&
		\includegraphics[width=\wvisual, height=\hvisualshort]{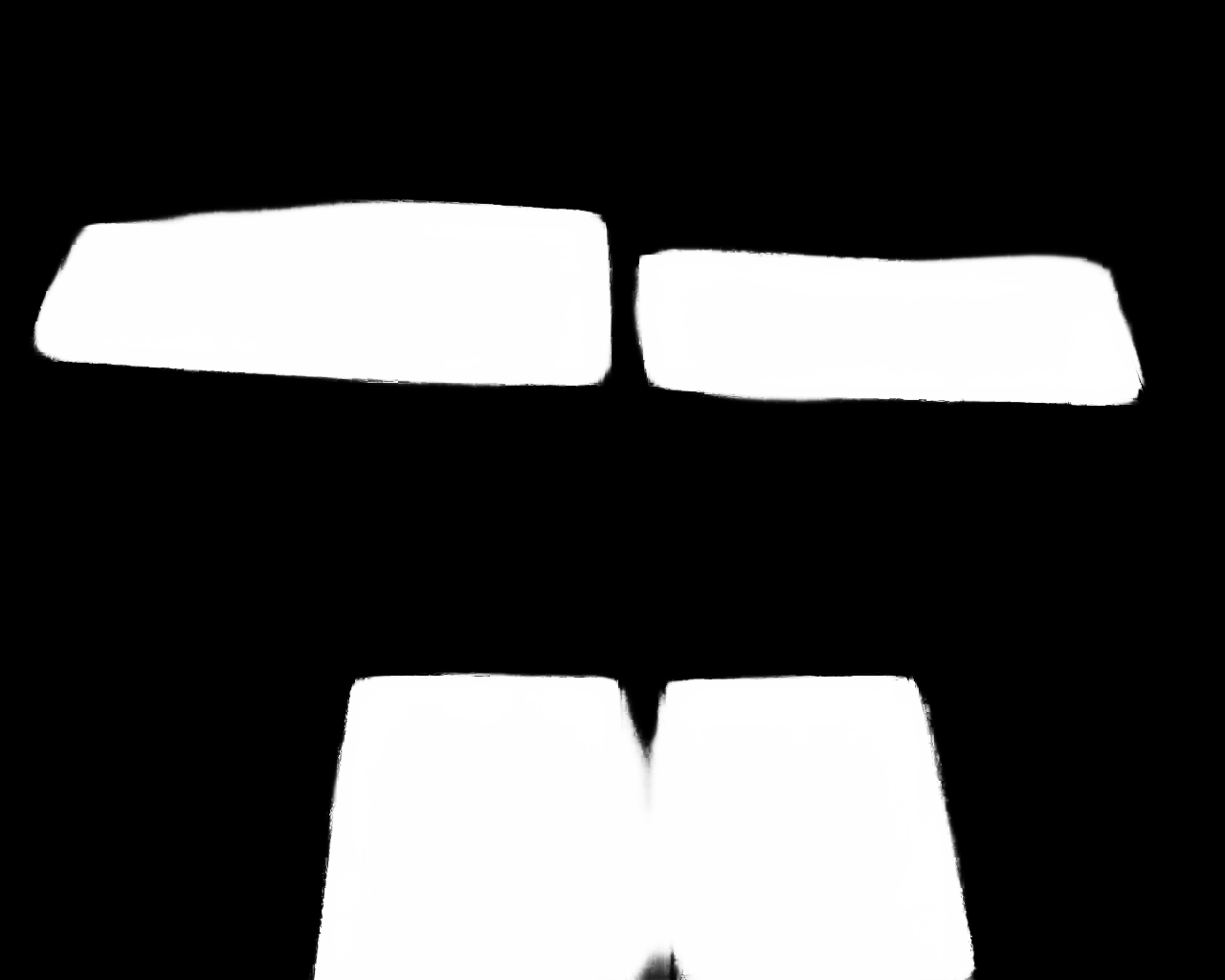}&
		\includegraphics[width=\wvisual, height=\hvisualshort]{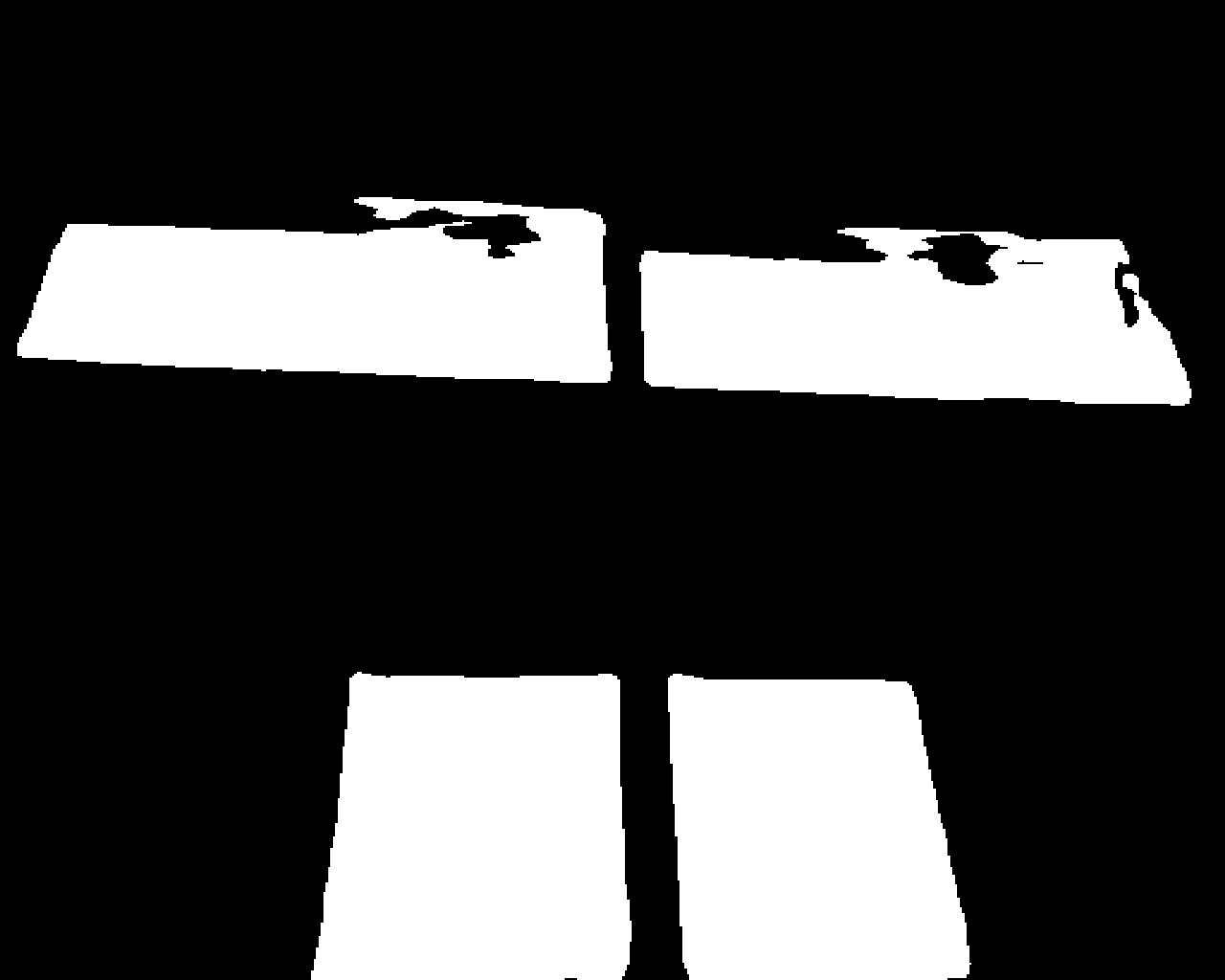}&
		\includegraphics[width=\wvisual, height=\hvisualshort]{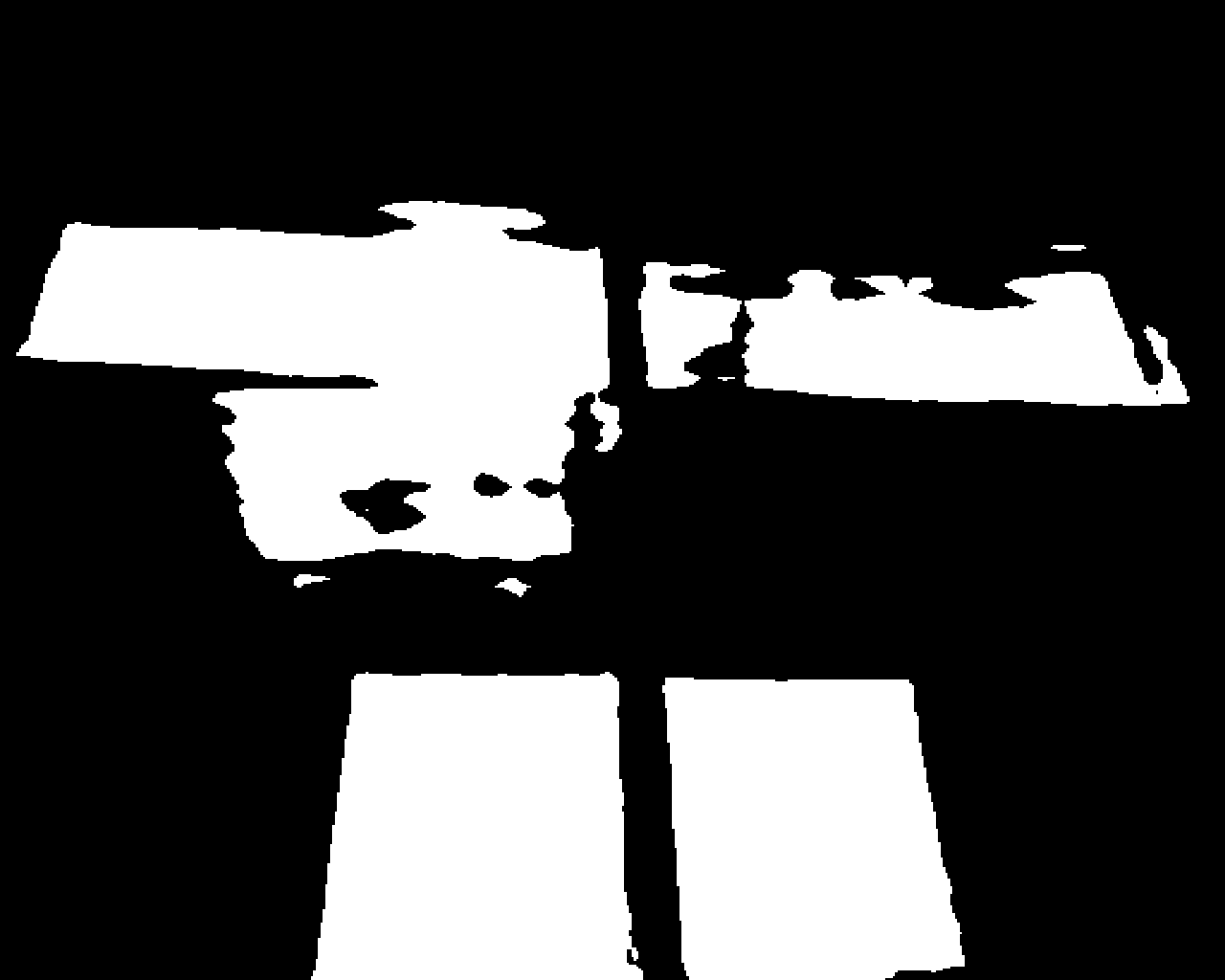}&
		\includegraphics[width=\wvisual, height=\hvisualshort]{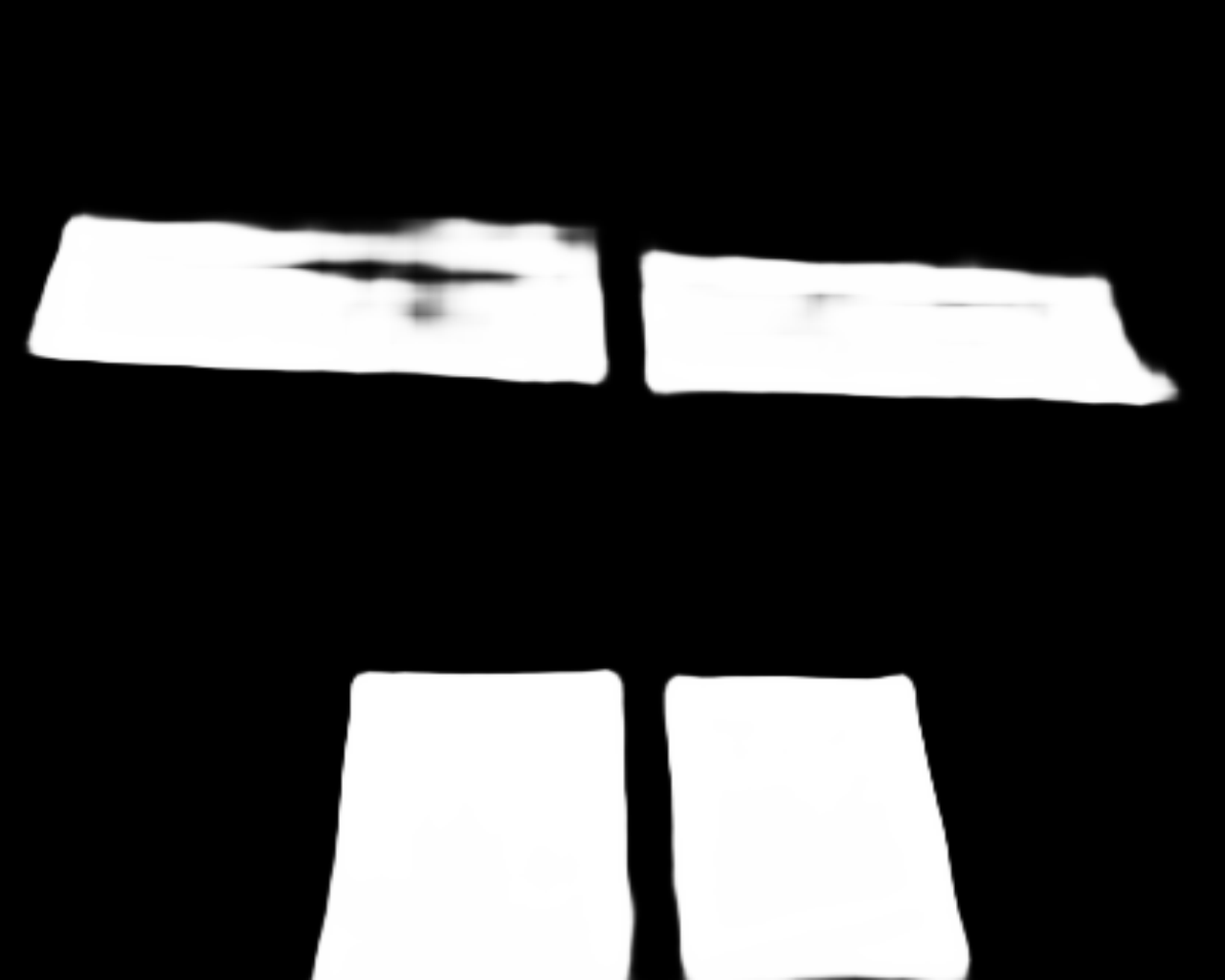}&
		\includegraphics[width=\wvisual, height=\hvisualshort]{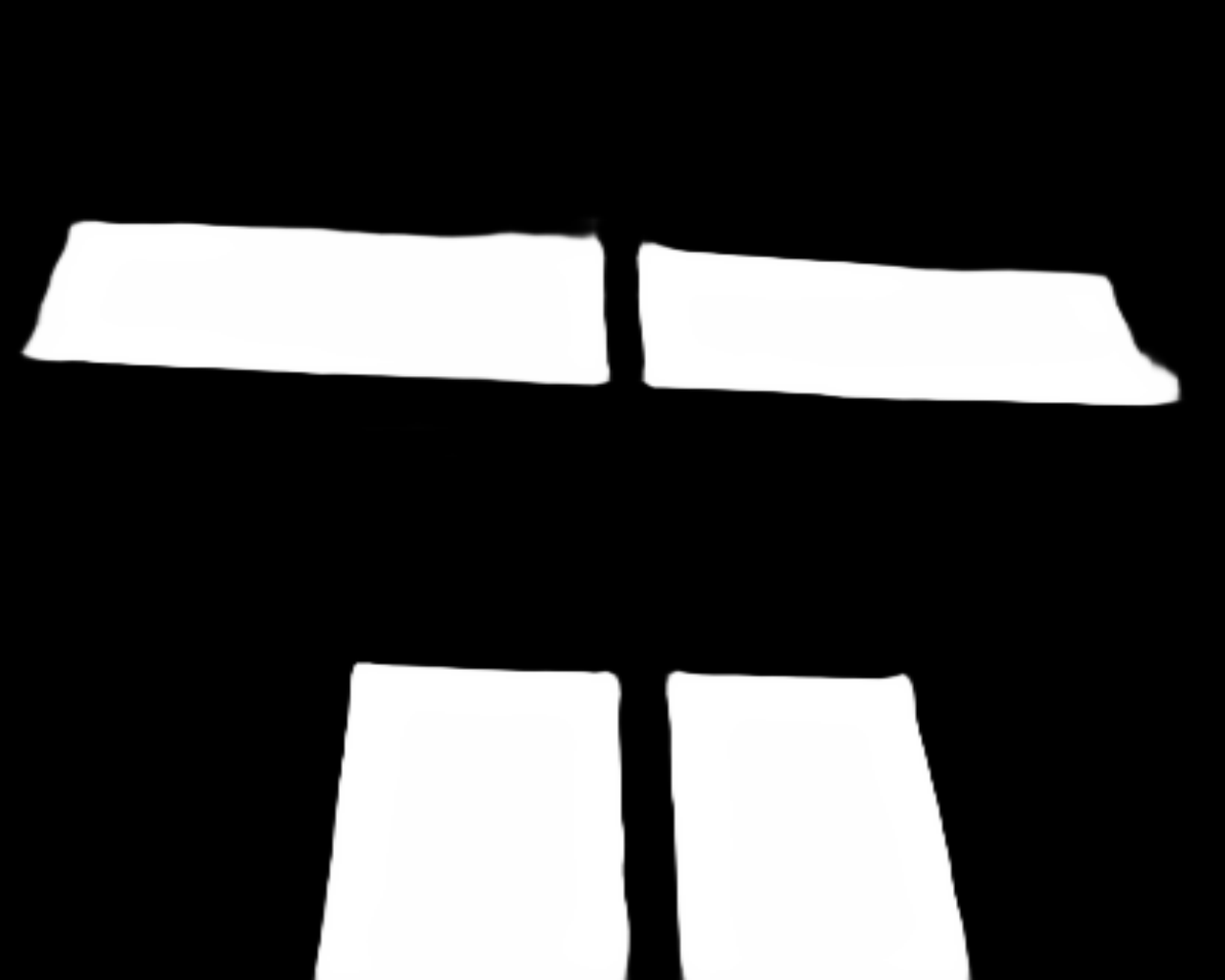}&
		\includegraphics[width=\wvisual, height=\hvisualshort]{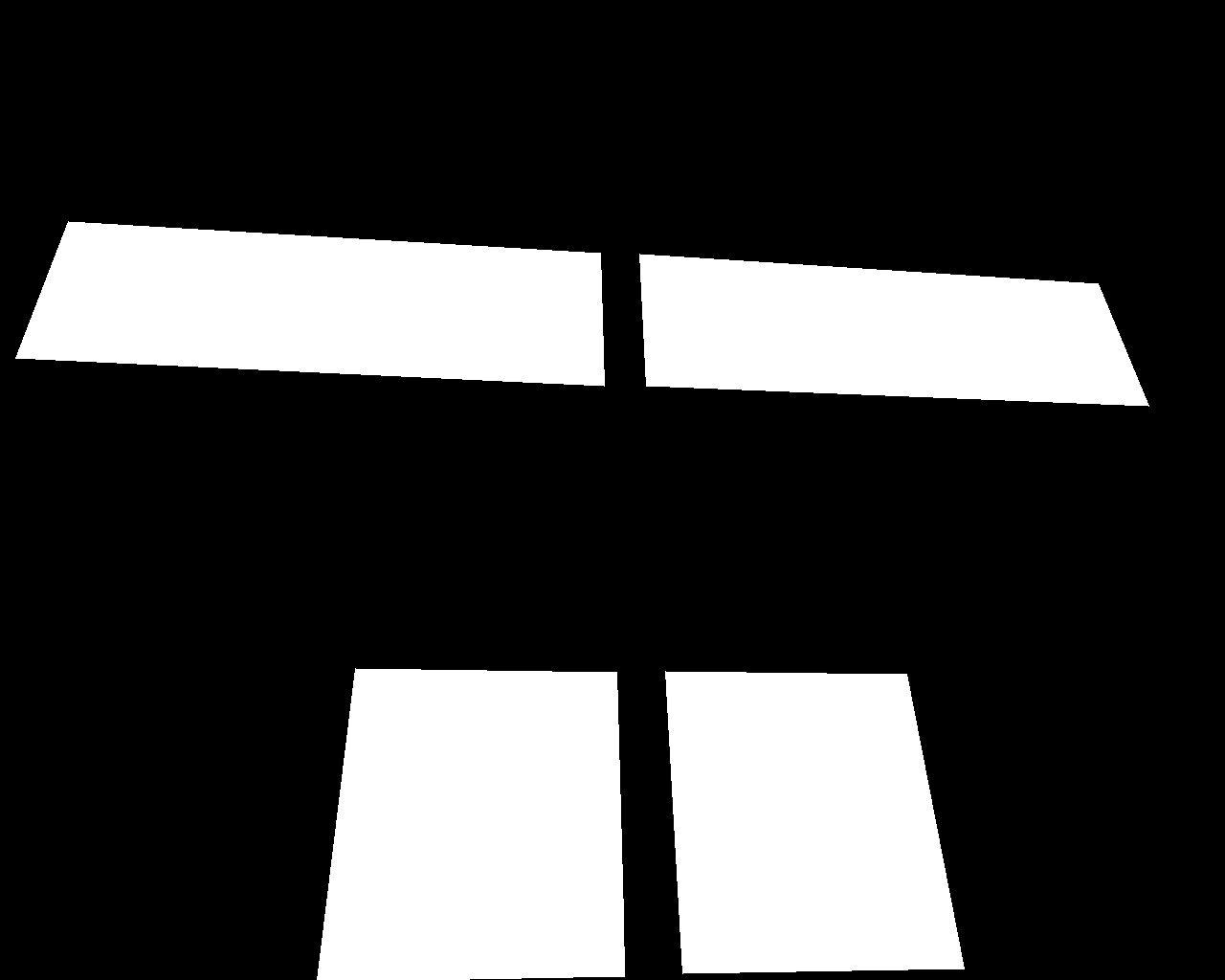} \\
		
		\includegraphics[width=\wvisual, height=\hvisualshort]{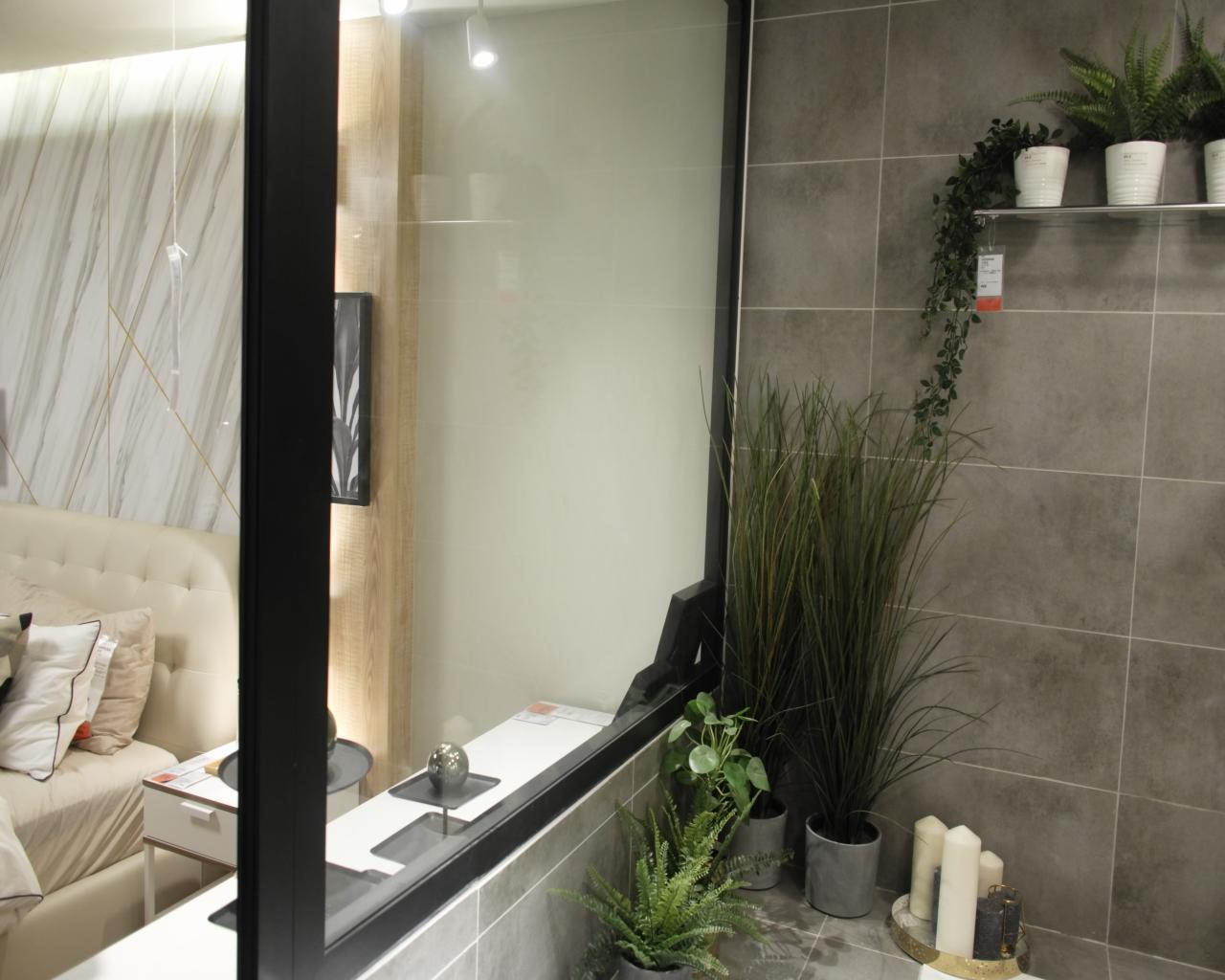}&
		\includegraphics[width=\wvisual, height=\hvisualshort]{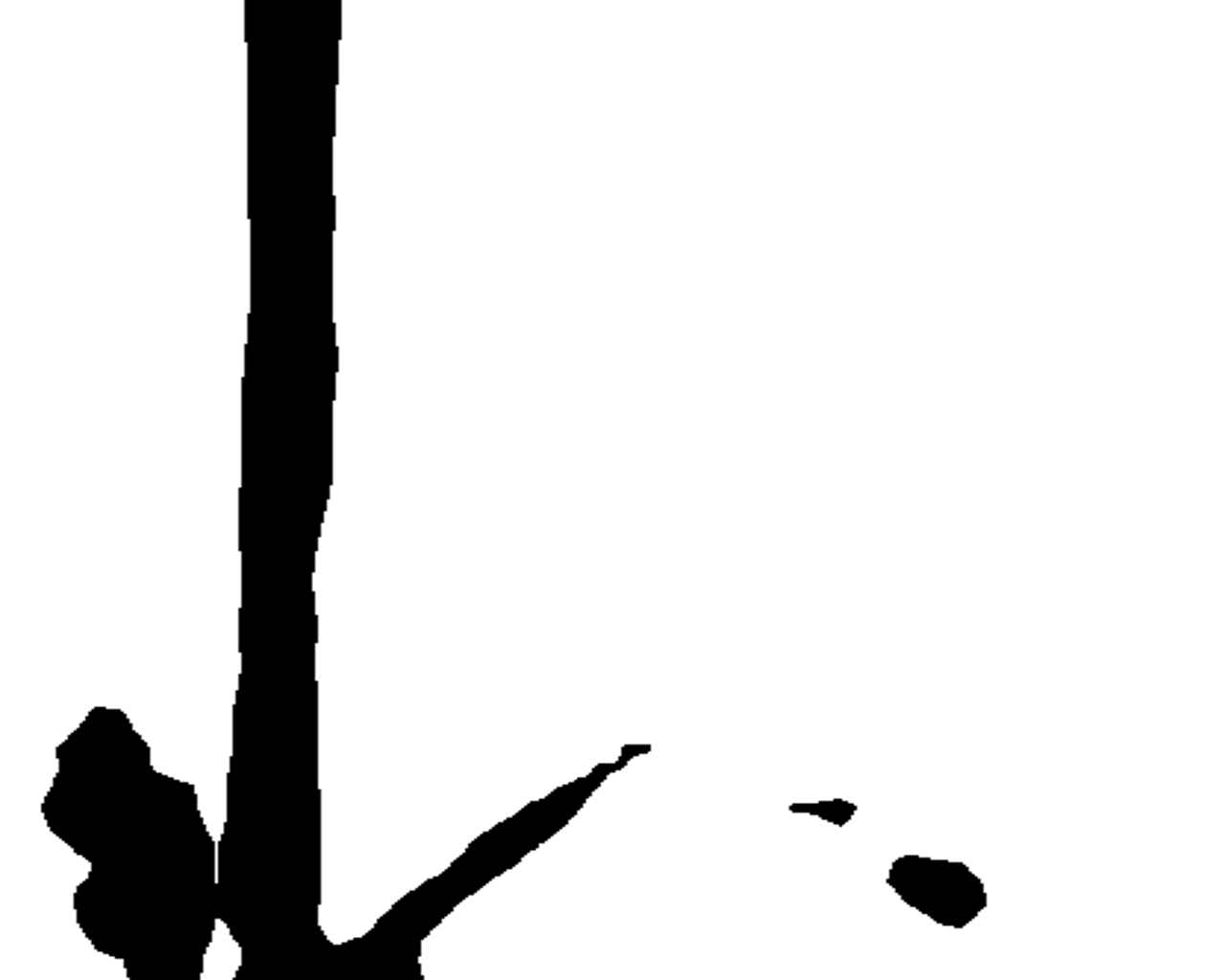}&
		\includegraphics[width=\wvisual, height=\hvisualshort]{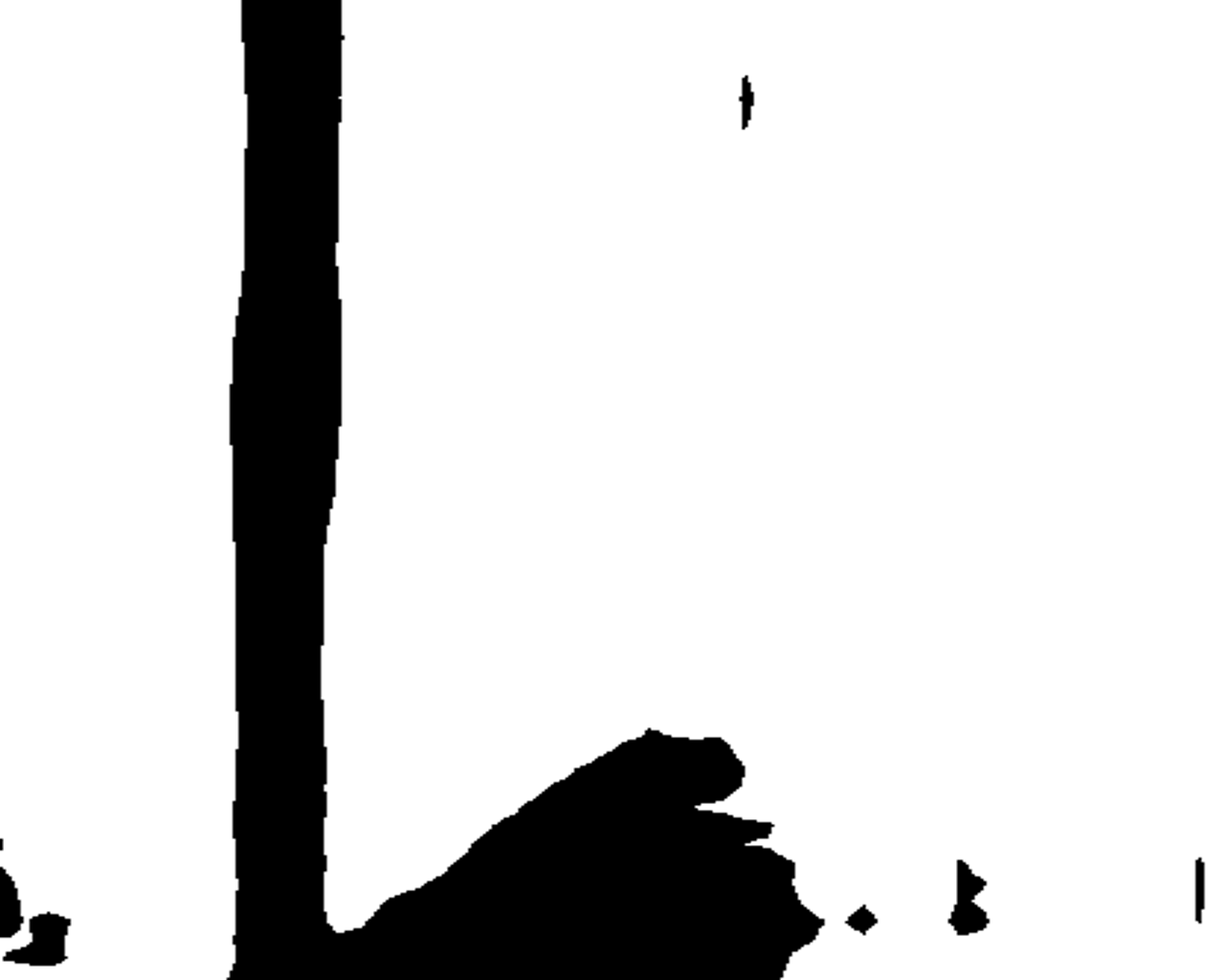}&
		\includegraphics[width=\wvisual, height=\hvisualshort]{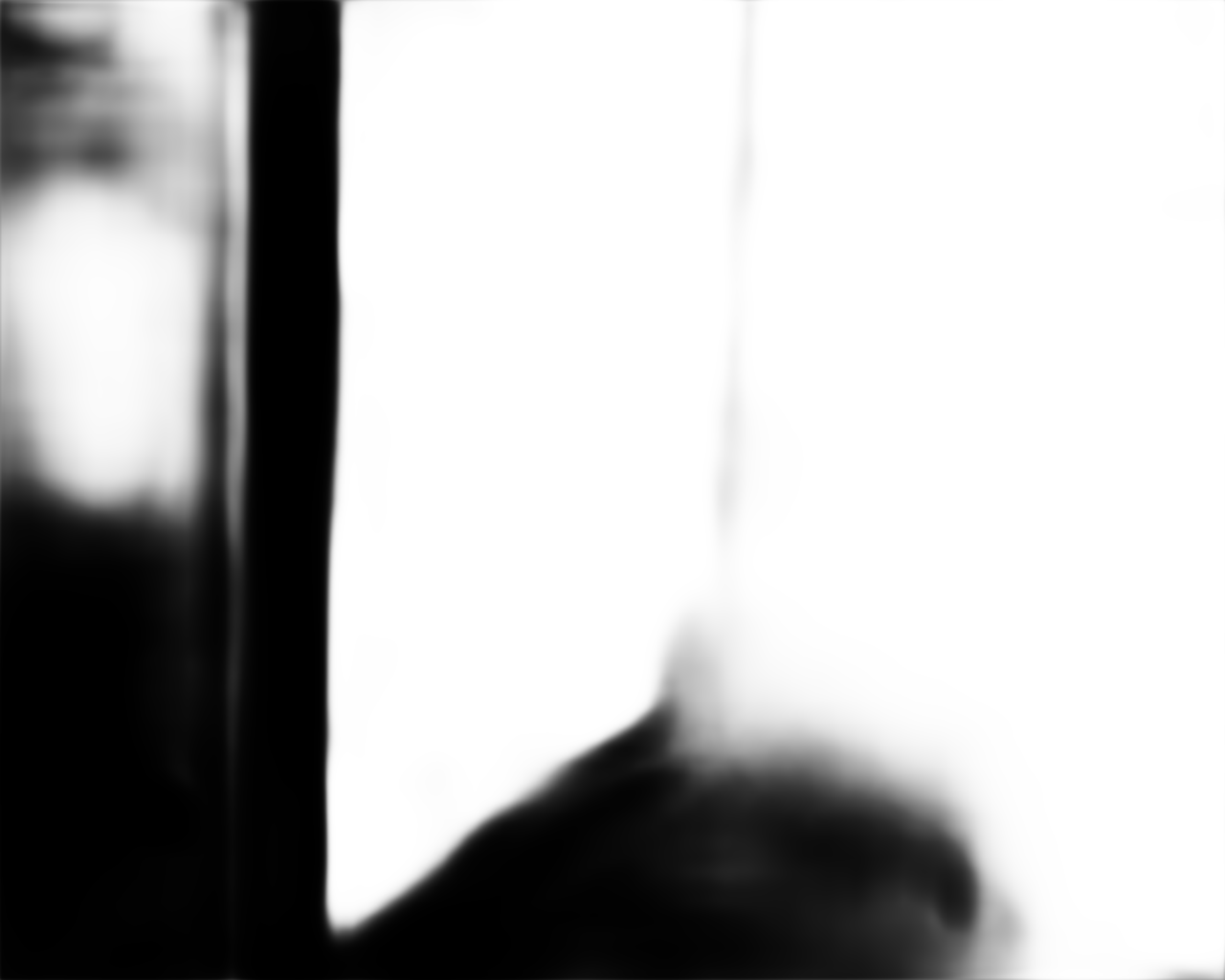}&
		\includegraphics[width=\wvisual, height=\hvisualshort]{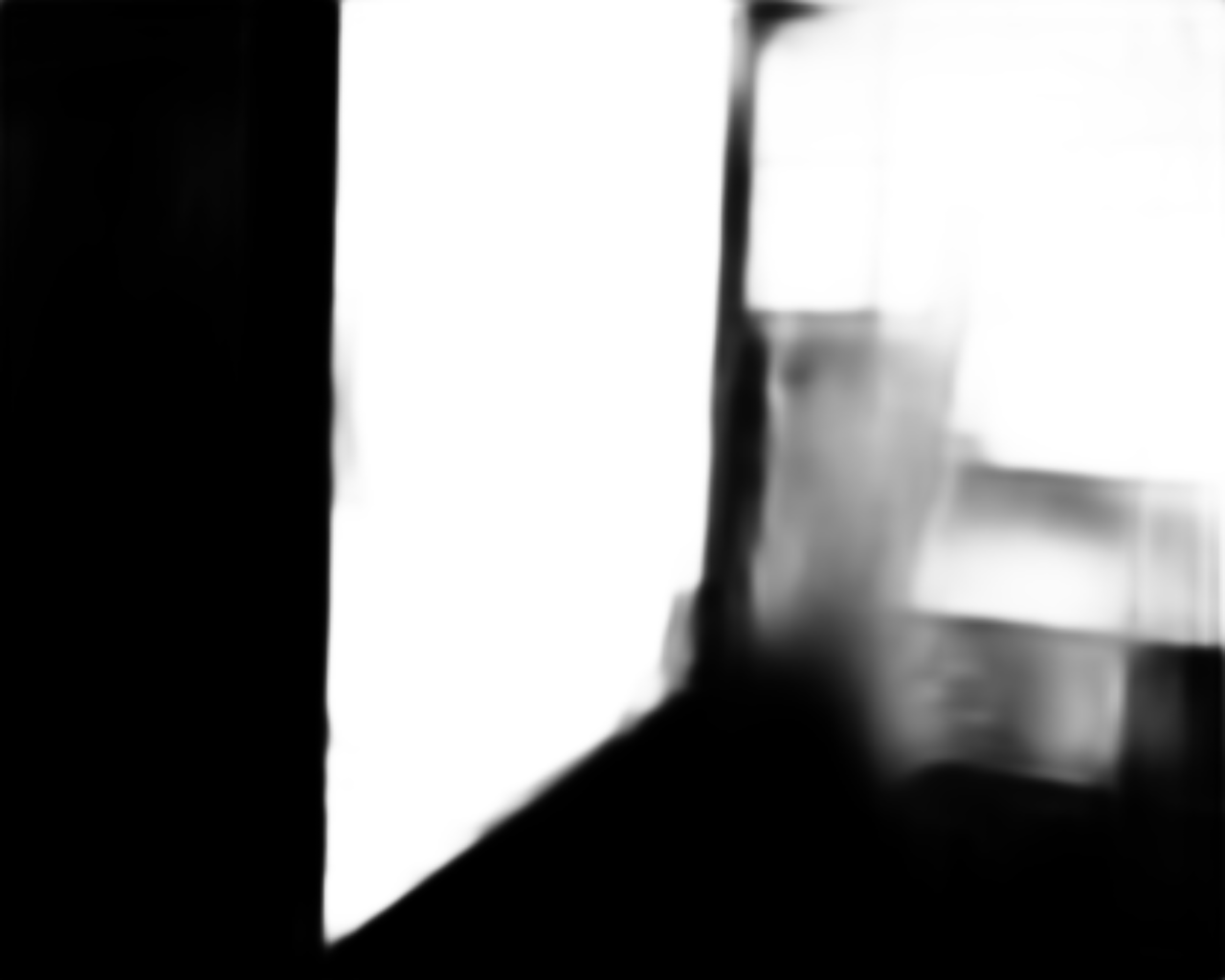}&
		\includegraphics[width=\wvisual, height=\hvisualshort]{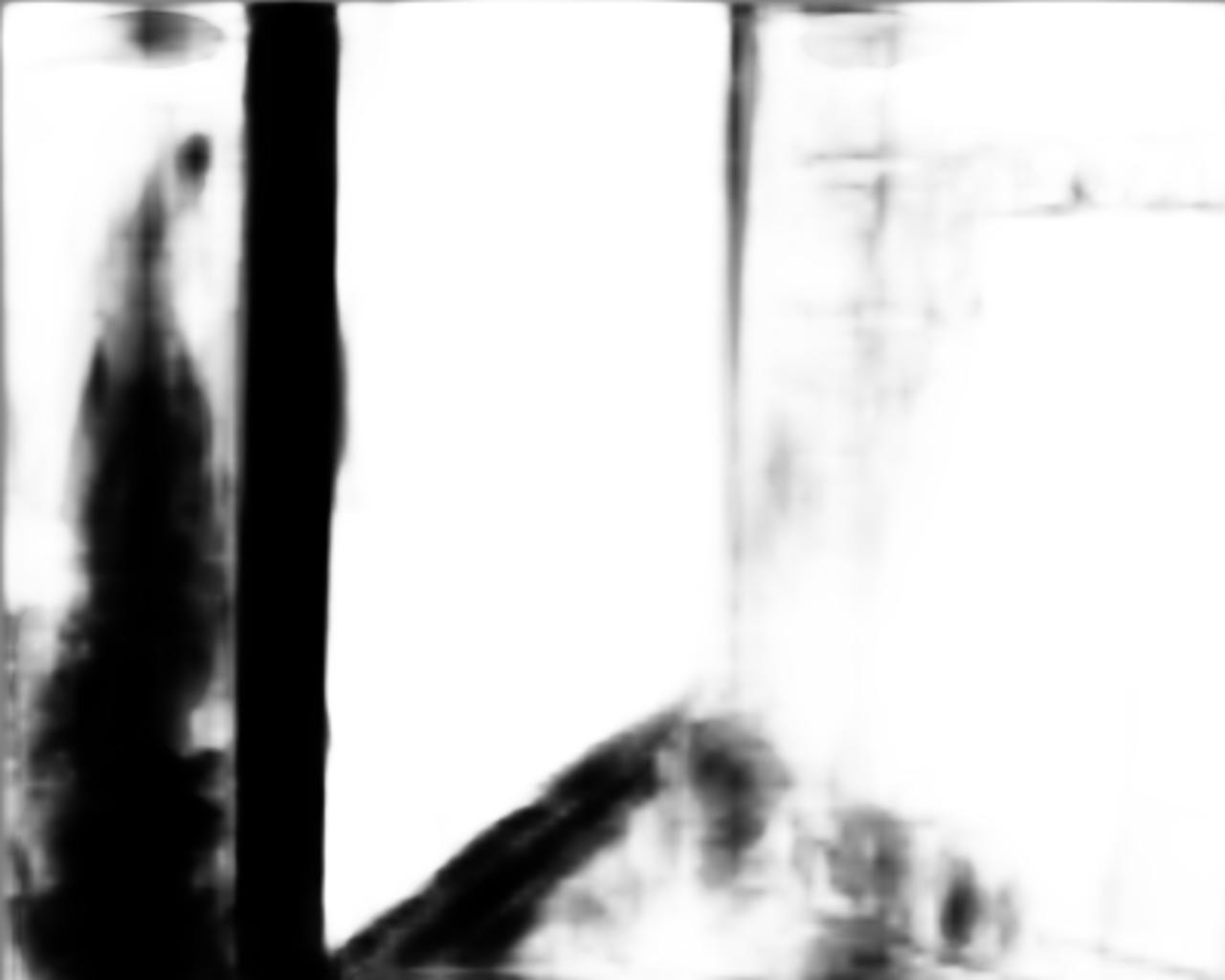}&
		\includegraphics[width=\wvisual, height=\hvisualshort]{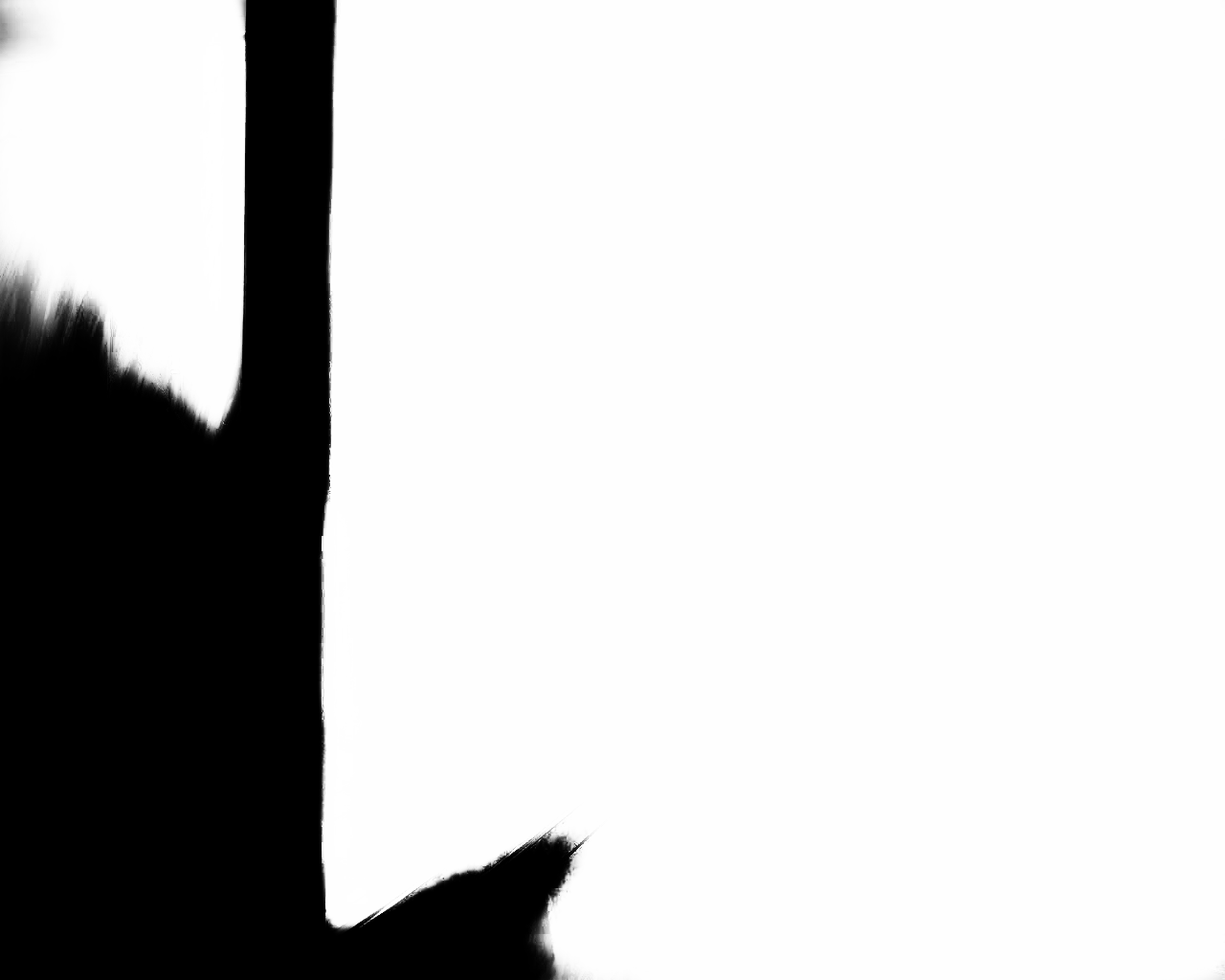}&
		\includegraphics[width=\wvisual, height=\hvisualshort]{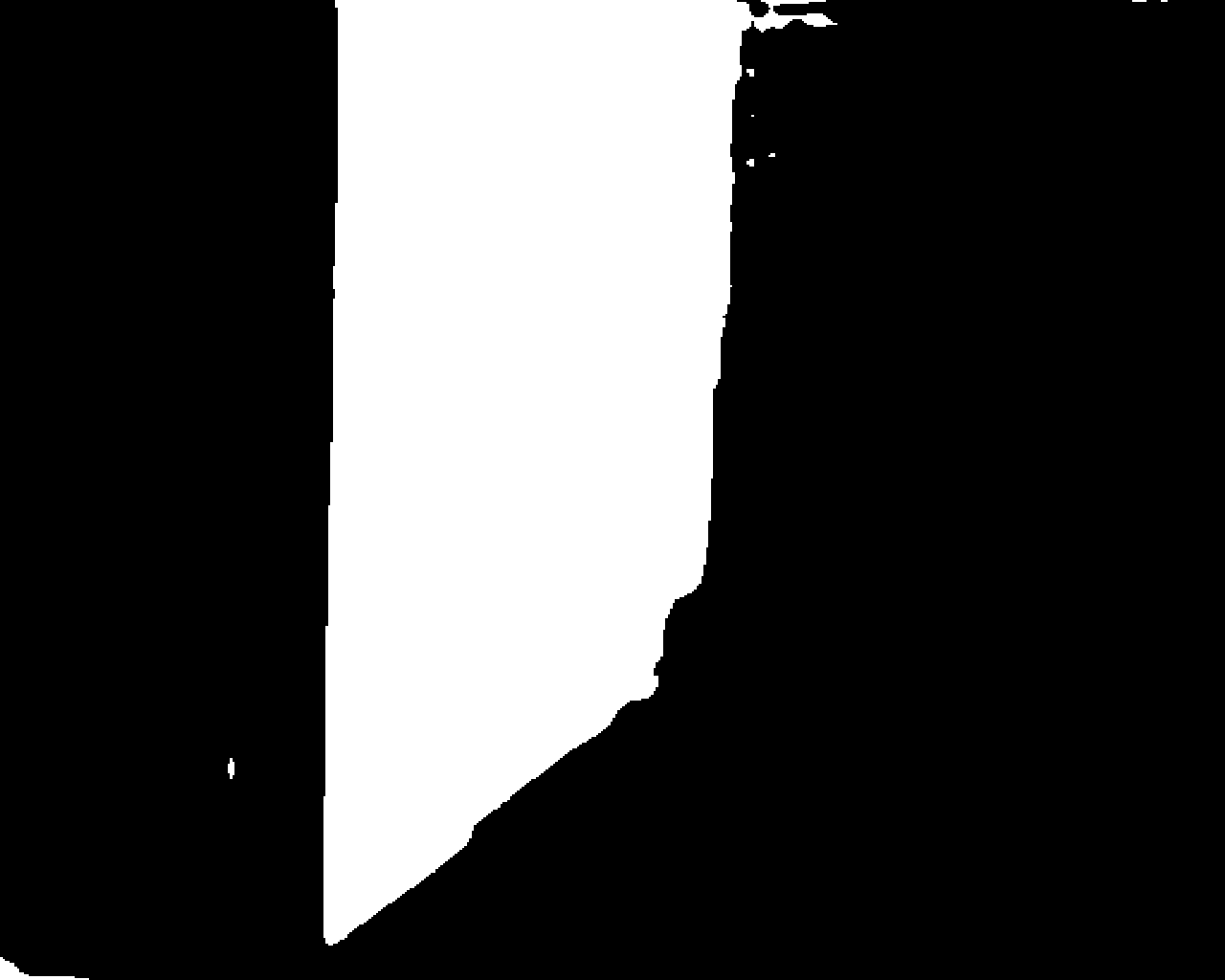}&
		\includegraphics[width=\wvisual, height=\hvisualshort]{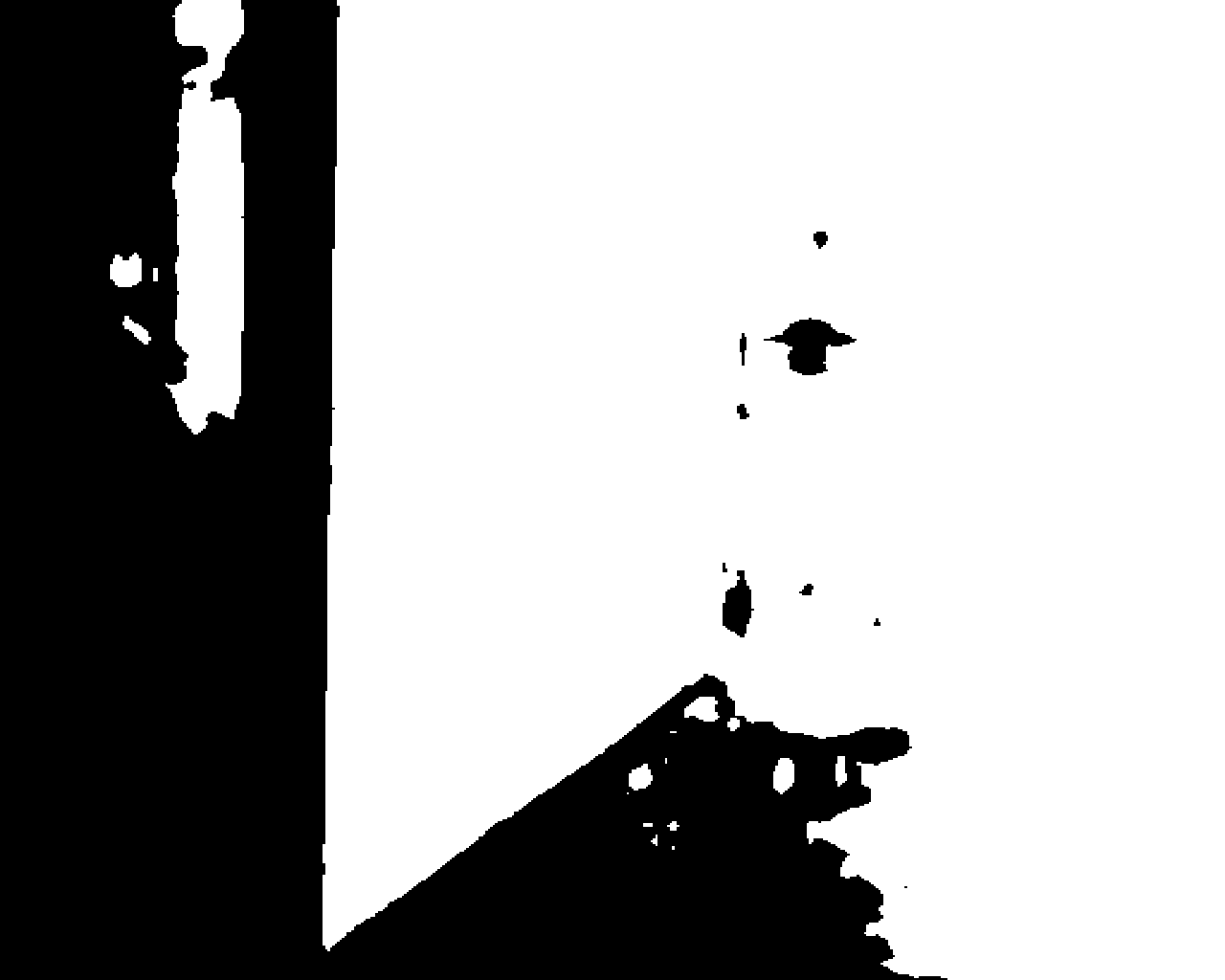}&
		\includegraphics[width=\wvisual, height=\hvisualshort]{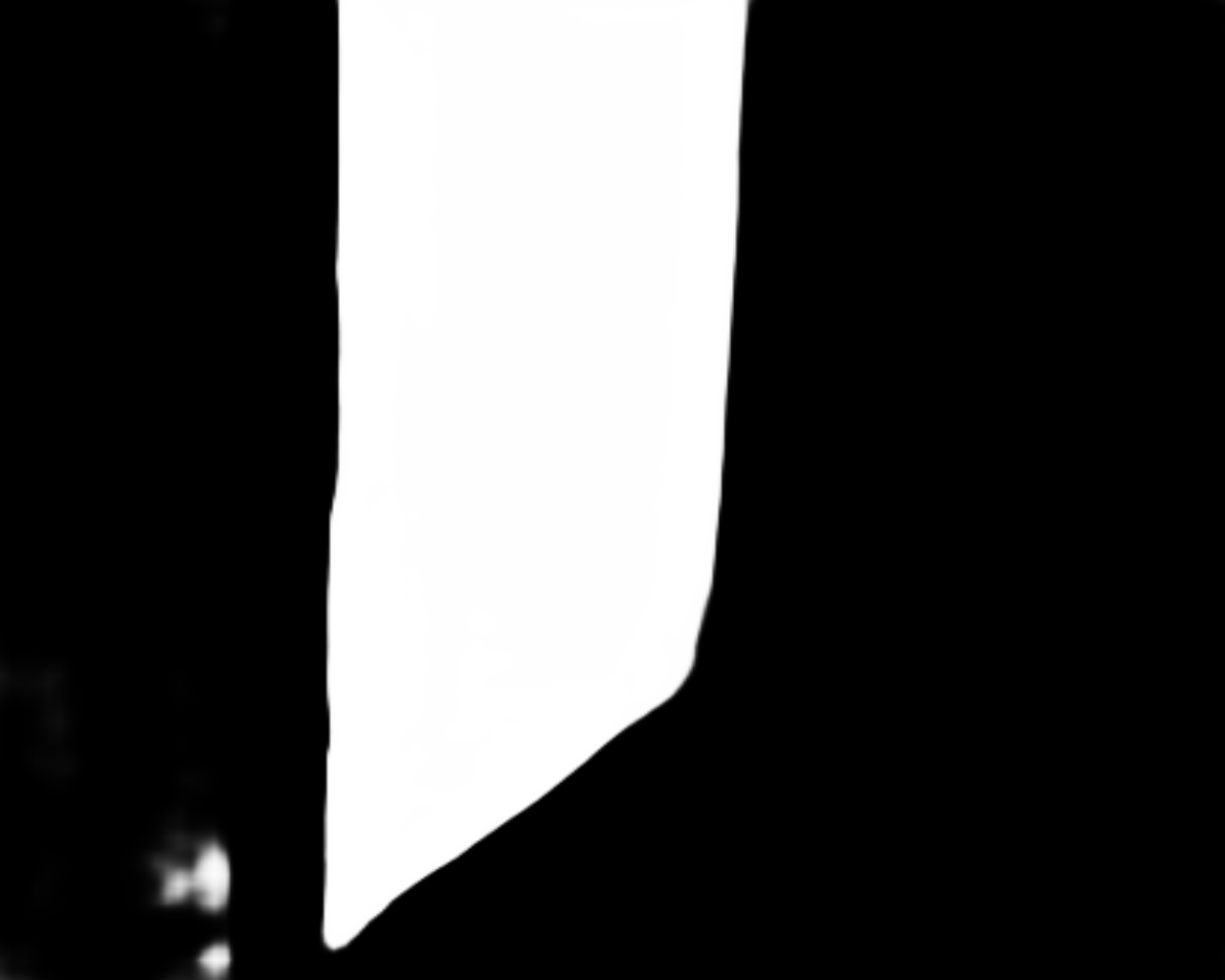}&
		\includegraphics[width=\wvisual, height=\hvisualshort]{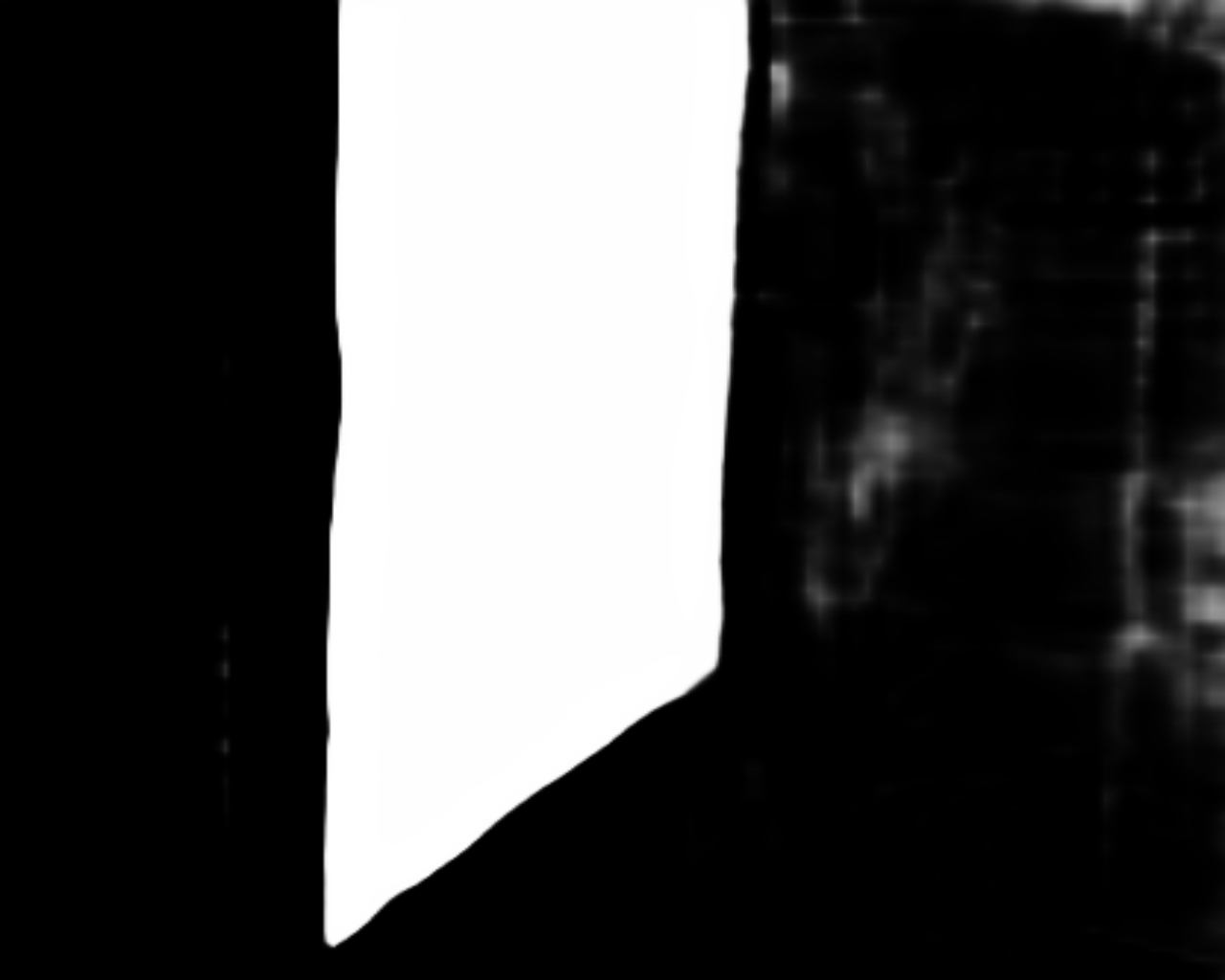}&
		\includegraphics[width=\wvisual, height=\hvisualshort]{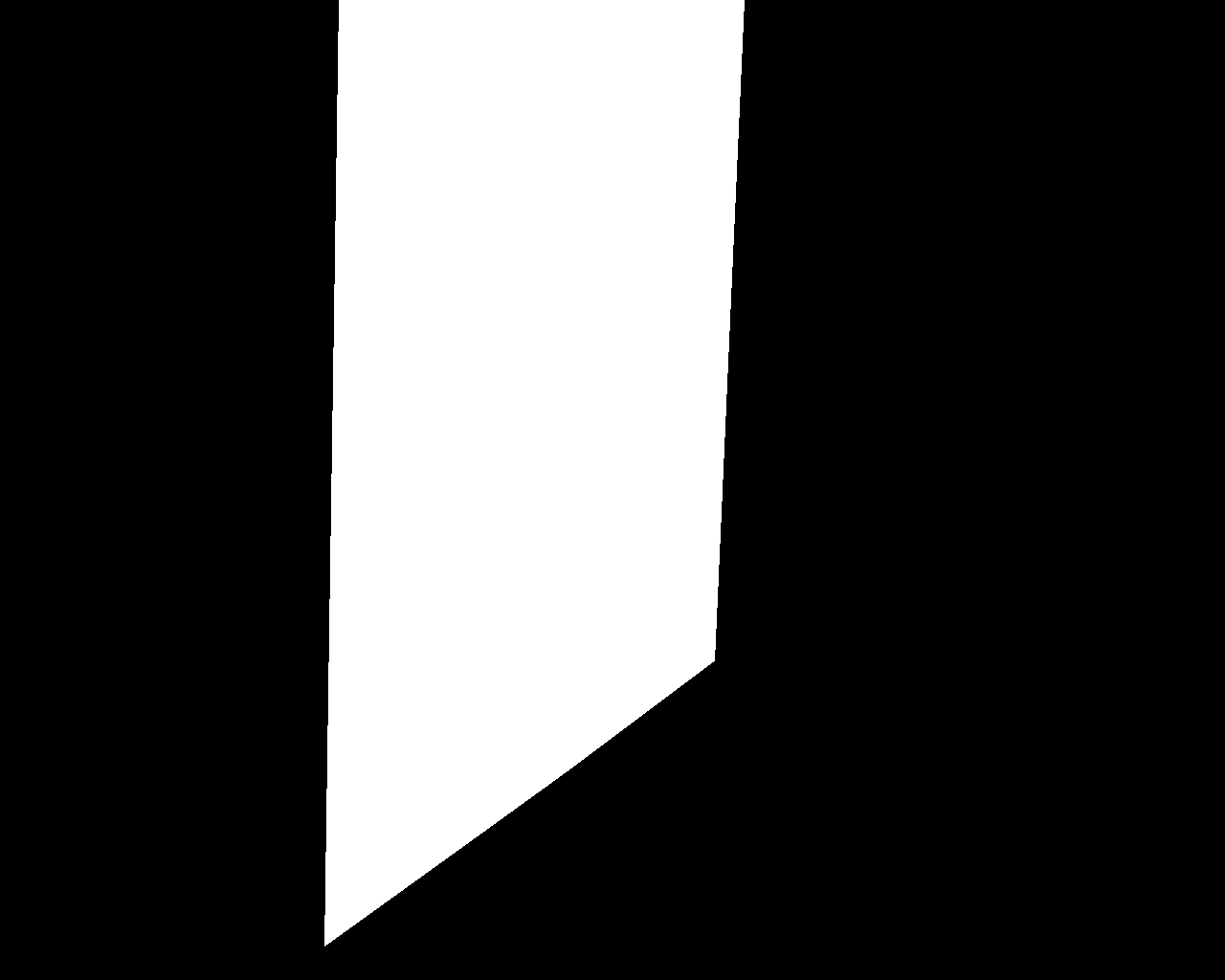} \\
		
		\includegraphics[width=\wvisual, height=\hvisualshort]{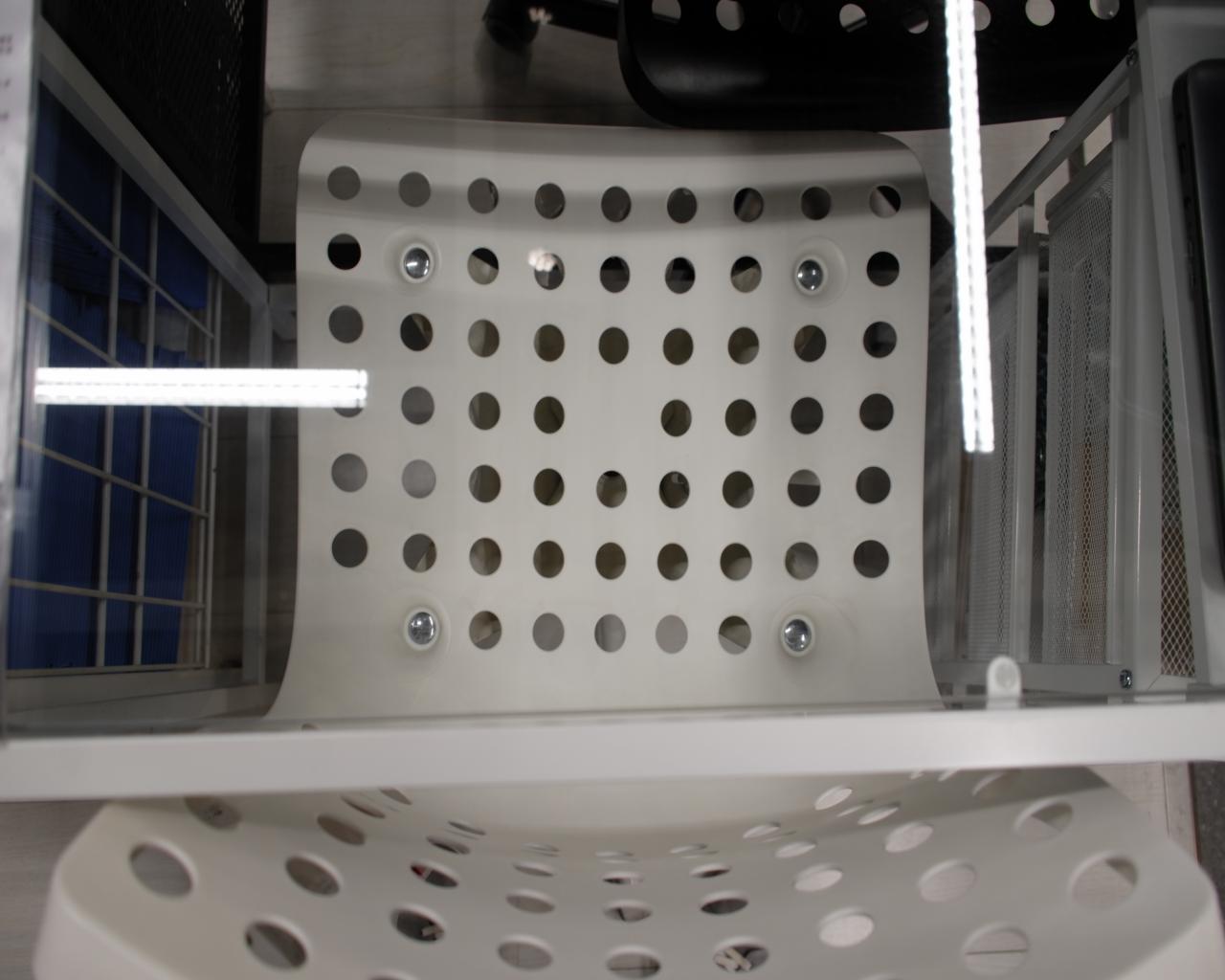}&
		\includegraphics[width=\wvisual, height=\hvisualshort]{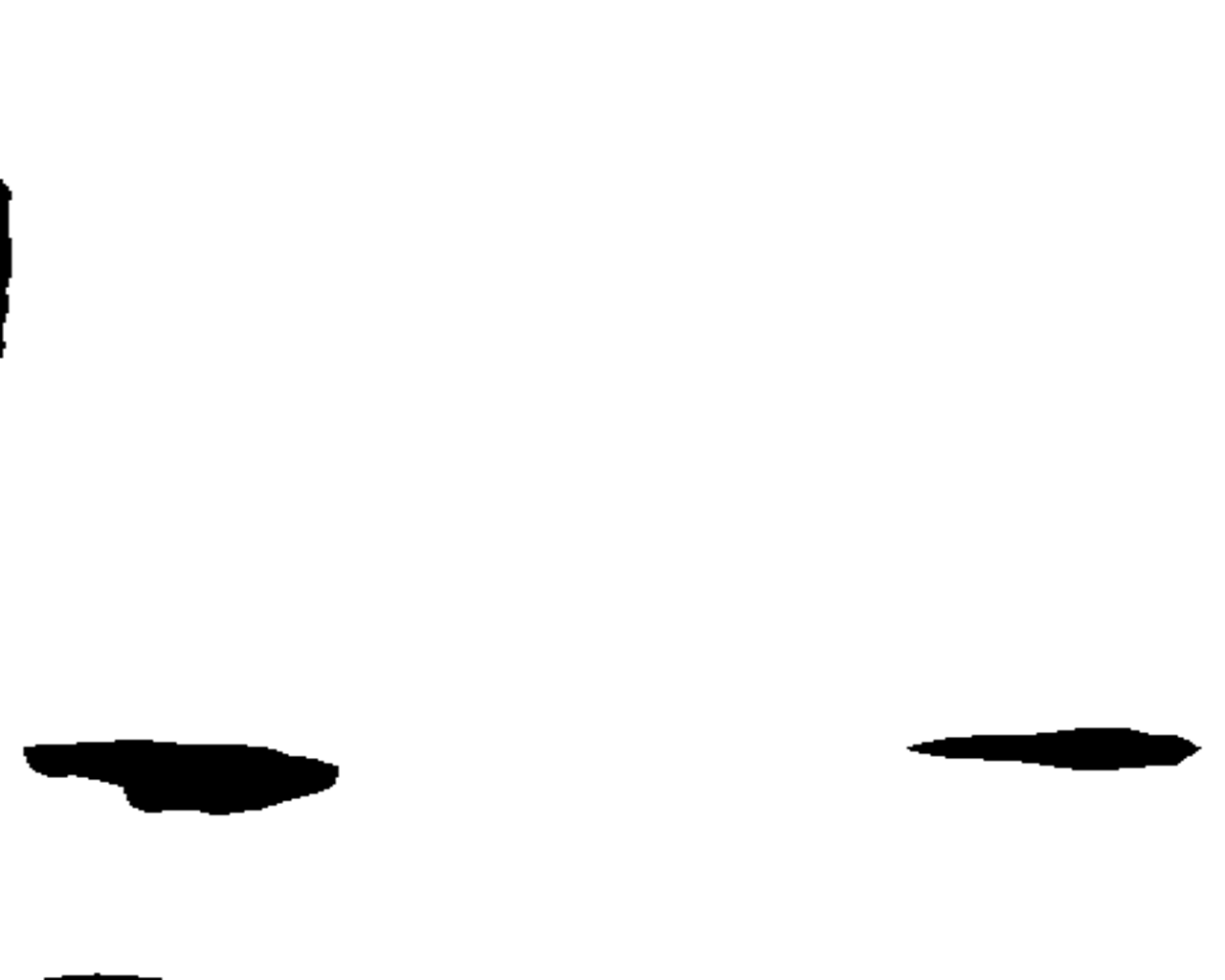}&
		\includegraphics[width=\wvisual, height=\hvisualshort]{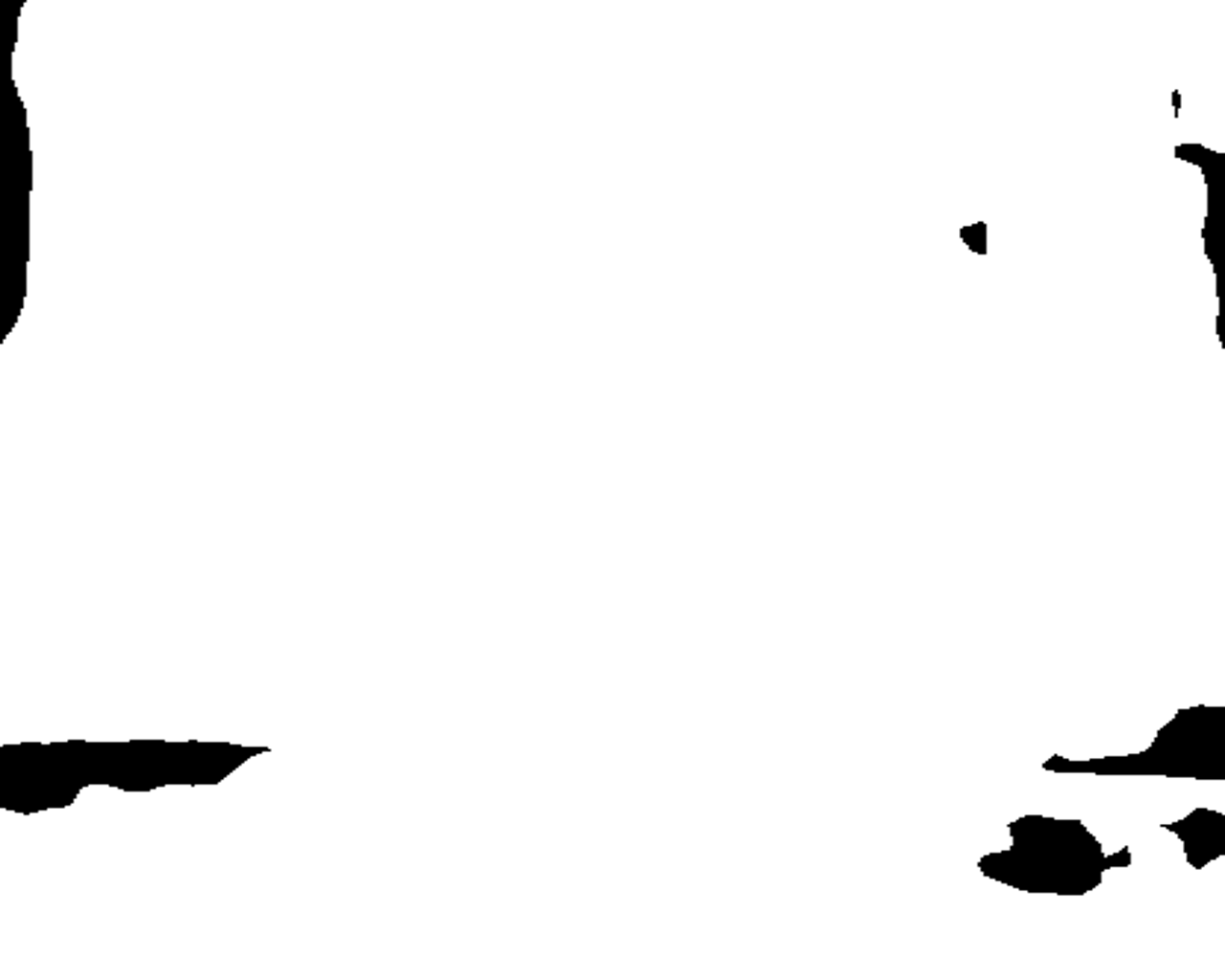}&
		\includegraphics[width=\wvisual, height=\hvisualshort]{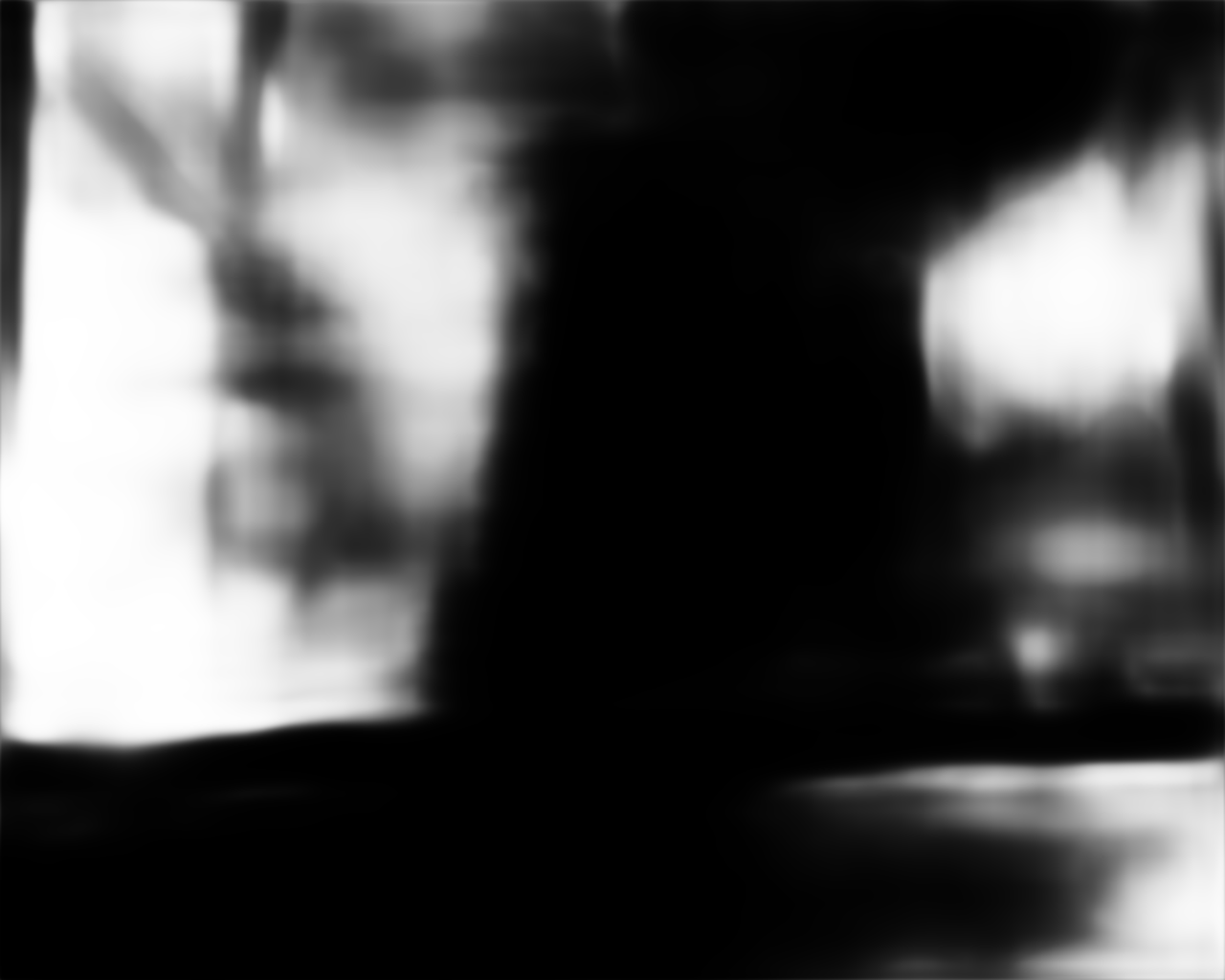}&
		\includegraphics[width=\wvisual, height=\hvisualshort]{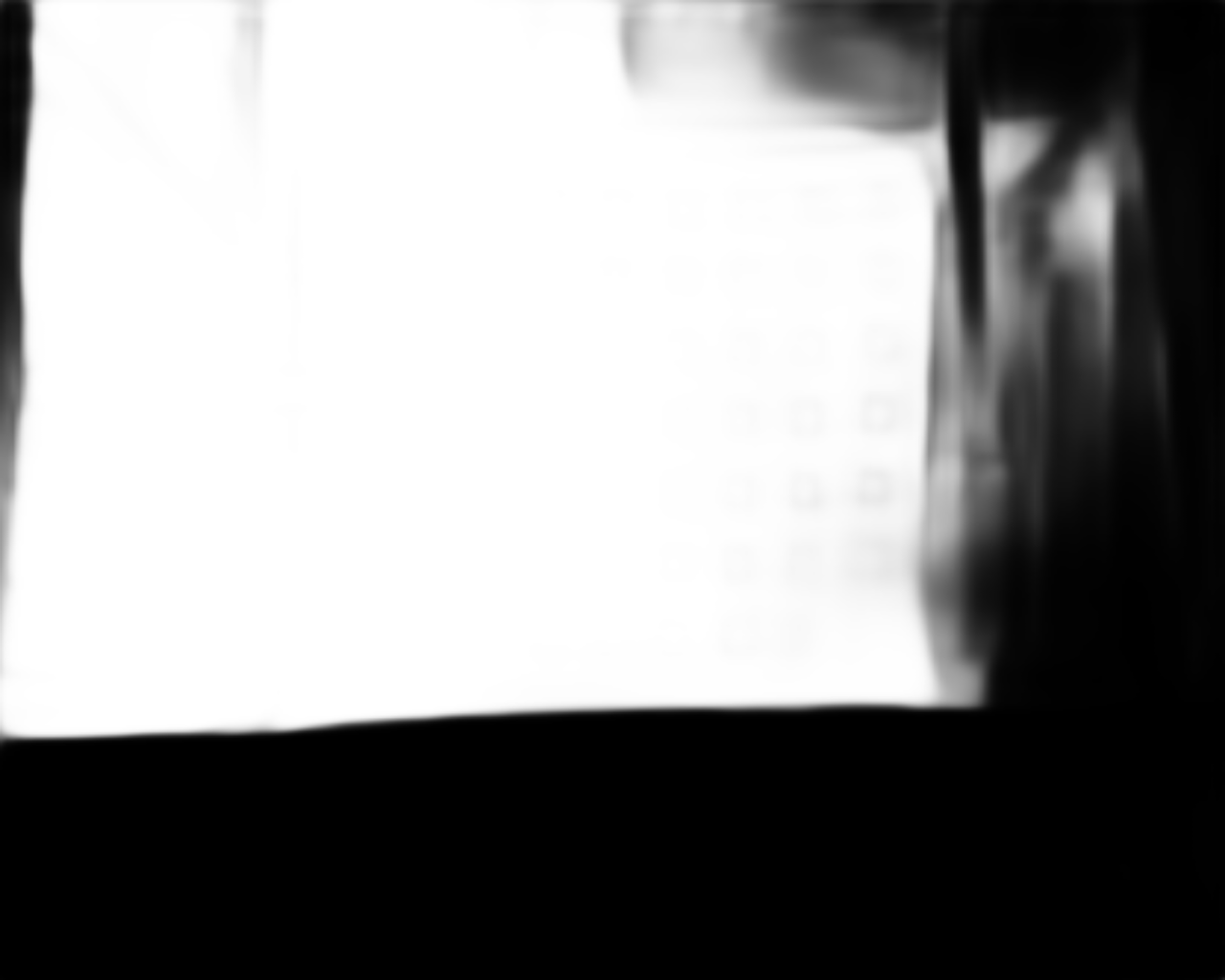}&
		\includegraphics[width=\wvisual, height=\hvisualshort]{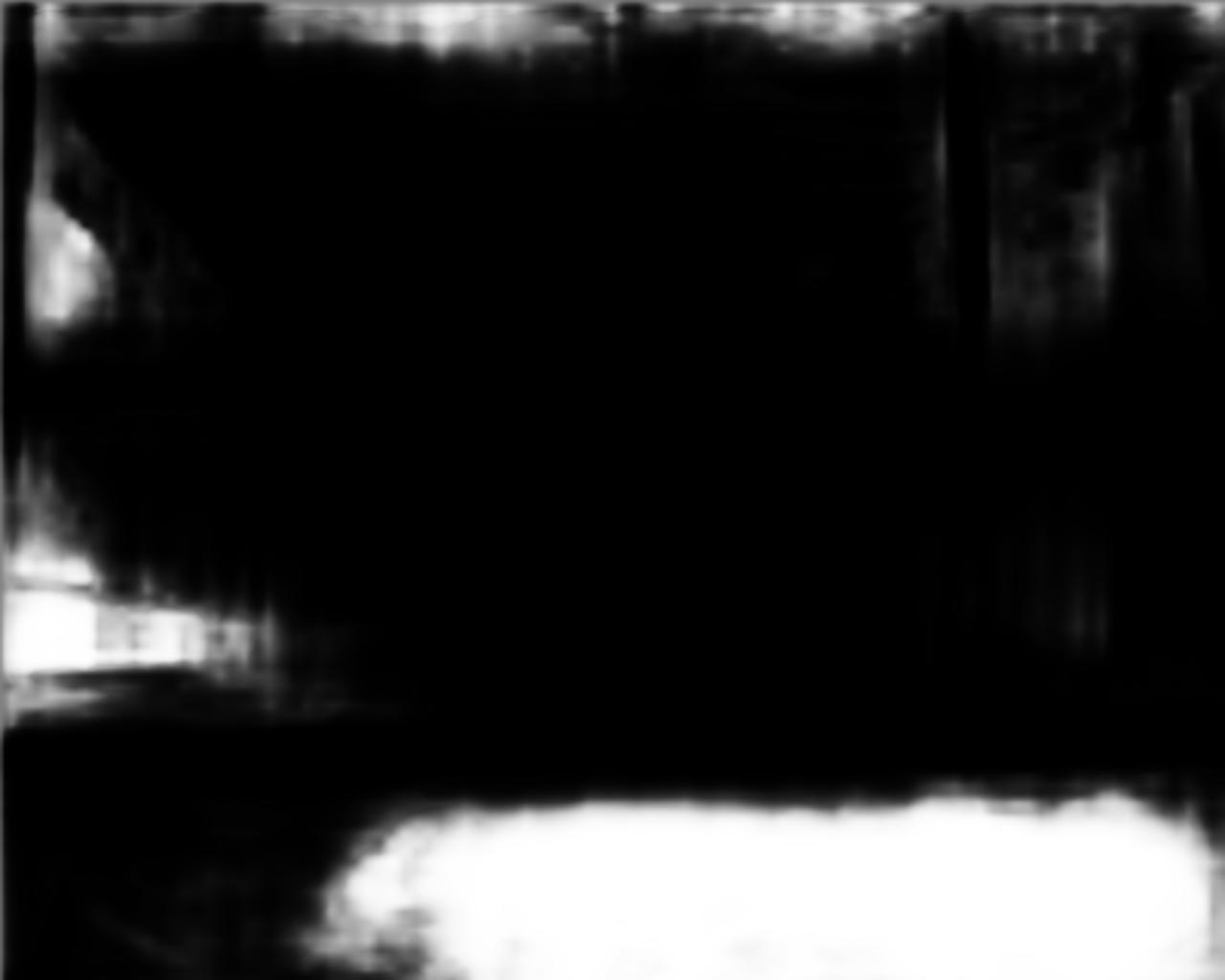}&		
		\includegraphics[width=\wvisual, height=\hvisualshort]{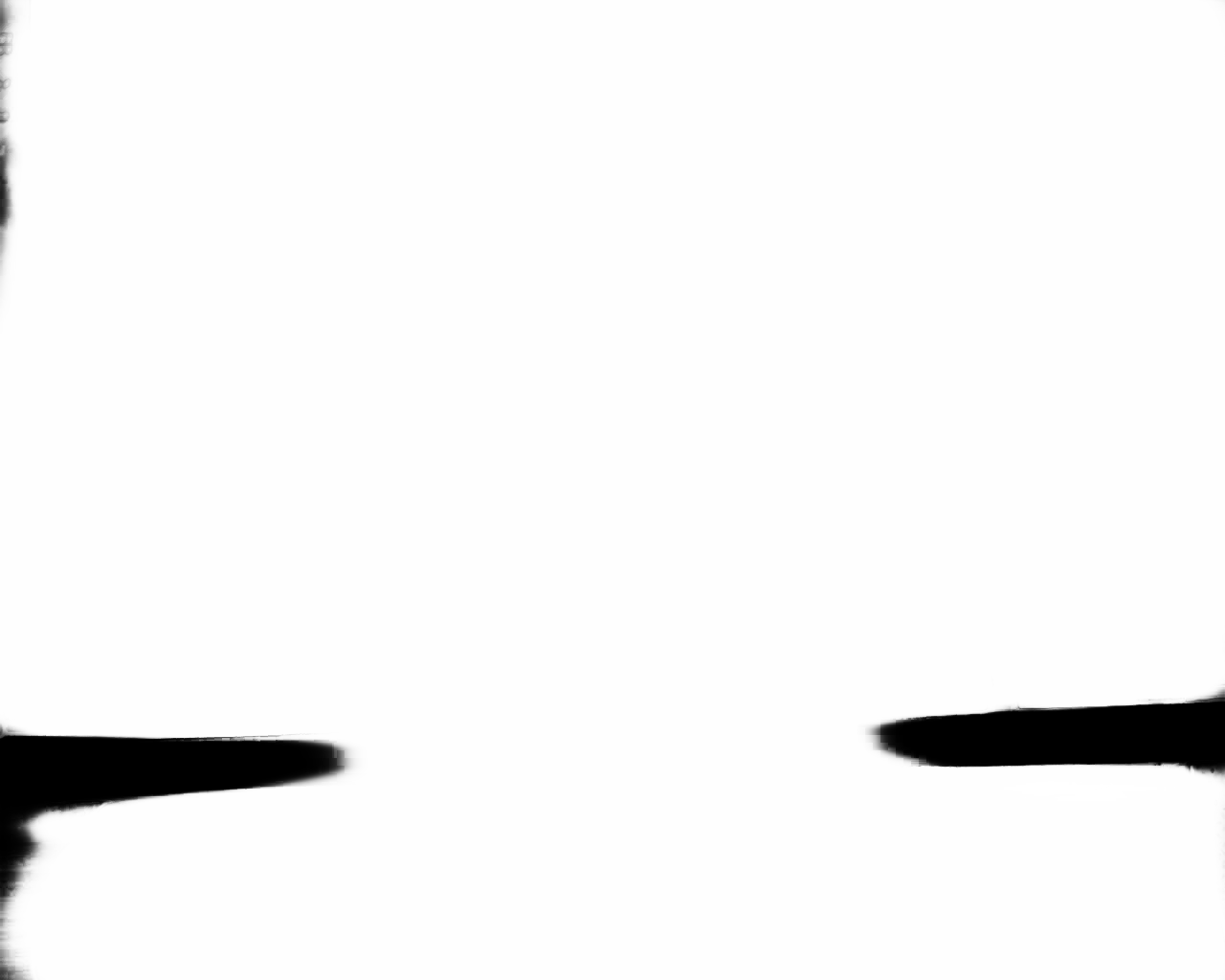}&
		\includegraphics[width=\wvisual, height=\hvisualshort]{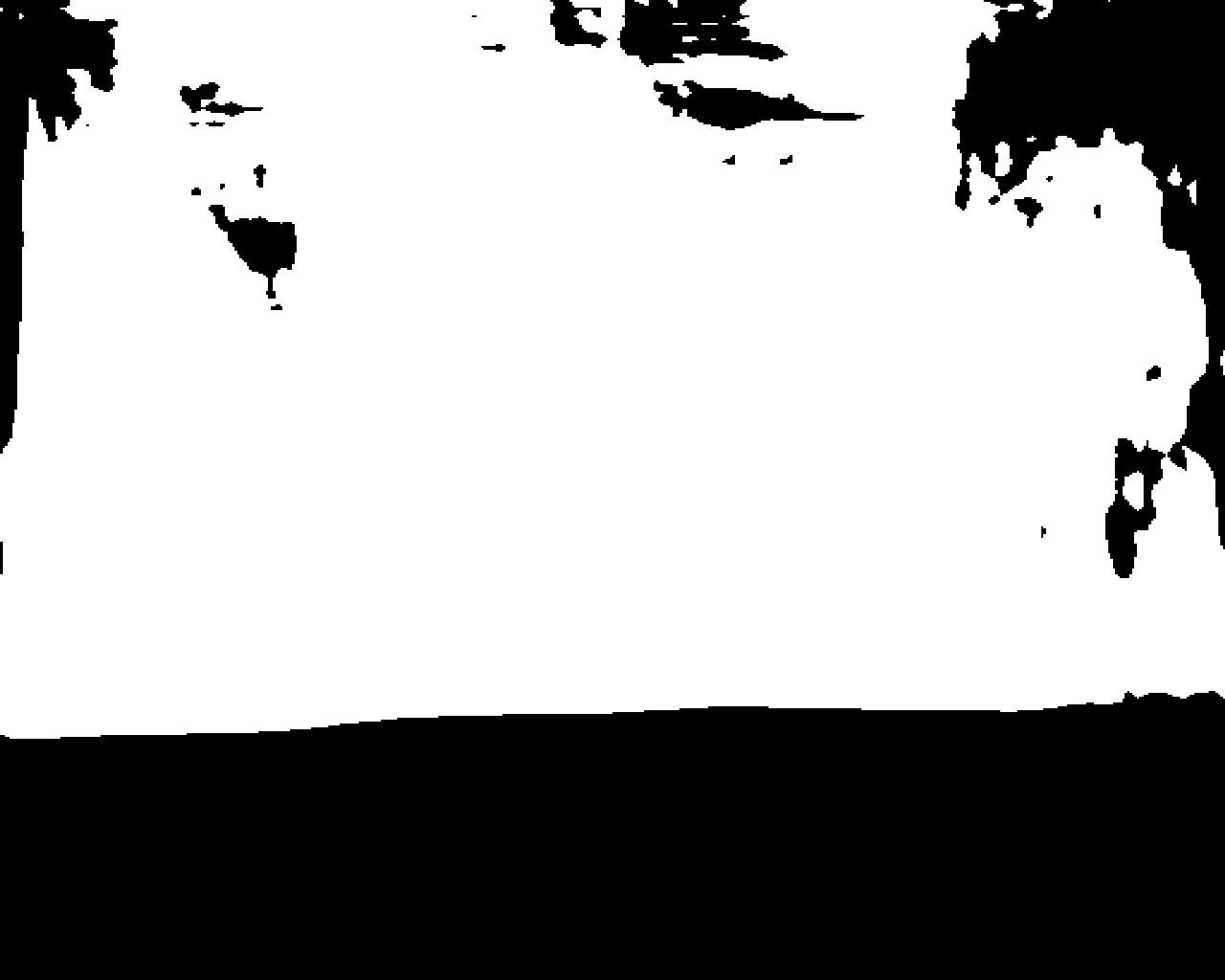}&
		\includegraphics[width=\wvisual, height=\hvisualshort]{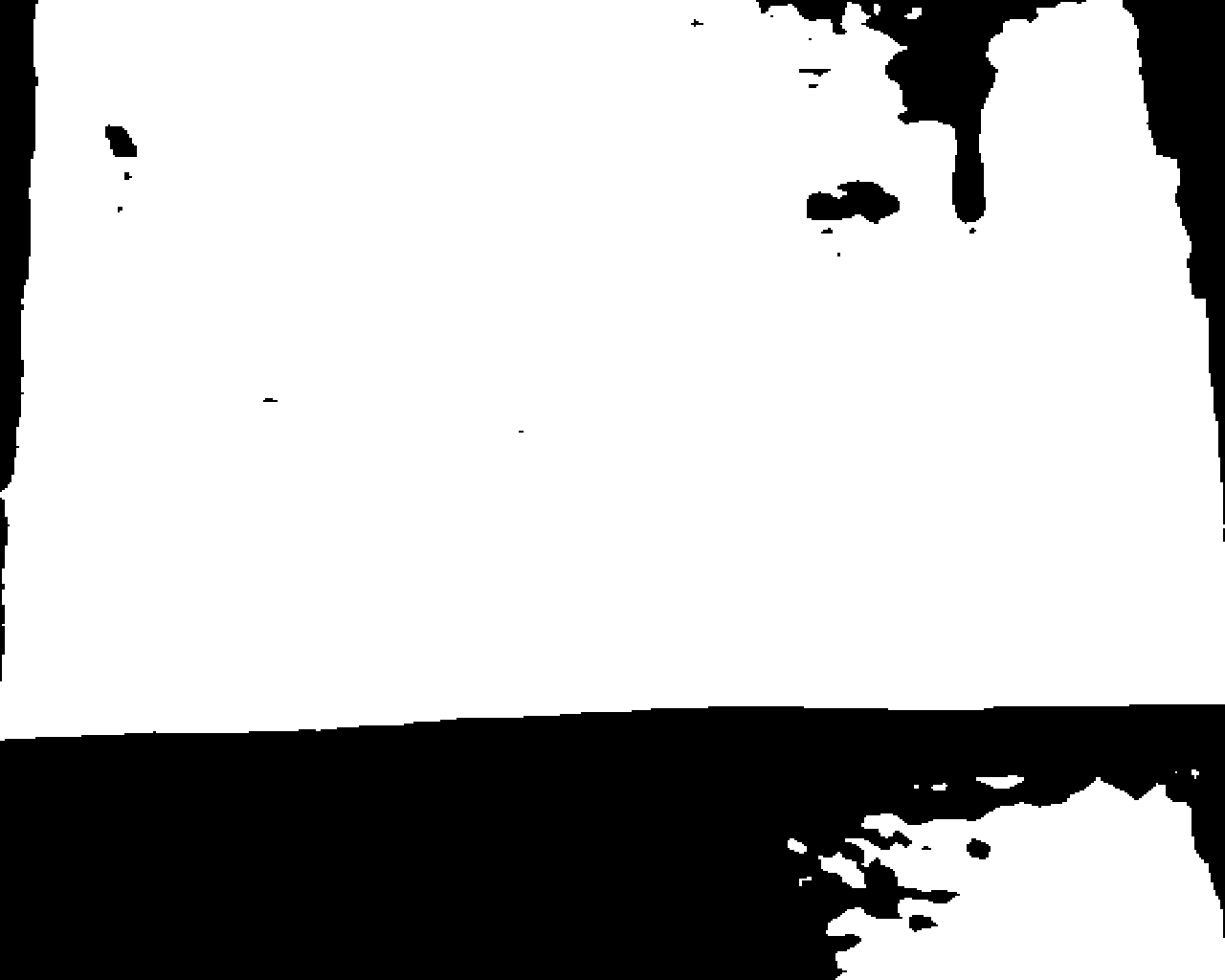}&
		\includegraphics[width=\wvisual, height=\hvisualshort]{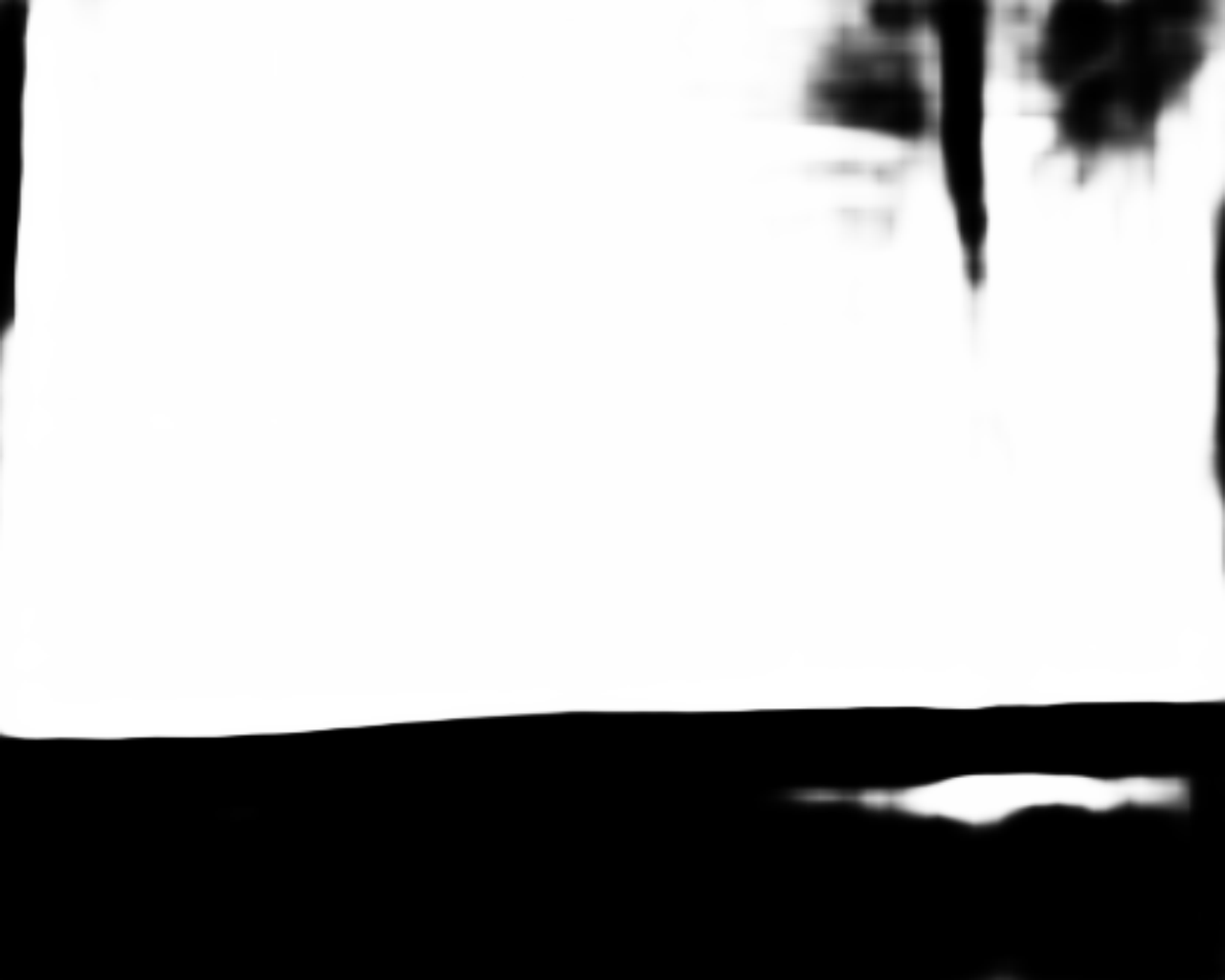}&
		\includegraphics[width=\wvisual, height=\hvisualshort]{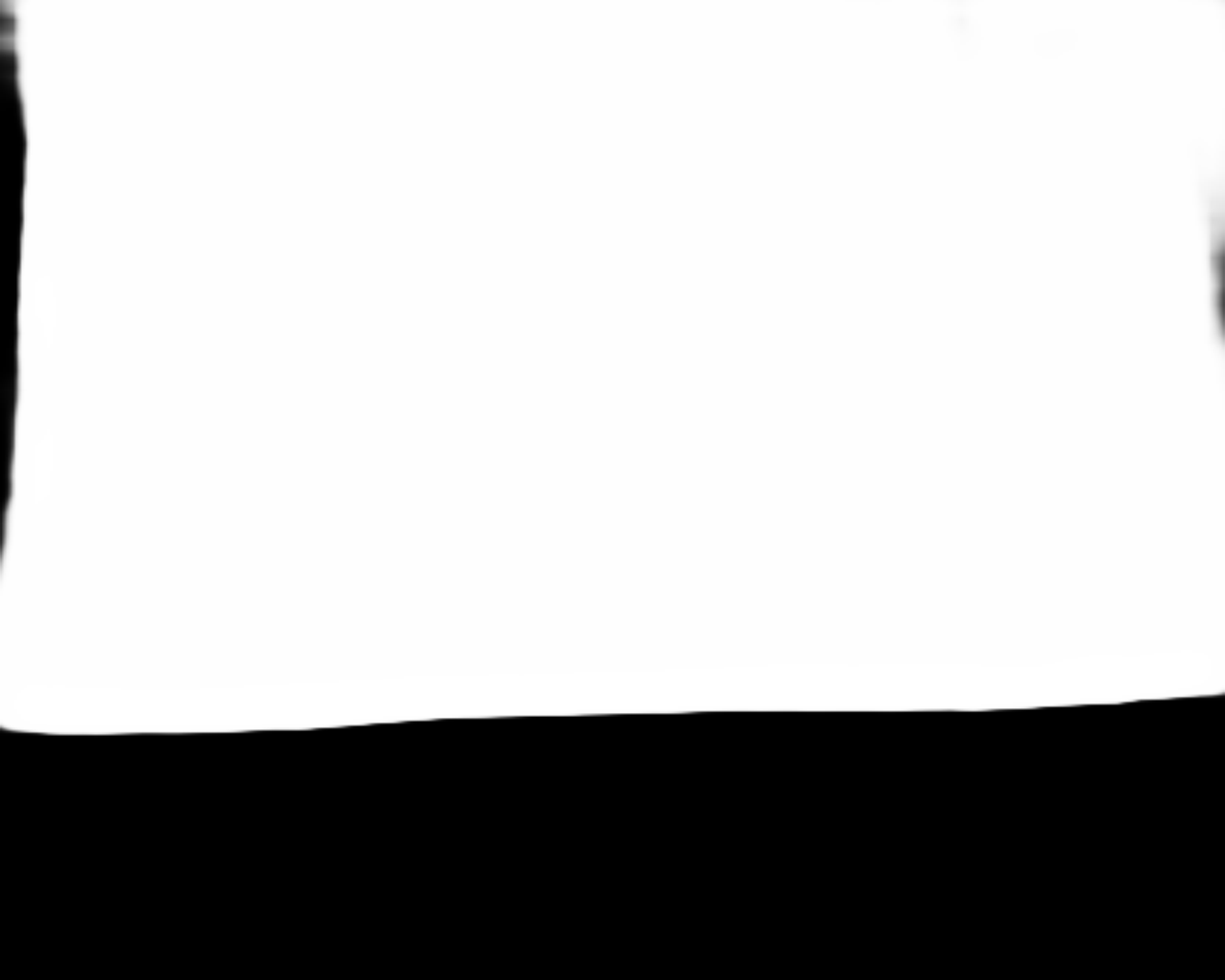}&
		\includegraphics[width=\wvisual, height=\hvisualshort]{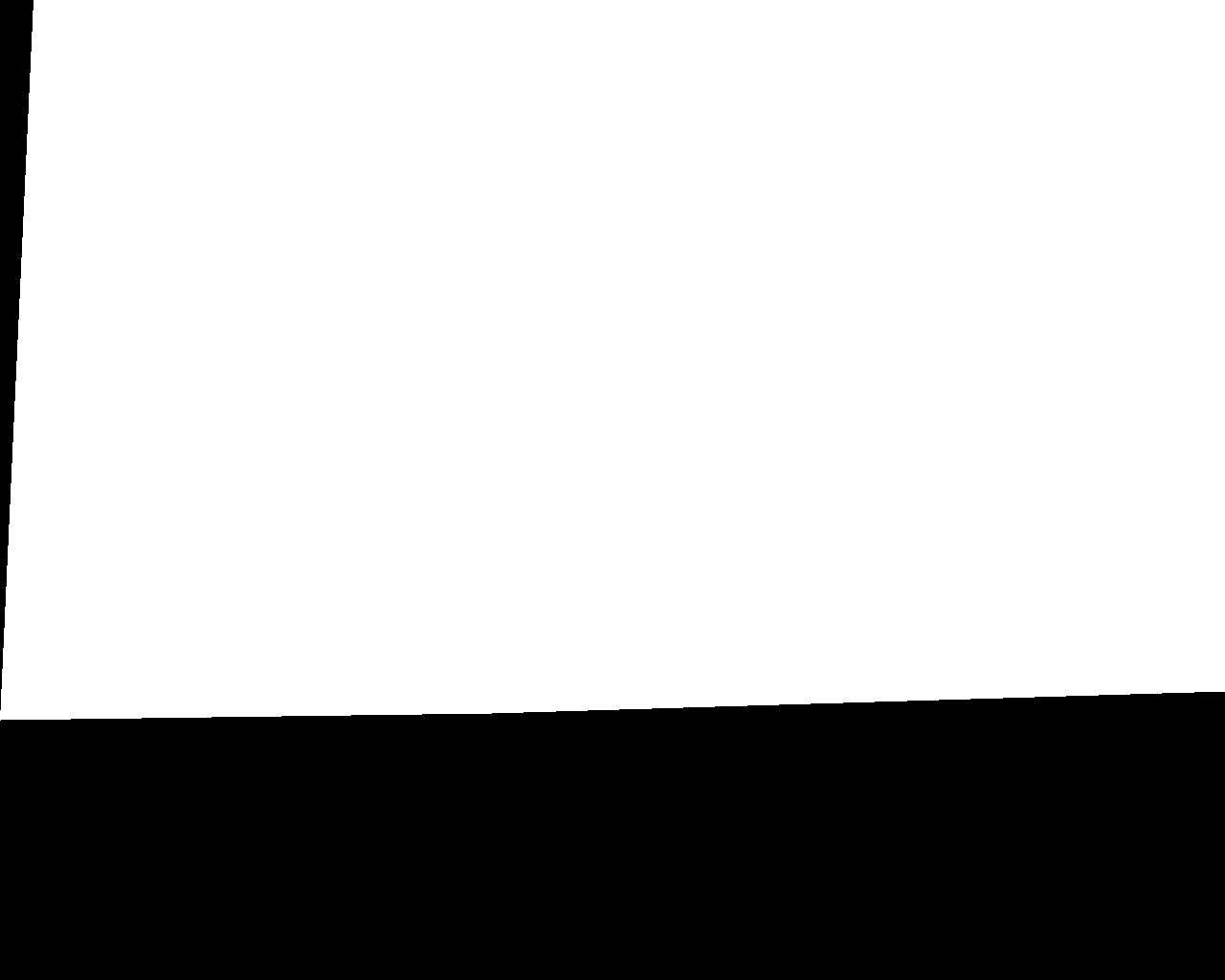} \\
		
		\includegraphics[width=\wvisual, height=\hvisualshort]{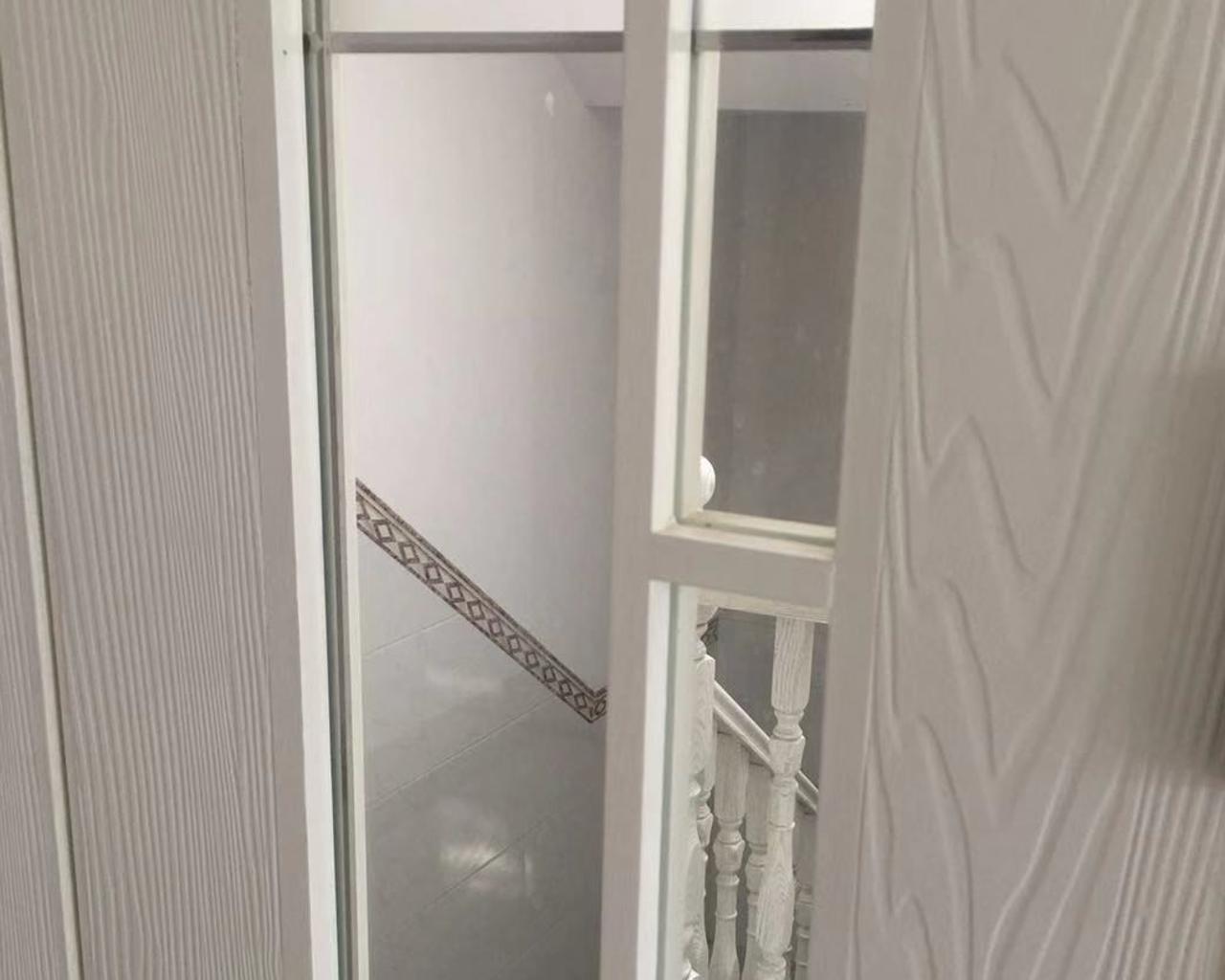}&
		\includegraphics[width=\wvisual, height=\hvisualshort]{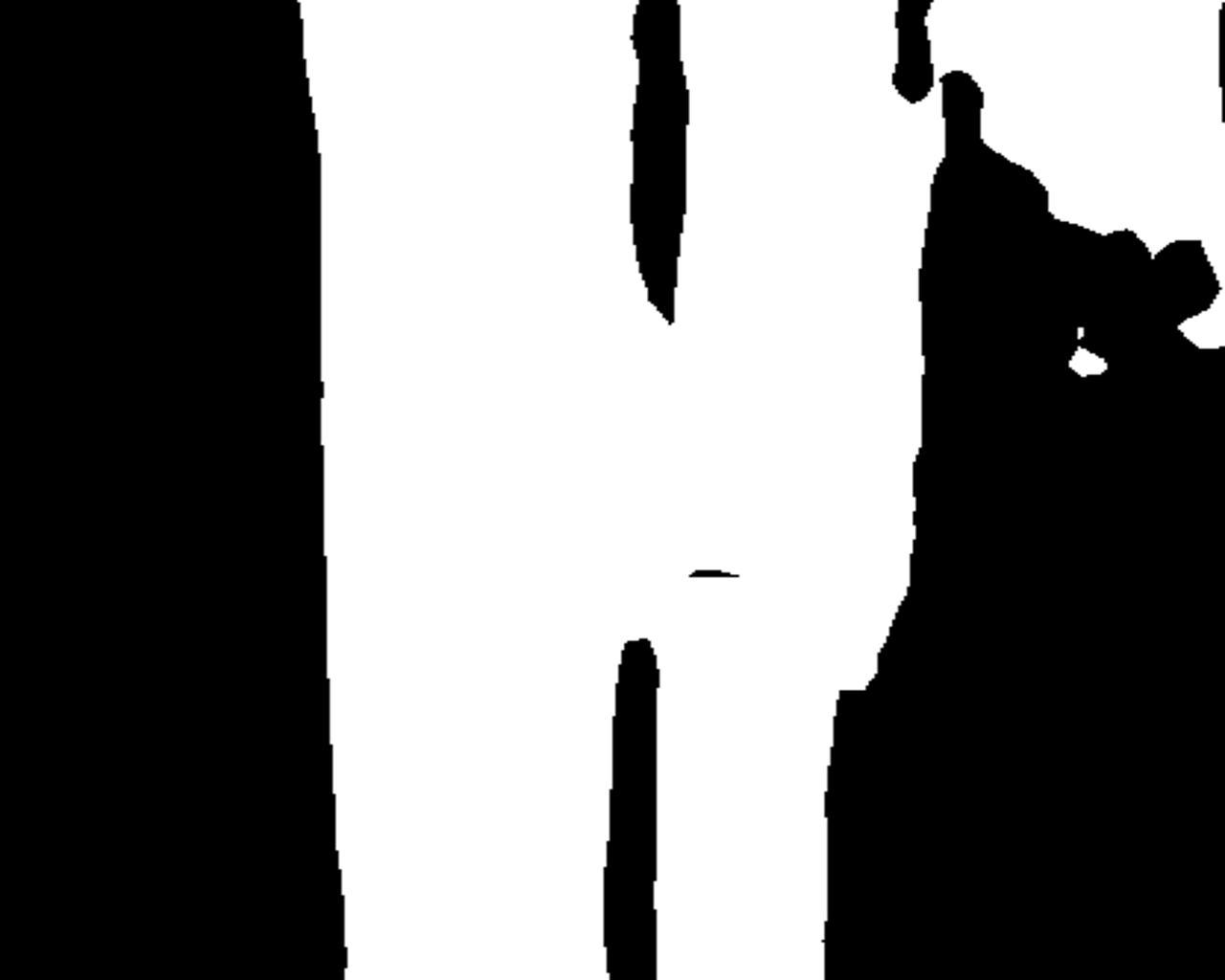}&
		\includegraphics[width=\wvisual, height=\hvisualshort]{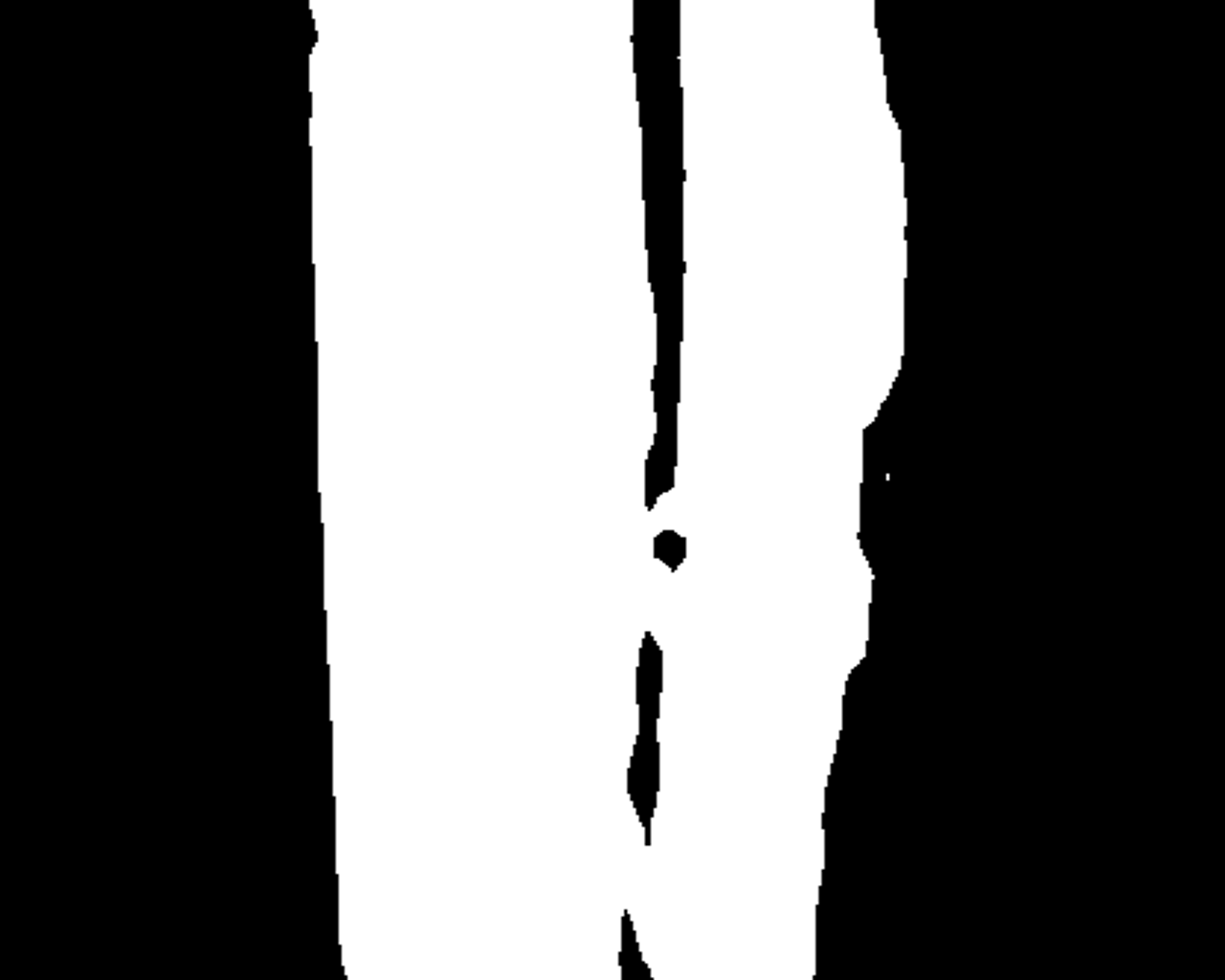}&
		\includegraphics[width=\wvisual, height=\hvisualshort]{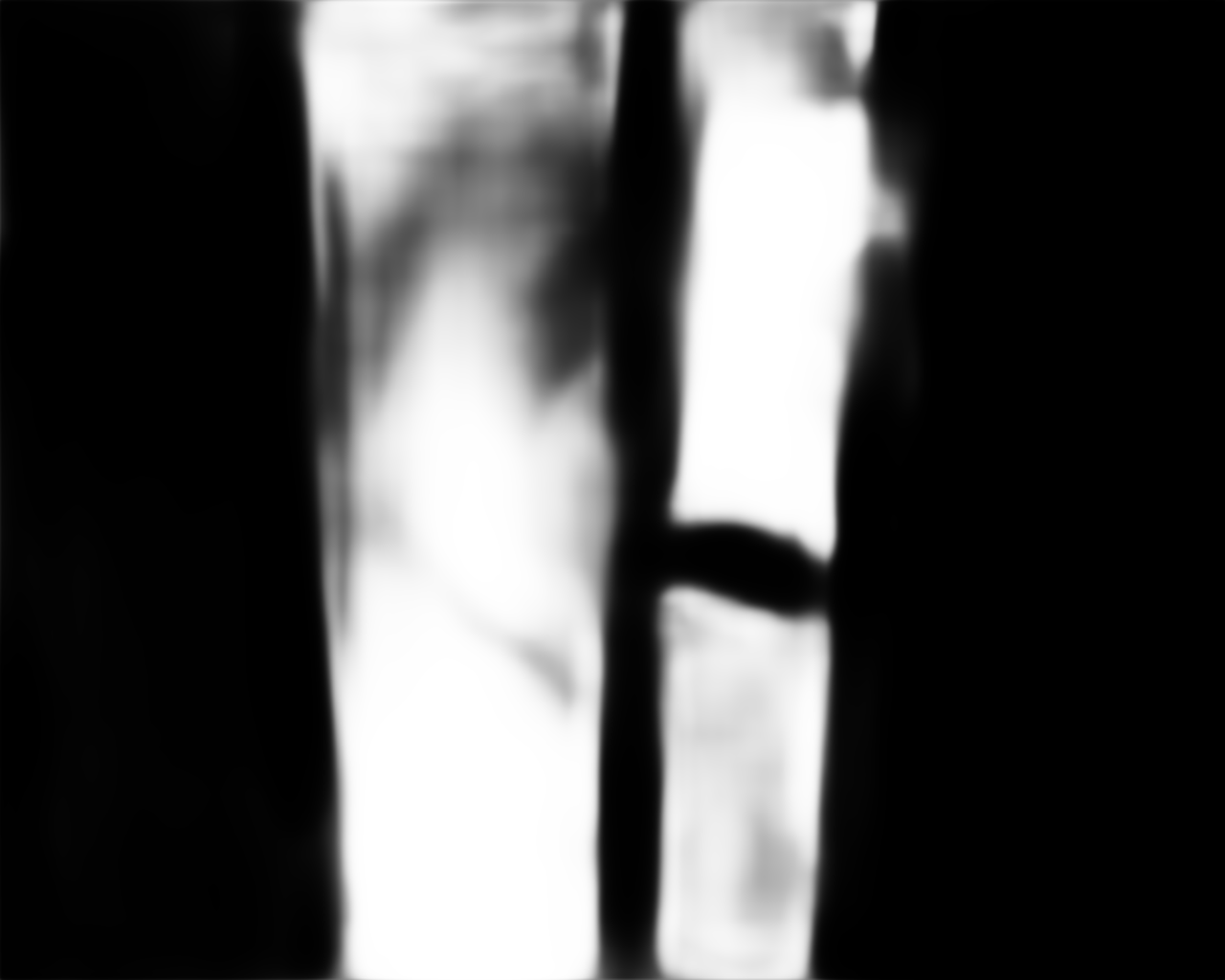}&
		\includegraphics[width=\wvisual, height=\hvisualshort]{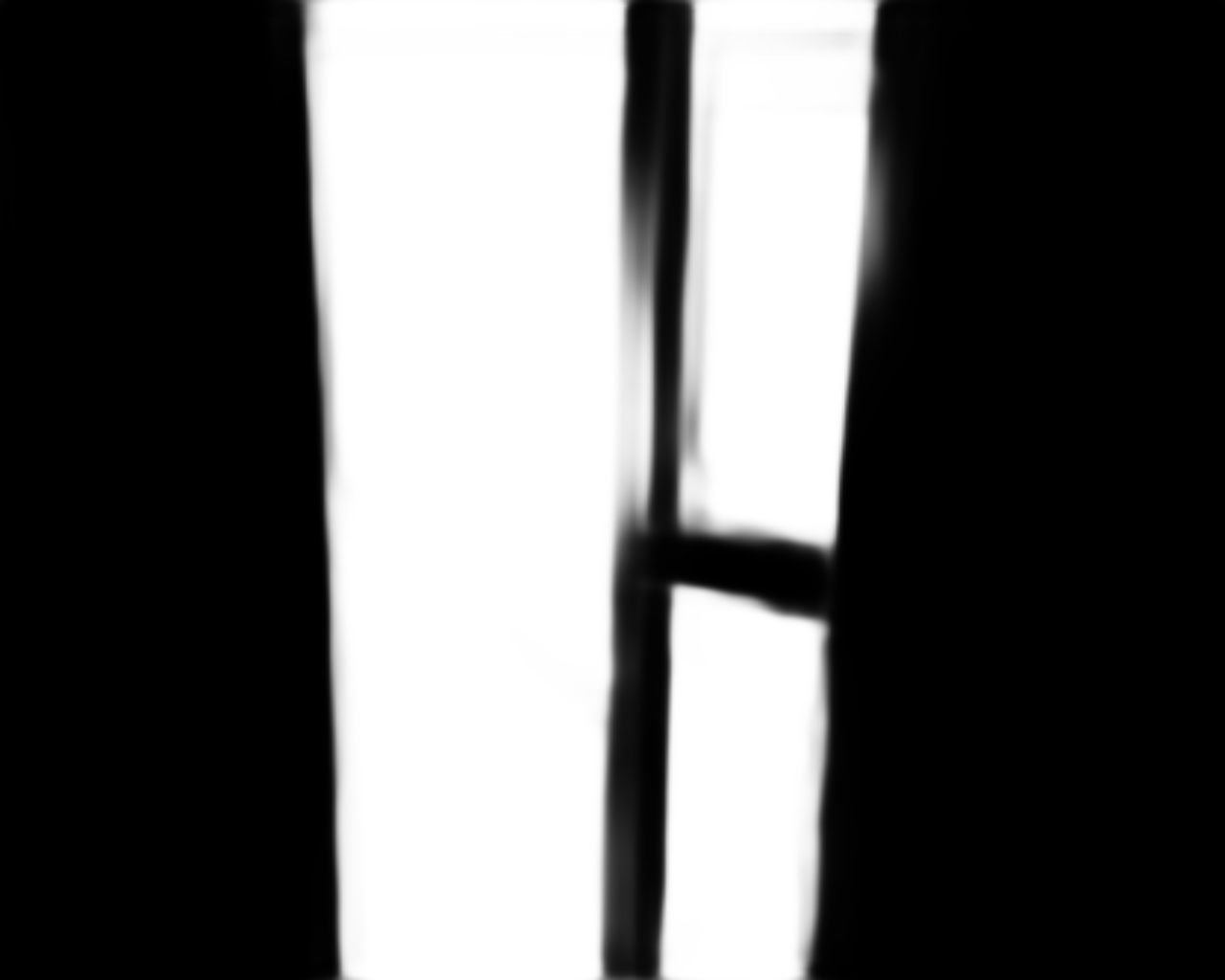}&
		\includegraphics[width=\wvisual, height=\hvisualshort]{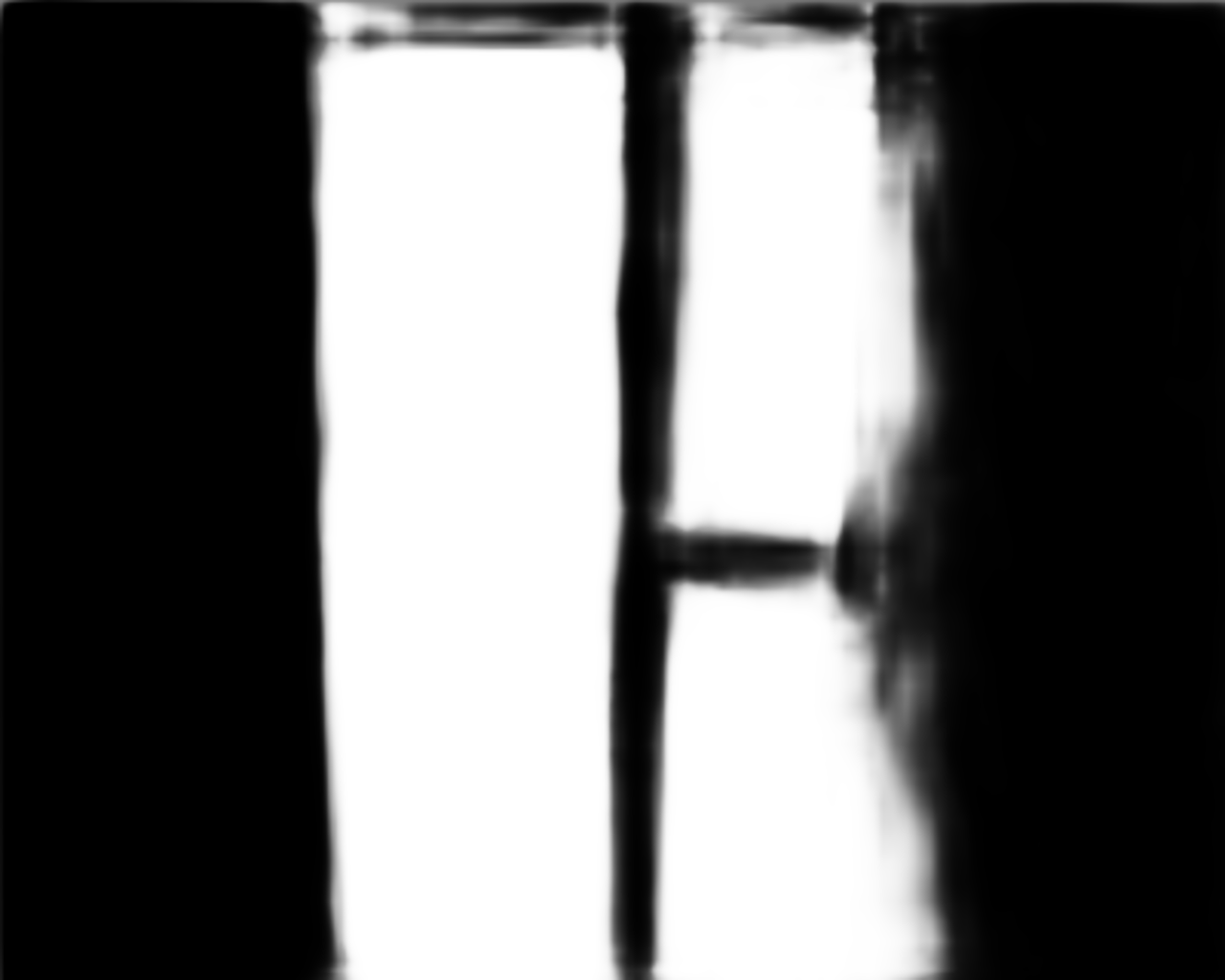}&
		\includegraphics[width=\wvisual, height=\hvisualshort]{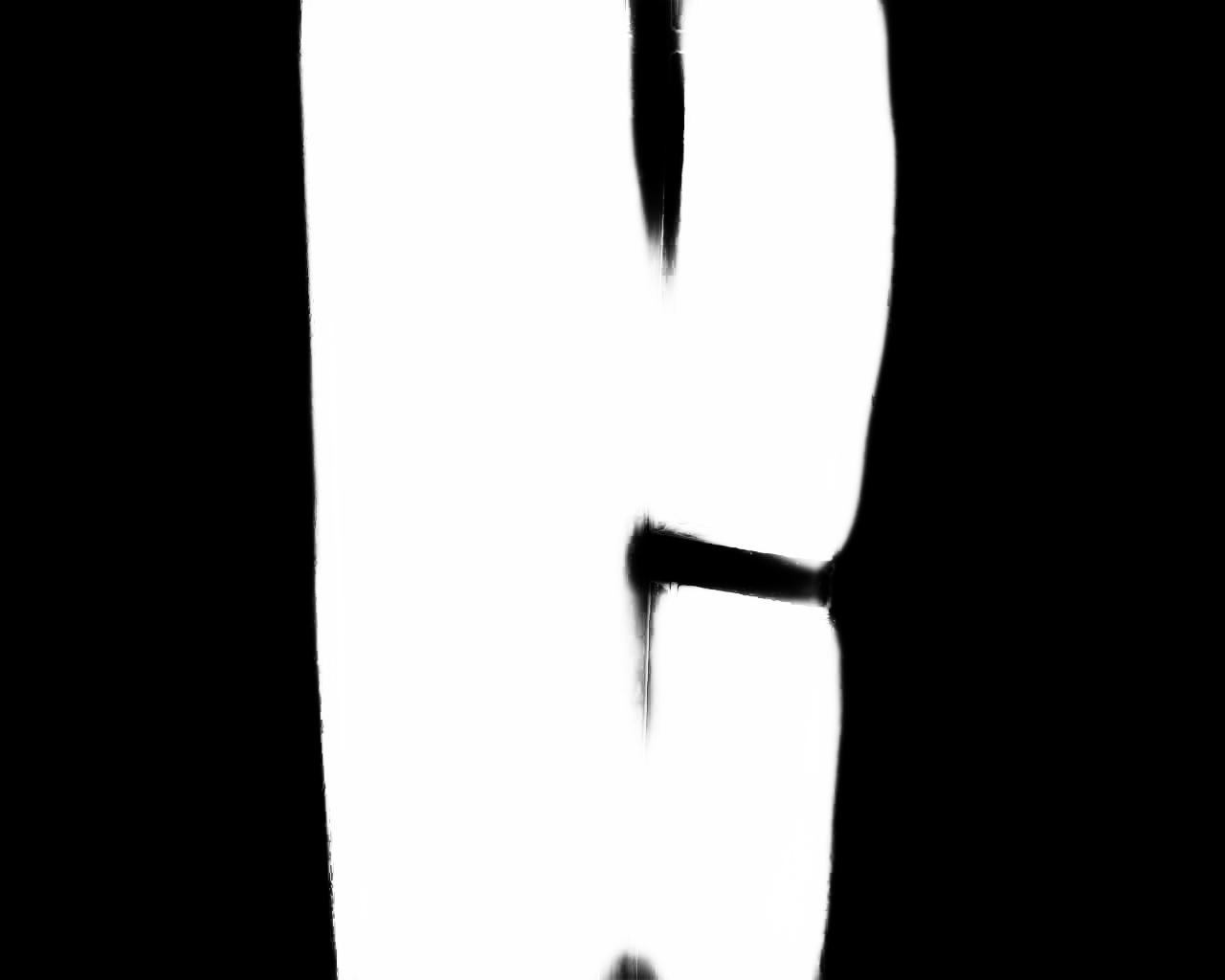}&
		\includegraphics[width=\wvisual, height=\hvisualshort]{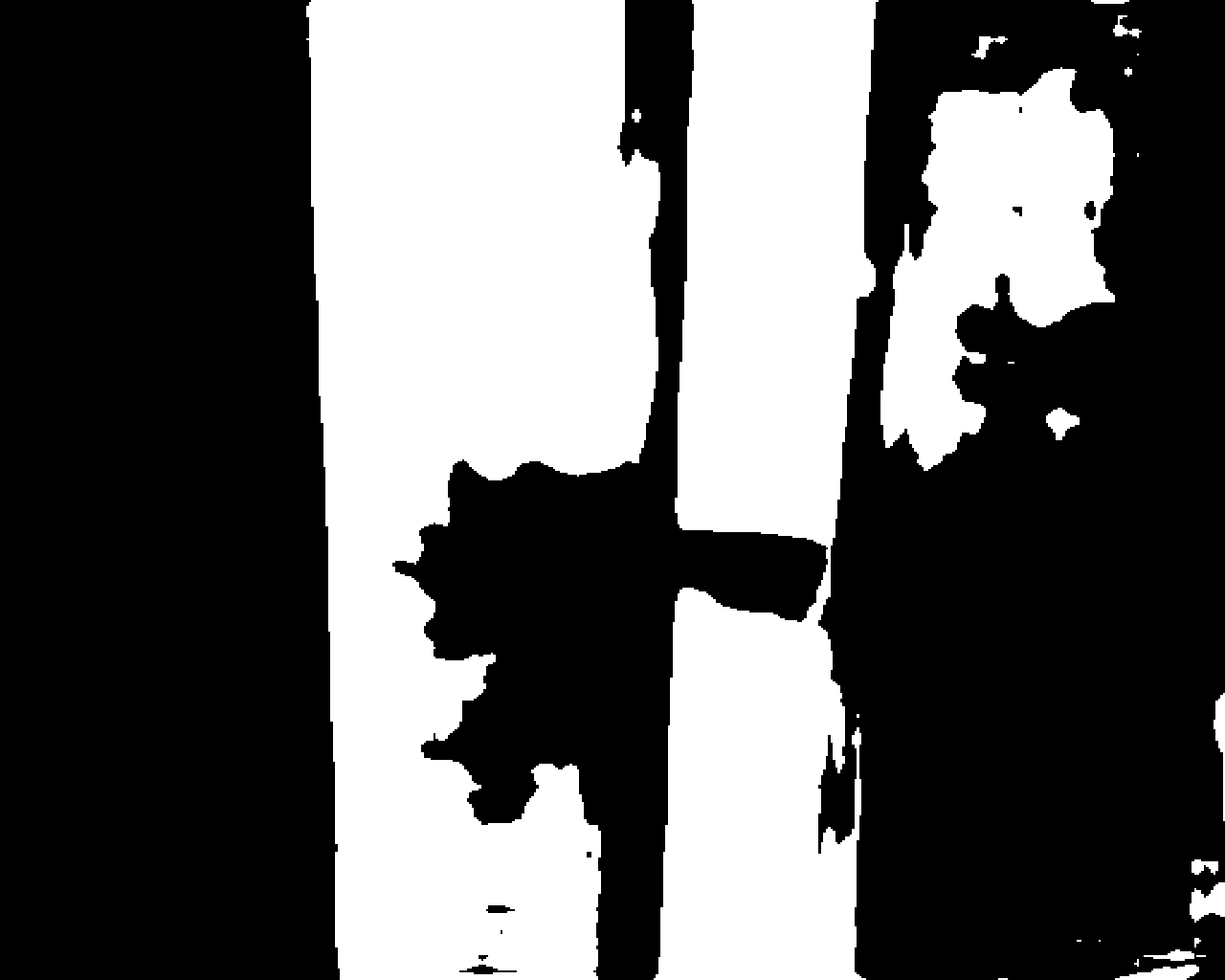}&
		\includegraphics[width=\wvisual, height=\hvisualshort]{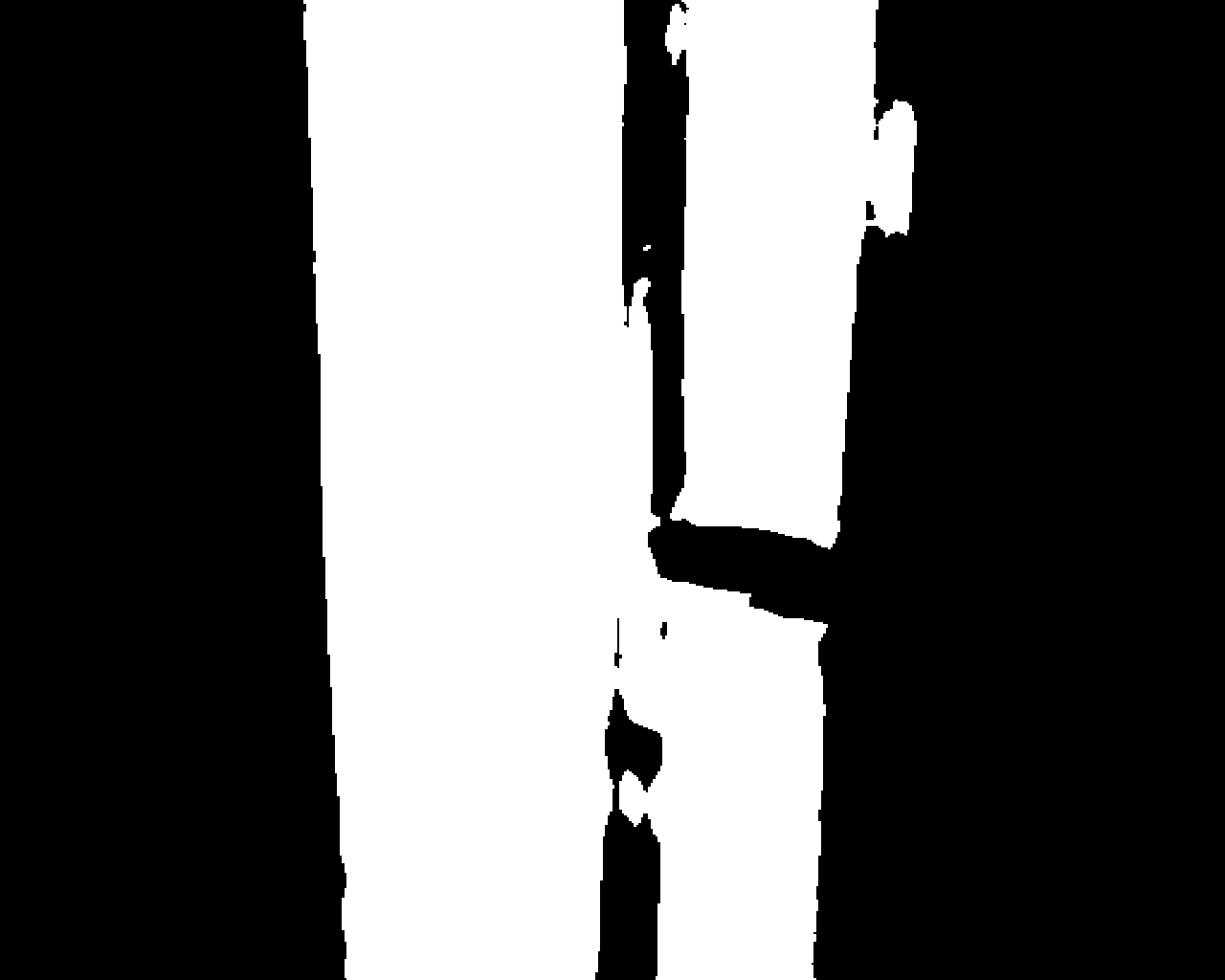}&
		\includegraphics[width=\wvisual, height=\hvisualshort]{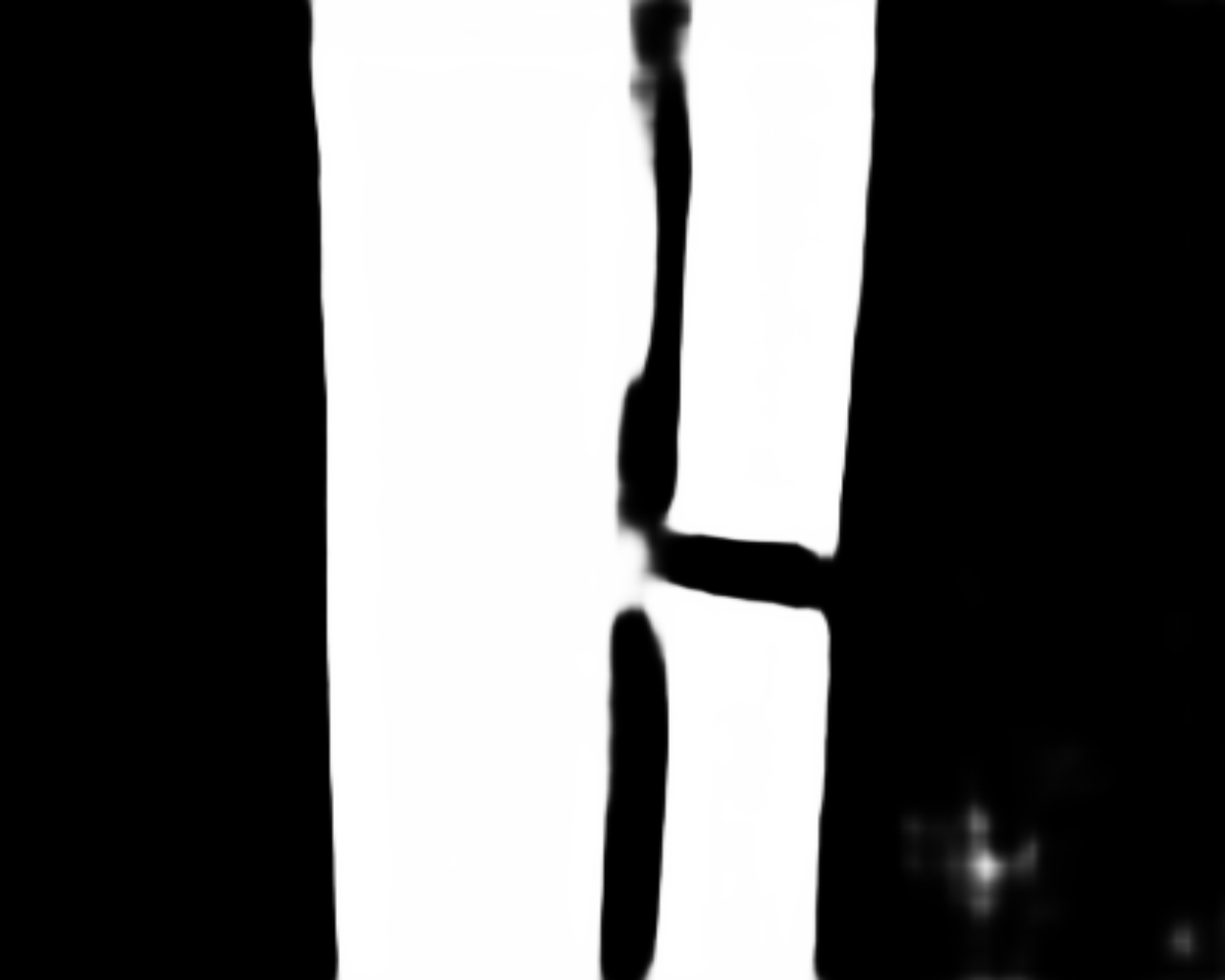}&
		\includegraphics[width=\wvisual, height=\hvisualshort]{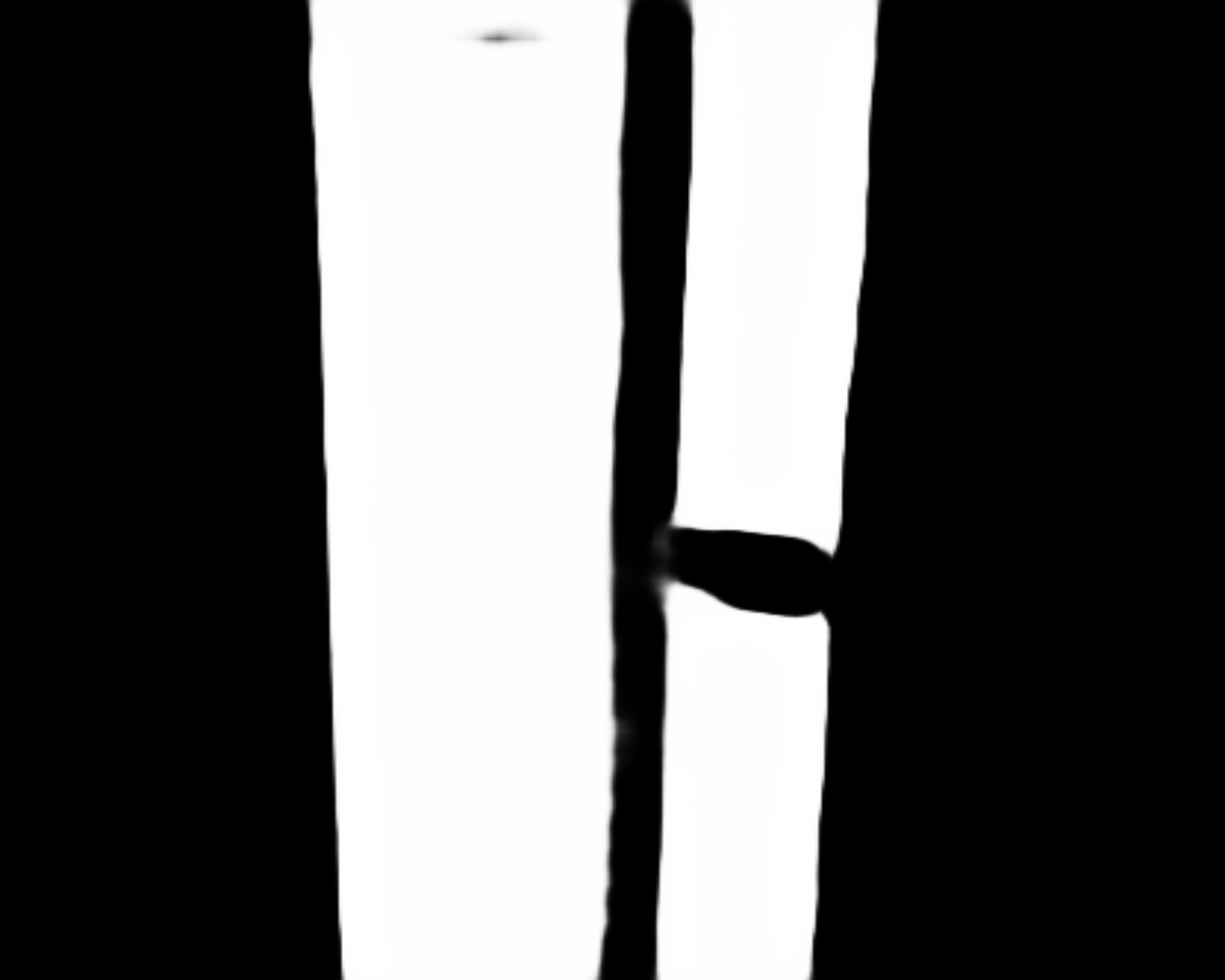}&
		\includegraphics[width=\wvisual, height=\hvisualshort]{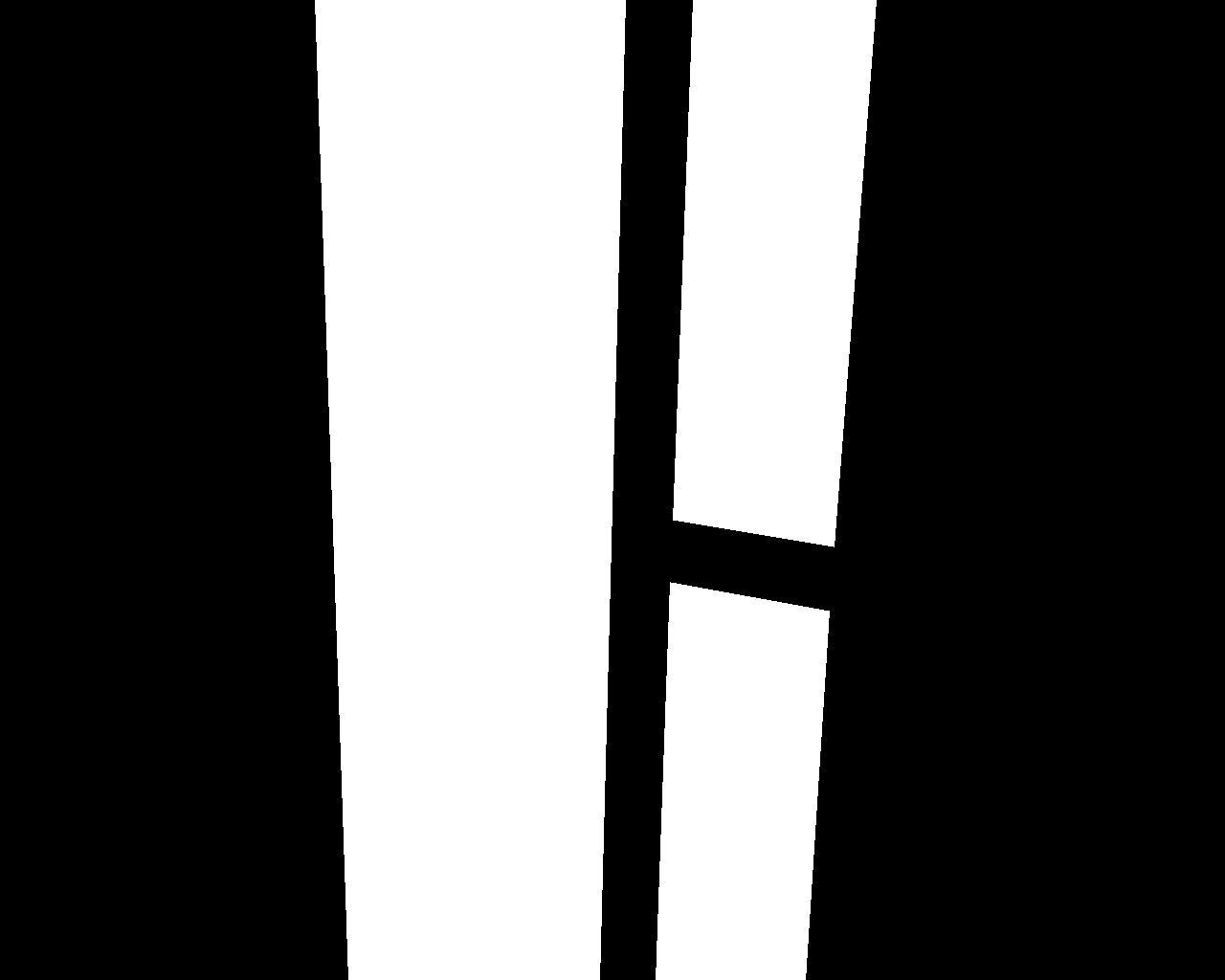} \\
		
		\includegraphics[width=\wvisual, height=\hvisualtall]{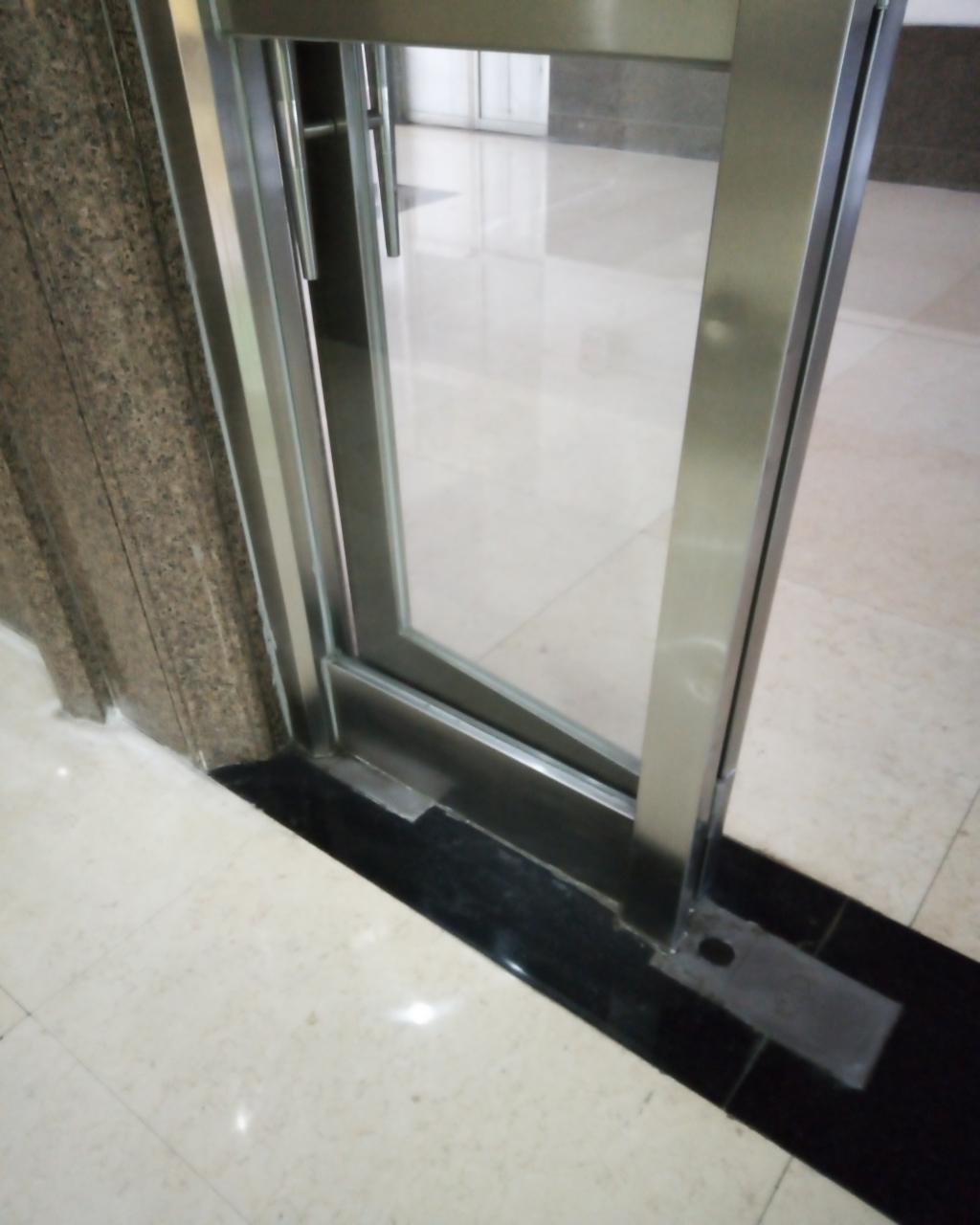}&
		\includegraphics[width=\wvisual, height=\hvisualtall]{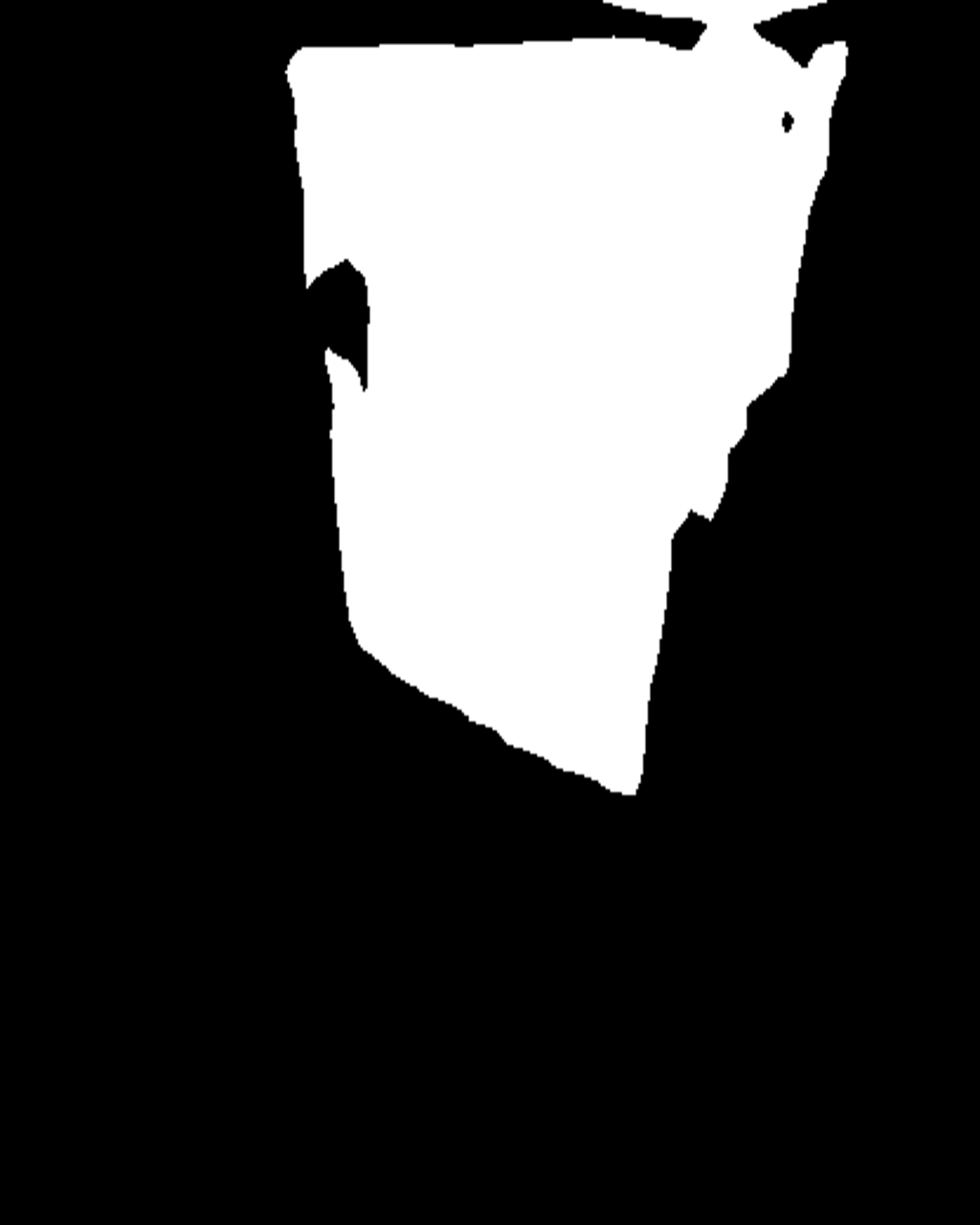}&
		\includegraphics[width=\wvisual, height=\hvisualtall]{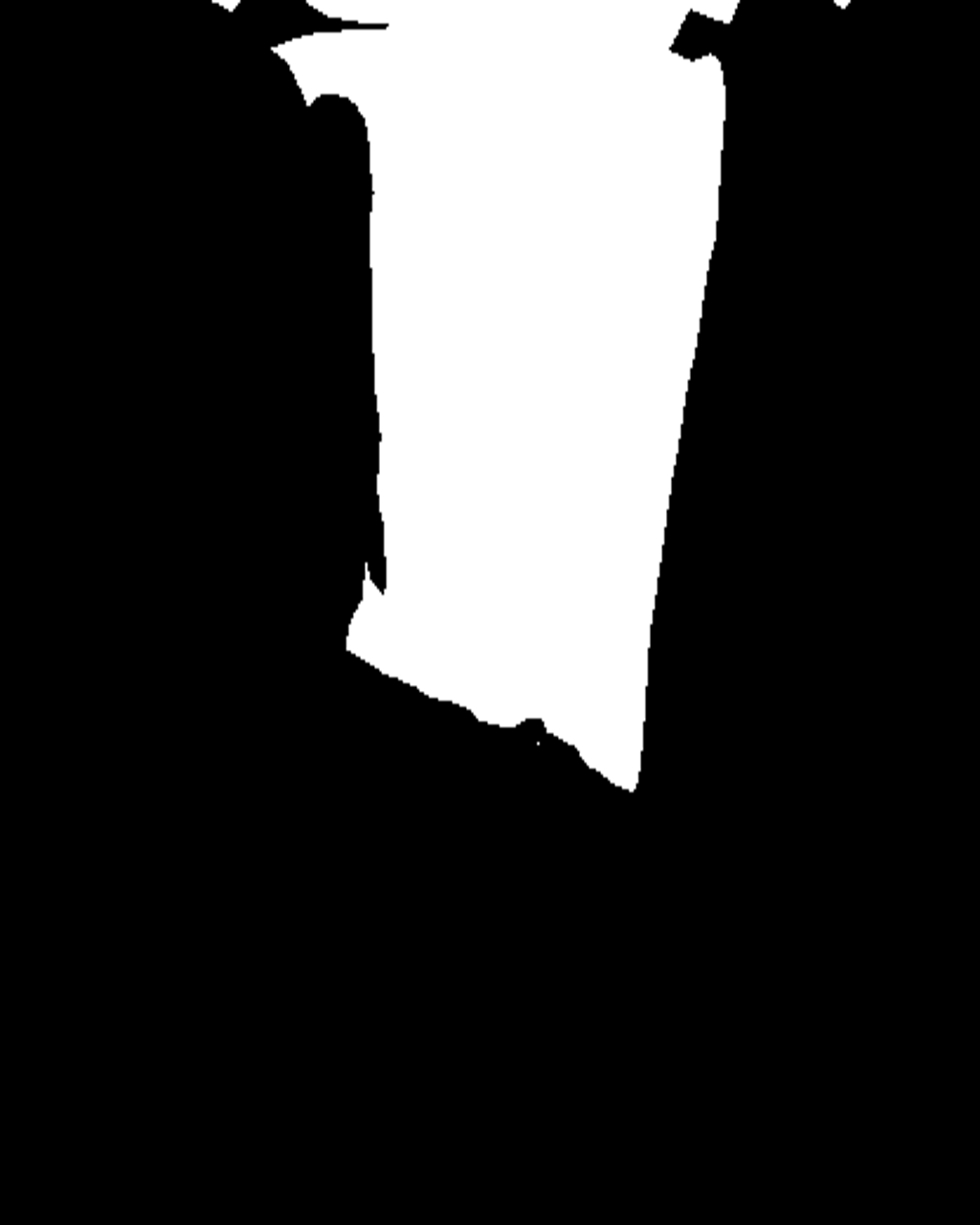}&
		\includegraphics[width=\wvisual, height=\hvisualtall]{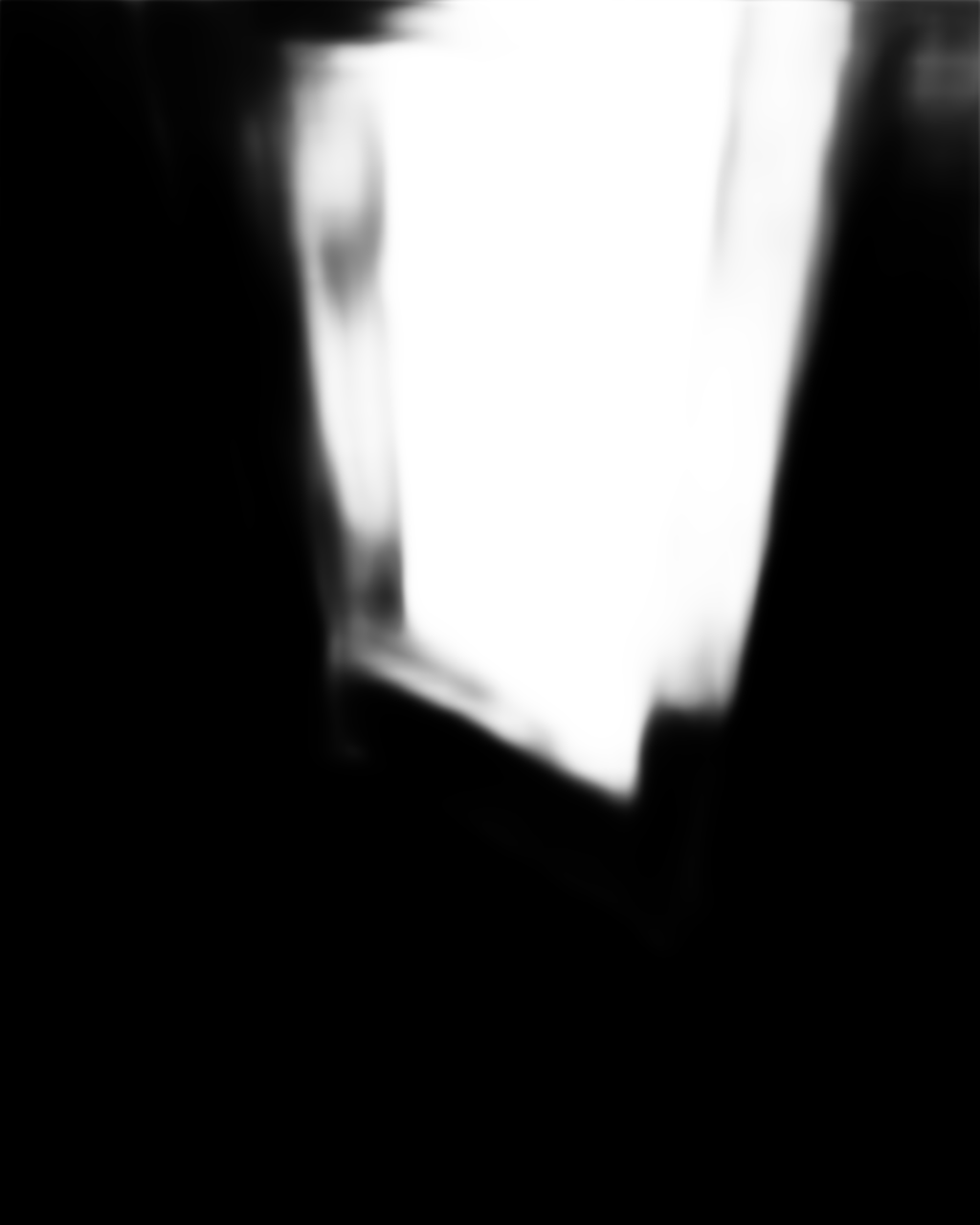}&
		\includegraphics[width=\wvisual, height=\hvisualtall]{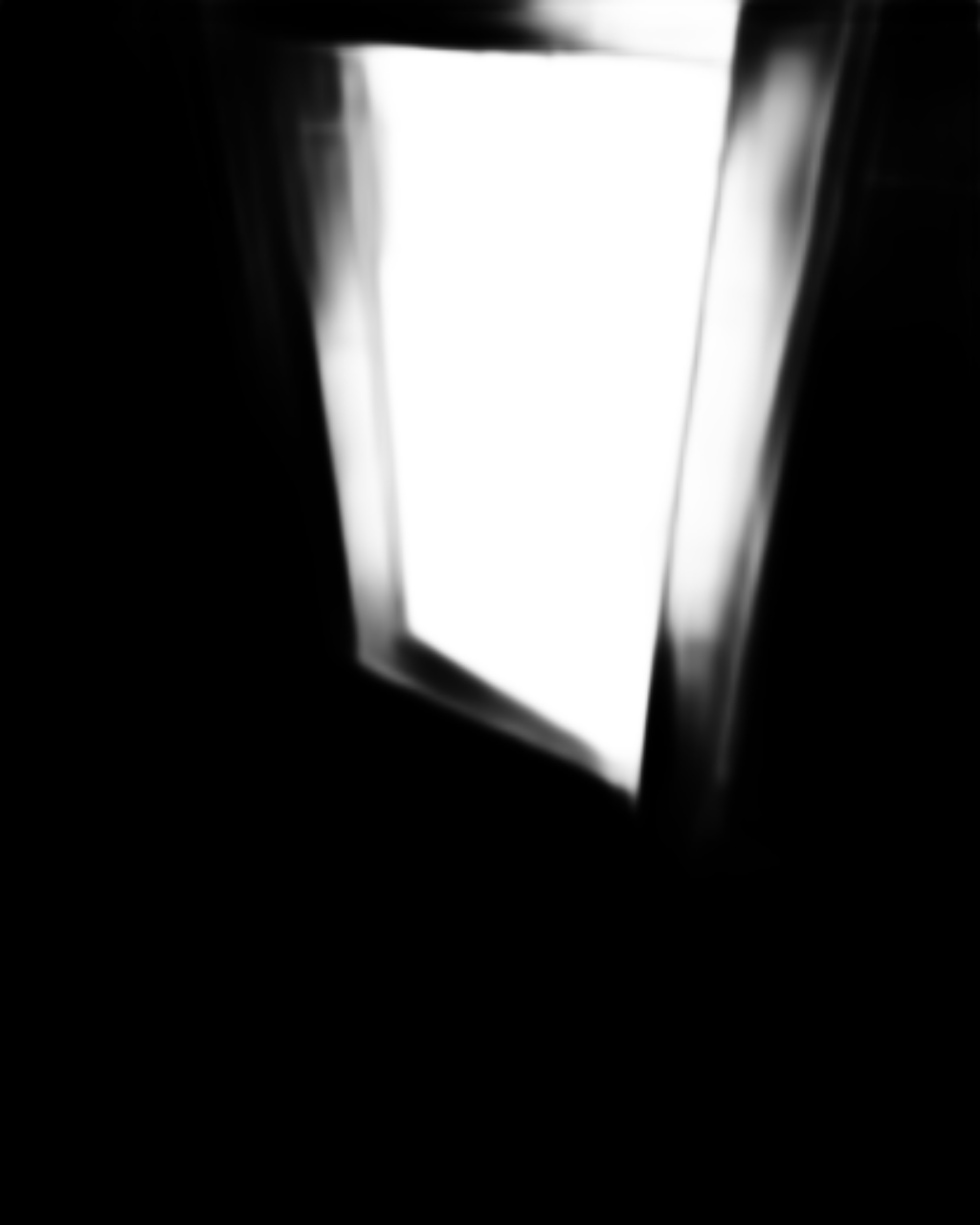}&
		\includegraphics[width=\wvisual, height=\hvisualtall]{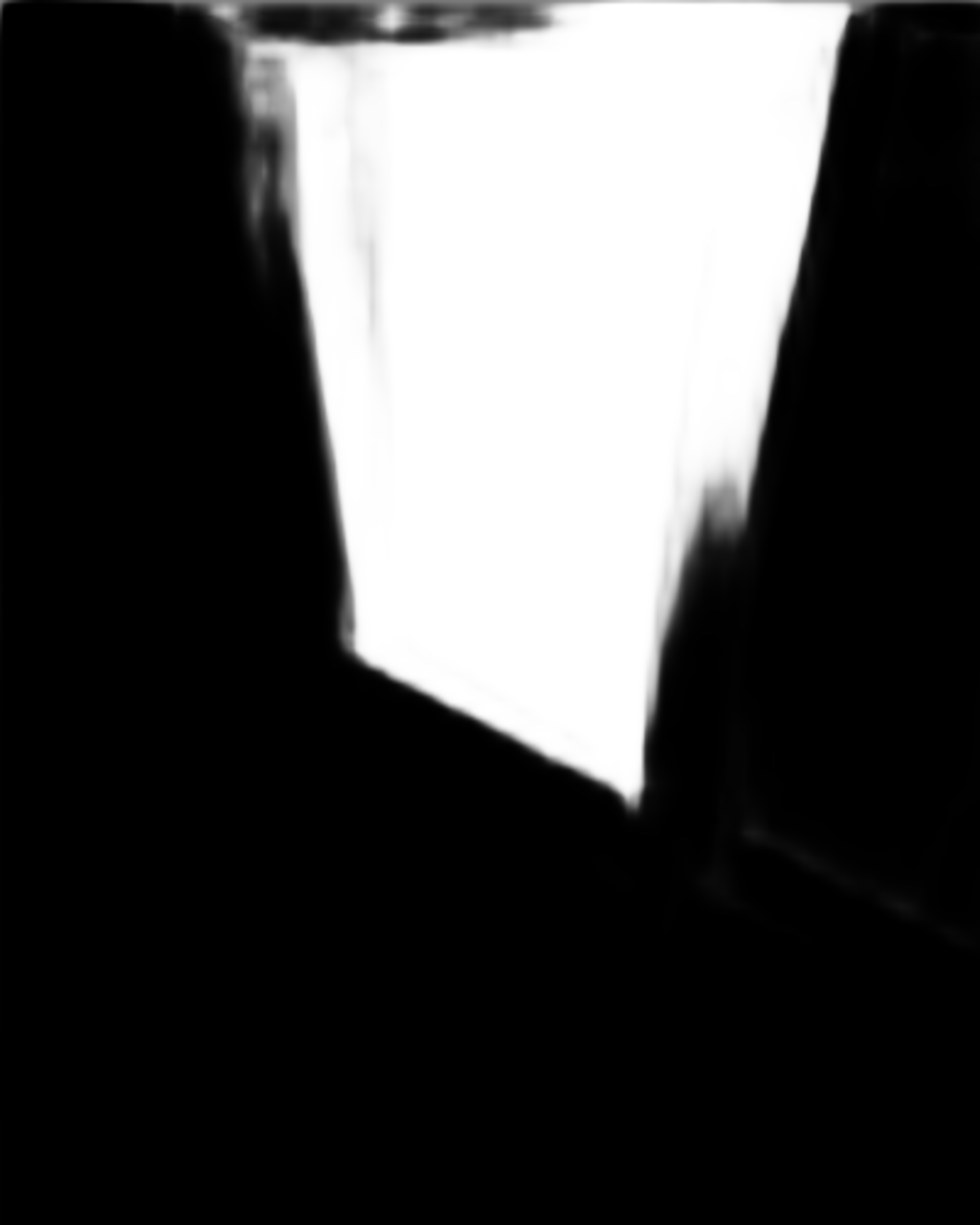}&
		\includegraphics[width=\wvisual, height=\hvisualtall]{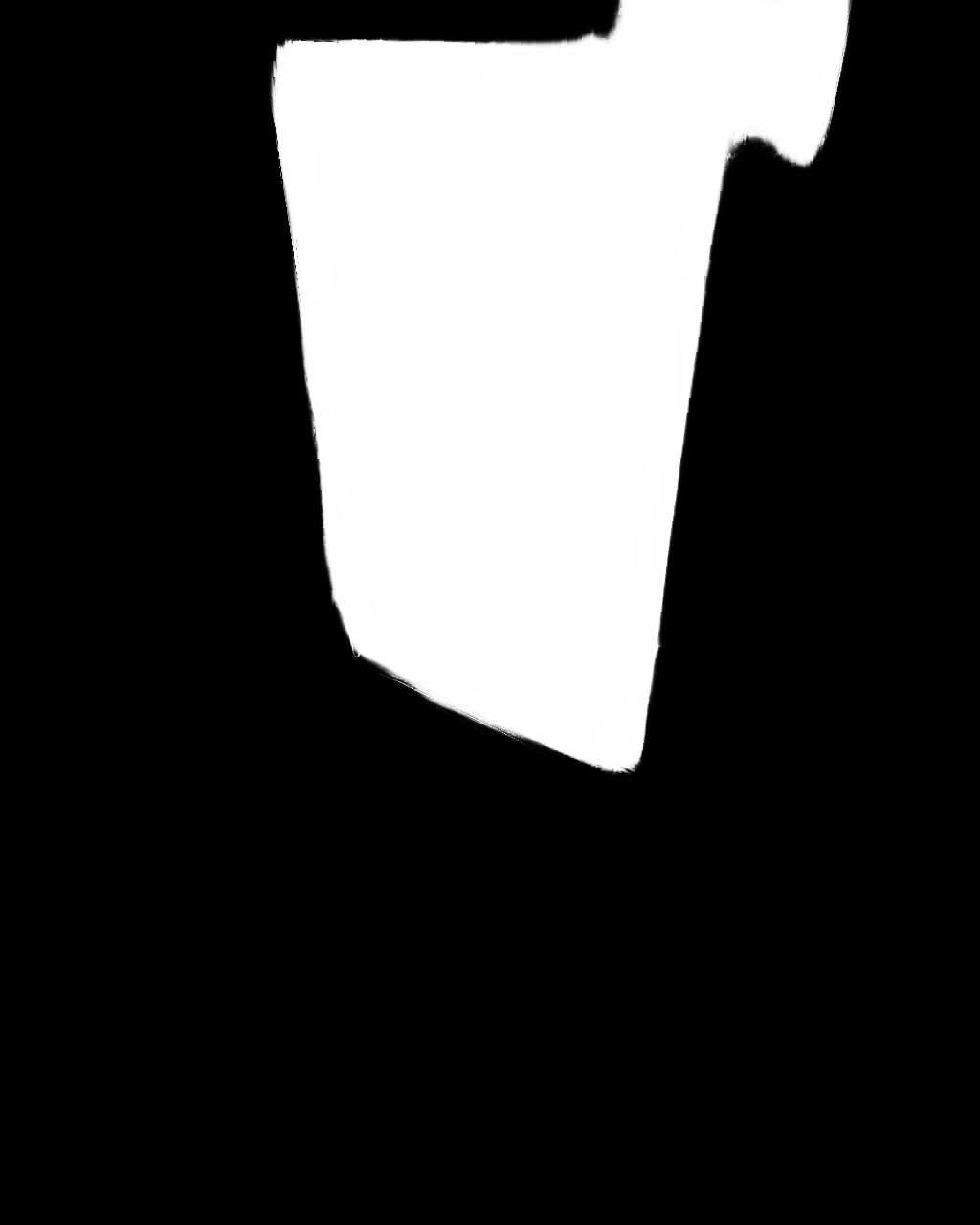}&
		\includegraphics[width=\wvisual, height=\hvisualtall]{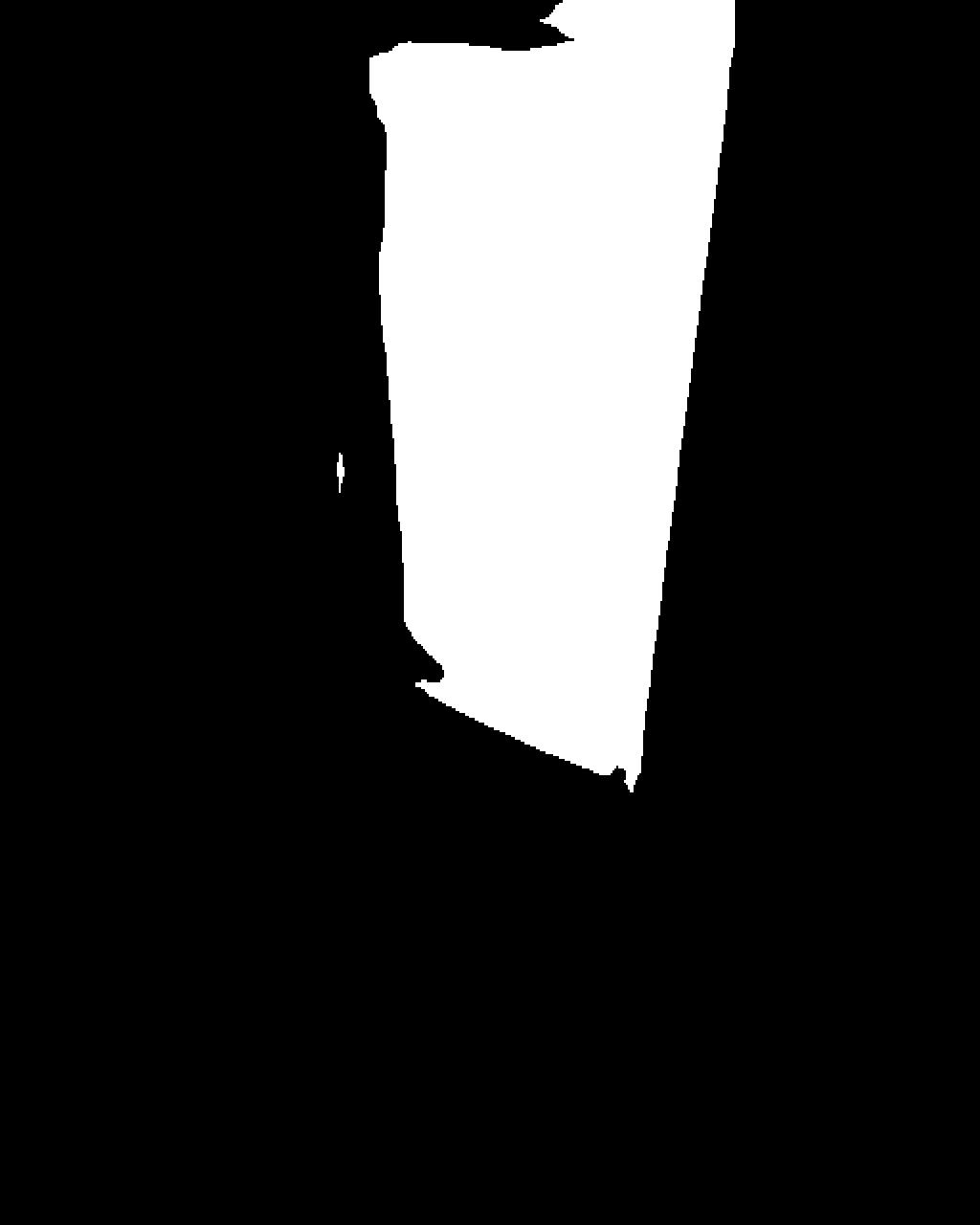}&
		\includegraphics[width=\wvisual, height=\hvisualtall]{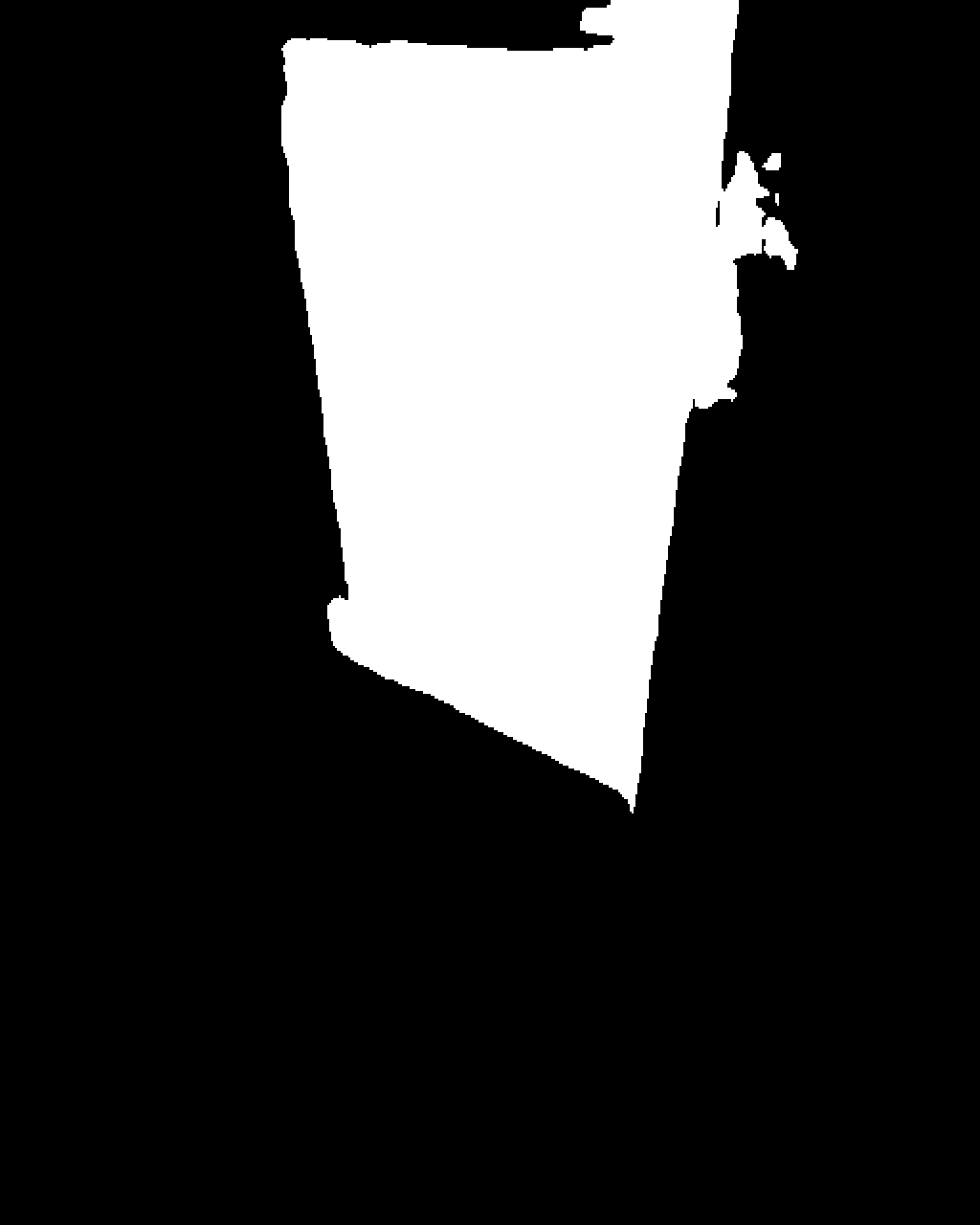}&
		\includegraphics[width=\wvisual, height=\hvisualtall]{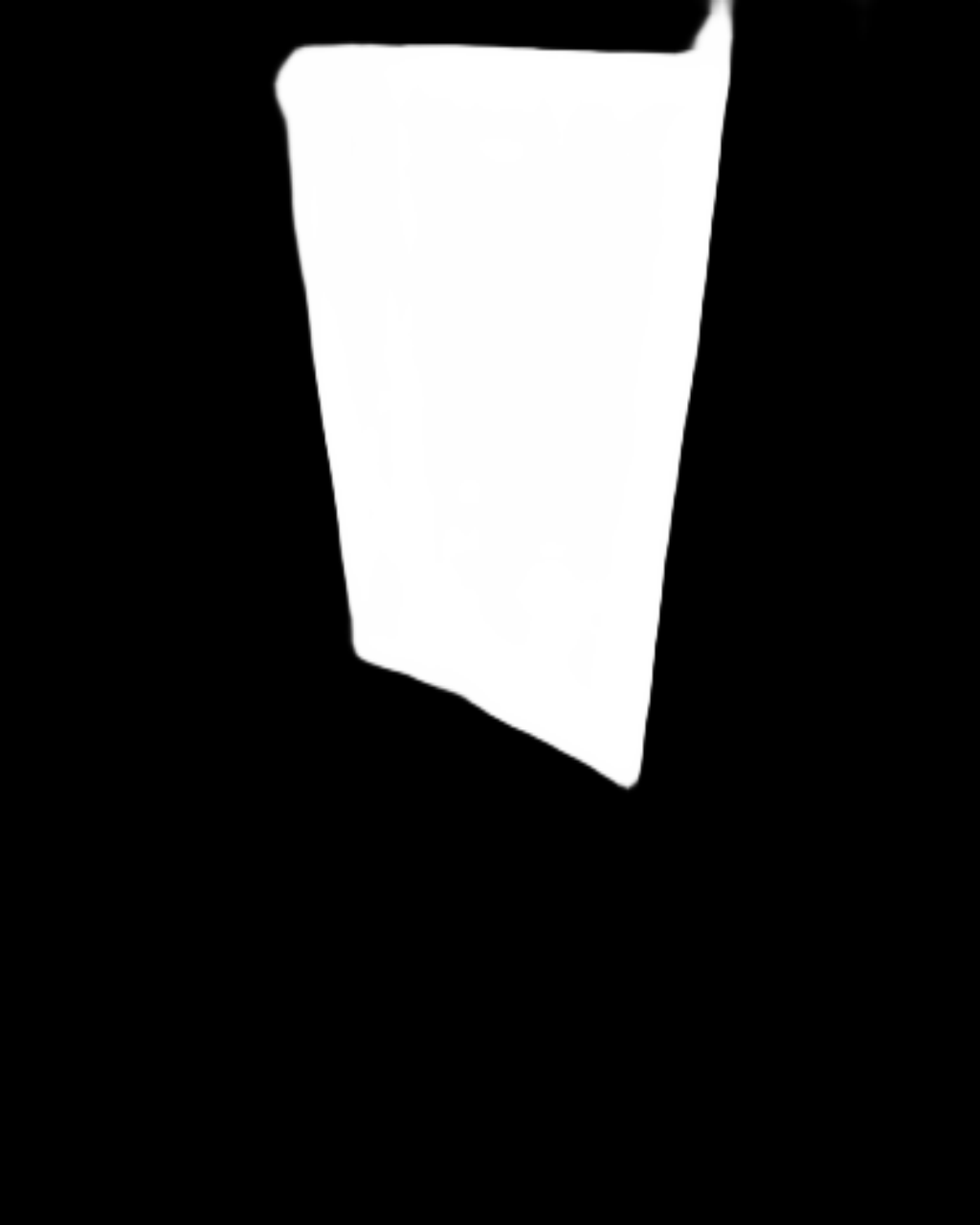}&
		\includegraphics[width=\wvisual, height=\hvisualtall]{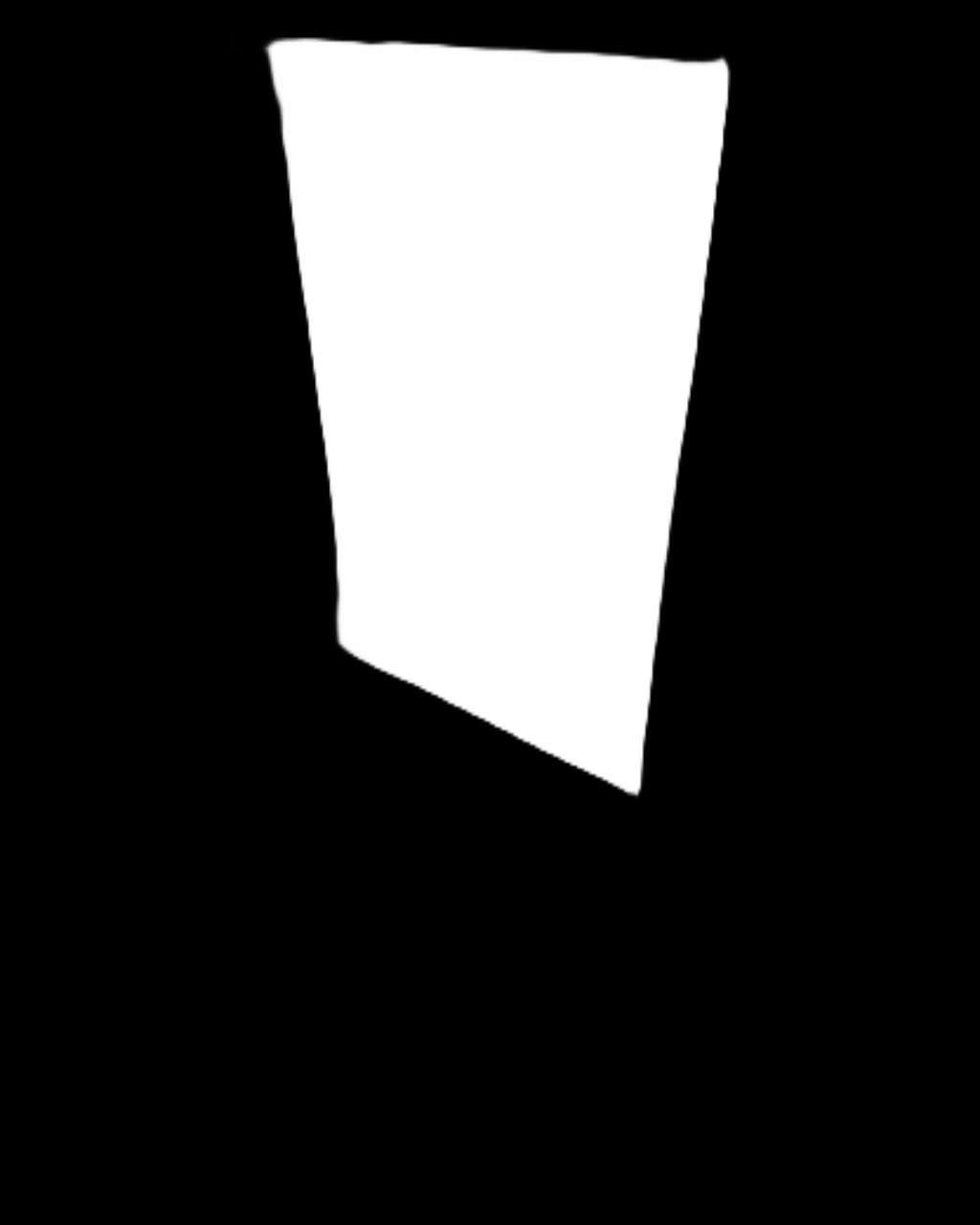}&
		\includegraphics[width=\wvisual, height=\hvisualtall]{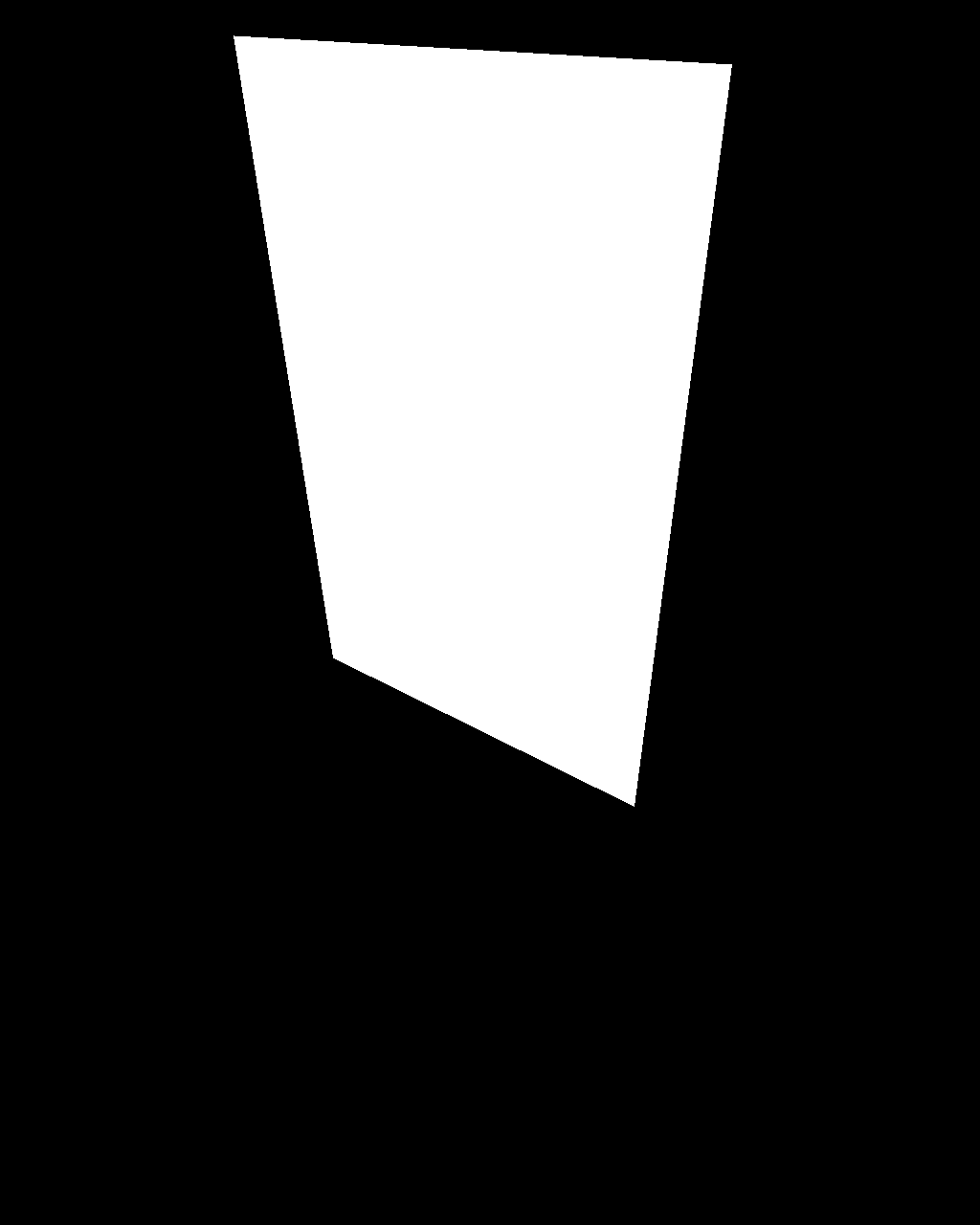} \\
		
		\includegraphics[width=\wvisual, height=\hvisualtall]{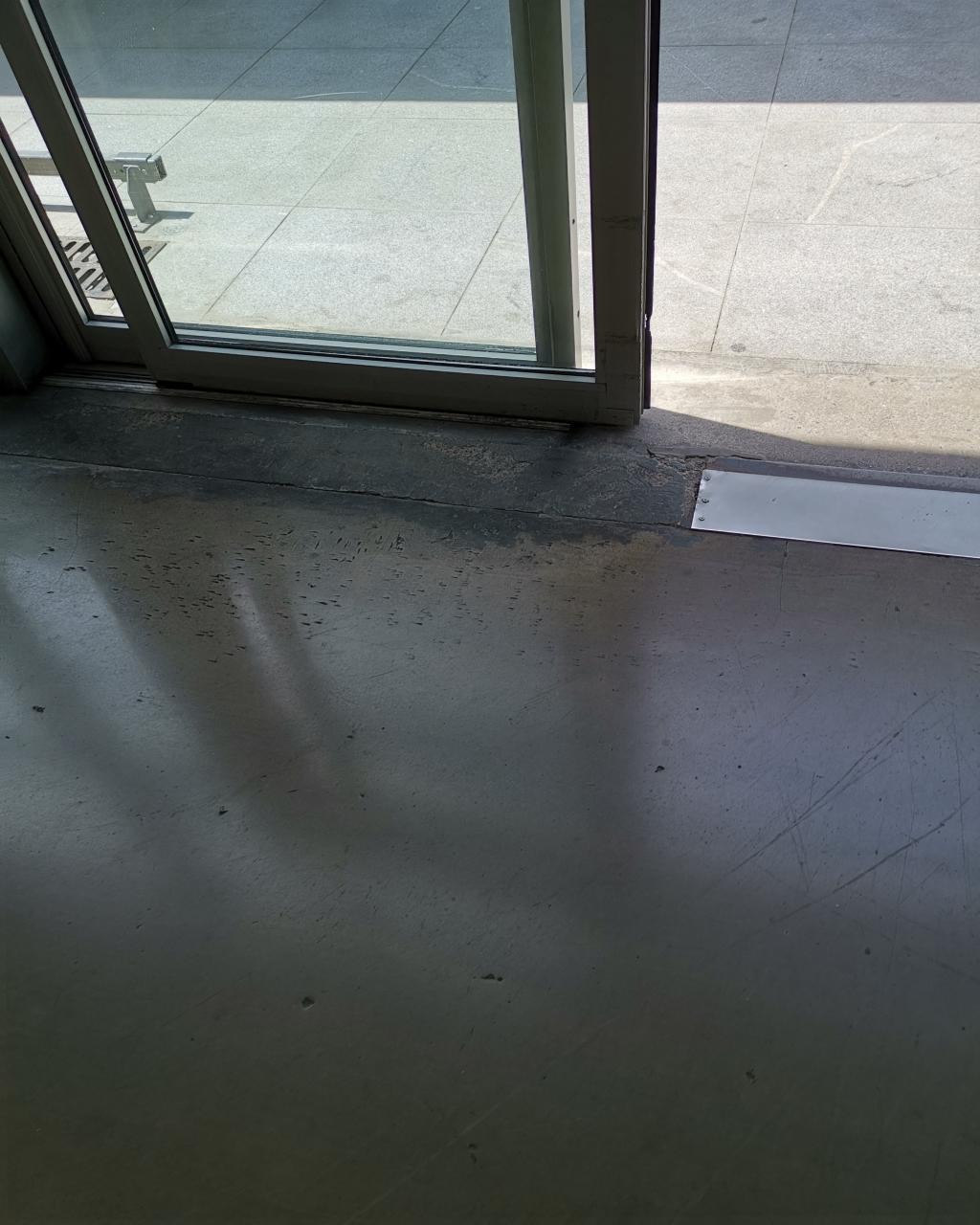}&
		\includegraphics[width=\wvisual, height=\hvisualtall]{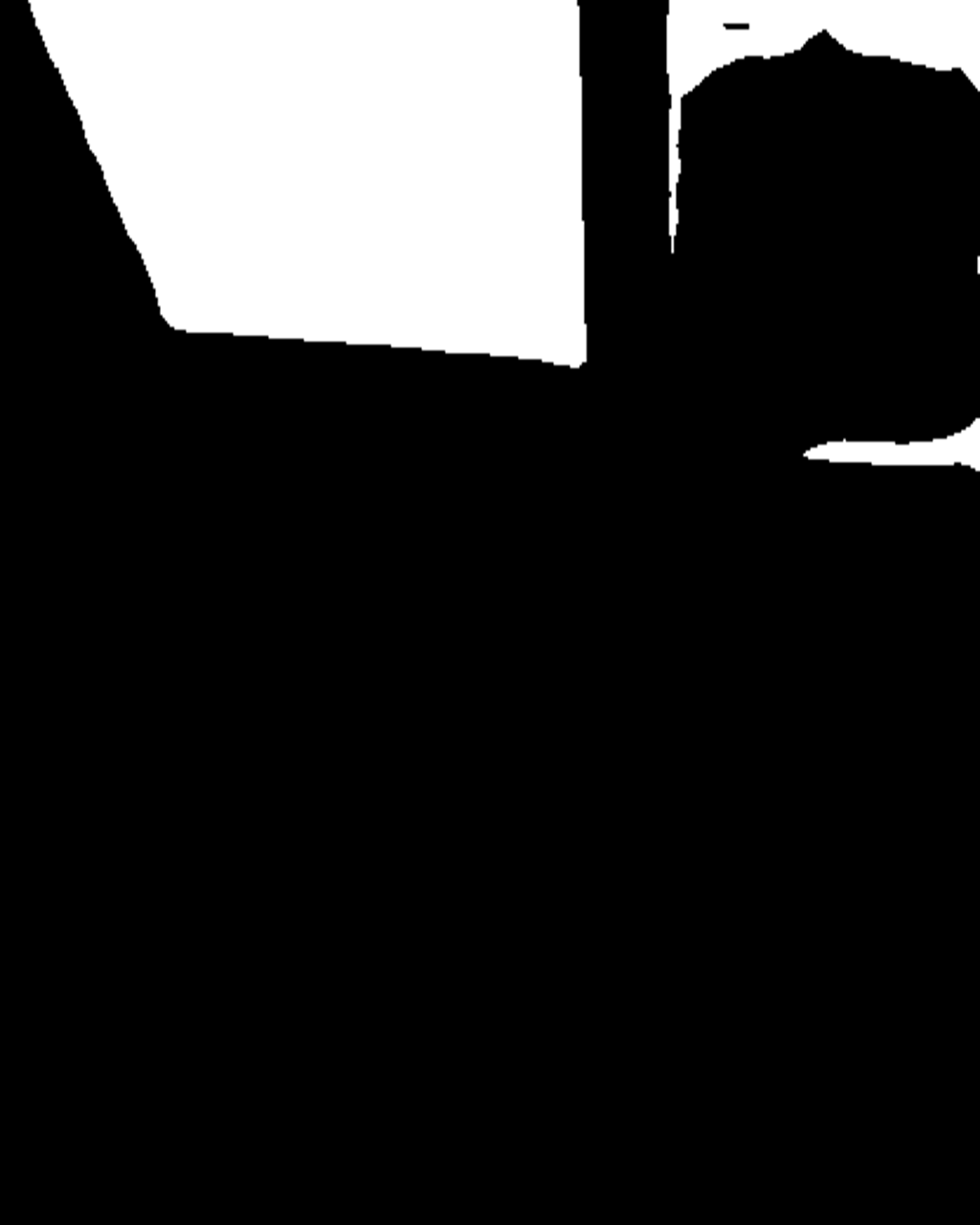}&
		\includegraphics[width=\wvisual, height=\hvisualtall]{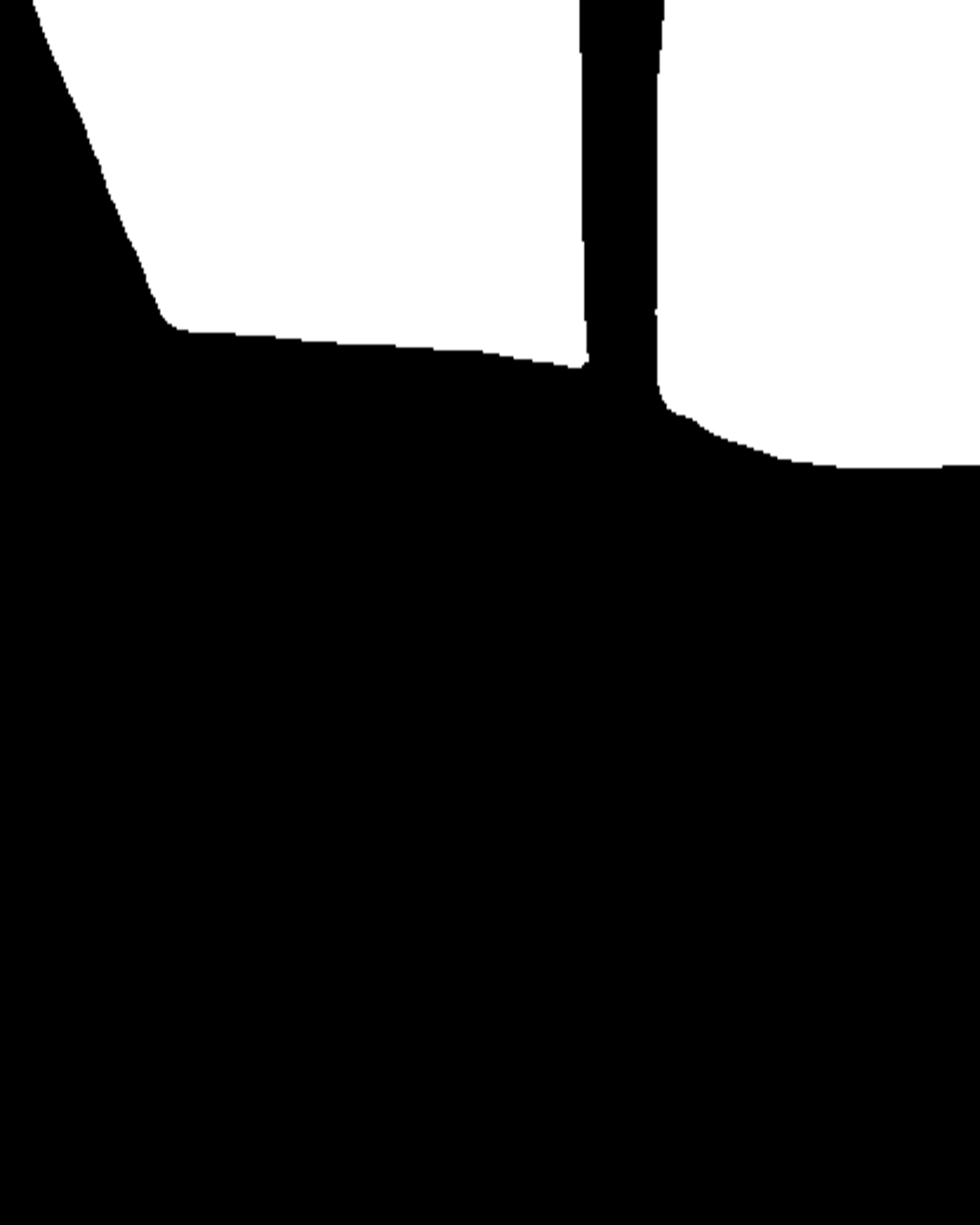}&
		\includegraphics[width=\wvisual, height=\hvisualtall]{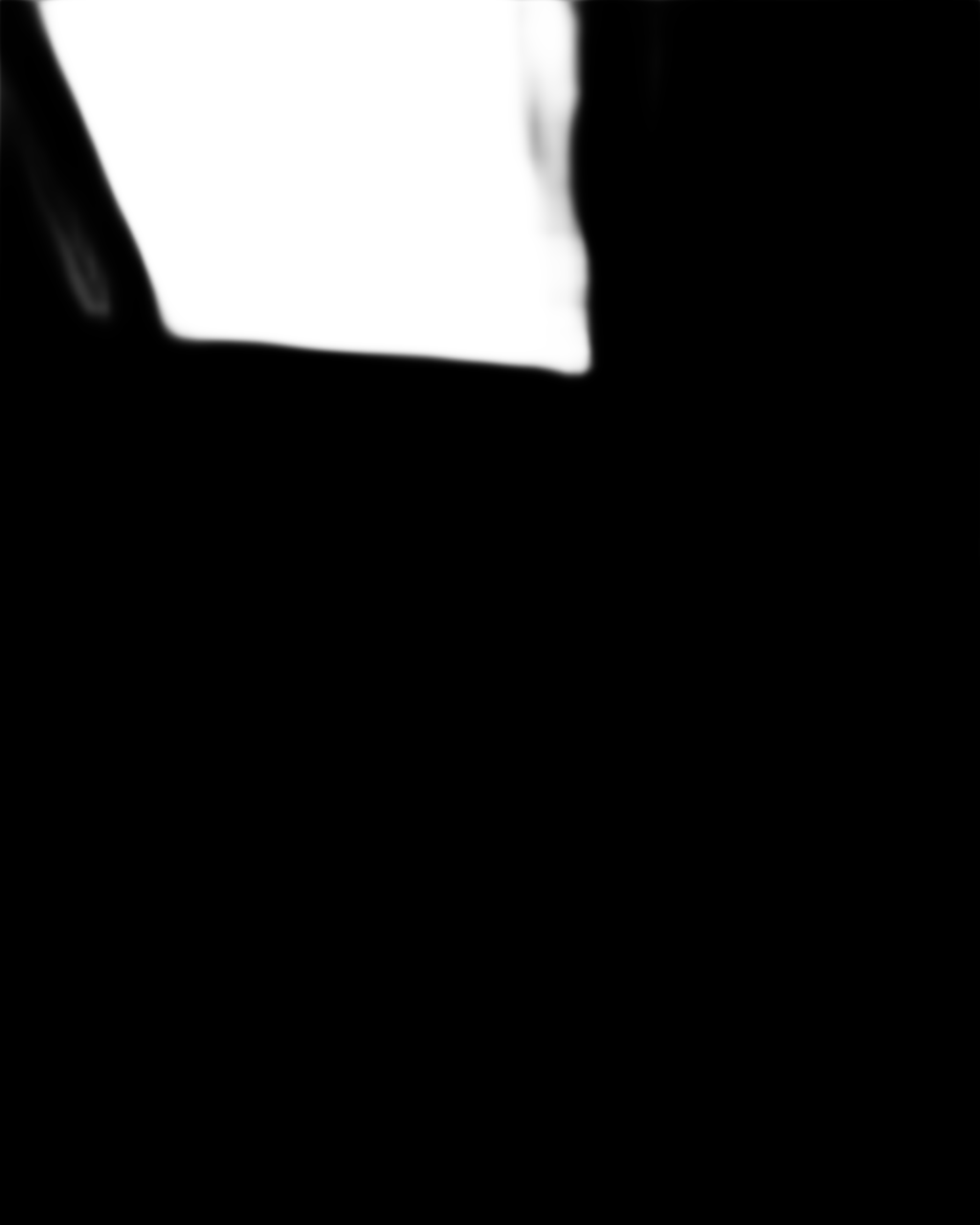}&
		\includegraphics[width=\wvisual, height=\hvisualtall]{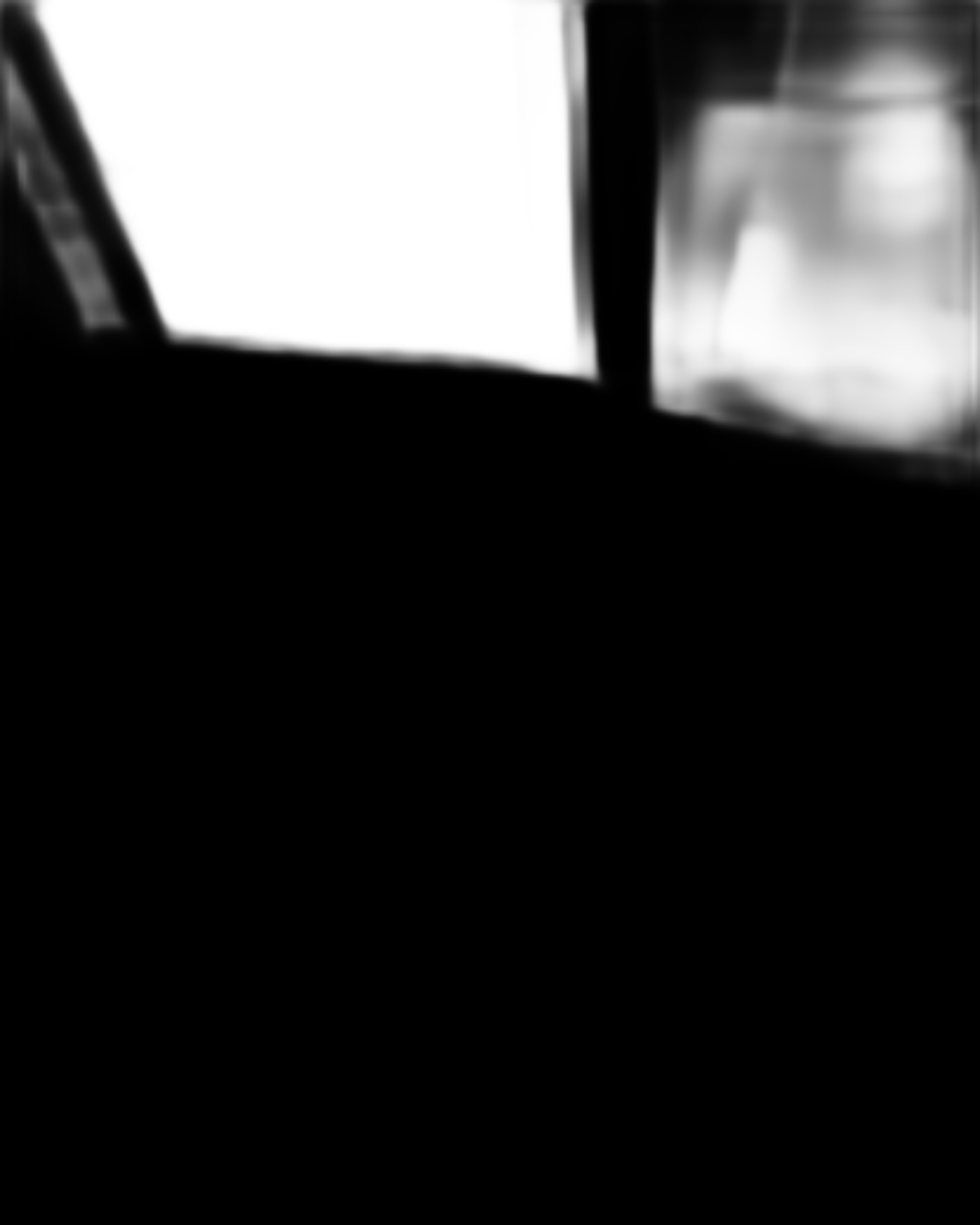}&
		\includegraphics[width=\wvisual, height=\hvisualtall]{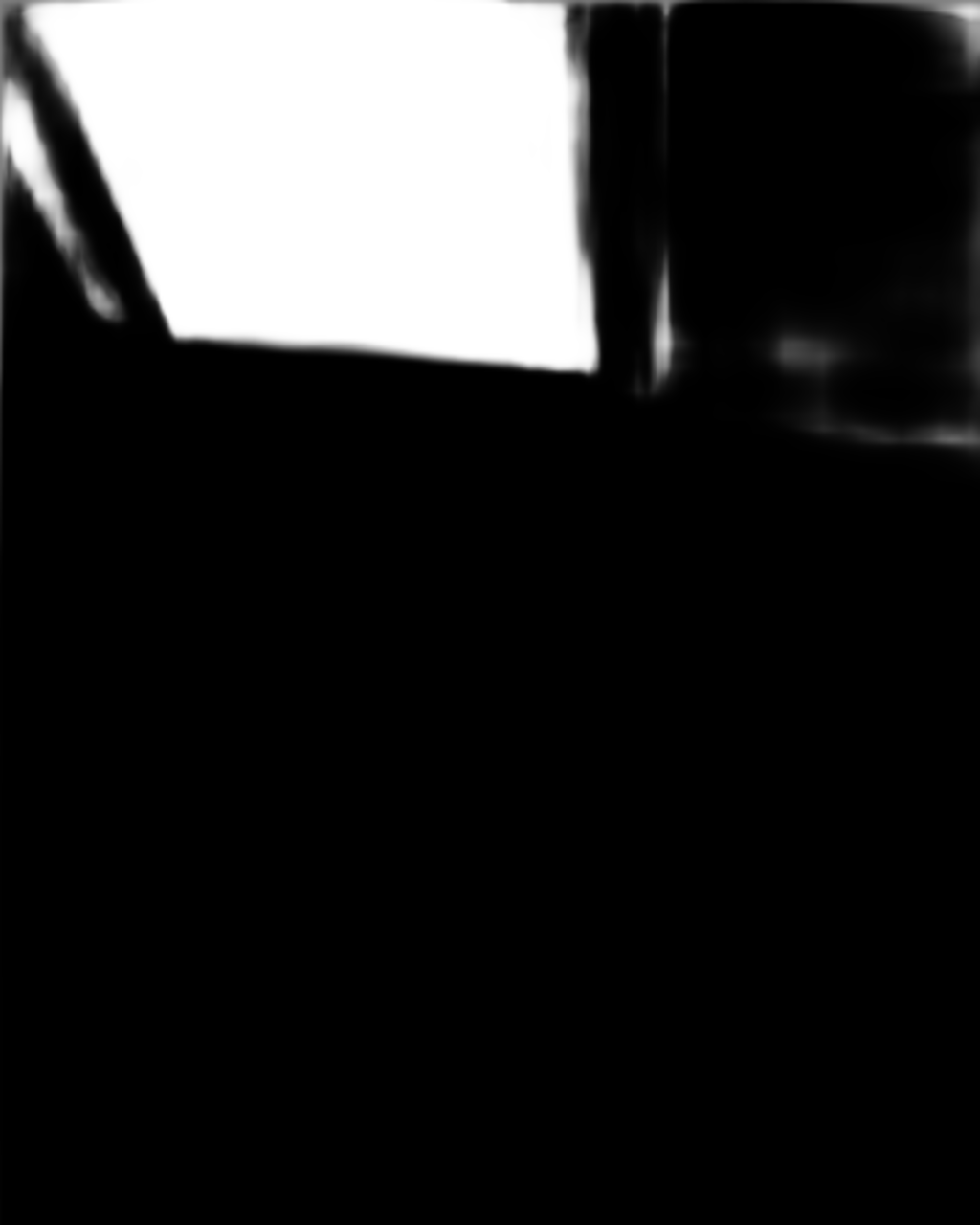}&
		\includegraphics[width=\wvisual, height=\hvisualtall]{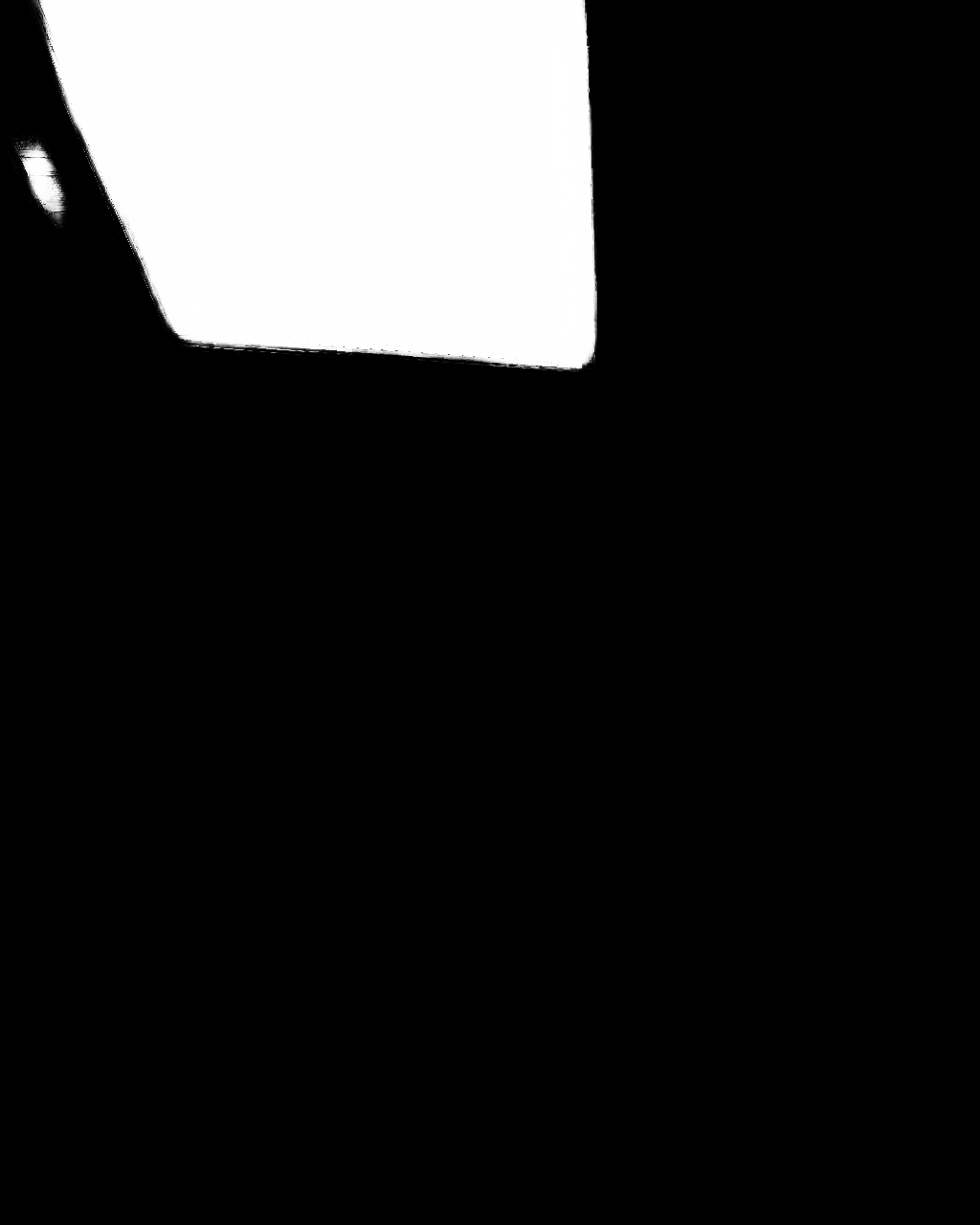}&
		\includegraphics[width=\wvisual, height=\hvisualtall]{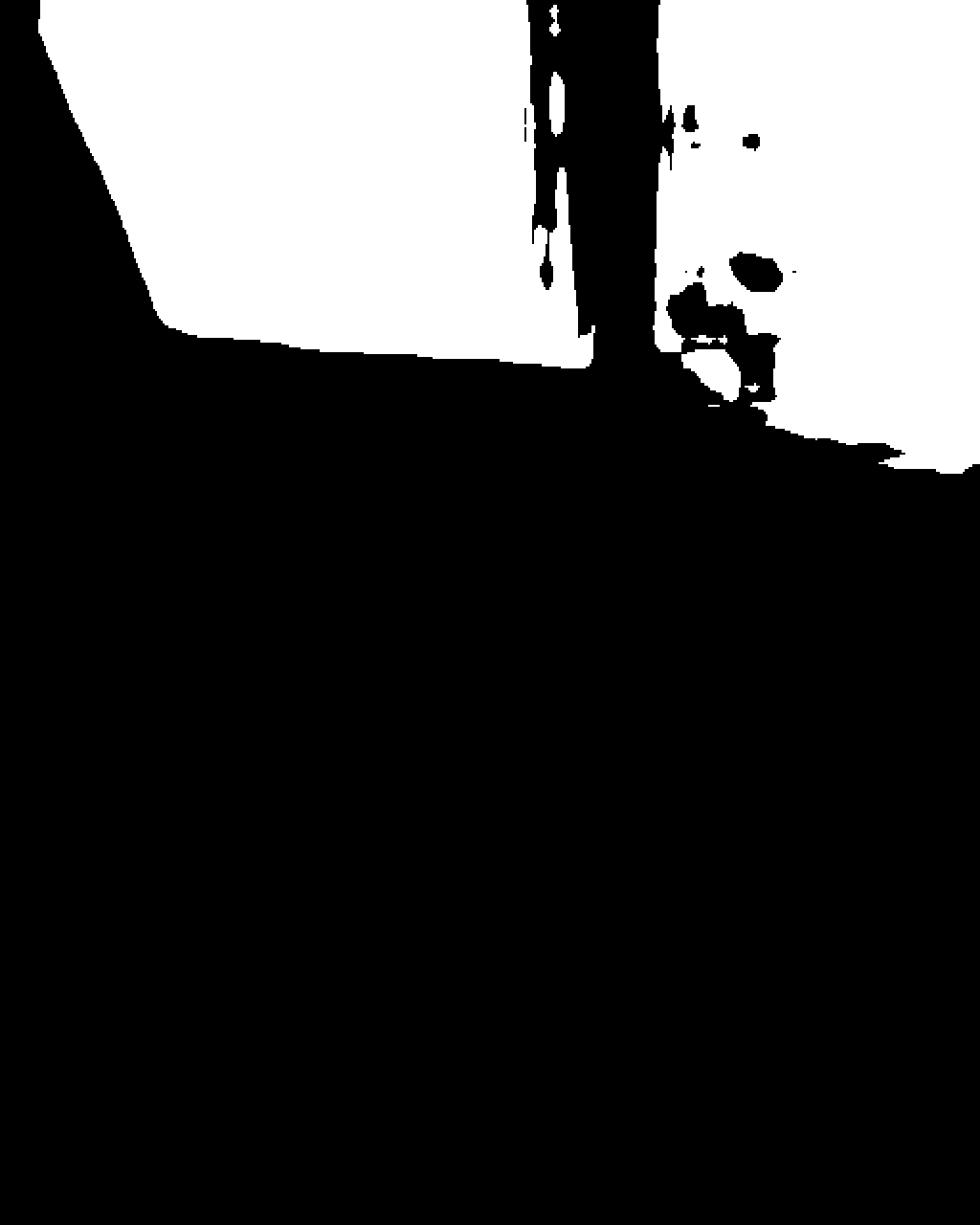}&
		\includegraphics[width=\wvisual, height=\hvisualtall]{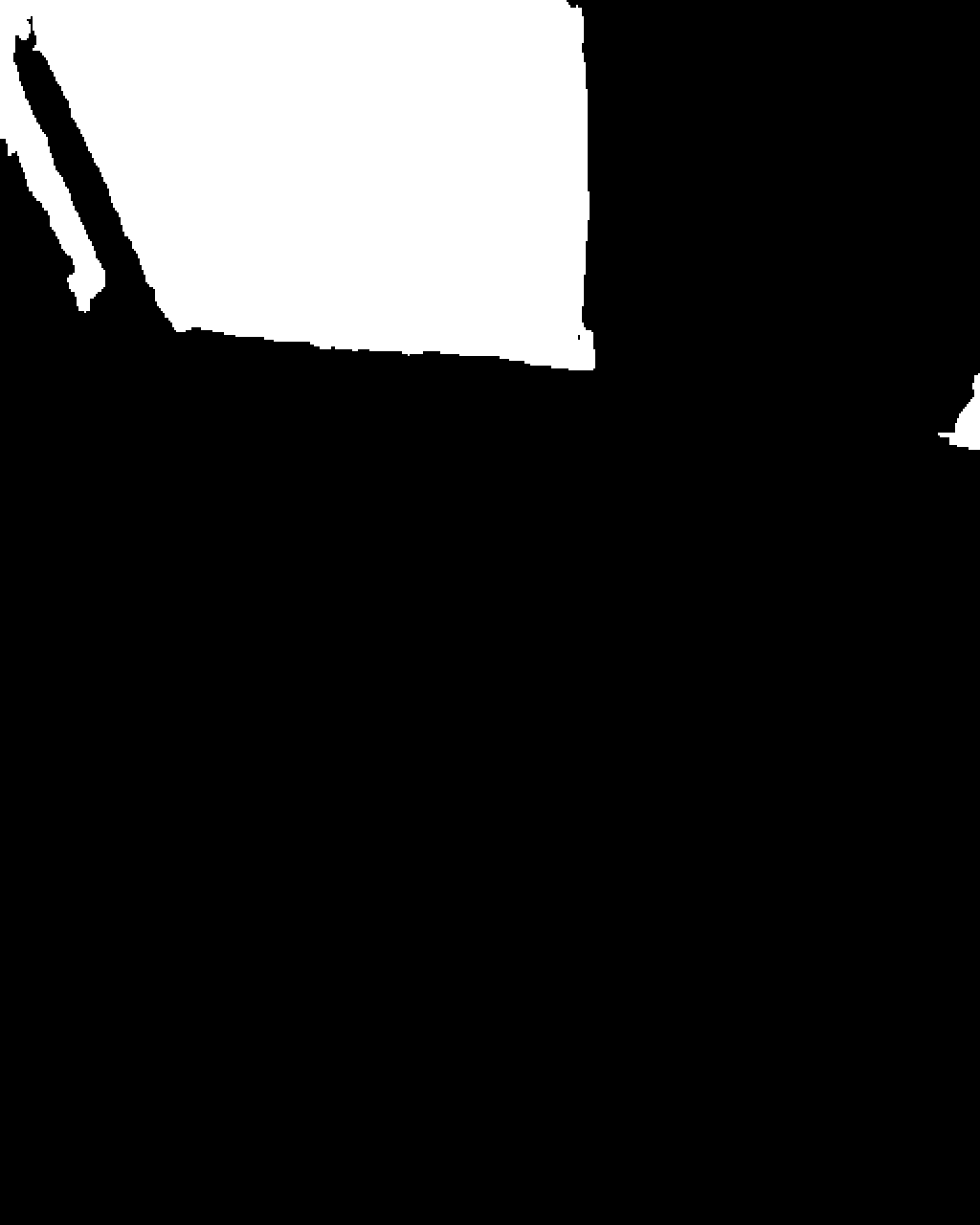}&
		\includegraphics[width=\wvisual, height=\hvisualtall]{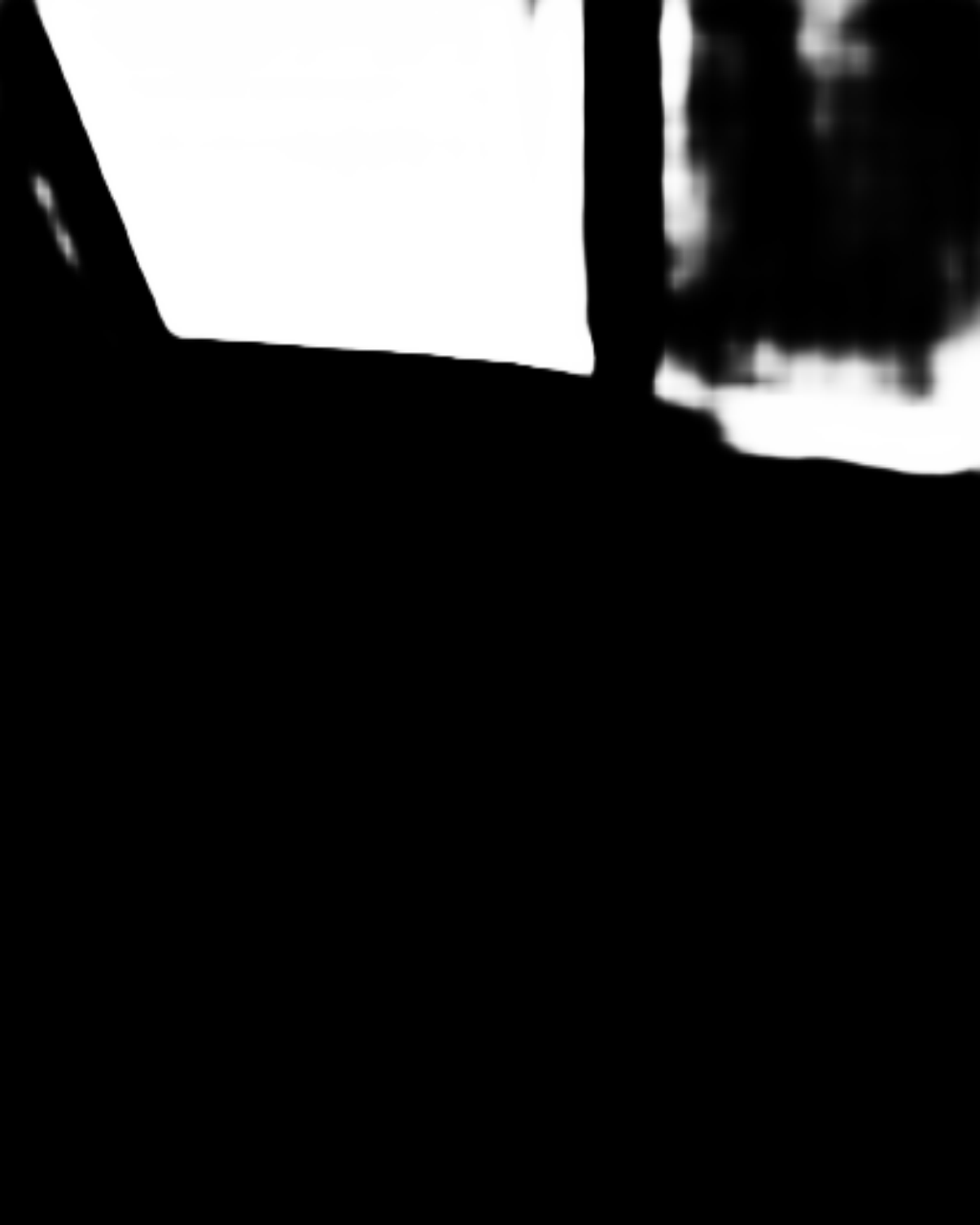}&
		\includegraphics[width=\wvisual, height=\hvisualtall]{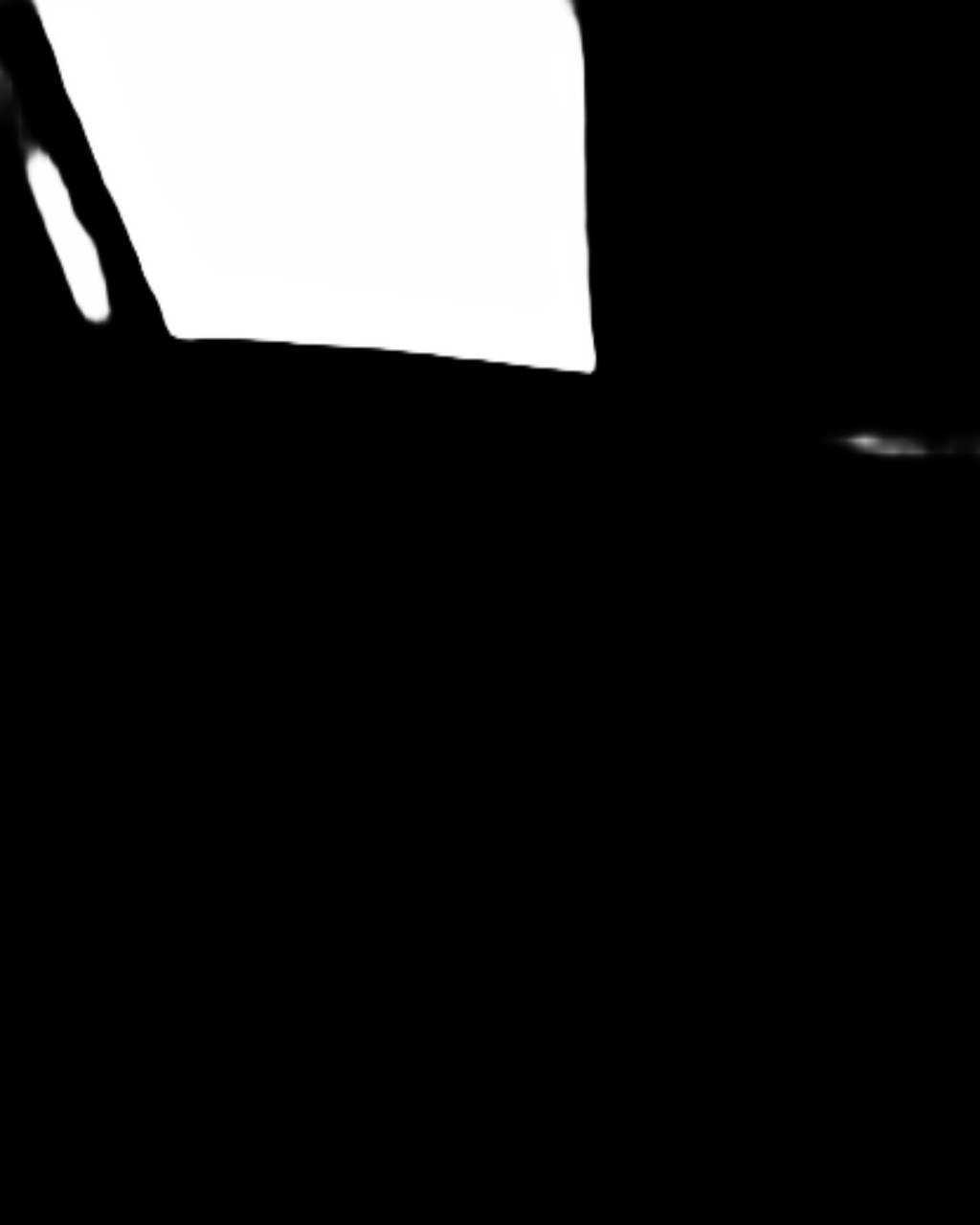}&
		\includegraphics[width=\wvisual, height=\hvisualtall]{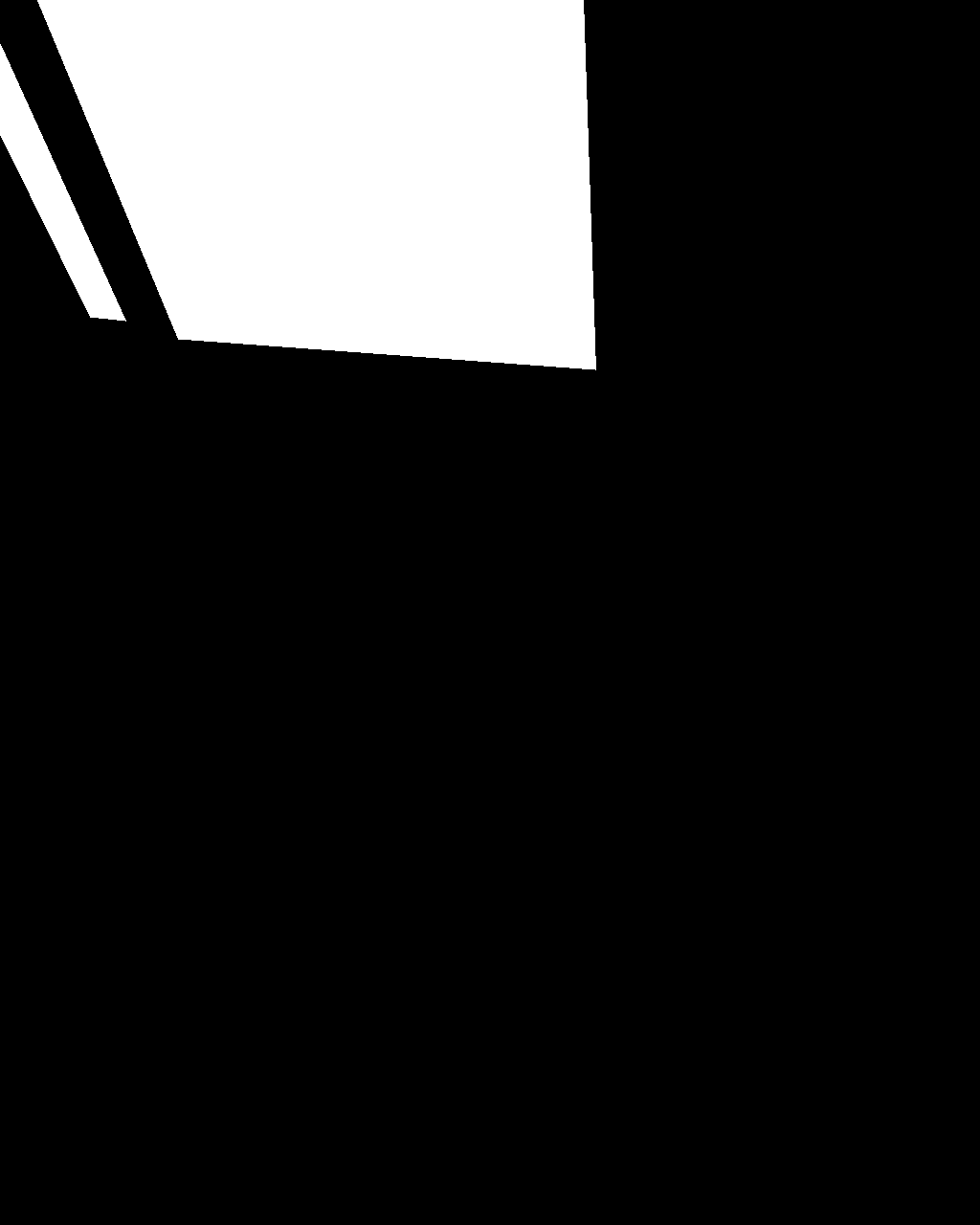} \\
		
		\includegraphics[width=\wvisual, height=\hvisualtall]{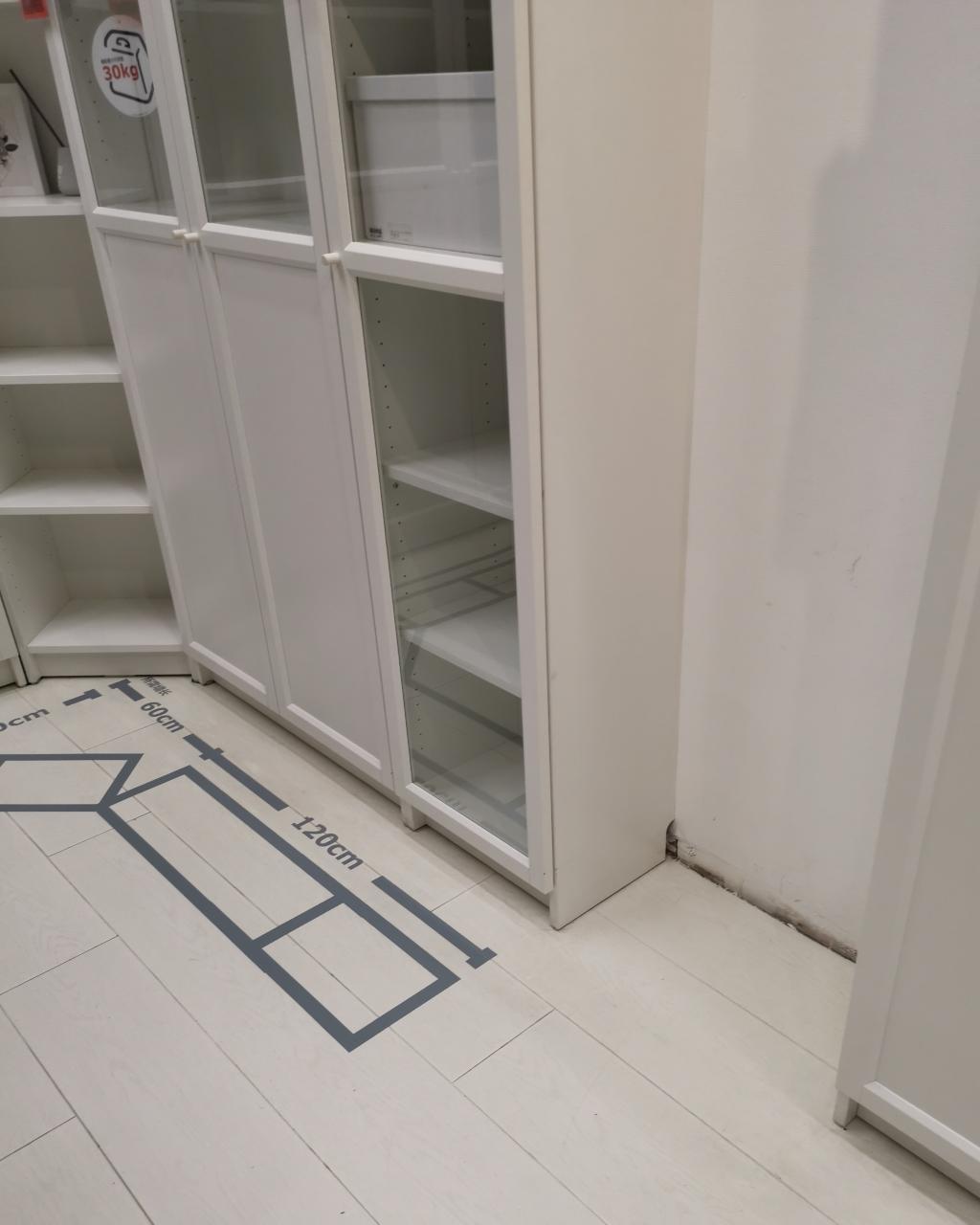}&
		\includegraphics[width=\wvisual, height=\hvisualtall]{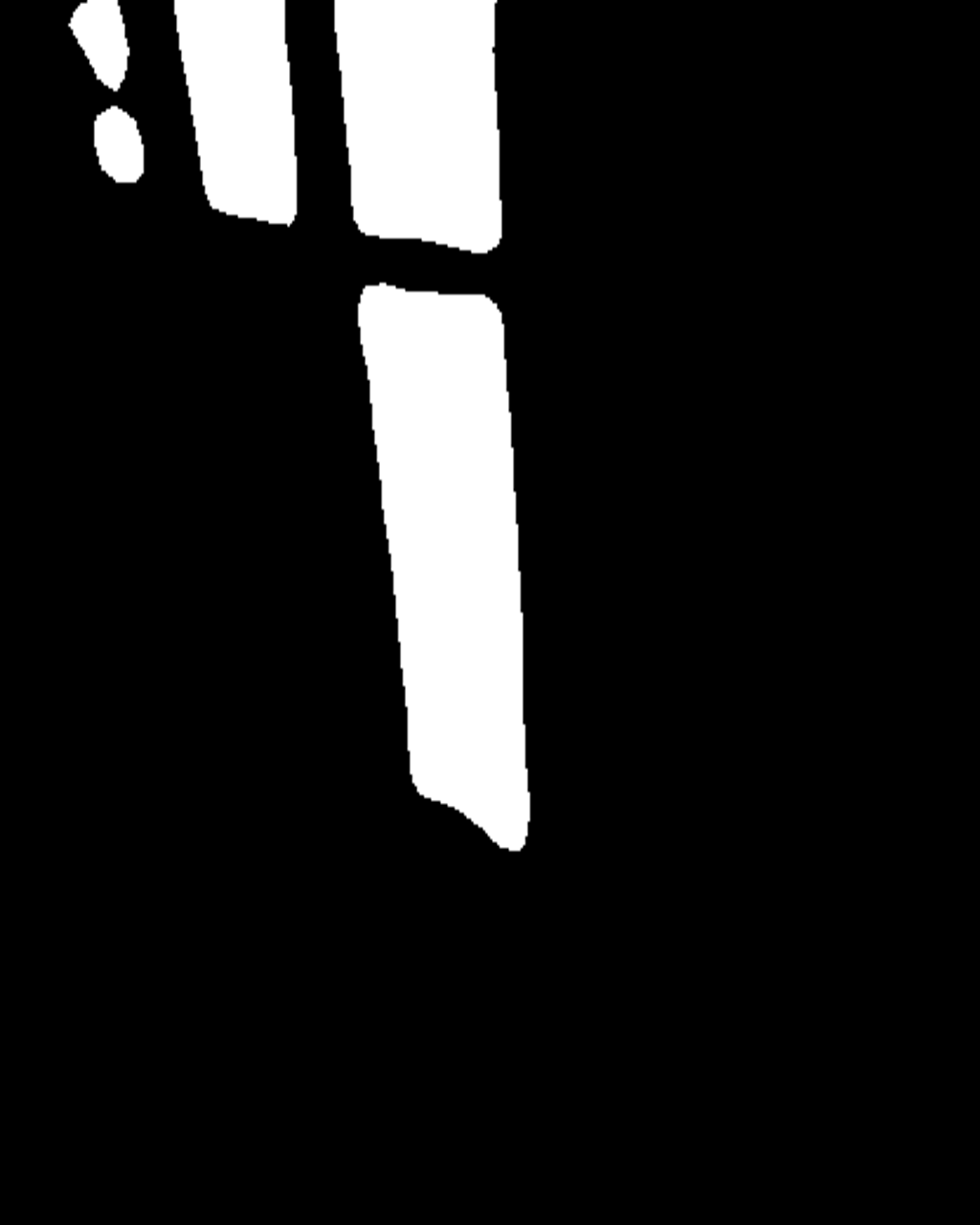}&
		\includegraphics[width=\wvisual, height=\hvisualtall]{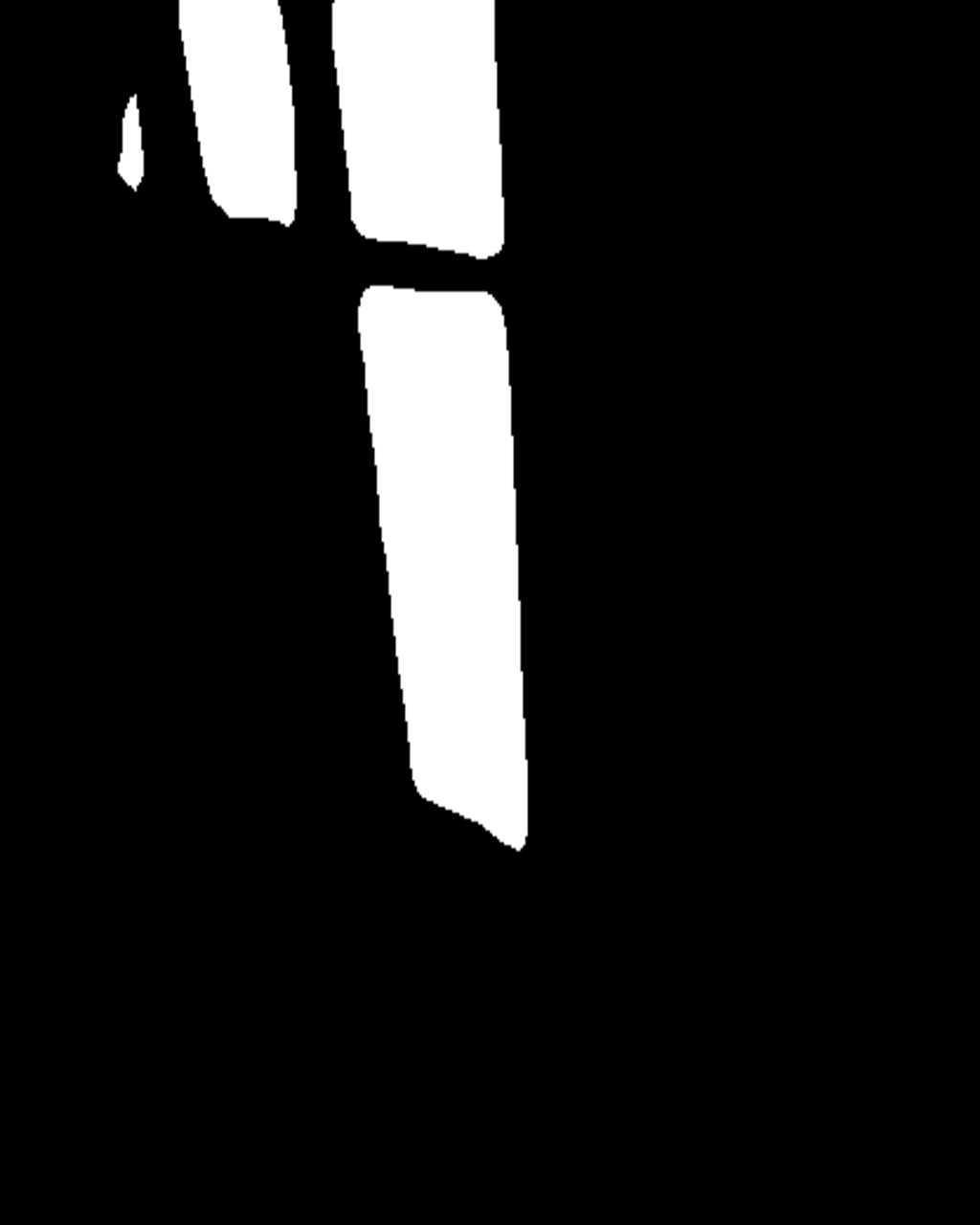}&
		\includegraphics[width=\wvisual, height=\hvisualtall]{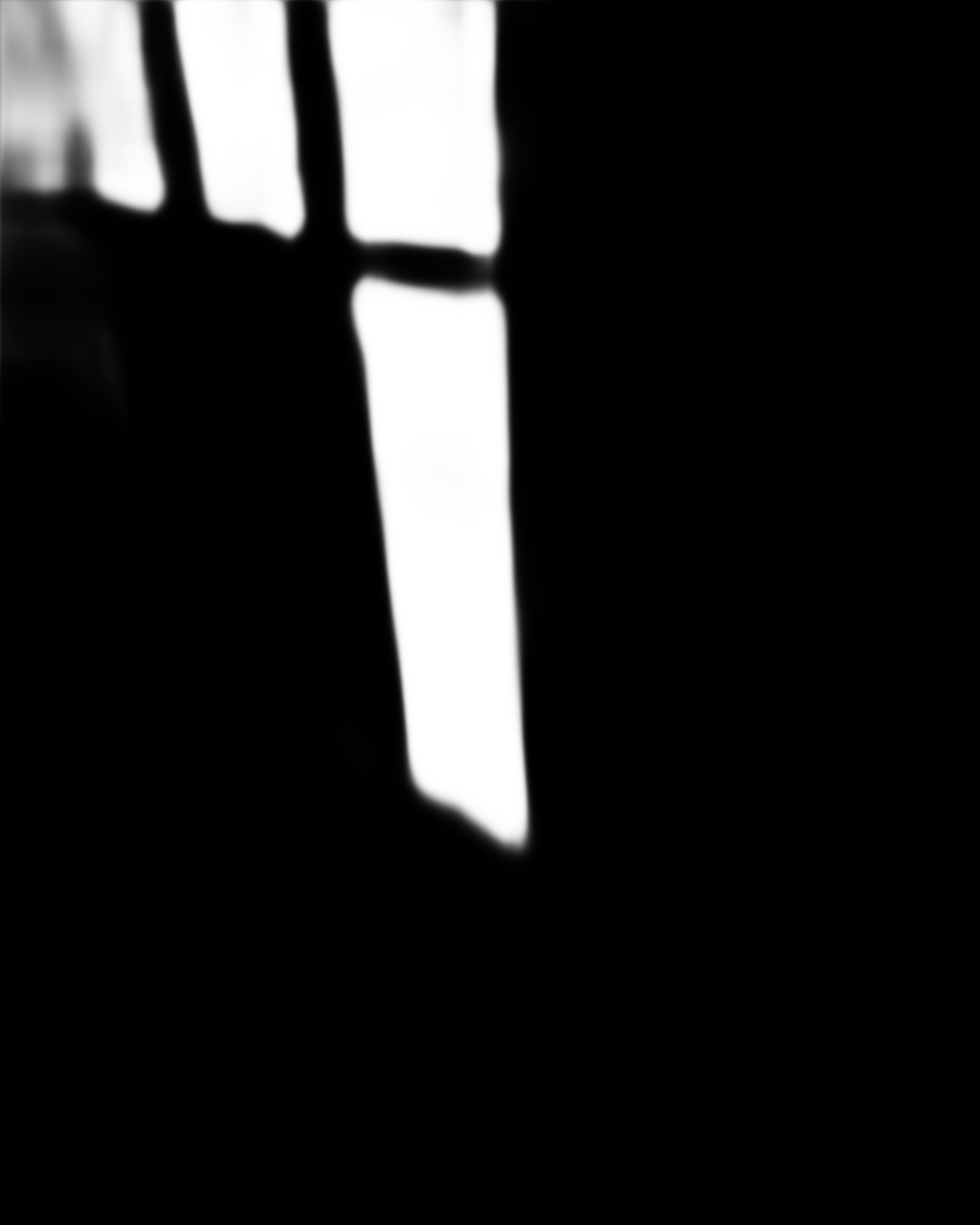}&
		\includegraphics[width=\wvisual, height=\hvisualtall]{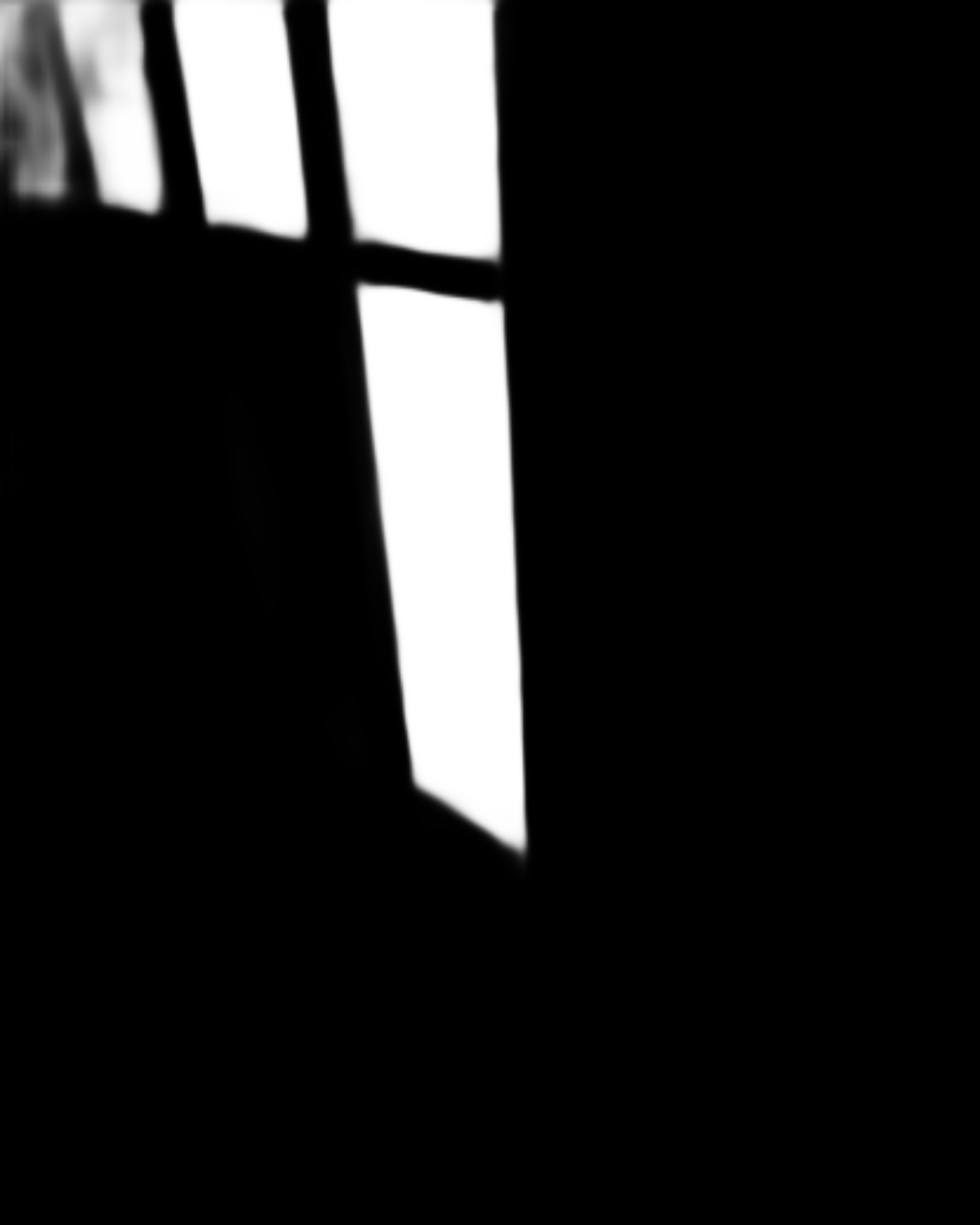}&
		\includegraphics[width=\wvisual, height=\hvisualtall]{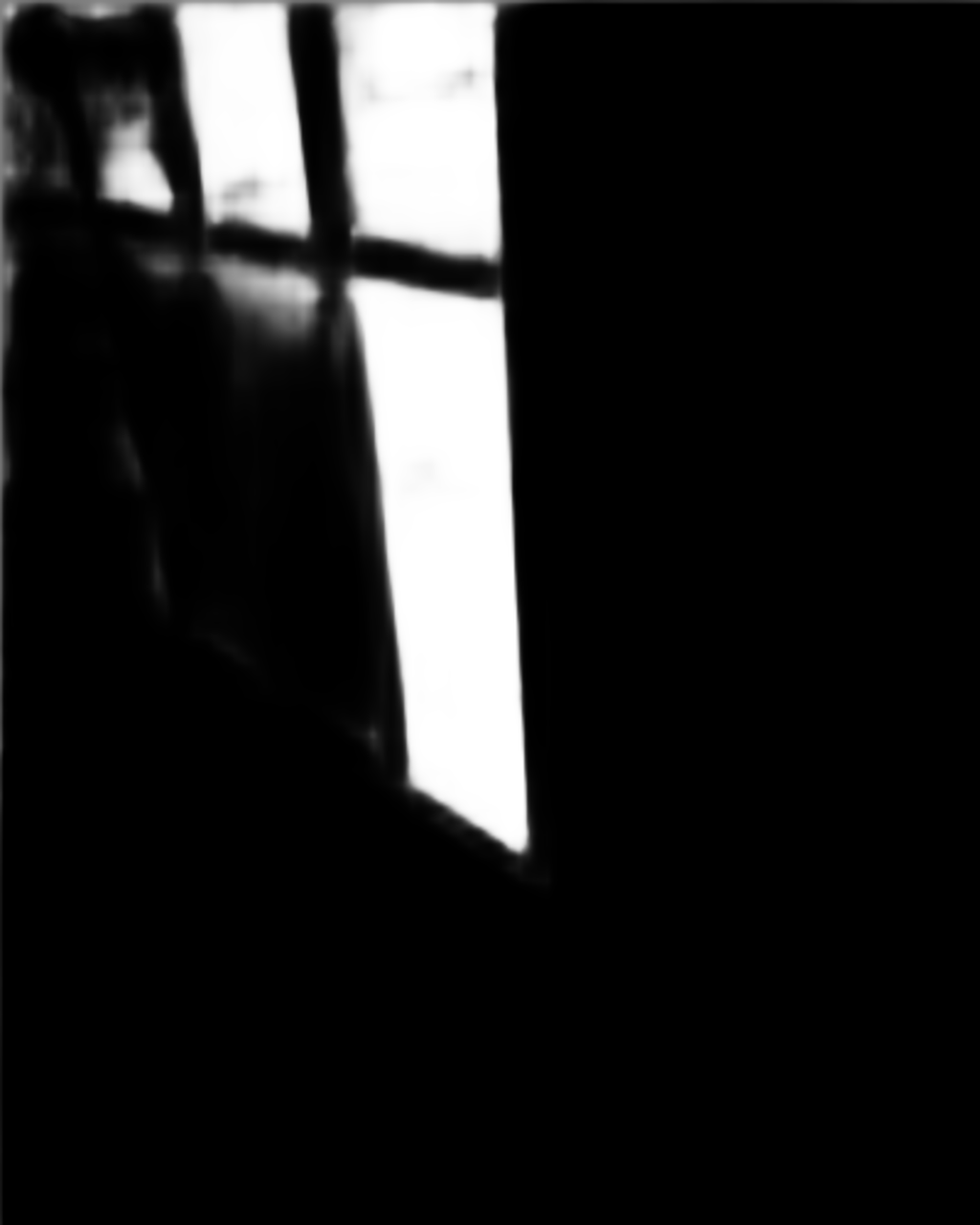}&
		\includegraphics[width=\wvisual, height=\hvisualtall]{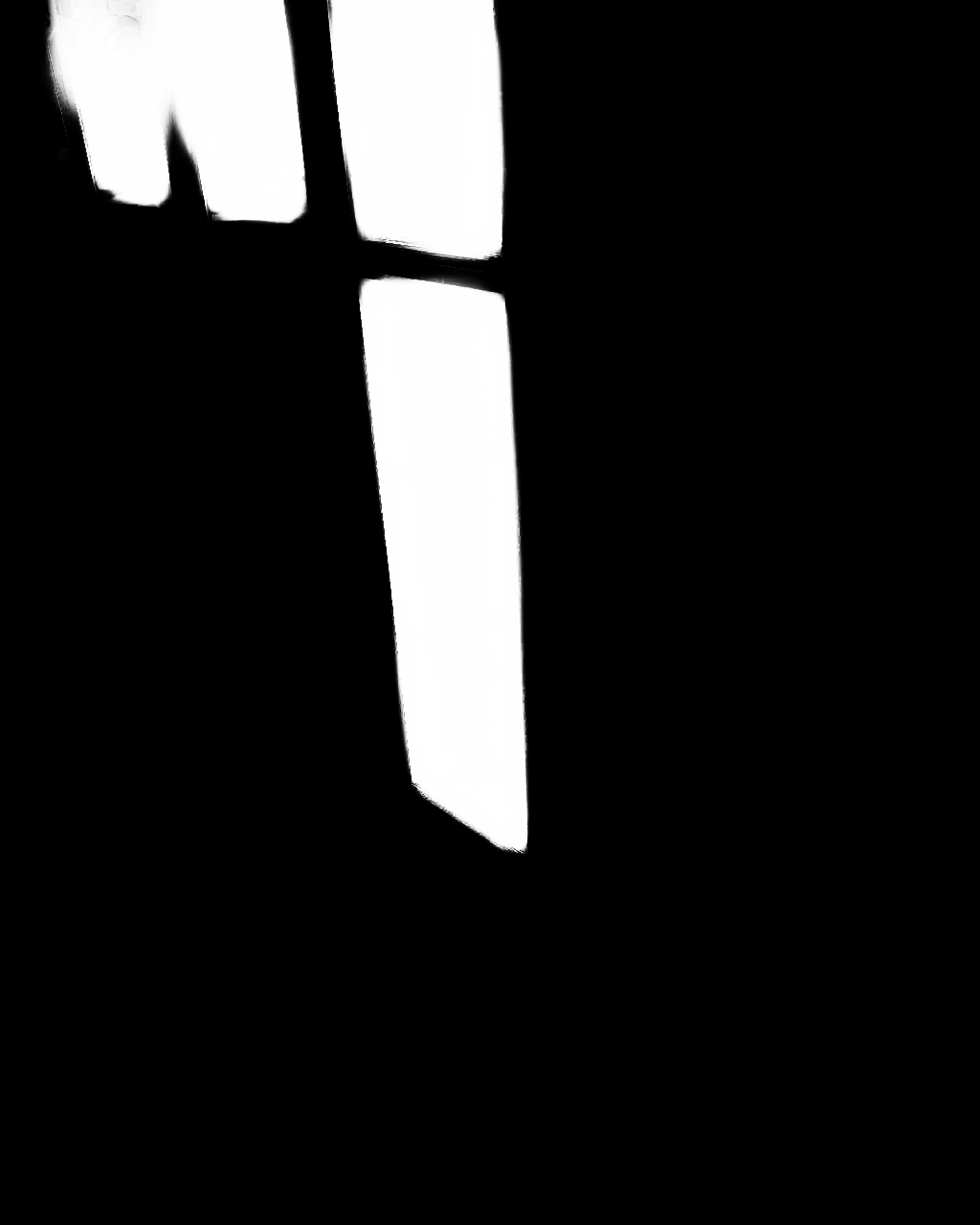}&
		\includegraphics[width=\wvisual, height=\hvisualtall]{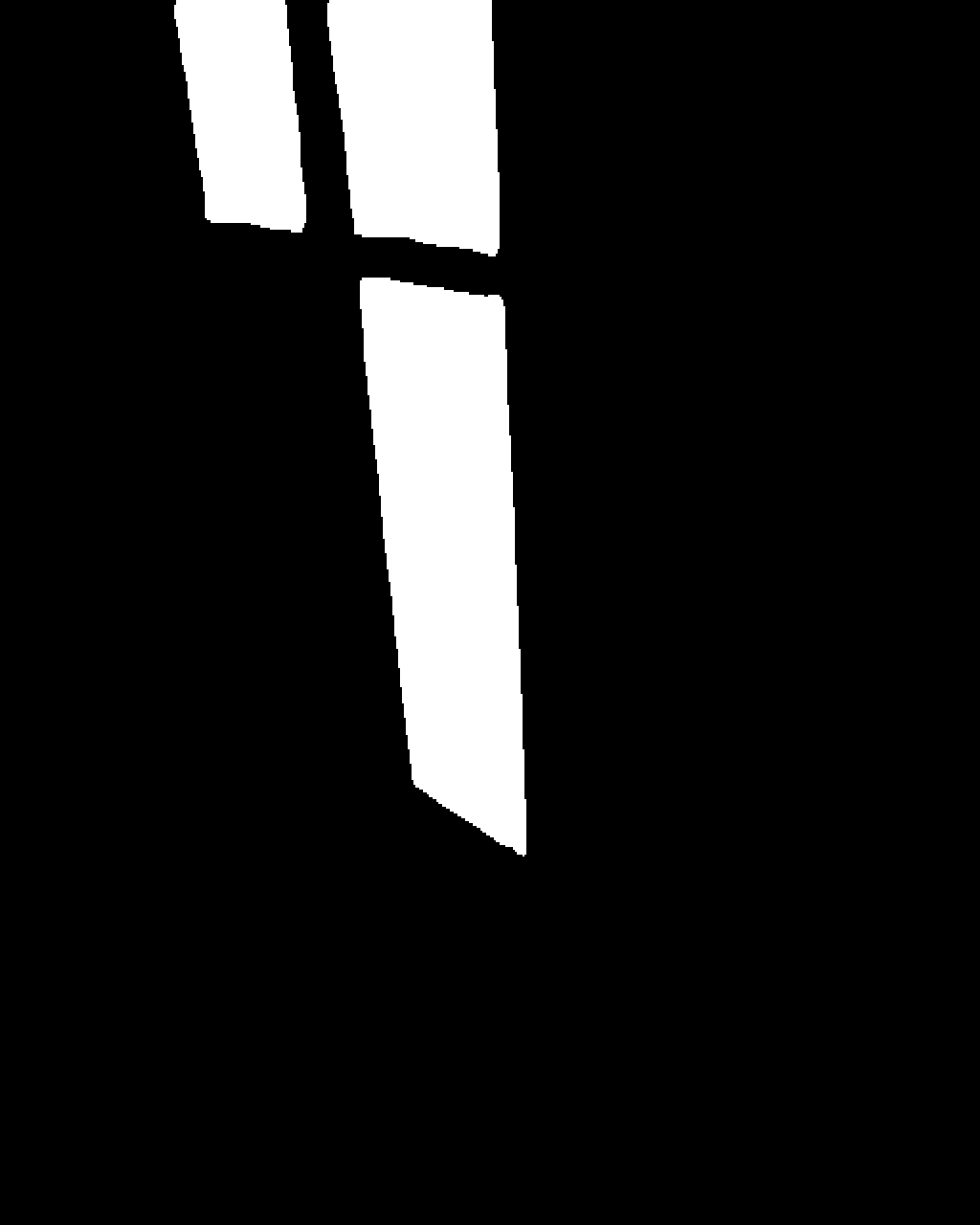}&
		\includegraphics[width=\wvisual, height=\hvisualtall]{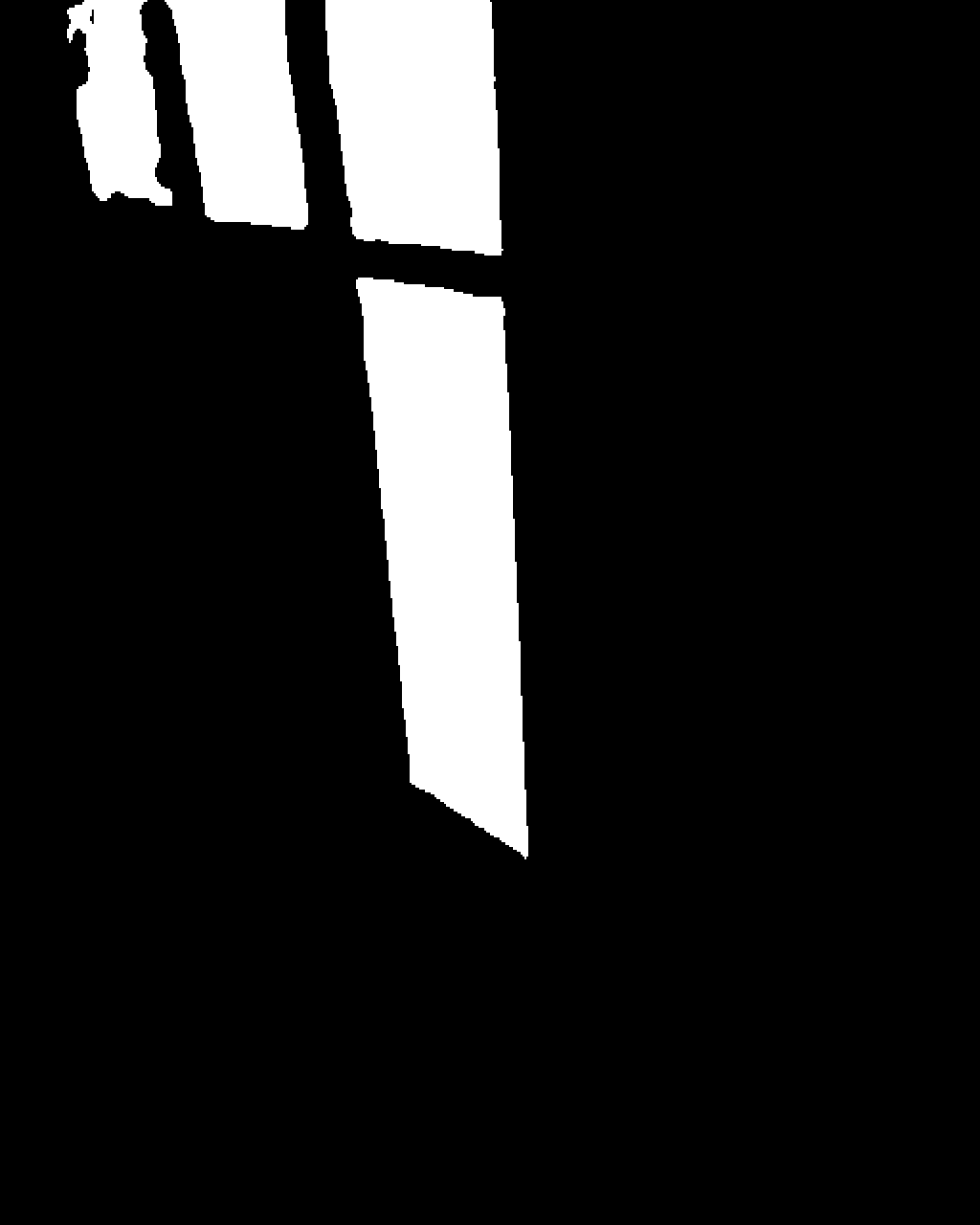}&
		\includegraphics[width=\wvisual, height=\hvisualtall]{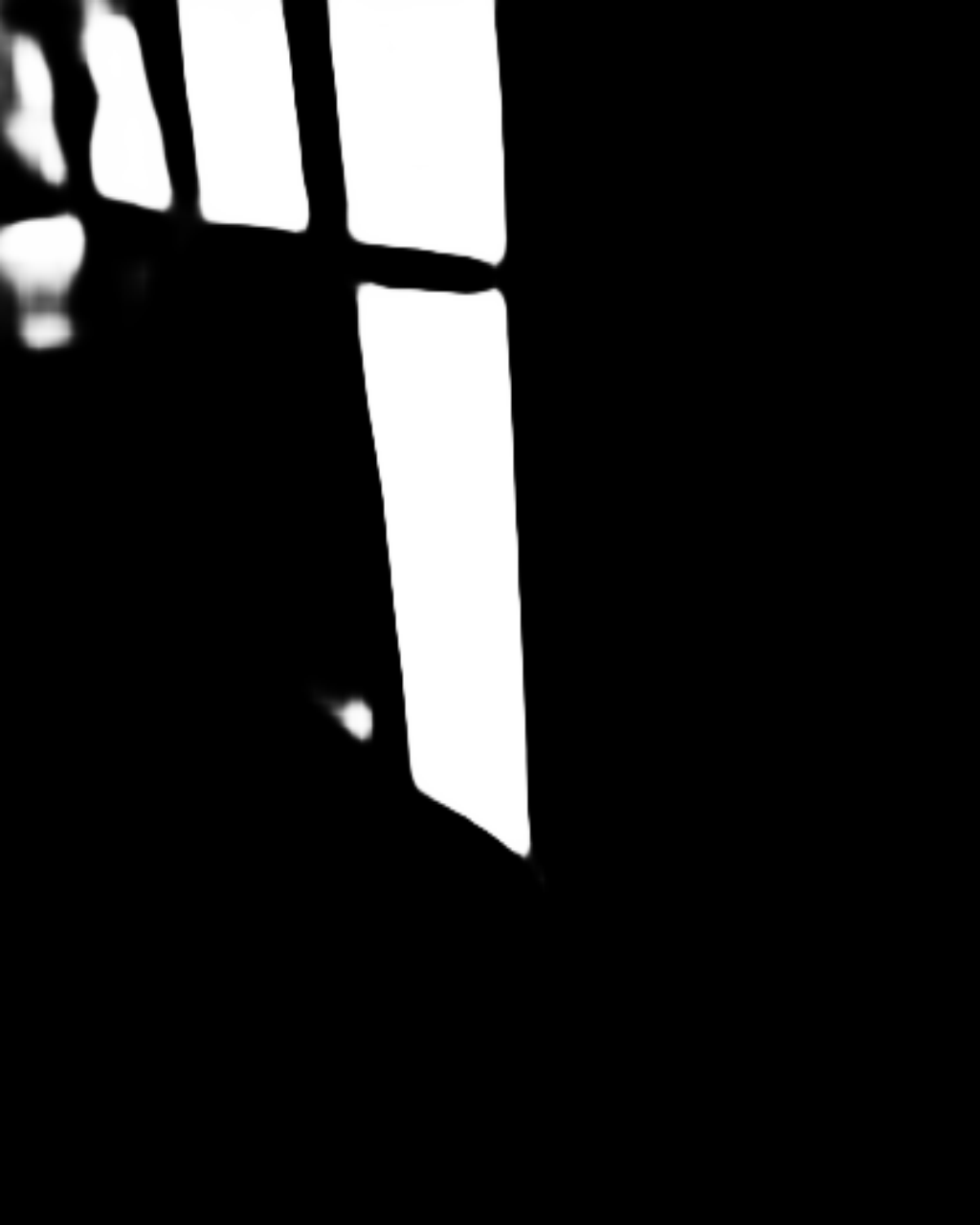}&
		\includegraphics[width=\wvisual, height=\hvisualtall]{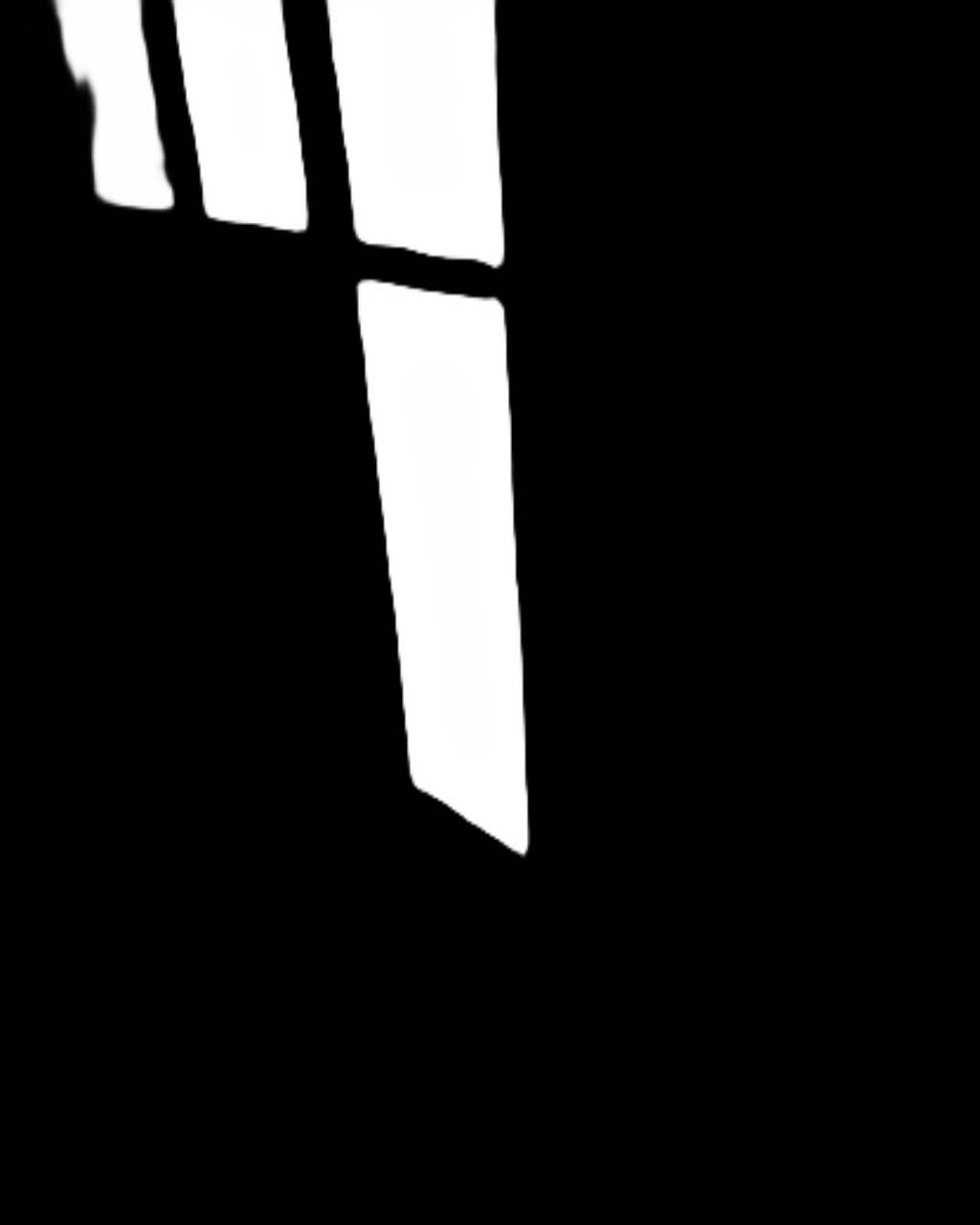}&
		\includegraphics[width=\wvisual, height=\hvisualtall]{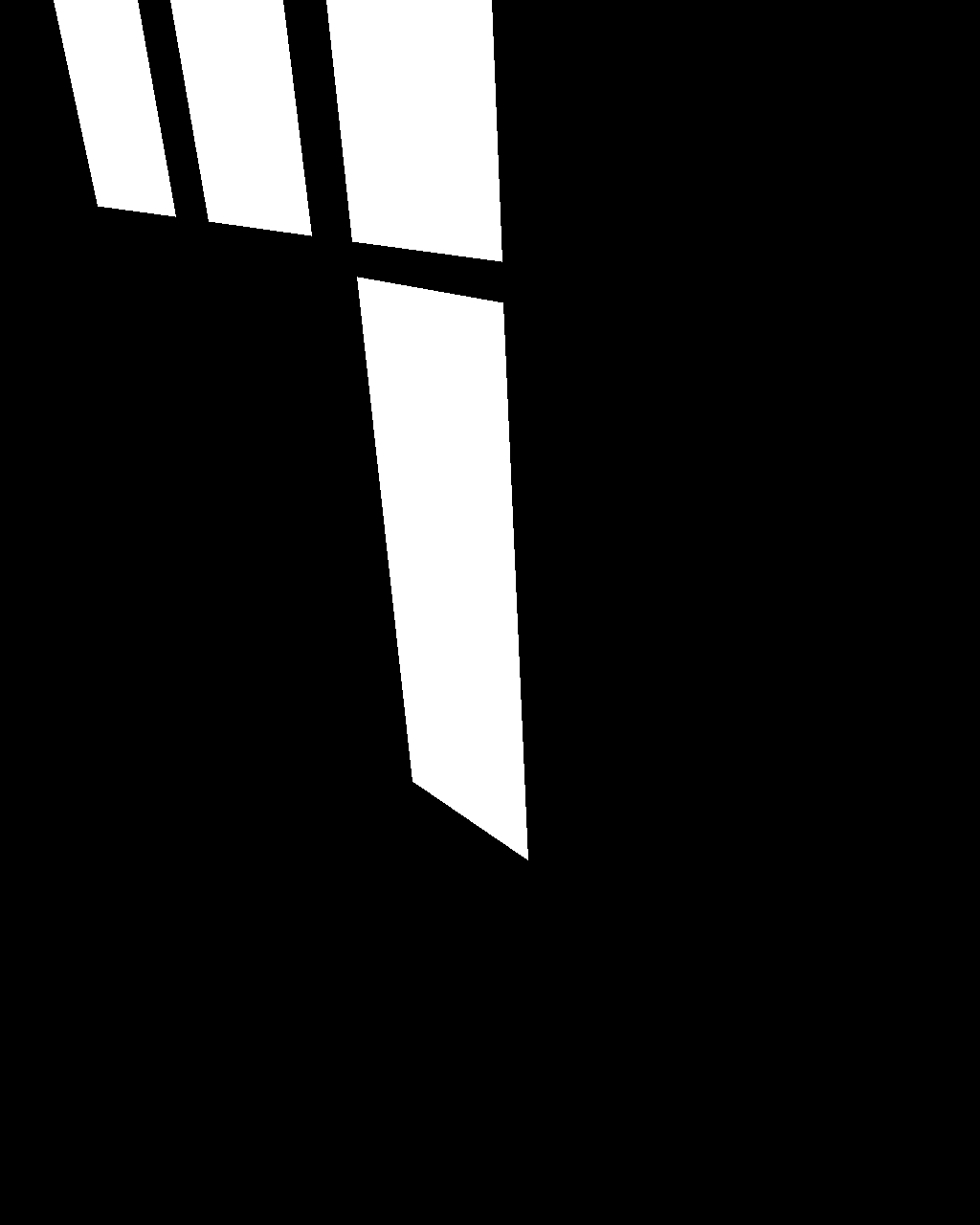} \\
		
		\includegraphics[width=\wvisual, height=\hvisualtall]{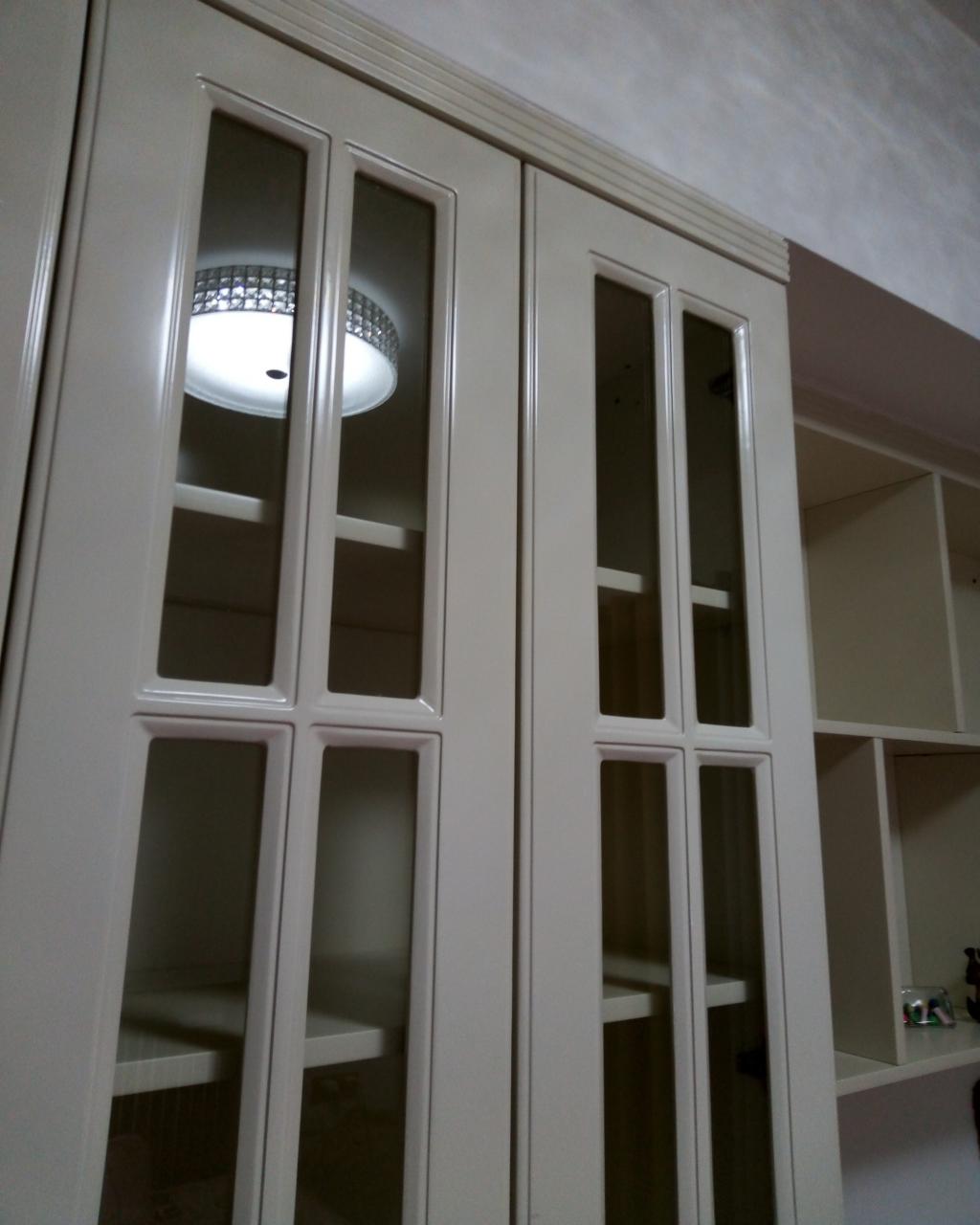}&
		\includegraphics[width=\wvisual, height=\hvisualtall]{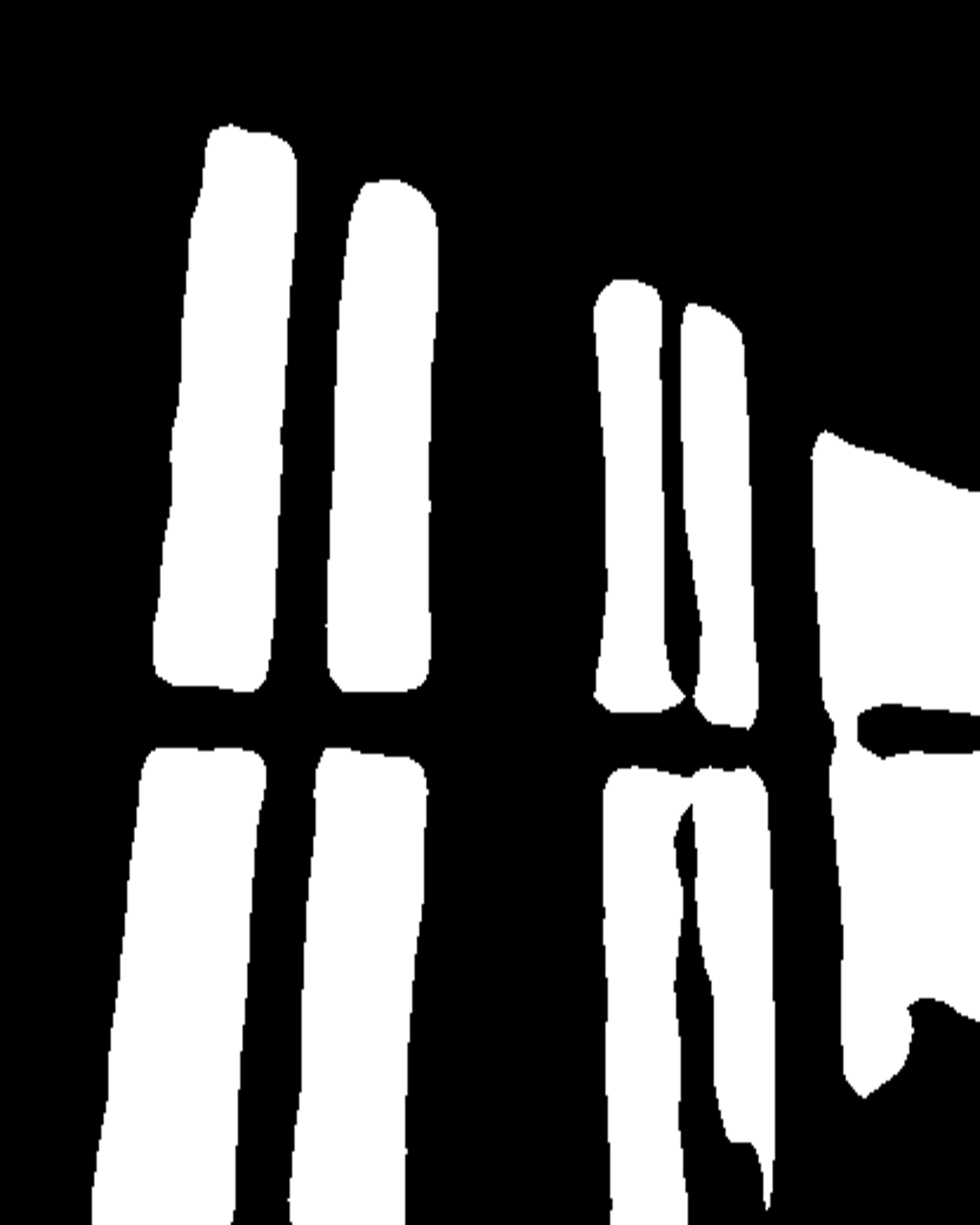}&
		\includegraphics[width=\wvisual, height=\hvisualtall]{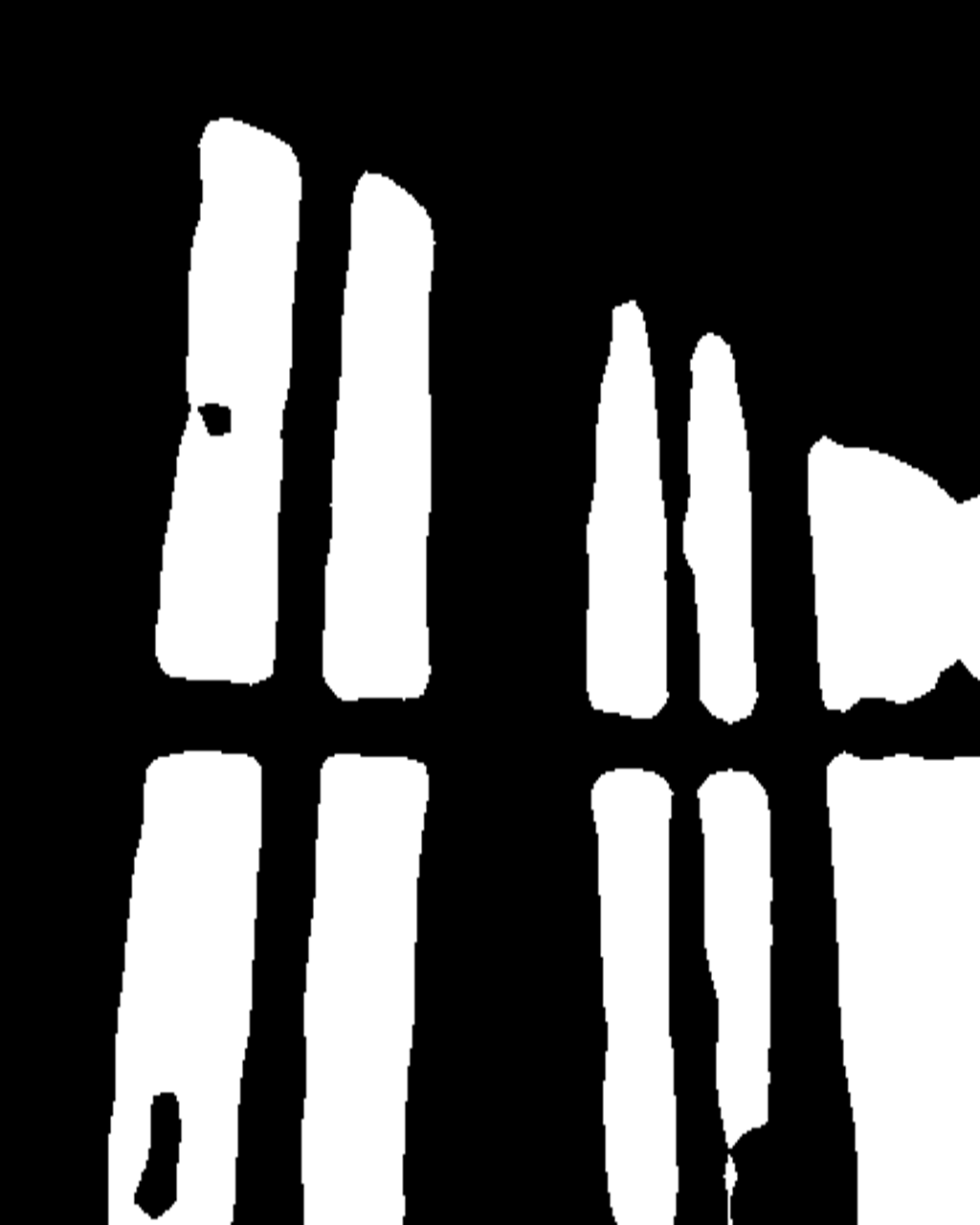}&
		\includegraphics[width=\wvisual, height=\hvisualtall]{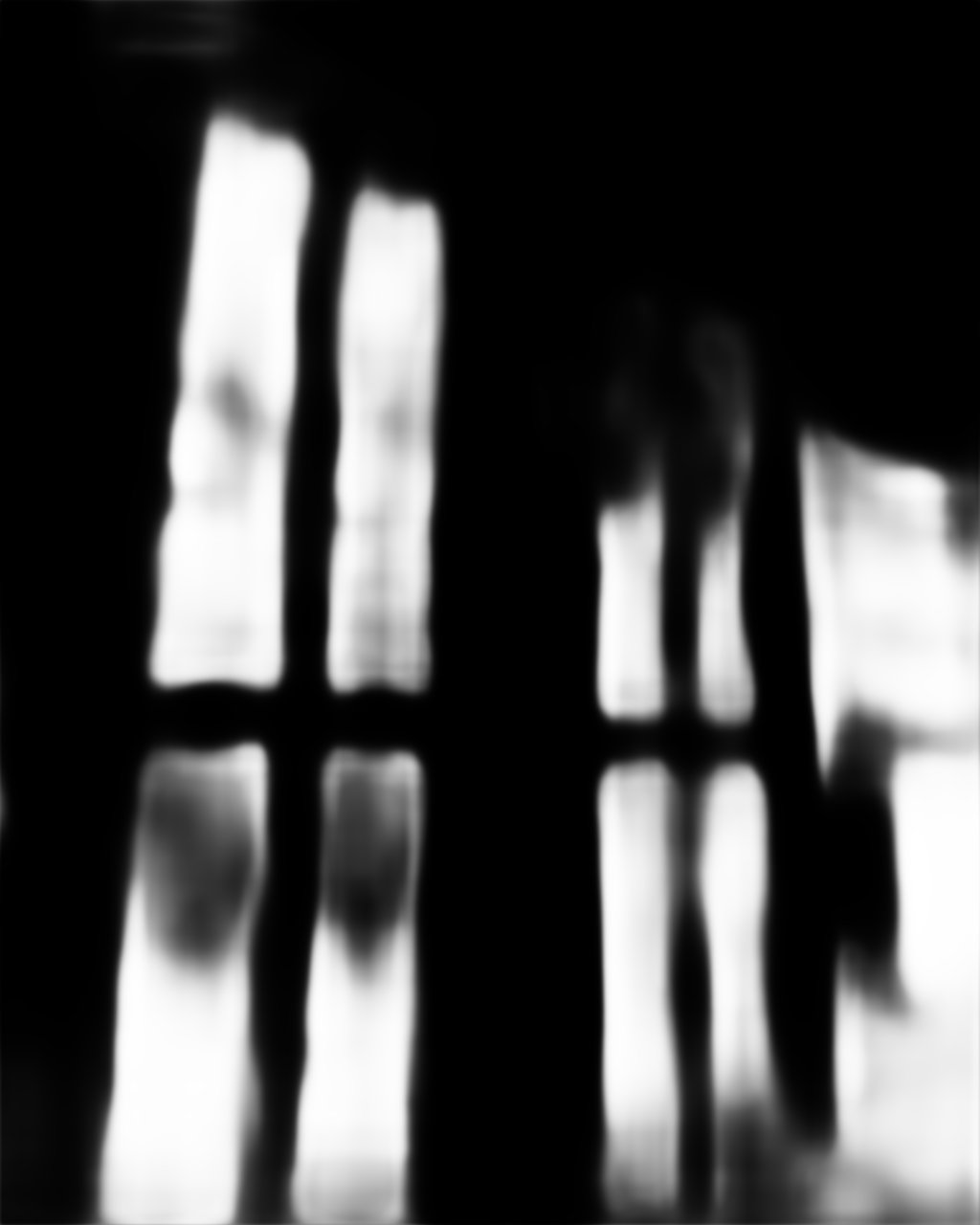}&
		\includegraphics[width=\wvisual, height=\hvisualtall]{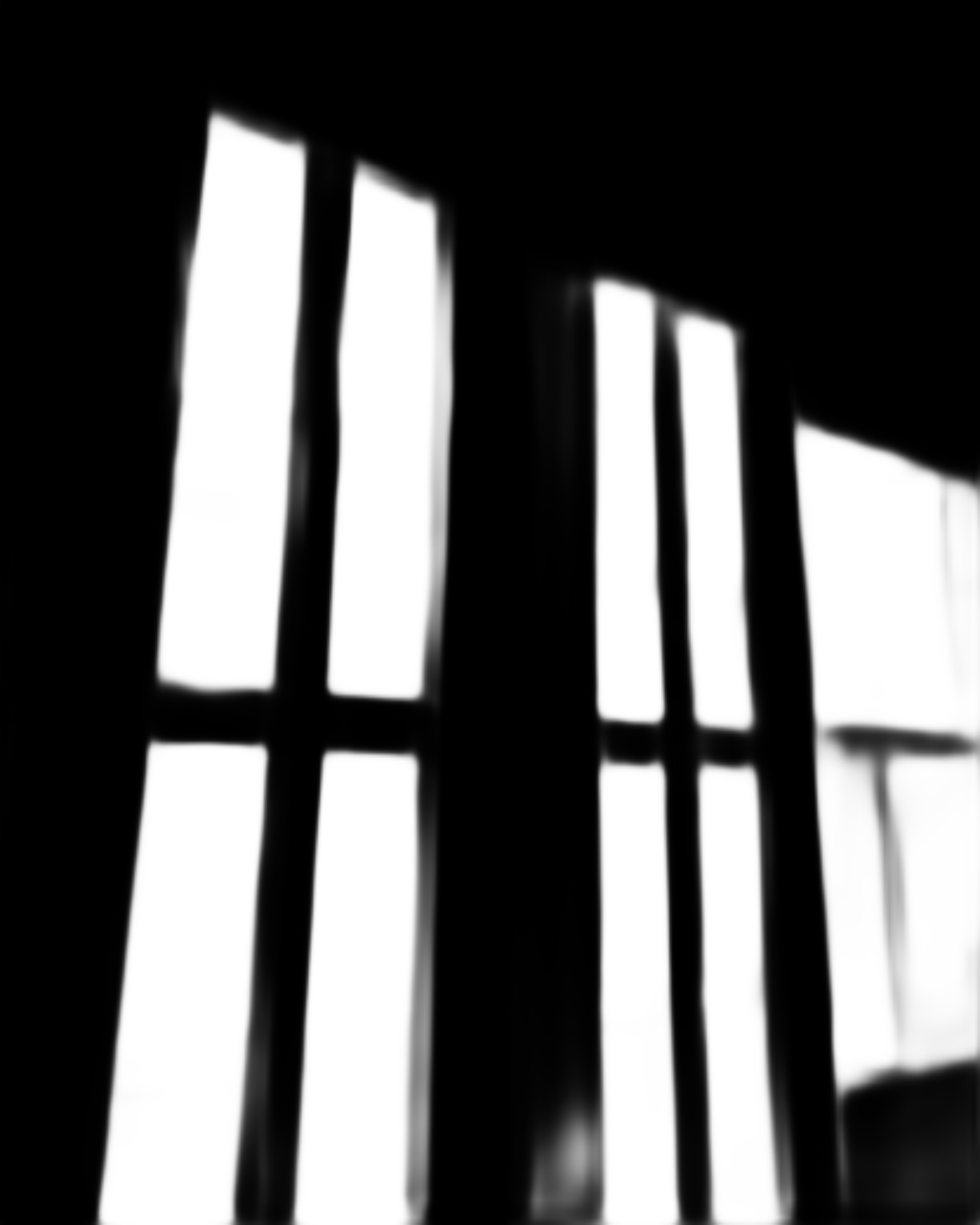}&
		\includegraphics[width=\wvisual, height=\hvisualtall]{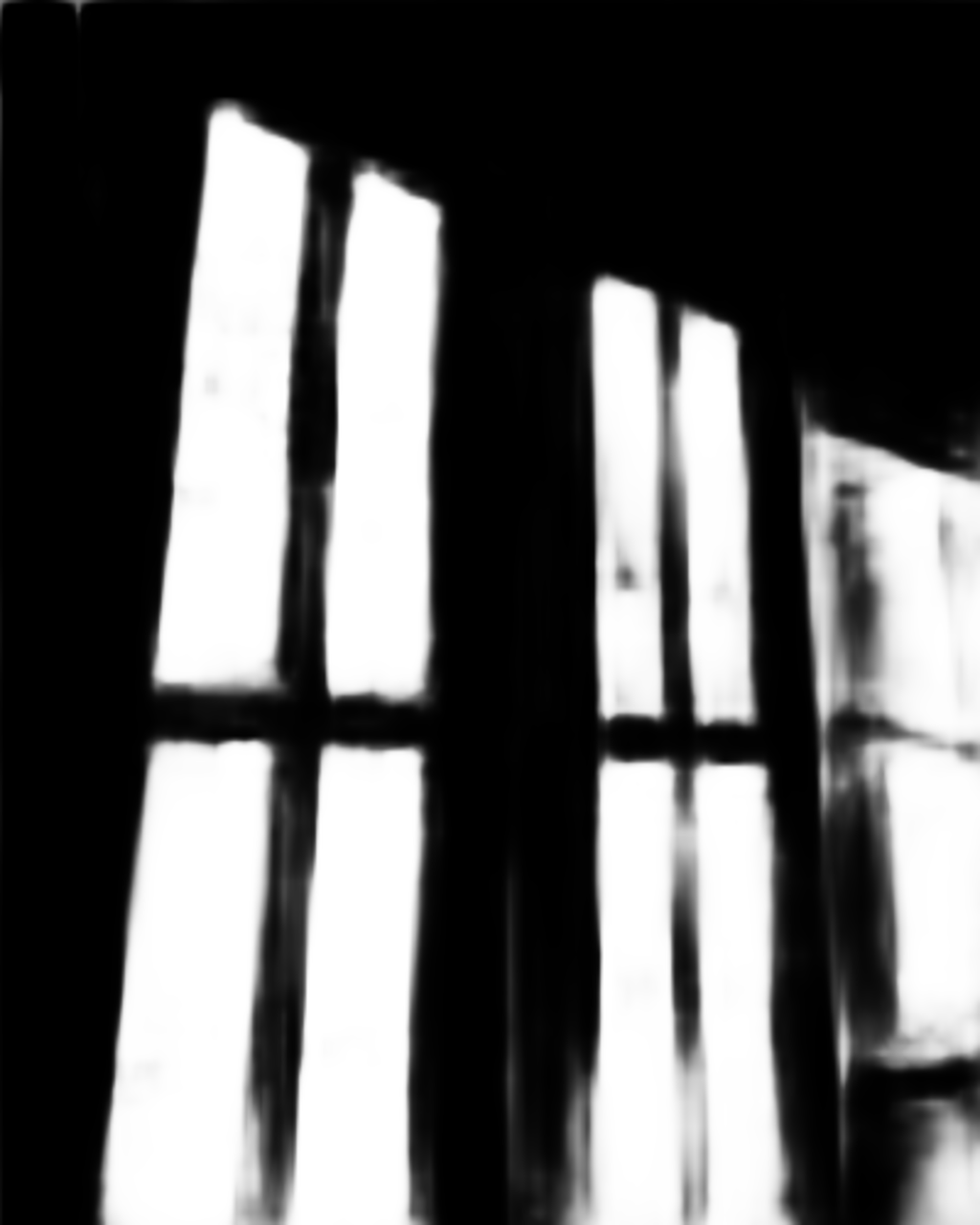}&
		\includegraphics[width=\wvisual, height=\hvisualtall]{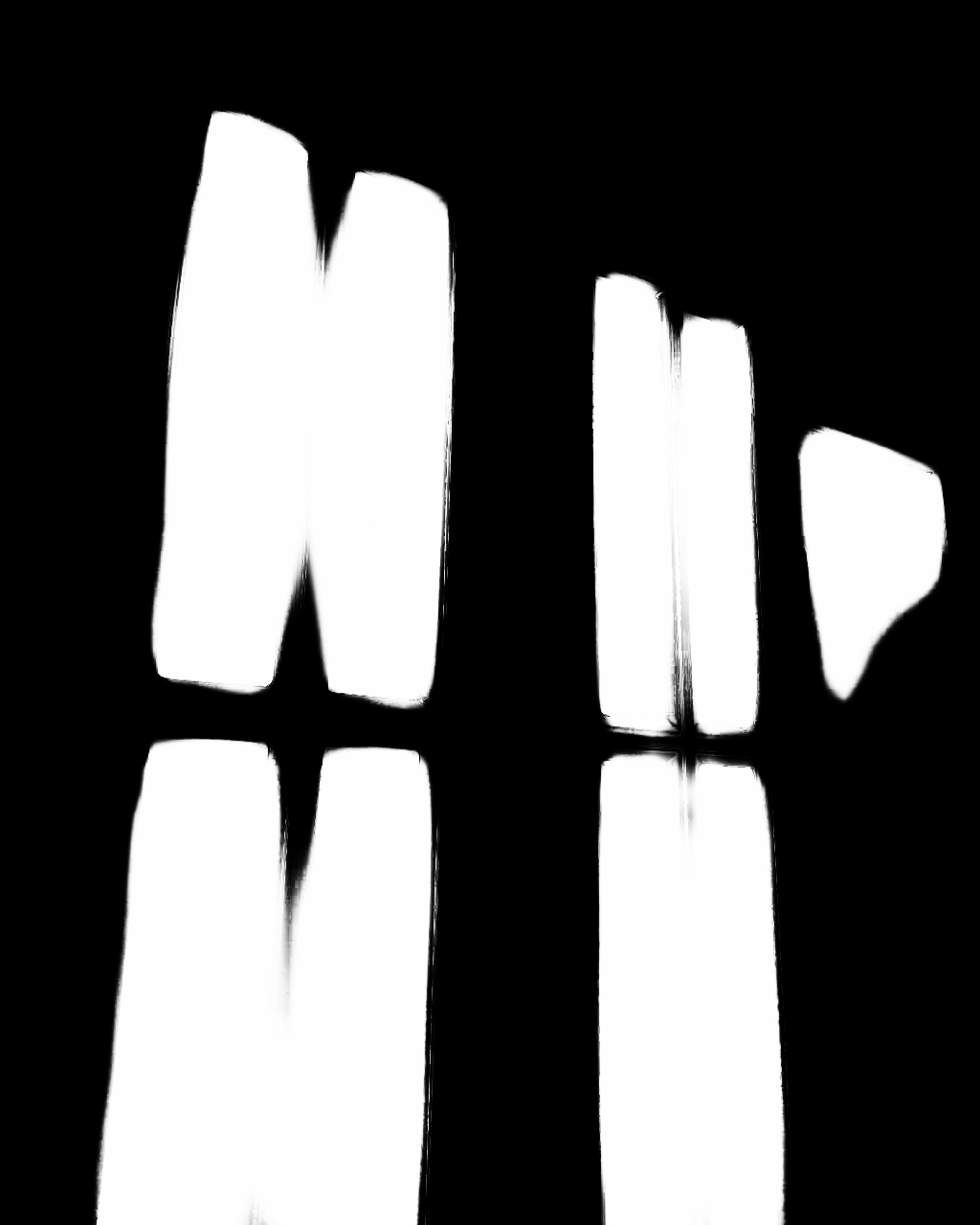}&
		\includegraphics[width=\wvisual, height=\hvisualtall]{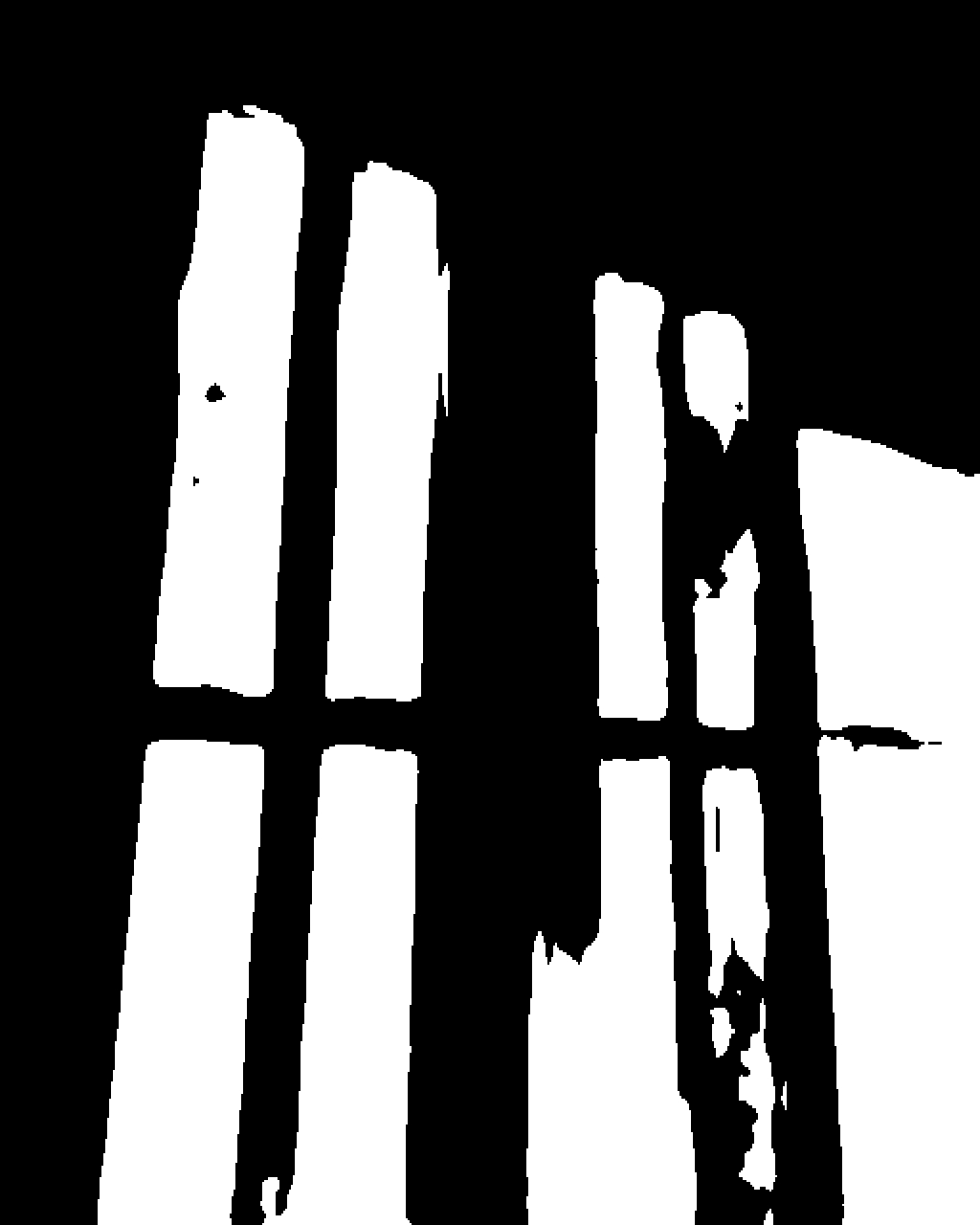}&
		\includegraphics[width=\wvisual, height=\hvisualtall]{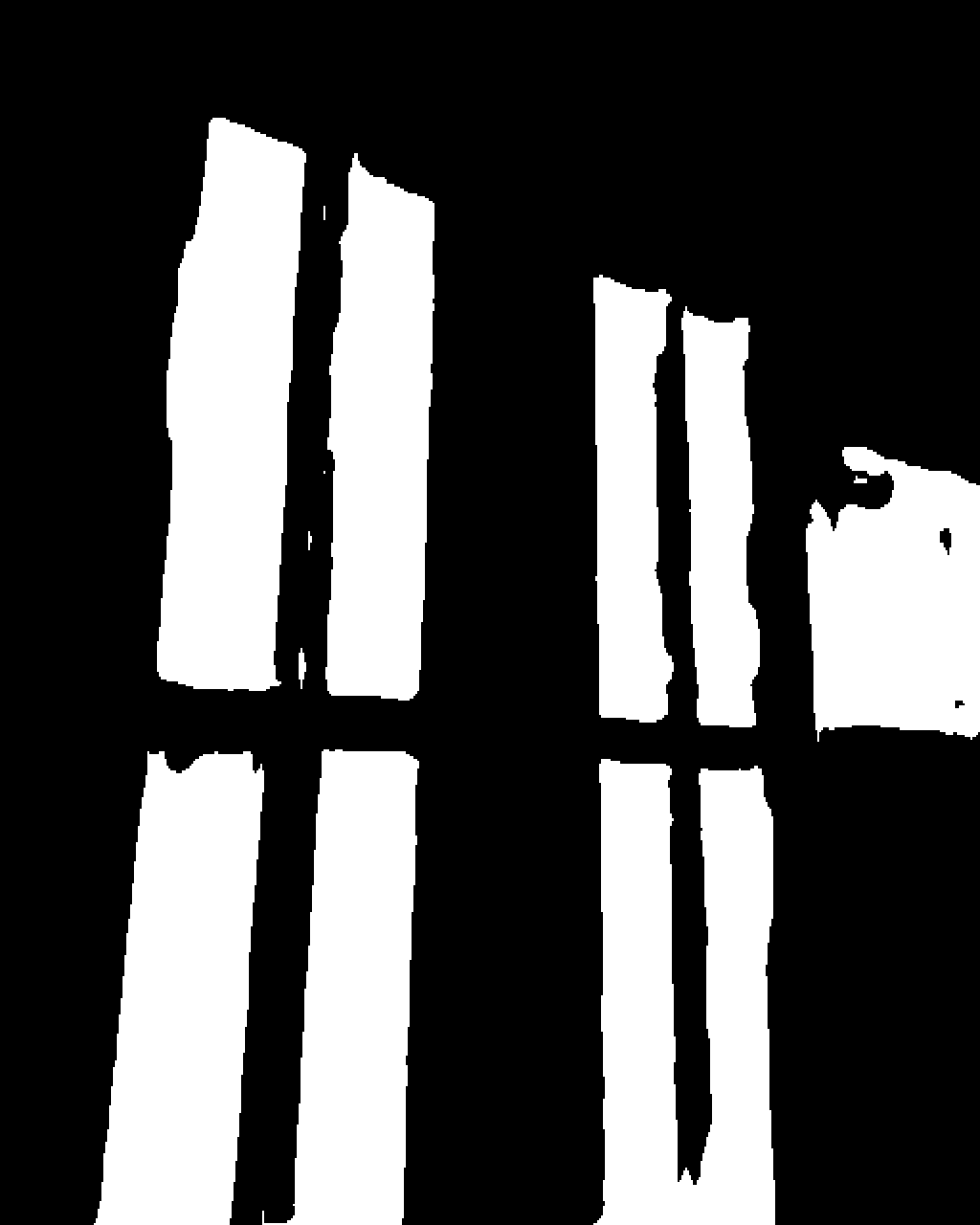}&
		\includegraphics[width=\wvisual, height=\hvisualtall]{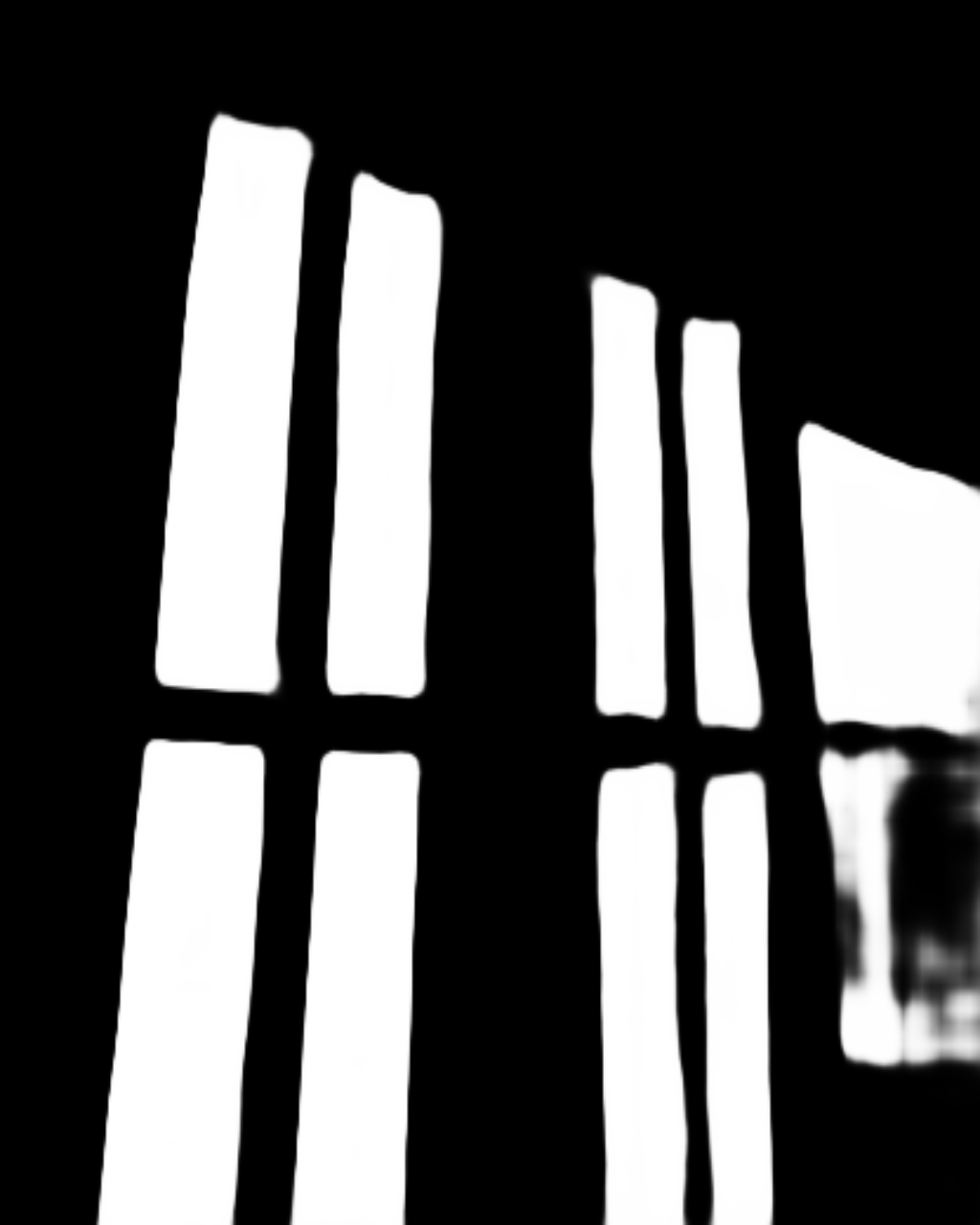}&
		\includegraphics[width=\wvisual, height=\hvisualtall]{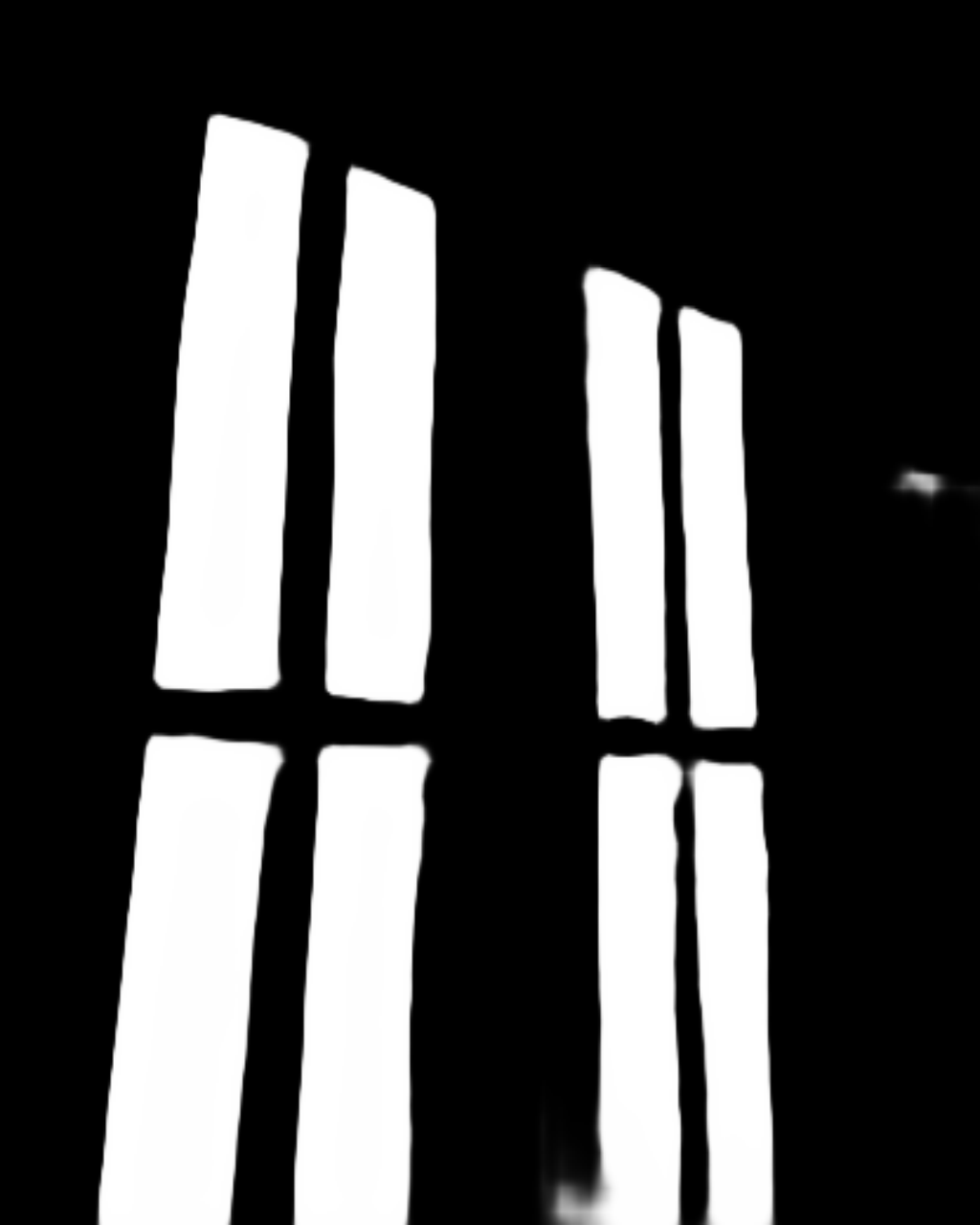}&
		\includegraphics[width=\wvisual, height=\hvisualtall]{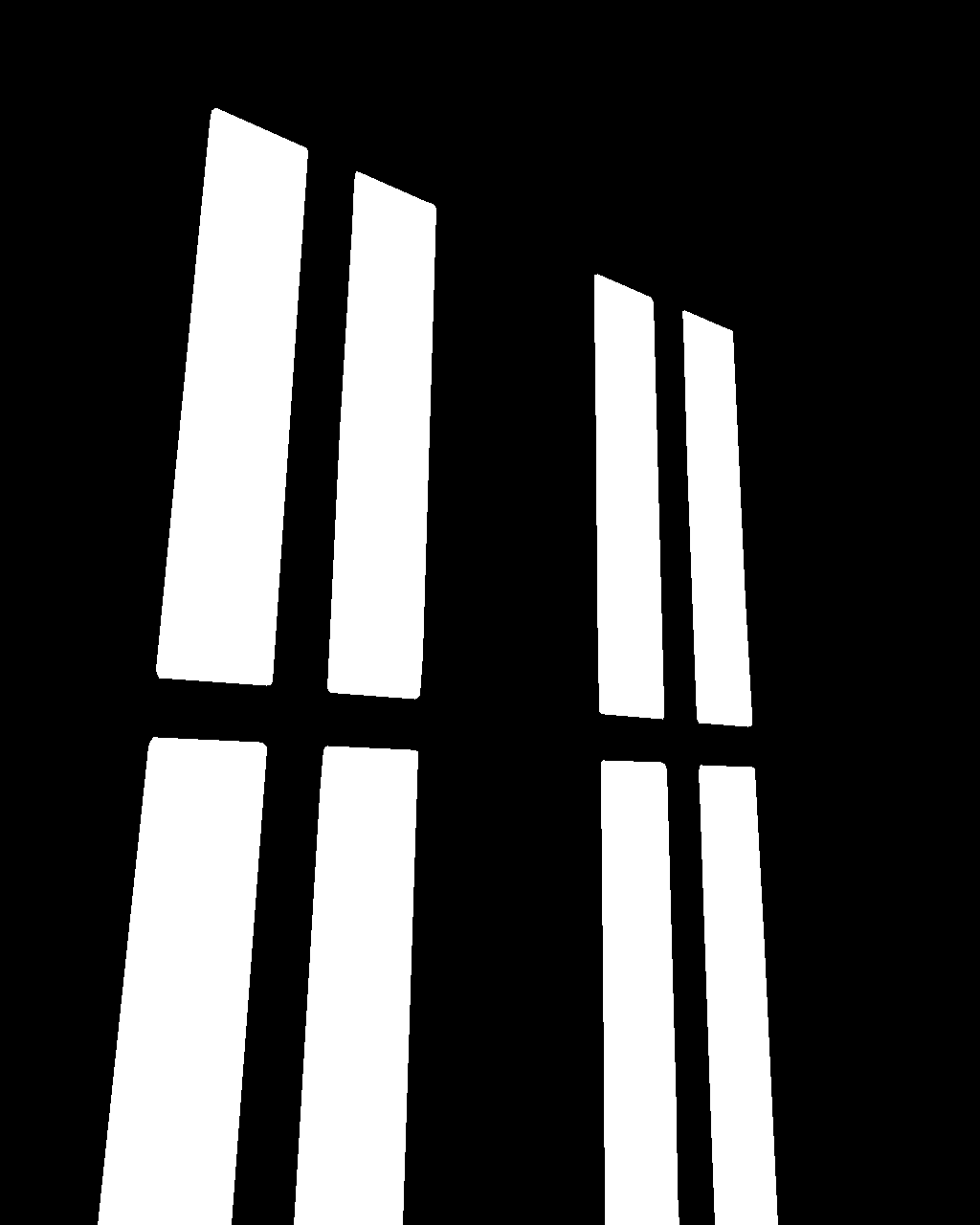} \\
		
		Image & \tabincell{c}{DANet\\\cite{Fu_2019_CVPR}} & \tabincell{c}{CCNet\\\cite{Huang_2019_ICCV}} & \tabincell{c}{PoolNet\\\cite{Liu_2019_CVPR}} & \tabincell{c}{EGNet\\\cite{zhao2019EGNet}} & \tabincell{c}{DSC\\\cite{Hu_2018_CVPR}} & \tabincell{c}{MirrorNet\\\cite{yang2019mirrornet}} & \tabincell{c}{\add{TransLab}\\\cite{xie2020segmenting}} & \tabincell{c}{\add{Trans2Seg}\\\cite{xie2021segmenting}} & \tabincell{c}{GDNet\\\cite{Mei_2020_CVPR}} & \tabincell{c}{GDNet-B\\(ours)} & GT \\
	\end{tabular}
	\caption{Visual comparison of GDNet-B to the state-of-the-art methods on the proposed GDD testing set.}
	\label{fig:visual}
\end{figure*}

\mhy{\textbf{Effectiveness of the LCFI module.}}
\mhy{In Table \ref{tab:ablation_lcfi}, by comparing the results of ``Base + LCFI w/ one scale'', ``Base + LCFI w/ two scales'' and ``Base + LCFI'' (also can be expressed as ``Base + LCFI w/ four scales''), we can see that multi-scale convolutions can help improve the detection performance.} In addition, using dilated convolution in the LCFI module (\emph{i.e.}, ``Base + LCFI w/ sparse'') performs better than using local one (\emph{i.e.}, ``Base + LCFI w/ local''), as contexts can be explored from a larger receptive field. With approximately the same number of parameters, using spatially separable convolutions (\emph{i.e.}, ``Base + LCFI w/ one path'') can harvest more contexts from a large field to further boost the performance. Finally, large-field contextual features with different characteristics can be obtained by two parallel spatially separable convolution paths and help ``Base + LCFI'' achieve more accurate glass detection results. Figure \ref{fig:visual_lcfi} shows a visual example. We can see that ``Base + LCFI'' successfully addresses the under-segmentation problem with the help of abundant contextual features extracted from the large field.

\mhy{\textbf{Effectiveness of the BFE module.}}
\mhy{By comparing the results of ``GDNet-B w/o BFE'' and ``GDNet-B'' in Table \ref{tab:ablation_bfe}, we can see that BFE can help improve the glass detection performance. Figure \ref{fig:visual_bfe} shows a visual comparison example where we can see that the BFE module can help the network detect the complete glass region (the top row) and generate glass detection result with more accurate glass boundary (the bottom row). This clearly demonstrates the effectiveness of our BFE module.}

\begin{table}[tbp]
	\centering
	\caption{\mhy{Ablation studies on the effectiveness of the BFE module.}}
	\begin{tabular}{p{3.0cm}|p{1cm}<{\centering}p{0.9cm}<{\centering}p{0.8cm}<{\centering}}
		\hline
		\hline
		Networks & IoU$\uparrow$ & $F_\beta^m$$\uparrow$ & BER$\downarrow$ \\
		\hline
		\hline
		GDNet-B w/o BFE  & 87.63 & 0.937 & 5.62 \\
		\hline
		GDNet-B      & \bf{87.83} & \bf{0.939} & \bf{5.52} \\
		\hline
		\hline
	\end{tabular}
	\label{tab:ablation_bfe}
\end{table}

\def\wbfe{0.19\linewidth}
\def\hbfe{.5in}
\begin{figure}[tbp]
	\setlength{\tabcolsep}{1.2pt}
	\centering
	\begin{tabular}{ccccc}
		\includegraphics[width=\wbfe, height=\hbfe]{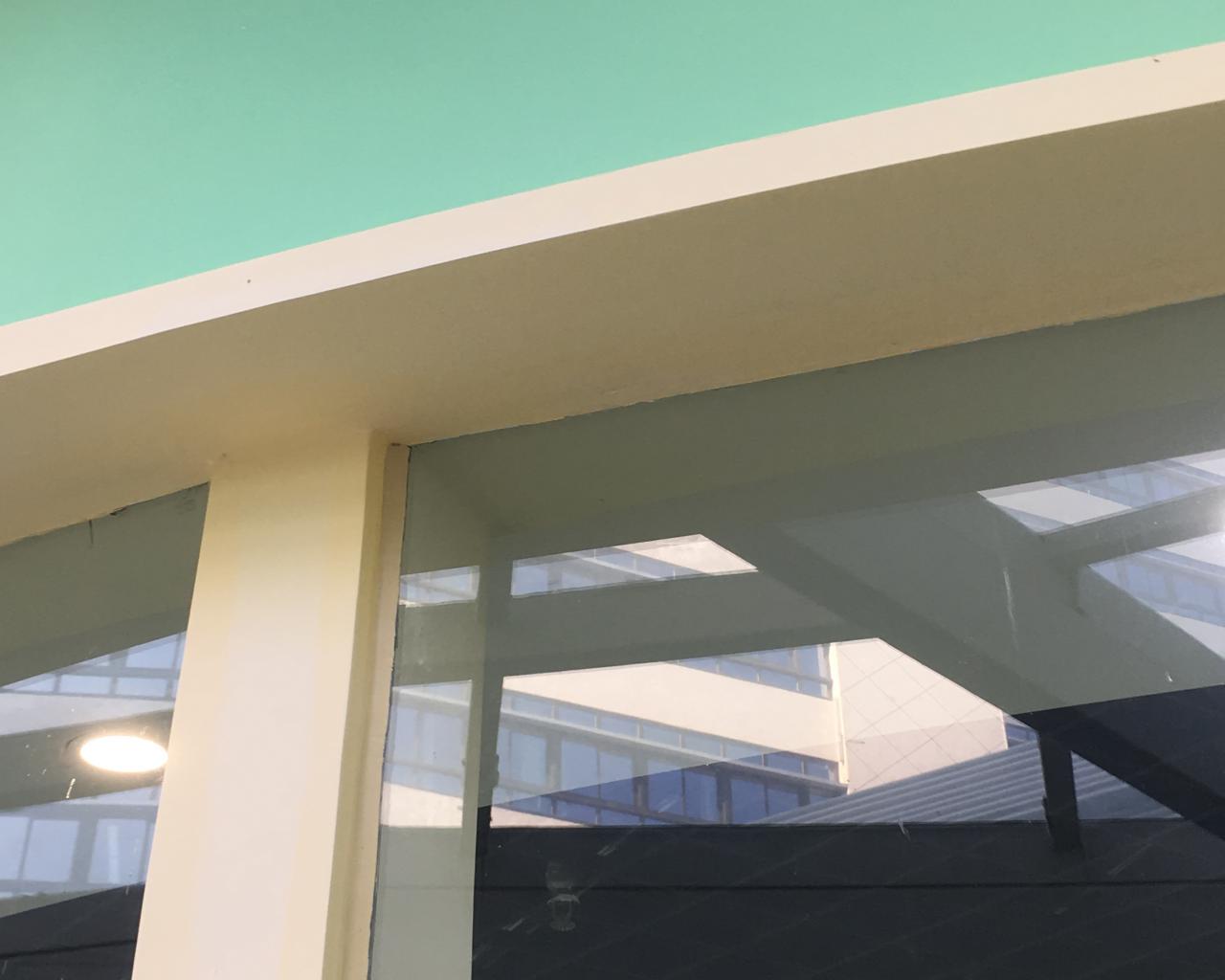}&
		\includegraphics[width=\wbfe, height=\hbfe]{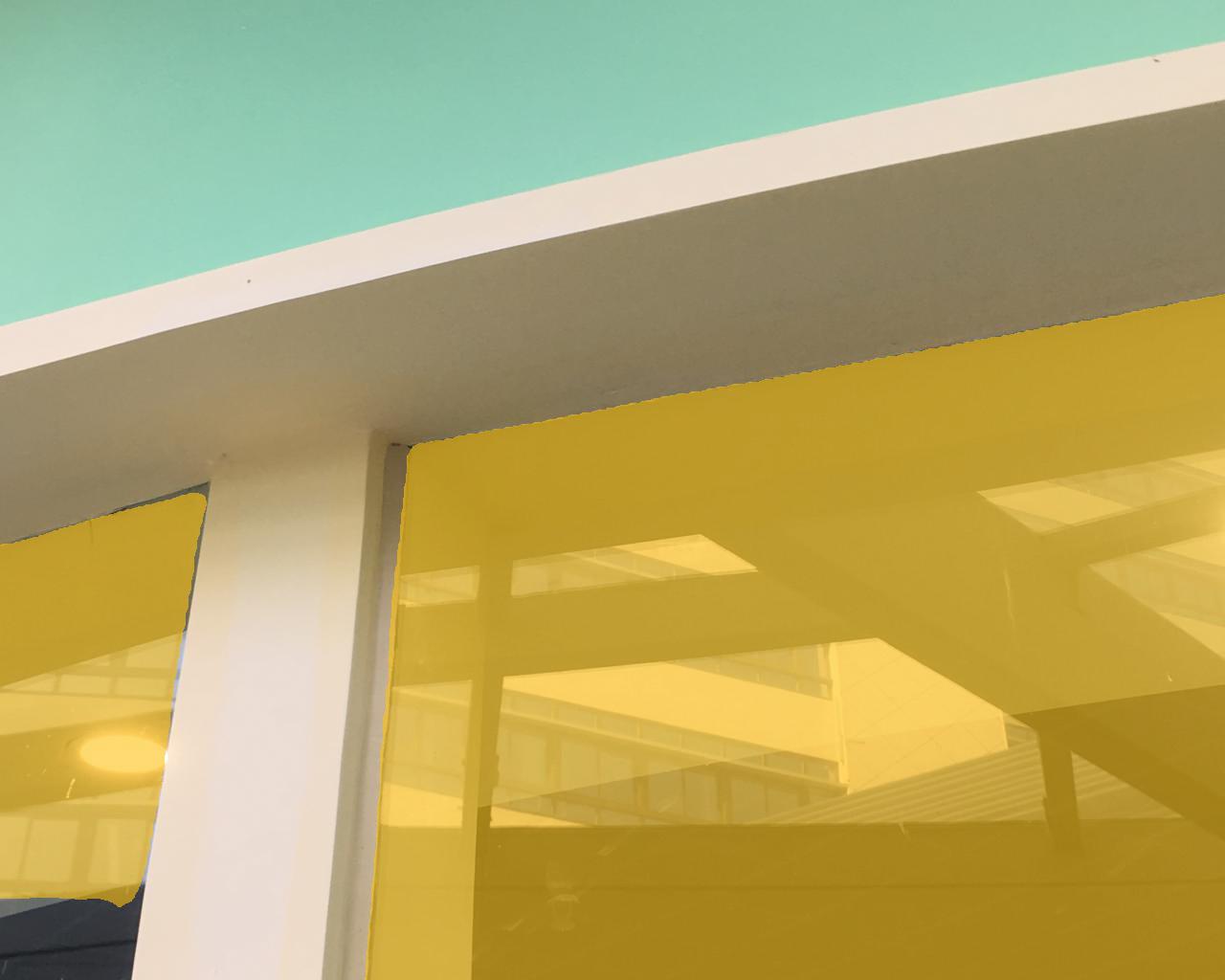}&
		\includegraphics[width=\wbfe, height=\hbfe]{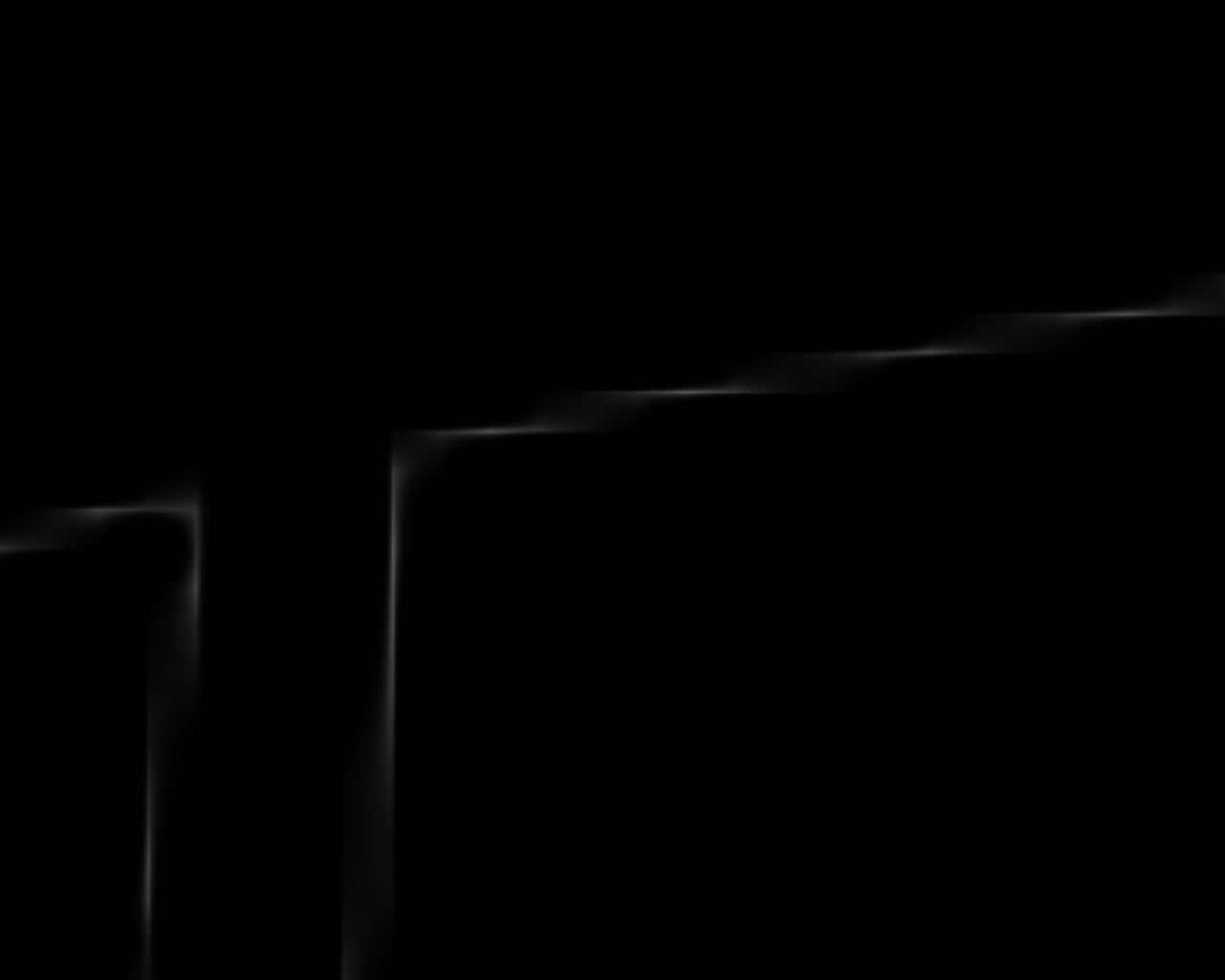}&
		\includegraphics[width=\wbfe, height=\hbfe]{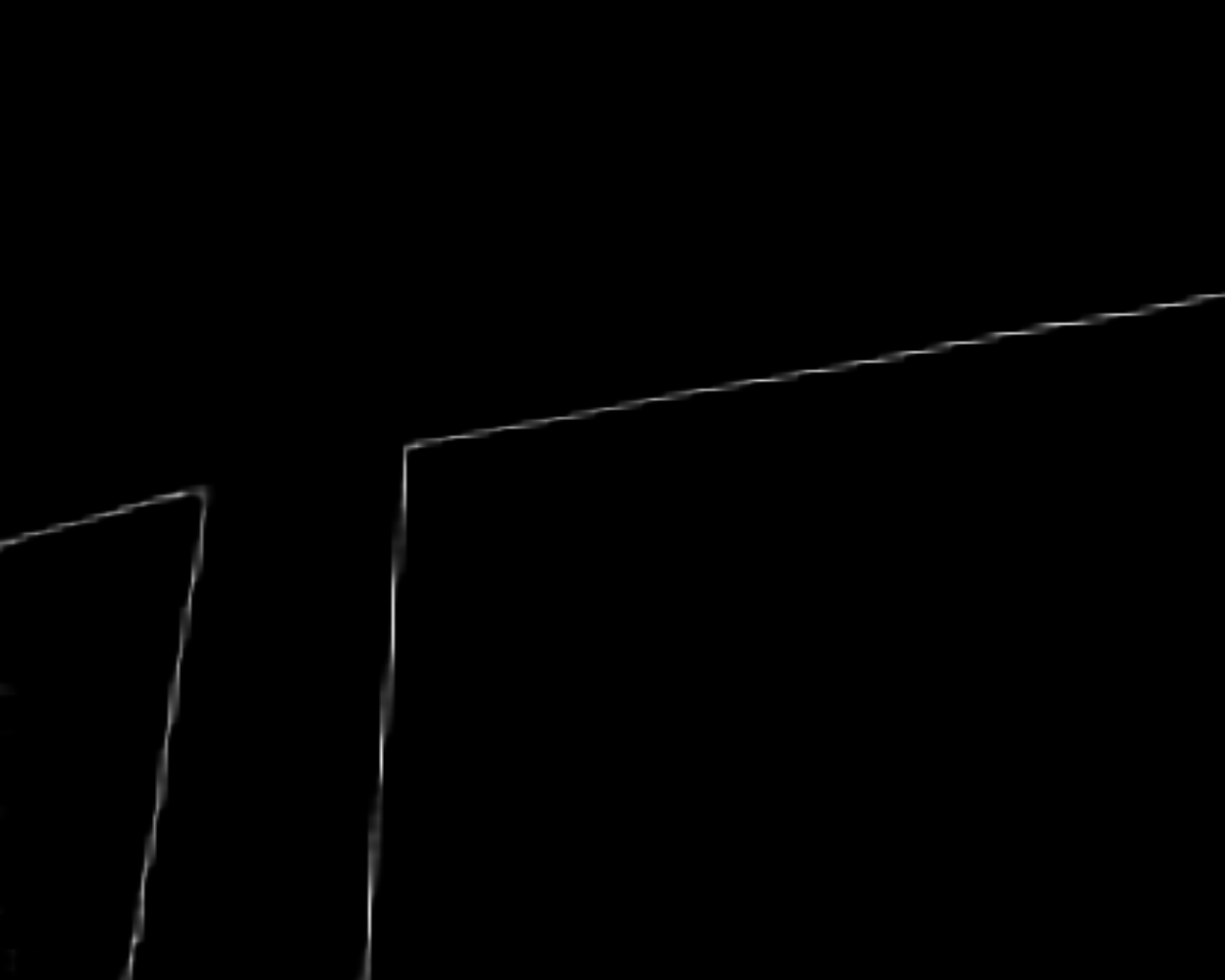}&
		\includegraphics[width=\wbfe, height=\hbfe]{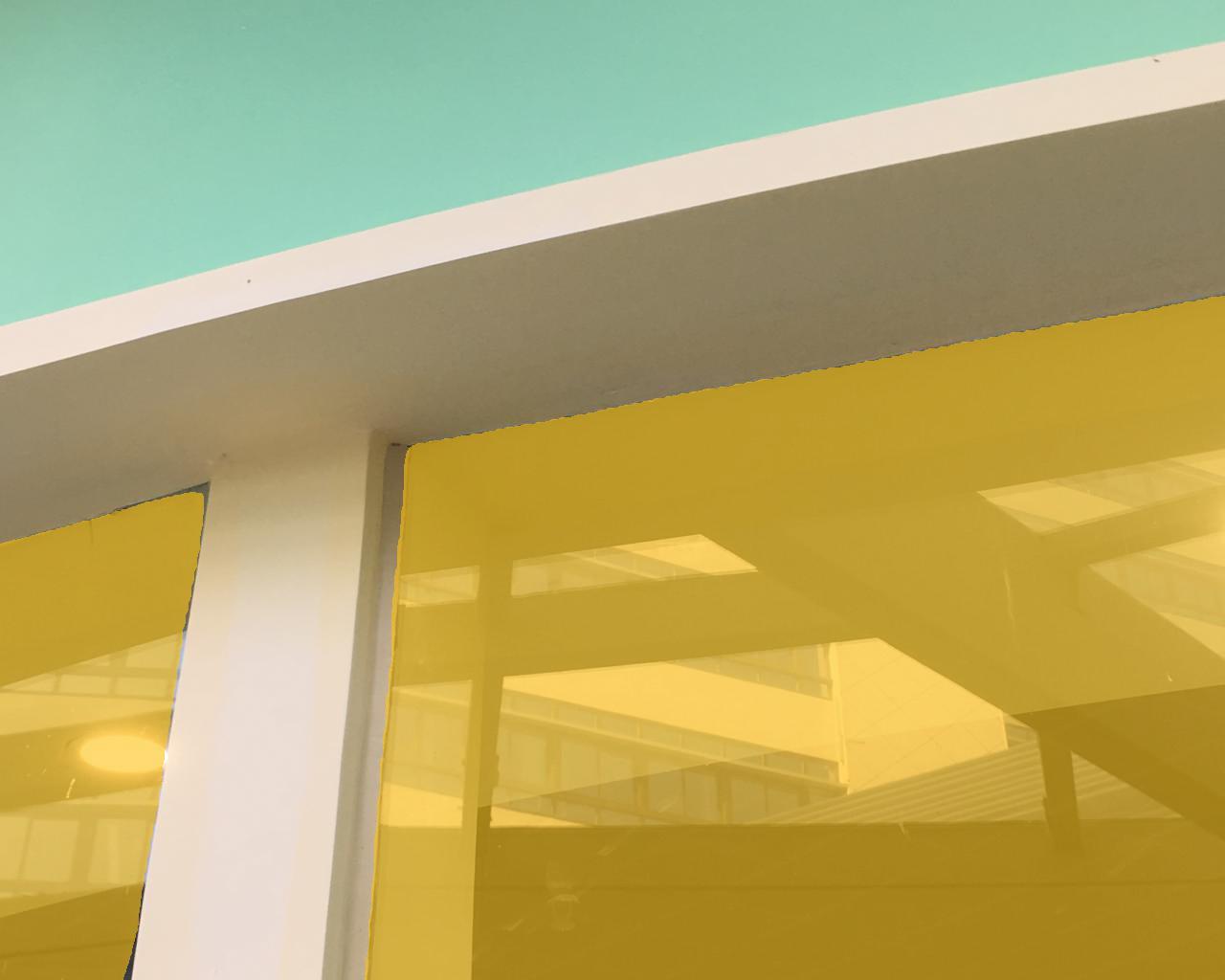} \\
		
		\includegraphics[width=\wbfe, height=\hbfe]{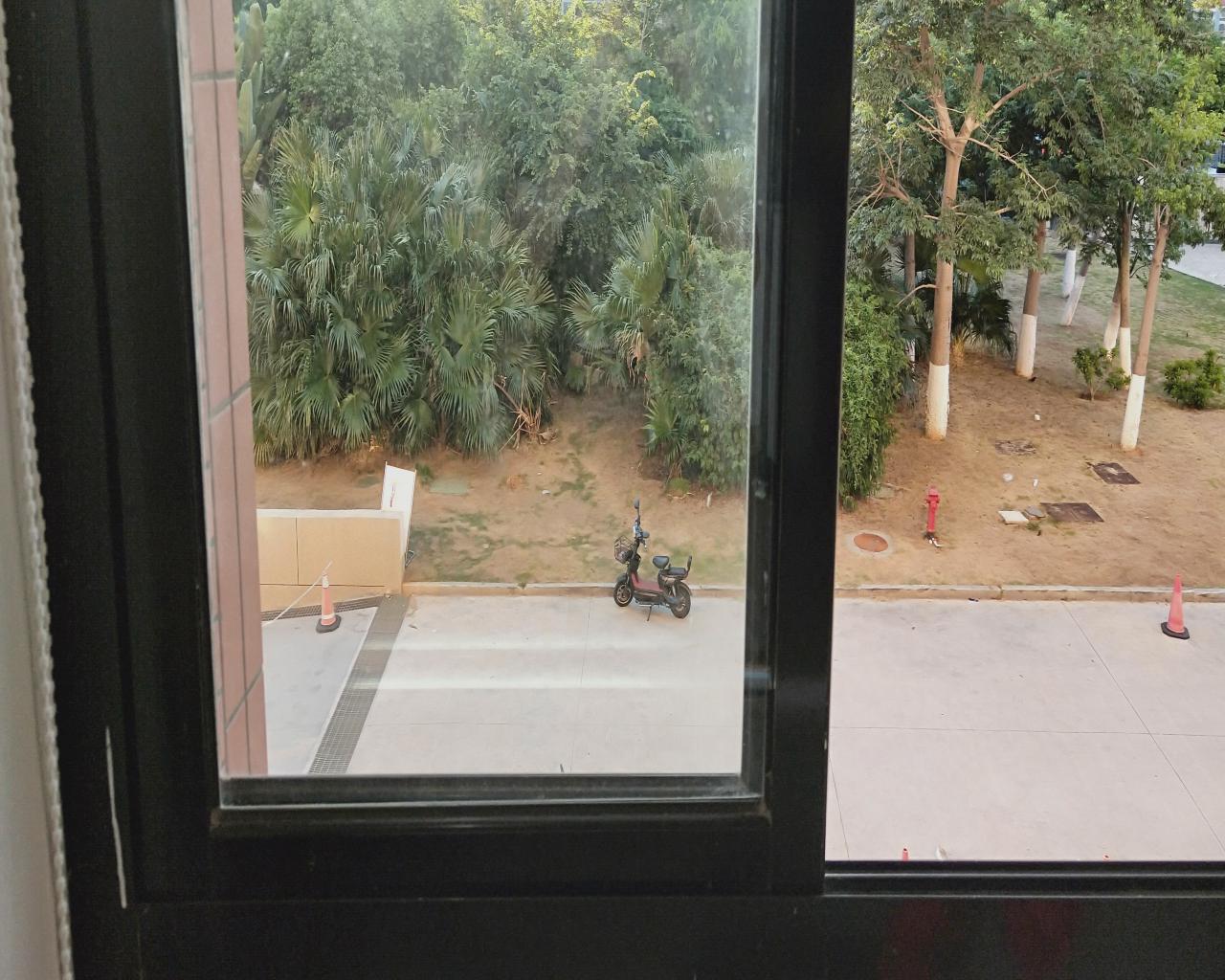}&
		\includegraphics[width=\wbfe, height=\hbfe]{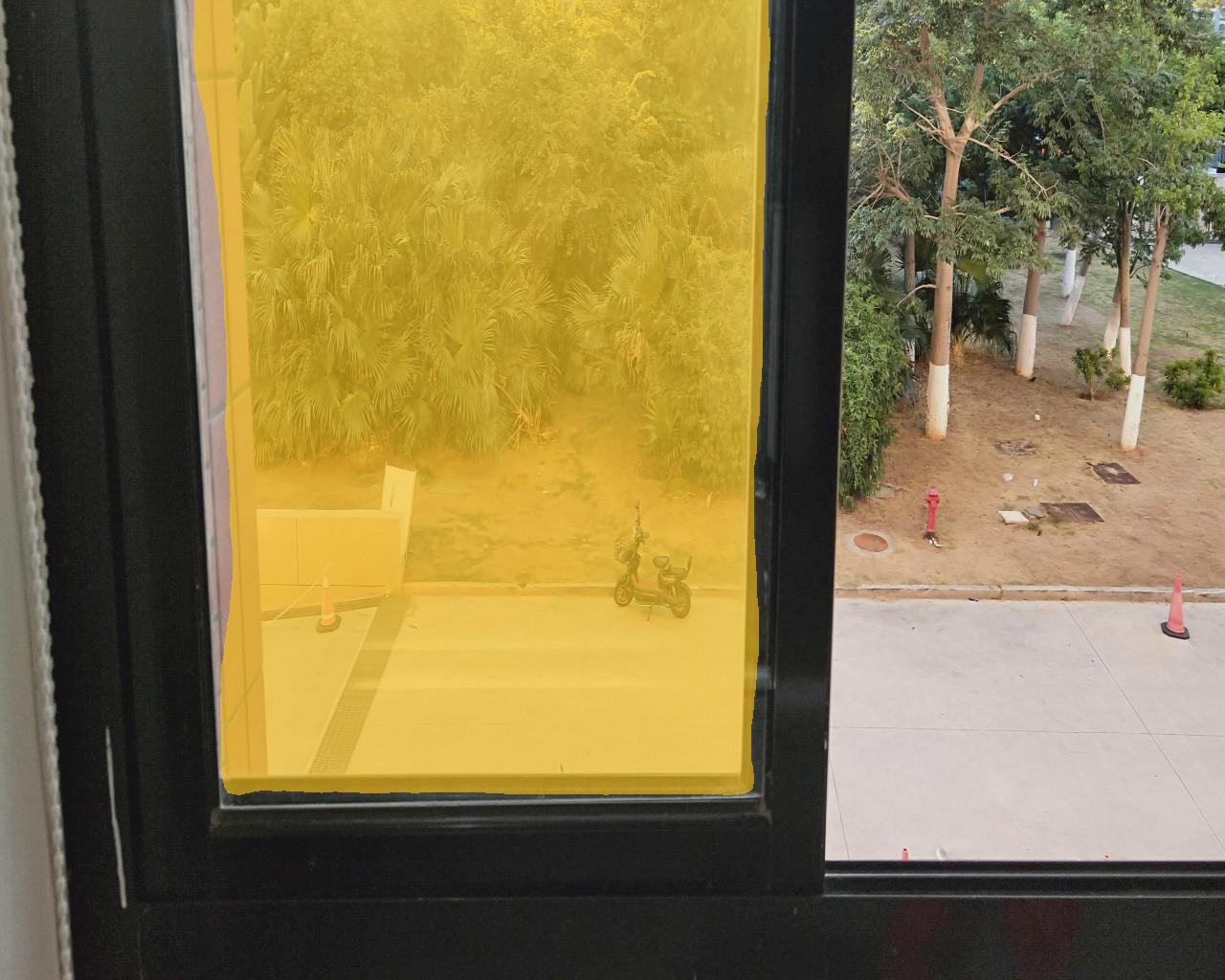}&
		\includegraphics[width=\wbfe, height=\hbfe]{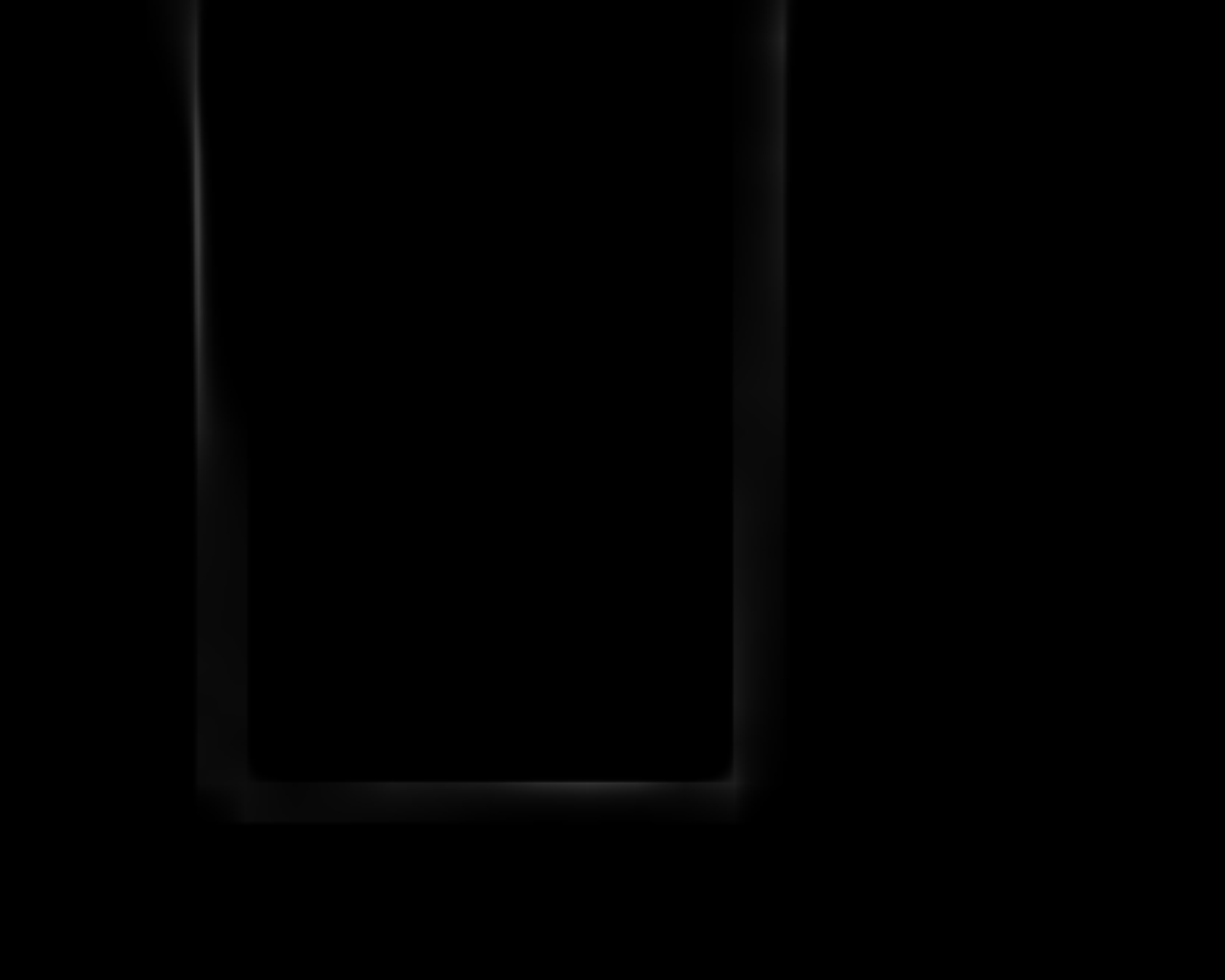}&
		\includegraphics[width=\wbfe, height=\hbfe]{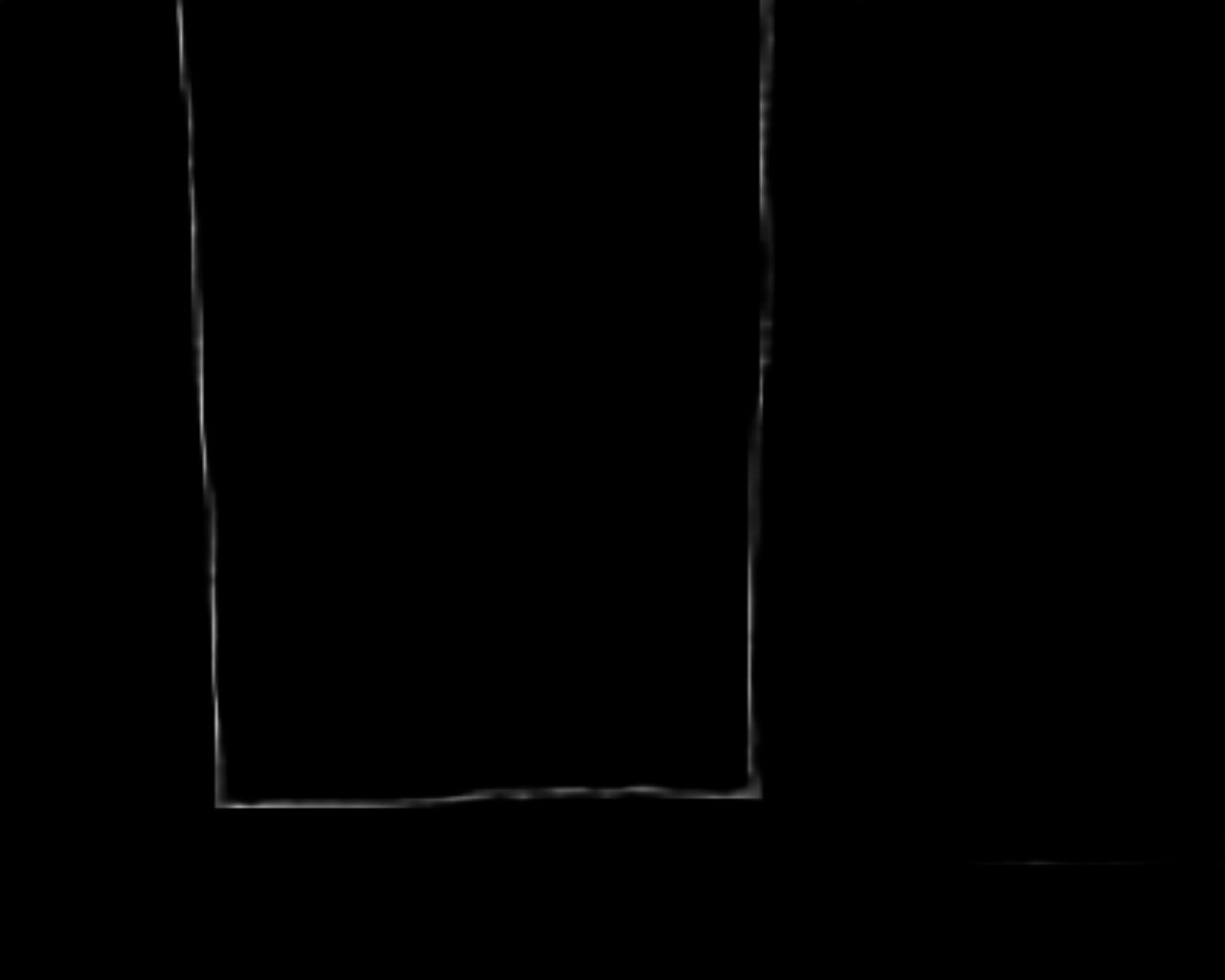}&
		\includegraphics[width=\wbfe, height=\hbfe]{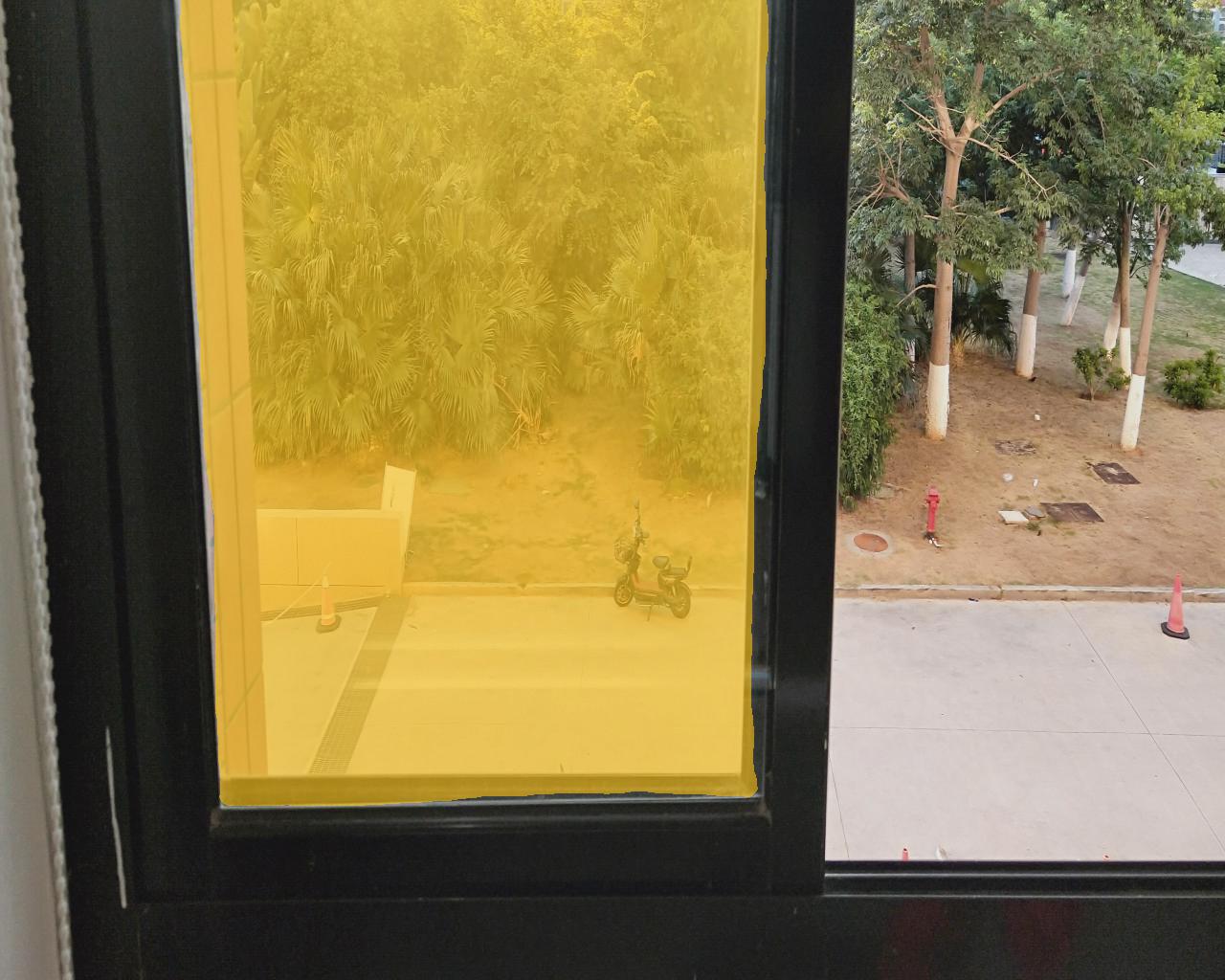} \\
		
		\footnotesize{Image} & \footnotesize{\tabincell{c}{GDNet-B\\w/o BFE}} & \footnotesize{HL map} & \footnotesize{LL map} & \footnotesize{GDNet-B} \\
		
	\end{tabular}
	\caption{\mhy{Ablation analysis on the effectiveness of the BFE module. ``HL map'' and ``LL map'' are the output boundary maps of the BFE module applied on the high-level features and attentive low-level features in GDNet-B, respectively.}}
\label{fig:visual_bfe}
\end{figure}

\def\wmore{0.081\linewidth}
\def\hmore{.45in}
\begin{figure*}[htb]
	\setlength{\tabcolsep}{0.6pt}
	\centering
	\begin{tabular}{cccccccccccc}
		\includegraphics[width=\wmore, height=\hmore]{figure/more/ade_124.jpg}&
		\includegraphics[width=\wmore, height=\hmore]{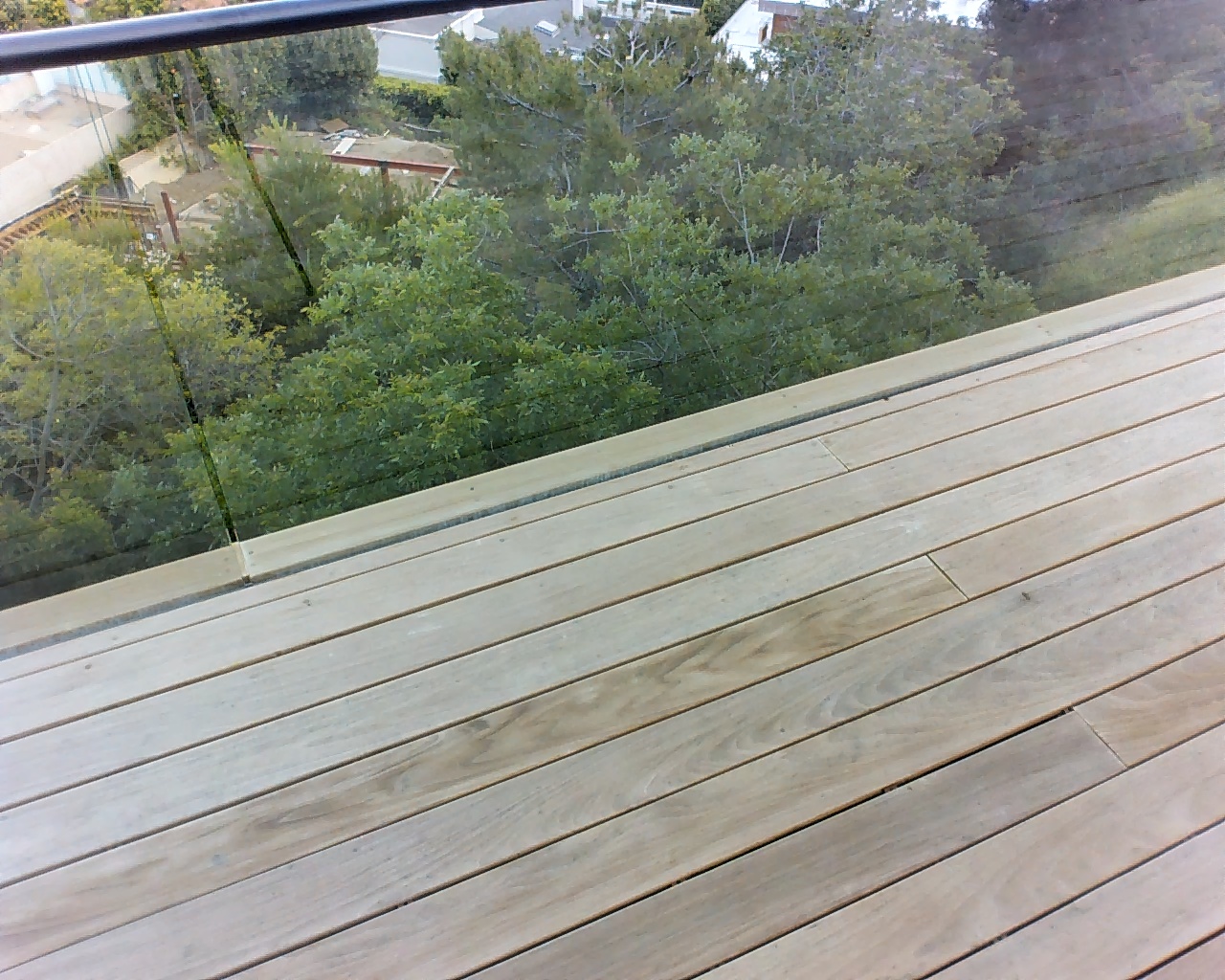}&
		\includegraphics[width=\wmore, height=\hmore]{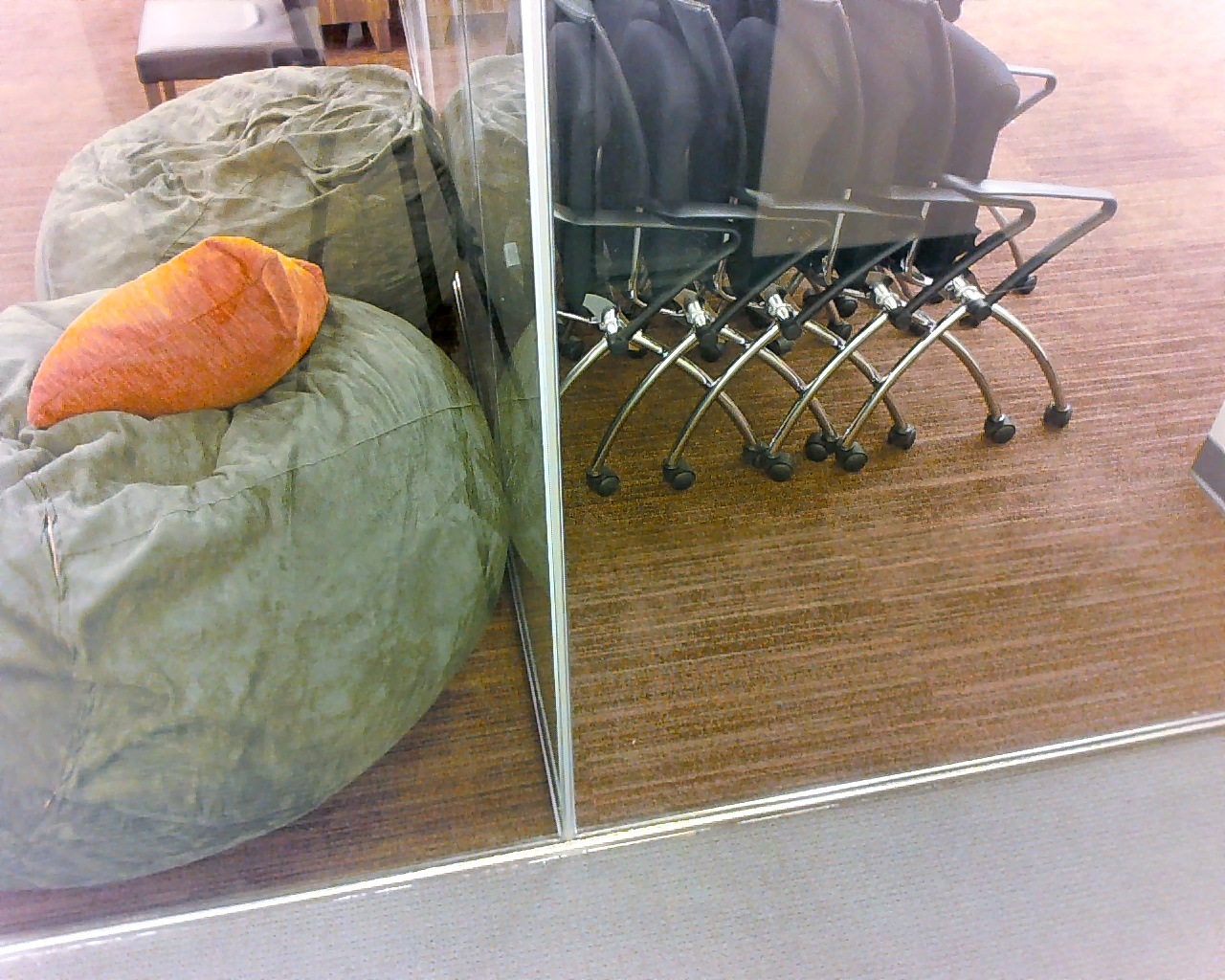}&
		\includegraphics[width=\wmore, height=\hmore]{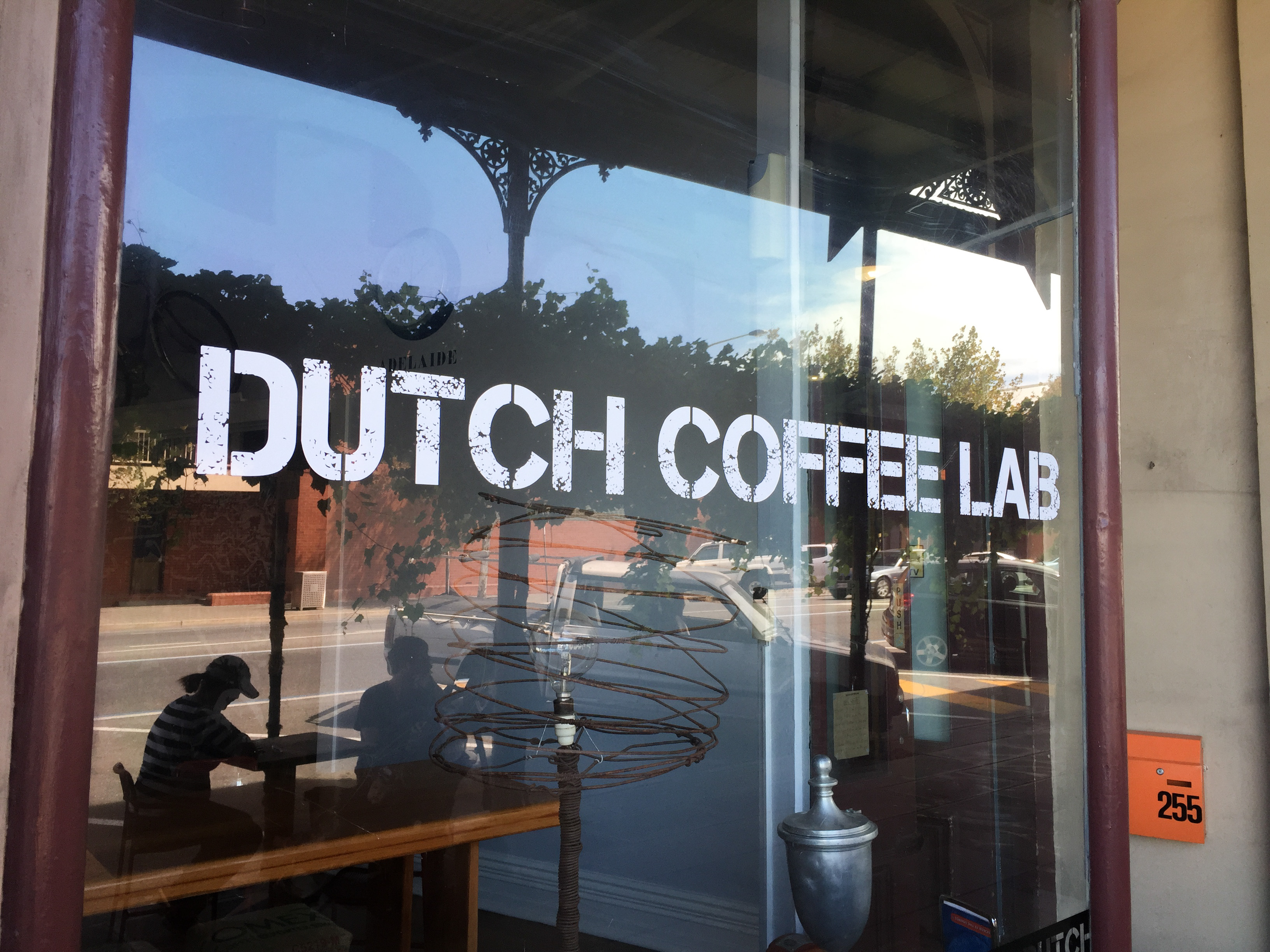}&
		\includegraphics[width=\wmore, height=\hmore]{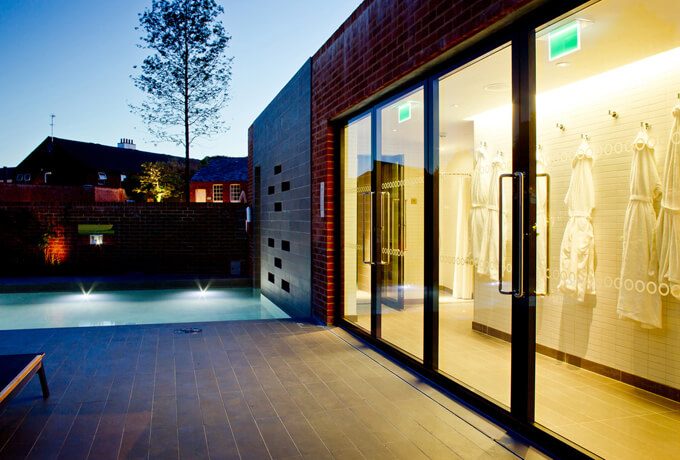}&
		\includegraphics[width=\wmore, height=\hmore]{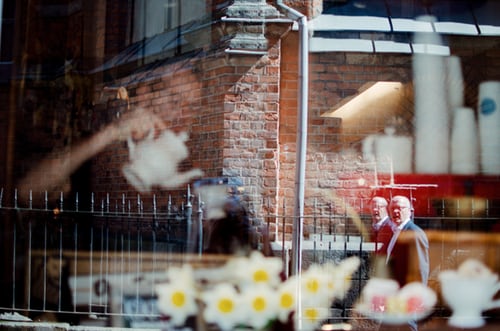}&
		\includegraphics[width=\wmore, height=\hmore]{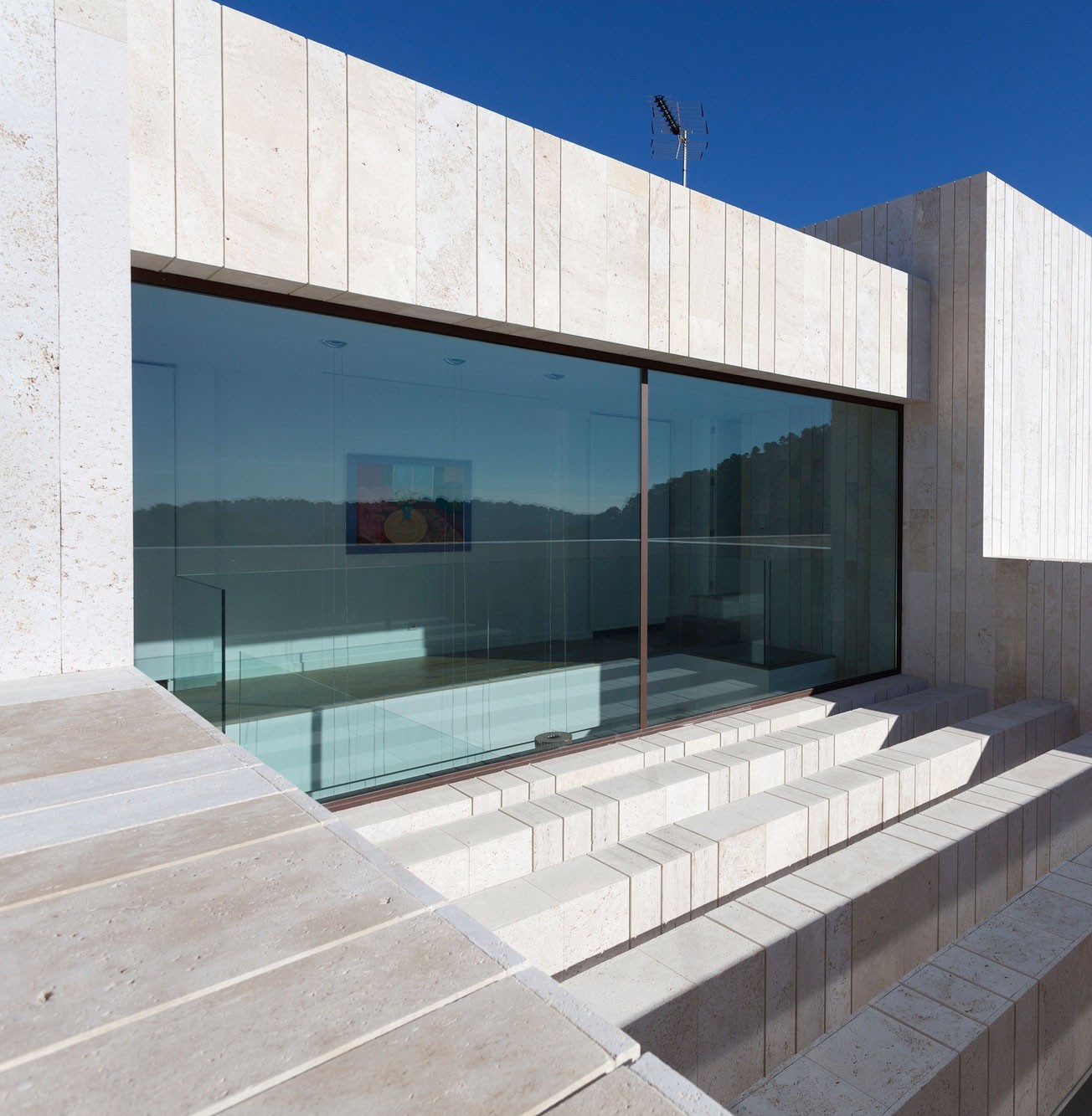}&
		\includegraphics[width=\wmore, height=\hmore]{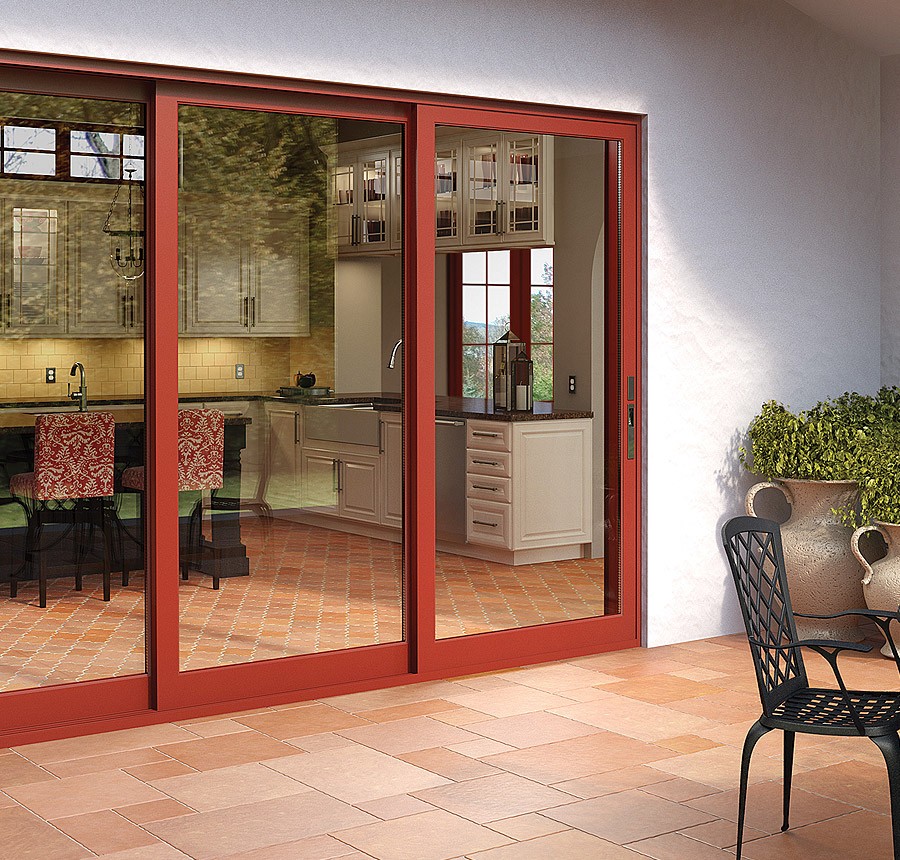}&
		\includegraphics[width=\wmore, height=\hmore]{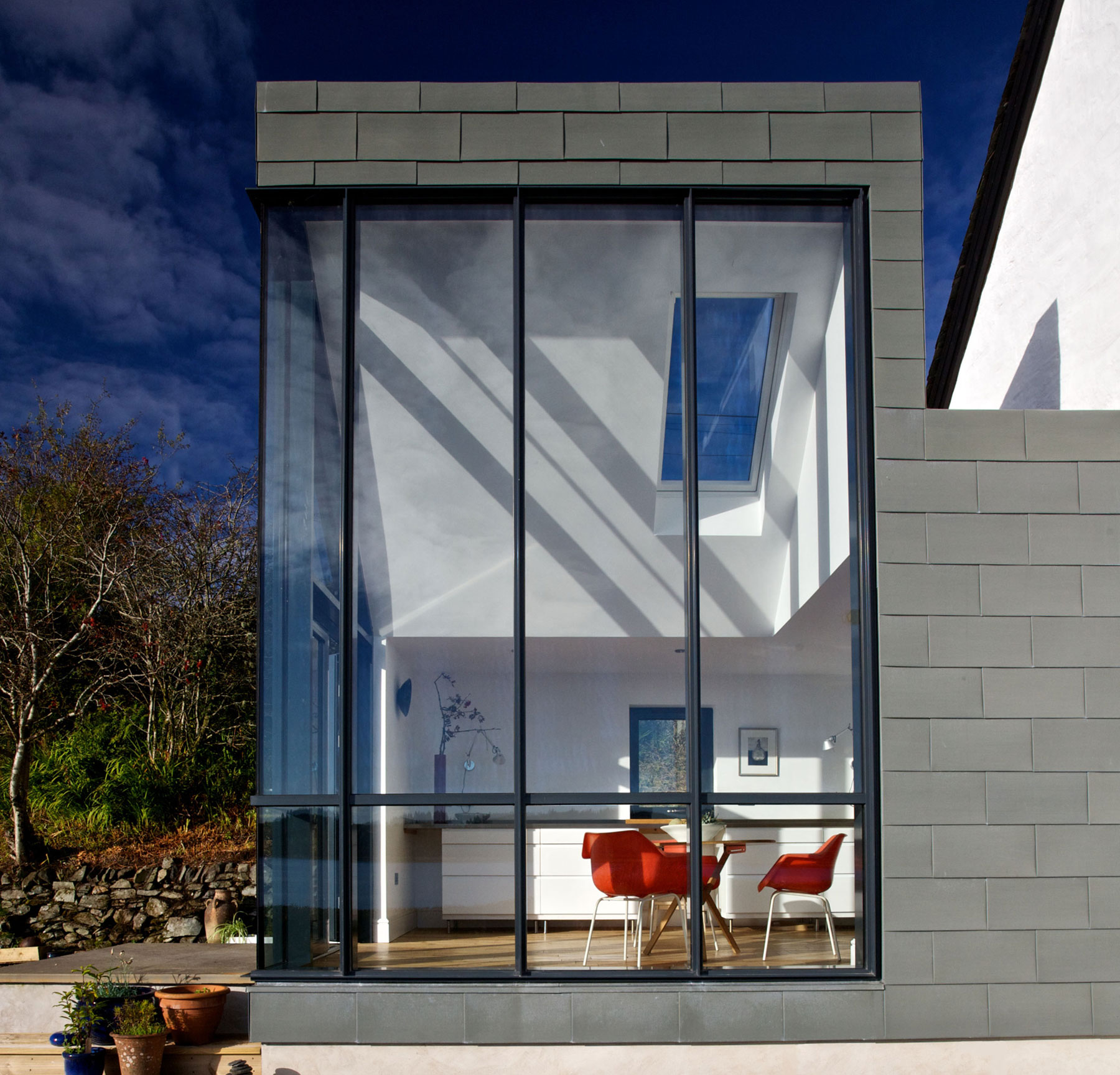}&
		\includegraphics[width=\wmore, height=\hmore]{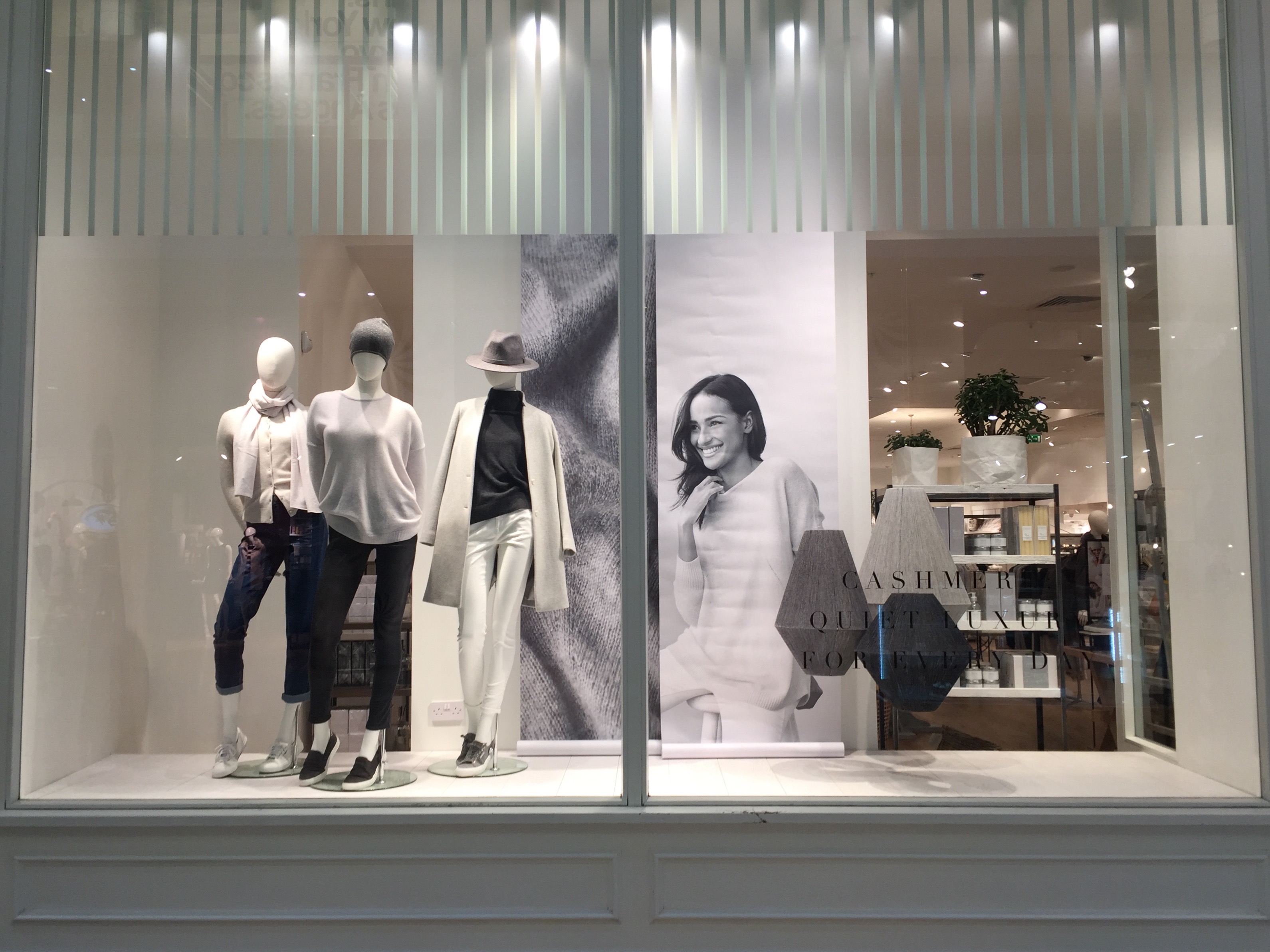}&
		\includegraphics[width=\wmore, height=\hmore]{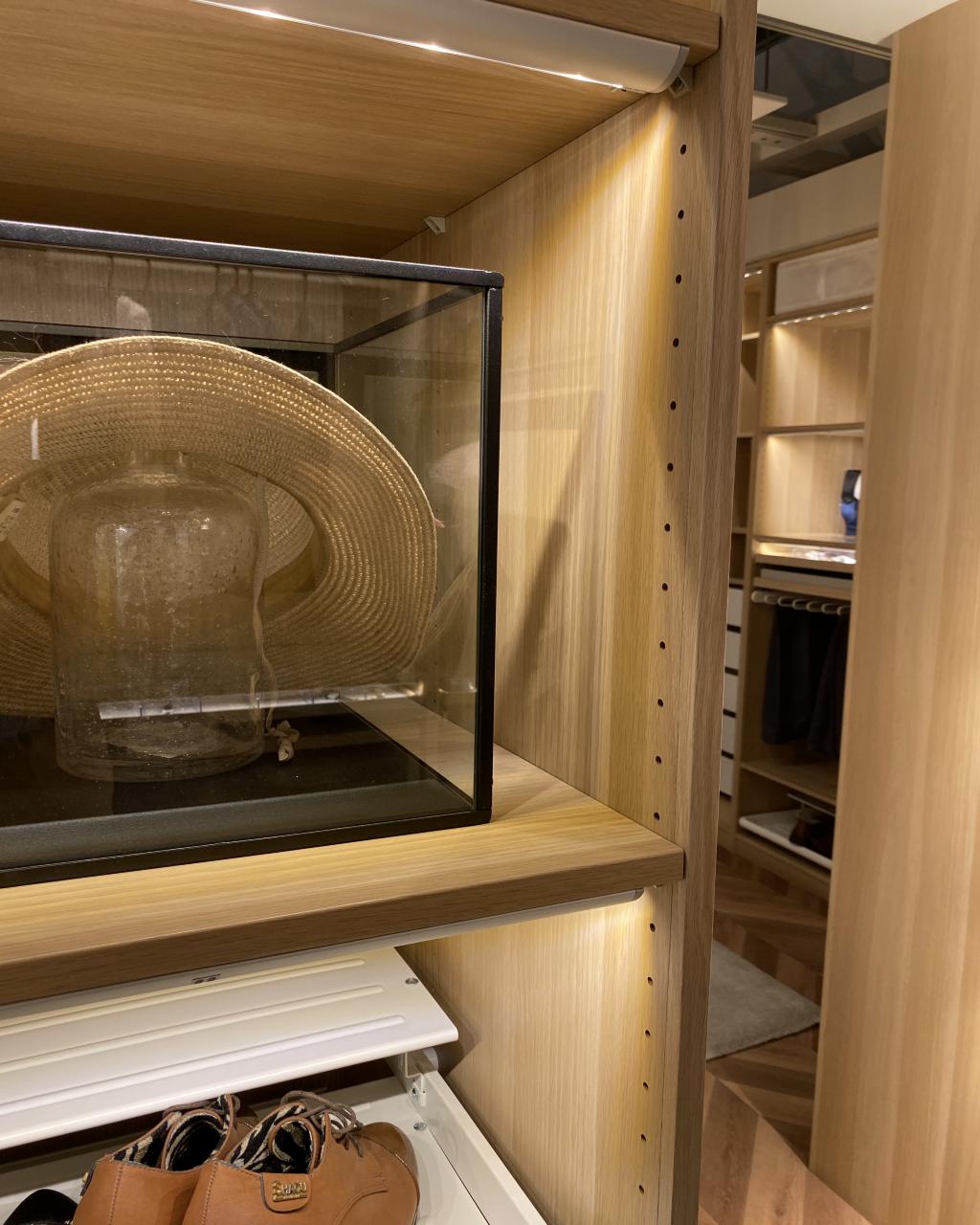}&
		\includegraphics[width=\wmore, height=\hmore]{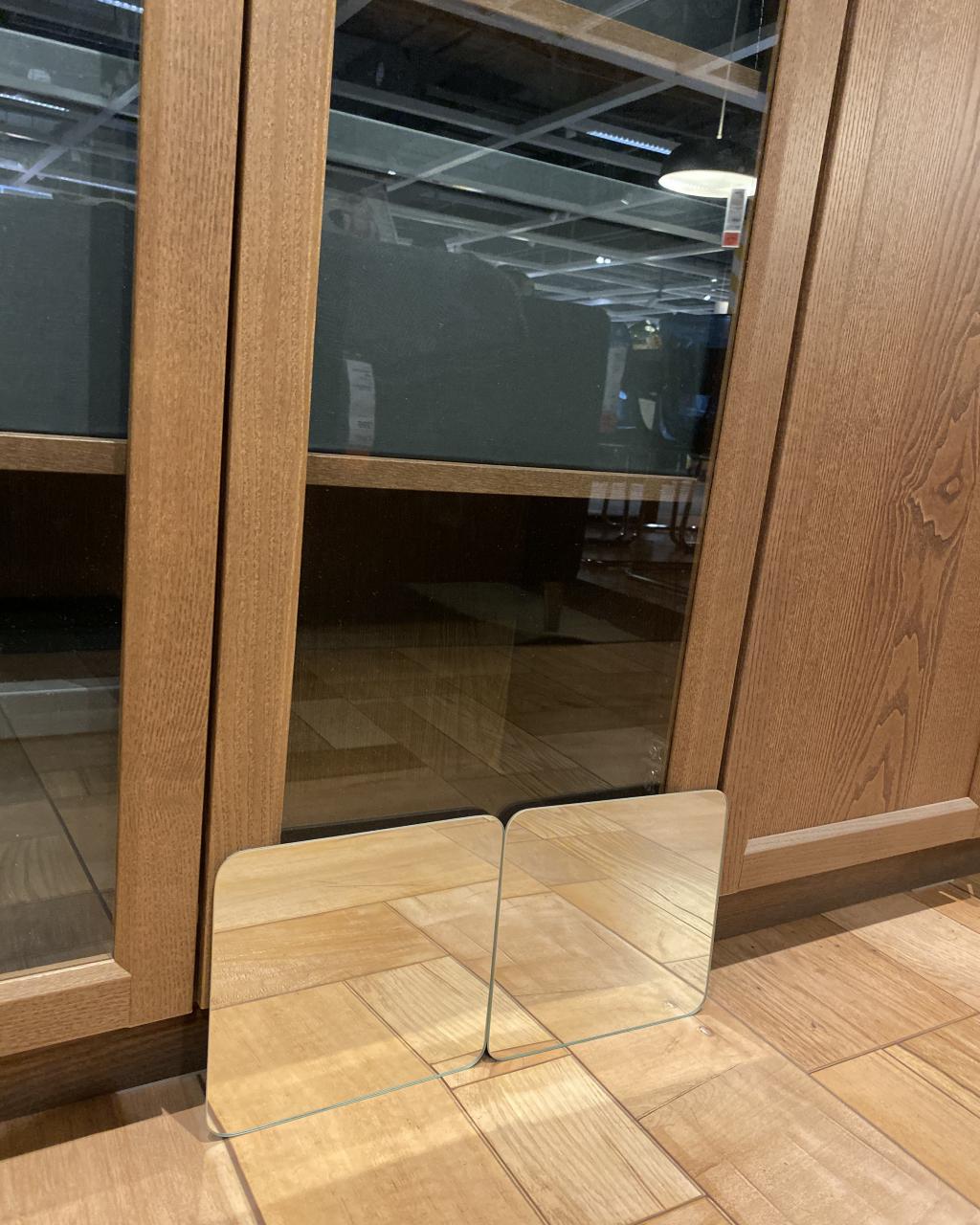} \\
		
		\includegraphics[width=\wmore, height=\hmore]{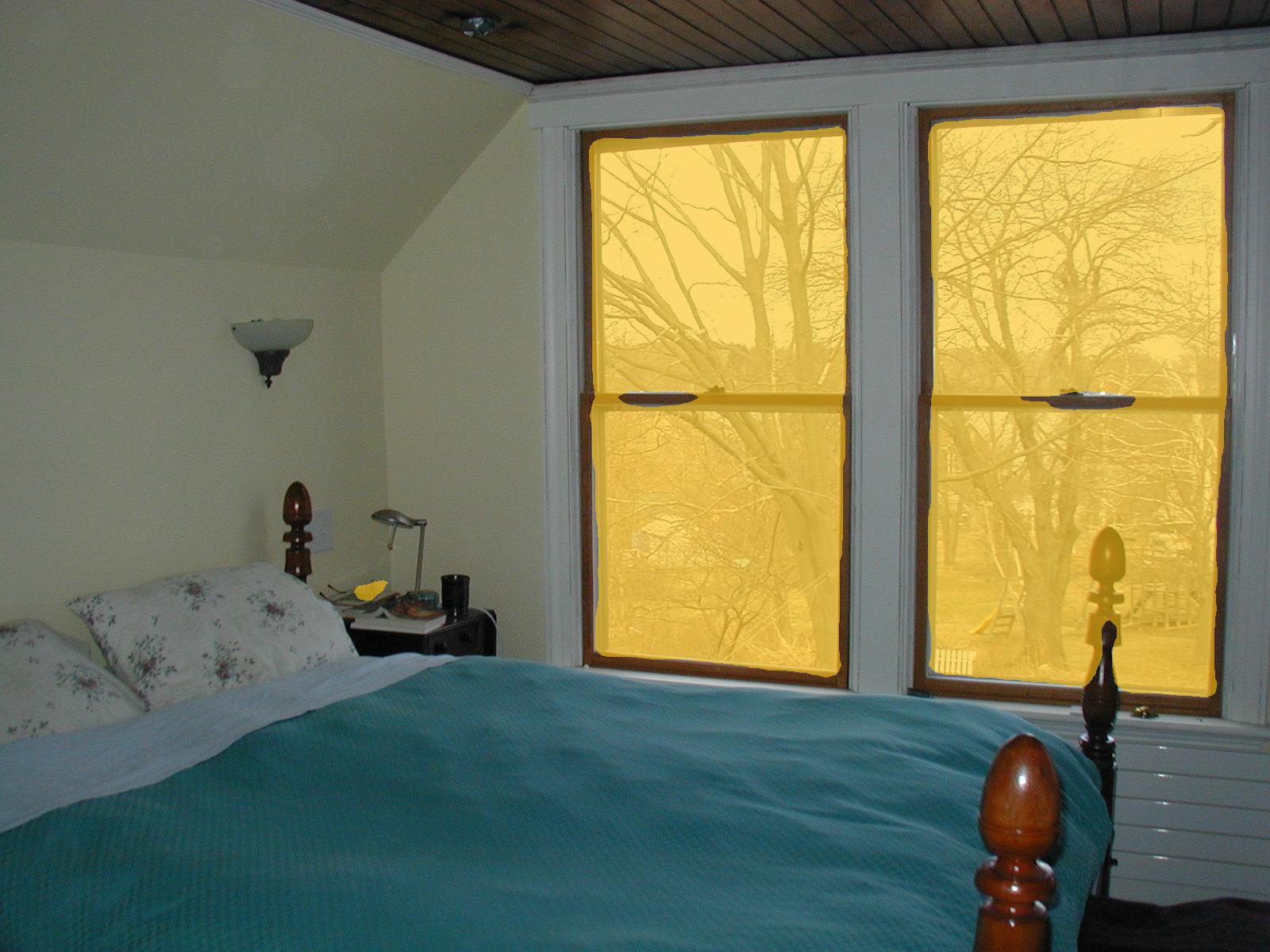}&
		\includegraphics[width=\wmore, height=\hmore]{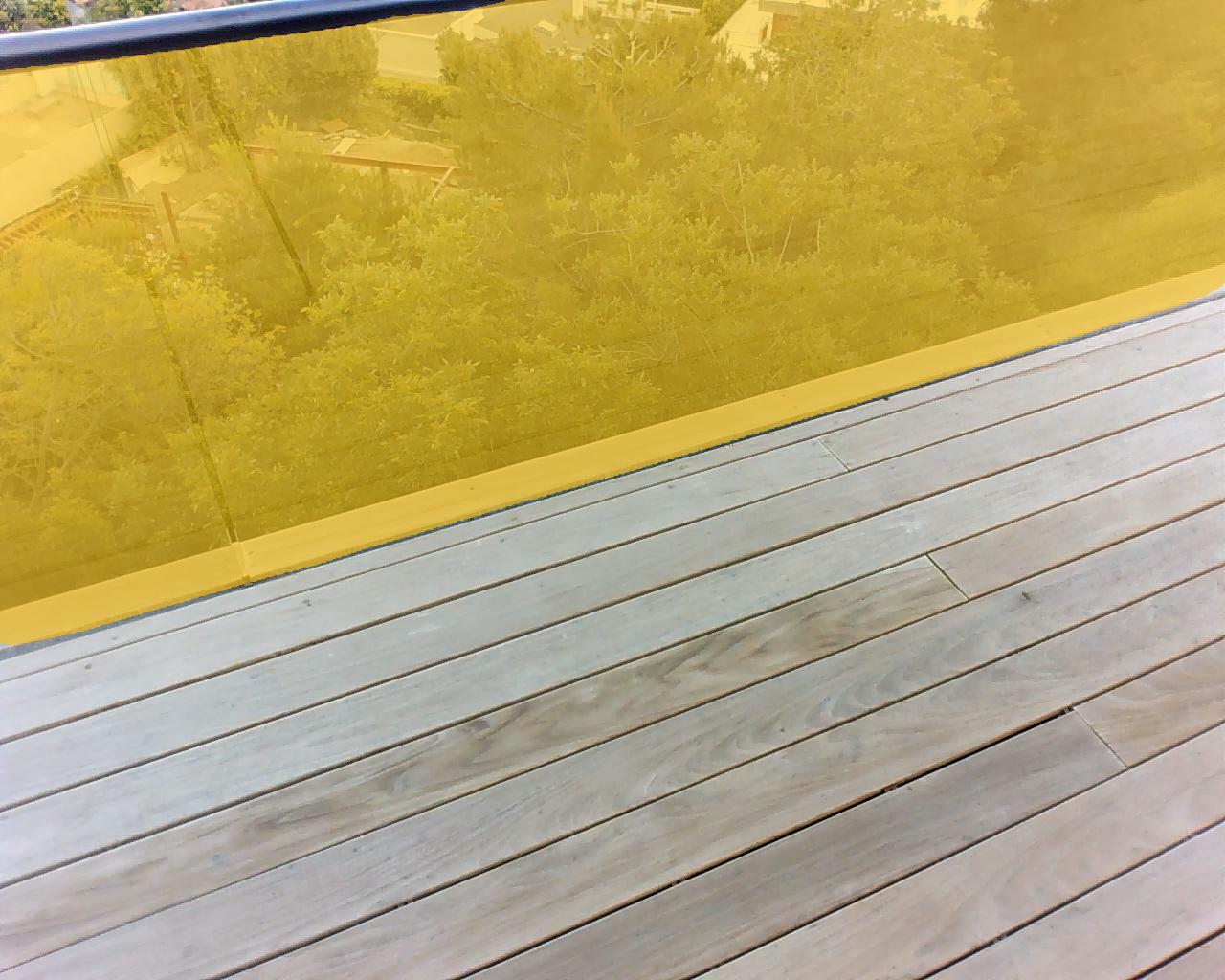}&
		\includegraphics[width=\wmore, height=\hmore]{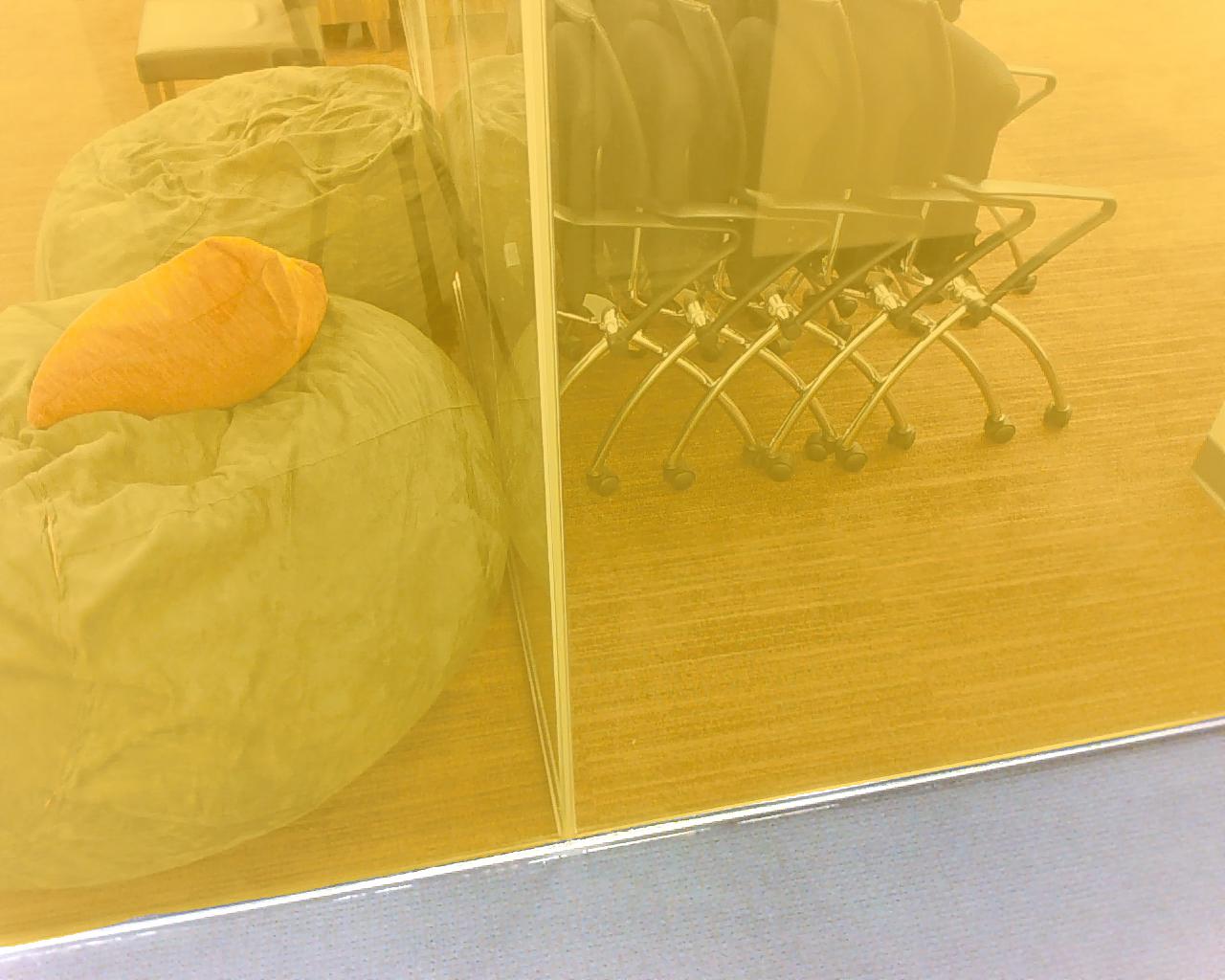}&
		\includegraphics[width=\wmore, height=\hmore]{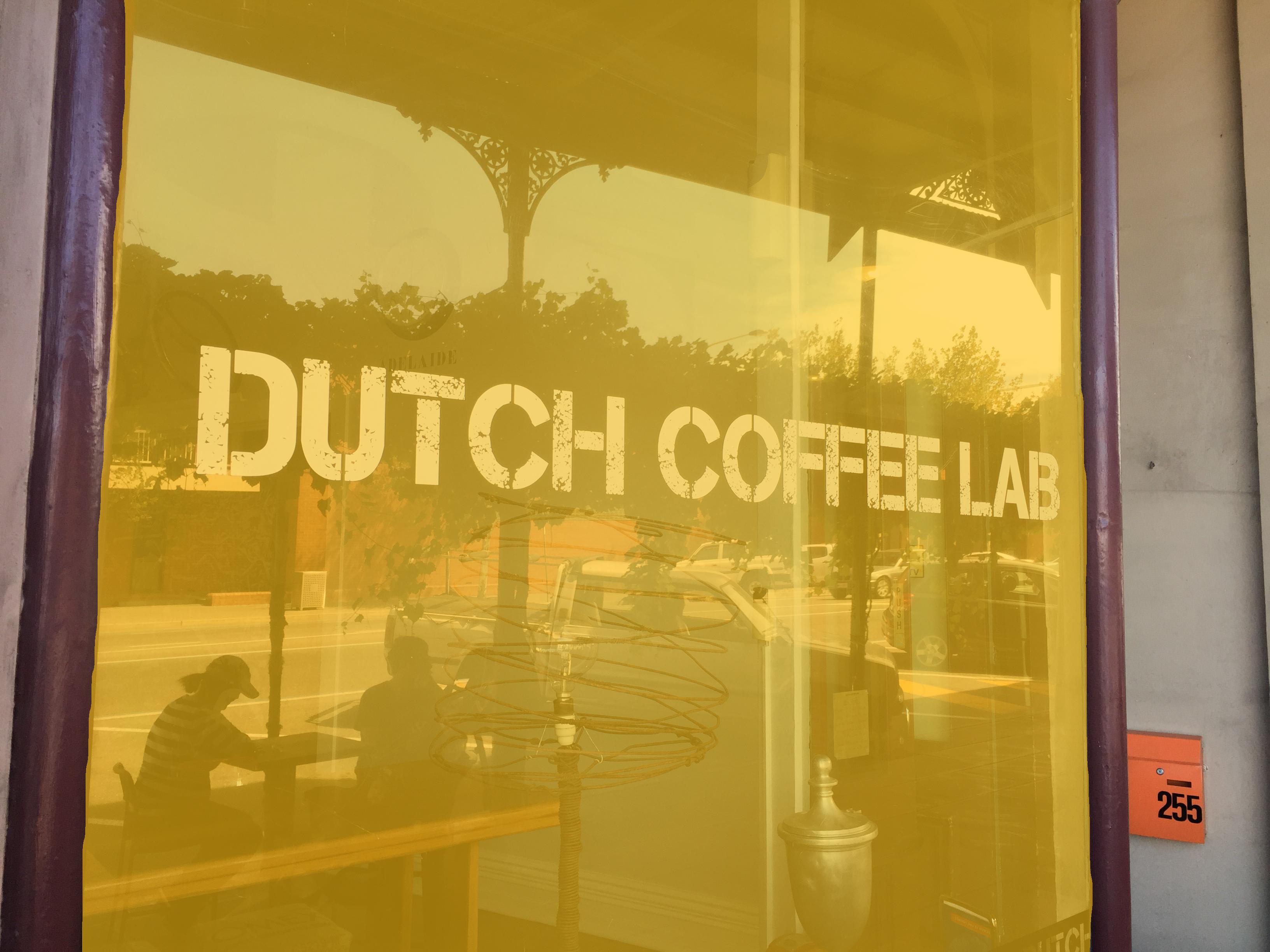}&
		\includegraphics[width=\wmore, height=\hmore]{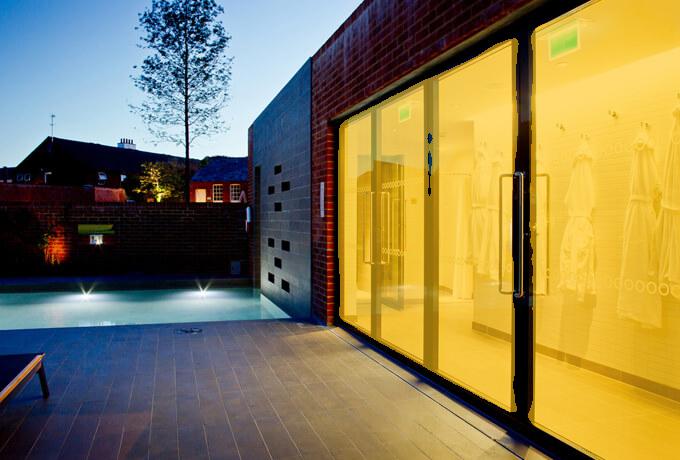}&
		\includegraphics[width=\wmore, height=\hmore]{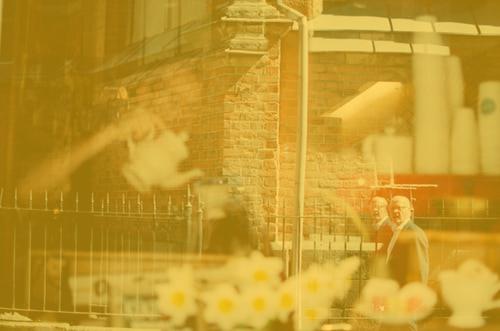}&
		\includegraphics[width=\wmore, height=\hmore]{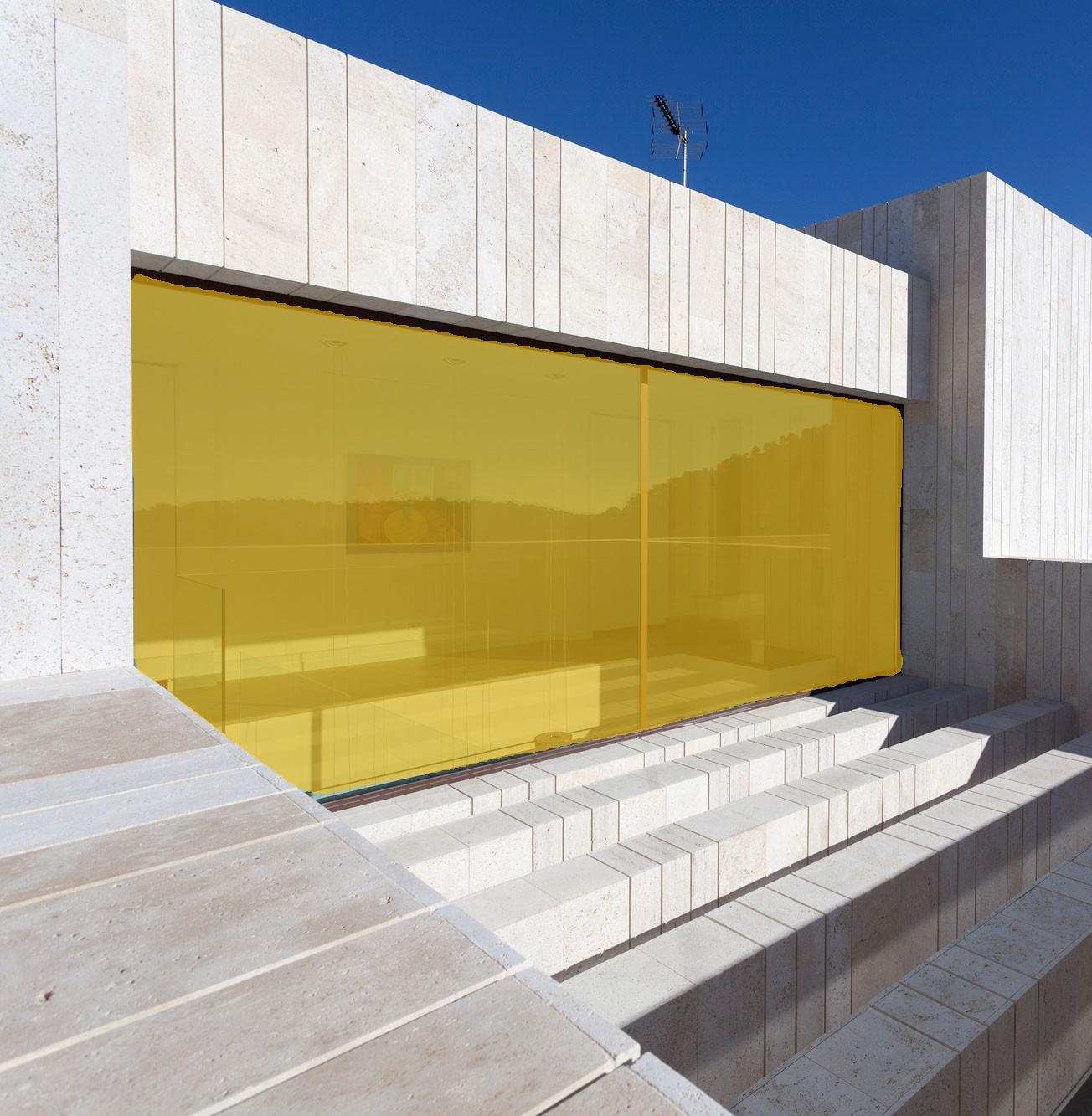}&
		\includegraphics[width=\wmore, height=\hmore]{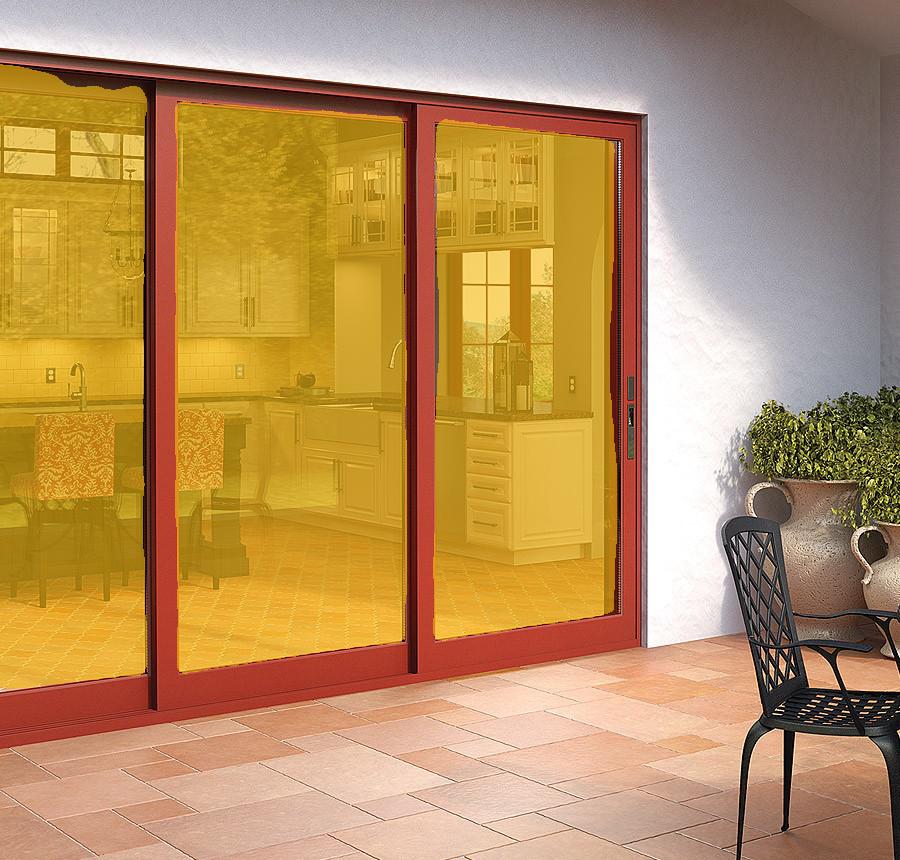}&
		\includegraphics[width=\wmore, height=\hmore]{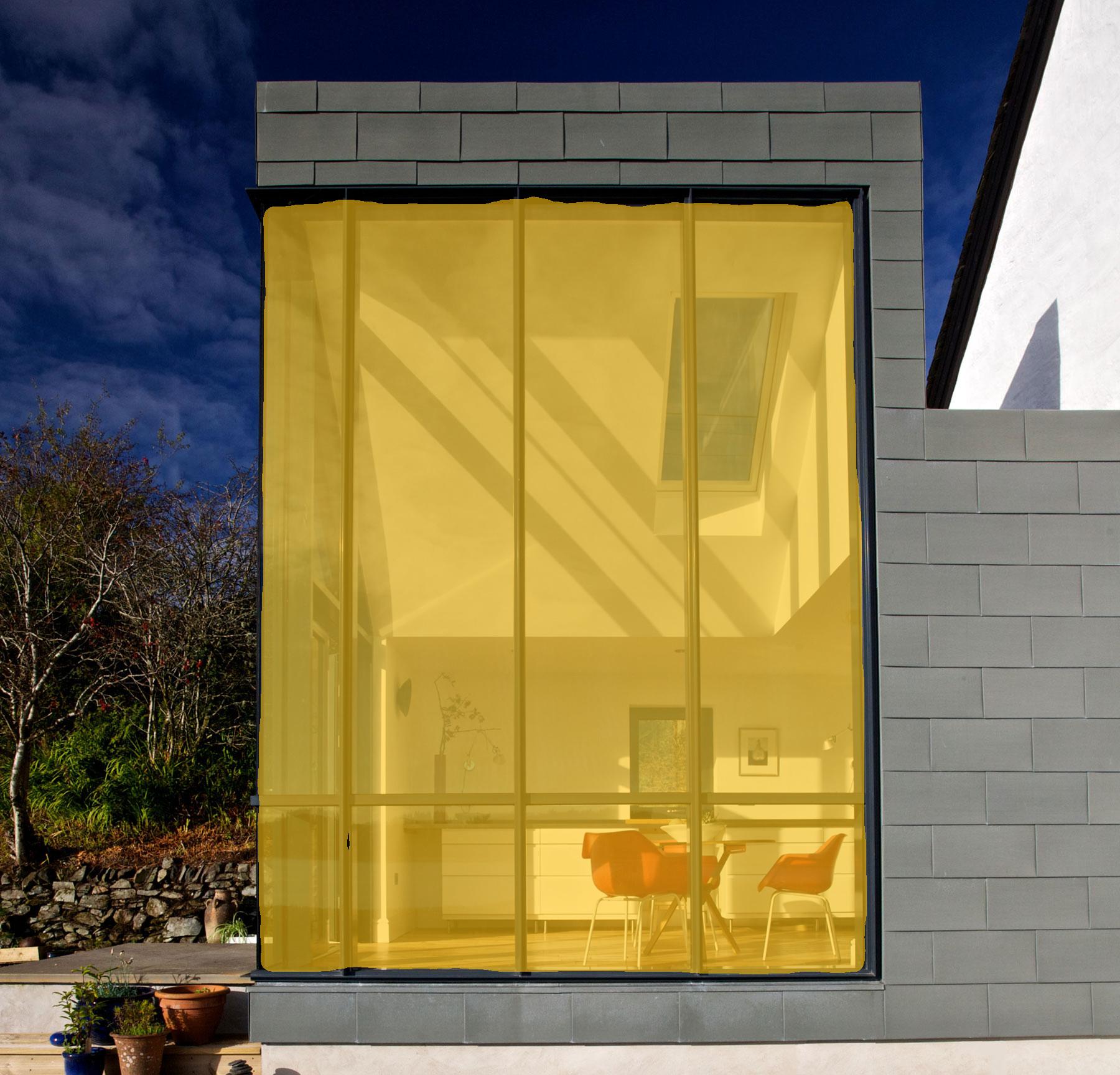}&
		\includegraphics[width=\wmore, height=\hmore]{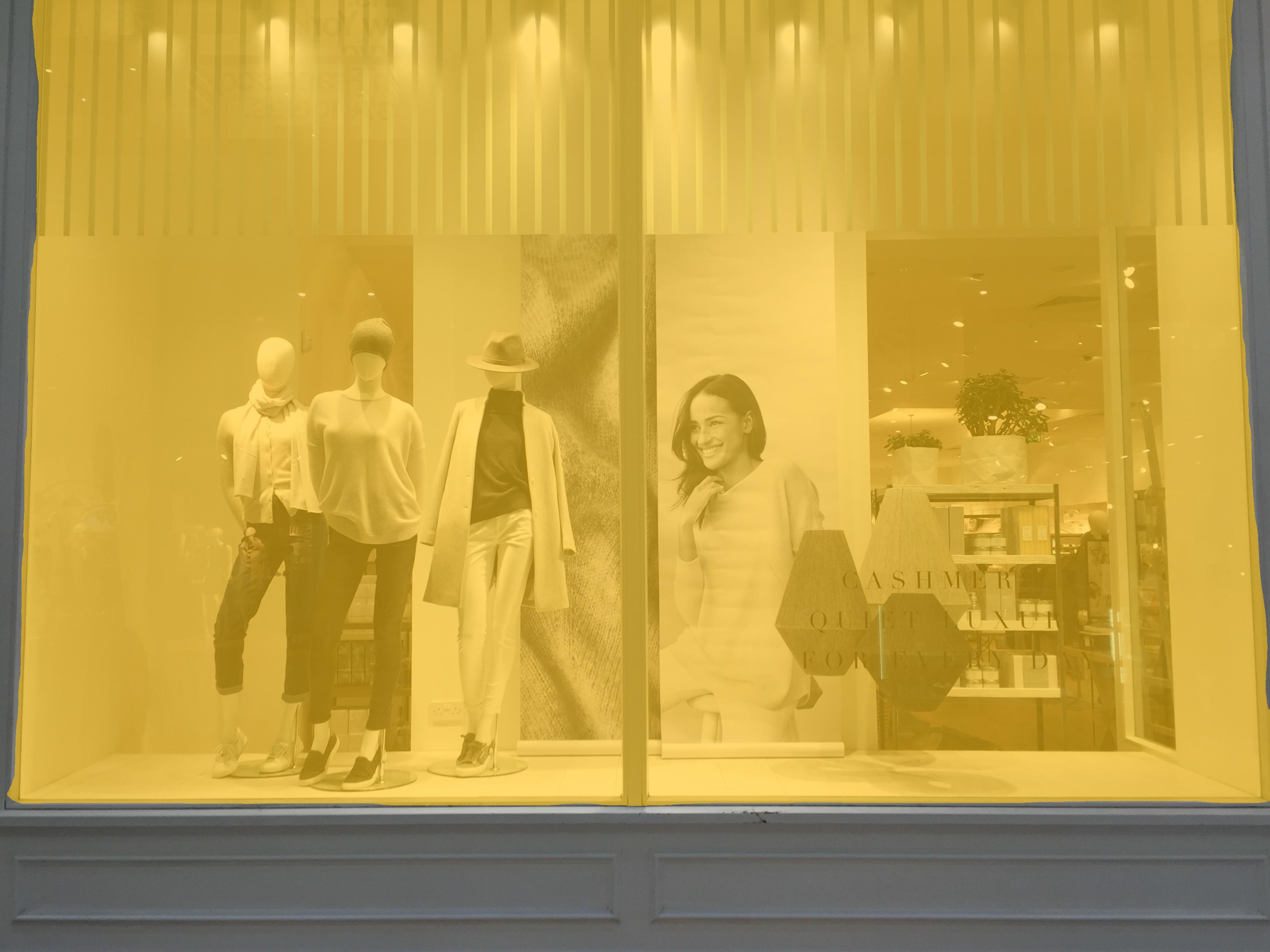}&
		\includegraphics[width=\wmore, height=\hmore]{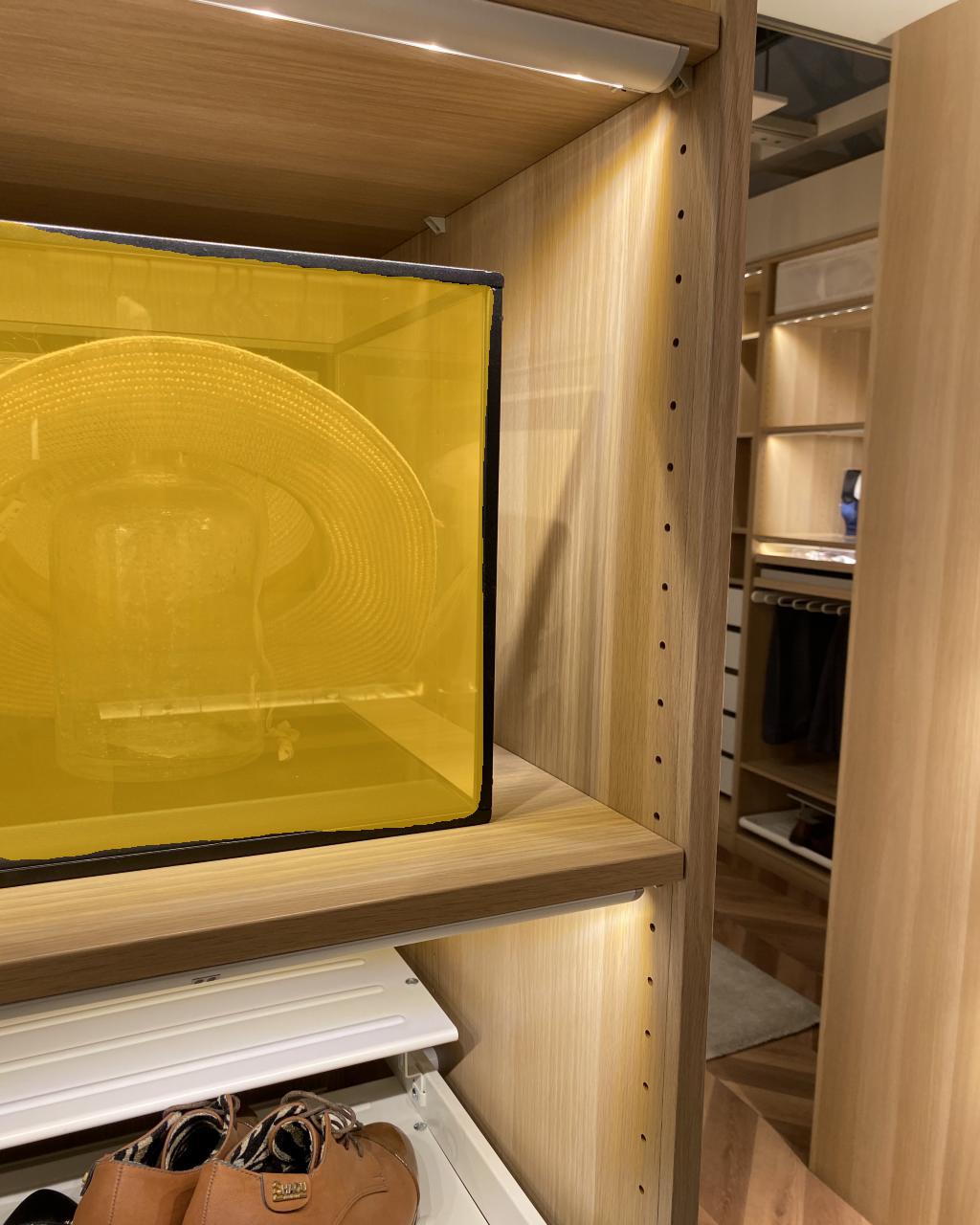}&
		\includegraphics[width=\wmore, height=\hmore]{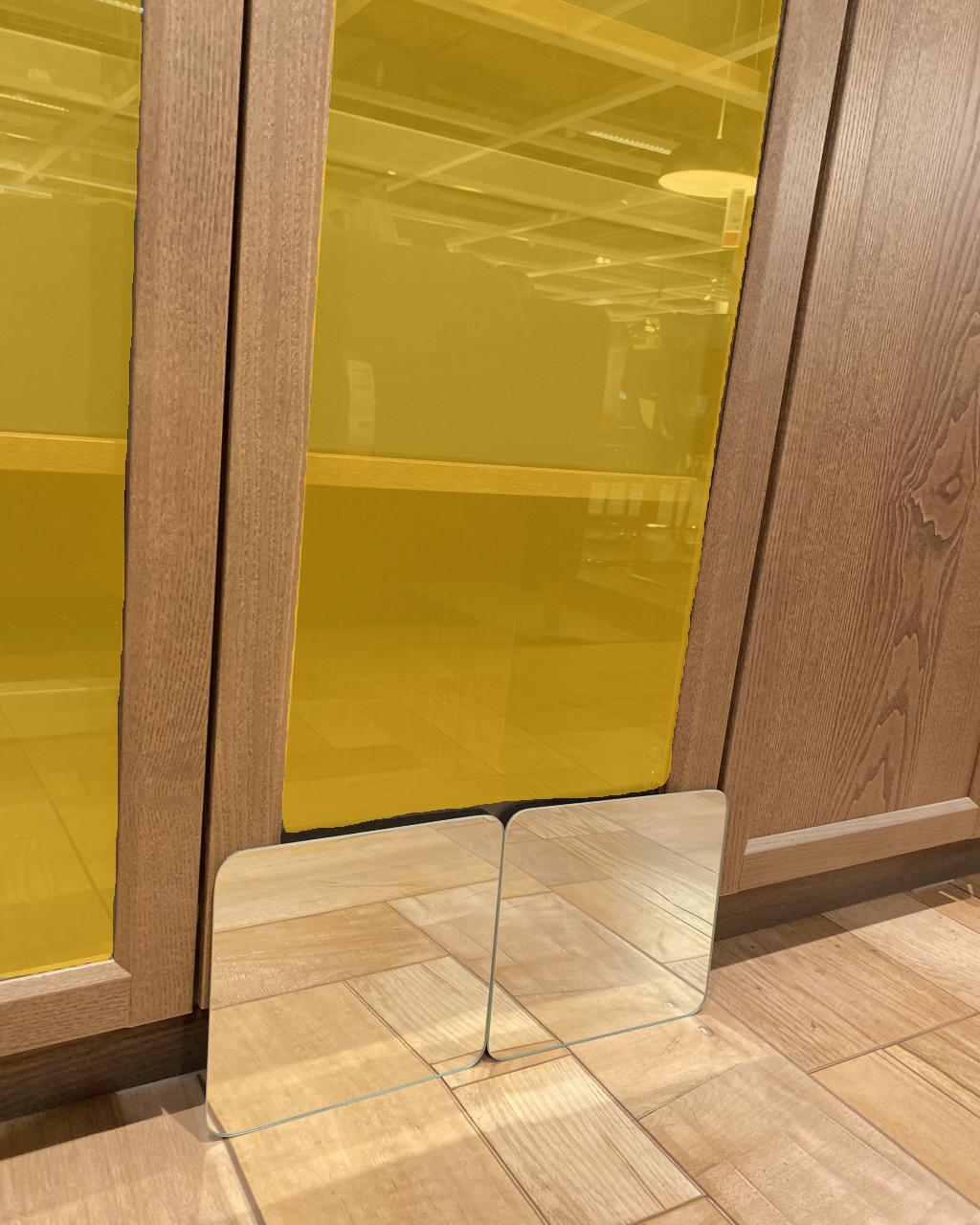} \\
	\end{tabular}
	\caption{More glass detection results of our GDNet-B on images beyond the GDD testing set.}
	\label{fig:more}
\end{figure*}

\add{\textbf{Impact of different multi-level feature extractors.}}
\add{We further explore the impact of different multi-level feature extractors in GDNet-B for glass detection via adopting different backbone architectures. In our experiment, three popular architectures are used (\textit{i.e.}, ResNet-101 \cite{he2016deep_resnet}, ResNeXt-101 \cite{Xie_2017_CVPR}, and SENet-154 \cite{hu2018squeeze}) and the weights of these backbones are initialized with the corresponding pre-trained models on ImageNet. From Table \ref{tab:ablation_backbone}, we can see that the performance of the multi-level feature extractor has a direct impact on the glass detection performance.}

\subsection{More Glass Detection Results}
Figure \ref{fig:more} further shows some glass detection results of GDNet-B on images beyond the GDD testing set, \emph{i.e.}, images selected from the ADE20K dataset \cite{Zhou_2017_CVPR} (the 1\textit{st} column), Matterport3D dataset \cite{Matterport3D} (the 2\textit{nd} column), and 2D3DS \cite{armeni2017joint} dataset (the 3\textit{rd} column), and images downloaded from the Internet (4-12\emph{th} columns). We can see that GDNet-B performs well under these various cases, demonstrating its effectiveness.

\begin{table}[t]
	\centering
	\caption{\add{Ablation studies on the impact of different multi-level feature extractors.}}
	\begin{tabular}{p{4.2cm}|p{1cm}<{\centering}p{0.9cm}<{\centering}p{0.8cm}<{\centering}}
		\hline
		\hline
		Multi-Level Feature Extractors & IoU$\uparrow$ & $F_\beta^m$$\uparrow$ & BER$\downarrow$ \\
		\hline
		\hline
		GDNet-B w/ ResNet-101 \cite{he2016deep_resnet}  & 86.84 & 0.934 & 5.95 \\
		GDNet-B w/ ResNeXt-101 \cite{Xie_2017_CVPR}     & 87.83 & \textbf{0.939} & 5.52 \\
		GDNet-B w/ SENet-154 \cite{hu2018squeeze}       & \textbf{88.09} & \textbf{0.939} & \textbf{5.48} \\
		\hline
		\hline
	\end{tabular}
	\label{tab:ablation_backbone}
\end{table}

\section{Experiments on Mirror Segmentation}\label{sec:mirror}
With the help of the well-designed large-field contextual feature integration module, our GDNet-B can explore abundant contextual information from a large field-of-view, and thus has the potential to handle other challenging vision tasks. In this section, we consider the mirror segmentation task. \mhy{It has been pointed out by previous work \cite{yang2019mirrornet} that there typically exists high-level and low-level contextual contrast between inside and outside the mirror. As our method can perceive rich contextual cues, we believe that the contextual contrast could be modeled implicitly by our GDNet-B in the end-to-end training process based on the mirror segmentation dataset.}


\subsection{Experimental Settings}
\mhy{\textbf{Benchmark datasets.} We conduct the experiments on the benchmark mirror segmentation dataset (MSD) \cite{yang2019mirrornet}. MSD \cite{yang2019mirrornet} is the first large-scale mirror dataset in the deep learning era. It consists of 4,018 images containing mirrors and their corresponding manually annotated mirror masks, taken from diverse daily life scenes. It is divided into 3,063 training images and 955 testing images.}

\mhy{\textbf{Implementation details.} We re-train our GDNet-B on the MSD \cite{yang2019mirrornet} training set and dubbed this model as GDNet-B-M. The training process takes about 23 hours for 200 epochs and the inference takes 0.034 seconds on each image.}

\mhy{\textbf{Evaluation metrics.} Following \cite{yang2019mirrornet}, we adopt five metrics to quantitatively evaluate the mirror segmentation performance of GDNet-B-M, \emph{i.e.}, intersection over union ($IoU$), pixel accuracy ($PA$), maximum F-measure ($F_\beta^m$), mean absolute error ($M$), and balance error rate ($BER$).}

\subsection{Comparison with the State-of-the-arts}
\mhy{We compare our GDNet-B-M against 10 state-of-the-art related methods, including PSPNet \cite{zhao2017pyramid}, ICNet \cite{Zhao_2018_ECCV_icnet}, Mask RCNN~\cite{He_2017_ICCV}, DSS \cite{HouPami19Dss}, PiCANet \cite{liu2018picanet}, RAS \cite{Chen_2018_ECCV}, R\textsuperscript{3}Net \cite{deng2018r3net}, DSC \cite{Hu_2018_CVPR}, BDRAR \cite{Zhu_2018_ECCV}, and MirrorNet \cite{yang2019mirrornet}.}

\mhy{Table \ref{tab:mirror} shows the quantitative comparison results, in which we can see that GDNet-B-M outperforms all the other counterparts. This demonstrates that large-field contextual information and boundary cues can effectively boost the performance of mirror segmentation. In addition, Figure \ref{fig:mirror} shows some qualitative comparison results. Benefited by the rich contextual features extracted by the LCFI module and the boundary information perceived by the BFE module, GDNet-B-M successfully suppresses the mirror-like regions (\emph{e.g.}, the first row) and handles the large variations inside the mirror regions (\emph{e.g.}, the last row), leading a more robust mirror detection.}

\begin{table}[tbp]
	\centering
	\caption{\mhy{Quantitative comparison to state-of-the-art methods on the MSD \cite{yang2019mirrornet} testing set. All methods are re-trained on the MSD \cite{yang2019mirrornet} training set. * denotes using CRF~\cite{krahenbuhl2011efficient} for post-processing. ``Statistics" refers to thresholding mirror location statistics from the MSD \cite{yang2019mirrornet} training set as a mirror mask for detection. The first, second, and third best results are marked in \textcolor[rgb]{1,0,0}{\textbf{red}}, \textcolor[rgb]{0,1,0}{\textbf{green}}, and \textcolor[rgb]{0,0,1}{\textbf{blue}}, respectively. Our GDNet-B-M achieves superior performance on all five evaluation metrics.}}
	\setlength{\tabcolsep}{.8pt}
	\centering
	\begin{tabular}{p{2.0cm}|p{1.3cm}<{\centering}|p{1.0cm}<{\centering}|p{1.0cm}<{\centering}|p{1.0cm}<{\centering}|p{1.0cm}<{\centering}|p{1.0cm}<{\centering}}
		\hline
		\hline
		method & Pub.'Year & IoU$\uparrow$ & PA$\uparrow$ & $F_\beta^m$$\uparrow$ & M$\downarrow$ & BER$\downarrow$ \\
		\hline
		\hline
		Statistics & - & 30.83 & 0.701 & 0.438 & 0.358 & 32.89 \\
		\hline
		\hline
		PSPNet \cite{zhao2017pyramid} & CVPR'17 & 63.21 & 0.892 & 0.746 & 0.117 & 15.82 \\
		ICNet \cite{Zhao_2018_ECCV_icnet} & ECCV'18 & 57.25 & 0.878 & 0.710 & 0.124 & 18.75 \\
		\hline
		\hline
		Mask RCNN \cite{He_2017_ICCV} & ICCV'17 & 63.18 & 0.884 & 0.819 & 0.095 & 14.35 \\
		\hline
		\hline
		DSS \cite{HouPami19Dss} & TPAMI'19 & 59.11 & 0.896 & 0.743 & 0.125 & 18.81 \\
		PiCANet \cite{liu2018picanet} & CVPR'18 & 71.78 & 0.918 & 0.808 & 0.088 & 10.99 \\
		RAS \cite{Chen_2018_ECCV} & ECCV'18 & 60.48 & 0.896 & 0.758 & 0.111 & 17.60 \\
		R\textsuperscript{3}Net* \cite{deng2018r3net} & IJCAI'18 & \textcolor[rgb]{0,0,1}{\textbf{73.21}} & \textcolor[rgb]{0,0,1}{\textbf{0.933}} & \textcolor[rgb]{0,0,1}{\textbf{0.847}} & \textcolor[rgb]{0,0,1}{\textbf{0.068}} & \textcolor[rgb]{0,0,1}{\textbf{11.39}} \\
		\hline
		\hline
		DSC \cite{Hu_2018_CVPR} & CVPR'18 & 69.71 & 0.917 & 0.812 & 0.087 & 11.77 \\
		BDRAR* \cite{Zhu_2018_ECCV} & ECCV'18 & 67.43 & 0.907 & 0.804 & 0.093 & 12.41 \\
		\hline
		\hline
		MirrorNet* \cite{yang2019mirrornet} & ICCV'19 & \textcolor[rgb]{0,1,0}{\textbf{78.95}} & \textcolor[rgb]{0,1,0}{\textbf{0.935}} & \textcolor[rgb]{0,1,0}{\textbf{0.857}} & \textcolor[rgb]{0,1,0}{\textbf{0.065}} &\textcolor[rgb]{0,1,0}{\textbf{6.39}} \\
		\hline
		\hline
		\textbf{GDNet-B-M} & \textbf{Ours} & \textcolor[rgb]{1,0,0}{\textbf{81.70}} & \textcolor[rgb]{1,0,0}{\textbf{0.952}} & \textcolor[rgb]{1,0,0}{\textbf{0.894}} & \textcolor[rgb]{1,0,0}{\textbf{0.050}} & \textcolor[rgb]{1,0,0}{\textbf{6.17}} \\
		\hline
		\hline
	\end{tabular}
	\label{tab:mirror}
\end{table}

\def\wvisual{0.078\linewidth}
\def\hvisualshort{.45in}
\def\hvisualtall{.58in}
\begin{figure*}
	\setlength{\tabcolsep}{1.6pt}
	\centering
	\scriptsize
	\begin{tabular}{cccccccccccc}
		\includegraphics[width=\wvisual, height=\hvisualtall]{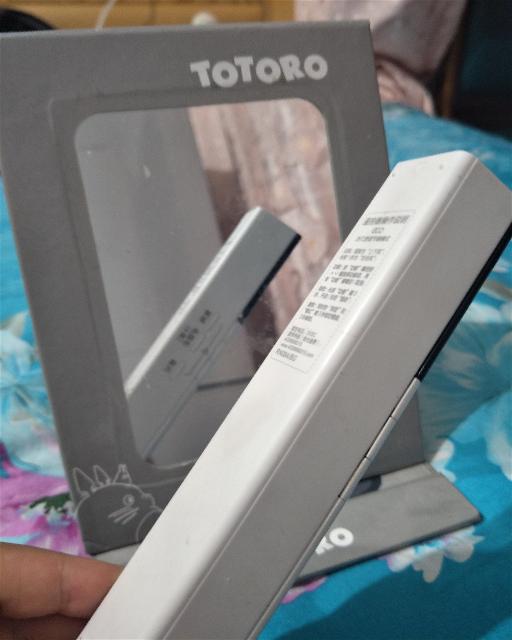}&
		\includegraphics[width=\wvisual, height=\hvisualtall]{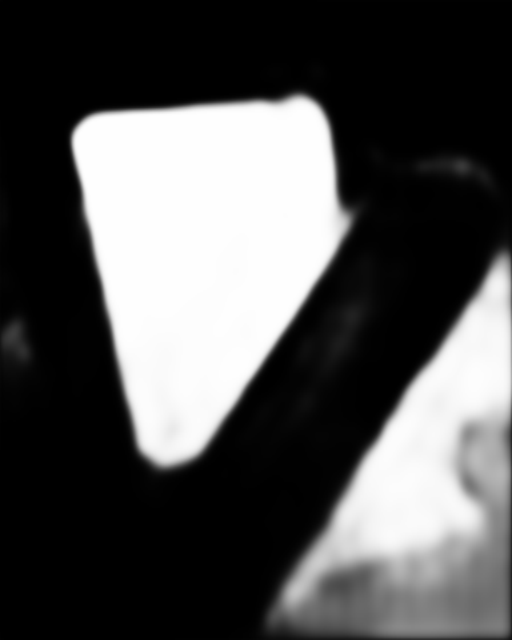}&
		\includegraphics[width=\wvisual, height=\hvisualtall]{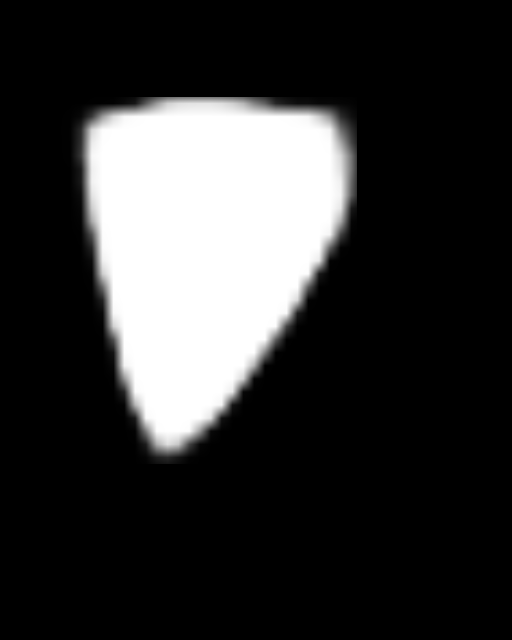}&
		\includegraphics[width=\wvisual, height=\hvisualtall]{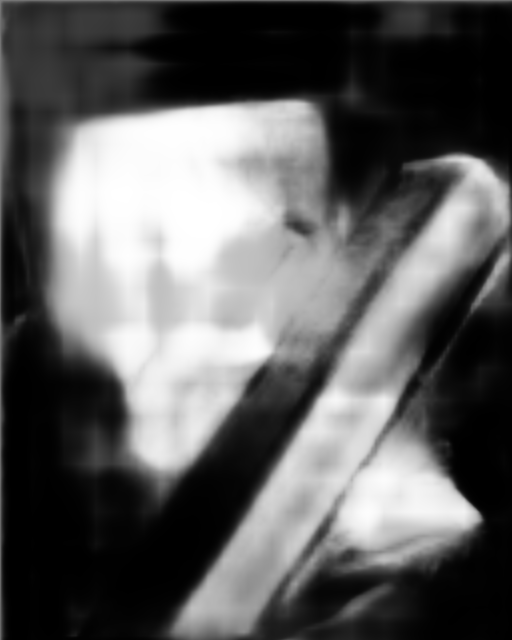}&
		\includegraphics[width=\wvisual, height=\hvisualtall]{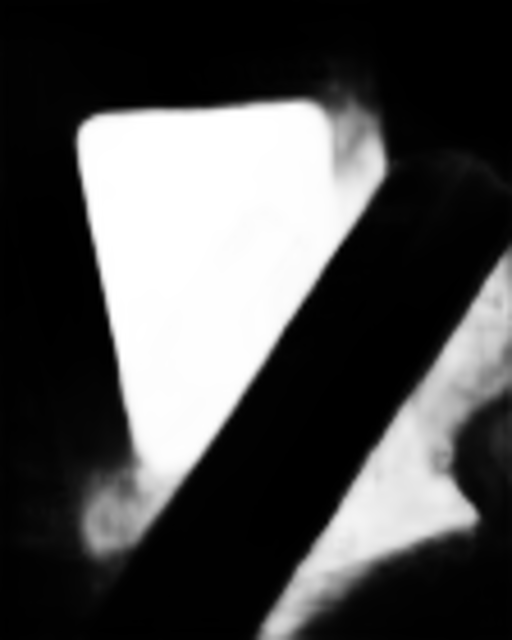}&
		\includegraphics[width=\wvisual, height=\hvisualtall]{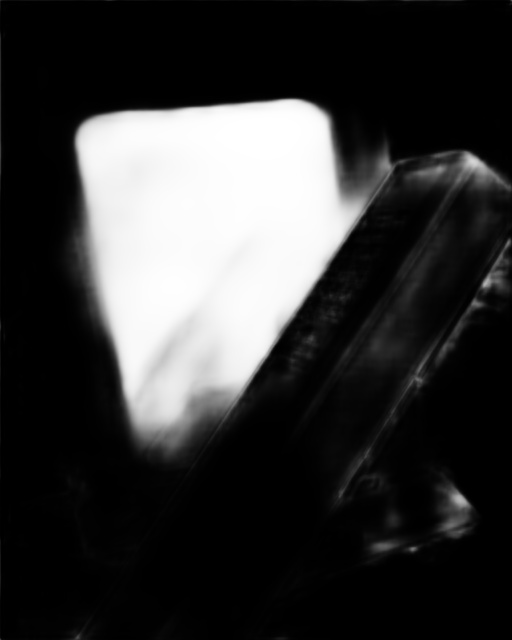}&
		\includegraphics[width=\wvisual, height=\hvisualtall]{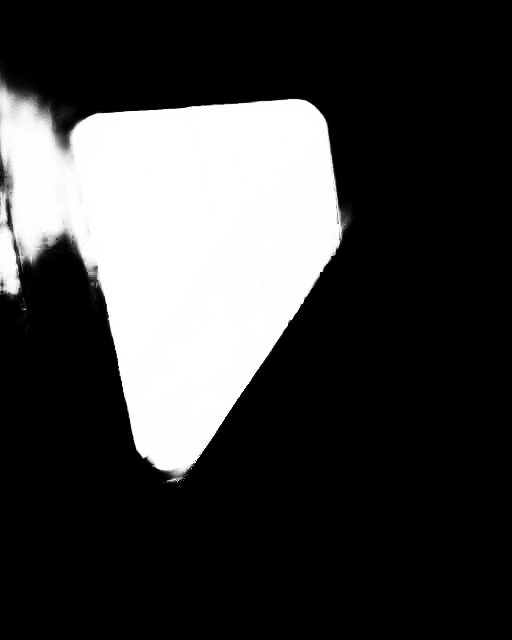}&
		\includegraphics[width=\wvisual, height=\hvisualtall]{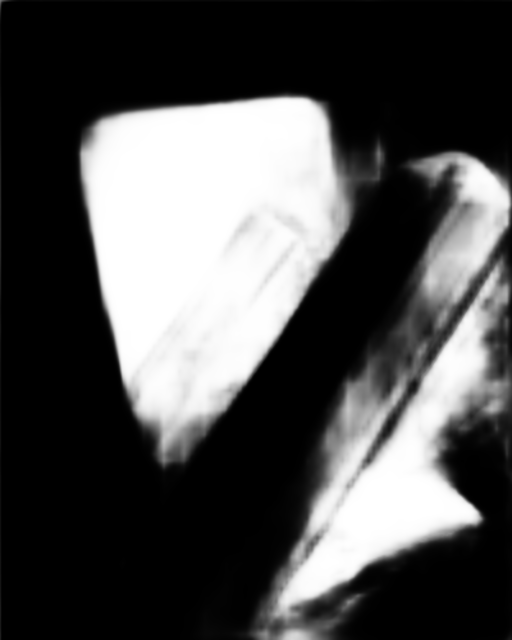}&
		\includegraphics[width=\wvisual, height=\hvisualtall]{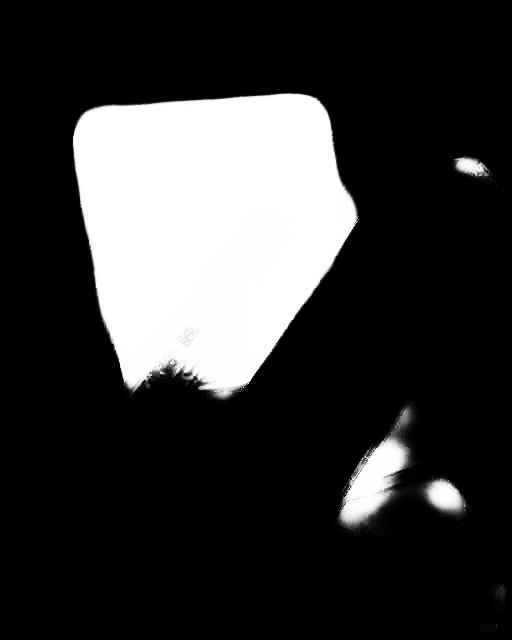}&
		\includegraphics[width=\wvisual, height=\hvisualtall]{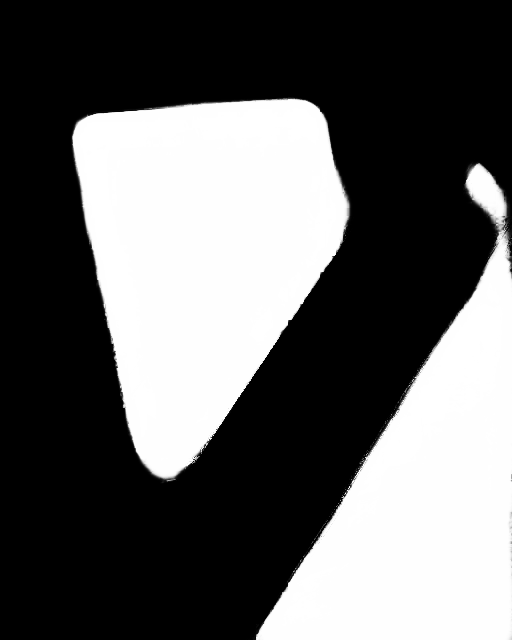}&
		\includegraphics[width=\wvisual, height=\hvisualtall]{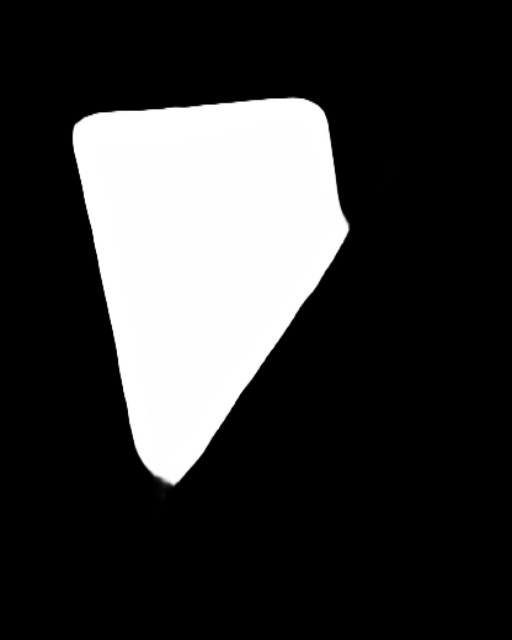}&
		\includegraphics[width=\wvisual, height=\hvisualtall]{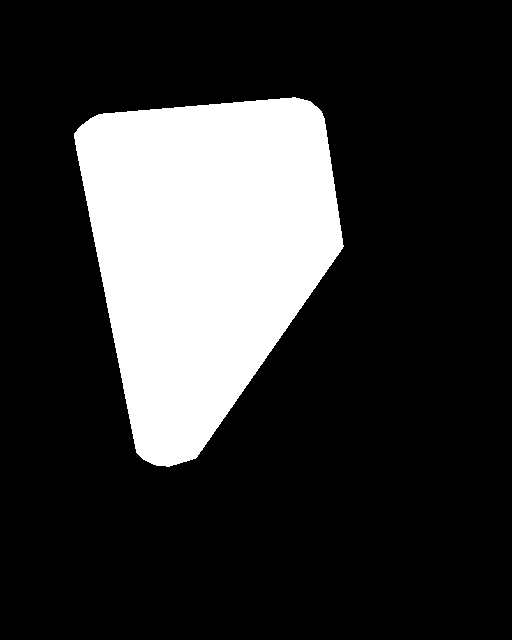} \\
		
		\includegraphics[width=\wvisual, height=\hvisualshort]{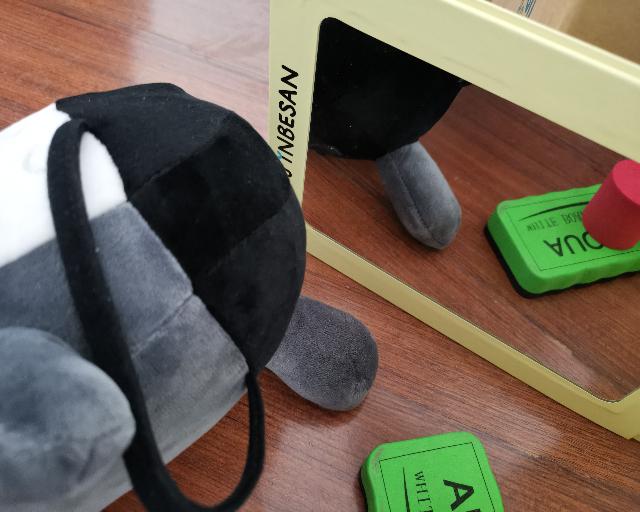}&
		\includegraphics[width=\wvisual, height=\hvisualshort]{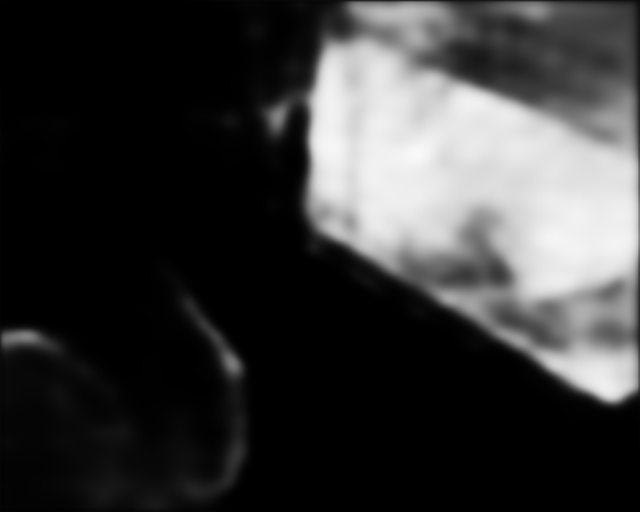}&
		\includegraphics[width=\wvisual, height=\hvisualshort]{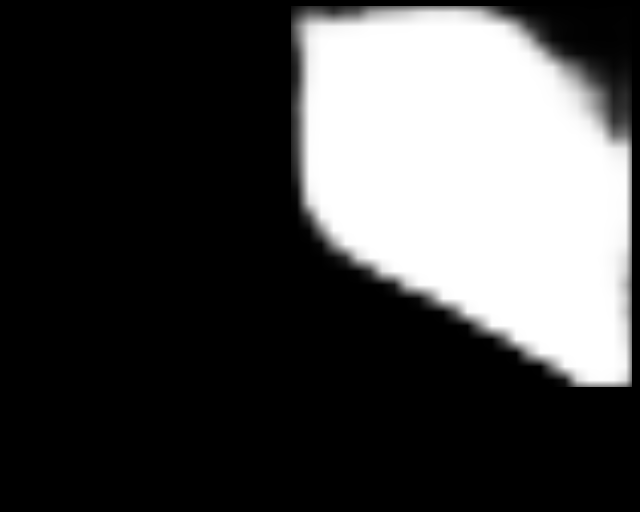}&
		\includegraphics[width=\wvisual, height=\hvisualshort]{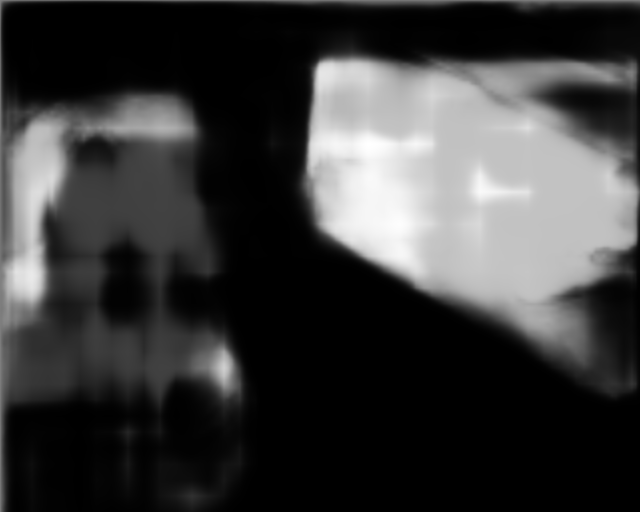}&
		\includegraphics[width=\wvisual, height=\hvisualshort]{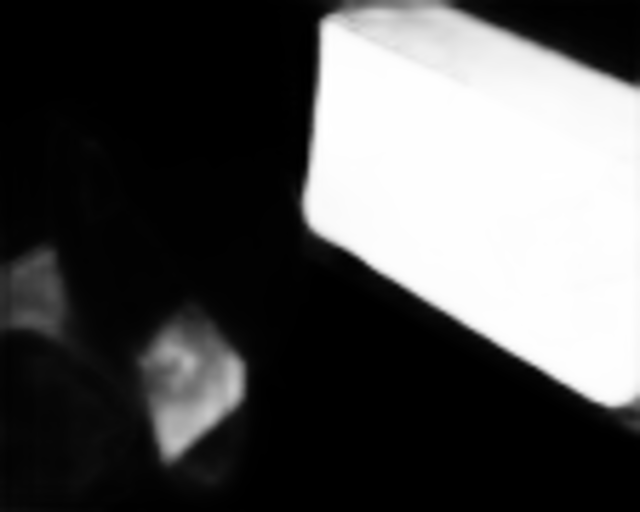}&
		\includegraphics[width=\wvisual, height=\hvisualshort]{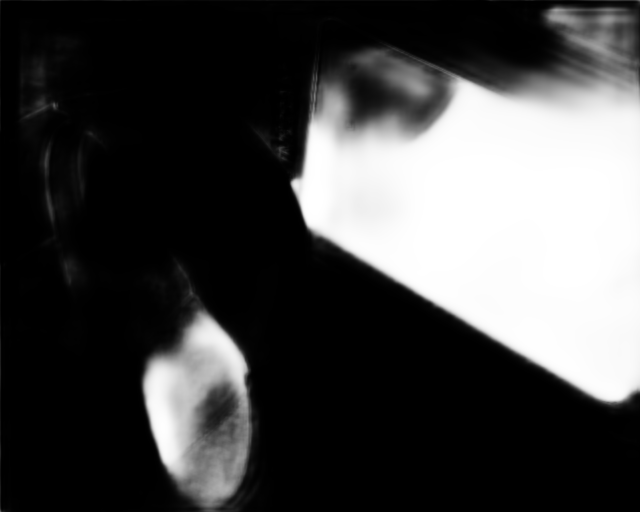}&
		\includegraphics[width=\wvisual, height=\hvisualshort]{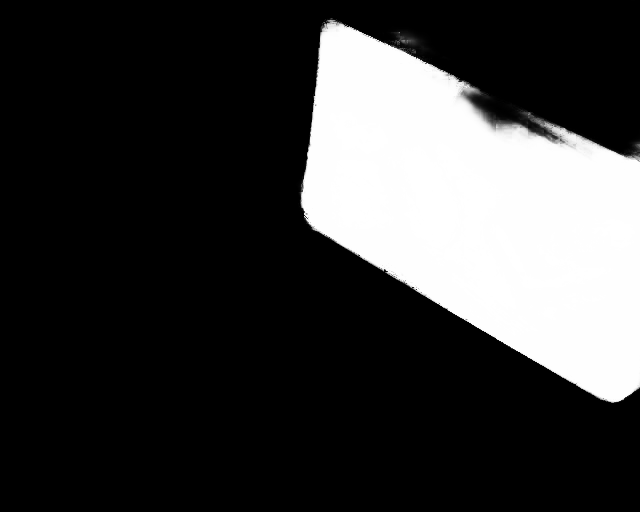}&
		\includegraphics[width=\wvisual, height=\hvisualshort]{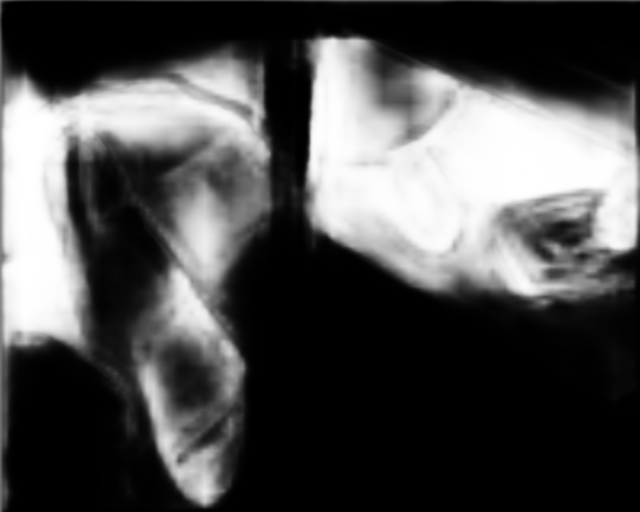}&
		\includegraphics[width=\wvisual, height=\hvisualshort]{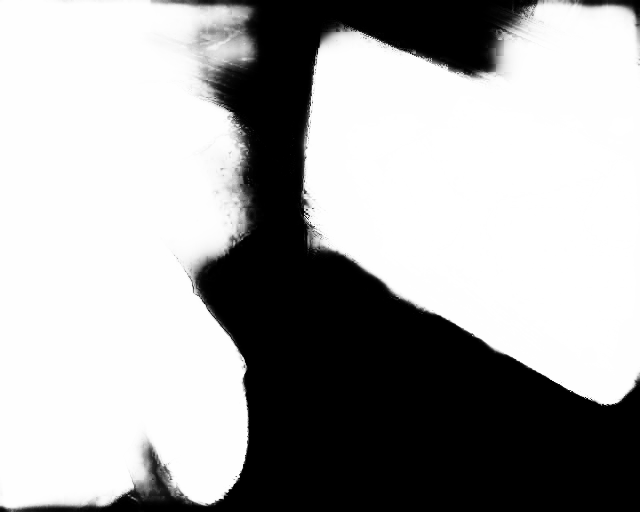}&
		\includegraphics[width=\wvisual, height=\hvisualshort]{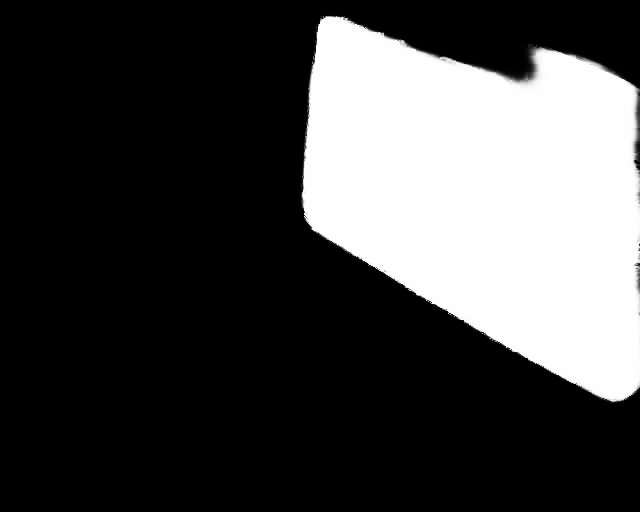}&
		\includegraphics[width=\wvisual, height=\hvisualshort]{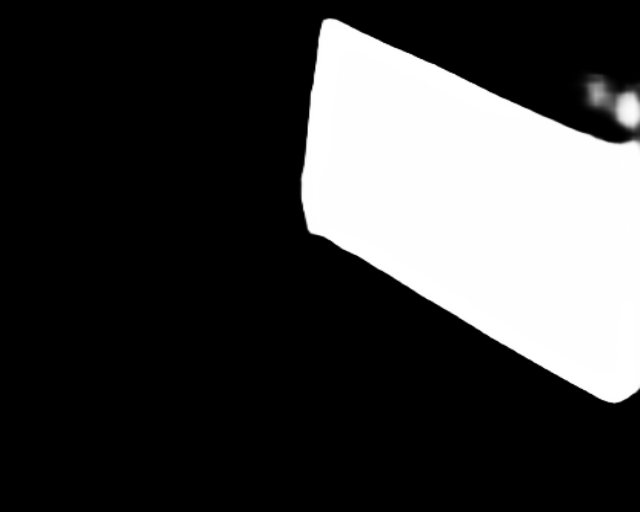}&
		\includegraphics[width=\wvisual, height=\hvisualshort]{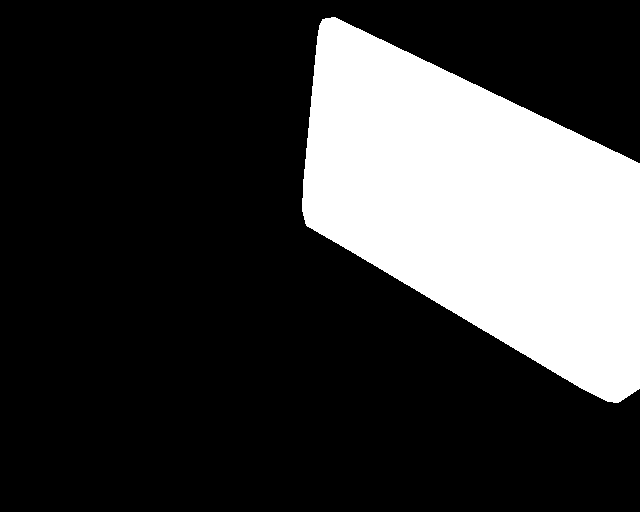} \\
		
		\includegraphics[width=\wvisual, height=\hvisualtall]{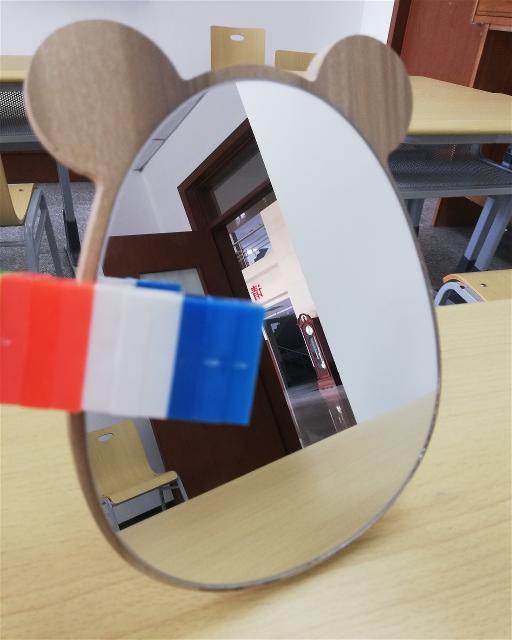}&
		\includegraphics[width=\wvisual, height=\hvisualtall]{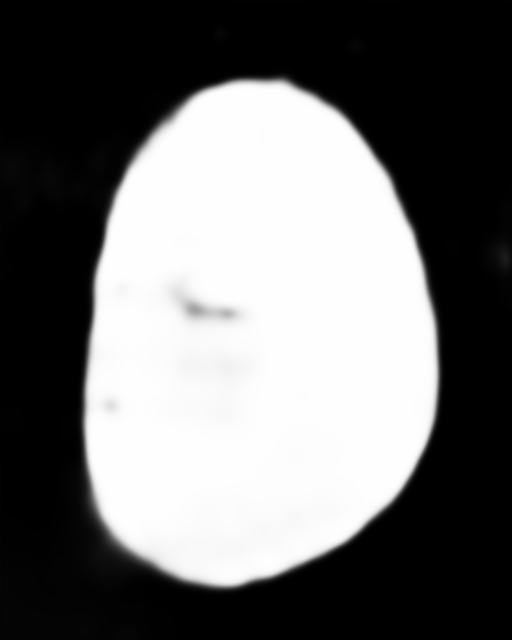}&
		\includegraphics[width=\wvisual, height=\hvisualtall]{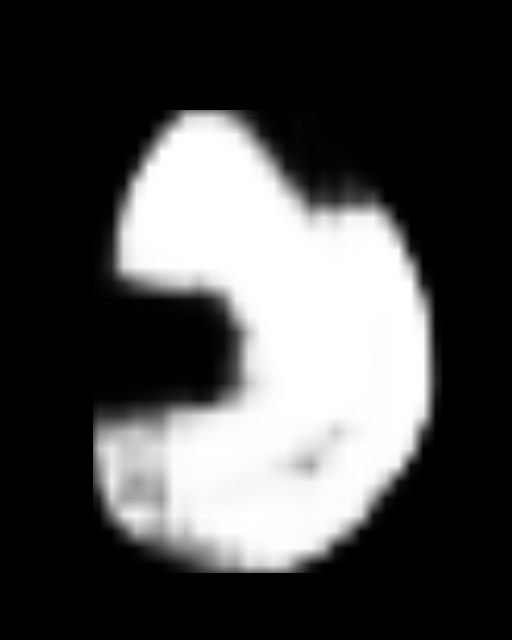}&
		\includegraphics[width=\wvisual, height=\hvisualtall]{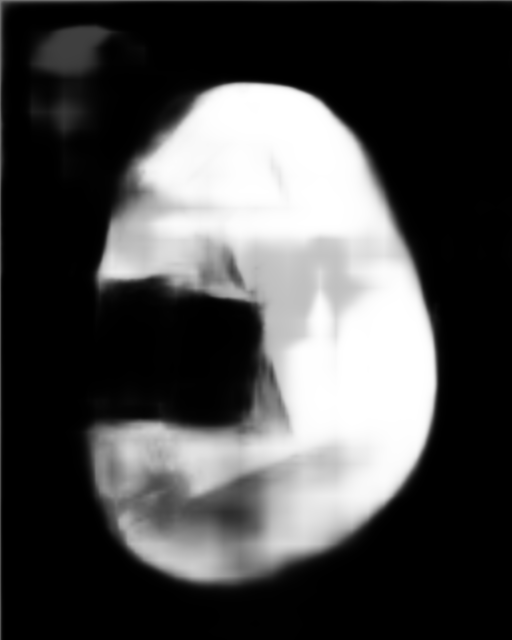}&
		\includegraphics[width=\wvisual, height=\hvisualtall]{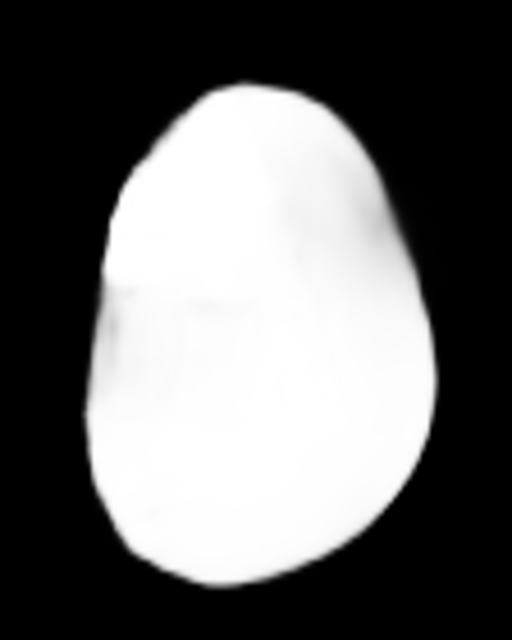}&
		\includegraphics[width=\wvisual, height=\hvisualtall]{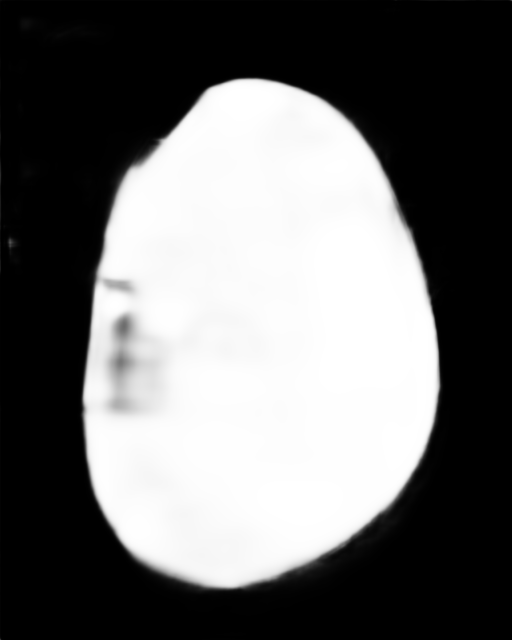}&
		\includegraphics[width=\wvisual, height=\hvisualtall]{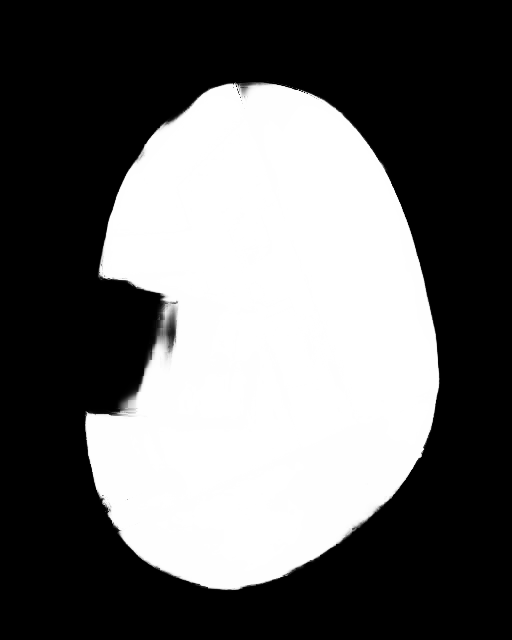}&
		\includegraphics[width=\wvisual, height=\hvisualtall]{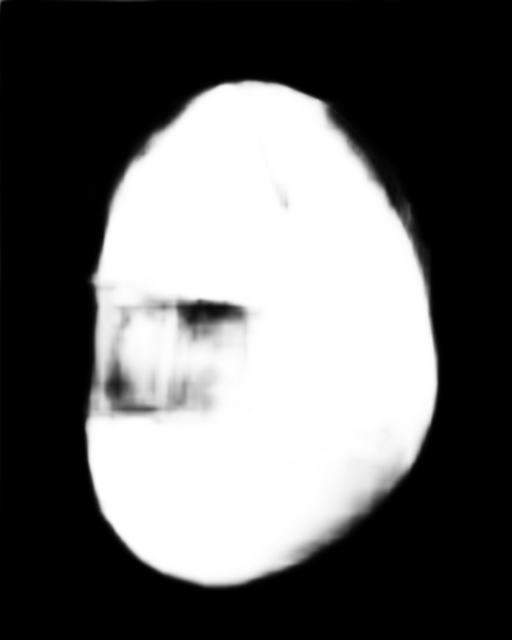}&
		\includegraphics[width=\wvisual, height=\hvisualtall]{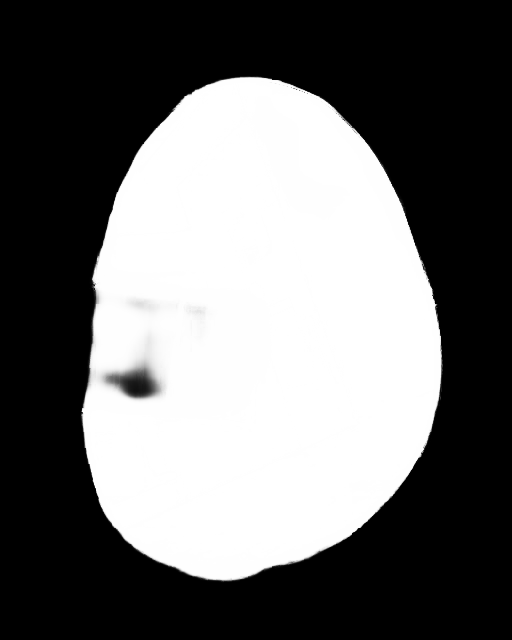}&
		\includegraphics[width=\wvisual, height=\hvisualtall]{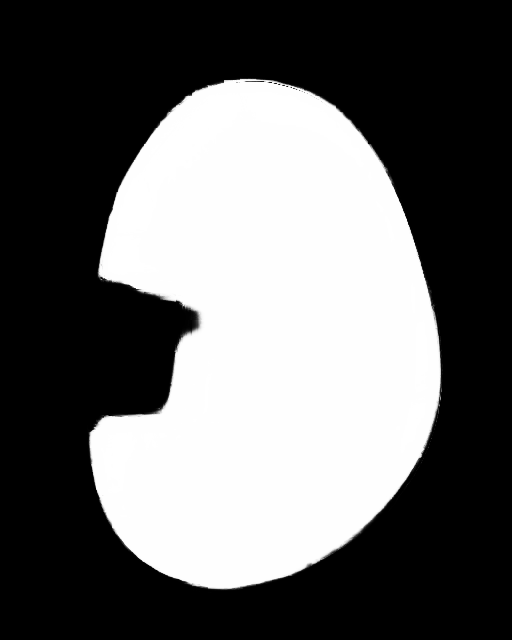}&
		\includegraphics[width=\wvisual, height=\hvisualtall]{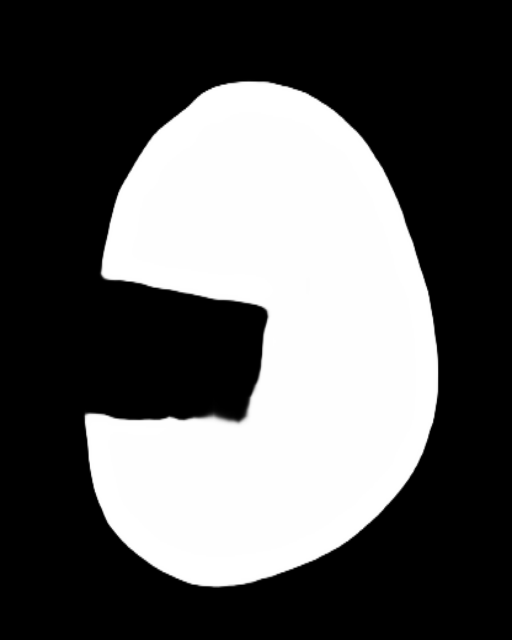}&
		\includegraphics[width=\wvisual, height=\hvisualtall]{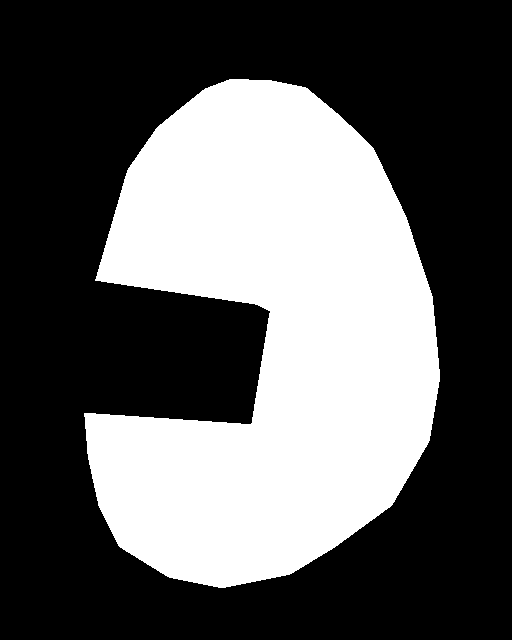} \\
		
		\includegraphics[width=\wvisual, height=\hvisualtall]{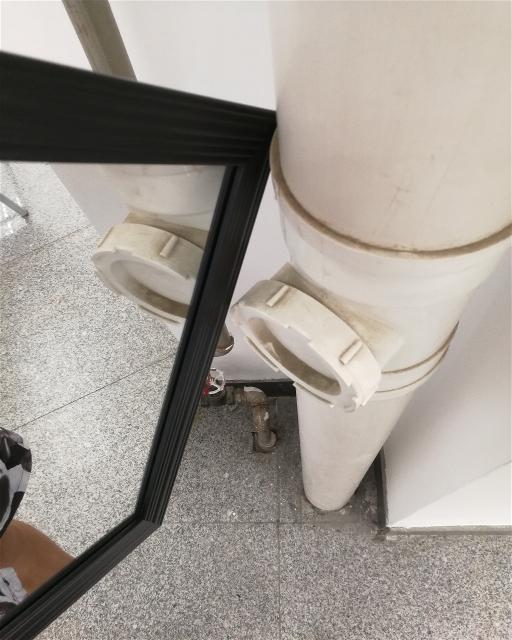}&
		\includegraphics[width=\wvisual, height=\hvisualtall]{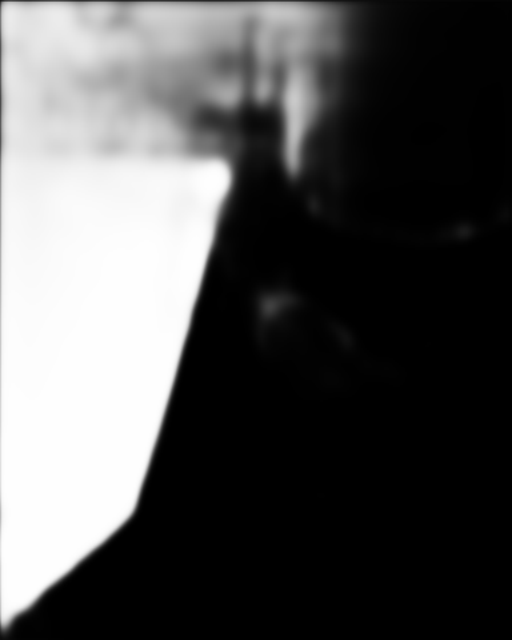}&
		\includegraphics[width=\wvisual, height=\hvisualtall]{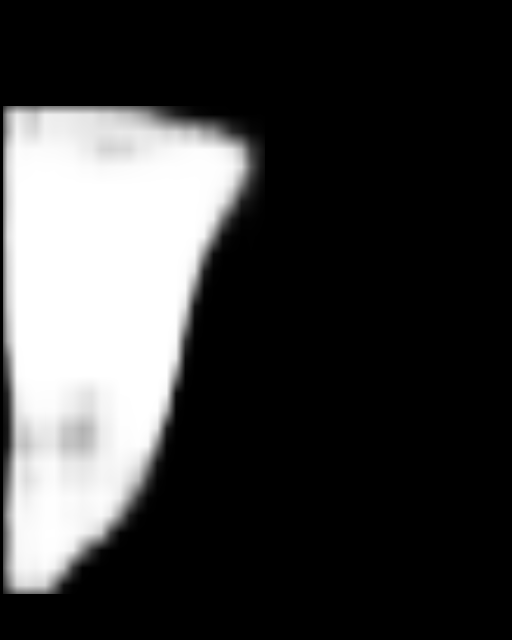}&
		\includegraphics[width=\wvisual, height=\hvisualtall]{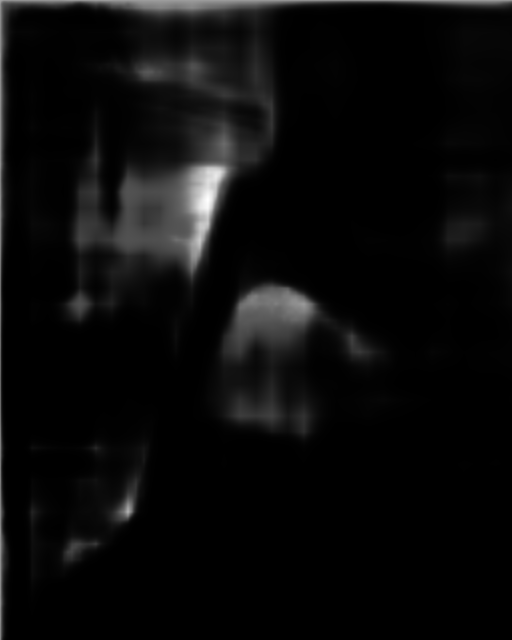}&
		\includegraphics[width=\wvisual, height=\hvisualtall]{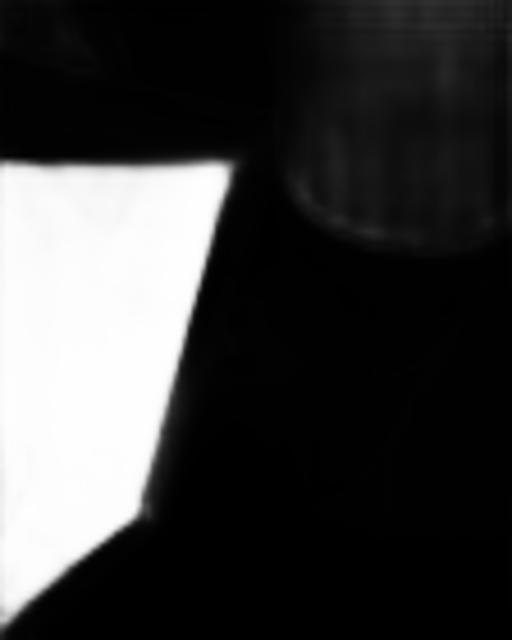}&
		\includegraphics[width=\wvisual, height=\hvisualtall]{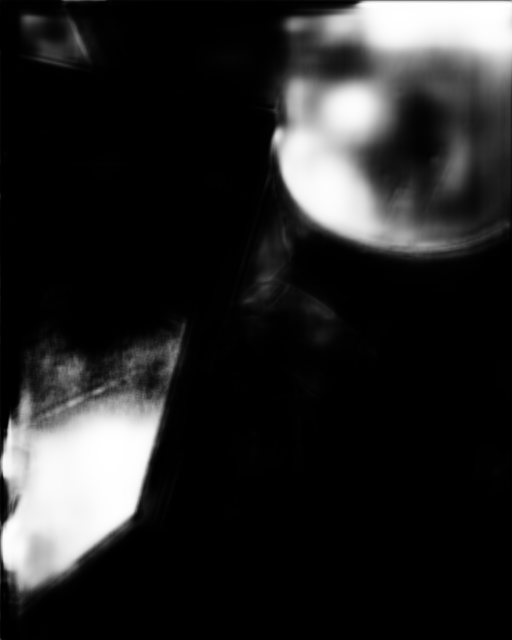}&
		\includegraphics[width=\wvisual, height=\hvisualtall]{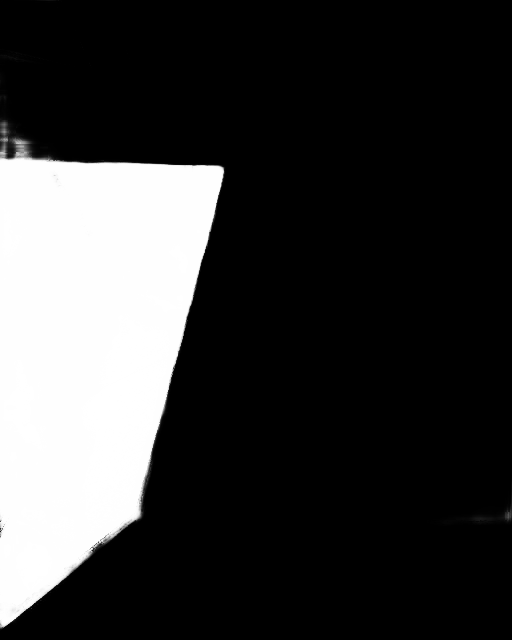}&
		\includegraphics[width=\wvisual, height=\hvisualtall]{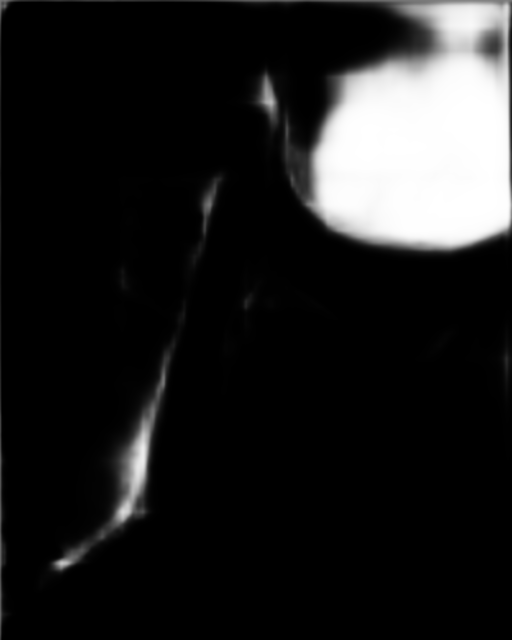}&
		\includegraphics[width=\wvisual, height=\hvisualtall]{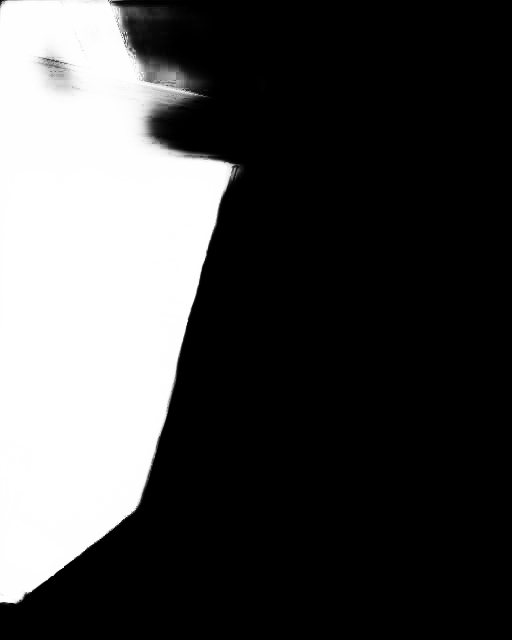}&
		\includegraphics[width=\wvisual, height=\hvisualtall]{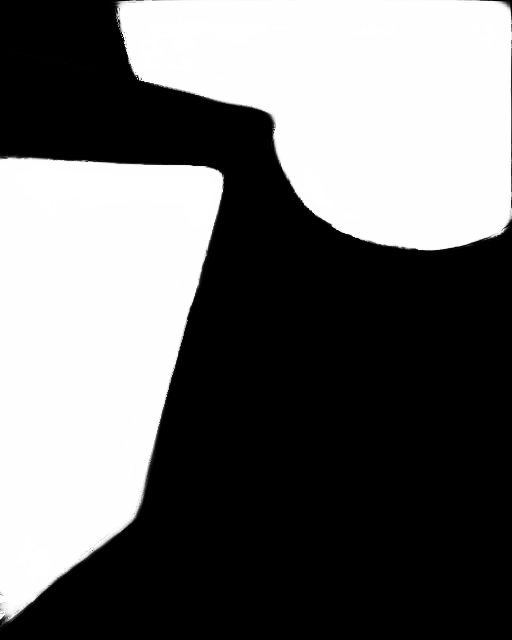}&
		\includegraphics[width=\wvisual, height=\hvisualtall]{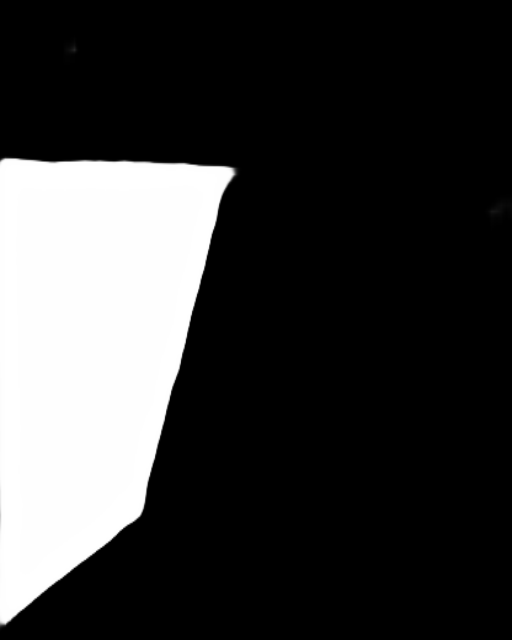}&
		\includegraphics[width=\wvisual, height=\hvisualtall]{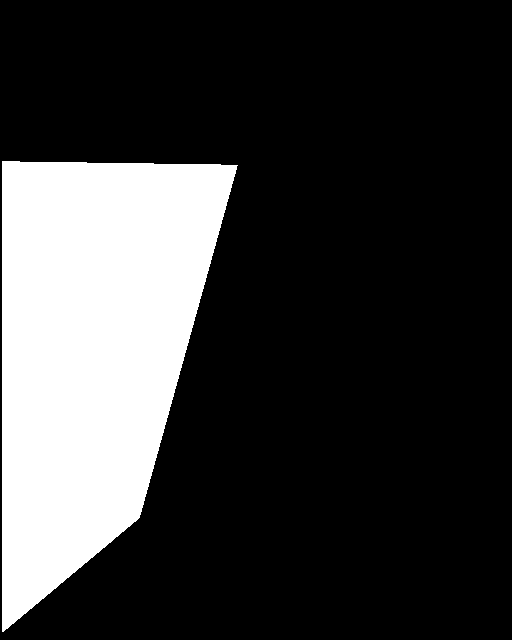} \\
		
		\includegraphics[width=\wvisual, height=\hvisualtall]{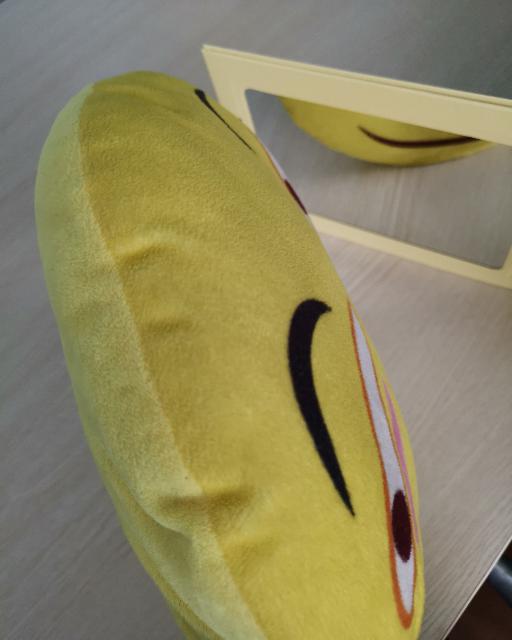}&
		\includegraphics[width=\wvisual, height=\hvisualtall]{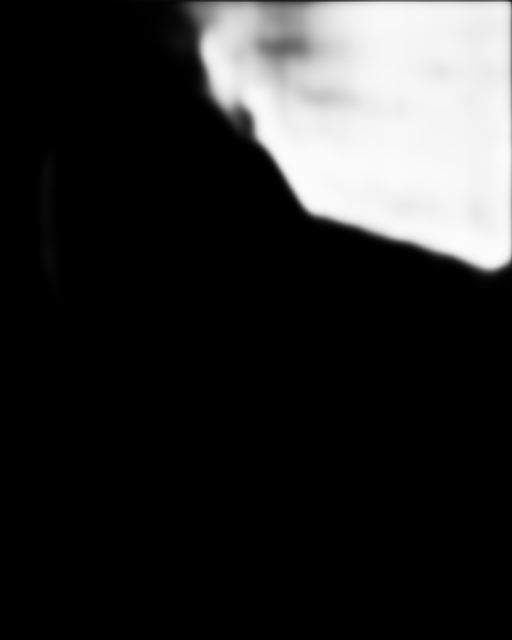}&
		\includegraphics[width=\wvisual, height=\hvisualtall]{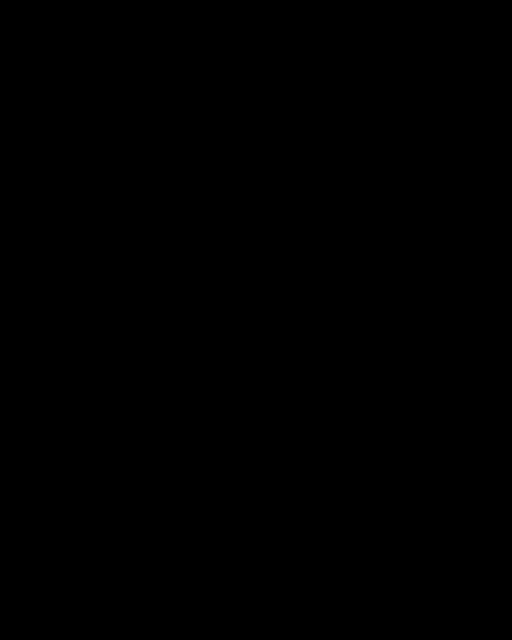}&
		\includegraphics[width=\wvisual, height=\hvisualtall]{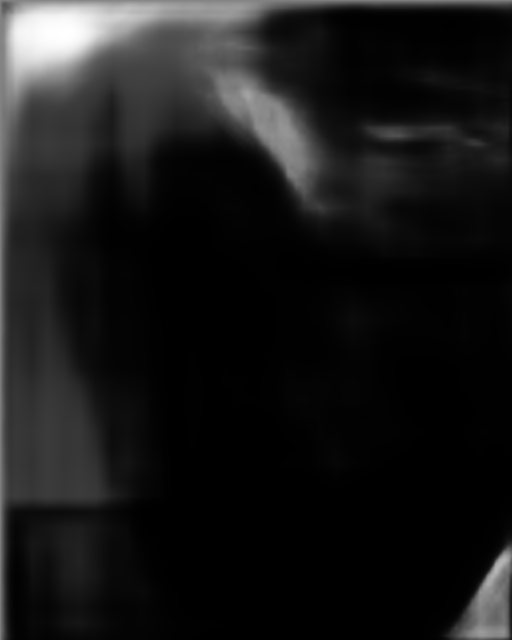}&
		\includegraphics[width=\wvisual, height=\hvisualtall]{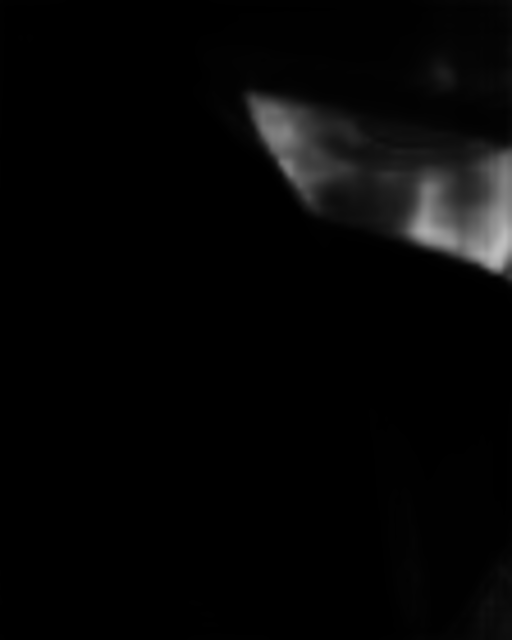}&
		\includegraphics[width=\wvisual, height=\hvisualtall]{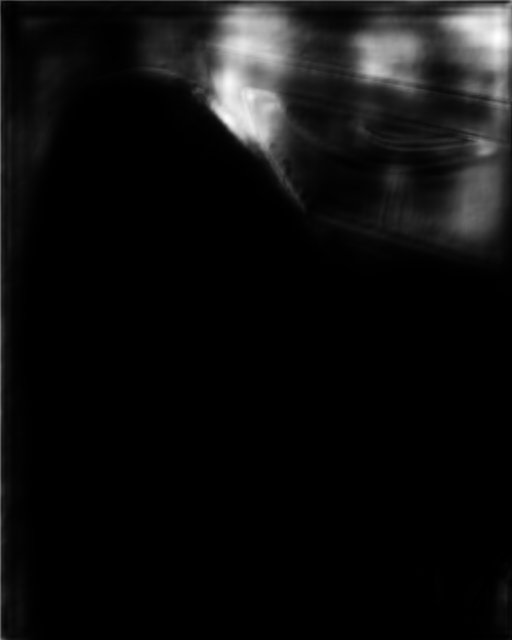}&
		\includegraphics[width=\wvisual, height=\hvisualtall]{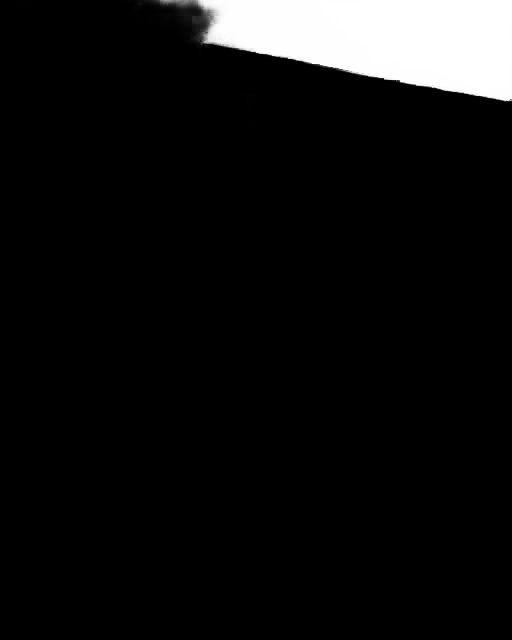}&
		\includegraphics[width=\wvisual, height=\hvisualtall]{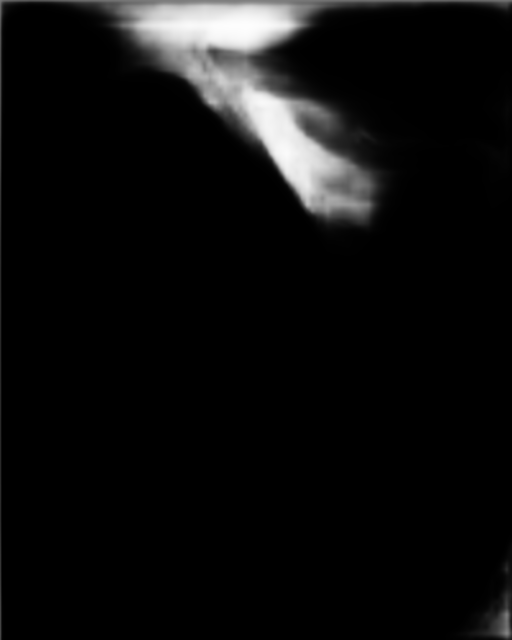}&
		\includegraphics[width=\wvisual, height=\hvisualtall]{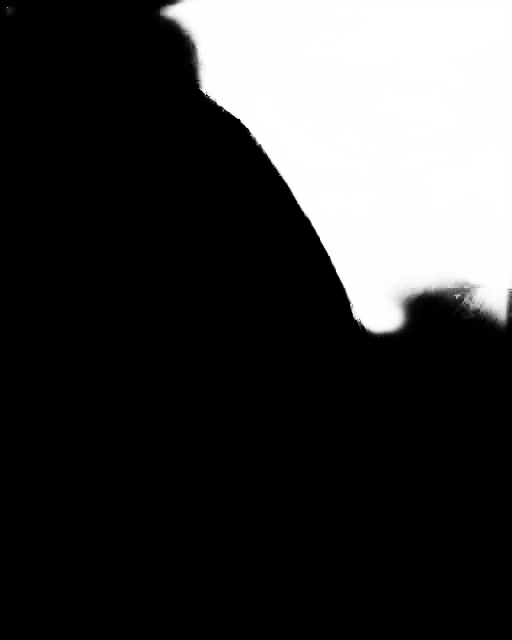}&
		\includegraphics[width=\wvisual, height=\hvisualtall]{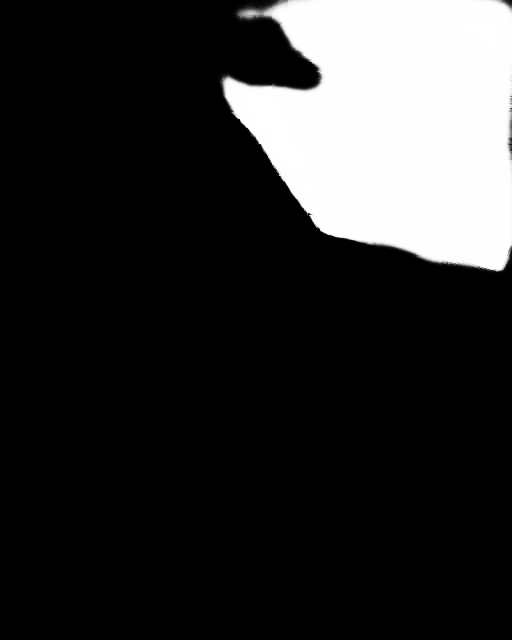}&
		\includegraphics[width=\wvisual, height=\hvisualtall]{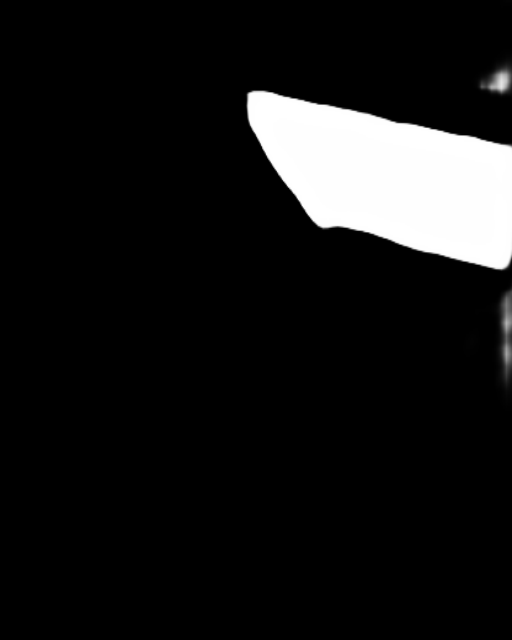}&
		\includegraphics[width=\wvisual, height=\hvisualtall]{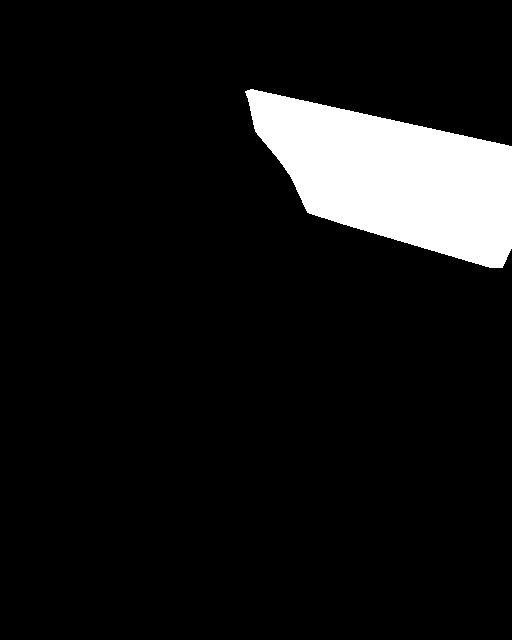} \\
		
		\includegraphics[width=\wvisual, height=\hvisualtall]{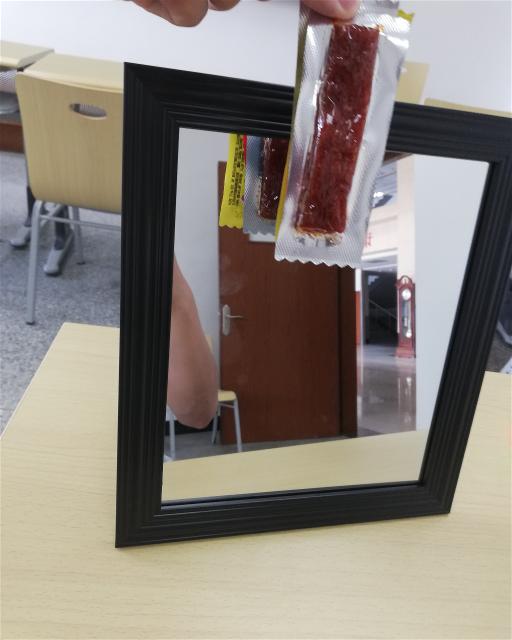}&
		\includegraphics[width=\wvisual, height=\hvisualtall]{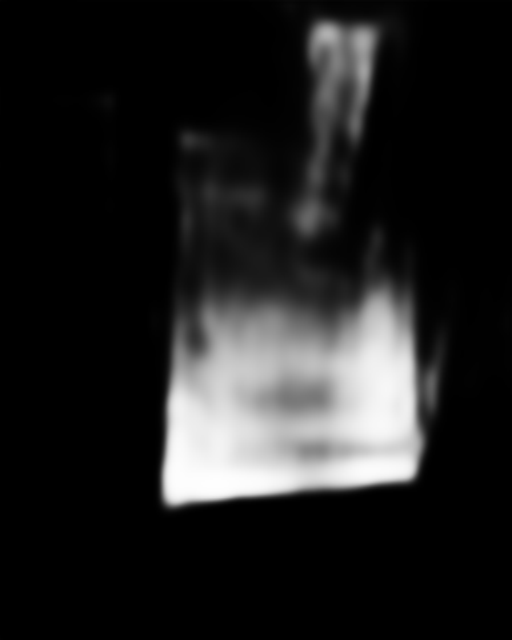}&
		\includegraphics[width=\wvisual, height=\hvisualtall]{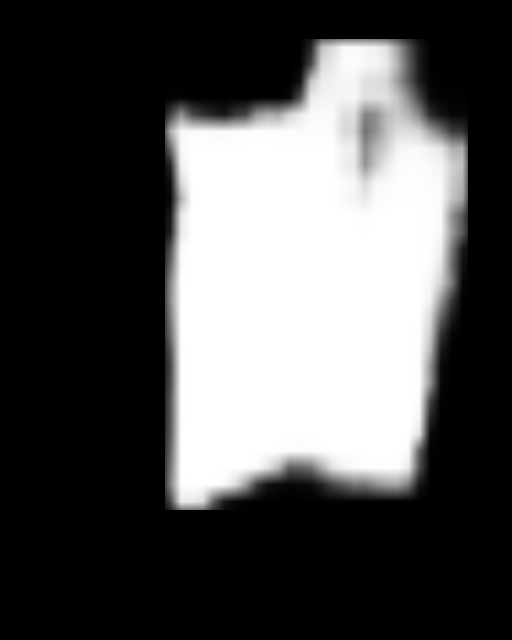}&
		\includegraphics[width=\wvisual, height=\hvisualtall]{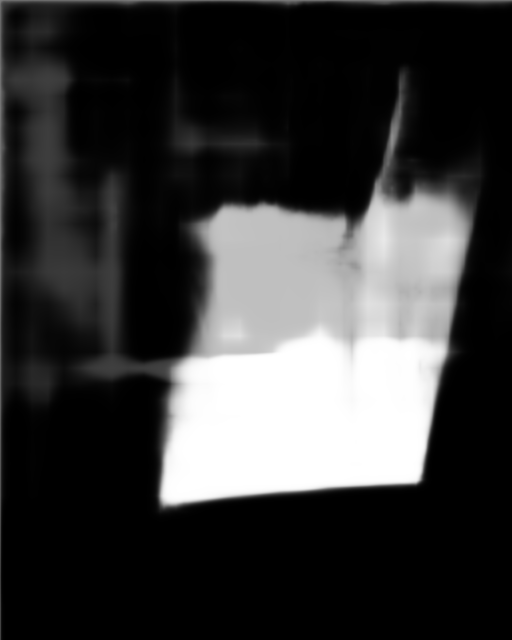}&
		\includegraphics[width=\wvisual, height=\hvisualtall]{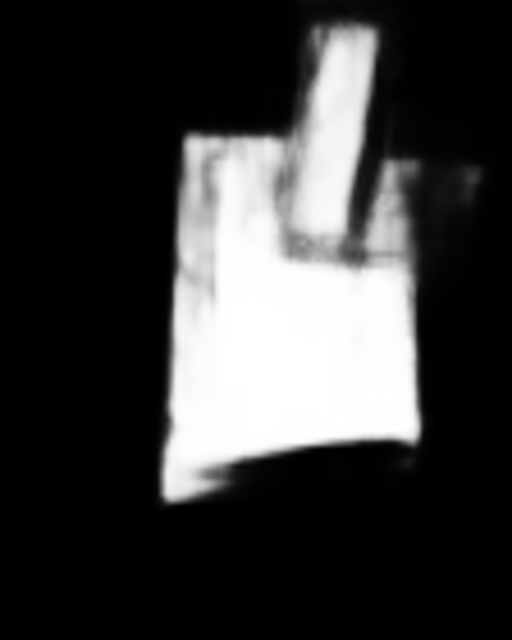}&
		\includegraphics[width=\wvisual, height=\hvisualtall]{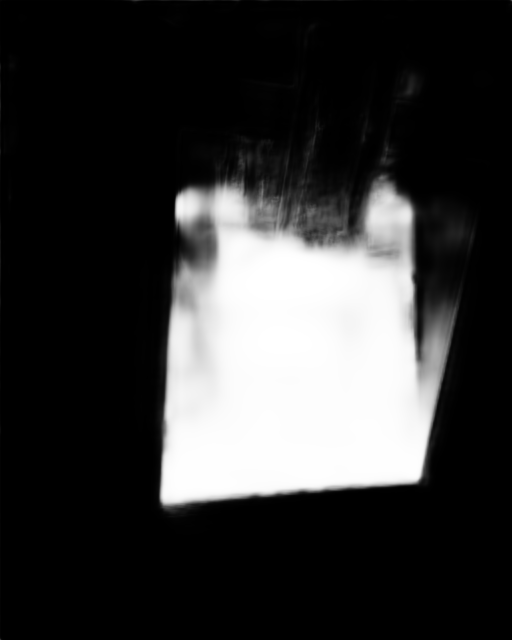}&
		\includegraphics[width=\wvisual, height=\hvisualtall]{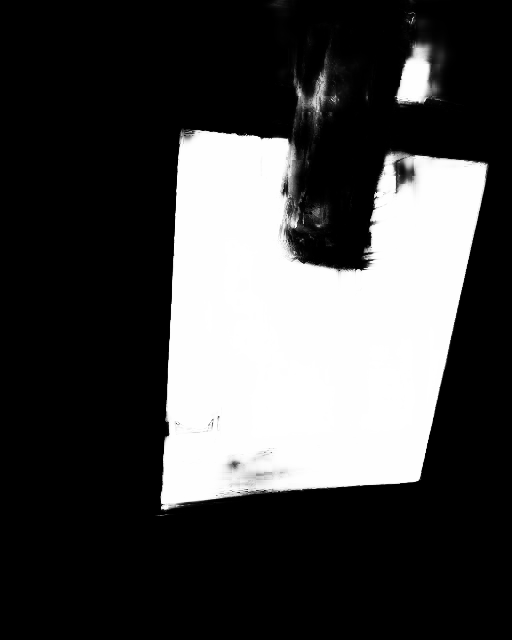}&
		\includegraphics[width=\wvisual, height=\hvisualtall]{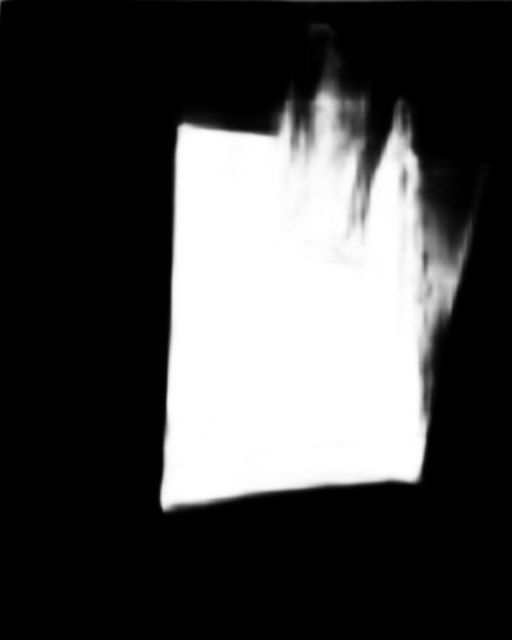}&
		\includegraphics[width=\wvisual, height=\hvisualtall]{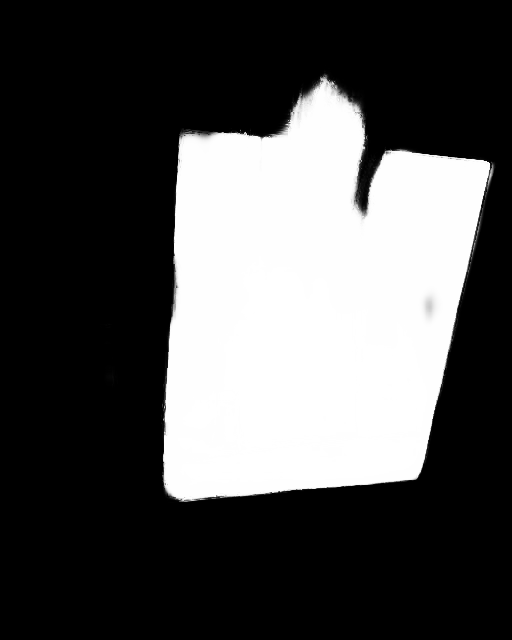}&
		\includegraphics[width=\wvisual, height=\hvisualtall]{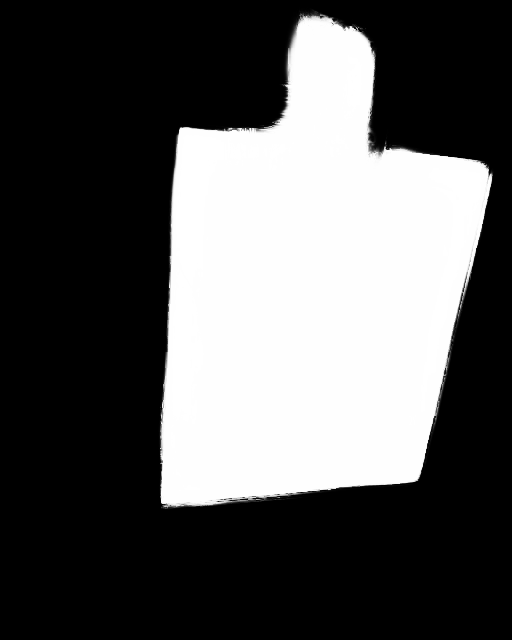}&
		\includegraphics[width=\wvisual, height=\hvisualtall]{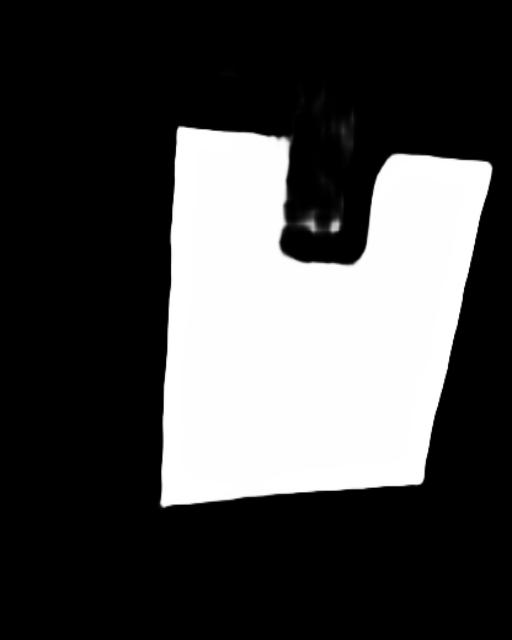}&
		\includegraphics[width=\wvisual, height=\hvisualtall]{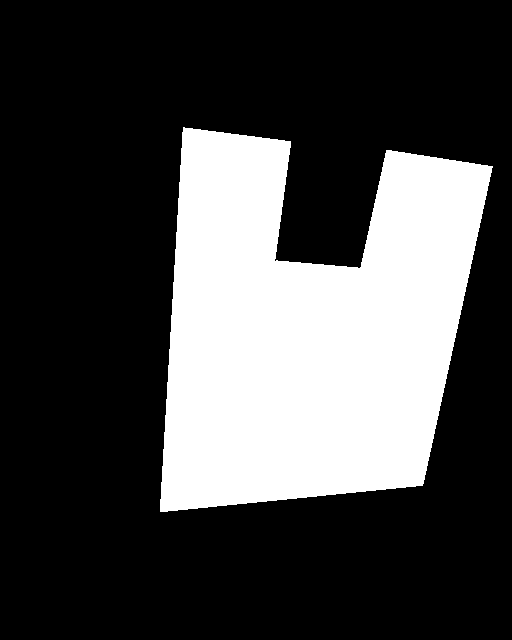} \\
		
		\includegraphics[width=\wvisual, height=\hvisualshort]{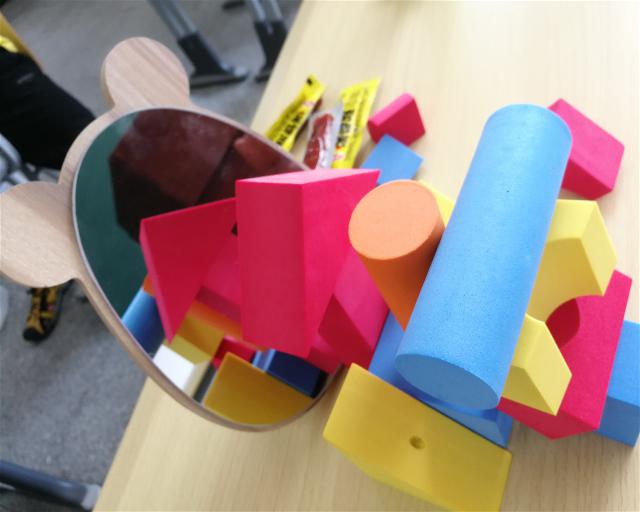}&
		\includegraphics[width=\wvisual, height=\hvisualshort]{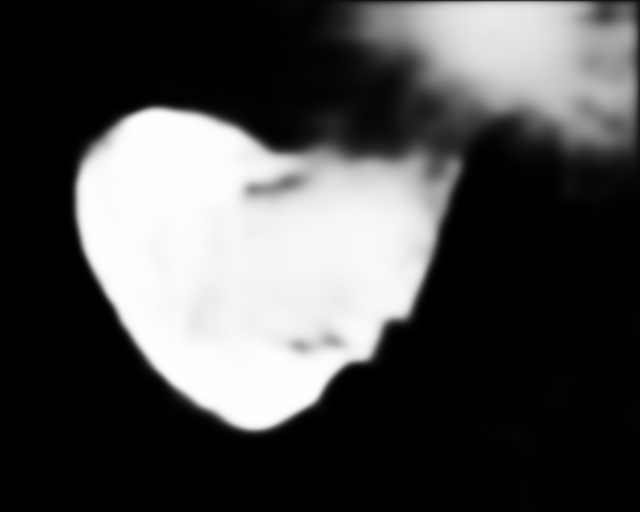}&
		\includegraphics[width=\wvisual, height=\hvisualshort]{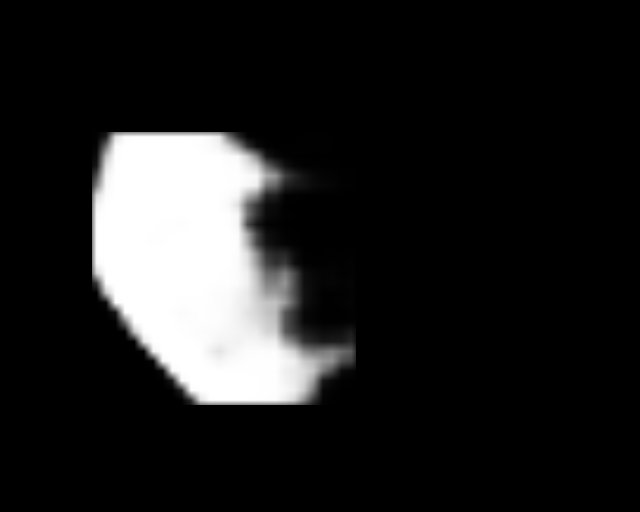}&
		\includegraphics[width=\wvisual, height=\hvisualshort]{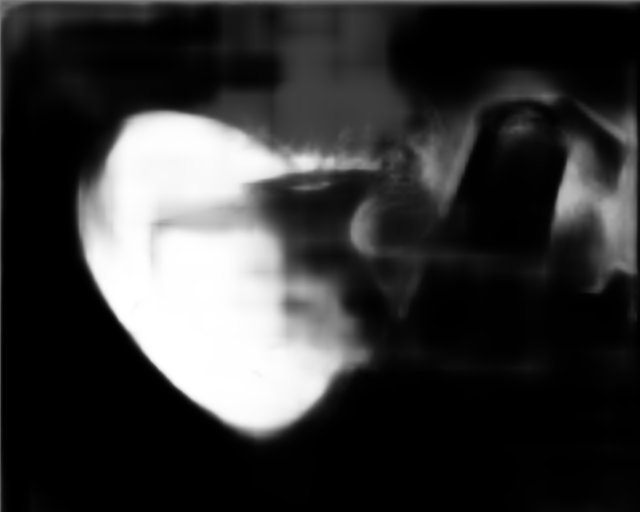}&
		\includegraphics[width=\wvisual, height=\hvisualshort]{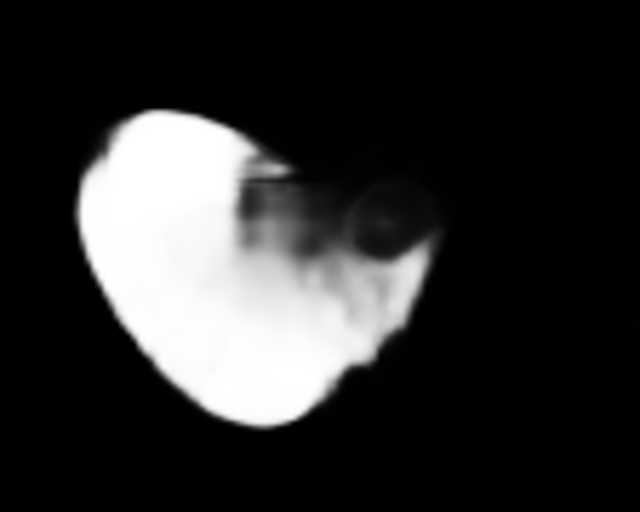}&
		\includegraphics[width=\wvisual, height=\hvisualshort]{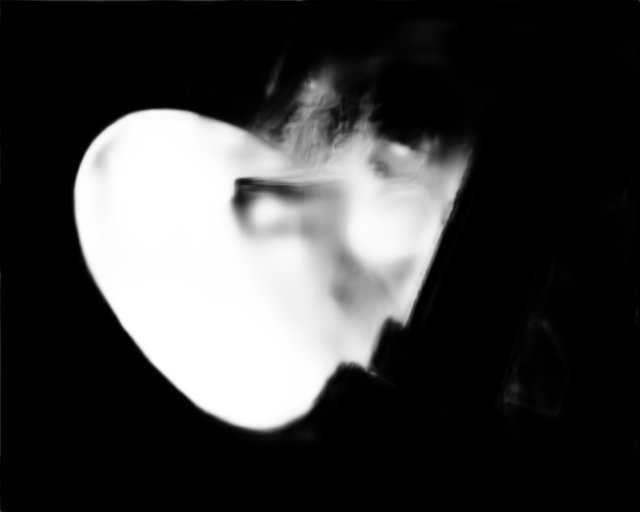}&
		\includegraphics[width=\wvisual, height=\hvisualshort]{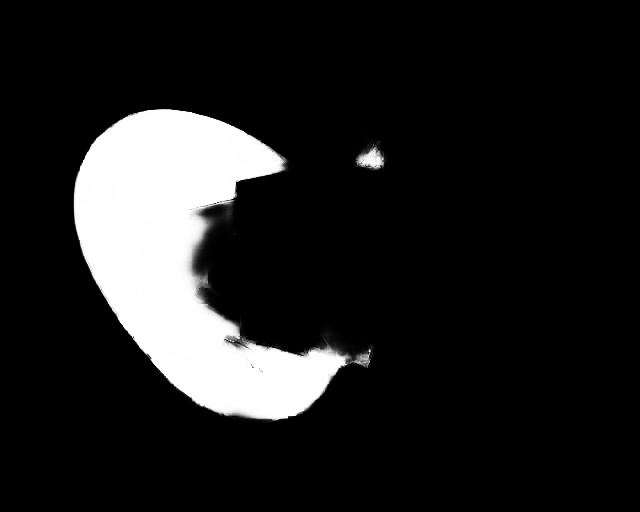}&
		\includegraphics[width=\wvisual, height=\hvisualshort]{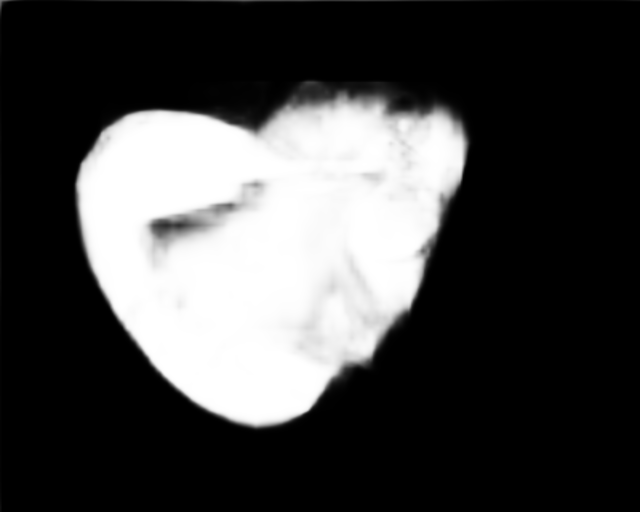}&
		\includegraphics[width=\wvisual, height=\hvisualshort]{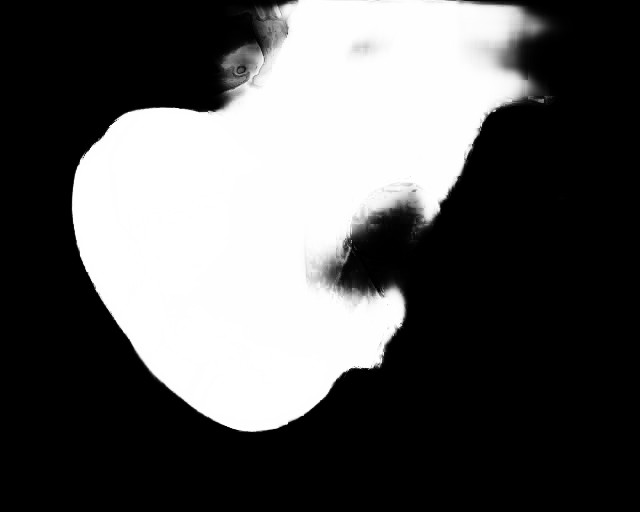}&
		\includegraphics[width=\wvisual, height=\hvisualshort]{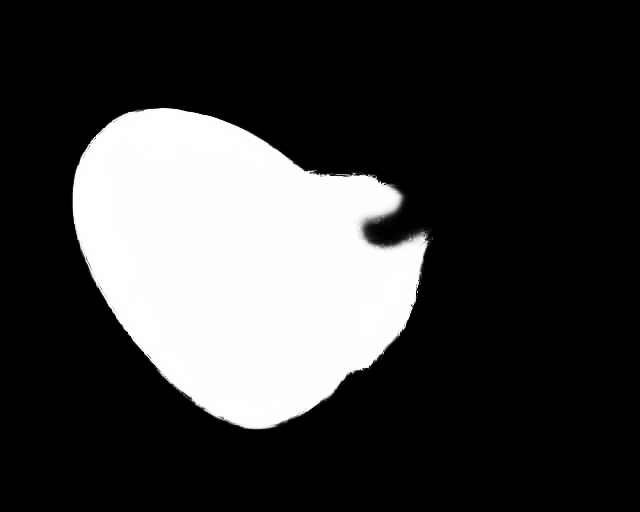}&
		\includegraphics[width=\wvisual, height=\hvisualshort]{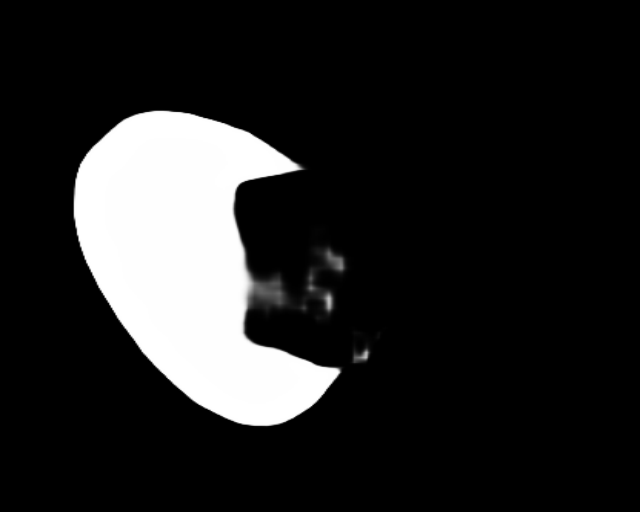}&
		\includegraphics[width=\wvisual, height=\hvisualshort]{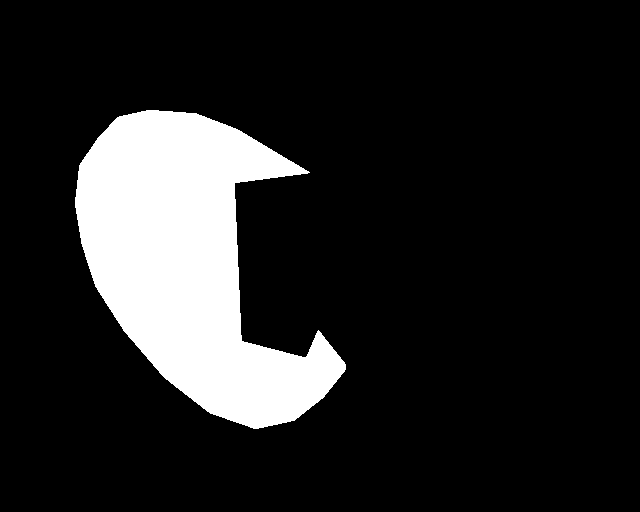} \\
		
		\includegraphics[width=\wvisual, height=\hvisualshort]{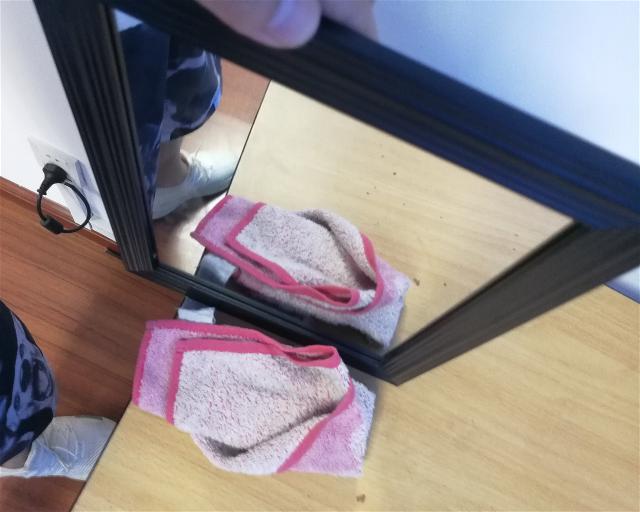}&
		\includegraphics[width=\wvisual, height=\hvisualshort]{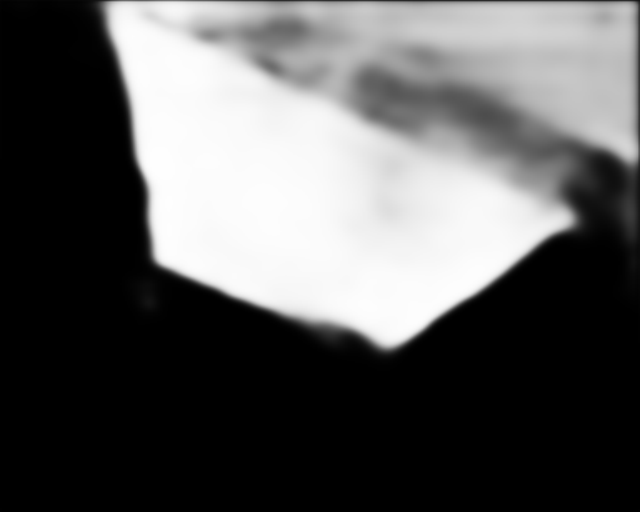}&
		\includegraphics[width=\wvisual, height=\hvisualshort]{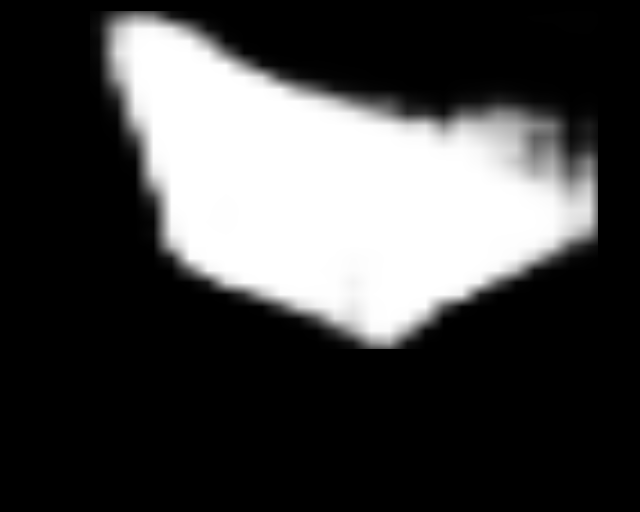}&
		\includegraphics[width=\wvisual, height=\hvisualshort]{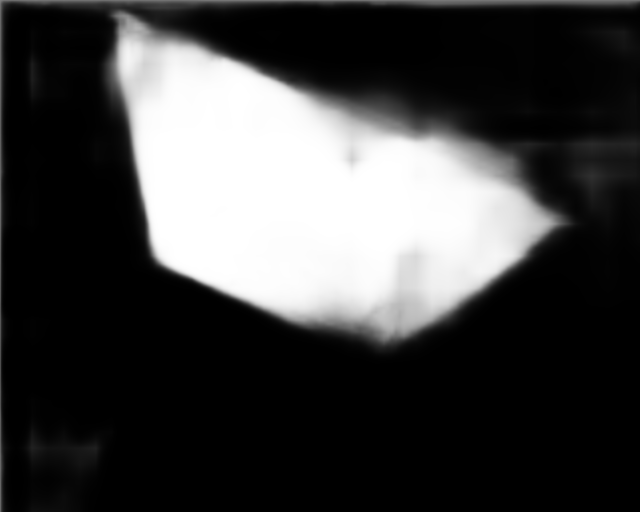}&
		\includegraphics[width=\wvisual, height=\hvisualshort]{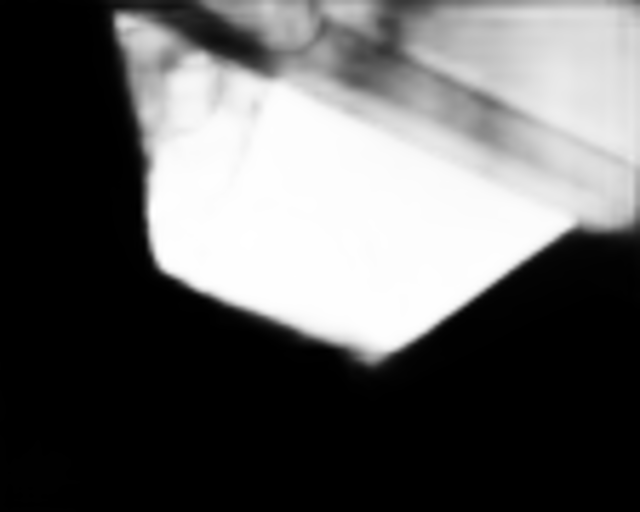}&
		\includegraphics[width=\wvisual, height=\hvisualshort]{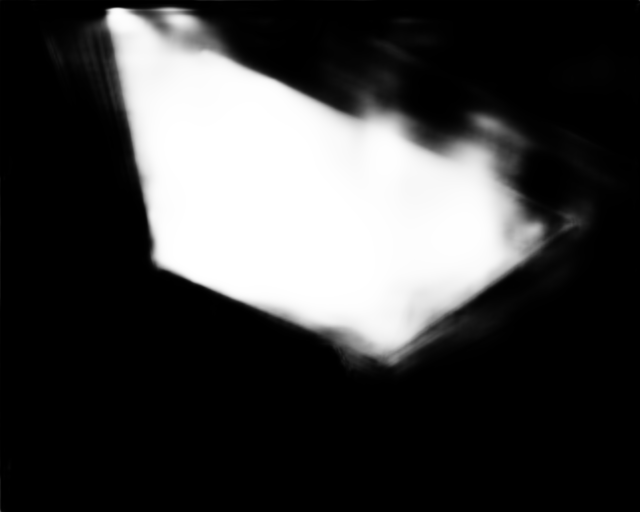}&
		\includegraphics[width=\wvisual, height=\hvisualshort]{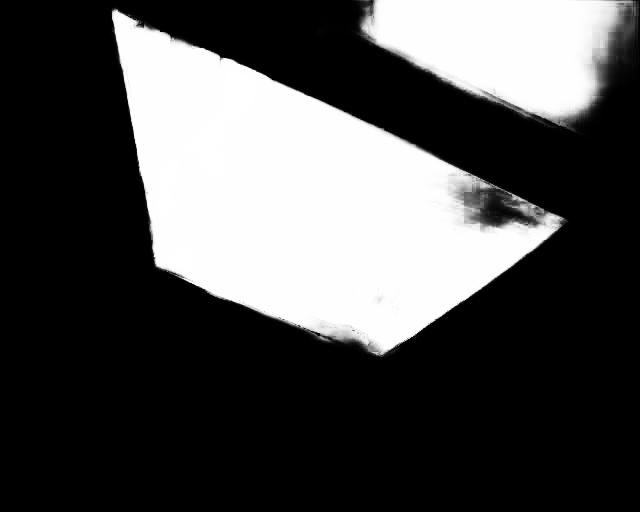}&
		\includegraphics[width=\wvisual, height=\hvisualshort]{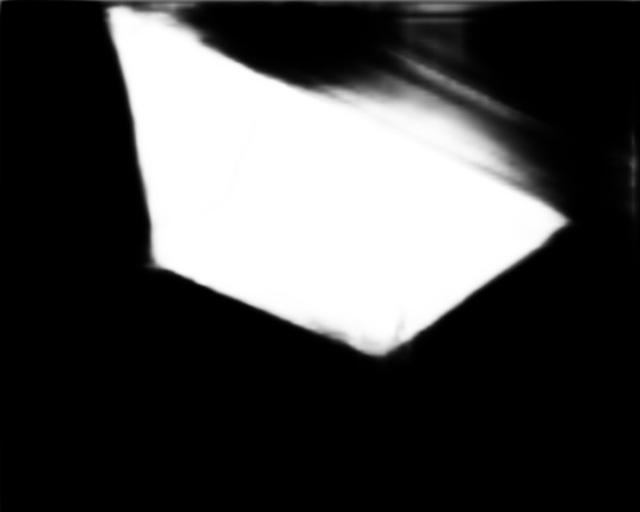}&
		\includegraphics[width=\wvisual, height=\hvisualshort]{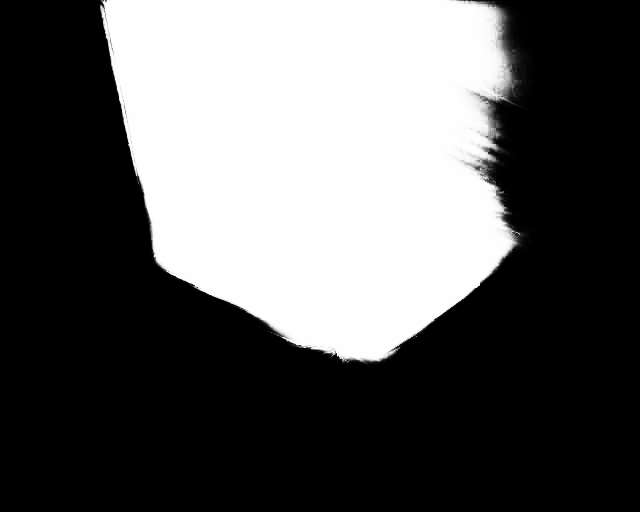}&
		\includegraphics[width=\wvisual, height=\hvisualshort]{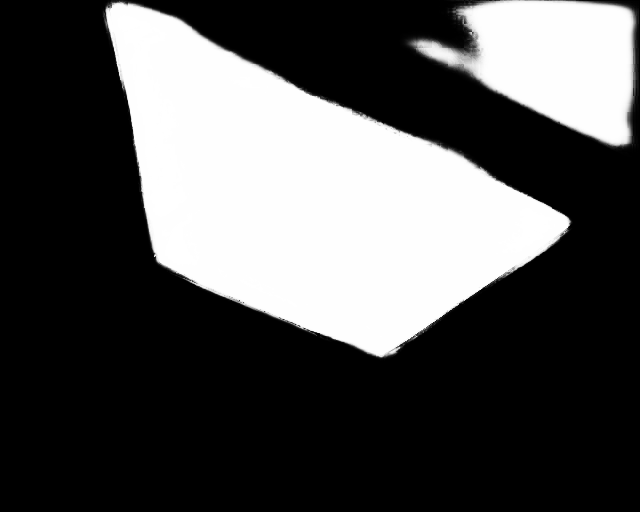}&
		\includegraphics[width=\wvisual, height=\hvisualshort]{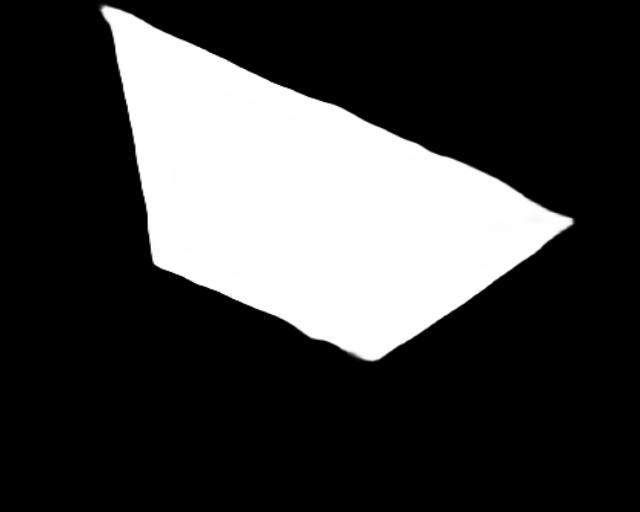}&
		\includegraphics[width=\wvisual, height=\hvisualshort]{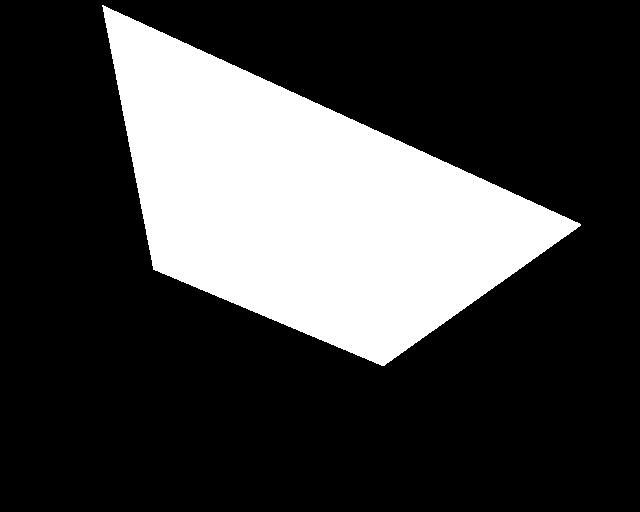} \\
		
		\includegraphics[width=\wvisual, height=\hvisualtall]{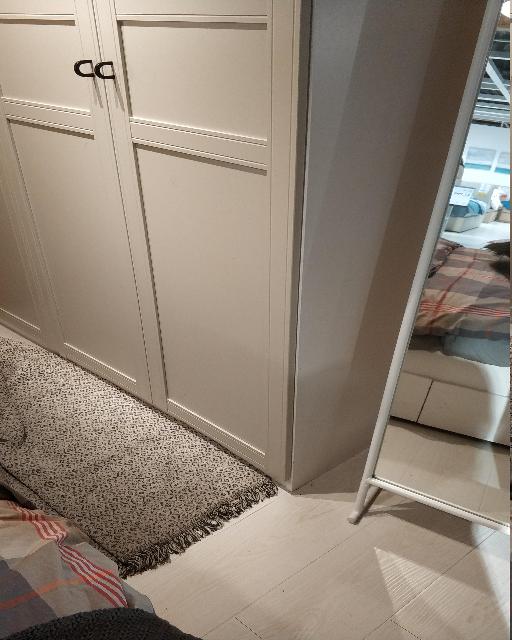}&
		\includegraphics[width=\wvisual, height=\hvisualtall]{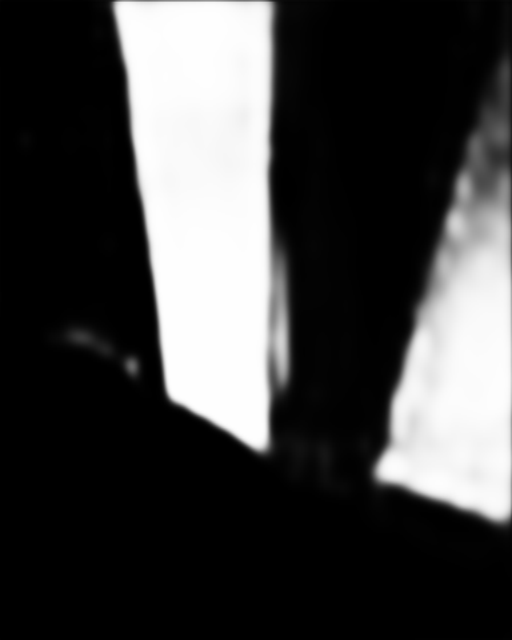}&
		\includegraphics[width=\wvisual, height=\hvisualtall]{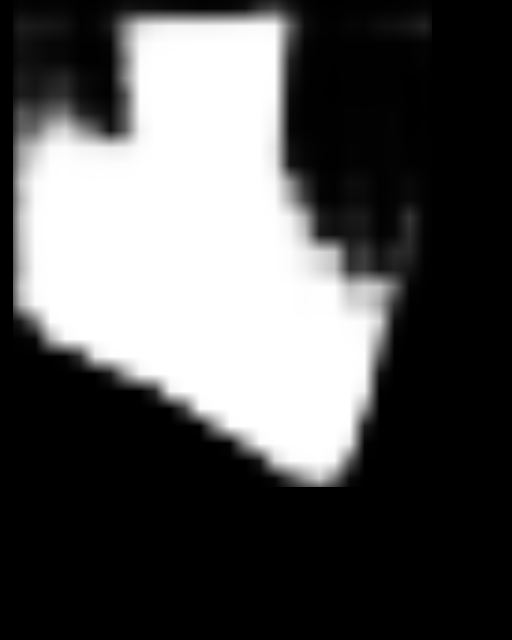}&
		\includegraphics[width=\wvisual, height=\hvisualtall]{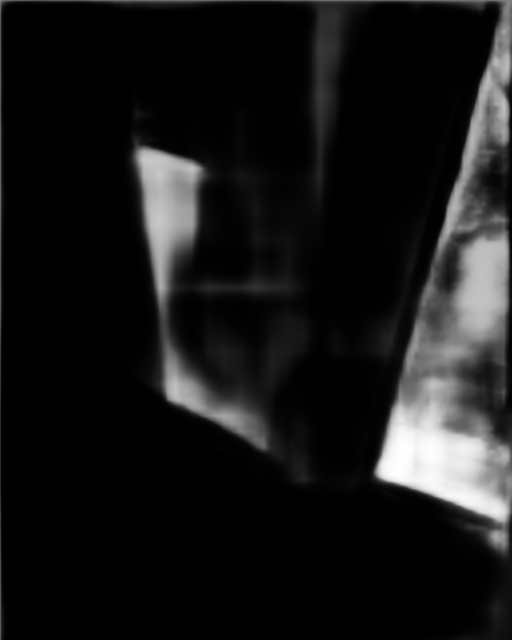}&
		\includegraphics[width=\wvisual, height=\hvisualtall]{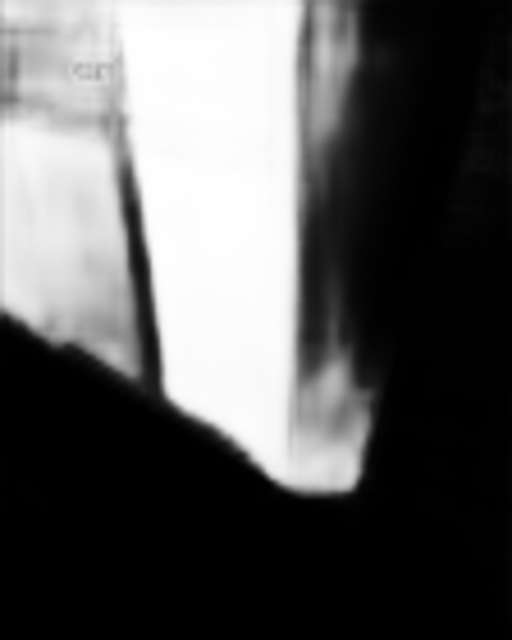}&
		\includegraphics[width=\wvisual, height=\hvisualtall]{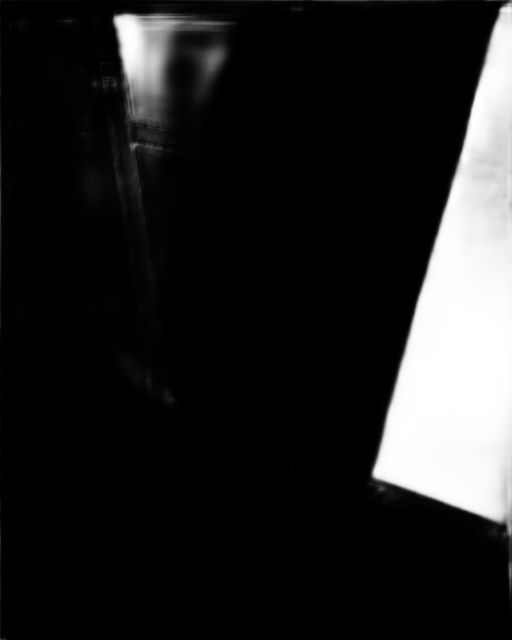}&
		\includegraphics[width=\wvisual, height=\hvisualtall]{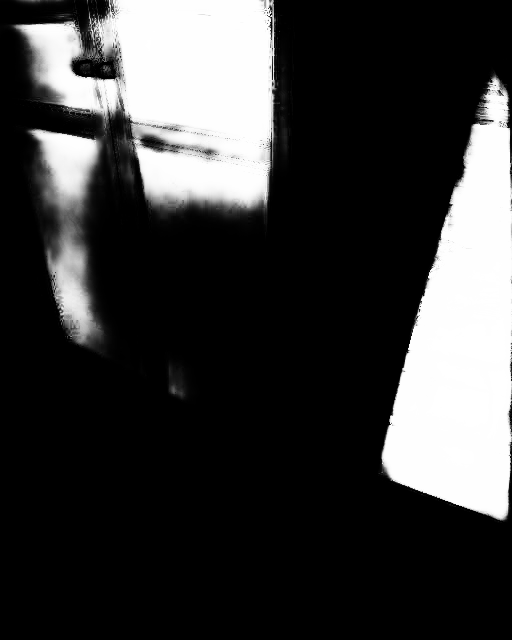}&
		\includegraphics[width=\wvisual, height=\hvisualtall]{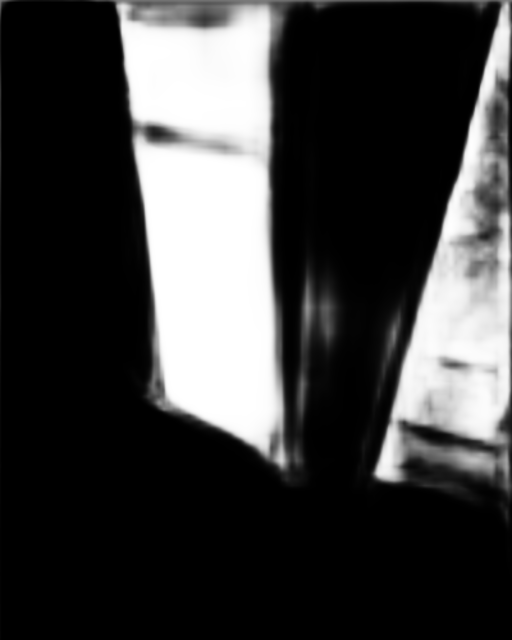}&
		\includegraphics[width=\wvisual, height=\hvisualtall]{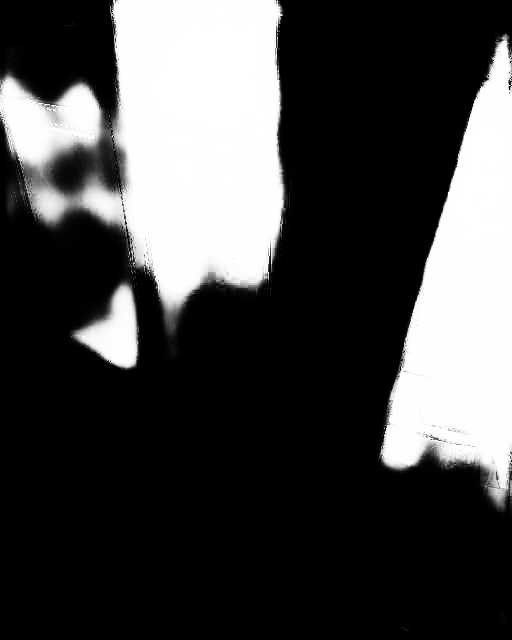}&
		\includegraphics[width=\wvisual, height=\hvisualtall]{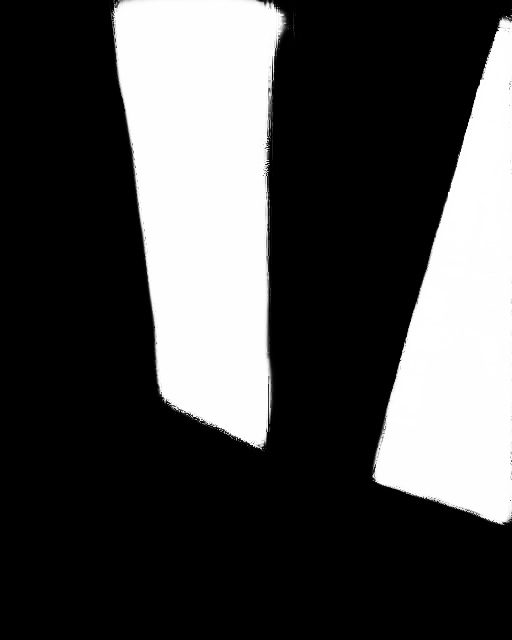}&
		\includegraphics[width=\wvisual, height=\hvisualtall]{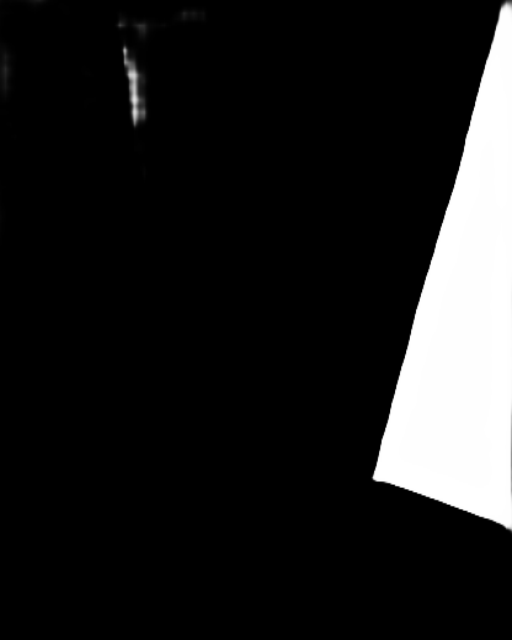}&
		\includegraphics[width=\wvisual, height=\hvisualtall]{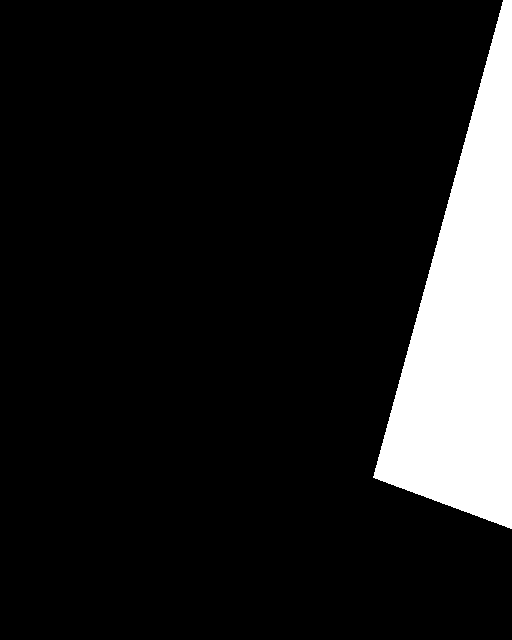} \\
		
		\includegraphics[width=\wvisual, height=\hvisualtall]{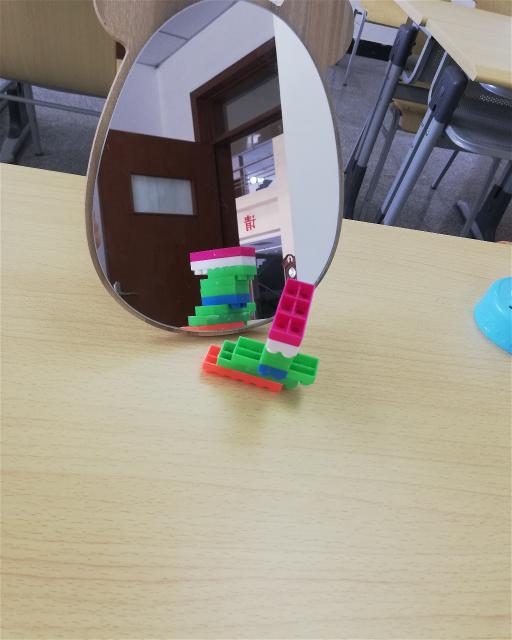}&
		\includegraphics[width=\wvisual, height=\hvisualtall]{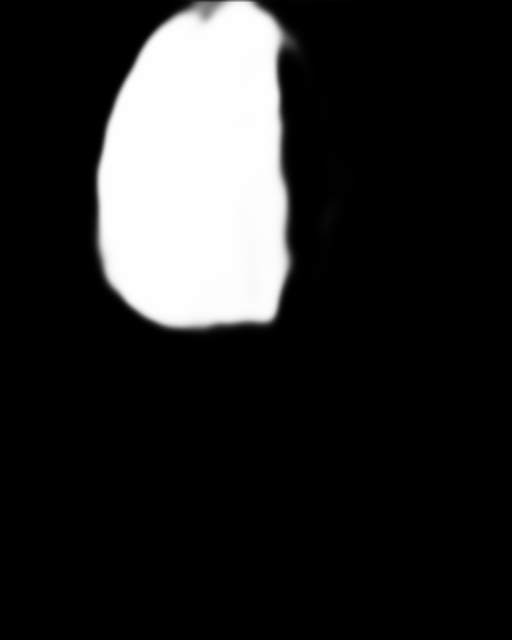}&
		\includegraphics[width=\wvisual, height=\hvisualtall]{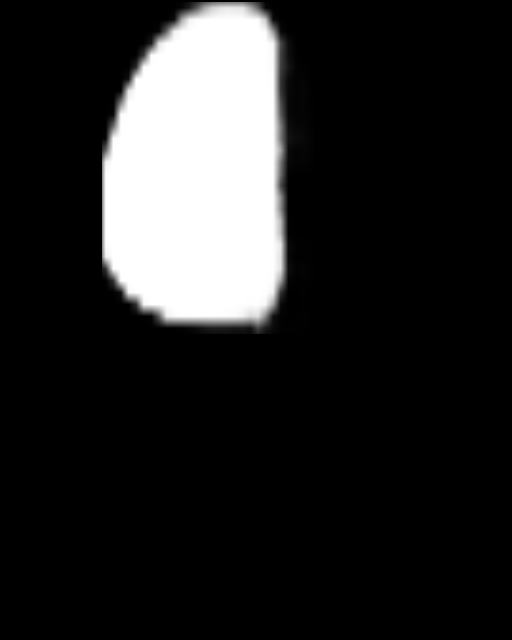}&
		\includegraphics[width=\wvisual, height=\hvisualtall]{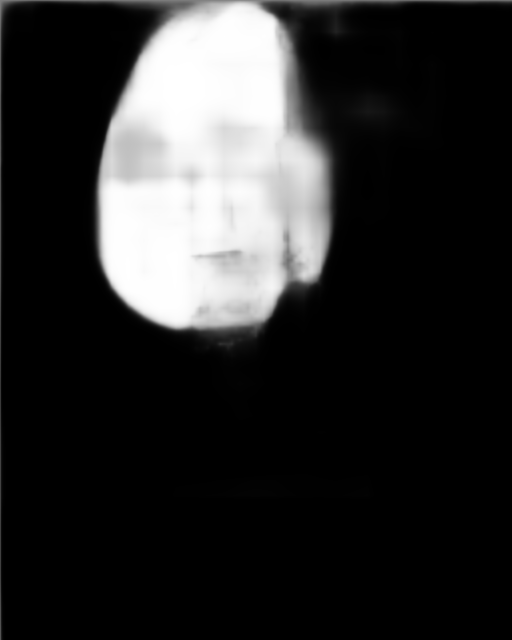}&
		\includegraphics[width=\wvisual, height=\hvisualtall]{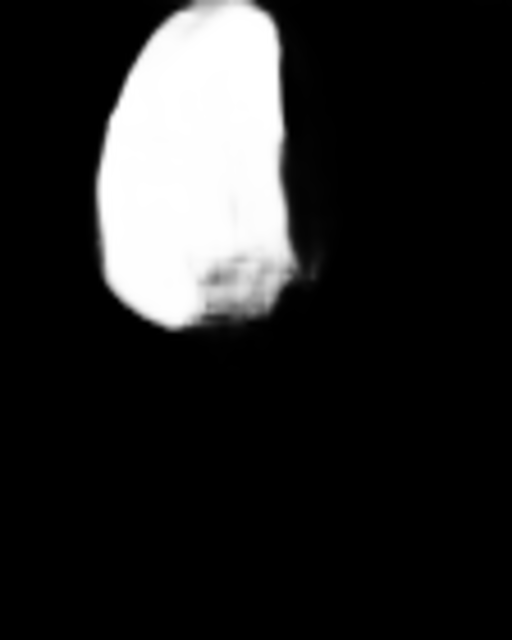}&
		\includegraphics[width=\wvisual, height=\hvisualtall]{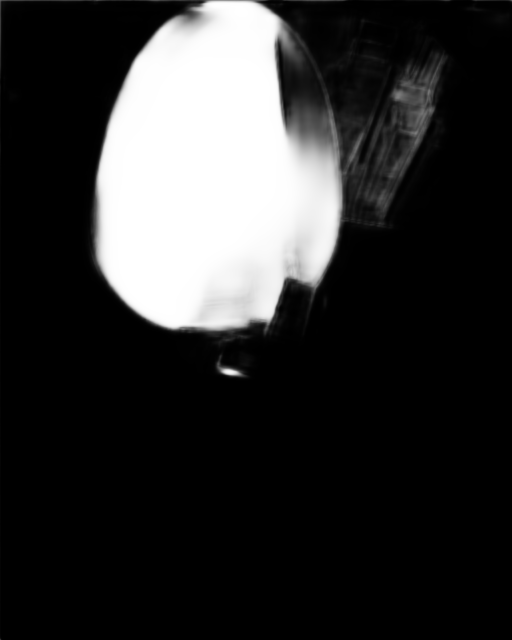}&
		\includegraphics[width=\wvisual, height=\hvisualtall]{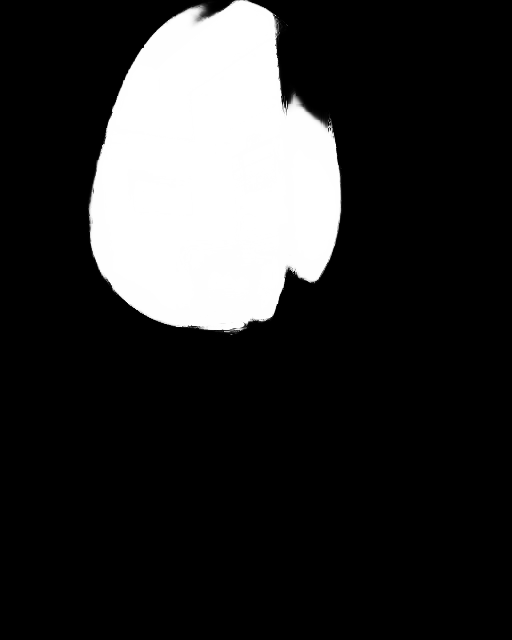}&
		\includegraphics[width=\wvisual, height=\hvisualtall]{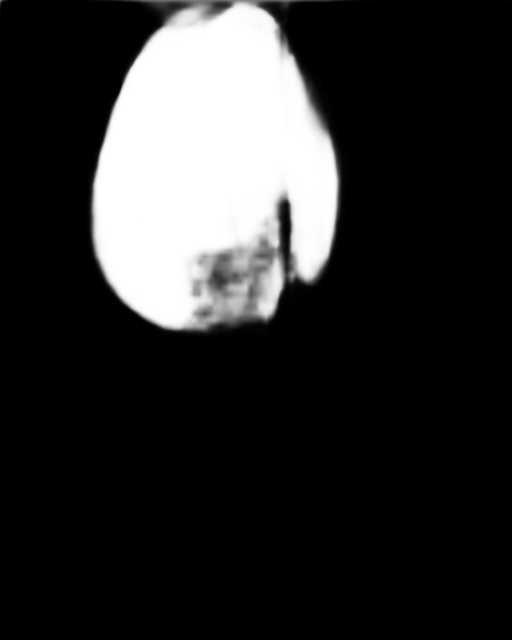}&
		\includegraphics[width=\wvisual, height=\hvisualtall]{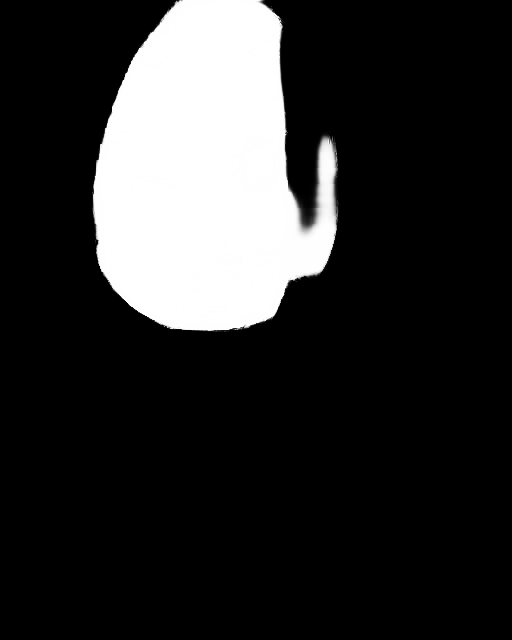}&
		\includegraphics[width=\wvisual, height=\hvisualtall]{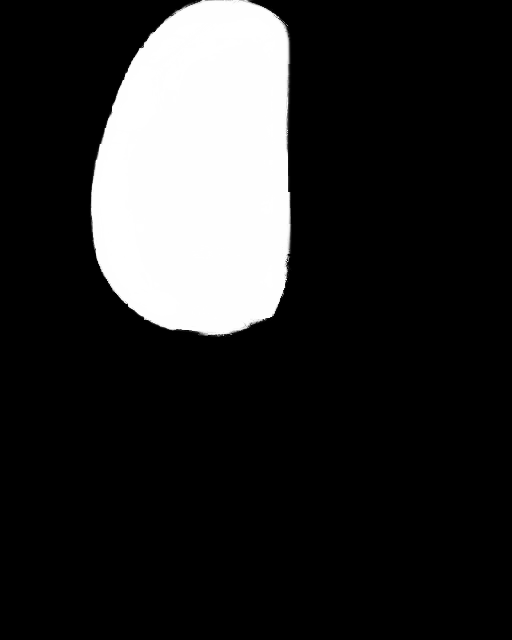}&
		\includegraphics[width=\wvisual, height=\hvisualtall]{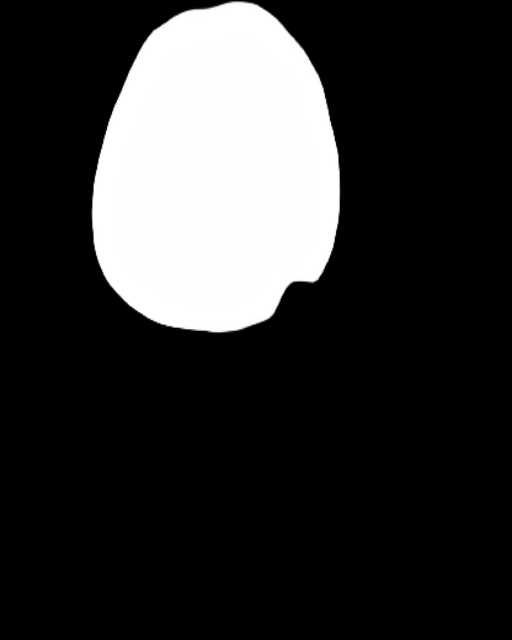}&
		\includegraphics[width=\wvisual, height=\hvisualtall]{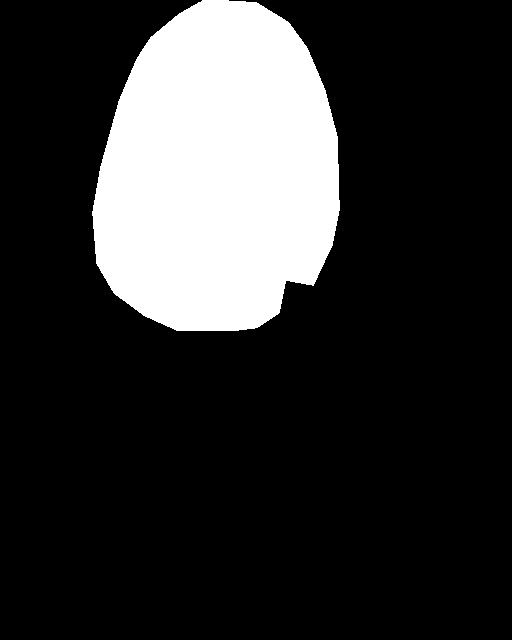} \\
		
		Image  & \tabincell{c}{PSPNet\\\cite{zhao2017pyramid}} & \tabincell{c}{Mask RCNN\\\cite{He_2017_ICCV}} & \tabincell{c}{DSS\\\cite{HouPami19Dss}} & \tabincell{c}{PiCANet\\\cite{liu2018picanet}} & \tabincell{c}{RAS\\\cite{Chen_2018_ECCV}} &
		\tabincell{c}{R\textsuperscript{3}Net\\\cite{deng2018r3net}} & \tabincell{c}{DSC\\\cite{Hu_2018_CVPR}} & \tabincell{c}{BDRAR\\\cite{Zhu_2018_ECCV}} & \tabincell{c}{MirrorNet\\\cite{yang2019mirrornet}} & \tabincell{c}{GDNet-B-M\\(ours)} & GT \\
	\end{tabular}
	\caption{\mhy{Visual comparison of GDNet-B-M to the state-of-the-art methods on the MSD \cite{yang2019mirrornet} testing set.}}
	\label{fig:mirror}
\end{figure*}

\section{Experiments on Saliency Detection}\label{sec:saliency}
\mhy{In addition to glass detection and mirror segmentation, contexts also play an important role in salient object detection (SOD). High-level contexts describe the relations between different parts/objects and thus are helpful for discovering the specific locations of salient objects while low-level contexts could provide fine detail information for delineating the salient object boundaries. Benefiting from the exploration of rich contexts in the large field-of-view, our network has the potential to handle the salient object detection task.}

\subsection{Experimental Settings}
\mhy{\textbf{Benchmark datasets.}}
\mhy{We conduct the experiments on five frequently used benchmark datasets: PASCAL-S \cite{Li_2014_CVPR_pascals}, DUT-OMRON \cite{Yang_2013_ICCV_Workshops}, ECSSD \cite{Yan_2013_ICCV_Workshops}, HKU-IS \cite{Li_2015_CVPR_hkuis}, and DUTS \cite{Wang_2017_CVPR_duts}. PASCAL-S \cite{Li_2014_CVPR_pascals} includes 850 images with cluttered backgrounds and complex foreground objects. DUT-OMRON \cite{Yang_2013_ICCV_Workshops} has 5,168 images with one or two objects in each image. Most of the foreground objects are structurally complex. ECSSD \cite{Yan_2013_ICCV_Workshops} consists of 1,000 images with complex and semantically meaningful objects. HKU-IS \cite{Li_2015_CVPR_hkuis} contains 4,447 images, which usually contain multiple disconnected salient objects or salient objects that overlap with the image boundaries. DUTS \cite{Wang_2017_CVPR_duts} is currently the largest salient object detection dataset, which are divided into 10,553 training images (DUTS-TR) and 5,019 testing images (DUTS-TE). We follow previous works \cite{Qin_2019_CVPR_basnet,Zhao_2019_CVPR_pfan,Wu_2019_CVPR_cpd,Su_2019_ICCV_banet,Zhao_2019_ICCV_egnet,mei2021exploring} to use DUTS-TR as the training set and others as testing sets.}

\begin{table*}[tbp]
	\caption{\mhy{Comparison of the proposed method on the saliency detection task with 17 state-of-the-art methods on five benchmark datasets in terms of the structure-measure $S_\alpha$ (larger is better), the adaptive E-measure $E_\phi^{a}$ (larger is better), the weighted F-measure $F_\beta^w$ (larger is better), and the mean absolute error $M$ (smaller is better). The first, second, and third best results are marked in \textcolor[rgb]{1, 0, 0}{\textbf{red}}, \textcolor[rgb]{0, 1, 0}{\textbf{green}}, and \textcolor[rgb]{0, 0, 1}{\textbf{blue}}, respectively. ``$^*$'' means that the results are achieved with post-processing by CRF \cite{krahenbuhl2011efficient}. Our method achieves the state-of-the-art on five benchmark datasets under all four evaluation metrics.}}
	\centering
	\setlength{\tabcolsep}{2.7pt}
	\begin{tabular}{l|c|cccc|cccc|cccc|cccc|cccc}
		\hline
		\hline
		\multirow{3}{*}{Method} & \multirow{3}{*}{Pub.'Year} & \multicolumn{4}{c|}{PASCAL-S \cite{Li_2014_CVPR_pascals}} & \multicolumn{4}{c|}{DUT-OMRON \cite{Yang_2013_ICCV_Workshops}} & \multicolumn{4}{c|}{ECSSD \cite{Yan_2013_ICCV_Workshops}} & \multicolumn{4}{c|}{HKU-IS \cite{Li_2015_CVPR_hkuis}} & \multicolumn{4}{c}{DUTS-TE \cite{Wang_2017_CVPR_duts}}\\
		& & \multicolumn{4}{c|}{850 images} & \multicolumn{4}{c|}{5,016 images} & \multicolumn{4}{c|}{1,000 images} & \multicolumn{4}{c|}{4,447 images} & \multicolumn{4}{c}{5,168 images} \\
		\cline{3-22}
		& & $S_\alpha$$\uparrow$ & $E_\phi^{a}$$\uparrow$ & $F_\beta^w$$\uparrow$ & M$\downarrow$ & $S_\alpha$$\uparrow$ & $E_\phi^{a}$$\uparrow$ & $F_\beta^w$$\uparrow$ & M$\downarrow$ & $S_\alpha$$\uparrow$ & $E_\phi^{a}$$\uparrow$ & $F_\beta^w$$\uparrow$ & M$\downarrow$ & $S_\alpha$$\uparrow$ & $E_\phi^{a}$$\uparrow$ & $F_\beta^w$$\uparrow$ & M$\downarrow$ & $S_\alpha$$\uparrow$ & $E_\phi^{a}$$\uparrow$ & $F_\beta^w$$\uparrow$ & M$\downarrow$  \\
		
		\hline
		\hline
		
		C2SNet \cite{Li_2018_ECCV_c2s}             & ECCV'18 & .830 & .837 & .756 & .085 & .782 & .817 & .643 & .079 & .886 & .909 & .844 & .057 & .878 & .923 & .824 & .049 & .820 & .842 & .705 & .065    \\
		RAS \cite{Chen_2018_ECCV_ras}              & ECCV'18 & .801 & .840 & .735 & .100 & .812 & .843 & .695 & .063 & .894 & .916 & .860 & .055 & .889 & .932 & .849 & .045 & .839 & .861 & .740 & .059    \\
		PAGRN \cite{Zhang_2018_CVPR_progressive}            & CVPR'18 & .822 & .853 & .732 & .089 & .775 & .842 & .622 & .071 & .889 & .914 & .833 & .061 & .887 & .939 & .820 & .048 & .838 & .880 & .724 & .055   \\
		DGRL \cite{Wang_2018_CVPR_dgrl}             & CVPR'18 & .836 & .848 & .791 & .072 & .806 & .848 & .709 & .062 & .903 & .918 & .891 & .041 & .895 & .944 & .876 & .035 & .842 & .879 & .774 & .050   \\
		R\textsuperscript{3}Net* \cite{deng2018r3net}           & IJCAI'18 & .807 & .842 & .755 & .092 & .817 & .851 & .728 & .063 & .910 & \textcolor[rgb]{1,0,0}{\textbf{.929}} & .902 & .040 & .900 & .943 & .884 & .034 & .835 & .868 & .765 & .057   \\
		BMPM \cite{Zhang_2018_CVPR_bmpm}             & CVPR'18 & .845 & .843 & .774 & .074 & .809 & .838 & .681 & .064 & .911 & .914 & .871 & .045 & .907 & .937 & .859 & .039 & .862 & .860 & .761 & .049   \\
		PiCANet-R \cite{Liu_2018_CVPR_picanet}           & CVPR'18 & .854 & .838 & .772 & .076 & .832 & .841 & .695 & .065 & .917 & .913 & .867 & .046 & .904 & .936 & .840 & .043 & .869 & .862 & .755 & .051   \\
		DSS* \cite{HouPami19Dss}            & TPAMI'19 & .799 & .848 & .754 & .094 & .788 & .844 & .690 & .066 & .882 & .917 & .871 & .053 & .881 & .936 & .866 & .040 & .824 & .882 & .755 & .056   \\
		BASNet \cite{Qin_2019_CVPR_basnet}          & CVPR'19 & .838 & .853 & .793 & .076 & .836 & \textcolor[rgb]{0,0,1}{\textbf{.869}} & \textcolor[rgb]{0,1,0}{\textbf{.751}} & \textcolor[rgb]{0,0,1}{\textbf{.056}} & .916 & .921 & .904 & .037 & .909 & .946 & .889 & .032 & .866 & .884 & .803 & .048   \\
		CPD \cite{Wu_2019_CVPR_cpd}             & CVPR'19 & .848 & .855 & .794 & .071 & .825 & .866 & .719 & \textcolor[rgb]{0,0,1}{\textbf{.056}} & .918 & .925 & .898 & .037 & .909 & .946 & .879 & .033 & .869 & .887 & .795 & .043   \\
		PAGE-Net \cite{Wang_2019_CVPR_pagenet}        & CVPR'19 & .842 & .847 & .783 & .076 & .824 & .853 & .722 & .062 & .912 & .920 & .886 & .042 & .903 & .940 & .865 & .037 & .854 & .869 & .769 & .052   \\
		AFNet \cite{Feng_2019_CVPR_afnet}           & CVPR'19 & .850 & .851 & .792 & .070 & .826 & .853 & .717 & .057 & .914 & .918 & .887 & .042 & .905 & .942 & .870 & .036 & .867 & .879 & .785 & .046   \\
		BANet \cite{Su_2019_ICCV_banet}           & ICCV'19 & .852 & \textcolor[rgb]{0,0,1}{\textbf{.859}} & .802 & .070 & .832 & .860 & .736 & .059 & .924 & \textcolor[rgb]{0,1,0}{\textbf{.928}} & .908 & .035 & .913 & .950 & .886 & .032 & .879 & .892 & .811 & .040   \\
		
		GCPANet \cite{chen2020global}          & AAAI'20 & \textcolor[rgb]{1,0,0}{\textbf{.864}} & .853 & .808 & \textcolor[rgb]{0,0,1}{\textbf{.062}} & \textcolor[rgb]{0,0,1}{\textbf{.839}} & .860 & .734 & \textcolor[rgb]{0,0,1}{\textbf{.056}} & \textcolor[rgb]{1,0,0}{\textbf{.927}} & .920 & .903 & .035 & \textcolor[rgb]{0,1,0}{\textbf{.920}} & .949 & .889 & .031 & \textcolor[rgb]{0,1,0}{\textbf{.891}} & .891 & .821 & .038   \\
		F3Net \cite{wei2020f3net}          & AAAI'20 & \textcolor[rgb]{0,1,0}{\textbf{.861}} & \textcolor[rgb]{0,1,0}{\textbf{.865}} & \textcolor[rgb]{0,1,0}{\textbf{.816}} & \textcolor[rgb]{0,1,0}{\textbf{.061}} & .838 & \textcolor[rgb]{0,1,0}{\textbf{.870}} & .747 & \textcolor[rgb]{0,1,0}{\textbf{.053}} & .924 & \textcolor[rgb]{0,0,1}{\textbf{.927}} & \textcolor[rgb]{0,1,0}{\textbf{.912}} & \textcolor[rgb]{0,1,0}{\textbf{.033}} & \textcolor[rgb]{0,1,0}{\textbf{.920}} & \textcolor[rgb]{0,1,0}{\textbf{.955}} & \textcolor[rgb]{0,1,0}{\textbf{.905}} & \textcolor[rgb]{0,1,0}{\textbf{.027}} & \textcolor[rgb]{0,0,1}{\textbf{.888}} & \textcolor[rgb]{0,1,0}{\textbf{.902}} & \textcolor[rgb]{0,1,0}{\textbf{.835}} & \textcolor[rgb]{0,1,0}{\textbf{.035}}   \\
		MINet-R \cite{Pang_2020_CVPR_minet}           & CVPR'20 & .858 & \textcolor[rgb]{0,0,1}{\textbf{.859}} & .810 & .064 & .834 & .866 & .737 & \textcolor[rgb]{0,0,1}{\textbf{.056}} & \textcolor[rgb]{0,1,0}{\textbf{.926}} & \textcolor[rgb]{0,1,0}{\textbf{.928}} & \textcolor[rgb]{0,0,1}{\textbf{.910}} & \textcolor[rgb]{0,0,1}{\textbf{.034}} & \textcolor[rgb]{0,1,0}{\textbf{.920}} & \textcolor[rgb]{0,0,1}{\textbf{.954}} & \textcolor[rgb]{0,0,1}{\textbf{.896}} & \textcolor[rgb]{0,0,1}{\textbf{.029}} & .885 & \textcolor[rgb]{0,0,1}{\textbf{.899}} & \textcolor[rgb]{0,0,1}{\textbf{.824}} & \textcolor[rgb]{0,0,1}{\textbf{.037}}   \\
		ITSD \cite{zhou2020interactive}           & CVPR'20 & \textcolor[rgb]{0,0,1}{\textbf{.859}} & .856 & \textcolor[rgb]{0,0,1}{\textbf{.812}} & .066 & \textcolor[rgb]{0,1,0}{\textbf{.840}} & .863 & \textcolor[rgb]{0,0,1}{\textbf{.750}} & .061 & \textcolor[rgb]{0,0,1}{\textbf{.925}} & \textcolor[rgb]{0,0,1}{\textbf{.927}} & \textcolor[rgb]{0,0,1}{\textbf{.910}} & \textcolor[rgb]{0,0,1}{\textbf{.034}} & \textcolor[rgb]{0,0,1}{\textbf{.917}} & .952 & .894 & .031 & .885 & .895 & \textcolor[rgb]{0,0,1}{\textbf{.824}} & .041   \\

		\hline
		\hline
		
		\textbf{GDNet-B-S} & \textbf{Ours} & \textcolor[rgb]{1,0,0}{\textbf{.864}} & \textcolor[rgb]{1,0,0}{\textbf{.872}} & \textcolor[rgb]{1,0,0}{\textbf{.826}} & \textcolor[rgb]{1,0,0}{\textbf{.058}} & \textcolor[rgb]{1,0,0}{\textbf{.848}} & \textcolor[rgb]{1,0,0}{\textbf{.879}} & \textcolor[rgb]{1,0,0}{\textbf{.769}} & \textcolor[rgb]{1,0,0}{\textbf{.047}} & \textcolor[rgb]{1,0,0}{\textbf{.927}} & \textcolor[rgb]{0,0,1}{\textbf{.927}} & \textcolor[rgb]{1,0,0}{\textbf{.920}} & \textcolor[rgb]{1,0,0}{\textbf{.031}} & \textcolor[rgb]{1,0,0}{\textbf{.923}} & \textcolor[rgb]{1,0,0}{\textbf{.958}} & \textcolor[rgb]{1,0,0}{\textbf{.911}} & \textcolor[rgb]{1,0,0}{\textbf{.026}} & \textcolor[rgb]{1,0,0}{\textbf{.892}} & \textcolor[rgb]{1,0,0}{\textbf{.913}} & \textcolor[rgb]{1,0,0}{\textbf{.850}} & \textcolor[rgb]{1,0,0}{\textbf{.032}}  \\
		
		\hline
		\hline
	\end{tabular}

	\label{tab:sod}
\end{table*}

\def\wprf{0.2\linewidth}
\def\hprf{1.3in}
\begin{figure*}[tbp]
	\setlength{\tabcolsep}{.2pt}
	\centering
	\begin{tabular}{ccccc}
		
		\includegraphics[width=\wprf, height=\hprf]{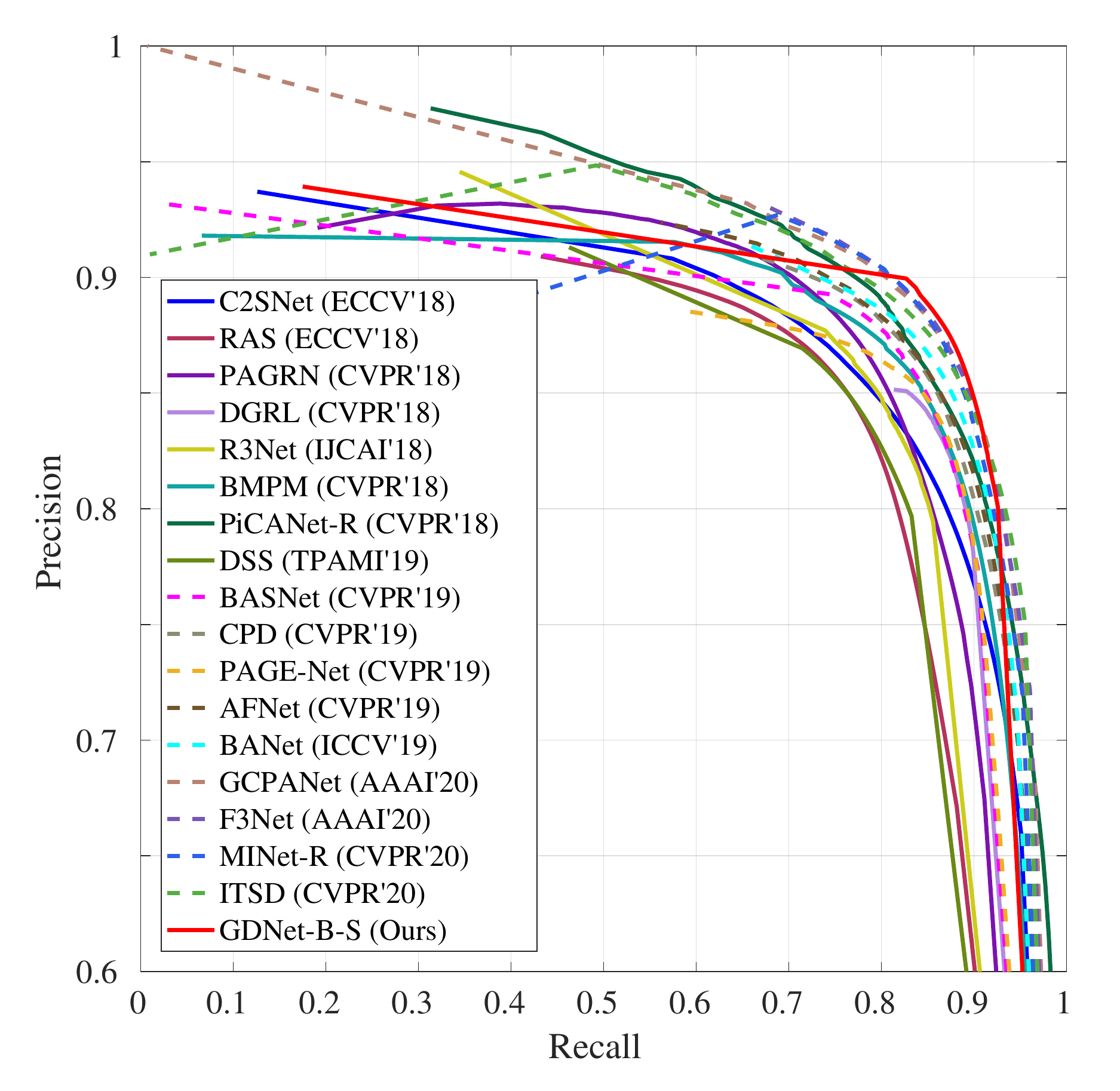} &
		\includegraphics[width=\wprf, height=\hprf]{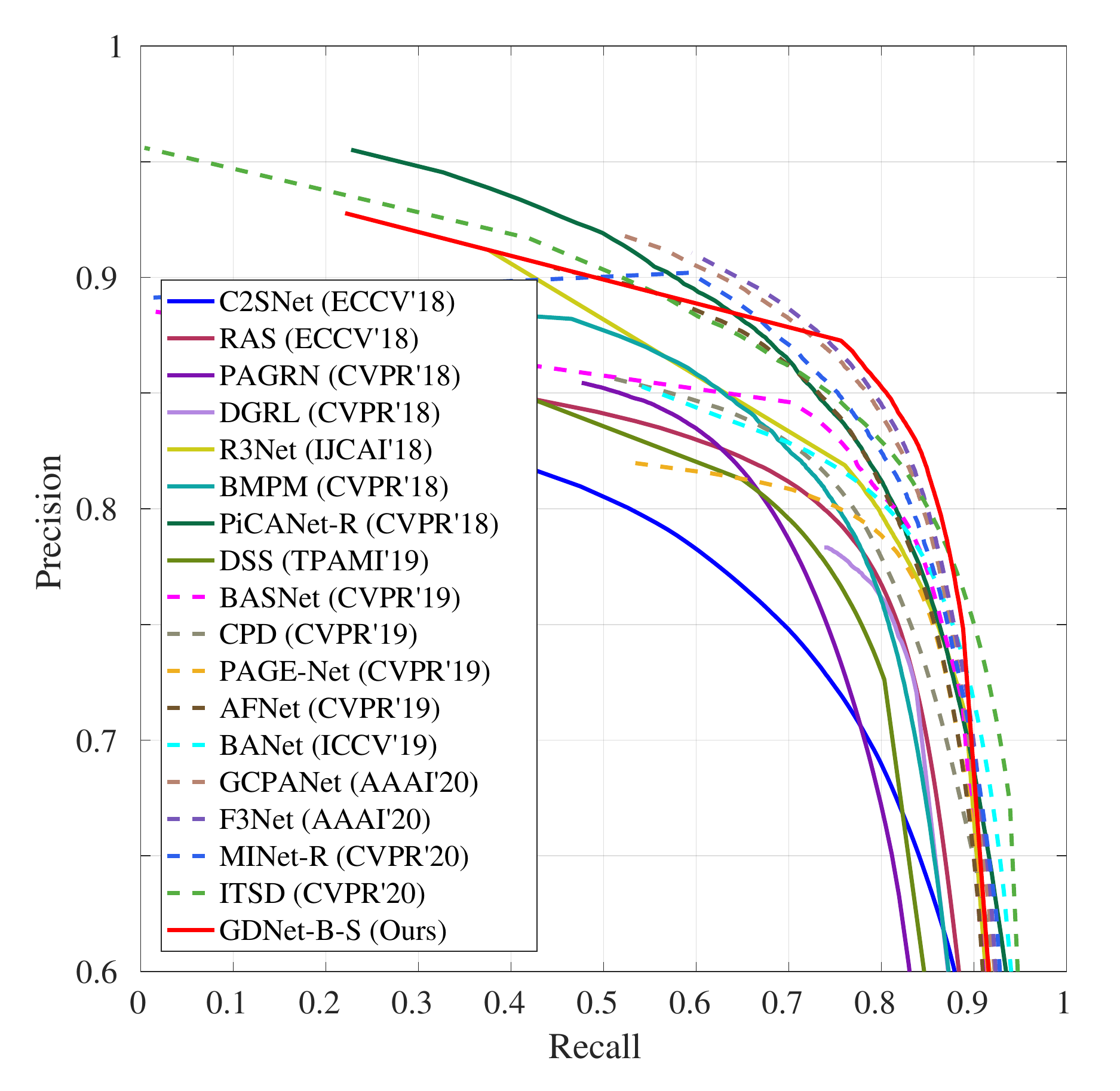} &
		\includegraphics[width=\wprf, height=\hprf]{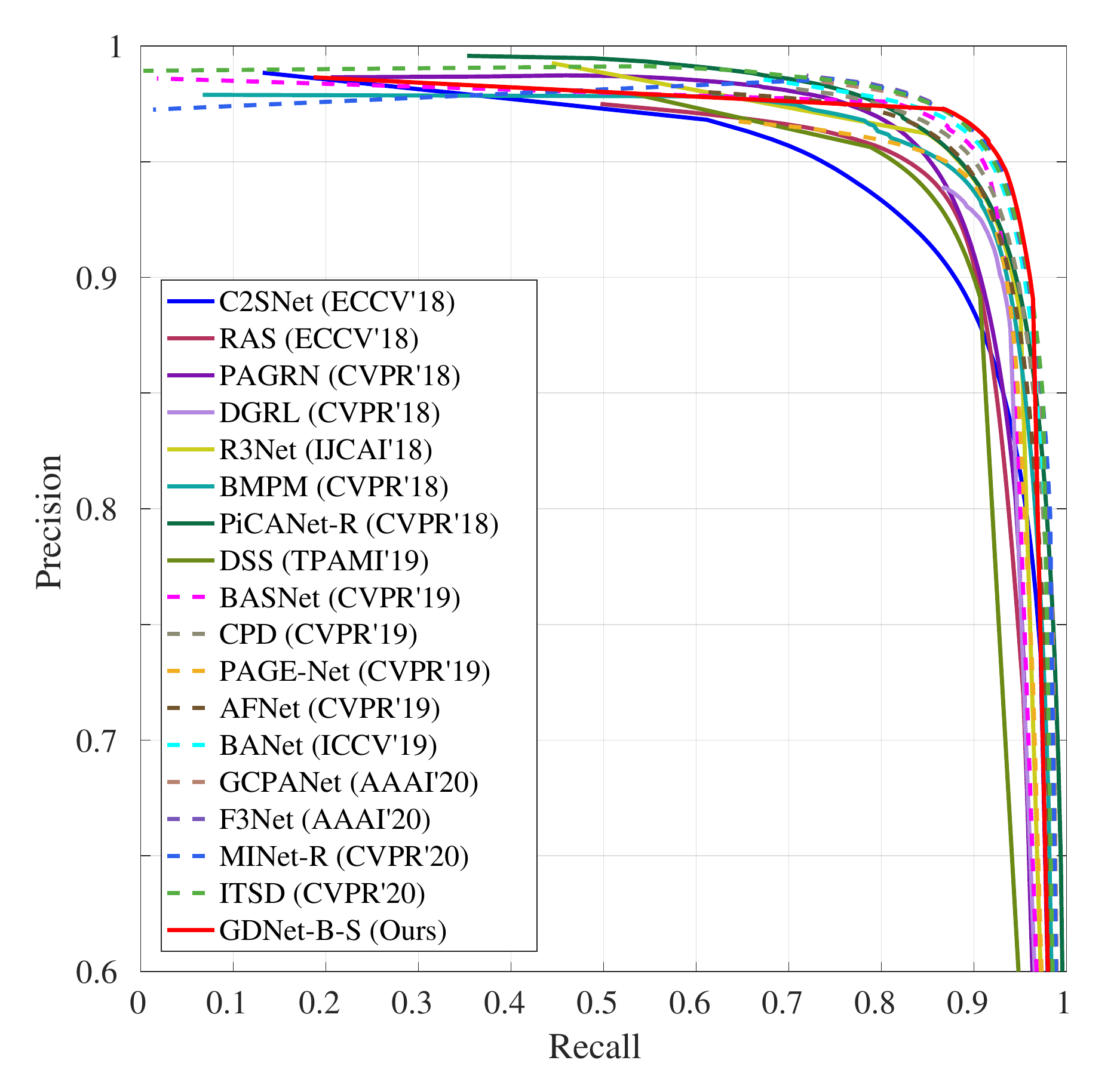} &
		\includegraphics[width=\wprf, height=\hprf]{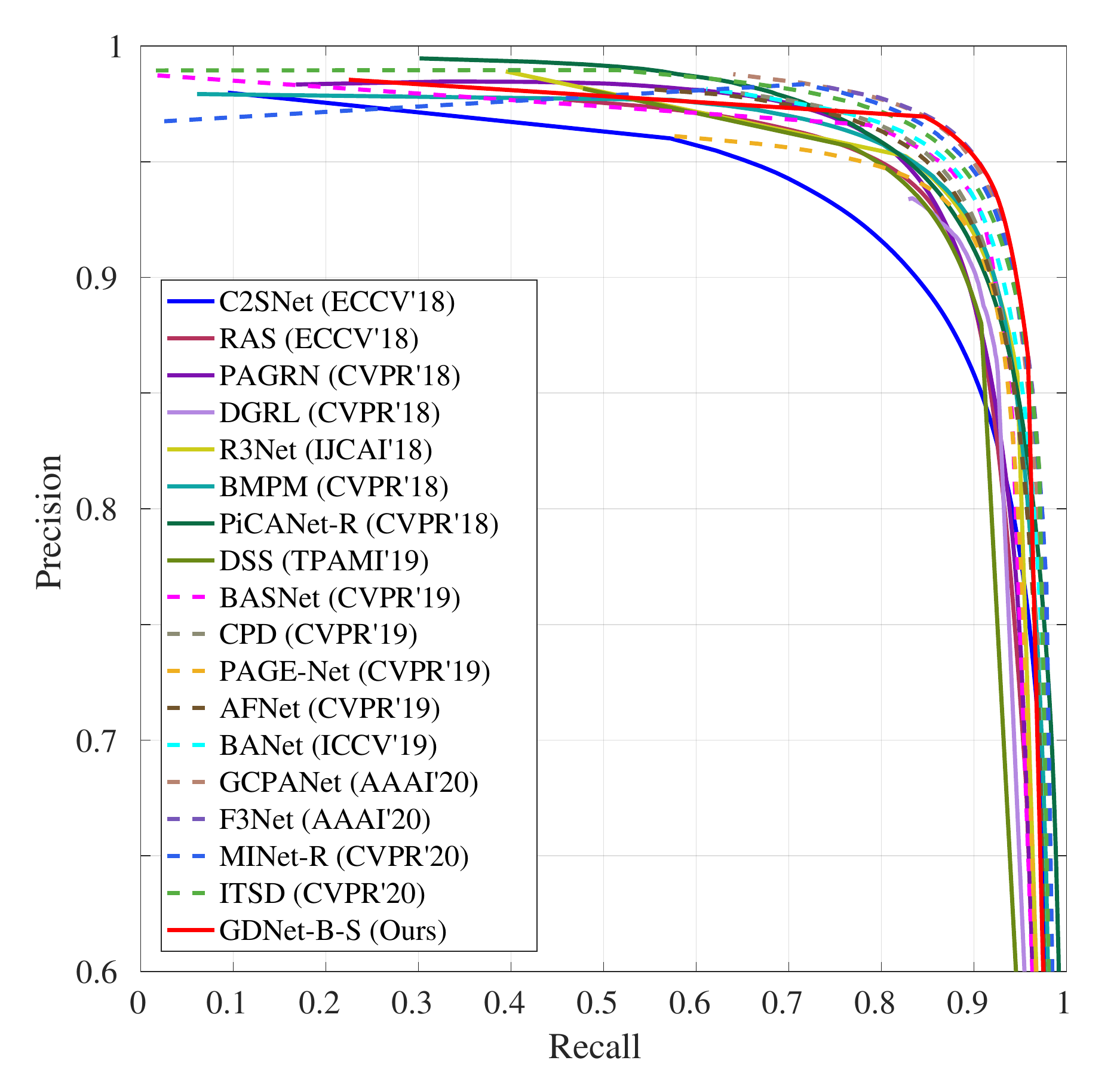} &
		\includegraphics[width=\wprf, height=\hprf]{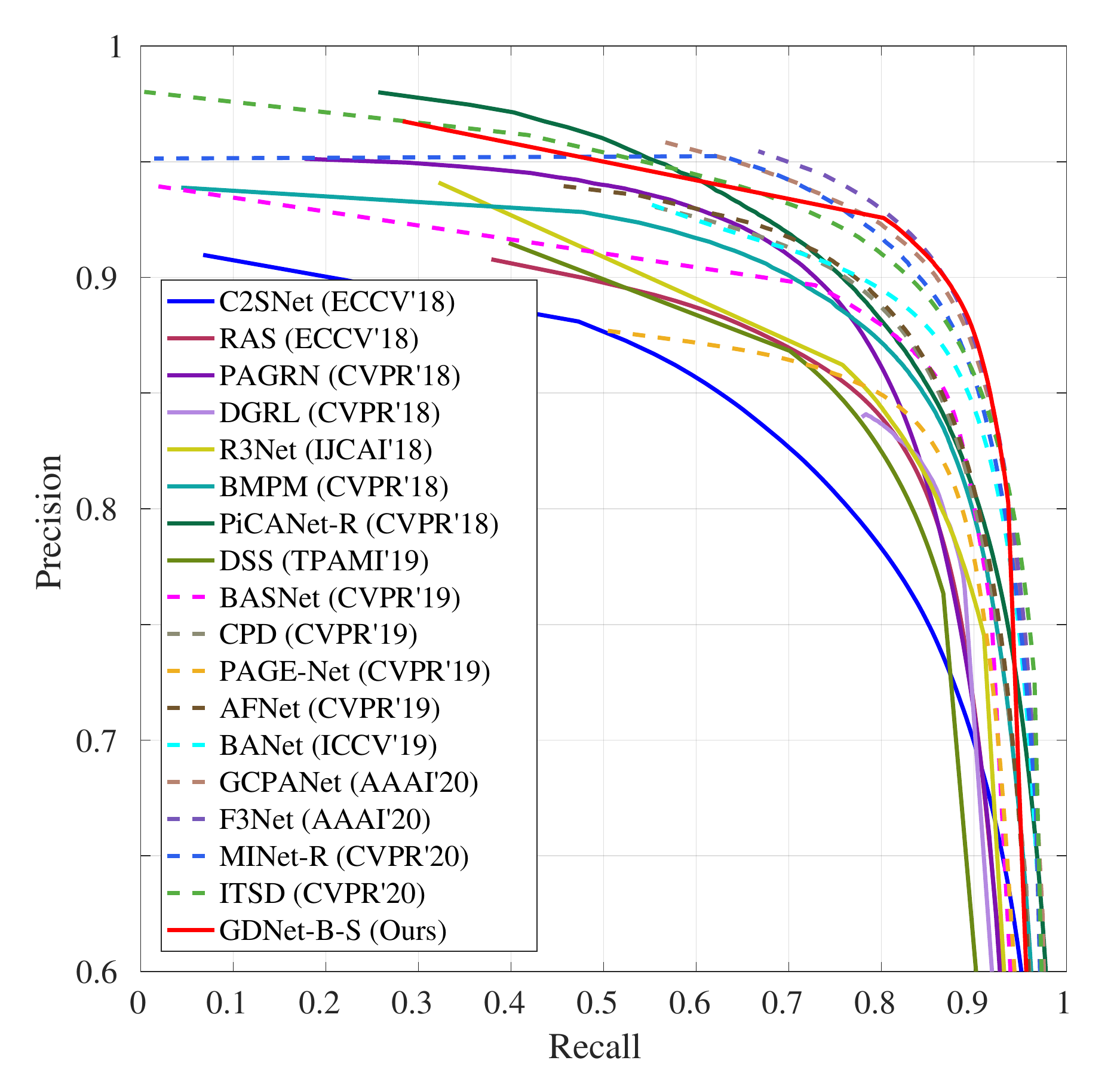} \\
		
		\small{PASCAL-S \cite{Li_2014_CVPR_pascals}} & \small{DUT-OMRON \cite{Yang_2013_ICCV_Workshops}} & \small{ECSSD \cite{Yan_2013_ICCV_Workshops}} & \small{HKU-IS \cite{Li_2015_CVPR_hkuis}} & \small{DUTS-TE \cite{Wang_2017_CVPR_duts}} \\
		
		\includegraphics[width=\wprf, height=\hprf]{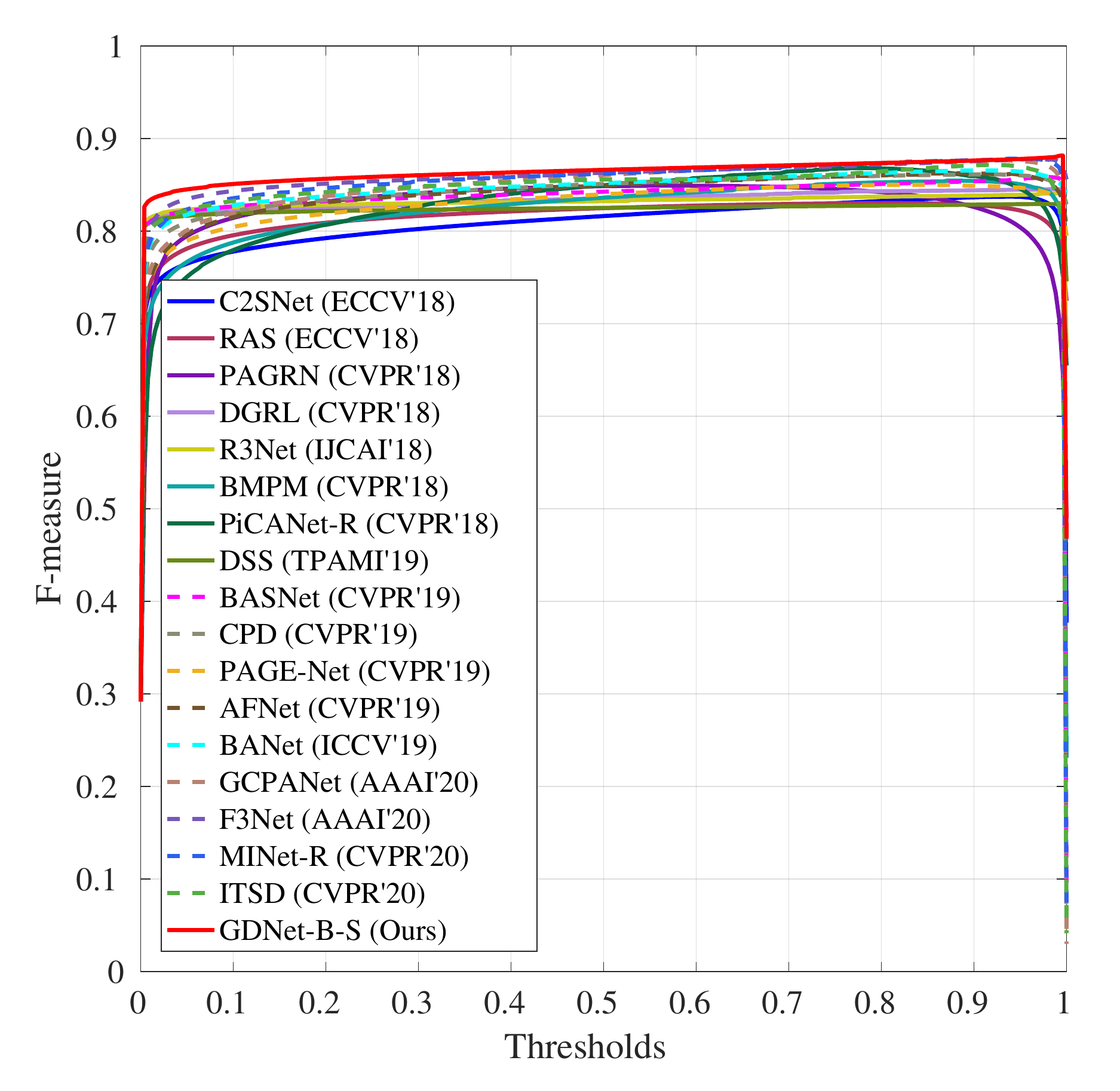} &
		\includegraphics[width=\wprf, height=\hprf]{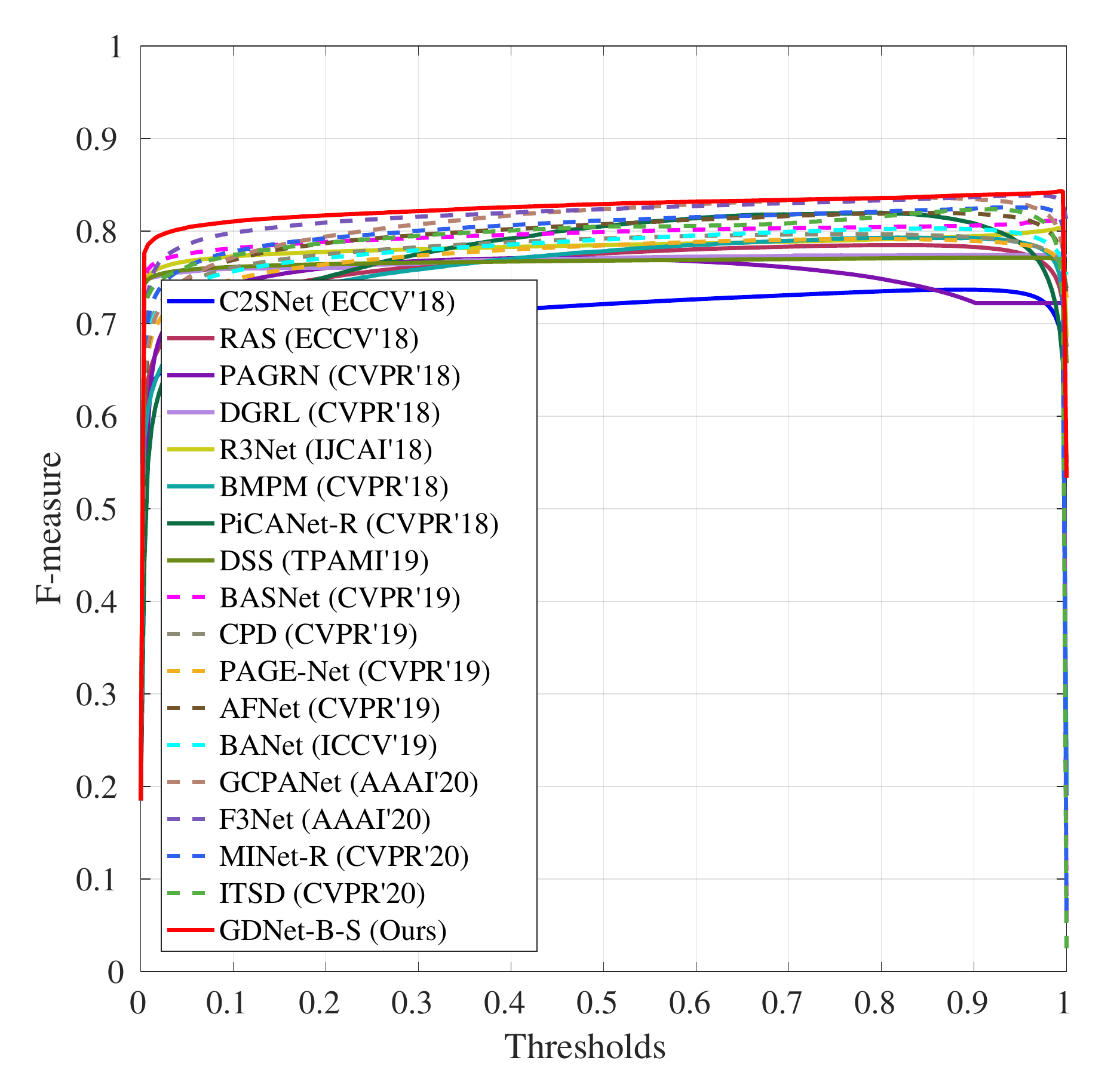} &
		\includegraphics[width=\wprf, height=\hprf]{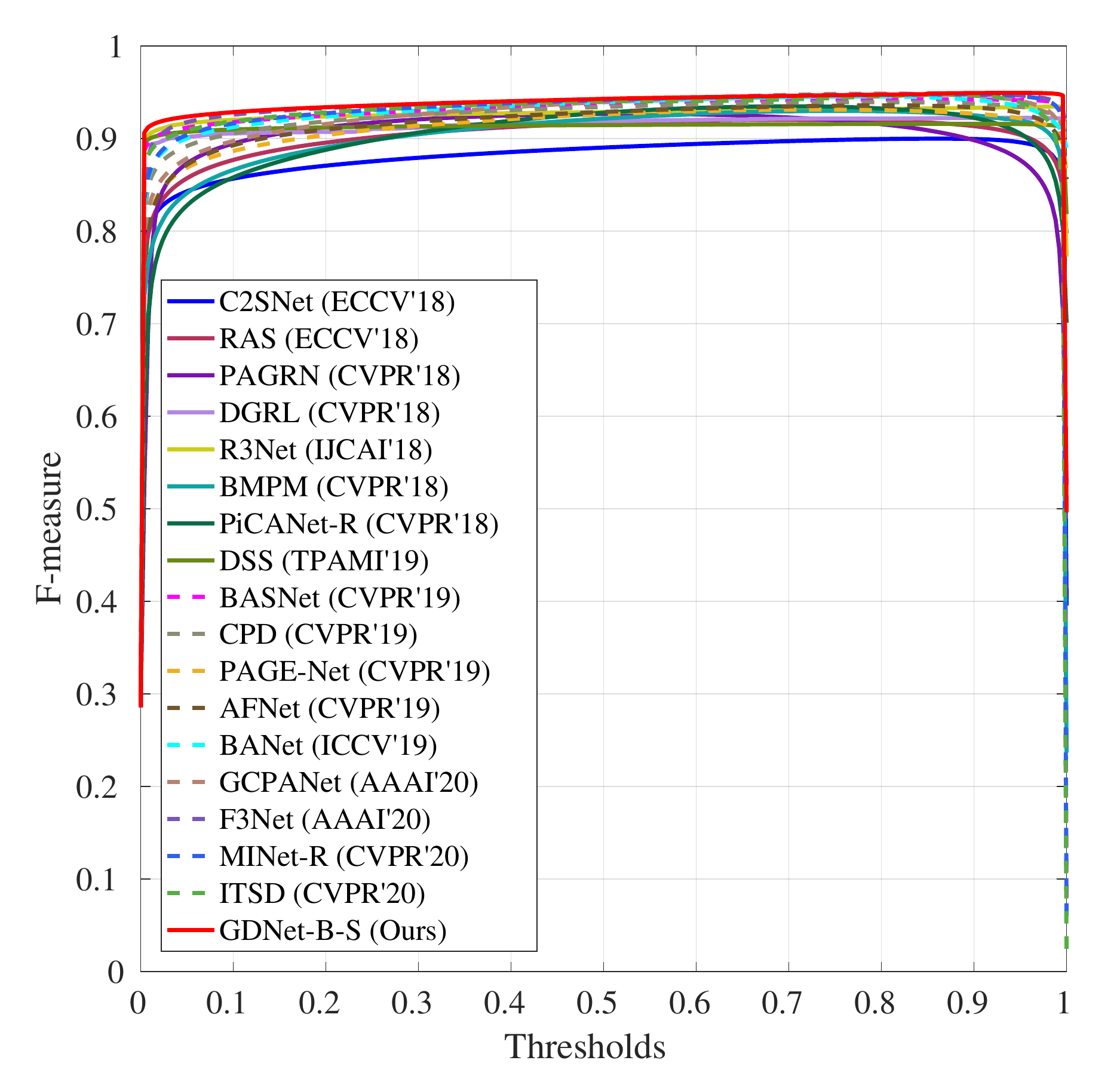} &
		\includegraphics[width=\wprf, height=\hprf]{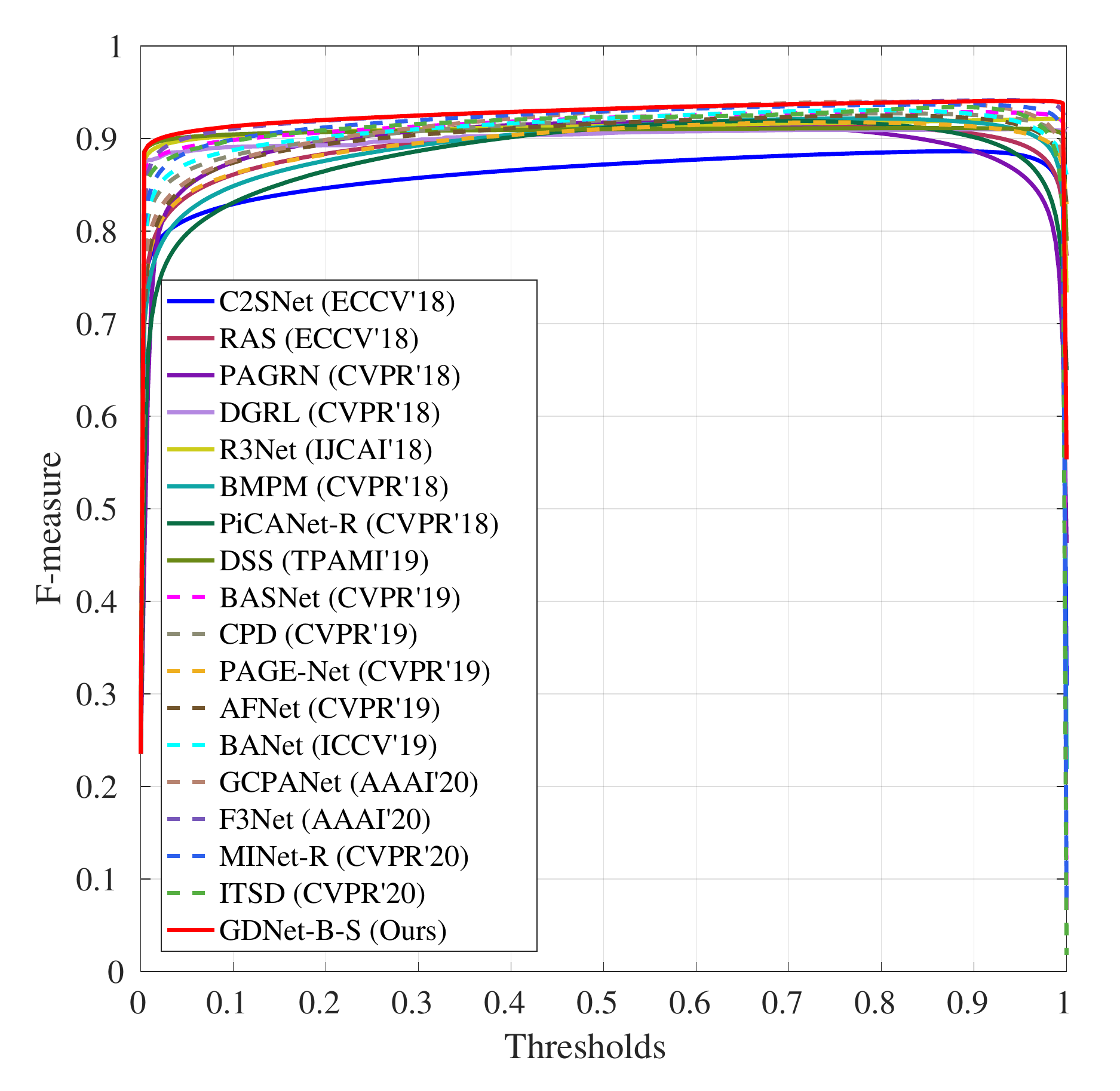} &
		\includegraphics[width=\wprf, height=\hprf]{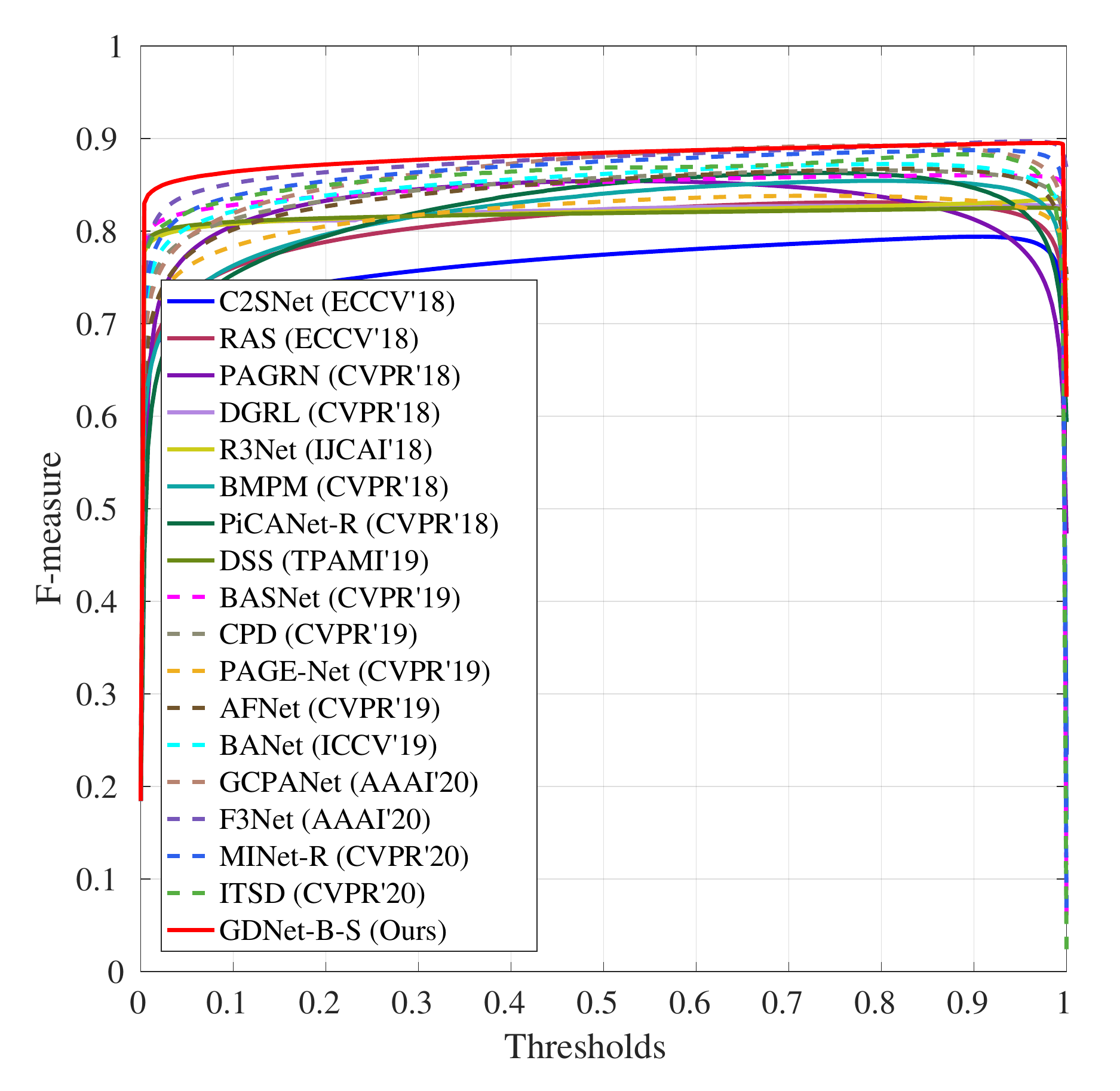} \\
		
		\small{PASCAL-S \cite{Li_2014_CVPR_pascals}} & \small{DUT-OMRON \cite{Yang_2013_ICCV_Workshops}} & \small{ECSSD \cite{Yan_2013_ICCV_Workshops}} & \small{HKU-IS \cite{Li_2015_CVPR_hkuis}} & \small{DUTS-TE \cite{Wang_2017_CVPR_duts}} \\
		
	\end{tabular}
	\caption{\mhy{The $PR$ curves (the top row) and the F-measure curves (the bottom row) on five benchmark datasets. It can be seen that the proposed method performs favorably against the state-of-the-art methods.}}
	\label{fig:curve}
\end{figure*}

\mhy{\textbf{Implementation details.}}
\mhy{We name the GDNet-B re-trained for salient object detection as GDNet-B-S. For training, input images are resized to a resolution of $320\times320$ and are augmented by randomly horizontal flipping. We set the batch size to 10 and keep other settings the same as the experiments on glass detection. It takes about 36 hours for the network to converge for 150 epochs. For testing, the image is first resized to $320\times320$ for network inference, and the output saliency map is then resized back to the original size of the input image. Both resizing processes use bilinear interpolation. There is no post-processing (such as the fully connected conditional random field (CRF) \cite{krahenbuhl2011efficient}) applied to further enhance the final output. The inference time for a $320\times320$ image is only about 0.033 seconds.}

\mhy{\textbf{Evaluation metrics.}}
\mhy{We use six popular metrics to evaluate GDNet-B-S: precision-recall ($PR$) curve, F-measure curve,  structure-measure ($S_\alpha$) \cite{fan2017structure}, E-measure ($E_\phi$) \cite{fan2018enhanced}, weighted F-measure ($F_\beta^w$) \cite{Margolin_2014_CVPR}, and mean absolute error ($M$)}.

\mhy{The $PR$ curve is a standard measure to evaluate the predicted saliency probability maps. The precision and recall of a saliency map are computed by comparing the binarized saliency map against the ground truth map. Under a specific binarizing threshold, we can obtain a pair of average precision and recall over all binarized saliency maps in a dataset. Varying the threshold from 0 to 1 with a step of $\frac{1}{255}$ produces $256$ precision-recall pairs, which are plotted as the $PR$ curve.}

\def\wtallvisual{0.078\linewidth}
\def\htallvisual{.6in}
\def\wwidevisual{0.078\linewidth}
\def\hwidevisual{.385in}
\begin{figure*}[!t]
	\setlength{\tabcolsep}{1.6pt}
	\centering
	\scriptsize
	\begin{tabular}{cccccccccccc}
		\includegraphics[width=\wwidevisual, height=\hwidevisual]{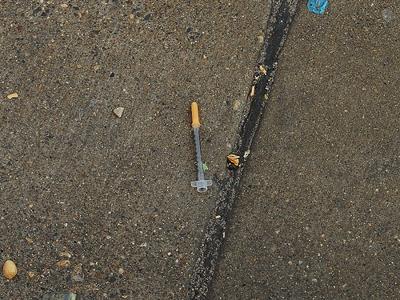}&
		\includegraphics[width=\wwidevisual, height=\hwidevisual]{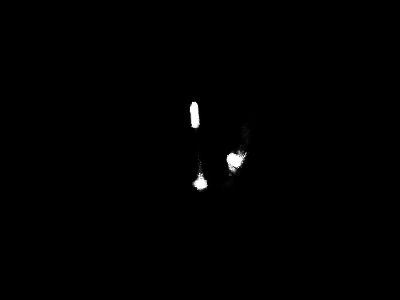}&
		\includegraphics[width=\wwidevisual, height=\hwidevisual]{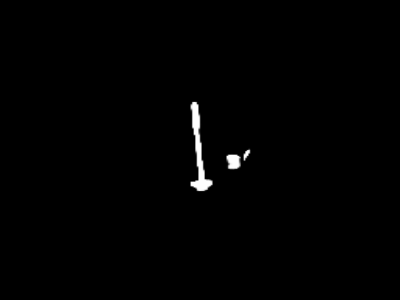}&
		\includegraphics[width=\wwidevisual, height=\hwidevisual]{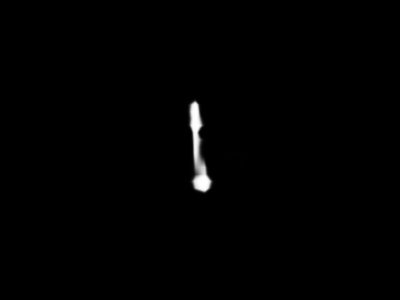}&
		\includegraphics[width=\wwidevisual, height=\hwidevisual]{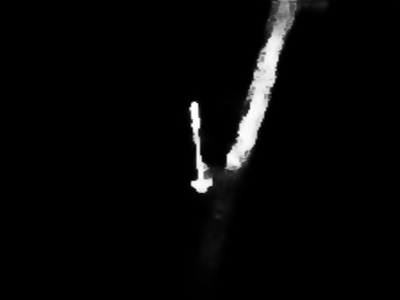}&
		\includegraphics[width=\wwidevisual, height=\hwidevisual]{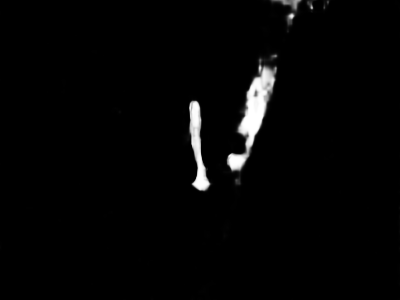}&
		\includegraphics[width=\wwidevisual, height=\hwidevisual]{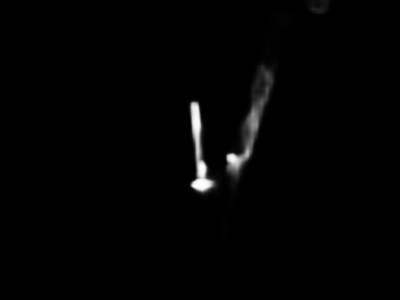}&
		\includegraphics[width=\wwidevisual, height=\hwidevisual]{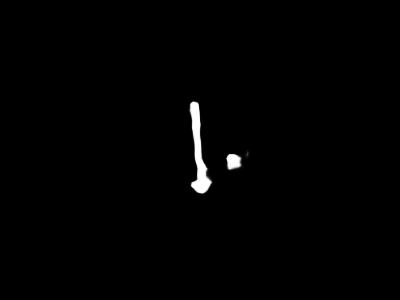}&
		\includegraphics[width=\wwidevisual, height=\hwidevisual]{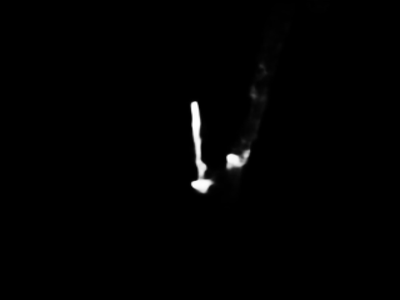}&
		\includegraphics[width=\wwidevisual, height=\hwidevisual]{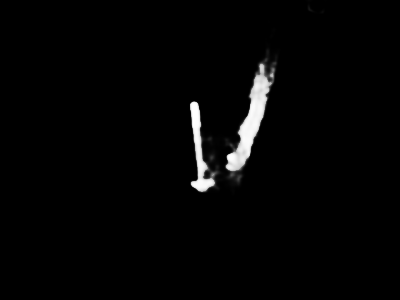}&
		\includegraphics[width=\wwidevisual, height=\hwidevisual]{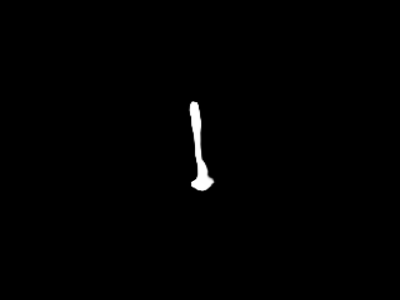}&
		\includegraphics[width=\wwidevisual, height=\hwidevisual]{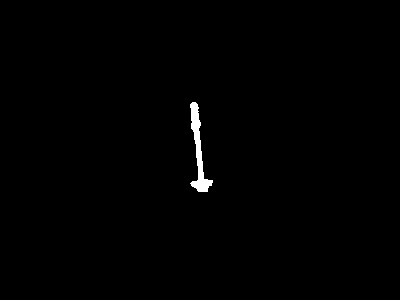} \\
		
		\includegraphics[width=\wtallvisual, height=\htallvisual]{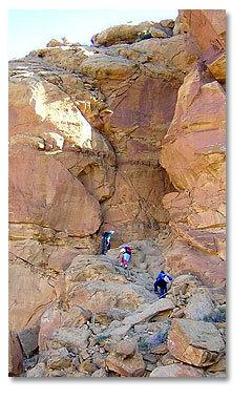}&
		\includegraphics[width=\wtallvisual, height=\htallvisual]{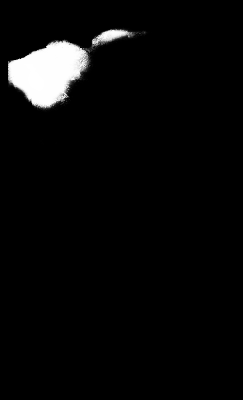}&
		\includegraphics[width=\wtallvisual, height=\htallvisual]{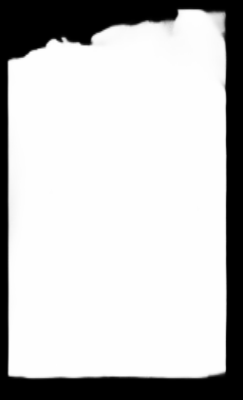}&
		\includegraphics[width=\wtallvisual, height=\htallvisual]{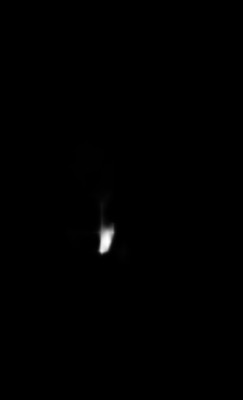}&
		\includegraphics[width=\wtallvisual, height=\htallvisual]{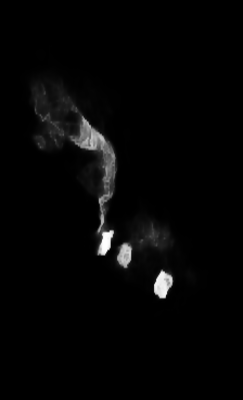}&
		\includegraphics[width=\wtallvisual, height=\htallvisual]{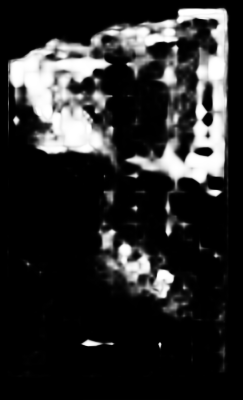}&
		\includegraphics[width=\wtallvisual, height=\htallvisual]{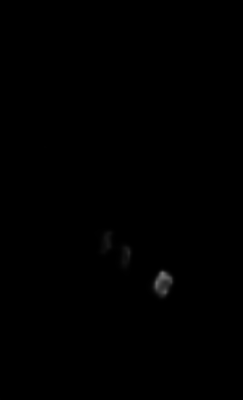}&
		\includegraphics[width=\wtallvisual, height=\htallvisual]{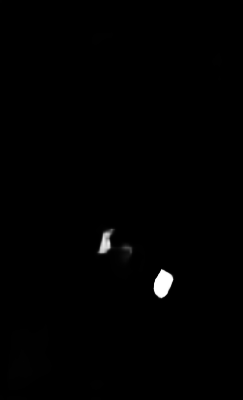}&
		\includegraphics[width=\wtallvisual, height=\htallvisual]{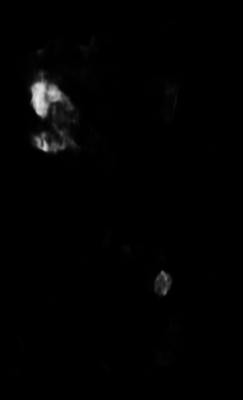}&
		\includegraphics[width=\wtallvisual, height=\htallvisual]{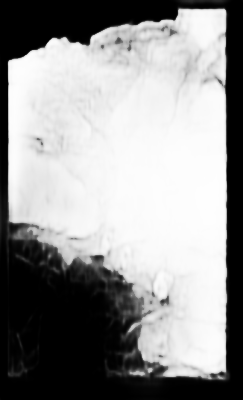}&
		\includegraphics[width=\wtallvisual, height=\htallvisual]{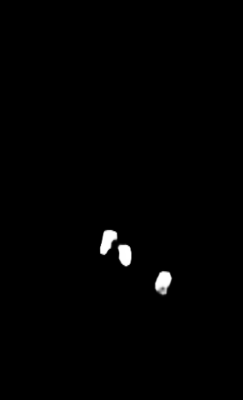}&
		\includegraphics[width=\wtallvisual, height=\htallvisual]{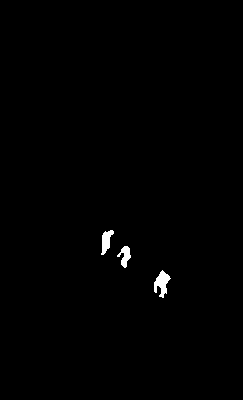} \\
		
		\includegraphics[width=\wwidevisual, height=\hwidevisual]{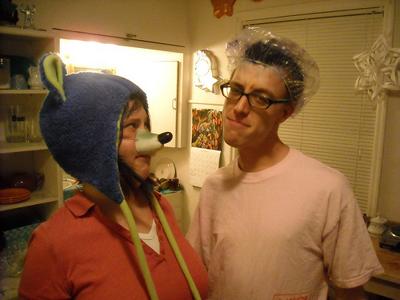}&
		\includegraphics[width=\wwidevisual, height=\hwidevisual]{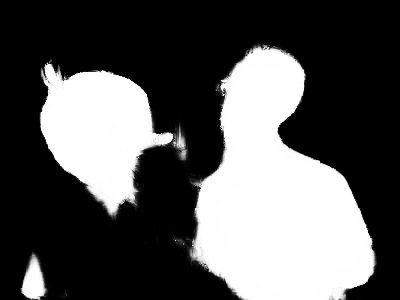}&
		\includegraphics[width=\wwidevisual, height=\hwidevisual]{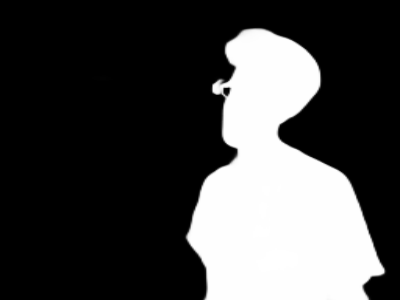}&
		\includegraphics[width=\wwidevisual, height=\hwidevisual]{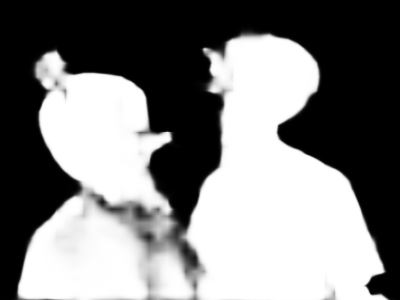}&
		\includegraphics[width=\wwidevisual, height=\hwidevisual]{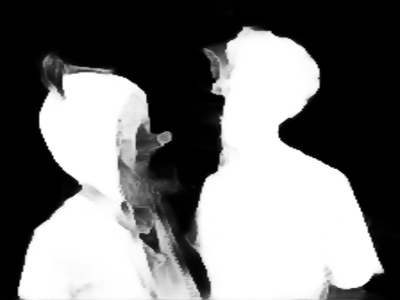}&
		\includegraphics[width=\wwidevisual, height=\hwidevisual]{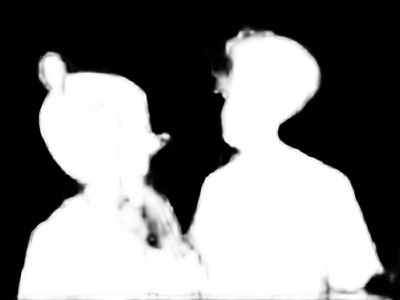}&
		\includegraphics[width=\wwidevisual, height=\hwidevisual]{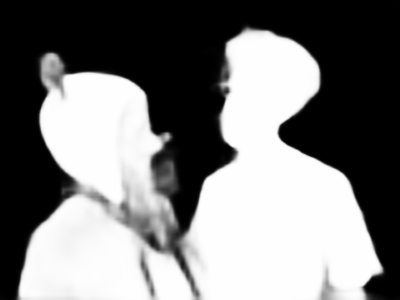}&
		\includegraphics[width=\wwidevisual, height=\hwidevisual]{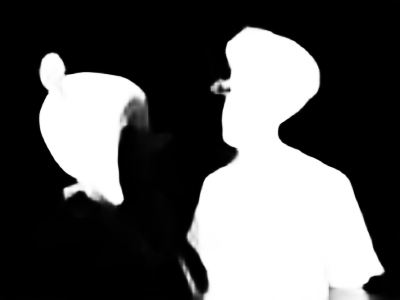}&
		\includegraphics[width=\wwidevisual, height=\hwidevisual]{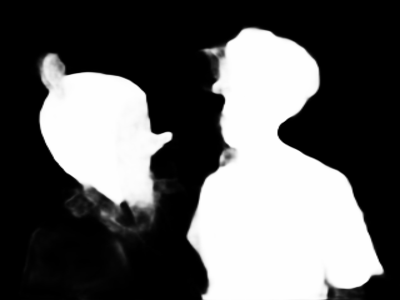}&
		\includegraphics[width=\wwidevisual, height=\hwidevisual]{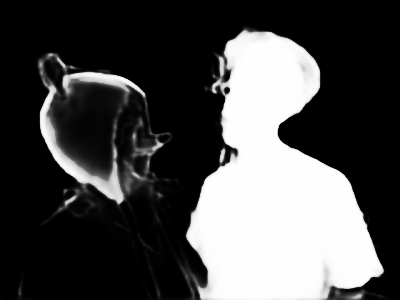}&
		\includegraphics[width=\wwidevisual, height=\hwidevisual]{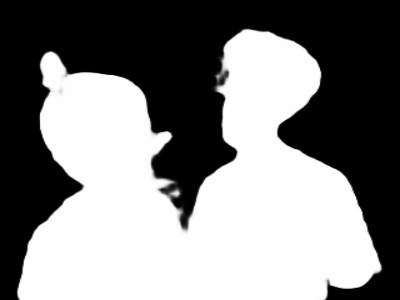}&
		\includegraphics[width=\wwidevisual, height=\hwidevisual]{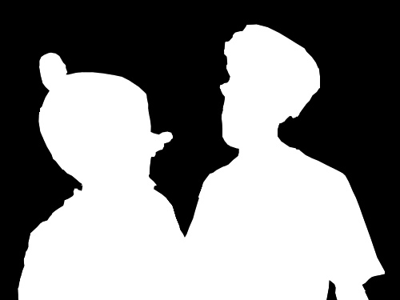} \\
		
		\includegraphics[width=\wwidevisual, height=\hwidevisual]{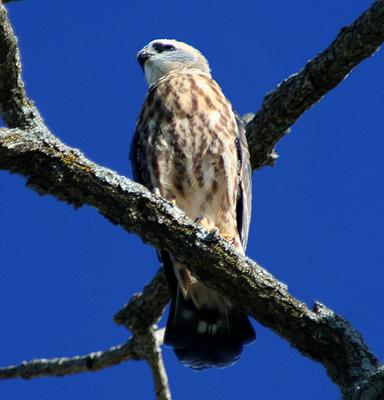}&
		\includegraphics[width=\wwidevisual, height=\hwidevisual]{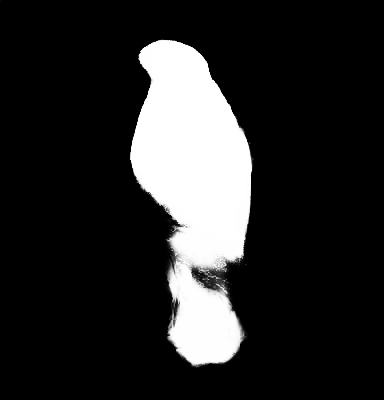}&
		\includegraphics[width=\wwidevisual, height=\hwidevisual]{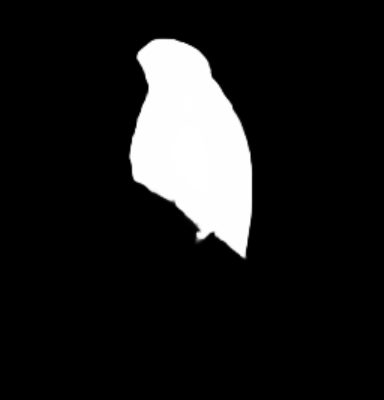}&
		\includegraphics[width=\wwidevisual, height=\hwidevisual]{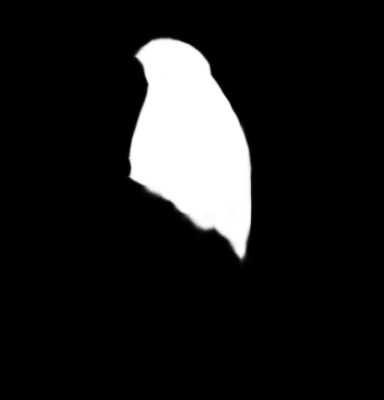}&
		\includegraphics[width=\wwidevisual, height=\hwidevisual]{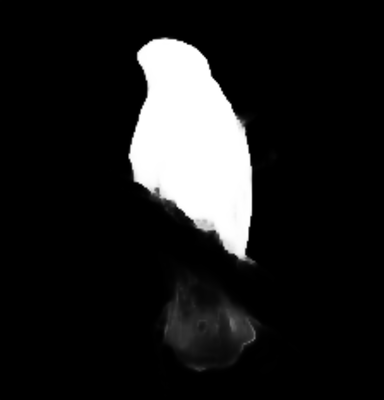}&
		\includegraphics[width=\wwidevisual, height=\hwidevisual]{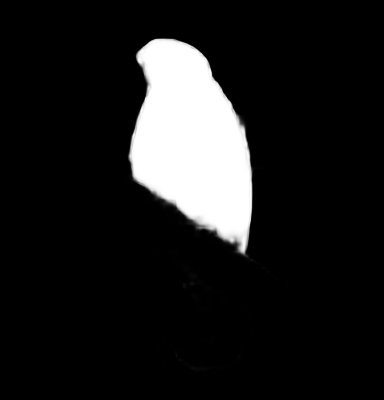}&
		\includegraphics[width=\wwidevisual, height=\hwidevisual]{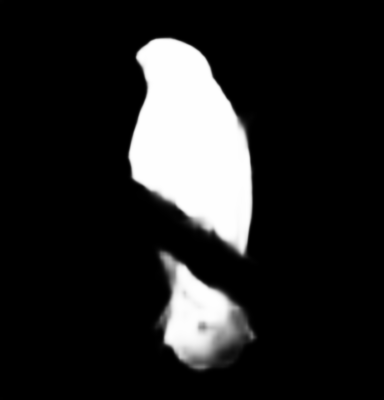}&
		\includegraphics[width=\wwidevisual, height=\hwidevisual]{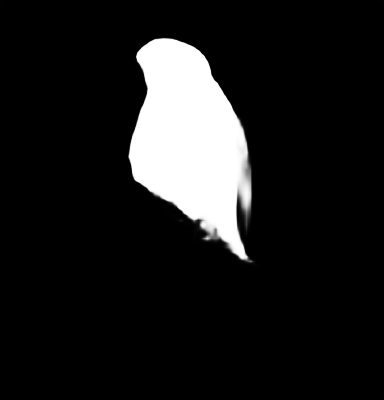}&
		\includegraphics[width=\wwidevisual, height=\hwidevisual]{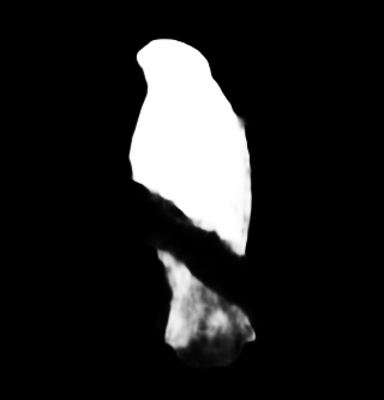}&
		\includegraphics[width=\wwidevisual, height=\hwidevisual]{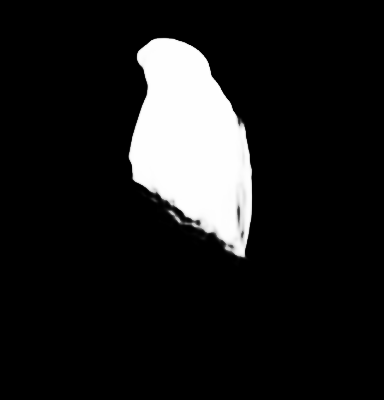}&
		\includegraphics[width=\wwidevisual, height=\hwidevisual]{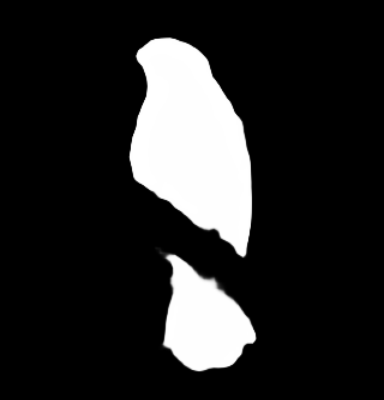}&
		\includegraphics[width=\wwidevisual, height=\hwidevisual]{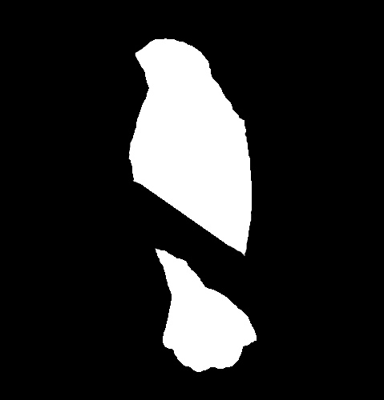} \\
		
		\includegraphics[width=\wwidevisual, height=\hwidevisual]{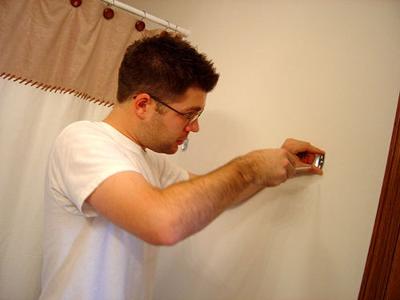}&
		\includegraphics[width=\wwidevisual, height=\hwidevisual]{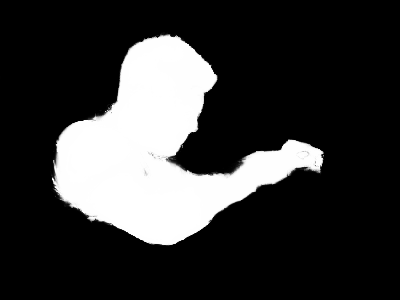}&
		\includegraphics[width=\wwidevisual, height=\hwidevisual]{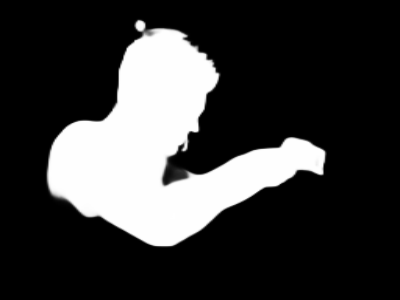}&
		\includegraphics[width=\wwidevisual, height=\hwidevisual]{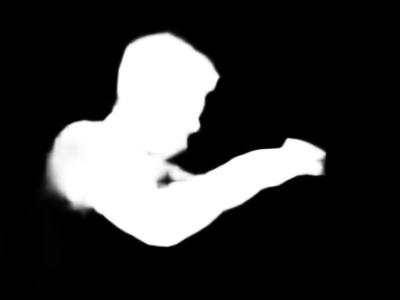}&
		\includegraphics[width=\wwidevisual, height=\hwidevisual]{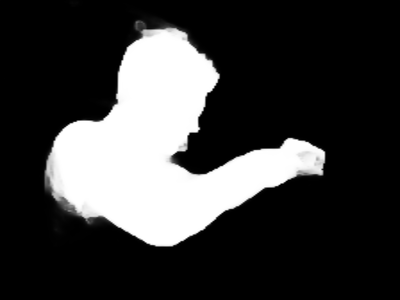}&
		\includegraphics[width=\wwidevisual, height=\hwidevisual]{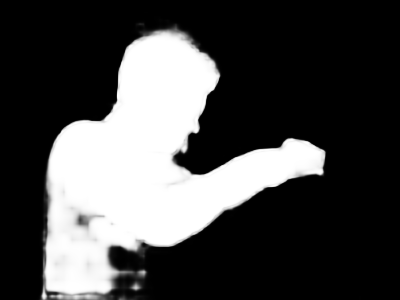}&
		\includegraphics[width=\wwidevisual, height=\hwidevisual]{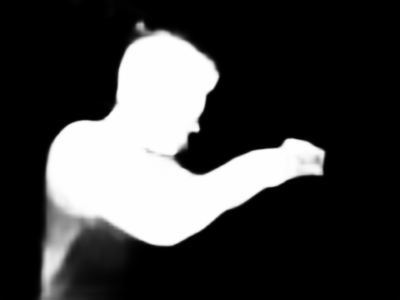}&
		\includegraphics[width=\wwidevisual, height=\hwidevisual]{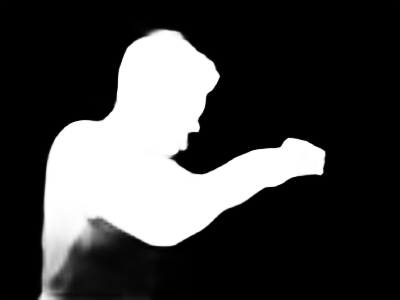}&
		\includegraphics[width=\wwidevisual, height=\hwidevisual]{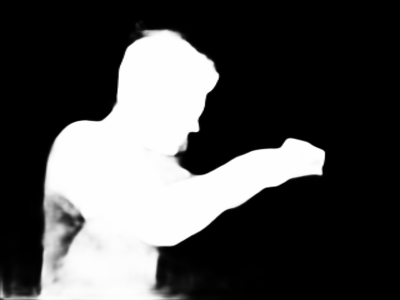}&
		\includegraphics[width=\wwidevisual, height=\hwidevisual]{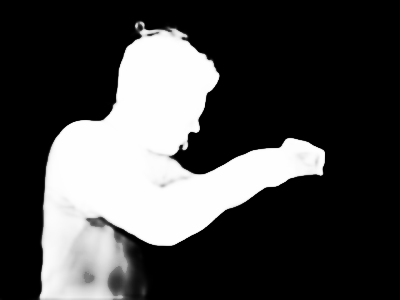}&
		\includegraphics[width=\wwidevisual, height=\hwidevisual]{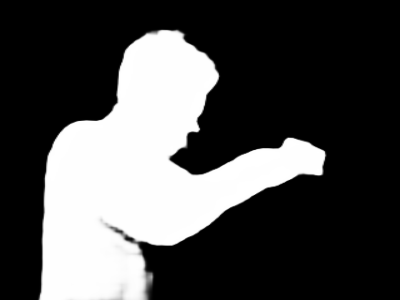}&
		\includegraphics[width=\wwidevisual, height=\hwidevisual]{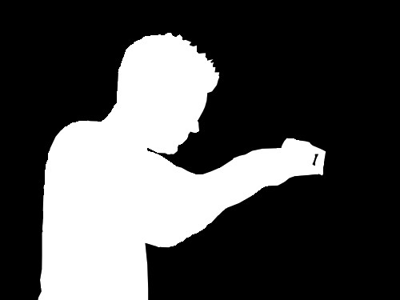} \\
		
		\includegraphics[width=\wwidevisual, height=\hwidevisual]{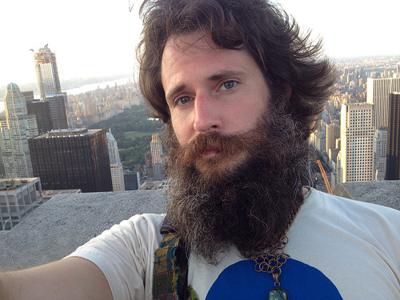}&
		\includegraphics[width=\wwidevisual, height=\hwidevisual]{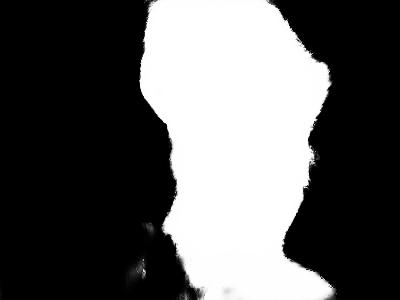}&
		\includegraphics[width=\wwidevisual, height=\hwidevisual]{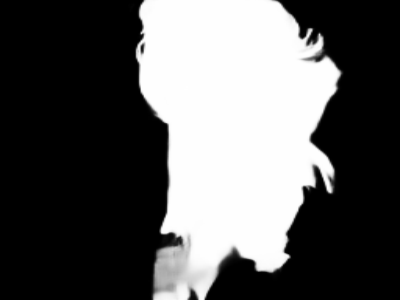}&
		\includegraphics[width=\wwidevisual, height=\hwidevisual]{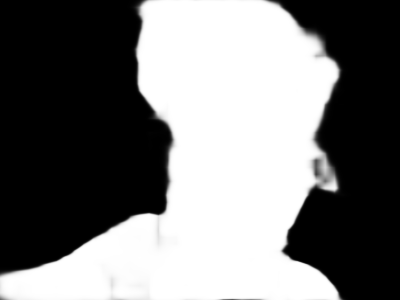}&
		\includegraphics[width=\wwidevisual, height=\hwidevisual]{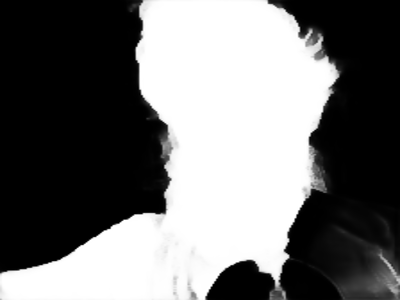}&
		\includegraphics[width=\wwidevisual, height=\hwidevisual]{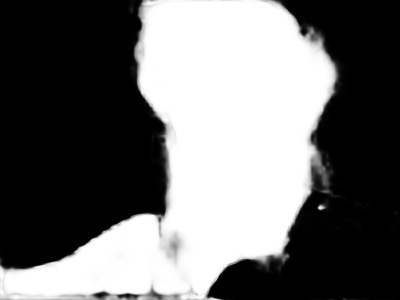}&
		\includegraphics[width=\wwidevisual, height=\hwidevisual]{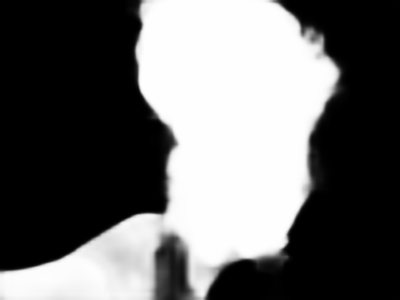}&
		\includegraphics[width=\wwidevisual, height=\hwidevisual]{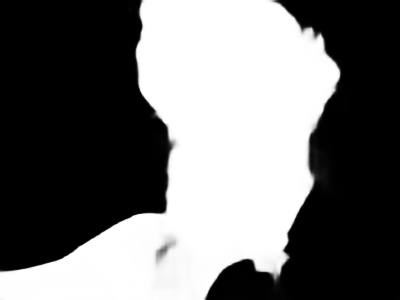}&
		\includegraphics[width=\wwidevisual, height=\hwidevisual]{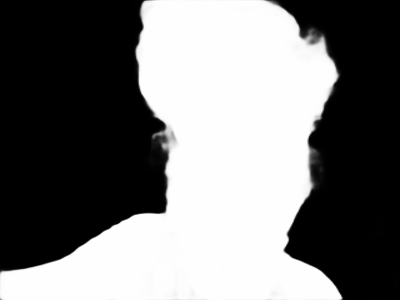}&
		\includegraphics[width=\wwidevisual, height=\hwidevisual]{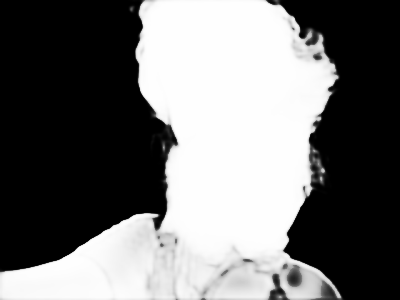}&
		\includegraphics[width=\wwidevisual, height=\hwidevisual]{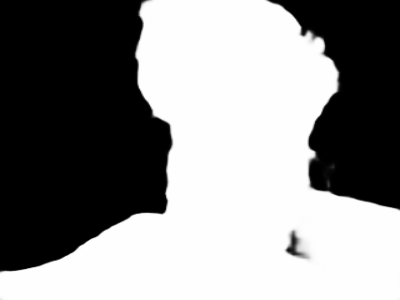}&
		\includegraphics[width=\wwidevisual, height=\hwidevisual]{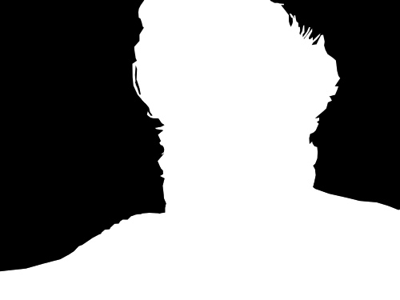} \\
		
		\includegraphics[width=\wtallvisual, height=\htallvisual]{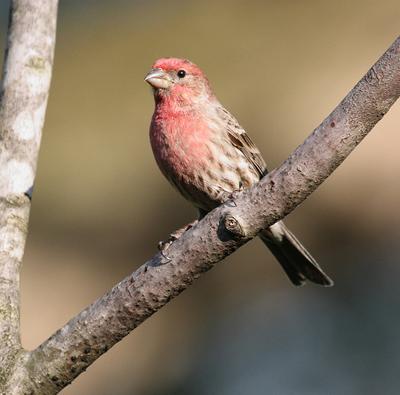}&
		\includegraphics[width=\wtallvisual, height=\htallvisual]{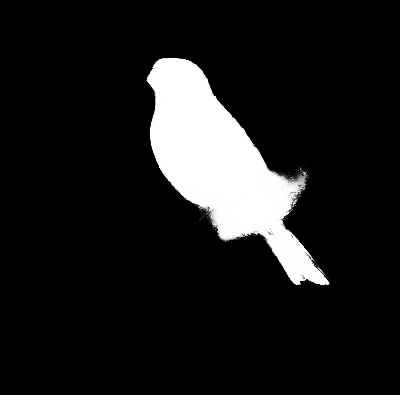}&
		\includegraphics[width=\wtallvisual, height=\htallvisual]{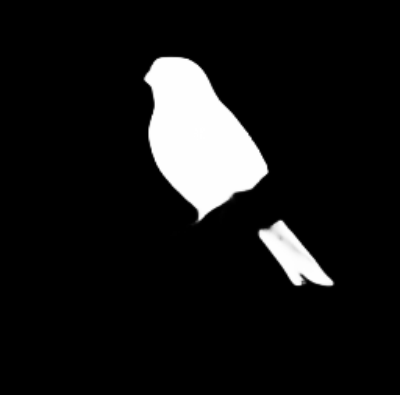}&
		\includegraphics[width=\wtallvisual, height=\htallvisual]{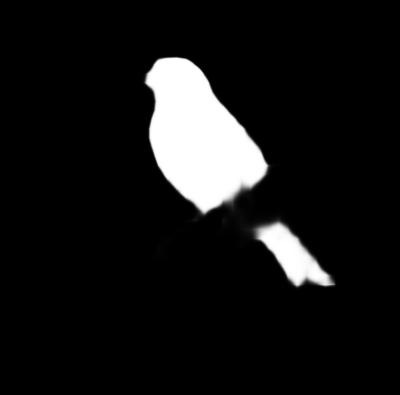}&
		\includegraphics[width=\wtallvisual, height=\htallvisual]{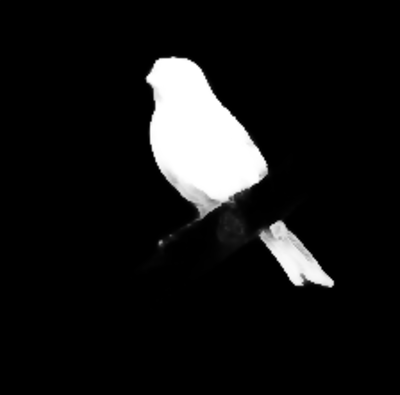}&
		\includegraphics[width=\wtallvisual, height=\htallvisual]{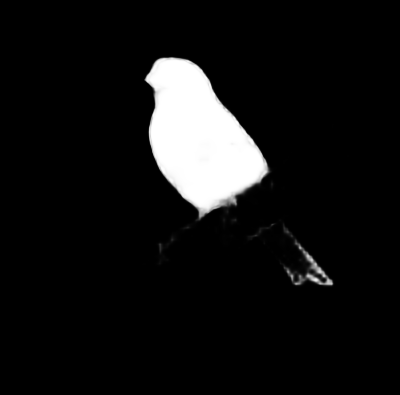}&
		\includegraphics[width=\wtallvisual, height=\htallvisual]{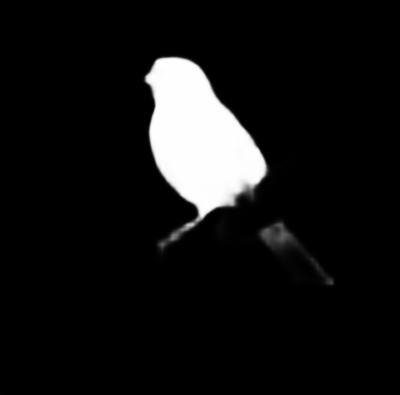}&
		\includegraphics[width=\wtallvisual, height=\htallvisual]{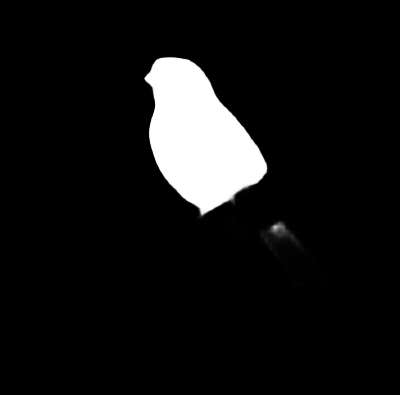}&
		\includegraphics[width=\wtallvisual, height=\htallvisual]{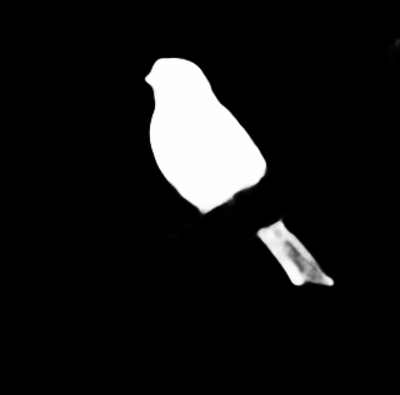}&
		\includegraphics[width=\wtallvisual, height=\htallvisual]{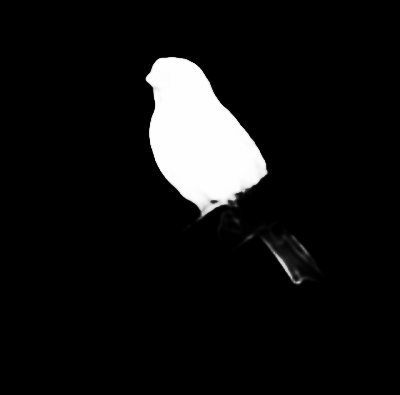}&
		\includegraphics[width=\wtallvisual, height=\htallvisual]{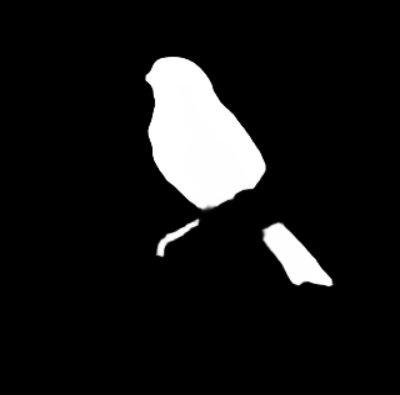}&
		\includegraphics[width=\wtallvisual, height=\htallvisual]{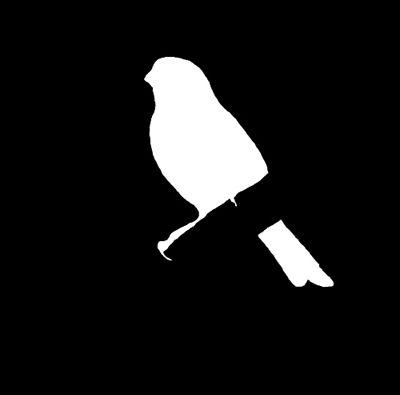} \\
		
		\includegraphics[width=\wwidevisual, height=\hwidevisual]{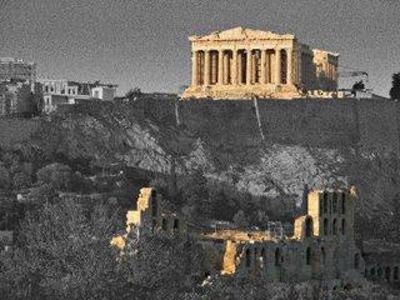}&
		\includegraphics[width=\wwidevisual, height=\hwidevisual]{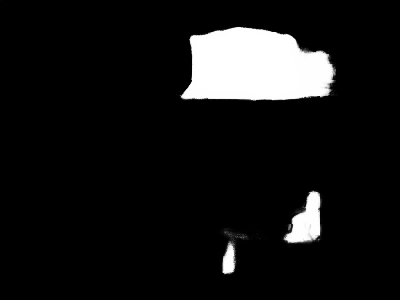}&
		\includegraphics[width=\wwidevisual, height=\hwidevisual]{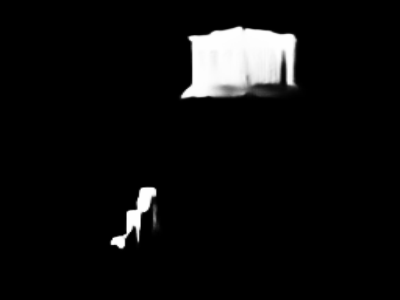}&
		\includegraphics[width=\wwidevisual, height=\hwidevisual]{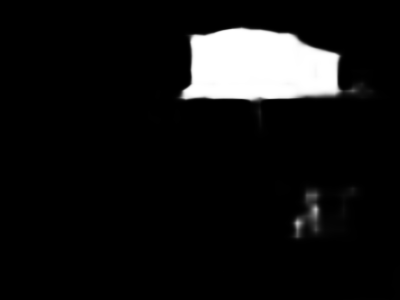}&
		\includegraphics[width=\wwidevisual, height=\hwidevisual]{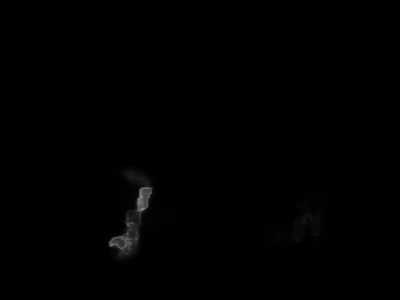}&
		\includegraphics[width=\wwidevisual, height=\hwidevisual]{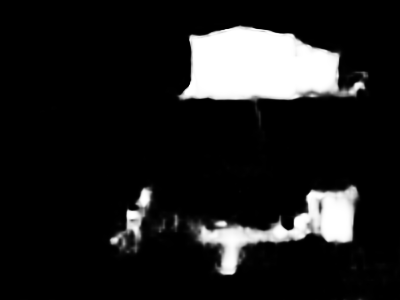}&
		\includegraphics[width=\wwidevisual, height=\hwidevisual]{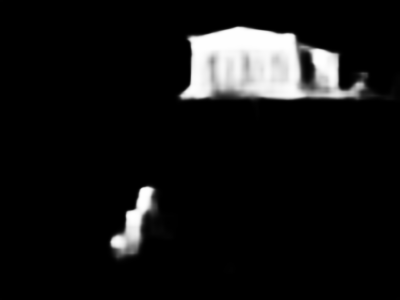}&
		\includegraphics[width=\wwidevisual, height=\hwidevisual]{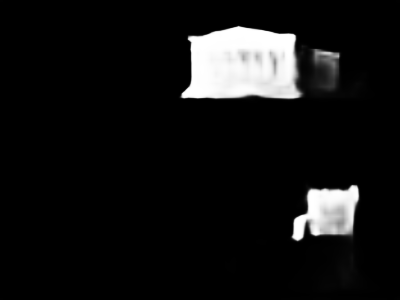}&
		\includegraphics[width=\wwidevisual, height=\hwidevisual]{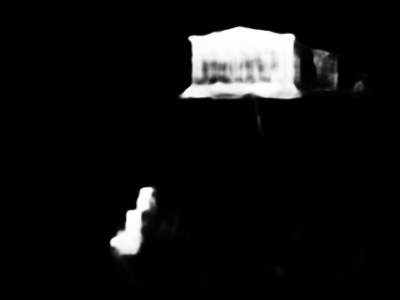}&
		\includegraphics[width=\wwidevisual, height=\hwidevisual]{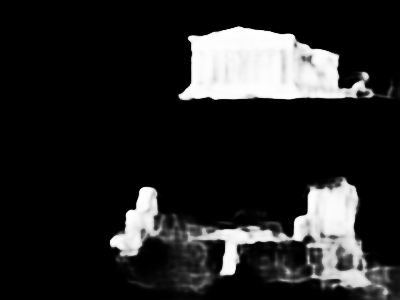}&
		\includegraphics[width=\wwidevisual, height=\hwidevisual]{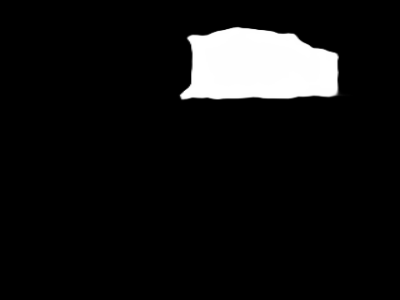}&
		\includegraphics[width=\wwidevisual, height=\hwidevisual]{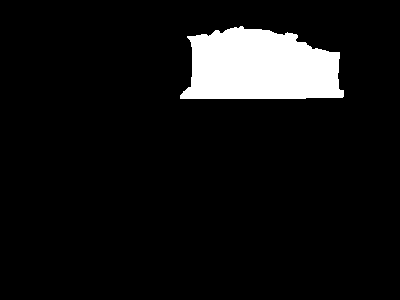} \\
		
		\includegraphics[width=\wwidevisual, height=\hwidevisual]{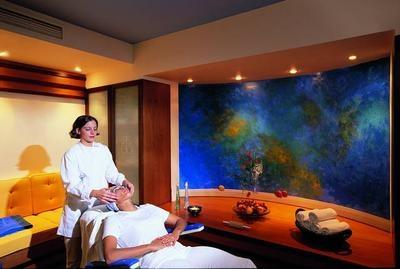}&
		\includegraphics[width=\wwidevisual, height=\hwidevisual]{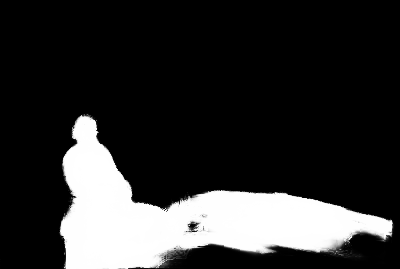}&
		\includegraphics[width=\wwidevisual, height=\hwidevisual]{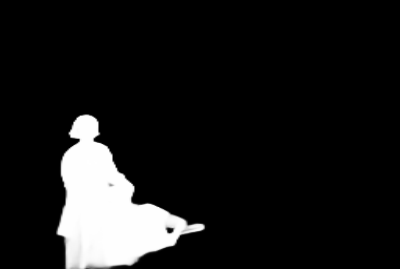}&
		\includegraphics[width=\wwidevisual, height=\hwidevisual]{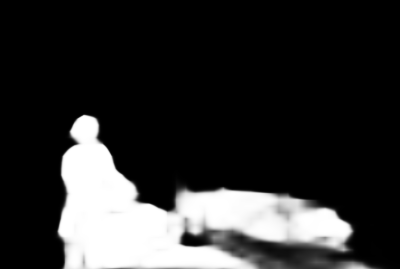}&
		\includegraphics[width=\wwidevisual, height=\hwidevisual]{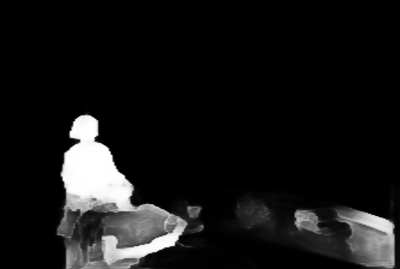}&
		\includegraphics[width=\wwidevisual, height=\hwidevisual]{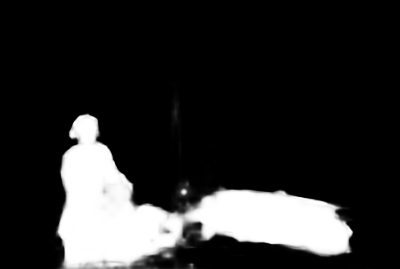}&
		\includegraphics[width=\wwidevisual, height=\hwidevisual]{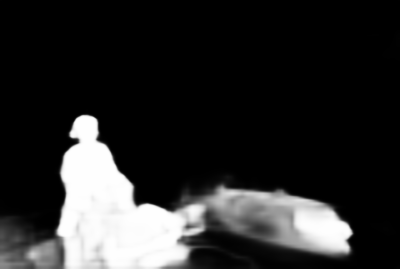}&
		\includegraphics[width=\wwidevisual, height=\hwidevisual]{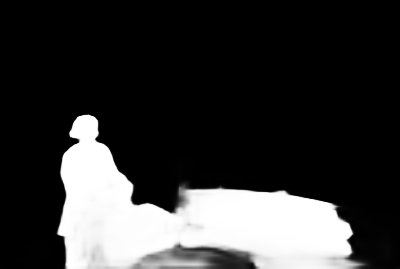}&
		\includegraphics[width=\wwidevisual, height=\hwidevisual]{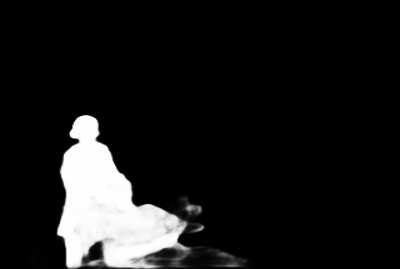}&
		\includegraphics[width=\wwidevisual, height=\hwidevisual]{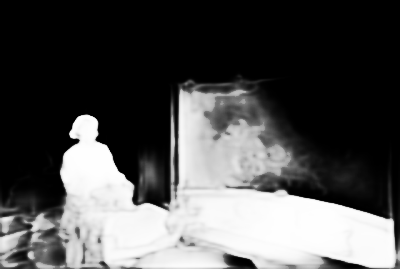}&
		\includegraphics[width=\wwidevisual, height=\hwidevisual]{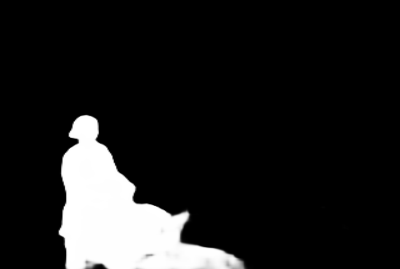}&
		\includegraphics[width=\wwidevisual, height=\hwidevisual]{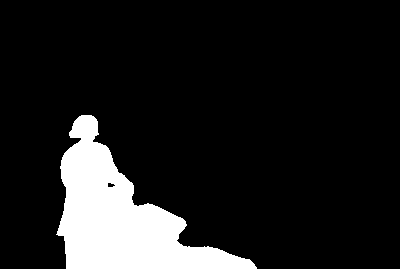} \\	
		
		\includegraphics[width=\wwidevisual, height=\hwidevisual]{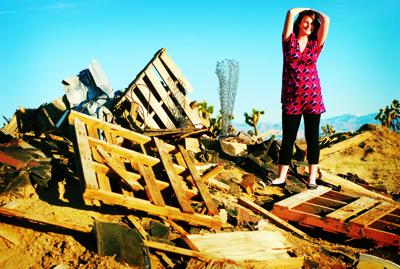}&
		\includegraphics[width=\wwidevisual, height=\hwidevisual]{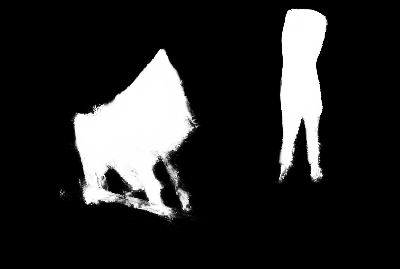}&
		\includegraphics[width=\wwidevisual, height=\hwidevisual]{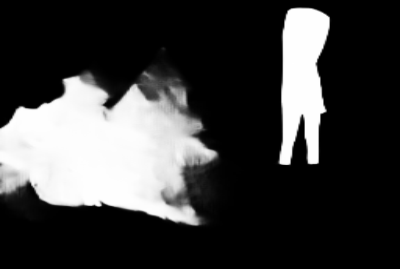}&
		\includegraphics[width=\wwidevisual, height=\hwidevisual]{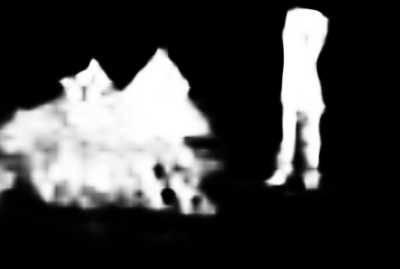}&
		\includegraphics[width=\wwidevisual, height=\hwidevisual]{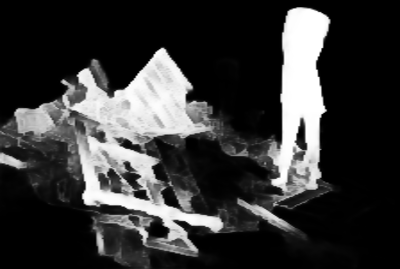}&
		\includegraphics[width=\wwidevisual, height=\hwidevisual]{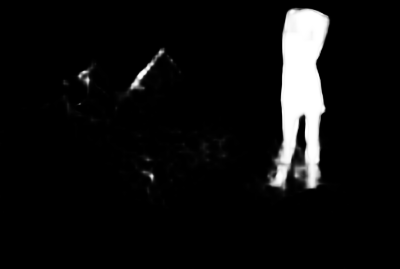}&
		\includegraphics[width=\wwidevisual, height=\hwidevisual]{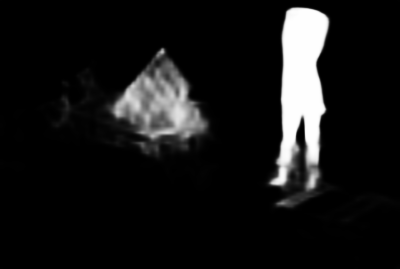}&
		\includegraphics[width=\wwidevisual, height=\hwidevisual]{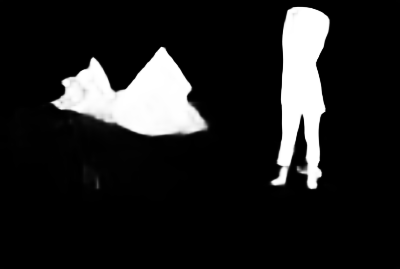}&
		\includegraphics[width=\wwidevisual, height=\hwidevisual]{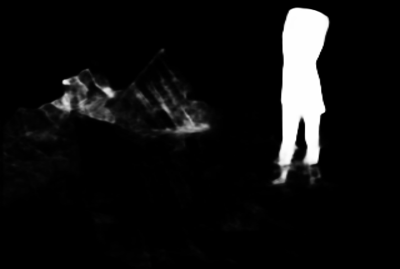}&
		\includegraphics[width=\wwidevisual, height=\hwidevisual]{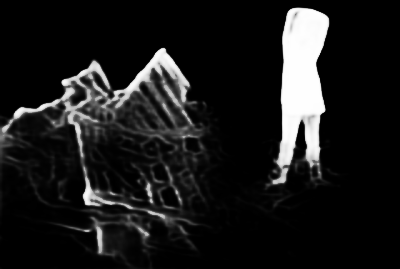}&
		\includegraphics[width=\wwidevisual, height=\hwidevisual]{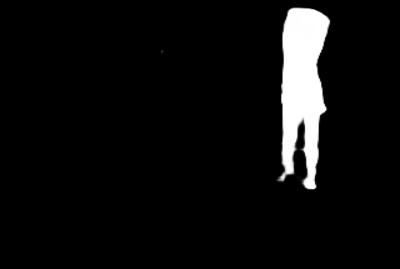}&
		\includegraphics[width=\wwidevisual, height=\hwidevisual]{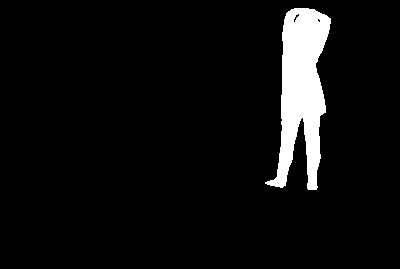} \\
		
		\includegraphics[width=\wwidevisual, height=\hwidevisual]{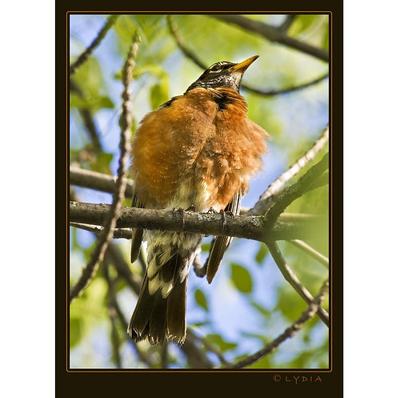}&
		\includegraphics[width=\wwidevisual, height=\hwidevisual]{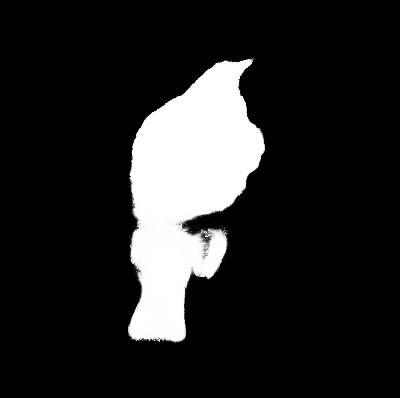}&
		\includegraphics[width=\wwidevisual, height=\hwidevisual]{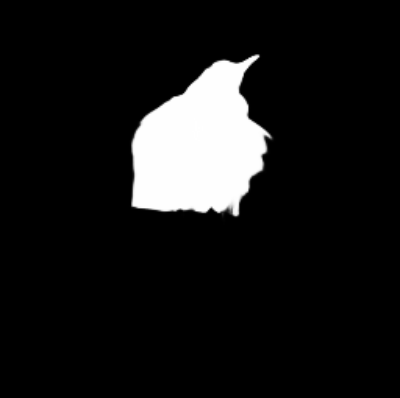}&
		\includegraphics[width=\wwidevisual, height=\hwidevisual]{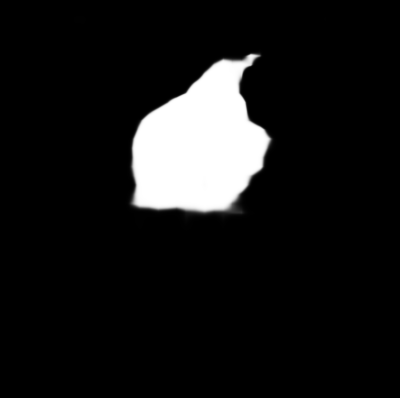}&
		\includegraphics[width=\wwidevisual, height=\hwidevisual]{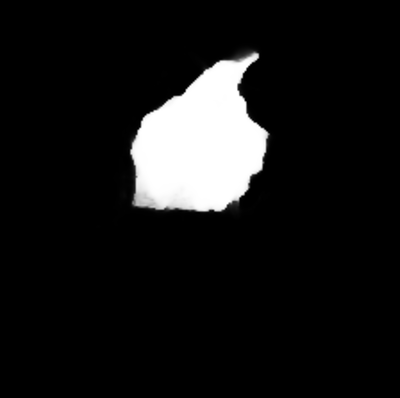}&
		\includegraphics[width=\wwidevisual, height=\hwidevisual]{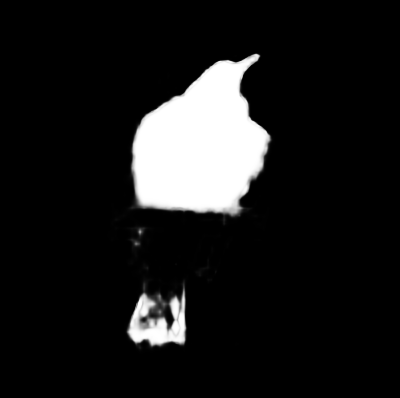}&
		\includegraphics[width=\wwidevisual, height=\hwidevisual]{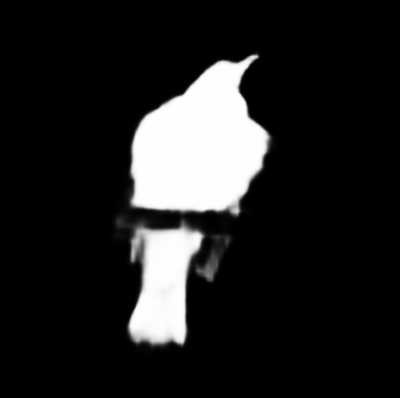}&
		\includegraphics[width=\wwidevisual, height=\hwidevisual]{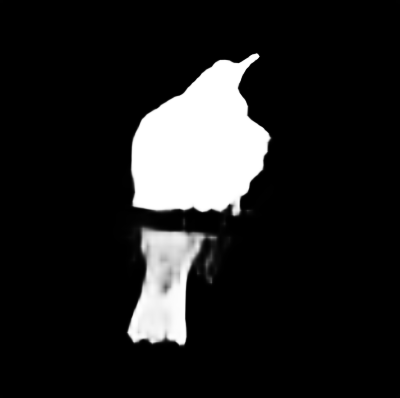}&
		\includegraphics[width=\wwidevisual, height=\hwidevisual]{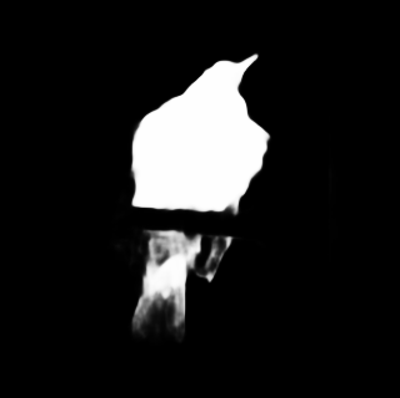}&
		\includegraphics[width=\wwidevisual, height=\hwidevisual]{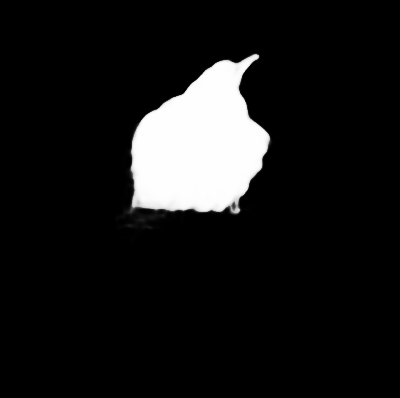}&
		\includegraphics[width=\wwidevisual, height=\hwidevisual]{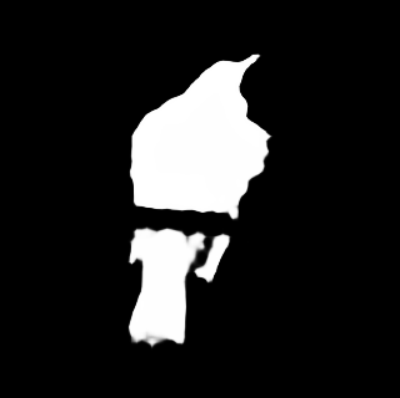}&
		\includegraphics[width=\wwidevisual, height=\hwidevisual]{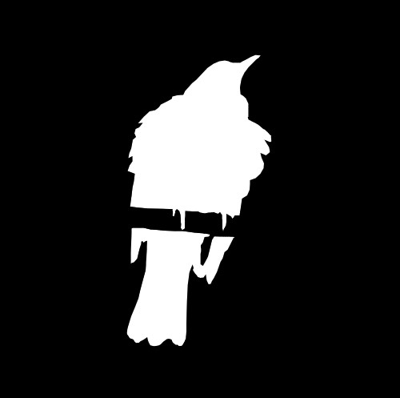} \\
		
		\includegraphics[width=\wwidevisual, height=\hwidevisual]{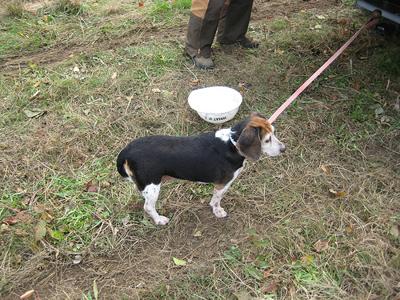}&
		\includegraphics[width=\wwidevisual, height=\hwidevisual]{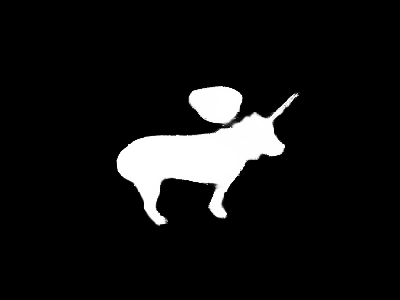}&
		\includegraphics[width=\wwidevisual, height=\hwidevisual]{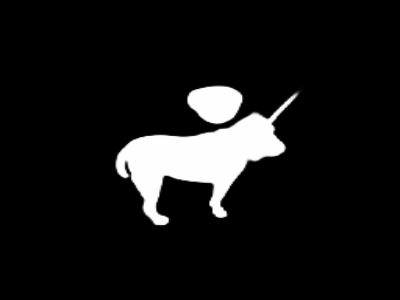}&
		\includegraphics[width=\wwidevisual, height=\hwidevisual]{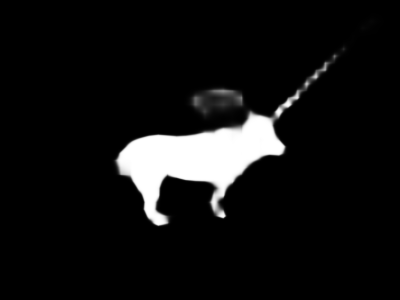}&
		\includegraphics[width=\wwidevisual, height=\hwidevisual]{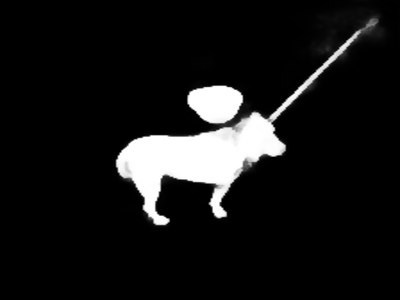}&
		\includegraphics[width=\wwidevisual, height=\hwidevisual]{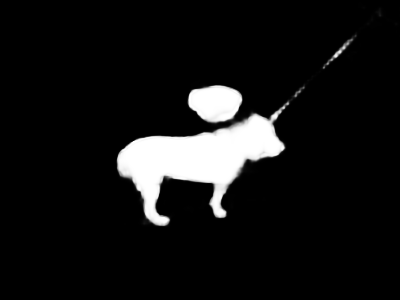}&
		\includegraphics[width=\wwidevisual, height=\hwidevisual]{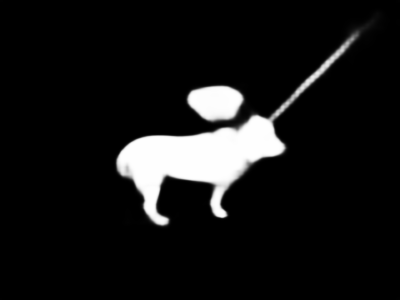}&
		\includegraphics[width=\wwidevisual, height=\hwidevisual]{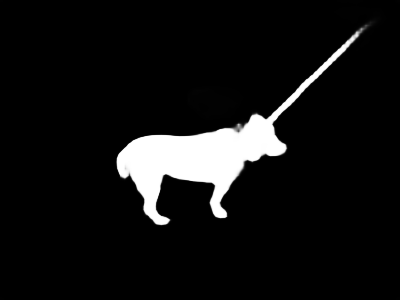}&
		\includegraphics[width=\wwidevisual, height=\hwidevisual]{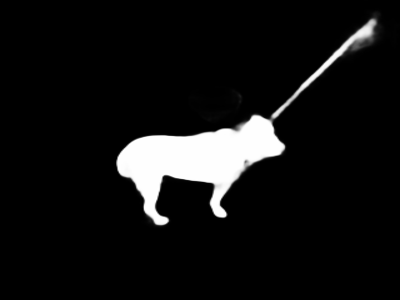}&
		\includegraphics[width=\wwidevisual, height=\hwidevisual]{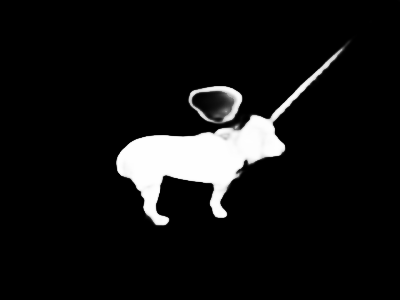}&
		\includegraphics[width=\wwidevisual, height=\hwidevisual]{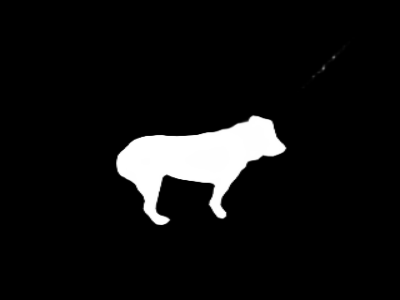}&
		\includegraphics[width=\wwidevisual, height=\hwidevisual]{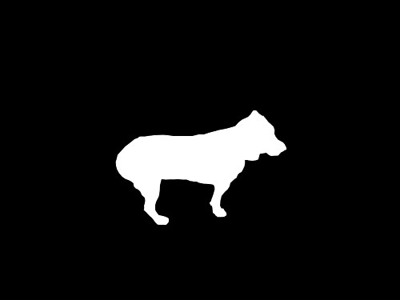} \\
		
		\includegraphics[width=\wwidevisual, height=\hwidevisual]{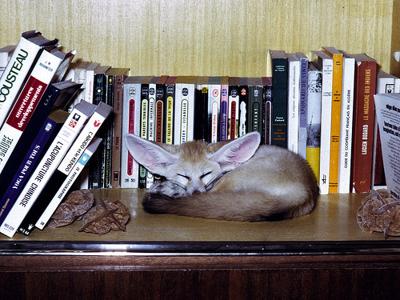}&
		\includegraphics[width=\wwidevisual, height=\hwidevisual]{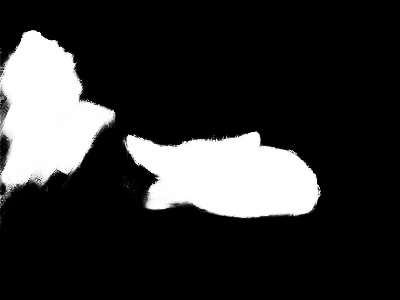}&
		\includegraphics[width=\wwidevisual, height=\hwidevisual]{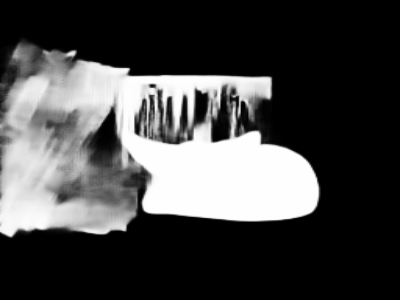}&
		\includegraphics[width=\wwidevisual, height=\hwidevisual]{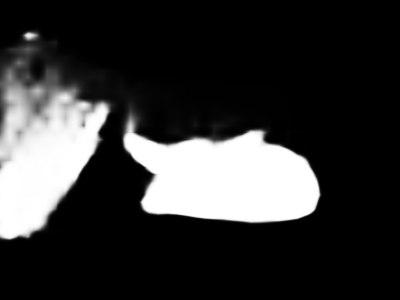}&
		\includegraphics[width=\wwidevisual, height=\hwidevisual]{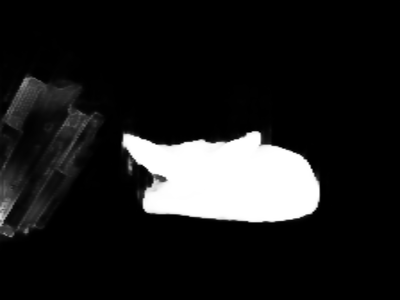}&
		\includegraphics[width=\wwidevisual, height=\hwidevisual]{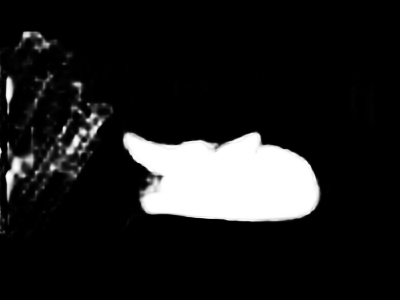}&
		\includegraphics[width=\wwidevisual, height=\hwidevisual]{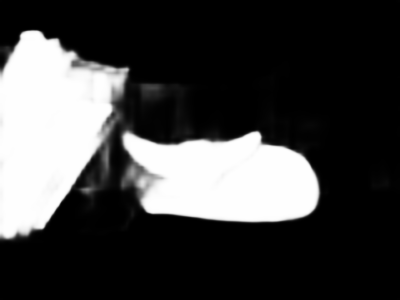}&
		\includegraphics[width=\wwidevisual, height=\hwidevisual]{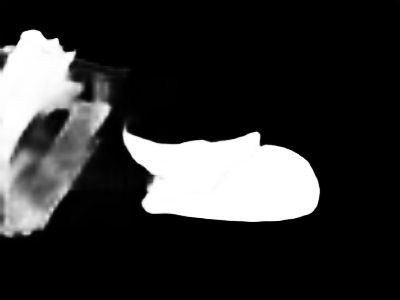}&
		\includegraphics[width=\wwidevisual, height=\hwidevisual]{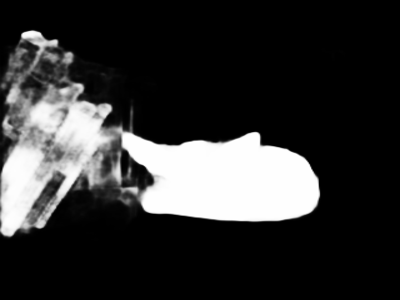}&
		\includegraphics[width=\wwidevisual, height=\hwidevisual]{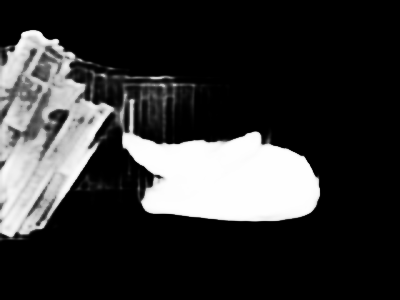}&
		\includegraphics[width=\wwidevisual, height=\hwidevisual]{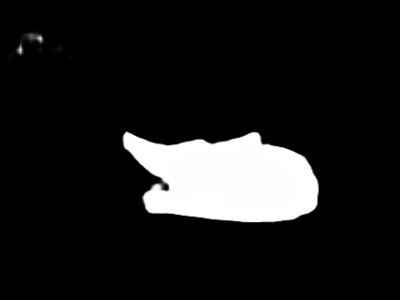}&
		\includegraphics[width=\wwidevisual, height=\hwidevisual]{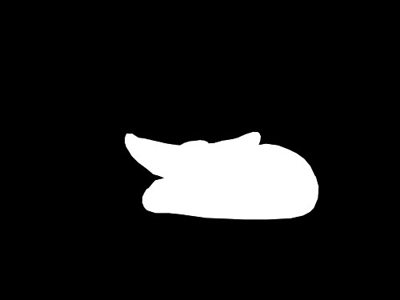} \\
		
		\includegraphics[width=\wtallvisual, height=\htallvisual]{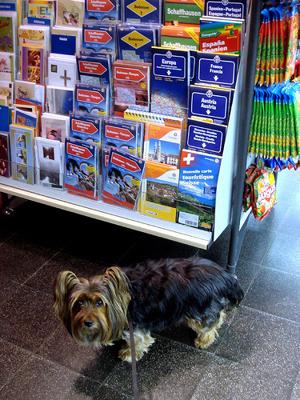}&
		\includegraphics[width=\wtallvisual, height=\htallvisual]{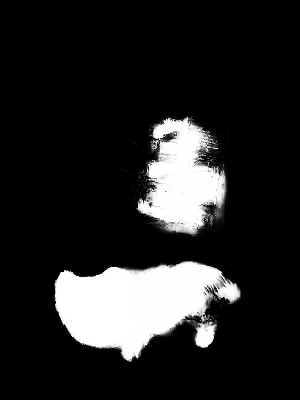}&
		\includegraphics[width=\wtallvisual, height=\htallvisual]{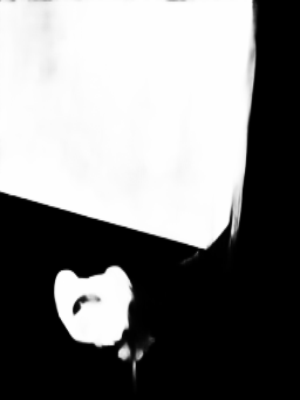}&
		\includegraphics[width=\wtallvisual, height=\htallvisual]{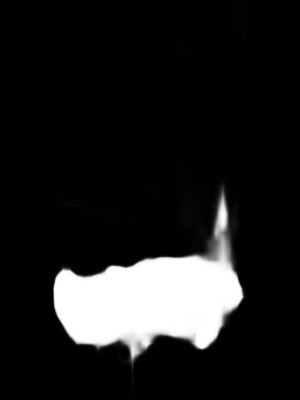}&
		\includegraphics[width=\wtallvisual, height=\htallvisual]{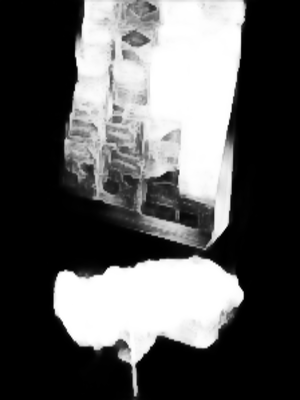}&
		\includegraphics[width=\wtallvisual, height=\htallvisual]{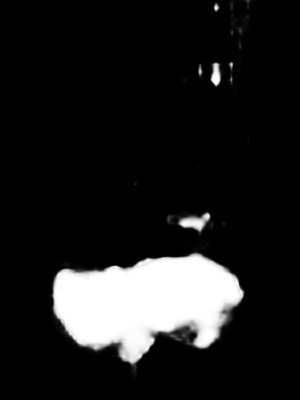}&
		\includegraphics[width=\wtallvisual, height=\htallvisual]{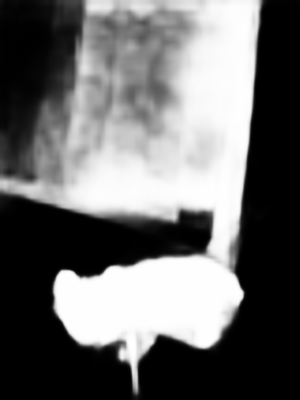}&
		\includegraphics[width=\wtallvisual, height=\htallvisual]{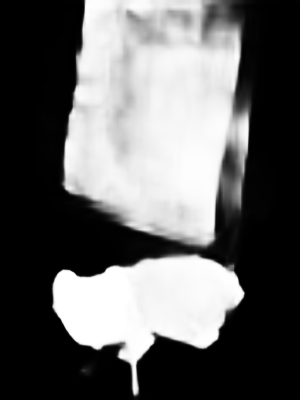}&
		\includegraphics[width=\wtallvisual, height=\htallvisual]{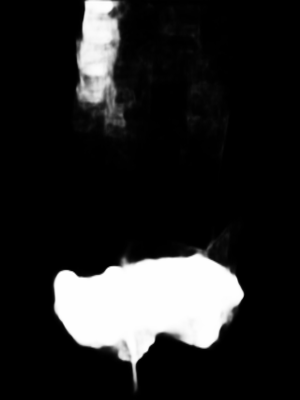}&
		\includegraphics[width=\wtallvisual, height=\htallvisual]{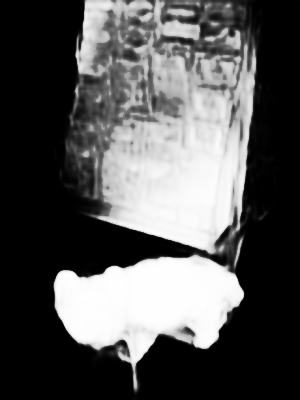}&
		\includegraphics[width=\wtallvisual, height=\htallvisual]{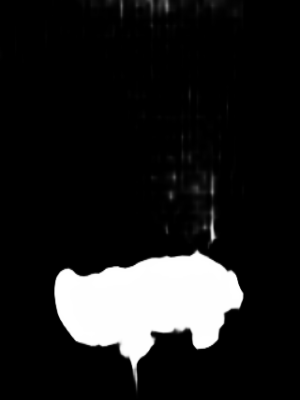}&
		\includegraphics[width=\wtallvisual, height=\htallvisual]{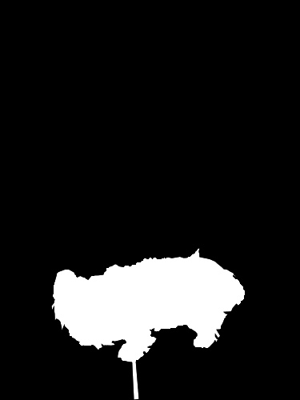} \\
		
		\scriptsize{Image} & \tabincell{c}{DSS\\\cite{HouPami19Dss}} & \tabincell{c}{BASNet\\\cite{Qin_2019_CVPR_basnet}} & \tabincell{c}{CPD\\\cite{Wu_2019_CVPR_cpd}} & \tabincell{c}{AFNet\\\cite{Feng_2019_CVPR_afnet}} & \tabincell{c}{BANet\\\cite{Su_2019_ICCV_banet}} & \tabincell{c}{GCPANet\\\cite{chen2020global}} &  \tabincell{c}{F3Net\\\cite{wei2020f3net}} & \tabincell{c}{MINet-R\\\cite{Pang_2020_CVPR_minet}} & \tabincell{c}{ITSD\\\cite{zhou2020interactive}} & \tabincell{c}{GDNet-B-S\\(ours)} & GT \\
		
	\end{tabular}
	\caption{\mhy{Visual comparison of GDNet-B-S with state-of-the-art SOD methods. We can see that GDNet-B-S is capable of segmenting salient objects of different scales in various scenes more accurately.}}
	\label{fig:sod}
\end{figure*}

\mhy{For a comprehensive measure on both precision and recall, F-measure ($F_\beta$) is computed by Eq. \ref{equ:f}. Following most salient object detection works \cite{Qin_2019_CVPR_basnet,Zhao_2019_ICCV_egnet}, we plot the F-measure curve from all precision-recall pairs under different thresholds. Recent studies \cite{fan2017structure,fan2018enhanced} have suggested that the weighted F-measure ($F_\beta^w$) \cite{Margolin_2014_CVPR} can provide more reliable evaluation results than the traditional $F_\beta$. Thus, we also consider this metric in the comparison.}

\mhy{Structure-measure ($S_\alpha$) focuses on evaluating the structural information of the saliency maps, and is closer to the human visual system than F-measure. Thus, we include structure-measure for a more comprehensive evaluation. Structure-measure ($S_\alpha$) is defined as:
	\begin{equation}\label{equ:s}
		S_\alpha = \alpha S_o + (1-\alpha)S_r,
	\end{equation}
	where $S_o$ and $S_r$ denote the object-aware and region-aware structural similarity, respectively. $\alpha$ is set as 0.5 by default. More details can be found in \cite{fan2017structure}.}

\mhy{Recently, Fan \emph{et al.} \cite{fan2018enhanced} propose a human visual perception based E-measure ($E_\phi$), which simultaneously evaluates pixel-level matching and image-level statistics. This metric can be used to assess the overall and localized accuracy of the salient object detection results.}

\subsection{Comparison with the State-of-the-arts}
\mhy{To demonstrate the effectiveness of GDNet-B-S, we compare it against 17 state-of-the-art SOD methods, including C2SNet \cite{Li_2018_ECCV_c2s}, RAS \cite{Chen_2018_ECCV_ras}, PAGRN \cite{Zhang_2018_CVPR_progressive}, DGRL \cite{Wang_2018_CVPR_dgrl}, R\textsuperscript{3}Net \cite{deng2018r3net}, BMPM \cite{Zhang_2018_CVPR_bmpm}, PiCANet-R \cite{Liu_2018_CVPR_picanet}, DSS \cite{HouPami19Dss}, BASNet \cite{Qin_2019_CVPR_basnet}, CPD \cite{Wu_2019_CVPR_cpd}, PAGE-Net \cite{Wang_2019_CVPR_pagenet}, AFNet \cite{Feng_2019_CVPR_afnet}, BANet \cite{Su_2019_ICCV_banet}, GCPANet \cite{chen2020global}, F3Net \cite{wei2020f3net}, MINet-R \cite{Pang_2020_CVPR_minet}, and ITSD \cite{zhou2020interactive}. For fair comparison, all saliency maps from the above methods are either provided by the authors or produced by running their released models. Besides, all saliency maps are evaluated with the same code.}

\mhy{Figure \ref{fig:curve} shows the precision-recall ($PR$) curves and the F-measure curves on the five benchmark datasets. We can see from these results that our GDNet-B-S (denoted by a solid red line) performs favorably against all the other methods. In addition, Table \ref{tab:sod} summarizes the comparison results in terms of the structure-measure ($S_\alpha$), the adaptive E-measure ($E_\phi^a$), the weighted F-measure ($F_\beta^w$), and the mean absolute error ($M$). As we can see, our method outperforms other state-of-the-art SOD methods on almost all five datasets with respect to all four metrics. This clearly demonstrates the superior performance of GDNet-B-S. To further illustrate the advantages of our method for salient object detection, we show some qualitative comparison results in Figure \ref{fig:sod}. We can see that GDNet-B-S is capable of accurately detecting small objects in the scenes (\emph{e.g.}, the first two rows) and segmenting out the complete salient objects from cluttered backgrounds (\emph{e.g.}, 3\emph{rd}, 4\emph{th}, 5\emph{th}, 6\emph{th}, and 7\emph{th} rows). While the state-of-the-art methods are typically confused by the messy background objects (\emph{e.g.}, 8-12\textit{th} rows) or the high similarities between foreground and background (\emph{e.g.}, the last two rows), GDNet-B-S successfully handles such ambiguities and avoids false positive prediction. This is mainly contributed to the large-field contextual feature learning, which could provide rich contextual information for context inference and salient object localization.}

\section{Discussion}\label{sec:discussion}
\mhy{In this section, we analysis the limitations of our method, and discuss the potential applications of glass detection and possible future research directions.}

\mhy{\textbf{Limitations.}}
\mhy{Although the proposed method has achieved state-of-the-art glass detection performance on the GDD testing set, it does have its limitations. In particular, the proposed method may fail in cases where the scene is very complex or provides insufficient contexts both inside and outside the glass, as shown in Figure \ref{fig:failure}. We believe that a deeper research is needed to study these extreme cases.}

\mhy{\textbf{Potential applications.}}
\mhy{Automatic glass detection has various possible applications. Here, we envision two potential ones: obstacle avoidance for drones and intelligence photography\add{/editing \cite{sheng2013video,li2012interactive}}. If a drone is equipped with the ability to detect glass, it can easily find out where the glass is and avoid crashing into the glass (\emph{e.g.}, Figure \ref{fig:app}(b)). In addition, if a phone camera is equipped with the ability to detect glass, it can identify the glass and then remove the reflections \cite{Yang_2018_ECCV_teaser} from the glass region to obtain the desired result (\emph{e.g.}, Figure \ref{fig:app}(d)).}

\mhy{\textbf{Future work.}}
\mhy{People may sometimes bump into glass, due to the transparency nature of glass. As the first attempt to address the glass detection problem with a computational method, we focus in this paper on detecting various types of glass in our daily life, from a single input image. This can benefit a wide range of applications. However, we have not addressed the extreme cases where even humans may not be able to identify where the glass is correctly. Studying these extreme cases can be an interesting future work.}
\mhy{Besides, it would also be interesting to study glass detection in videos and the instance glass detection problem. New techniques such as weakly supervised learning \cite{shen2018generative}, zero-shot learning \cite{zhu2018a}, and multi-scale backbone \cite{gao2019res2net} for glass detection may also be explored.}

\def\wfailure{0.45\linewidth}
\def\hfailure{.5in}
\begin{figure}
	\setlength{\tabcolsep}{2.5pt}
	\centering
	\begin{tabular}{cc}
		
		\includegraphics[width=\wfailure, height=\hfailure]{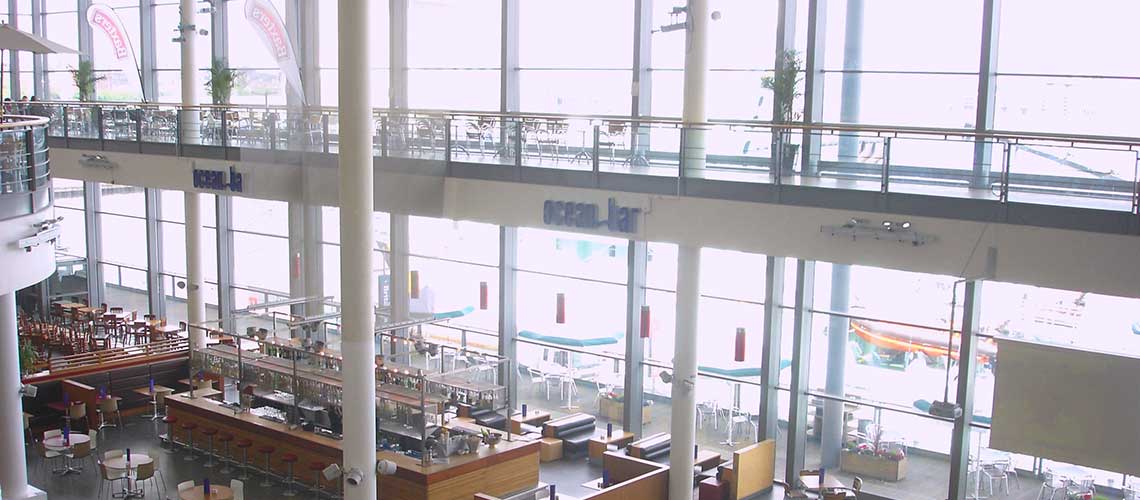}&
		\includegraphics[width=\wfailure, height=\hfailure]{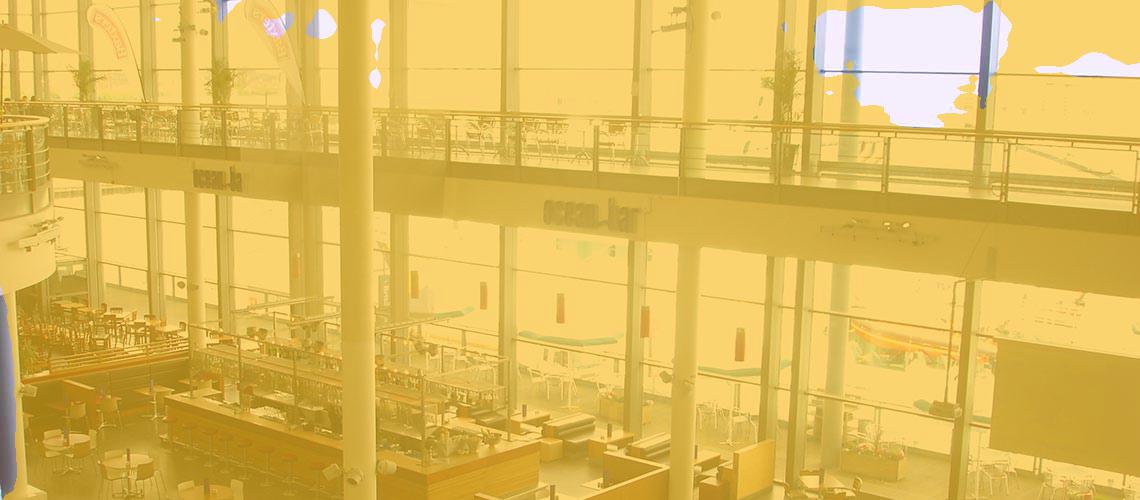} \\

		\includegraphics[width=\wfailure, height=\hfailure]{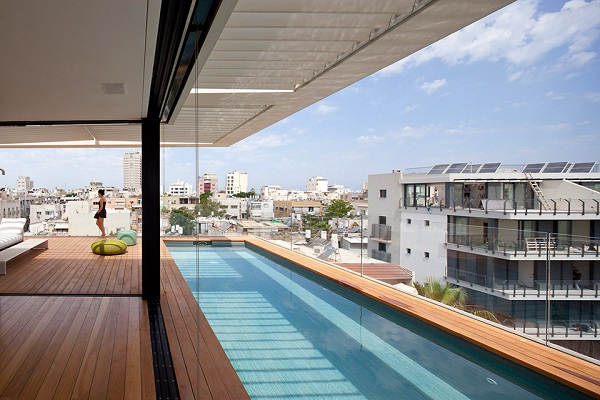}&
		\includegraphics[width=\wfailure, height=\hfailure]{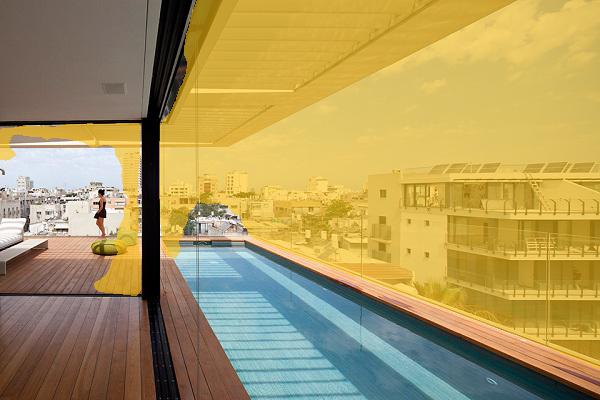} \\
		
		\small{input images} & \small{our results} \\
		
	\end{tabular}
	\caption{Failure cases. \mhy{Our glass detection method may fail in cases where the scene is extremely complicated (the top row) or contains insufficient contexts both inside and outside the glass (the bottom row).}}
	\label{fig:failure}
\end{figure}

\section{Conclusion}
In this paper, we have proposed an important problem of detecting glass from a single input image and contributed the first large-scale glass detection dataset (GDD) covering diverse scenes in our daily life. A novel network is proposed to address this challenging task. It leverages both high-level and low-level contexts extracted from a large field-of-view and the boundary cue to detect glass of different sizes in various scenes. Extensive evaluations on the images within and beyond the GDD testing set demonstrate the strong capability of our network to accurately detect glass. \mhy{We have further applied our network to the other vision tasks (\emph{i.e.}, mirror segmentation and salient object detection), validating the effectiveness and generalization capability of the proposed method. Finally, we have discussed some potential applications of glass detection and possible future research directions.}

\def\wapp{0.45\linewidth}
\def\happ{1in}
\begin{figure}
	\setlength{\tabcolsep}{2.5pt}
	\centering
	\begin{tabular}{cc}
		
		\includegraphics[width=\wapp, height=\happ]{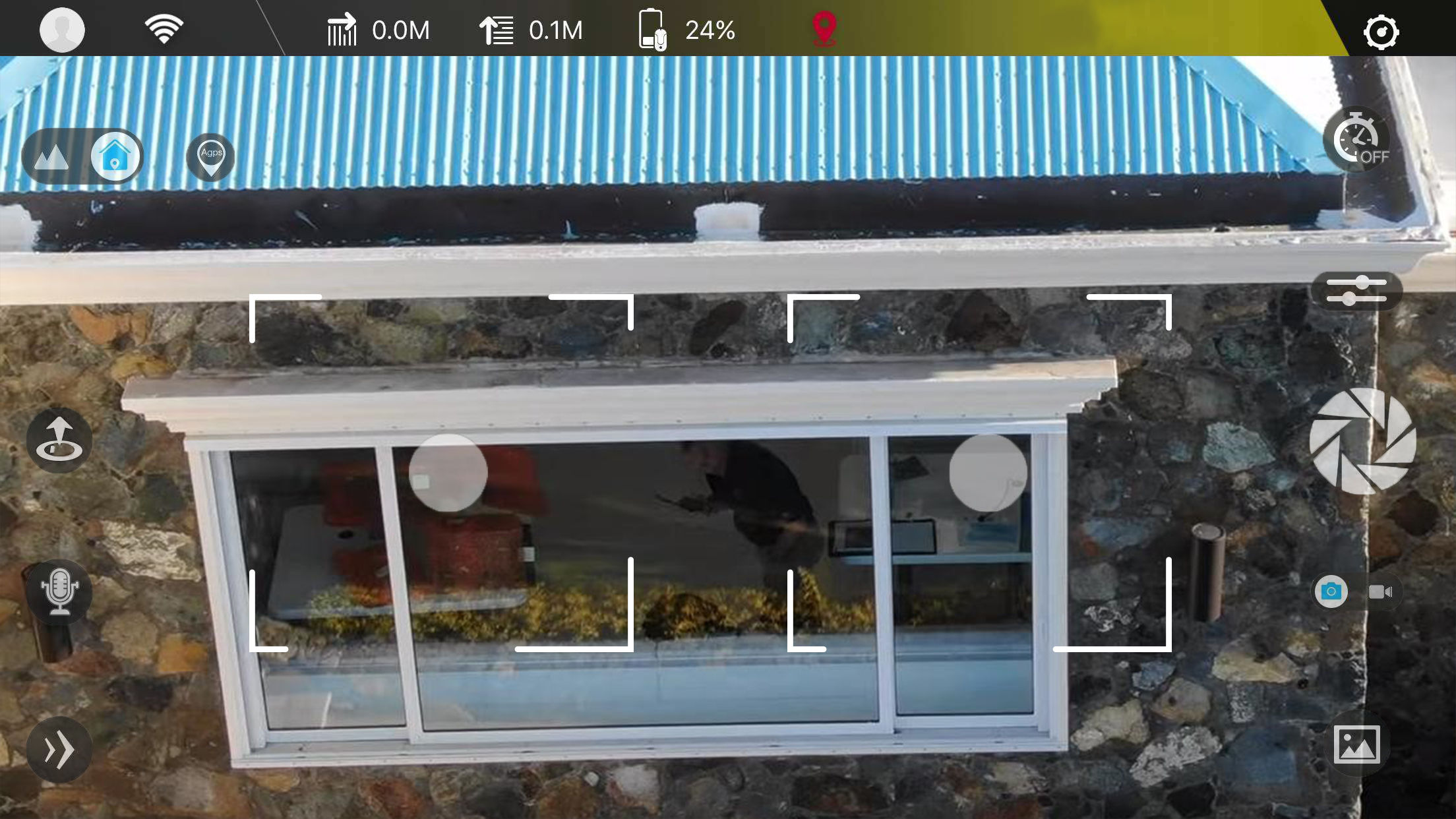}&
		\includegraphics[width=\wapp, height=\happ]{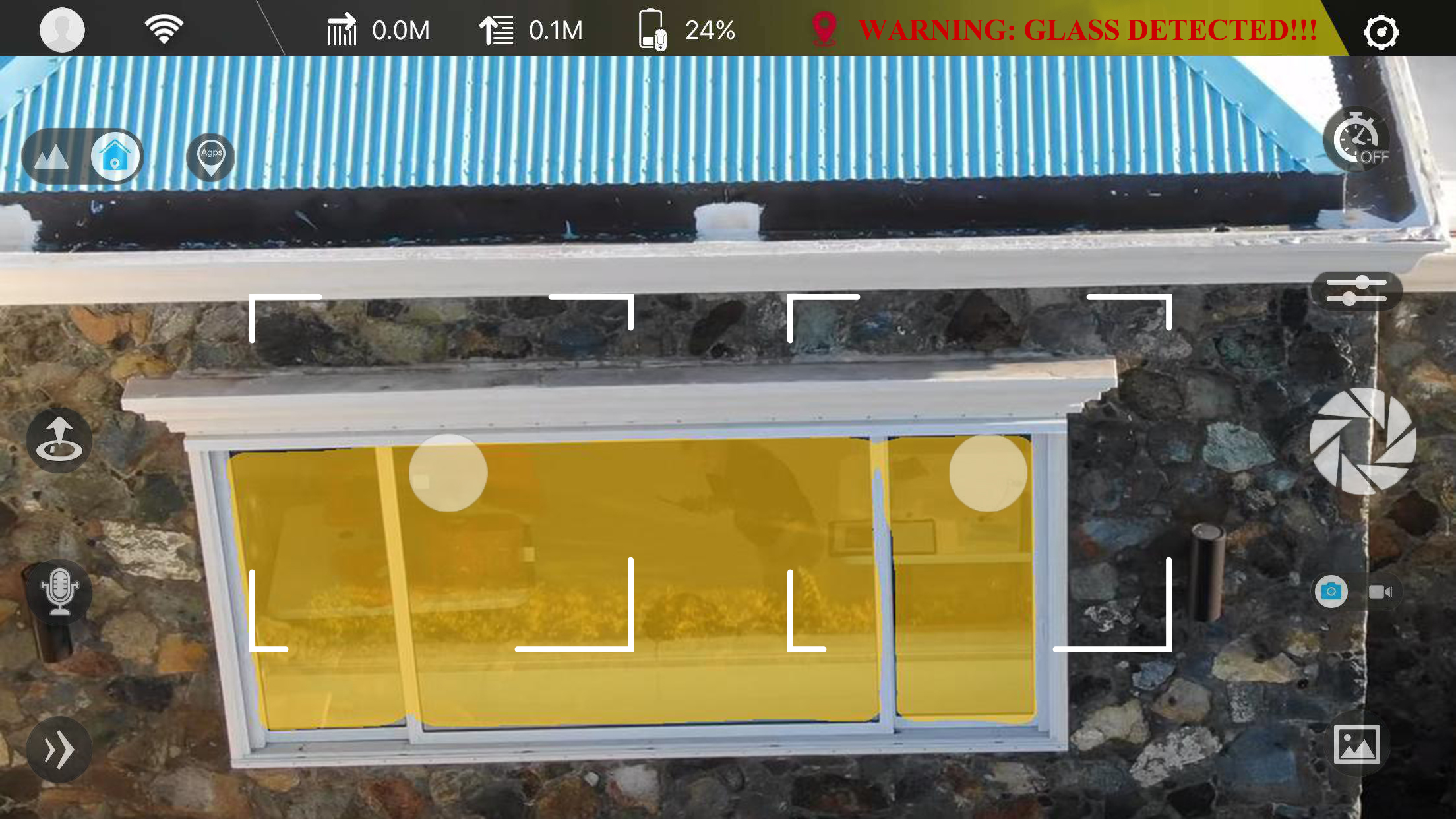} \\
		
		\small{(a) delivery drone w/o GDS} & \small{(b) delivery drone w/ GDS} \\

		\includegraphics[width=\wapp, height=\happ]{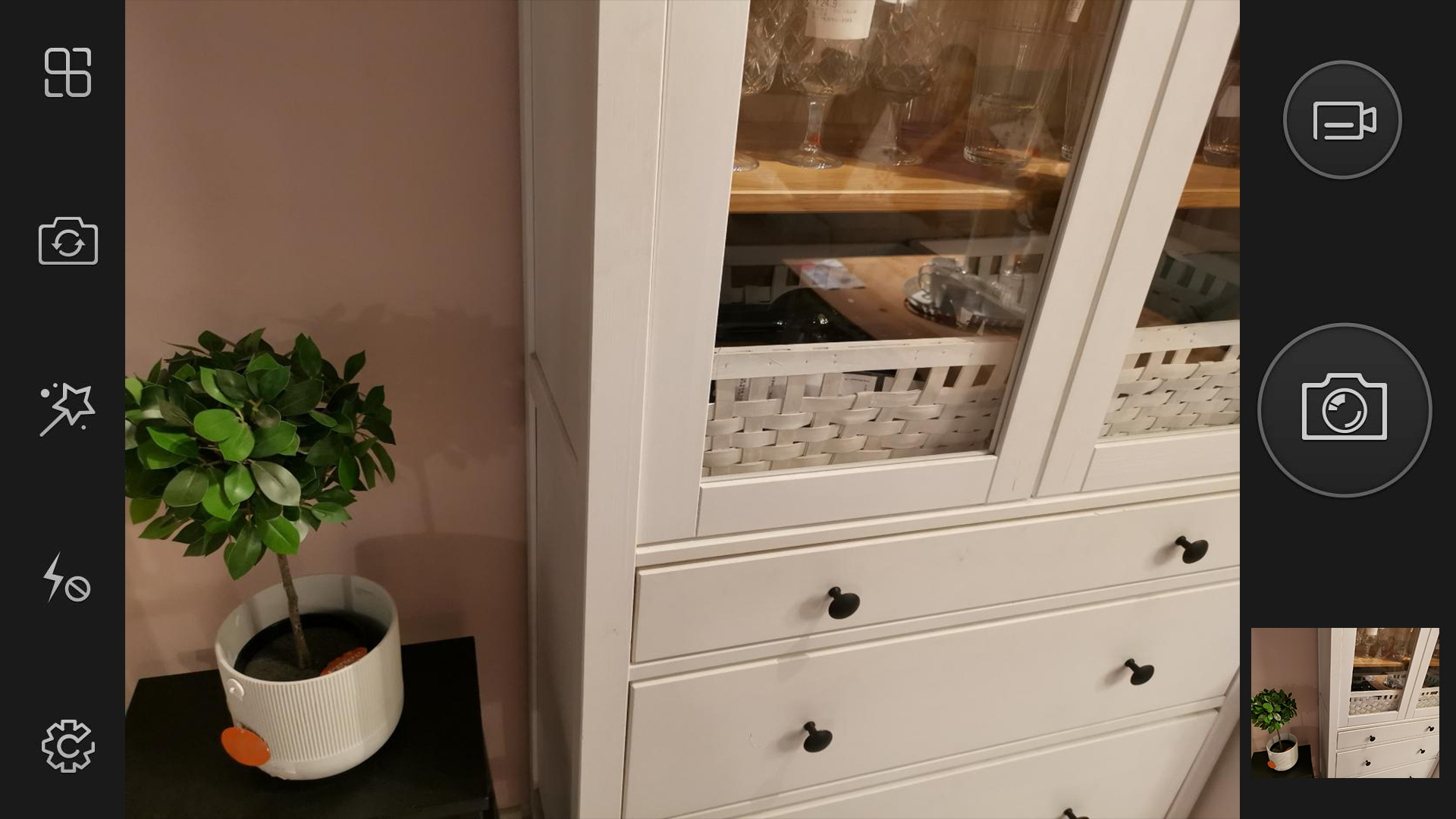}&
		\includegraphics[width=\wapp, height=\happ]{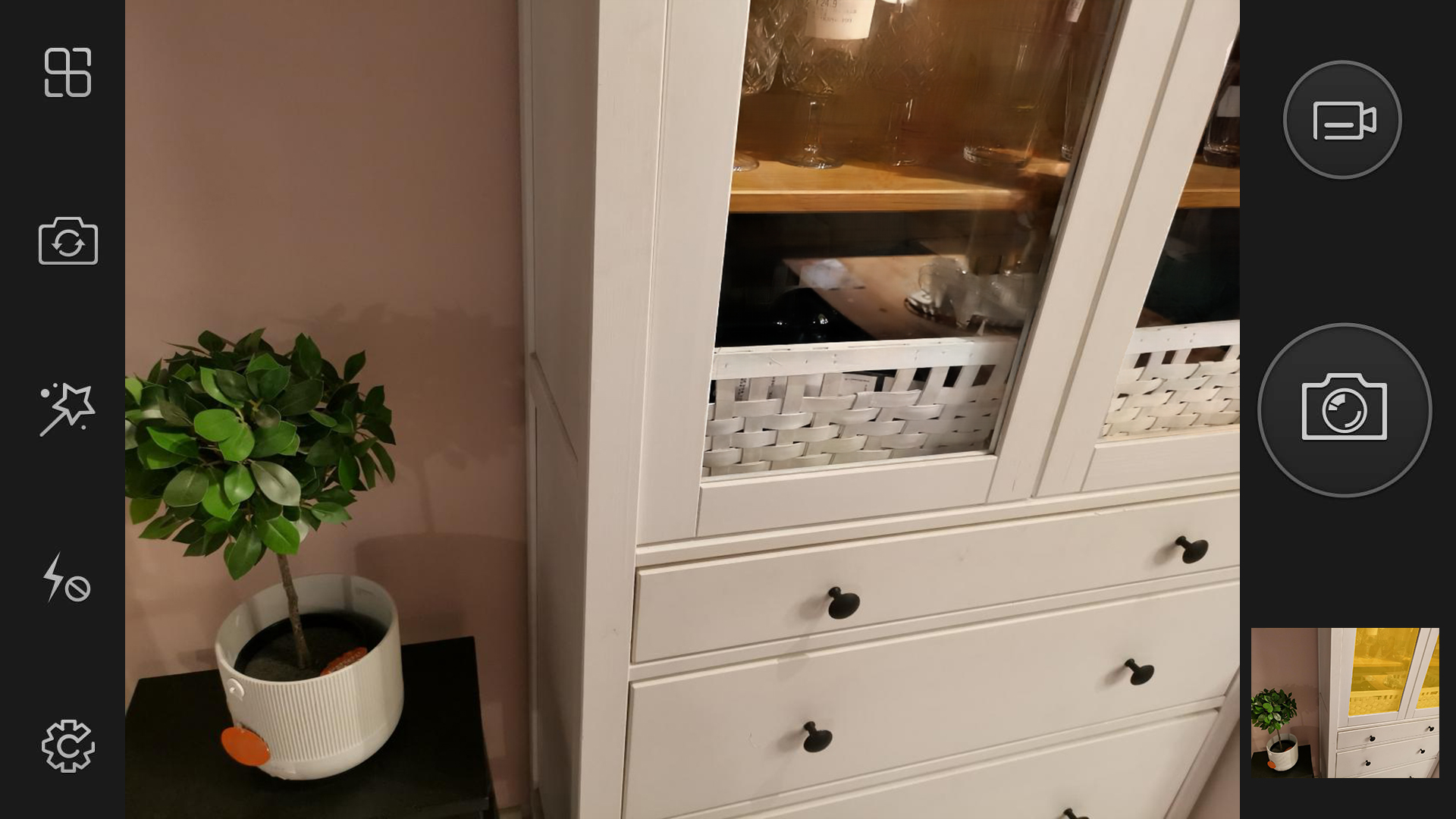} \\
		
		\small{(c) camera w/o GDS} & \small{(d) camera w/ GDS} \\
		
	\end{tabular}
	\caption{\mhy{Potential applications of glass detection. If a drone is equipped with the ability to detect glass, it may easily find out where the glass is and avoid crashing into the glass (b). If a phone camera is equipped with the ability to detect glass, it may identify the glass region and then remove the reflections \cite{Yang_2018_ECCV_teaser} only from the glass region to get the desired result (d).} }
	\label{fig:app}
\end{figure}

\ifCLASSOPTIONcompsoc
  \section*{Acknowledgments}
\else
  \section*{Acknowledgment}
\fi

This work was supported in part by the National Key Research and Development Program of China (No. 2022ZD0210500/2021ZD0112400), and the National Natural Science Foundation of China under Grant 61972067/U21A20491/U1908214, the Innovation Technology Funding of Dalian (2020JJ26GX036), the Research Grants Council of Hong Kong (Grant No.: 11205620), and a Strategic Research Grant from City University of Hong Kong (Ref.: 7005674).

\ifCLASSOPTIONcaptionsoff
  \newpage
\fi

\bibliographystyle{IEEEtran}
\bibliography{main}

\begin{thebibliography}{100}
\providecommand{\url}[1]{#1}
\csname url@samestyle\endcsname
\providecommand{\newblock}{\relax}
\providecommand{\bibinfo}[2]{#2}
\providecommand{\BIBentrySTDinterwordspacing}{\spaceskip=0pt\relax}
\providecommand{\BIBentryALTinterwordstretchfactor}{4}
\providecommand{\BIBentryALTinterwordspacing}{\spaceskip=\fontdimen2\font plus
\BIBentryALTinterwordstretchfactor\fontdimen3\font minus
  \fontdimen4\font\relax}
\providecommand{\BIBforeignlanguage}[2]{{%
\expandafter\ifx\csname l@#1\endcsname\relax
\typeout{** WARNING: IEEEtran.bst: No hyphenation pattern has been}%
\typeout{** loaded for the language `#1'. Using the pattern for}%
\typeout{** the default language instead.}%
\else
\language=\csname l@#1\endcsname
\fi
#2}}
\providecommand{\BIBdecl}{\relax}
\BIBdecl

\bibitem{Li_2018_CVPR_teaser}
Z.~Li and N.~Snavely, ``Megadepth: Learning single-view depth prediction from
  internet photos,'' in \emph{CVPR}, 2018.

\bibitem{He_2017_ICCV}
K.~He, G.~Gkioxari, P.~Dollar, and R.~Girshick, ``Mask r-cnn,'' in \emph{ICCV},
  2017.

\bibitem{Yang_2018_ECCV_teaser}
J.~Yang, D.~Gong, L.~Liu, and Q.~Shi, ``Seeing deeply and bidirectionally: A
  deep learning approach for single image reflection removal,'' in \emph{ECCV},
  2018.

\bibitem{Cai_2018_CVPR_cascadercnn}
Z.~Cai and N.~Vasconcelos, ``Cascade r-cnn: Delving into high quality object
  detection,'' in \emph{CVPR}, 2018.

\bibitem{huang2019mask_msrcnn}
Z.~Huang, L.~Huang, Y.~Gong, C.~Huang, and X.~Wang, ``Mask scoring r-cnn,'' in
  \emph{CVPR}, 2019.

\bibitem{Kalra_2020_CVPR}
A.~Kalra, V.~Taamazyan, S.~K. Rao, K.~Venkataraman, and R.~Raskar, ``Deep
  polarization cues for transparent object segmentation,'' in \emph{CVPR},
  2020.

\bibitem{liu2018picanet}
N.~Liu, J.~Han, and M.-H. Yang, ``Picanet: Learning pixel-wise contextual
  attention for saliency detection,'' in \emph{CVPR}, 2018.

\bibitem{qin2019basnet}
X.~Qin, Z.~Zhang, C.~Huang, C.~Gao, M.~Dehghan, and M.~Jagersand, ``Basnet:
  Boundary-aware salient object detection,'' in \emph{CVPR}, 2019.

\bibitem{yang2019mirrornet}
X.~Yang, H.~Mei, K.~Xu, X.~Wei, B.~Yin, and R.~W. Lau, ``Where is my mirror?''
  in \emph{ICCV}, 2019.

\bibitem{Mei_2020_CVPR}
H.~Mei, X.~Yang, Y.~Wang, Y.~Liu, S.~He, Q.~Zhang, X.~Wei, and R.~W. Lau,
  ``Don't hit me! glass detection in real-world scenes,'' in \emph{CVPR}, 2020.

\bibitem{Long_2015_CVPR}
J.~Long, E.~Shelhamer, and T.~Darrell, ``Fully convolutional networks for
  semantic segmentation,'' in \emph{CVPR}, 2015.

\bibitem{chen2017deeplab}
L.-C. Chen, G.~Papandreou, I.~Kokkinos, K.~Murphy, and A.~L. Yuille, ``Deeplab:
  Semantic image segmentation with deep convolutional nets, atrous convolution,
  and fully connected crfs,'' \emph{IEEE TPAMI}, 2017.

\bibitem{zhao2017pyramid}
H.~Zhao, J.~Shi, X.~Qi, X.~Wang, and J.~Jia, ``Pyramid scene parsing network,''
  in \emph{CVPR}, 2017.

\bibitem{Ding_2018_CVPR}
H.~Ding, X.~Jiang, B.~Shuai, A.~Qun~Liu, and G.~Wang, ``Context contrasted
  feature and gated multi-scale aggregation for scene segmentation,'' in
  \emph{CVPR}, 2018.

\bibitem{Zhang_2018_CVPR_enc}
H.~Zhang, K.~Dana, J.~Shi, Z.~Zhang, X.~Wang, A.~Tyagi, and A.~Agrawal,
  ``Context encoding for semantic segmentation,'' in \emph{CVPR}, 2018.

\bibitem{Fu_2019_CVPR}
J.~Fu, J.~Liu, H.~Tian, Y.~Li, Y.~Bao, Z.~Fang, and H.~Lu, ``Dual attention
  network for scene segmentation,'' in \emph{CVPR}, 2019.

\bibitem{Huang_2019_ICCV}
Z.~Huang, X.~Wang, L.~Huang, C.~Huang, Y.~Wei, and W.~Liu, ``Ccnet: Criss-cross
  attention for semantic segmentation,'' in \emph{ICCV}, 2019.

\bibitem{ren2017faster}
S.~Ren, K.~He, R.~Girshick, and J.~Sun, ``Faster r-cnn: towards real-time
  object detection with region proposal networks,'' \emph{IEEE TPAMI}, 2017.

\bibitem{liu2018path}
S.~Liu, L.~Qi, H.~Qin, J.~Shi, and J.~Jia, ``Path aggregation network for
  instance segmentation,'' in \emph{CVPR}, 2018.

\bibitem{achanta2009frequency}
R.~{Achanta}, S.~{Hemami}, F.~{Estrada}, and S.~{Susstrunk}, ``Frequency-tuned
  salient region detection,'' in \emph{CVPR}, 2009.

\bibitem{klein2011center}
D.~A. {Klein} and S.~{Frintrop}, ``Center-surround divergence of feature
  statistics for salient object detection,'' in \emph{ICCV}, 2011.

\bibitem{cheng2014global}
M.-M. Cheng, N.~J. Mitra, X.~Huang, P.~H. Torr, and S.-M. Hu, ``Global contrast
  based salient region detection,'' \emph{IEEE TPAMI}, 2014.

\bibitem{Liu_2016_CVPR}
N.~Liu and J.~Han, ``Dhsnet: Deep hierarchical saliency network for salient
  object detection,'' in \emph{CVPR}, 2016.

\bibitem{Zhang_2017_ICCV}
P.~Zhang, D.~Wang, H.~Lu, H.~Wang, and X.~Ruan, ``Amulet: Aggregating
  multi-level convolutional features for salient object detection,'' in
  \emph{ICCV}, 2017.

\bibitem{Zhang_2018_CVPR_progressive}
X.~Zhang, T.~Wang, J.~Qi, H.~Lu, and G.~Wang, ``Progressive attention guided
  recurrent network for salient object detection,'' in \emph{CVPR}, 2018.

\bibitem{Zhang_2018_CVPR_bd}
L.~Zhang, J.~Dai, H.~Lu, Y.~He, and G.~Wang, ``A bi-directional message passing
  model for salient object detection,'' in \emph{CVPR}, 2018.

\bibitem{Wang_2019_CVPR}
W.~Wang, J.~Shen, M.-M. Cheng, and L.~Shao, ``An iterative and cooperative
  top-down and bottom-up inference network for salient object detection,'' in
  \emph{CVPR}, 2019.

\bibitem{Hu_2018_CVPR}
X.~Hu, L.~Zhu, C.-W. Fu, J.~Qin, and P.-A. Heng, ``Direction-aware spatial
  context features for shadow detection,'' in \emph{CVPR}, 2018.

\bibitem{Zhu_2018_ECCV}
L.~Zhu, Z.~Deng, X.~Hu, C.-W. Fu, X.~Xu, J.~Qin, and P.-A. Heng,
  ``Bidirectional feature pyramid network with recurrent attention residual
  modules for shadow detection,'' in \emph{ECCV}, 2018.

\bibitem{zheng2019distraction}
Q.~Zheng, X.~Qiao, Y.~Cao, and R.~W. Lau, ``Distraction-aware shadow
  detection,'' in \emph{CVPR}, 2019.

\bibitem{Han_2018_ECCV}
X.~Han, C.~Nguyen, S.~You, and J.~Lu, ``Single image water hazard detection
  using fcn with reflection attention units,'' in \emph{ECCV}, 2018.

\bibitem{levin2007user}
A.~Levin and Y.~Weiss, ``User assisted separation of reflections from a single
  image using a sparsity prior,'' \emph{IEEE TPAMI}, 2007.

\bibitem{wan2016depth}
R.~Wan, B.~Shi, T.~A. Hwee, and A.~C. Kot, ``Depth of field guided reflection
  removal,'' in \emph{ICIP}, 2016.

\bibitem{Li_2014_CVPR}
Y.~Li and M.~S. Brown, ``Single image layer separation using relative
  smoothness,'' in \emph{CVPR}, 2014.

\bibitem{Shih_2015_CVPR}
Y.~Shih, D.~Krishnan, F.~Durand, and W.~T. Freeman, ``Reflection removal using
  ghosting cues,'' in \emph{CVPR}, 2015.

\bibitem{Fan_2017_ICCV}
Q.~Fan, J.~Yang, G.~Hua, B.~Chen, and D.~Wipf, ``A generic deep architecture
  for single image reflection removal and image smoothing,'' in \emph{ICCV},
  2017.

\bibitem{Wan_2018_CVPR}
R.~Wan, B.~Shi, L.-Y. Duan, A.-H. Tan, and A.~C. Kot, ``Crrn: Multi-scale
  guided concurrent reflection removal network,'' in \emph{CVPR}, 2018.

\bibitem{Zhang_2018_CVPR_sirr}
X.~Zhang, R.~Ng, and Q.~Chen, ``Single image reflection separation with
  perceptual losses,'' in \emph{CVPR}, 2018.

\bibitem{Wei_2019_CVPR}
K.~Wei, J.~Yang, Y.~Fu, D.~Wipf, and H.~Huang, ``Single image reflection
  removal exploiting misaligned training data and network enhancements,'' in
  \emph{CVPR}, 2019.

\bibitem{Qiao_2020_CVPR_Matting}
Y.~Qiao, Y.~Liu, X.~Yang, D.~Zhou, M.~Xu, Q.~Zhang, and X.~Wei,
  ``Attention-guided hierarchical structure aggregation for image matting,'' in
  \emph{CVPR}, 2020.

\bibitem{liu2021prior}
Y.~Liu, J.~Xie, Y.~Qiao, Y.~Tang, and X.~Yang, ``Prior-induced information
  alignment for image matting,'' \emph{IEEE TMM}, 2021.

\bibitem{Liu_2021_ICCV_Matting}
Y.~Liu, J.~Xie, X.~Shi, Y.~Qiao, Y.~Huang, Y.~Tang, and X.~Yang, ``Tripartite
  information mining and integration for image matting,'' in \emph{ICCV}, 2021.

\bibitem{qiao2021cpral}
Y.~Qiao, J.~Zhu, C.~Long, Z.~Zhang, Y.~Wang, Z.~Du, and X.~Yang, ``Cpral:
  Collaborative panoptic-regional active learning for semantic segmentation,''
  in \emph{AAAI}, 2022.

\bibitem{Zhao_2017_CVPR_pspnet}
H.~Zhao, J.~Shi, X.~Qi, X.~Wang, and J.~Jia, ``Pyramid scene parsing network,''
  in \emph{CVPR}, 2017.

\bibitem{liu2018receptive}
S.~Liu, D.~Huang \emph{et~al.}, ``Receptive field block net for accurate and
  fast object detection,'' in \emph{ECCV}, 2018.

\bibitem{He_2019_CVPR_adaptive}
J.~He, Z.~Deng, L.~Zhou, Y.~Wang, and Y.~Qiao, ``Adaptive pyramid context
  network for semantic segmentation,'' in \emph{CVPR}, 2019.

\bibitem{He_2019_ICCV_dynamic}
J.~He, Z.~Deng, and Y.~Qiao, ``Dynamic multi-scale filters for semantic
  segmentation,'' in \emph{ICCV}, 2019.

\bibitem{wu2016cartoon}
H.~Wu, Y.~Wu, S.~Zhang, P.~Li, and Z.~Wen, ``Cartoon image segmentation based
  on improved slic superpixels and adaptive region propagation merging,'' in
  \emph{ICSIP}, 2016.

\bibitem{yang2019drfn}
X.~{Yang}, H.~{Mei}, J.~{Zhang}, K.~{Xu}, B.~{Yin}, Q.~{Zhang}, and X.~{Wei},
  ``Drfn: Deep recurrent fusion network for single-image super-resolution with
  large factors,'' \emph{IEEE TMM}, 2019.

\bibitem{Zhang_2020_ICME}
J.~{Zhang}, C.~{Long}, Y.~{Wang}, X.~{Yang}, H.~{Mei}, and B.~{Yin},
  ``Multi-context and enhanced reconstruction network for single image super
  resolution,'' in \emph{ICME}, 2020.

\bibitem{nie2012compact}
Y.~Nie, C.~Xiao, H.~Sun, and P.~Li, ``Compact video synopsis via global
  spatiotemporal optimization,'' \emph{IEEE TVCG}, 2012.

\bibitem{tian2020weakly}
X.~Tian, K.~Xu, X.~Yang, B.~Yin, and R.~W. Lau, ``Weakly-supervised salient
  instance detection,'' in \emph{BMVC}, 2020.

\bibitem{tian2021learning}
------, ``Learning to detect instance-level salient objects using complementary
  image labels,'' \emph{IJCV}, 2021.

\bibitem{lin2018scn}
D.~Lin, R.~Zhang, Y.~Ji, P.~Li, and H.~Huang, ``Scn: switchable context network
  for semantic segmentation of rgb-d images,'' \emph{IEEE Transactions on
  Cybernetics}, 2018.

\bibitem{peng2017large}
C.~Peng, X.~Zhang, G.~Yu, G.~Luo, and J.~Sun, ``Large kernel matters--improve
  semantic segmentation by global convolutional network,'' in \emph{CVPR},
  2017.

\bibitem{Hu_2018_CVPR_dsc}
X.~Hu, L.~Zhu, C.-W. Fu, J.~Qin, and P.-A. Heng, ``Direction-aware spatial
  context features for shadow detection,'' in \emph{CVPR}, 2018.

\bibitem{wang2019spatial}
T.~Wang, X.~Yang, K.~Xu, S.~Chen, Q.~Zhang, and R.~W. Lau, ``Spatial attentive
  single-image deraining with a high quality real rain dataset,'' in
  \emph{CVPR}, 2019.

\bibitem{Ding_2018_CVPR_contrast}
H.~Ding, X.~Jiang, B.~Shuai, A.~Qun~Liu, and G.~Wang, ``Context contrasted
  feature and gated multi-scale aggregation for scene segmentation,'' in
  \emph{CVPR}, 2018.

\bibitem{Mei_2021_CVPR_PDNet}
H.~Mei, B.~Dong, W.~Dong, P.~Peers, X.~Yang, Q.~Zhang, and X.~Wei,
  ``Depth-aware mirror segmentation,'' in \emph{CVPR}, 2021.

\bibitem{Lin_2020_CVPR_PMD}
J.~Lin, G.~Wang, and R.~W. Lau, ``Progressive mirror detection,'' in
  \emph{CVPR}, 2020.

\bibitem{Lin_2021_CVPR_GSD}
J.~Lin, Z.~He, and R.~W. Lau, ``Rich context aggregation with reflection prior
  for glass surface detection,'' in \emph{CVPR}, 2021.

\bibitem{Wang_2018_CVPR_non-local}
X.~Wang, R.~Girshick, A.~Gupta, and K.~He, ``Non-local neural networks,'' in
  \emph{CVPR}, 2018.

\bibitem{Fu_2019_CVPR_danet}
J.~Fu, J.~Liu, H.~Tian, Y.~Li, Y.~Bao, Z.~Fang, and H.~Lu, ``Dual attention
  network for scene segmentation,'' in \emph{CVPR}, 2019.

\bibitem{Huang_2019_ICCV_ccnet}
Z.~Huang, X.~Wang, L.~Huang, C.~Huang, Y.~Wei, and W.~Liu, ``Ccnet: Criss-cross
  attention for semantic segmentation,'' in \emph{ICCV}, 2019.

\bibitem{Mei_2021_CVPR_PFNet}
H.~Mei, G.-P. Ji, Z.~Wei, X.~Yang, X.~Wei, and D.-P. Fan, ``Camouflaged object
  segmentation with distraction mining,'' in \emph{CVPR}, 2021.

\bibitem{xie2021segmenting}
E.~Xie, W.~Wang, W.~Wang, P.~Sun, H.~Xu, D.~Liang, and P.~Luo, ``Segmenting
  transparent object in the wild with transformer,'' in \emph{IJCAI}, 2021.

\bibitem{li2018non}
G.~{Li}, X.~{He}, W.~{Zhang}, H.~{Chang}, L.~{Dong}, and L.~{Lin},
  ``Non-locally enhanced encoder-decoder network for single image de-raining,''
  in \emph{ACM MM}, 2018.

\bibitem{liu2018non}
D.~Liu, B.~Wen, Y.~Fan, C.~C. Loy, and T.~S. Huang, ``Non-local recurrent
  network for image restoration,'' in \emph{NeurIPS}, 2018.

\bibitem{Dai_2019_CVPR_second-order}
T.~Dai, J.~Cai, Y.~Zhang, S.-T. Xia, and L.~Zhang, ``Second-order attention
  network for single image super-resolution,'' in \emph{CVPR}, 2019.

\bibitem{Cordts_2016_CVPR_cityscape}
M.~Cordts, M.~Omran, S.~Ramos, T.~Rehfeld, M.~Enzweiler, R.~Benenson,
  U.~Franke, S.~Roth, and B.~Schiele, ``The cityscapes dataset for semantic
  urban scene understanding,'' in \emph{CVPR}, 2016.

\bibitem{everingham2015the}
M.~{Everingham}, S.~M. {Eslami}, L.~{Gool}, C.~K. {Williams}, J.~{Winn}, and
  A.~{Zisserman}, ``The pascal visual object classes challenge: A
  retrospective,'' \emph{IJCV}, 2015.

\bibitem{Zhou_2017_CVPR}
B.~Zhou, H.~Zhao, X.~Puig, S.~Fidler, A.~Barriuso, and A.~Torralba, ``Scene
  parsing through ade20k dataset,'' in \emph{CVPR}, 2017.

\bibitem{tan2021night}
X.~Tan, K.~Xu, Y.~Cao, Y.~Zhang, L.~Ma, and R.~W. Lau, ``Night-time scene
  parsing with a large real dataset,'' \emph{IEEE TIP}, 2021.

\bibitem{deng2009imagenet}
J.~{Deng}, W.~{Dong}, R.~{Socher}, L.-J. {Li}, K.~{Li}, and L.~{Fei-Fei},
  ``Imagenet: A large-scale hierarchical image database,'' in \emph{CVPR},
  2009.

\bibitem{Xie_2017_CVPR}
S.~Xie, R.~Girshick, P.~Dollar, Z.~Tu, and K.~He, ``Aggregated residual
  transformations for deep neural networks,'' in \emph{CVPR}, 2017.

\bibitem{Woo_2018_ECCV}
S.~Woo, J.~Park, J.-Y. Lee, and I.~So~Kweon, ``Cbam: Convolutional block
  attention module,'' in \emph{ECCV}, 2018.

\bibitem{Wang_2018_CVPR}
X.~Wang, R.~Girshick, A.~Gupta, and K.~He, ``Non-local neural networks,'' in
  \emph{CVPR}, 2018.

\bibitem{Su_2019_ICCV}
J.~Su, J.~Li, Y.~Zhang, C.~Xia, and Y.~Tian, ``Selectivity or invariance:
  Boundary-aware salient object detection,'' in \emph{ICCV}, 2019.

\bibitem{xie2020segmenting}
E.~Xie, W.~Wang, W.~Wang, M.~Ding, C.~Shen, and P.~Luo, ``Segmenting
  transparent objects in the wild,'' in \emph{ECCV}, 2020.

\bibitem{Su_2019_ICCV_banet}
J.~Su, J.~Li, Y.~Zhang, C.~Xia, and Y.~Tian, ``Selectivity or invariance:
  Boundary-aware salient object detection,'' in \emph{ICCV}, 2019.

\bibitem{boer2005a}
P.-T. de~{Boer}, D.~P. {Kroese}, S.~{Mannor}, and R.~Y. {Rubinstein}, ``A
  tutorial on the cross-entropy method,'' \emph{Annals of Operations Research},
  2005.

\bibitem{Zhao_2019_CVPR}
T.~Zhao and X.~Wu, ``Pyramid feature attention network for saliency
  detection,'' in \emph{CVPR}, 2019.

\bibitem{paszke2019pytorch}
A.~Paszke, S.~Gross, F.~Massa, A.~Lerer, J.~Bradbury, G.~Chanan, T.~Killeen,
  Z.~Lin, N.~Gimelshein, L.~Antiga \emph{et~al.}, ``Pytorch: An imperative
  style, high-performance deep learning library,'' in \emph{NeurIPS}, 2019.

\bibitem{liu2015parsenet}
W.~Liu, A.~Rabinovich, and A.~C. Berg, ``Parsenet: Looking wider to see
  better,'' \emph{arXiv:1506.04579}, 2015.

\bibitem{krahenbuhl2011efficient}
P.~Kr{\"a}henb{\"u}hl and V.~Koltun, ``Efficient inference in fully connected
  crfs with gaussian edge potentials,'' in \emph{NeurIPS}, 2011.

\bibitem{Zhao_2018_ECCV_icnet}
H.~Zhao, X.~Qi, X.~Shen, J.~Shi, and J.~Jia, ``Icnet for real-time semantic
  segmentation on high-resolution images,'' in \emph{ECCV}, 2018.

\bibitem{Yang_2018_CVPR}
M.~Yang, K.~Yu, C.~Zhang, Z.~Li, and K.~Yang, ``Denseaspp for semantic
  segmentation in street scenes,'' in \emph{CVPR}, 2018.

\bibitem{Yu_2018_ECCV}
C.~Yu, J.~Wang, C.~Peng, C.~Gao, G.~Yu, and N.~Sang, ``Bisenet: Bilateral
  segmentation network for real-time semantic segmentation,'' in \emph{ECCV},
  2018.

\bibitem{Zhao_2018_ECCV_psanet}
H.~Zhao, Y.~Zhang, S.~Liu, J.~Shi, C.~Change~Loy, D.~Lin, and J.~Jia, ``Psanet:
  Point-wise spatial attention network for scene parsing,'' in \emph{ECCV},
  2018.

\bibitem{HouPami19Dss}
Q.~Hou, M.-M. Cheng, X.~Hu, A.~Borji, Z.~Tu, and P.~Torr, ``Deeply supervised
  salient object detection with short connections,'' \emph{IEEE TPAMI}, 2019.

\bibitem{Chen_2018_ECCV}
S.~Chen, X.~Tan, B.~Wang, and X.~Hu, ``Reverse attention for salient object
  detection,'' in \emph{ECCV}, 2018.

\bibitem{deng2018r3net}
Z.~Deng, X.~Hu, L.~Zhu, X.~Xu, J.~Qin, G.~Han, and P.-A. Heng, ``R3net:
  Recurrent residual refinement network for saliency detection,'' in
  \emph{IJCAI}, 2018.

\bibitem{Wu_2019_CVPR}
Z.~Wu, L.~Su, and Q.~Huang, ``Cascaded partial decoder for fast and accurate
  salient object detection,'' in \emph{CVPR}, 2019.

\bibitem{Liu_2019_CVPR}
J.-J. Liu, Q.~Hou, M.-M. Cheng, J.~Feng, and J.~Jiang, ``A simple pooling-based
  design for real-time salient object detection,'' in \emph{CVPR}, 2019.

\bibitem{zhao2019EGNet}
J.-X. Zhao, J.-J. Liu, D.-P. Fan, Y.~Cao, J.~Yang, and M.-M. Cheng,
  ``Egnet:edge guidance network for salient object detection,'' in \emph{ICCV},
  2019.

\bibitem{he2016deep_resnet}
K.~He, X.~Zhang, S.~Ren, and J.~Sun, ``Deep residual learning for image
  recognition,'' in \emph{CVPR}, 2016.

\bibitem{hu2018squeeze}
J.~Hu, L.~Shen, and G.~Sun, ``Squeeze-and-excitation networks,'' in
  \emph{CVPR}, 2018.

\bibitem{Matterport3D}
A.~Chang, A.~Dai, T.~Funkhouser, M.~Halber, M.~Niessner, M.~Savva, S.~Song,
  A.~Zeng, and Y.~Zhang, ``{Matterport3D}: Learning from {RGB-D} data in indoor
  environments,'' in \emph{3DV}, 2017.

\bibitem{armeni2017joint}
I.~Armeni, S.~Sax, A.~R. Zamir, and S.~Savarese, ``Joint 2d-3d-semantic data
  for indoor scene understanding,'' \emph{arXiv:1702.01105}, 2017.

\bibitem{Li_2014_CVPR_pascals}
Y.~Li, X.~Hou, C.~Koch, J.~M. Rehg, and A.~L. Yuille, ``The secrets of salient
  object segmentation,'' in \emph{CVPR}, 2014.

\bibitem{Yang_2013_ICCV_Workshops}
C.~Yang, L.~Zhang, H.~Lu, X.~Ruan, and M.-H. Yang, ``Saliency detection via
  graph-based manifold ranking,'' in \emph{CVPR}, 2013.

\bibitem{Yan_2013_ICCV_Workshops}
Q.~Yan, L.~Xu, J.~Shi, and J.~Jia, ``Hierarchical saliency detection,'' in
  \emph{CVPR}, 2013.

\bibitem{Li_2015_CVPR_hkuis}
G.~Li and Y.~Yu, ``Visual saliency based on multiscale deep features,'' in
  \emph{CVPR}, 2015.

\bibitem{Wang_2017_CVPR_duts}
L.~Wang, H.~Lu, Y.~Wang, M.~Feng, D.~Wang, B.~Yin, and X.~Ruan, ``Learning to
  detect salient objects with image-level supervision,'' in \emph{CVPR}, 2017.

\bibitem{Qin_2019_CVPR_basnet}
X.~Qin, Z.~Zhang, C.~Huang, C.~Gao, M.~Dehghan, and M.~Jagersand, ``Basnet:
  Boundary-aware salient object detection,'' in \emph{CVPR}, 2019.

\bibitem{Zhao_2019_CVPR_pfan}
T.~Zhao and X.~Wu, ``Pyramid feature attention network for saliency
  detection,'' in \emph{CVPR}, 2019.

\bibitem{Wu_2019_CVPR_cpd}
Z.~Wu, L.~Su, and Q.~Huang, ``Cascaded partial decoder for fast and accurate
  salient object detection,'' in \emph{CVPR}, 2019.

\bibitem{Zhao_2019_ICCV_egnet}
J.-X. Zhao, J.-J. Liu, D.-P. Fan, Y.~Cao, J.~Yang, and M.-M. Cheng, ``Egnet:
  Edge guidance network for salient object detection,'' in \emph{ICCV}, 2019.

\bibitem{mei2021exploring}
H.~Mei, Y.~Liu, Z.~Wei, D.~Zhou, X.~Xiaopeng, Q.~Zhang, and X.~Yang,
  ``Exploring dense context for salient object detection,'' \emph{IEEE TCSVT},
  2021.

\bibitem{Li_2018_ECCV_c2s}
X.~Li, F.~Yang, H.~Cheng, W.~Liu, and D.~Shen, ``Contour knowledge transfer for
  salient object detection,'' in \emph{ECCV}, 2018.

\bibitem{Chen_2018_ECCV_ras}
S.~Chen, X.~Tan, B.~Wang, and X.~Hu, ``Reverse attention for salient object
  detection,'' in \emph{ECCV}, 2018.

\bibitem{Wang_2018_CVPR_dgrl}
T.~Wang, L.~Zhang, S.~Wang, H.~Lu, G.~Yang, X.~Ruan, and A.~Borji, ``Detect
  globally, refine locally: A novel approach to saliency detection,'' in
  \emph{CVPR}, 2018.

\bibitem{Zhang_2018_CVPR_bmpm}
L.~Zhang, J.~Dai, H.~Lu, Y.~He, and G.~Wang, ``A bi-directional message passing
  model for salient object detection,'' in \emph{CVPR}, 2018.

\bibitem{Liu_2018_CVPR_picanet}
N.~Liu, J.~Han, and M.-H. Yang, ``Picanet: Learning pixel-wise contextual
  attention for saliency detection,'' in \emph{CVPR}, 2018.

\bibitem{Wang_2019_CVPR_pagenet}
W.~Wang, S.~Zhao, J.~Shen, S.~C.~H. Hoi, and A.~Borji, ``Salient object
  detection with pyramid attention and salient edges,'' in \emph{CVPR}, 2019.

\bibitem{Feng_2019_CVPR_afnet}
M.~Feng, H.~Lu, and E.~Ding, ``Attentive feedback network for boundary-aware
  salient object detection,'' in \emph{CVPR}, 2019.

\bibitem{chen2020global}
Z.~Chen, Q.~Xu, R.~Cong, and Q.~Huang, ``Global context-aware progressive
  aggregation network for salient object detection,'' in \emph{AAAI}, 2020.

\bibitem{wei2020f3net}
J.~{Wei}, S.~{Wang}, and Q.~{Huang}, ``F3net: Fusion, feedback and focus for
  salient object detection,'' in \emph{AAAI}, 2020.

\bibitem{Pang_2020_CVPR_minet}
Y.~Pang, X.~Zhao, L.~Zhang, and H.~Lu, ``Multi-scale interactive network for
  salient object detection,'' in \emph{CVPR}, 2020.

\bibitem{zhou2020interactive}
H.~Zhou, X.~Xie, J.-H. Lai, Z.~Chen, and L.~Yang, ``Interactive two-stream
  decoder for accurate and fast saliency detection,'' in \emph{CVPR}, 2020.

\bibitem{fan2017structure}
D.-P. Fan, M.-M. Cheng, Y.~Liu, T.~Li, and A.~Borji, ``Structure-measure: A new
  way to evaluate foreground maps,'' in \emph{ICCV}, 2017.

\bibitem{fan2018enhanced}
D.-P. {Fan}, C.~{Gong}, Y.~{Cao}, B.~{Ren}, M.-M. {Cheng}, and A.~{Borji},
  ``Enhanced-alignment measure for binary foreground map evaluation,'' in
  \emph{IJCAI}, 2018.

\bibitem{Margolin_2014_CVPR}
R.~Margolin, L.~Zelnik-Manor, and A.~Tal, ``How to evaluate foreground maps?''
  in \emph{CVPR}, 2014.

\bibitem{sheng2013video}
B.~Sheng, H.~Sun, M.~Magnor, and P.~Li, ``Video colorization using parallel
  optimization in feature space,'' \emph{IEEE TCSVT}, 2013.

\bibitem{li2012interactive}
P.~Li, H.~Sun, C.~Huang, J.~Shen, and Y.~Nie, ``Interactive image/video
  retexturing using gpu parallelism,'' \emph{Computers \& Graphics}, 2012.

\bibitem{shen2018generative}
Y.~{Shen}, R.~{Ji}, S.~{Zhang}, W.~{Zuo}, and Y.~{Wang}, ``Generative
  adversarial learning towards fast weakly supervised detection,'' in
  \emph{CVPR}, 2018.

\bibitem{zhu2018a}
Y.~{Zhu}, M.~{Elhoseiny}, B.~{Liu}, X.~{Peng}, and A.~{Elgammal}, ``A
  generative adversarial approach for zero-shot learning from noisy texts,'' in
  \emph{CVPR}, 2018.

\bibitem{gao2019res2net}
S.~{Gao}, M.-M. {Cheng}, K.~{Zhao}, X.-Y. {Zhang}, M.-H. {Yang}, and P.~H.
  {Torr}, ``Res2net: A new multi-scale backbone architecture.'' \emph{IEEE
  TPAMI}, 2019.

\end{thebibliography}

\begin{IEEEbiography}[{\includegraphics[width=1in,height=1.2in,clip]{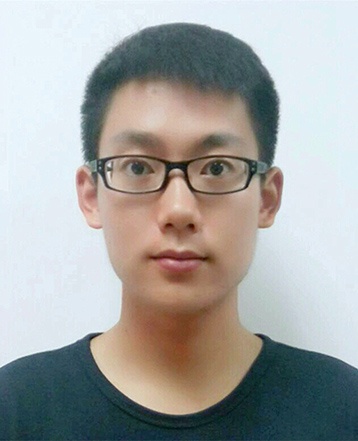}}]{Haiyang Mei}
	received the B.Eng. degree in the Automation from Dalian University of Technology, Dalian China, in 2017. He is currently working toward the Ph.D. degree with the Department of Computer Science and Technology, Dalian University of Technology, Dalian, China. His research interests include image processing and computer vision.
\end{IEEEbiography}

\begin{IEEEbiography}[{\includegraphics[width=1in,height=1.3in,clip]{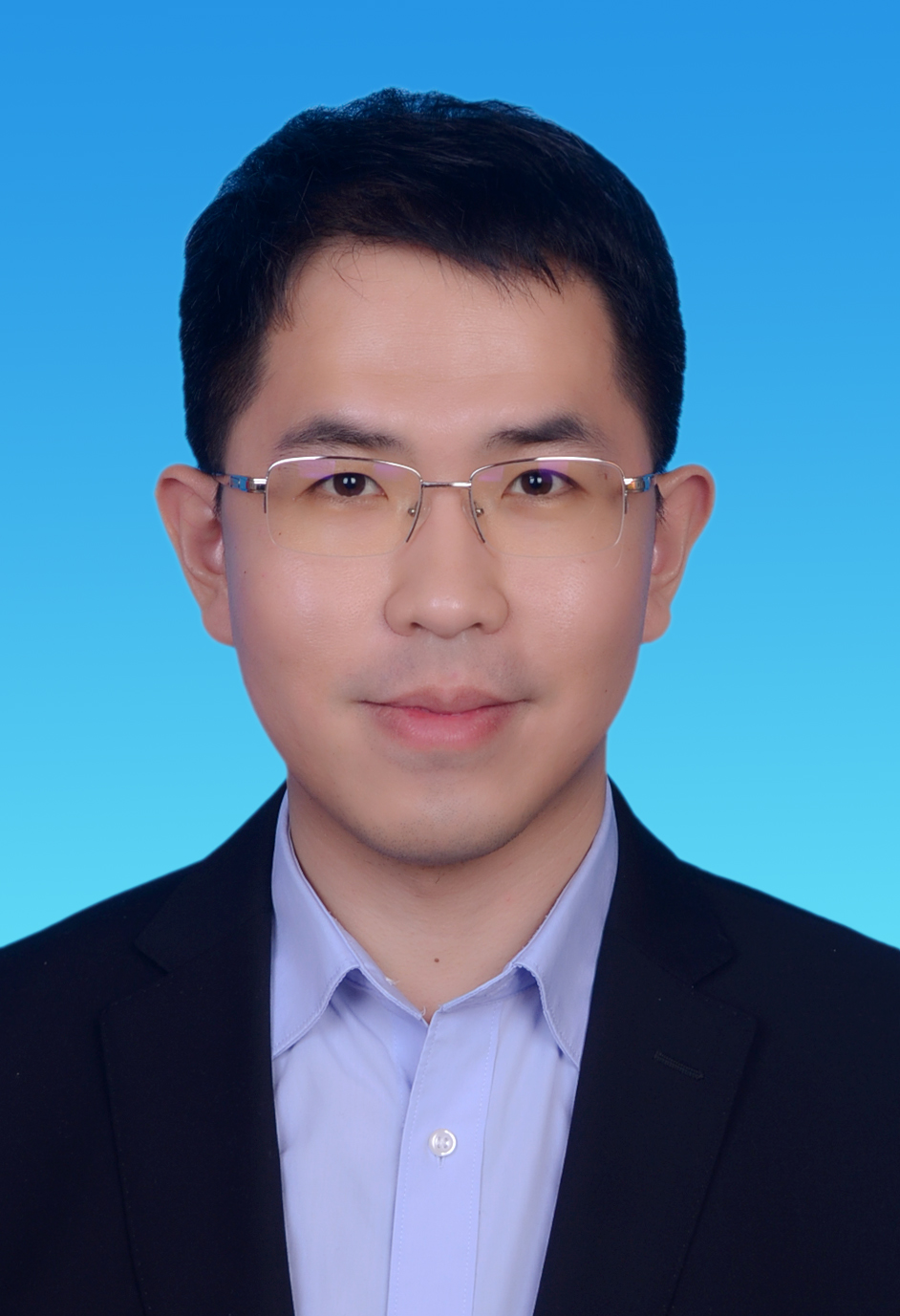}}]{Xin Yang}
	is a Professor in the Department of Computer Science at Dalian University of Technology, China. Yang received his B.S. degree in Computer Science from Jilin University in 2007. From 2007 to June 2012, he was a joint Ph.D. student at Zhejiang University and UC Davis for Graphics and received his Ph.D. degree in July 2012. His research interests include computer graphics and robotic vision.
\end{IEEEbiography}

\begin{IEEEbiography}[{\includegraphics[width=1in,height=1.3in,clip]{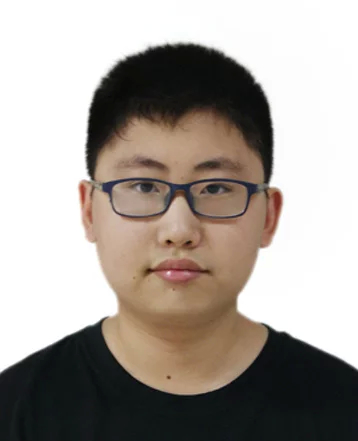}}]{Letian Yu}
	is a Ph.D. student in the Department of Computer Science at Dalian University of Technology. He received his B.S. degree in Computer Science and Technology in 2019. His research interests are computer vision and image processing.
\end{IEEEbiography}

\vfill

\begin{IEEEbiography}[{\includegraphics[width=1in,height=1.2in,clip]{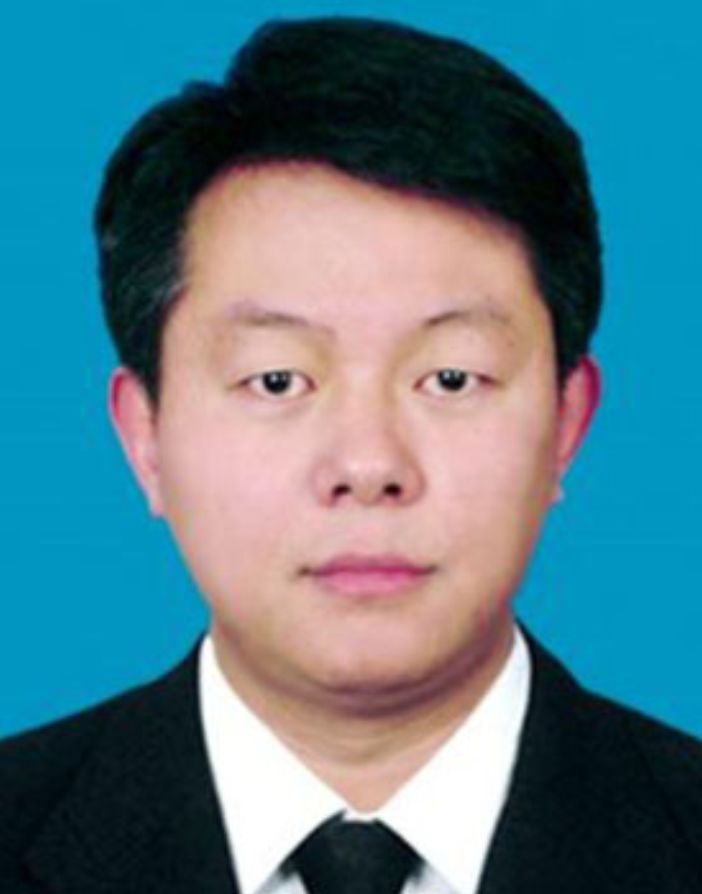}}]{Qiang Zhang}
	was born in Xian, China, in 1971. He received his M.Eng. degree in economic engineering and Ph.D. degree in circuits and systems from Xidian University, Xian, China, in 1999 and 2002, respectively. He was a lecturer at the Center of Advanced Design Technology, Dalian University, Dalian, China, in 2003 and was a professor in 2005. His research interests are bio-inspired computing and its applications. He has authored more than 70 papers in the above fields. Thus far, he has served on the editorial board of seven international journals and has edited special issues in journals such as \emph{Neurocomputing} and \emph{International Journal of Computer Applications in Technology}.
\end{IEEEbiography}

\begin{IEEEbiography}[{\includegraphics[width=1in,height=1.2in,clip]{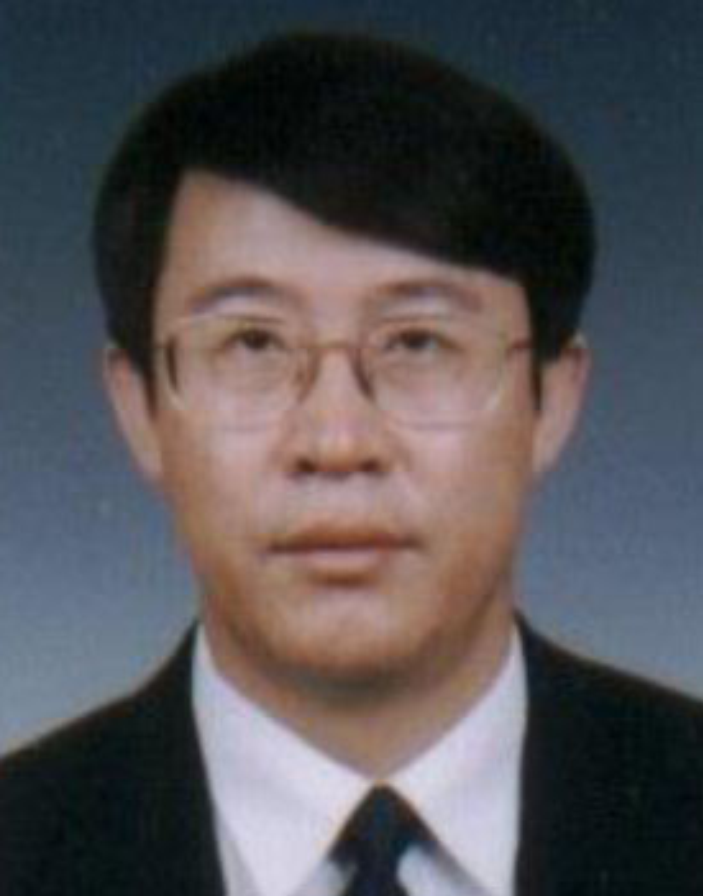}}]{Xiaopeng Wei}
	was born in Dalian, China, in 1959. He received his Ph.D. degree from Dalian University of Technology in 1993. He is a professor at Dalian University of Technology. His research areas include computer animation, computer vision, robots, and intelligent CAD. So far, he has (co-)authored approximately 200 published papers.
\end{IEEEbiography}

\begin{IEEEbiography}[{\includegraphics[width=1in,height=1.3in,clip]{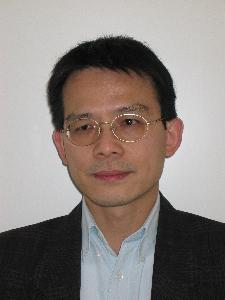}}]{Rynson W.H. Lau} received the Ph.D. degree from the University of Cambridge. He was on the Faculty of Durham University. He is currently with the City University of Hong Kong. 

He serves on the Editorial Board of International Journal of Computer Vision (IJCV). He served as the Guest Editor of a number of journal special issues, including ACM Trans. on Internet Technology, the IEEE Trans. on Multimedia, IEEE Trans. on Visualization and Computer Graphics, and IEEE Computer Graphics \& Applications. He also served in the committee of a number of conferences, including Program Co-chair of ACM VRST 2004, ACM MTDL 2009, IEEE U-MEDIA 2010, and Conference Co-Chair of CASA 2005, ACM VRST 2005, ACM MDI 2009, ACM VRST 2014. His research interests include computer graphics and computer vision.
\end{IEEEbiography}

\vfill

\end{document}